%% file: main.tex
\pdfoutput=1
\documentclass[10pt,twocolumn,letterpaper]{article}
\usepackage[pagenumbers]{cvpr}
\usepackage{amsthm}
\usepackage{multirow}
\usepackage{listings}
\usepackage{longtable}
\usepackage{caption}%
\definecolor{linkblue}{rgb}{0.21,0.49,0.74}
\usepackage[pagebackref,breaklinks,colorlinks,allcolors=linkblue]{hyperref}
\hypersetup{pdftitle={Seeing vs. Counting: Complexity Twins Show Where Multimodal LLMs Fail on Diagrams},pdfauthor={Yin Liu},pdfsubject={},pdfkeywords={}}
\setlist{nosep,leftmargin=1.2em}
\newtheorem{proposition}{Proposition}

\newtheorem{corollary}{Corollary}
\theoremstyle{definition}
\newtheorem{definition}{Definition}
\theoremstyle{remark}

\newcommand{\TC}{\mathsf{TC}^0}
\newcommand{\NC}[1]{\mathsf{NC}^{#1}}
\newcommand{\Lsp}{\mathsf{L}}
\newcommand{\FL}{\mathsf{FL}}
\newcommand{\GapL}{\mathsf{GapL}}
\newcommand{\FP}{\mathsf{FP}}
\newcommand{\shP}{\#\mathsf{P}}
\newcommand{\PH}{\mathsf{PH}}
\newcommand{\bench}{\textup{\textsc{VisCount-Complexity}}}
\newcommand{\car}{\textup{\textsc{CAR}}}
\newcommand{\sST}{\#\mathrm{ST}}
\newcommand{\sPM}{\#\mathrm{PM}}
\newcommand{\sAM}{\#\mathrm{M}}
\newcommand{\sPATH}{\#\mathrm{Paths}}
\newcommand{\sECd}{\#\mathrm{EC}_{\to}}
\newcommand{\sEC}{\#\mathrm{EC}}
\newcommand{\sCC}{\#\mathrm{CC}}

\newcommand{\sDOM}{\#\mathrm{Dom}}
\newcommand{\sSHORT}{\#\mathrm{ShortestPaths}}
\renewcommand{\topfraction}{0.95}
\renewcommand{\bottomfraction}{0.9}
\renewcommand{\textfraction}{0.05}
\renewcommand{\floatpagefraction}{0.85}
\renewcommand{\dbltopfraction}{0.95}
\renewcommand{\dblfloatpagefraction}{0.85}
\input{gen_v8p/gen_v5r/numbers.tex}

\input{gen_v3/numbers_v3.tex}

\input{gen_v4/numbers_v4.tex}
\input{gen_v4/numbers_v41.tex}

\input{gen_v5/numbers_v5.tex}
\input{gen_v5/numbers_v5_503.tex}
\input{gen_v5/numbers_v5_503_slope.tex}

\input{gen_v6/numbers_v6_b7.tex}
\ifdefined\vSixBSevenTVExact\ifnum\vSixBSevenTVExact=\vSixBSevenRotExact\relax\else
  \errmessage{numbers_v6_b7: TV and R180 exact counts differ; rewrite the "each exact" sentence in Sec. 6.6}\fi\fi
\input{gen_v6/numbers_v6_x8.tex}
\input{gen_v6/numbers_v6_t6.tex}
\input{gen_v6/numbers_v6_f6.tex}
\input{gen_v5/numbers_v5_c503.tex}
\input{gen_v8/numbers_g9.tex}

\newif\ifincludeQW\includeQWfalse
\input{gen_v8/qw_ready.tex}
\ifincludeQW
\input{gen_v8/numbers_qw.tex}
\fi
\input{gen_v8/numbers_trunc.tex}
\input{gen_v8/numbers_r9.tex}

\input{gen_v8/numbers_review.tex}

\input{gen_v5/numbers_v5_b6.tex}
\input{gen_v6/numbers_v6_d8.tex}

\newcommand{\checkfinalnumbers}[2]{\ifdefined#1\edef\tmpPartialFlag{#1}\ifx\tmpPartialFlag\empty\else
  \ifdefined\allowpartialnumbers\else\errmessage{#2 comes from an unfinished or mock run}\fi\fi\fi}
\checkfinalnumbers{\vSixXEightPartial}{gen_v6/numbers_v6_x8.tex}
\checkfinalnumbers{\vSixDEightPartial}{gen_v6/numbers_v6_d8.tex}
\providecommand{\vSixDEightImages}{--}
\providecommand{\vSixDEightPairs}{--}
\providecommand{\vSixDEightClusters}{--}
\checkfinalnumbers{\vSixTSixPartial}{gen_v6/numbers_v6_t6.tex}
\checkfinalnumbers{\vSixFPartial}{gen_v6/numbers_v6_f6.tex}
\checkfinalnumbers{\vFiveCPartial}{gen_v5/numbers_v5_c503.tex}
\checkfinalnumbers{\gNinePartial}{gen_v8/numbers_g9.tex}
\checkfinalnumbers{\rNinePartial}{gen_v8/numbers_r9.tex}

\newsavebox{\genfitbox}
\newlength{\genfitht}
\newcommand{\genfitsetinput}[1]{\sbox{\genfitbox}{#1}%
  \ifdim\wd\genfitbox>\linewidth\sbox{\genfitbox}{\setlength{\tabcolsep}{3pt}#1}\fi
  \ifdim\wd\genfitbox>\linewidth\sbox{\genfitbox}{\resizebox{\linewidth}{!}{\usebox{\genfitbox}}}\fi
  \setlength{\genfitht}{\ht\genfitbox}\addtolength{\genfitht}{\dp\genfitbox}%
  \ifdim\genfitht>0.8\textheight\typeout{GENFIT height-scaled}\resizebox*{!}{0.8\textheight}{\usebox{\genfitbox}}%
  \else\usebox{\genfitbox}\fi}
\newenvironment{keepblock}{\par\medskip\noindent\begin{minipage}{\linewidth}}{\end{minipage}\par\medskip}
\providecommand{\nTotalCalls}{--}
\providecommand{\nModels}{--}
\providecommand{\nConfigs}{--}

\providecommand{\nBenchPairs}{--}
\providecommand{\nXLPairs}{--}
\providecommand{\PerceptSummary}{[pending]}
\providecommand{\AttribSummary}{[pending]}
\providecommand{\TwinSummary}{[pending]}
\providecommand{\SizeSummary}{[pending]}
\providecommand{\BudgetSummary}{[pending]}
\providecommand{\TokenSummary}{[pending]}
\providecommand{\HintSummary}{[pending]}
\providecommand{\CodeSummary}{[pending]}
\providecommand{\CarSummary}{[pending]}
\providecommand{\CarClassAcc}{--}

\providecommand{\CarAblSummary}{[pending]}
\providecommand{\GenSummary}{[pending]}
\providecommand{\ClassifierNote}{}
\providecommand{\HeldoutSummary}{[pending]}
\providecommand{\CodeAttribSummary}{}
\providecommand{\MagnitudeSummary}{[pending]}
\providecommand{\BudgetDecadesSummary}{[pending]}
\providecommand{\ZoomSummary}{[pending]}

\providecommand{\OrientSummary}{[pending]}
\providecommand{\VerifyNumbers}{--}
\providecommand{\AlgoScaleSummary}{}
\providecommand{\TwoViewOthers}{}
\providecommand{\PerceptPooledN}{--}
\providecommand{\PerceptPooledF}{--}
\providecommand{\PerceptPooledExact}{--}

\providecommand{\AttribPooledPct}{--}

\title{Seeing vs.\ Counting: Complexity Twins Show Where\\Multimodal LLMs Fail on Diagrams}
\author{Yin Liu\\
Google\\
{\tt\small hughliu@google.com}
}

\begin{document}
\maketitle

\begin{abstract}
When a multimodal LLM miscounts in a diagram, does it fail at seeing or at counting? Existing visual graph and puzzle benchmarks vary visual density and algorithmic hardness together, confounding the two. We pair counting questions about the same drawing that sit in different counting-complexity classes, such as its spanning trees (a determinant) and its simple paths ($\shP$-complete), and use a transcription probe and a text-only control to separate reading from computing. We test \ifincludeQW{}Gemini, Gemma and Qwen3-VL\else{}Gemini and Gemma\fi{} models; for the Gemini models, the answer depends on how the model computes. In chain of thought, most of them read small drawings correctly and then fail at counting, at a rate set far more by the size of the answer than by its complexity class (for the primary model, the odds of an exact answer fall \vFourMagPooledNoxlVisualORinv-fold per decade of the answer). Given code and an exact transcription, the primary model counts correctly even on $10{\times}10$ grids, and reading becomes the bottleneck: a single view misses horizontal edges, and a second reading of a rotated copy with crop checks reads \vSixBSevenTVExact/\vSixBSevenTVNGrids{} large grids exactly, against \vSixBSevenBthreeExact/\vSixBSevenTVNGrids{} when the upright reading is simply repeated.
\end{abstract}

\section{Introduction}
\label{sec:intro}
Counting is the simplest quantitative question a vision system can be asked, and multimodal large language models (MLLMs) still get it wrong surprisingly often~\cite{paiss2023countbench,rahmanzadehgervi2024blind,vo2026biased,tamarapalli2025countqa}. The usual diagnosis is perceptual: clutter, occlusion, an encoder that cannot bind features~\cite{tong2024eyes,campbell2024binding}. But many counting questions asked of diagrams and graph drawings stay hard after the structure has been read off the pixels: they are \emph{computational}, and complexity theory makes a testable prediction about how hard they are. Existing visual graph and puzzle benchmarks~\cite{li2024visiongraph,wei2024gita,babaiee2025vga,fan2024nphardeval4v,ghosal2025algopuzzlevqa} vary visual density and algorithmic hardness together, confounding the two sources of error.

\begin{figure*}[!t]
\centering
\includegraphics[width=\linewidth]{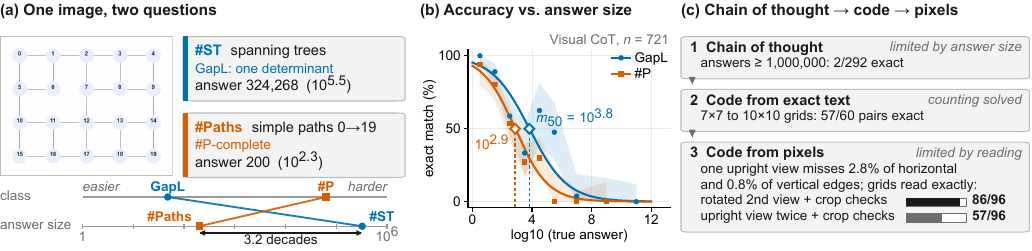}
\caption{\textbf{One image, two questions.} (a)~On this $4{\times}5$ drawing $\sST$ ($\GapL$, one determinant) is $324{,}268$ and $\sPATH$ from vertex~0 to vertex~19 ($\shP$-complete) is $200$: the harder question has the smaller answer. (b)~For \texttt{gemini-3.8-flash} (low; also in (c)), chain-of-thought exact match follows answer size far more than class (Visual, $n=\vFourMagPooledNoxlVisualN$; dashed: $m_{50}$, where accuracy crosses 50\%). (c)~Code from an exact edge list solves almost all extra-large pairs; from pixels, on \vSixBSevenTVNGrids{} fresh grids one upright view misses \vSixBSevenORUpzeroMissHRatePct\% of horizontal and \vSixBSevenORUpzeroMissVRatePct\% of vertical edges, and a second reading of a rotated copy with crop checks reads \vSixBSevenTVExact{} of them exactly, against \vSixBSevenBthreeExact{} for a repeated upright reading.}
\label{fig:teaser}
\end{figure*}

\begin{figure*}[!t]
\centering
\setlength{\tabcolsep}{3pt}
\resizebox{\linewidth}{!}{%
\begin{tabular}{ccccc}
\includegraphics[height=1.9cm]{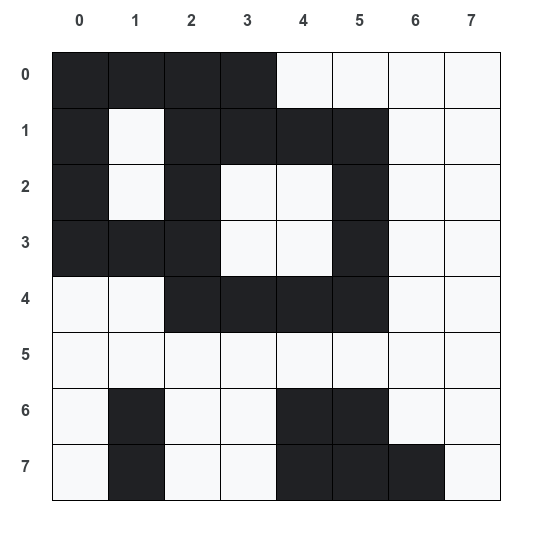} &
\includegraphics[height=1.9cm]{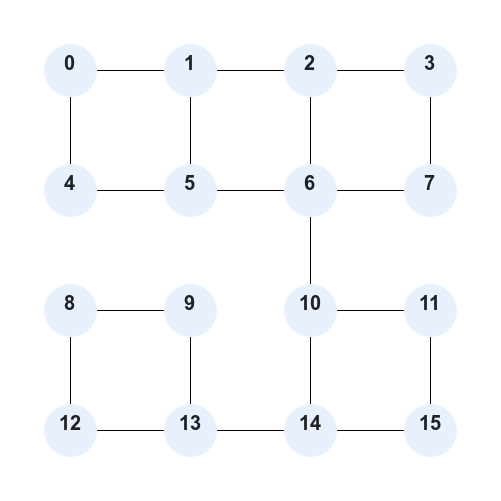} &
\includegraphics[height=1.9cm]{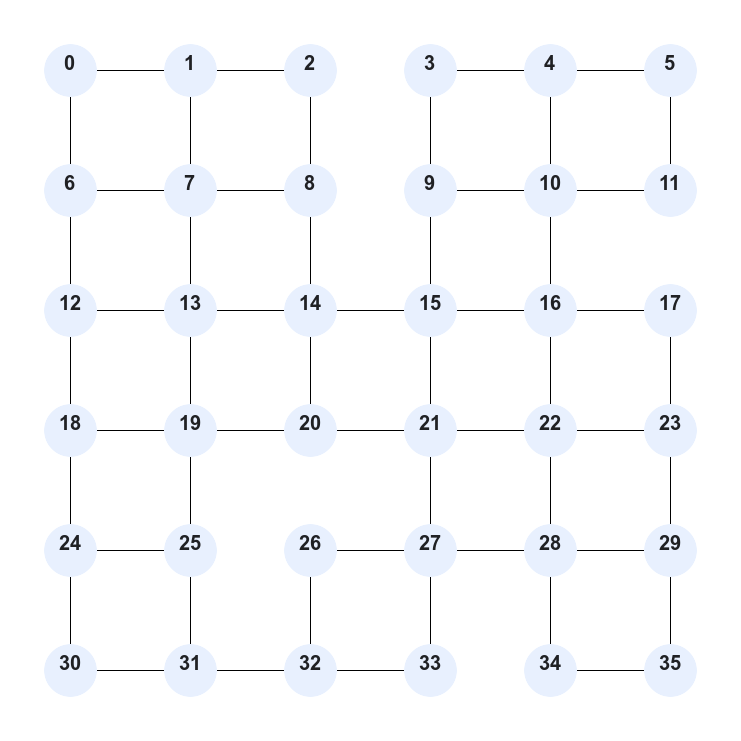} &
\includegraphics[height=1.9cm]{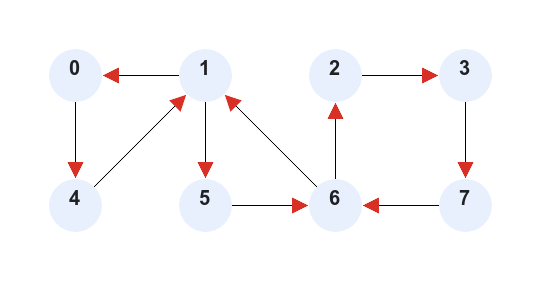} &
\includegraphics[height=1.9cm]{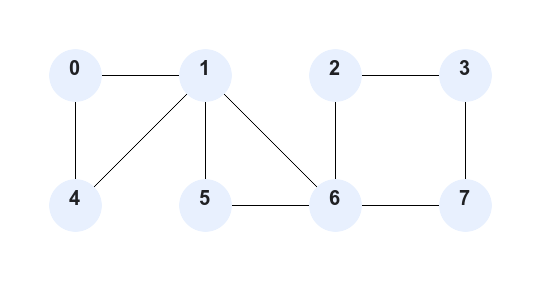} \\
{\footnotesize (a) binary image} & {\footnotesize (b) grid $4{\times}4$} & {\footnotesize (c) grid $6{\times}6$} & {\footnotesize (d) Eulerian, directed} & {\footnotesize (e) same drawing, undirected} \\
{\scriptsize $\chi$ ($\TC$) vs.\ $\sCC$ ($\Lsp$)} &
{\scriptsize $\sST,\sPM$ ($\GapL$) vs.\ $\sAM,\sPATH$ ($\shP$)} &
{\scriptsize same questions, $n=36$} &
{\scriptsize $\sECd$ ($\GapL$, BEST)} &
{\scriptsize $\sEC$ ($\shP$-complete)}
\end{tabular}}
\vspace{-4pt}
\caption{\textbf{Complexity twins.} Each of (a)--(c) supports two counting questions in different complexity classes on the same pixels; (d) and (e) show one drawing with and without arrowheads. On grids of bounded width both members are polynomial-time (Proposition~\ref{prop:car}), so the classes separate only as the width grows.}
\label{fig:twins}
\vspace{-6pt}
\end{figure*}

This paper separates them with a classical fact that, to our knowledge, has not been used to evaluate MLLMs: superficially similar counting questions about the \emph{same} structure sit on different rungs of the counting-complexity ladder. On a binary image the Euler characteristic is a weighted sum of $2{\times}2$ pattern counts~\cite{gray1971local,minsky1969perceptrons}, whereas the number of components needs global connectivity~\cite{reingold2008undirected,allender2009planar}. On a planar drawing the numbers of spanning trees and perfect matchings are determinants~\cite{kirchhoff1847ueber,kasteleyn1961statistics,temperley1961dimer}, whereas the numbers of all matchings and of simple $s$--$t$ paths are $\shP$-complete even for planar bounded-degree graphs~\cite{jerrum1987two,vadhan2001complexity,provan1986complexity}. On a balanced digraph Eulerian circuits are counted by one determinant (BEST~\cite{vanaardenne1951circuits,smith1941unicursal}); remove the arrowheads and the count becomes $\shP$-complete~\cite{brightwell2005counting,ge2012complexity}. We call such pairs \emph{complexity twins} (one in Fig.~\ref{fig:teaser}(a), all three families in Fig.~\ref{fig:twins}), the converse of CLIP-blind pairs~\cite{tong2024eyes}: a twin keeps the drawing and changes the question. The pixels are identical in the grid and binary-image families and differ only by arrowheads in the Eulerian family, but the reading can still depend on the question (the transcriptions embedded in task-specific routing-prompt programs are identical across tasks in \vFourDecPooledNoxlCarImagesAllTranscriptionsIdentical{} of \vFourDecPooledNoxlCarImagesWithGeTwoScorable{} images; Appendix~\ref{app:v4}), so we score the transcription of every call. A twin usually changes the size of the answer as well, and much of our analysis separates class from answer size.

Complexity classes are asymptotic, but the ladder is also a ladder of \emph{algorithms}: a determinant costs $O(n^3)$ steps; exact dynamic programs for the $\shP$-complete grid twins keep a number of states that is exponential only in the width of the drawing, so they are subexponential on lattice drawings and polynomial at bounded width (Proposition~\ref{prop:car}); and enumeration costs at least the answer itself, which on our grid drawings grows exponentially in $n$. Expressivity results for transformers~\cite{merrill2023parallelism,merrill2024expressive,li2024chain,feng2023towards,sanford2024transformers} tie the ladder to the length of the chain of thought.

Whether a model's errors follow the ladder is an empirical question. Twins let us ask it with the drawing held fixed and, through magnitude-matched pairs, with the answer size held fixed too; the most telling pair is the spanning-tree count $\sST$ against the simple-path count $\sPATH$, where class and answer size point in opposite directions (Fig.~\ref{fig:teaser}(a)). Our answer to the title question has two regimes:

\noindent\emph{In chain of thought, most Gemini API models read small diagrams correctly and then fail at counting, at a rate set mainly by the size of the answer rather than by its complexity class; given code and an exact transcription, the Gemini models we tested with code count correctly, and reading large diagrams becomes the failure.}

\paragraph{Contributions.}
\begin{enumerate}
\item \textbf{Complexity twins with an attribution protocol.} \bench{} asks two counting questions of one image on different rungs of the ladder in Eq.~\eqref{eq:ladder}, in three families with exact ground truth that an independent re-implementation reproduces on all \GTCheckN{} core, seed and extra-large instances. A transcription probe, a text-only control, and re-solving of the transcription embedded in generated code attribute errors to reading or to computing (\S\ref{sec:lens}--\S\ref{sec:res_percept}).
\item \textbf{In chain of thought, answer size, far more than class, sets success.} Exact answers become rarer with every decade of the answer, whereas at equal answer size the class shifts $m_{50}$, the answer size at which accuracy crosses 50\%, by only \vFourMagPooledNoxlVisualMfiftyDiff{} decade, concentrated in one twin pair; tenfold more thinking tokens move $m_{50}$ by about three decades. Pre-registered magnitude-matched twins show at most a small class advantage. Varying class without controlling for magnitude mostly measures answer size (\S\ref{sec:res_twins}--\S\ref{sec:res_hint}).
\item \textbf{With code, reading becomes the bottleneck.} Programs written from the exact text solve almost all extra-large pairs. From pixels, single views miss horizontal edges far more often than vertical ones, and the misses are systematic for a view: a second, decorrelated reading of a rotated copy with crop checks reads \vSixBSevenTVExact/\vSixBSevenTVNGrids{} fresh grids exactly, whereas repeating the identical upright reading reaches \vSixBSevenBthreeExact/\vSixBSevenTVNGrids{} (\S\ref{sec:res_code}, \S\ref{sec:car_results}).
\end{enumerate}

\section{Related Work}
\label{sec:related}
\paragraph{Visual counting and perceptual failures.} CountBench~\cite{paiss2023countbench}, PixMo-Count~\cite{deitke2025molmo} and CountQA~\cite{tamarapalli2025countqa} measure object counting in natural images, where the computation is a sum and the errors are perceptual; BlindTest~\cite{rahmanzadehgervi2024blind} and~\cite{vo2026biased} show that simple synthetic counts fail too, and BlindTest's tasks of counting the rows and columns of a grid and the single-colored paths between two subway stations are close to our grid reading and to $\sPATH$; MMVP~\cite{tong2024eyes} isolates encoder blind spots with image pairs CLIP cannot tell apart. We invert that design: our pairs keep the drawing and change the computation.

\paragraph{Graph and algorithmic benchmarks.} VisionGraph~\cite{li2024visiongraph}, GITA~\cite{wei2024gita}, VGCure~\cite{zhu2025vgcure}, Visual Graph Arena~\cite{babaiee2025vga} and GraphVerse~\cite{sun2026graphverse} test MLLMs on node-link drawings with decision and optimisation questions, and GraphVLM~\cite{liu2026graphvlm} on multimodal graph learning. VisionGraph reports that all tested models perceive graph structure poorly, which hurts problem solving; on our labelled lattice drawings, by contrast, the primary model reads most benchmark-size drawings exactly (\S\ref{sec:res_percept}). NPHardEval4V~\cite{fan2024nphardeval4v} grades NP-hard optimisation problems by class; AlgoPuzzleVQA and PuzzleVQA~\cite{ghosal2025algopuzzlevqa,chia2024puzzlevqa} inject ground-truth perception to attribute errors; IsoBench~\cite{fu2024isobench} and SpatialEval~\cite{wang2024spatialeval} compare image and text renderings. On the text side NLGraph~\cite{wang2023nlgraph}, GraphQA~\cite{fatemi2024talk} and GraphArena~\cite{tang2025grapharena} test graph problems, NPHardEval~\cite{fan2024nphardeval} grades by P/NP class and NPPC~\cite{yang2026nppc} scales NP-complete problems with progressive constraints. None uses counting classes, determinant-vs-permanent gaps or same-image twins, or measures the thinking budget as a function of the class.

\paragraph{Seeing versus solving.} Concurrently, Wang et al.~\cite{wang2026seeing} test, question by question, whether a vision-language model can extract an exactly scorable task state, solve the question from the ground-truth state, and answer the original question; they find composition failures, where extraction and solving both succeed but the direct answer fails, and reduce them by tuning the model to write out the state before its answer. Our Visual prompt already asks for the transcription before the answer; what differs here is the twin design, which varies the computation by counting class on one drawing and lets us ask how errors scale with answer size. In our attribution (Table~\ref{tab:attribution}), 115 of the primary model's 144 errors come with an exact transcription, and the other 29 also fail under Text.

\paragraph{Expressivity and test-time compute.} Log-precision fixed-depth transformers without chain of thought lie in uniform $\TC$~\cite{merrill2023parallelism,strobl2024formal}; $O(\log n)$ chain-of-thought steps stay within $\Lsp$ and polynomially many reach $\mathsf{P}$~\cite{merrill2024expressive,li2024chain,feng2023towards}; logarithmic depth suffices for connectivity~\cite{sanford2024transformers,merrill2025depth}. Lyu et al.~\cite{lyu2026intrinsic} test a $\TC$-vs-$\NC{1}$ split on ViT embeddings; we know of no empirical test of these predictions on counting problems or multimodal inputs. Faith and Fate~\cite{dziri2023faith} cast multiplication, logic grid puzzles and a dynamic program as computation graphs and find that transformer accuracy falls quickly as their complexity grows; our magnitude law is a multimodal, quantitative counterpart that measures the size of the answer. PAL and Program-of-Thoughts~\cite{gao2023pal,chen2023pot} offload arithmetic to code and test-time scaling~\cite{snell2024scaling} buys accuracy with tokens; we measure both along the ladder.

\section{Complexity Twins}
\label{sec:lens}
\subsection{The ladder}
We write $\TC$ for functions computed by uniform constant-depth threshold circuits, $\FL$ for logarithmic space, $\GapL$ for differences of two logspace-countable functions~\cite{mahajan1997determinant}, $\FP$ for polynomial time and $\shP$ for counting functions of polynomial-time nondeterministic machines~\cite{valiant1979permanent}. For $\mathbb N$-valued functions,
\begin{equation}
\TC\ \subseteq\ \FL\ \subseteq\ \GapL\ \subseteq\ \NC{2}\cap\FP\ \subseteq\ \shP,
\label{eq:ladder}
\end{equation}
every inclusion conjectured strict and none proved, and all rungs are populated by natural counting problems on images and drawings (Table~\ref{tab:ladder}). The integer determinant is complete for $\GapL$~\cite{mahajan1997determinant}, so every determinant-reducible count has a polylog-depth algorithm~\cite{berkowitz1984computing}.

\begin{table}[!t]
\centering\footnotesize
\setlength{\tabcolsep}{2.5pt}
\caption{The eight tasks of \bench{} and their classes. ``$\shP$-c'': $\shP$-complete on a graph class that contains our instances (Definition~\ref{def:twin}); grid graphs are subgraphs of the square lattice (planar, bipartite, maximum degree~4). At bounded grid width the $\sAM$ and $\sPATH$ rows are polynomial-time (Proposition~\ref{prop:car}), so these classes separate only as the width grows.}
\label{tab:ladder}
\resizebox{\linewidth}{!}{%
\begin{tabular}{llll}
\toprule
Task & Input & Class & Algorithm / hardness \\
\midrule
$\chi$ & binary image & $\TC$ & bit-quad formula~\cite{gray1971local}; $\notin\mathsf{AC}^0$ \\
$\sCC$ & binary image & $\Lsp$ & Reingold~\cite{reingold2008undirected}; $\NC{1}$-hard~\cite{allender2009planar} \\
$\sST$ & grid graph & $\GapL$ & matrix-tree det.~\cite{kirchhoff1847ueber} \\
$\sPM$ & grid graph & $\GapL$ & Kasteleyn/FKT~\cite{kasteleyn1961statistics,temperley1961dimer}; planar bipartite $\equiv$ det~\cite{datta2010planarity} \\
$\sECd$ & balanced digraph & $\GapL$ & BEST~\cite{vanaardenne1951circuits,smith1941unicursal} \\
$\sAM$ & grid graph & $\shP$-c & planar bip.\ bounded $\Delta$~\cite{jerrum1987two,vadhan2001complexity} \\
$\sPATH$ & grid graph & $\shP$-c & \cite{valiant1979enumeration}; planar~\cite{provan1986complexity}; grids~\cite{liskiewicz2003complexity} \\
$\sEC$ & Eulerian graph & $\shP$-c & \cite{brightwell2005counting}; planar 4-regular~\cite{ge2012complexity} \\
\bottomrule
\end{tabular}}
\end{table}

\subsection{Twins and the three families}
\label{sec:twins}
\begin{definition}[twin]
\label{def:twin}
Let $\mathcal I$ be a family of images with an injective combinatorial reading $G(I)$. A \emph{complexity twin} is a pair of counting functions $f_{\mathrm{lo}}, f_{\mathrm{hi}}:\mathcal I\to\mathbb N$ and classes $\mathcal C_{\mathrm{lo}}\subseteq\mathcal C_{\mathrm{hi}}$ (inclusion conjectured strict), taken from~\eqref{eq:ladder} or $\NC{1}$ (which lies between $\TC$ and $\FL$), such that $f_{\mathrm{lo}}\in\mathcal C_{\mathrm{lo}}$, $f_{\mathrm{hi}}$ is hard for $\mathcal C_{\mathrm{hi}}$ on a graph class that contains $G(\mathcal I)$, and $f_{\mathrm{lo}}(I)\neq f_{\mathrm{hi}}(I)$ for all $I$.
\end{definition}
Hardness on a graph class carries over to every class that contains it, so the classes of Table~\ref{tab:ladder} serve: planar bipartite graphs of bounded degree for $\sAM$~\cite{jerrum1987two,vadhan2001complexity}, subgraphs of the grid for $\sPATH$~\cite{liskiewicz2003complexity}, planar graphs of maximum degree~4 for $\sEC$, which contain the planar 4-regular graphs of~\cite{ge2012complexity}, and grid graphs for $\sCC$, which is $\NC{1}$-hard there~\cite{allender2009planar}. Hardness is asymptotic: on grids of bounded width both members are computable in polynomial time (Proposition~\ref{prop:car}) and our Eulerian drawings are small enough to enumerate, so on the drawings we test a twin separates the class of a question, not its cost. Our three families follow; proofs and conventions are in Appendix~\ref{app:proofs}.

\paragraph{Binary images: $\chi$ vs.\ $\sCC$.} For $I\in\{0,1\}^{R\times C}$ padded by white, let $C_4(I)$ be the number of 4-connected black components and $H_8(I)$ the number of 8-connected white holes. The Euler characteristic $\chi=C_4-H_8$ satisfies
\begin{equation}
\chi(I)=N_{\bullet}(I)-N_{\bullet\bullet}(I)+N_{\boxplus}(I),
\label{eq:gray}
\end{equation}
with $N_{\bullet}$ the number of black cells, $N_{\bullet\bullet}$ of adjacent black pairs and $N_{\boxplus}$ of all-black $2{\times}2$ blocks (Proposition~\ref{prop:gray} in Appendix~\ref{app:proofs}); each term is an iterated sum of local predicates, so $\chi\in\TC$ (and $\chi\notin\mathsf{AC}^0$, since it contains parity~\cite{furst1981parity,hastad1986almost}). Counting components is in $\Lsp$~\cite{reingold2008undirected} and $\NC{1}$-hard on grid graphs~\cite{allender2009planar}. We generate images with $H_8\ge1$, so $\chi<C_4$ always.

\paragraph{Grid drawings: determinants vs.\ $\shP$.} For a connected subgraph $G$ of the lattice, $\sST(G)$ is a cofactor of the Laplacian and, $G$ being planar and bipartite, $\sPM(G)=|\det K|$ for a Kasteleyn-signed biadjacency matrix~\cite{kasteleyn1967graph}; on the same $G$ the number of all matchings and the number of simple corner-to-corner paths are $\shP$-complete for planar bipartite bounded-degree graphs~\cite{jerrum1987two,vadhan2001complexity,provan1986complexity,liskiewicz2003complexity}. The median $\log_{10}$ counts of $\sST$ and $\sAM$ differ by at most \TwinMagSmallTwo{} decades up to $4{\times}5$ but drift apart on larger grids (\TwinMagFive{} and \TwinMagSix{} decades at $5{\times}5$ and $6{\times}6$; Appendix~\ref{app:data}), so the regression below controls for magnitude, and Appendix~\ref{app:t6} adds 80 twins matched within $0.25$ decades per graph. Every perfect matching is a matching, so $\sPM\le\sAM$ on every graph.

\paragraph{Eulerian drawings: arrowheads on or off.} Planar edge-disjoint unions of directed cycles with in-degree equal to out-degree in $\{1,2\}$. With arrowheads, BEST gives $\sECd=t_w\prod_v(d^+(v)-1)!=t_w$, one cofactor of the out-degree Laplacian (Proposition~\ref{prop:best} in Appendix~\ref{app:proofs}); without them the count is $\shP$-complete even for planar 4-regular graphs~\cite{ge2012complexity}. We reject pairs whose answers coincide. With the conventions of our prompts $\sECd\le\sEC$: each directed circuit, counted up to rotation, is an undirected circuit, counted up to rotation or reversal, and no directed circuit is the reversal of another (its arcs would run backwards), so the map is injective. In these two twins the $\GapL$ member therefore has the smaller answer, whereas in $\sST$/$\sPATH$ it has the larger one on every grid we test.

\subsection{What the ladder predicts}
\label{sec:predictions}
Without chain of thought a log-precision fixed-depth transformer computes a function in uniform $\TC$~\cite{merrill2023parallelism}; with $t(n)$ chain-of-thought tokens it stays within $\mathsf{TIME}(\mathrm{poly}(n+t(n)))$, $O(\log n)$ tokens keep it inside $\Lsp$, and polynomially many tokens simulate any polynomial-size circuit~\cite{merrill2024expressive,li2024chain}. Hence the determinant questions ($\sST$, $\sPM$, $\sECd$) are reachable with $\mathrm{poly}(n)$ thinking tokens, whereas their $\shP$-complete twins, on the graph classes of Table~\ref{tab:ladder}, are not unless $\shP\subseteq\FP/\mathrm{poly}$ (Proposition~\ref{prop:pred} in Appendix~\ref{app:proofs}, which also covers $\chi$ and $\sCC$ and gives the proofs). This is asymptotic: it says how the \emph{required} budget must scale with $n$ for each rung, not what a finite model does at $n=36$. In practice, the ladder predicts that a chain of thought can keep pace with the determinant questions at a cost polynomial in $n$ but, under standard complexity assumptions, not with their $\shP$-complete twins, so as drawings grow the $\shP$ member of a twin should go out of reach first, whatever the sizes of the two answers. For programs, the ladder fixes the cost of a correct route, that is, of choosing the right algorithm:
\begin{proposition}[cost of a correct route]
\label{prop:car}
On a subgraph of the $R\times C$ lattice with $n=RC$ vertices and $w=\min(R,C)$, a correctly routed program computes $\sST$ and $\sPM$ in $O(n^3)$ arithmetic operations (as it does $\sECd$ on a balanced digraph with $n$ vertices), and $\sAM$ and $\sPM$ in $O(n\,2^{w})$ and $\sPATH$ in $n\,2^{O(w)}$ by frontier dynamic programming, which is polynomial in $n$ for bounded $w$ and $2^{O(\sqrt n)}$ in general. Enumeration of matchings or simple paths visits every object it counts, so its cost is at least the answer, $\Omega(\mathrm{answer})$, which for a lattice subgraph with a perfect matching is at least $2^{n/2}$ matchings.
\end{proposition}
For $\sPATH$ the frontier states record which frontier vertices are joined by path fragments; by planarity these pairings do not cross, which keeps their number at $2^{O(w)}$, as in frontier-based counting with Knuth's SIMPATH~\cite{knuth2011taocp} (proof sketch in Appendix~\ref{app:proofs}). Our Eulerian drawings have at most 26 edges, so their circuits can simply be enumerated. Hence each grid twin is polynomial-time at bounded width, which covers our grids (width $w\le6$ on the benchmark and $w\le10$ on the extra-large grids), and the classes separate only as the width grows. The proposition also says where routing can matter: on the benchmark proper ($n\le36$) 265 of the 312 core grid and Eulerian answers are below $10^7$, and the larger ones are spanning-tree counts, which the programs compute as determinants at every size, and matching counts on $5{\times}6$ and $6{\times}6$ grids, where the programs already use dynamic programming (\S\ref{sec:res_code}), so every code condition should tie; on $10{\times}10$ grids the answers reach $10^{37}$, beyond any enumeration, while a correctly routed program stays polynomial.

\section{Benchmark and Protocol}
\label{sec:bench}
\label{sec:protocol}
\paragraph{Instances.} All instances come from a seeded generator, rendered from SVG at 96\,dpi. \emph{Grid drawings}: for $R\times C$ from $3{\times}3$ to $6{\times}6$ (seven sizes) we keep each lattice edge with probability $p\sim U[0.8,0.9]$, require connectivity, minimum degree~2 and at least two perfect matchings when $n$ is even; vertices are labelled $0..n{-}1$ and drawn as circles joined by segments (10 per size core, 25 pooled, 6 at $3{\times}3$). Four extra-large sizes, $7{\times}7$ to $10{\times}10$ (6 each, answers up to $10^{37}$), extend the scaling study with $\sST$, $\sPM$, $\sAM$. \emph{Binary images}: $6{\times}6$ to $12{\times}12$ grids with printed indices, built from one to three hollow squares and a few small blobs, with $H_8\ge1$, $2\le C_4\le 9$ and $\chi\ne C_4$ (black density \BinDensMin--\BinDensMax; 10 per size core, 25 pooled). \emph{Eulerian drawings}: planar unions of 3- and 4-cycles in four size buckets (8 pairs each core, 16 pooled), with and without arrowheads. Ground truth uses exact arithmetic; the generator checks $\sAM$ and $\sPM$ ($n\le 16$), undirected Eulerian circuits (at most 18 edges) and $\chi$ with a second solver, and an independent re-implementation of the eight task solvers reproduces every answer of the \GTCheckN{} core, seed and extra-large instances (Appendix~\ref{app:data}). The core slice has \nBenchPairs{} (instance, task) pairs, the pooled expansion has \vThreePooledPairs{} pairs across 320 instances (\vThreeSeedRecords{} extra calls), a five-style rendering suite re-renders all core grids and binary images (\vThreeStylePairs{} pairs, \vThreeStyleRecords{} calls; Appendix~\ref{app:v3_ext}), and the extra-large suite adds \nXLPairs{} pairs.

\paragraph{Conditions.} Each pair is evaluated under up to ten prompts (verbatim in Appendix~\ref{app:prompts}). \textbf{Visual}: image and task; the model must transcribe the structure first and end with a JSON object holding the transcription and the integer answer. \textbf{Text}: the ground-truth edge list or 0/1 matrix replaces the image. \textbf{Hint}: Visual plus a one-sentence statement of the relevant theorem or algorithm (Kirchhoff, Kasteleyn, BEST, the bit-quad formula, flood fill, or the exponential recurrence for the $\shP$ tasks). \textbf{Code}: the model returns one Python program that hard-codes what it sees and prints the answer, run with a 60\,s limit; \textbf{Text$\to$Code} and \textbf{Hint$\to$Code} are the text and hint analogues.

\paragraph{Code and multi-view reading.}\label{sec:car} A routing prompt, \car{} (complexity-aware routing), asks the model to transcribe the structure, state whether the counting problem has a known polynomial-time exact algorithm and name it, and then write the program (Text\,\car{} runs it on the text input); we use it only as a control, with four ablations such as Perceive+Code, which drops the classification step (prompts and ablations in Appendix~\ref{app:codecond}). To analyse reading at scale (\S\ref{sec:car_results}), \textbf{two-view} reading transcribes the image twice, upright and as a rotated copy ($90^\circ$, with $180^\circ$ as a second arm), accepts or rejects the vertex pairs on which the two readings agree, settles each disagreement by one further call showing a local crop of the two endpoint vertices alone, and solves the verified list by code; repeating the upright reading instead is the baseline. No ground truth enters; a crop-based alternative, Zoom-$k$, is described in Appendix~\ref{app:codecond}.

\begin{table}[!t]\centering\footnotesize
\setlength{\tabcolsep}{3pt}
\caption{Error attribution on the Visual condition for five configurations (all in Table~\ref{tab:attribution_all}). P: transcription exact; C: answer exact. Columns are percentages of instances in each class.}
\label{tab:attribution}
\input{gen_v8/tab_attribution_main_v82.tex}
\end{table}

\begin{table*}[!t]\centering\footnotesize
\caption{Paired same-image twin tests (Visual, core slice) for five configurations. Each cell gives acc$_A$/acc$_B$, exact match (\%) on the lower-class/higher-class member, and $b/c$, the pairs solved only by A / only by B; $^\dagger$: exact McNemar test significant after Holm adjustment over these \vFourTwinsMainTableVisualNTests{} tests; n/a: fewer than 10 paired instances. Intervals and raw and adjusted $p$ for these tests are in Table~\ref{tab:v4_twins_main}; all configurations and conditions in Table~\ref{tab:twins_all} and Table~\ref{tab:v4_twins_pooled}. Class and answer size are aligned in $\sPM$/$\sAM$ and $\sECd$/$\sEC$ (the $\GapL$ member has the smaller answer), opposed in $\sST$/$\sPATH$ and size-dependent in $\sST$/$\sAM$ (\S\ref{sec:twins}).}
\label{tab:twins}
\input{gen_v8/tab_twins_matrix.tex}
\end{table*}

\paragraph{Metrics, statistics, models.} \emph{Exact match} is the primary metric (unparsable answers count as wrong). Transcriptions are scored by edge F1 (cell accuracy and black-cell F1 for images) and by \emph{exact transcription}, which labels each error \emph{perceptual} (transcription wrong) or \emph{computational} (transcription exact, answer wrong). We record thinking and output tokens for every call and, for programs, the \emph{algorithm family} (algebraic, dynamic programming, backtracking, traversal, local formula; keyword classifier validated by hand, Appendix~\ref{app:code}), execution success and time-outs. Proportions carry Wilson 95\% intervals~\cite{wilson1927probable}; twin gaps use the exact McNemar test on paired same-image outcomes~\cite{mcnemar1947note} with the discordant counts $b/c$, a 95\% interval for the paired difference from a cluster bootstrap over images ($B=2000$), and, where a table reports many tests, Holm-adjusted $p$-values~\cite{holm1979simple} and the number of tests passing Benjamini--Hochberg at $q<0.05$~\cite{benjamini1995controlling}. Logistic regressions on the answer's magnitude test the class term by likelihood ratio and bootstrap its coefficient over images; they use the benchmark proper only, never the extra-large grids. Generations use temperature~0 (plus one $T{=}1$ control in Appendix~\ref{app:v5}, which also measures $T{=}0$ repeat drift). Via the Gemini API we evaluate \nConfigs{} configurations of \nModels{} models: \texttt{gemini-3.8-flash} (low/medium/high), \texttt{3.6-flash} (minimal/low/high), \texttt{3.7-flash}, \texttt{3.5-flash}, \texttt{3.1-flash-lite} and \texttt{3-flash-preview} (low), \texttt{3.1-pro-preview} (low/high), and open-weight \texttt{gemma-4-31b-it} (off/on) and \texttt{26b-a4b-it} (off)~\cite{gemini3models,gemma4}. Minimal/off means no thinking tokens (visible output allowed); public API model IDs~\cite{gemini3models} matched the logged \texttt{modelVersion} strings on the three main models, and the core and seed-expansion results rest on \vFourNUniqueCalls{} unique scored calls (Appendix~\ref{app:v4}). Under vLLM with exact token caps $b\in\{512,2048,8192,\text{free}\}$, we also self-host four Gemma~4 models (\texttt{31B}, \texttt{26B-A4B}, \texttt{12B}, \texttt{E4B-it}; \gNineTotalCalls{} calls, Appendix~\ref{app:g9})\ifincludeQW{} and Qwen3-VL (\texttt{32B-Thinking}, \texttt{8B-Thinking}, plus non-thinking \texttt{8B-Instruct} at \texttt{free}; \qWTotalCalls{} calls, Appendix~\ref{app:qw})\fi.

\section{Seeing at Benchmark Scale}
\label{sec:res_percept}
Numbers are exact-match percentages for \texttt{gemini-3.8-flash} (low), our primary configuration, unless noted; all configurations, per-size tables, intervals and counts are in Appendix~\ref{app:tables}. Over the \PerceptPooledN{} Visual calls with a scorable transcription (all configurations), mean edge or cell F1 is \PerceptPooledF{} and the transcription is exact in \PerceptPooledExact\% of calls (Table~\ref{tab:perception}); for the primary model it is exact in 88\% of grid drawings and in all binary images and Eulerian drawings. Reading depends on the model: \texttt{3.1-flash-lite}, the Gemma~4 API runs without thinking (Table~\ref{tab:perception}) and the self-hosted Gemma~4\ifincludeQW{} and Qwen3-VL\fi{} runs transcribe far fewer core images exactly, and the self-hosted runs' Visual cells are limited by reading (Appendix~\ref{app:g9}\ifincludeQW{}, Appendix~\ref{app:qw}\fi). Table~\ref{tab:attribution} splits the Visual errors: 115 of the primary model's 144 errors (80\%) come with an exact transcription. The other 29 also fail in the Text control, so this split cannot tell which step caused them. We therefore solved all \vFourOneAPairs{} pooled pairs from one task-independent transcription per image (Table~\ref{tab:v41_transcription}--Table~\ref{tab:v41_paired}): \vFourOneAImagesExact/\vFourOneAImagesParsed{} images are transcribed exactly, and of the errors \vFourOneACOnly{} come from computation alone, \vFourOneAPOnly{} from perception alone, \vFourOneABoth{} from both and \vFourOneACancel{} from error cancellation. Pooled over configurations, \AttribPooledPct\% of Visual errors come with an exact transcription, and removing the image changes accuracy by a few points at most (Table~\ref{tab:main_visual}, Table~\ref{tab:main_text}). At benchmark size most errors happen after the drawing has been read; \S\ref{sec:car_results} shows where this stops.

\section{Counting in Chain of Thought}
\label{sec:cot}
\subsection{Same image: accuracy follows answer size}
\label{sec:res_twins}
Table~\ref{tab:twins} gives the paired same-image tests on the core slice; Appendix~\ref{app:v3_ext} and Appendix~\ref{app:v4} repeat them on the pooled sample ($N=\vThreePooledPairs{}$ pairs). The most telling pair is $\sST$ against $\sPATH$, where class and answer size point in opposite directions: path counts are small (median 268 against 540{,}011 for spanning trees), so a procedure whose cost grows with the answer finds the $\shP$ member easy and the $\GapL$ member hard (Fig.~\ref{fig:acc_vs_n}), and in Table~\ref{tab:twins} the primary model's $\sST$ beats neither $\sPATH$ nor $\sAM$. On 80 pre-registered twins matched within $0.25$ decades per graph (Appendix~\ref{app:t6}), which hold answer size fixed, $\GapL$ wins, pooled over eight configurations, \vSixTSixPoolVisualShare\% of discordant pairs under Visual ($p={}$\vSixTSixPoolVisualP) and \vSixTSixPoolTextShare\% under Text ($p={}$\vSixTSixPoolTextP). In the two pairs that follow the ladder on the primary model, $\sPM$ over $\sAM$ and $\sECd$ over $\sEC$, the $\GapL$ member also has the smaller answer (\S\ref{sec:twins}), so these gaps do not separate class from size; of the \vFourTwinsMainTableVisualNTests{} tests in the table, \vFourTwinsMainTableVisualSigHolm{} survive Holm, both $\sPM$ vs.\ $\sAM$ (pooled $p=4.6{\times}10^{-7}$, and $p<10^{-4}$ on \texttt{3.6-flash} (low) and \texttt{3.1-pro} (low); all tests in Table~\ref{tab:v4_twins_pooled}). $\chi$ and $\sCC$ are both at 99--100\%.

A logistic regression of exact match on $\log_{10}$ of the true answer and on class, over the grid and Eulerian families (answers from $1$ to $6.6{\times}10^{12}$; Table~\ref{tab:v4_magnitude} adds the extra-large pairs), makes the role of answer size quantitative (Visual in Fig.~\ref{fig:teaser}(b); Visual and Text in Fig.~\ref{fig:magnitude}). Under Visual the odds of an exact answer fall by a factor \vFourMagPooledNoxlVisualORinv{} per decade ($n=\vFourMagPooledNoxlVisualN$; \vFourMagCoreNoxlVisualORinv{} on the core slice, $n=\vFourMagCoreNoxlVisualN$; \vFourMagPooledNoxlTextORinv{} under Text). At equal magnitude the $\shP$ member is still harder (LR $p={}$\vFourMagPooledNoxlVisualP; Text $p={}$\vFourMagPooledNoxlTextP): $m_{50}$, the answer at which accuracy crosses 50\%, is $10^{\vFourMagPooledNoxlVisualMfiftyPoly}$ for $\GapL$ and $10^{\vFourMagPooledNoxlVisualMfiftyHard}$ for $\shP$ ($10^{\vFourMagPooledNoxlTextMfiftyPoly}$ and $10^{\vFourMagPooledNoxlTextMfiftyHard}$ under Text), a gap of \vFourMagPooledNoxlVisualMfiftyDiff{} decade (bootstrap CI $[\vFourMagPooledNoxlVisualMfiftyDiffCIlo, \vFourMagPooledNoxlVisualMfiftyDiffCIhi]$), against answers that span thirteen decades. This class term is fragile (Table~\ref{tab:v4_covariates}, Table~\ref{tab:v4_discordant}). Instance covariates do not remove it, but leaving out $\sST$ or $\sPATH$ does; it is not detectable within the Eulerian family or on \texttt{3.6-flash} and \texttt{3.1-pro}, its sign is not stable across configurations (in 8 of the 14 per-configuration fits the $\shP$ member has the higher $m_{50}$; Table~\ref{tab:m50}), and it mostly says that $\sPATH$ is harder than its small answers predict. Within an image, when the two answers lie within a decade of each other, discordant pairs split 14 to 16 ($\shP$ to $\GapL$) under Visual and 7 to 24 under Text; once the $\shP$ answer is at least a decade larger they split 2 to 45 and 3 to 48.

\begin{table}[!t]\centering\scriptsize
\setlength{\tabcolsep}{3.5pt}\renewcommand{\arraystretch}{0.92}
\caption{Thinking budget within a family, Visual condition: class-average exact match (\%) on the instances shared by all levels (202 for \texttt{gemini-3.6-flash}, 152 for \texttt{gemini-3.8-flash}), on the calls that returned at first; mean thinking plus output tokens per call and $m_{50}$ from the per-configuration logistic fit (Table~\ref{tab:m50}). $^\ast$With the reruns of Appendix~\ref{app:rerun}; the calls that returned at first give $10^{\vFiveHiVisMfiftyNow}$.}
\label{tab:budget}
\begin{tabular}{llrrrrrr}
\toprule
Model & thinking & $\TC$ & $\Lsp$ & $\GapL$ & $\shP$ & tokens & $m_{50}$ \\
\midrule
3.6 Flash & minimal & 100 & 95 & 29 & 26 & 1.8k & $10^{1.2}$ \\
 & low & 100 & 100 & 47 & 41 & 7.0k & $10^{2.2}$ \\
 & high & 100 & 100 & 68 & 57 & 25.4k & $10^{3.9}$ \\
\midrule
3.8 Flash & low & 100 & 100 & 69 & 67 & 5.7k & $10^{3.2}$ \\
 & medium & 100 & 100 & 86 & 78 & 13.2k & $10^{4.5}$ \\
 & high & 100 & 100 & 92 & 83 & 27.4k & $10^{\vFiveCHiVisMfiftyMerged}$\,$^\ast$ \\
\bottomrule
\end{tabular}
\end{table}

\subsection{Thinking budget buys decades of answer size}
\label{sec:res_budget}
Within a family we vary only the thinking level (Table~\ref{tab:budget}; all configurations in Table~\ref{tab:budget_visual}). Both upper rungs rise, and across the \vSixFSlopeMergedCfiveK{} configurations of \vFourBudgetModels{} models with a fitted $m_{50}$, $m_{50}$ grows by \vSixFSlopeMergedCfiveTwo{} decades per tenfold increase in tokens (95\% CI $[\vSixFSlopeMergedCfiveCIlo, \vSixFSlopeMergedCfiveCIhi]$; Appendix~\ref{app:rerun}), counting the \vFiveHiVisLostPct\% of high-thinking \texttt{3.8-flash} Visual calls that returned only when rerun (\vSixFSlopeReturnedTwo{} without them; Table~\ref{tab:m50}). Fitted per class, $\GapL$ moves further on all three ladders (\texttt{3.8-flash}, \texttt{3.6-flash}, \texttt{3.1-pro}), but on the 80 matched twins more thinking does not favour $\GapL$ (both CIs include zero; Appendix~\ref{app:t6}), so this gap is also answer size. Any cost--answer power law gives a constant decade-per-budget slope, and tokens per call follow the rung only through answer size (Table~\ref{tab:tokens}). Under exact token caps on self-hosted models, the magnitude law holds in \gNineHOneSupported/\gNineNCells{} Gemma~4 cells\ifincludeQW{} and \qWHOneSupported/\qWHOneNCells{} Qwen3-VL cells\fi{}, and $m_{50}$ rises with budget in \gNineHTwoSupported/\gNineNCells{} Gemma~4 cells\ifincludeQW{} and in both Qwen3-VL Text cells\fi{} (Holm-corrected; Appendix~\ref{app:g9}\ifincludeQW{}, Appendix~\ref{app:qw}), partly because more calls finish within the cap (Appendix~\ref{app:trunc}\fi).

\subsection{Naming the theorem does not help}
\label{sec:res_hint}
If models enumerate because they do not know the polynomial-time theorems, naming them should help; it does not. For the primary model (391 instances) the hint changes exact match by at most 8 points on every task except $\sEC$ ($+22$), whose hint describes an enumeration the model can run; over the four hinted configurations the $\GapL$ average moves by $-3$ points and the $\shP$ average by $0$ (Table~\ref{tab:main_hint}). The theorems are known: asked to classify the problem before coding (the routing prompt), the models declare the right complexity class in \CarClassAcc{} of programs (Table~\ref{tab:car_class}), and in code they implement determinants unprompted (\S\ref{sec:res_code}). In chain of thought they rarely carry them out: 0 of 288 $6{\times}6$ spanning-tree calls are exact across all configurations and chain-of-thought conditions, and only \LargeAnsExact{} of \LargeAnsN{} Visual and Text calls with an answer of at least $10^6$ are exact (\LargeAnsExactMedHigh{} at medium or high thinking; \LargeAnsPrimExact{} of \LargeAnsPrimN{} for the primary model). Enumeration cannot reach such answers within the up to \LargeAnsMaxTok{} tokens used, and 11 of the 12 exact Visual $\sST$ answers name a Matrix-Tree determinant.

\section{Counting in Code, and Reading at Scale}
\label{sec:code}
\subsection{Code runs what chain of thought rarely finishes}
\label{sec:res_code}
Letting the model write and run code removes arithmetic and execution limits (Table~\ref{tab:code}). The primary model reaches 93\% with Code vs.\ 63\% with Visual on the 392 instances (0 time-outs); 62\% of $\sST$, $\sPM$ and $\sECd$ programs compute a determinant/Pfaffian and 27\% of $\shP$ programs use dynamic programming (the rest enumerate); from text (Text$\to$Code) the figures are 96\% exact, 64\% and 32\%. As instances grow, the models switch from enumeration to determinants or frontier dynamic programs (Table~\ref{tab:algo_size}): $\sAM$ dynamic-programming usage rises from 0/6 at $3{\times}3$ to 10/10 from $5{\times}6$ on ($100\%$ from $7{\times}7$ on); $\sST$ uses Kirchhoff at every size, while Kasteleyn for $\sPM$ appears only on the largest grids (3/6 at $10{\times}10$). On the benchmark proper the choice does not affect accuracy, as Proposition~\ref{prop:car} predicts: the routing prompt, a control that should beat Code only where enumeration is too slow, reaches \vFourCCCoreNoxlGThreeEightLowCarCodeAccA\% against \vFourCCCoreNoxlGThreeEightLowCarCodeAccB\% for Code on the same \vFourCCCoreNoxlGThreeEightLowCarCodeN{} instances ($p=\vFourCCCoreNoxlGThreeEightLowCarCodeP$; pooled $p=\vFourCCPooledNoxlGThreeEightLowCarCodeP$; per task in Table~\ref{tab:car}), \texttt{3.6-flash} and \texttt{3.1-pro} tie as well, and the only two significant differences point in opposite directions (Table~\ref{tab:v4_carcode}), although the declared class is right in \CarClassAcc{} of programs (Table~\ref{tab:car_class}). On extra-large grids the choice is decisive.

\begin{table}[!t]\centering\scriptsize
\setlength{\tabcolsep}{3pt}
\caption{Code conditions, primary model (392 instances); all columns in \% except tokens. exec: programs that run; t/o: time-outs at 60\,s; libs: programs that import a library; alg.: $\sST$, $\sPM$, $\sECd$ programs that compute a determinant/Pfaffian; DP: $\sAM$, $\sPATH$, $\sEC$ programs that use memoised/frontier dynamic programming rather than enumeration. All models and prompts: Table~\ref{tab:code_all}.}
\label{tab:code}
\input{gen_v8/tab_code_main_v84.tex}
\end{table}

\subsection{Reading large drawings}
\label{sec:car_results}
\paragraph{Code works from text; transcription fails from pixels.} Table~\ref{tab:xlgrid} collects the extra-large grids ($7{\times}7$ to $10{\times}10$; other models in Table~\ref{tab:xl_summary}), where Proposition~\ref{prop:car} says routing should matter. Chain of thought solves none, while programs written from the exact text solve almost all, including $\sAM$ at $10{\times}10$, because given a large input the models write a frontier dynamic program on their own. From pixels the same prompts solve few pairs, and on 96 fresh pre-registered grids (Appendix~\ref{app:x8}) the explicit transcription step of the routing prompt does not help (Holm $p={}$\vSixXEightHTwoPHolm). The failures are transcription failures: of the 101 wrong routing-prompt programs on grid instances, 53 compute the right answer for a wrongly transcribed graph, 9 a wrong answer for the right graph, 13 do both and 26 embed no edge list (Table~\ref{tab:attrib}), and on the extra-large grids all \vFourDecXlCarTZeroAZero{} wrong routing-prompt programs embed a wrong graph, \vFourDecXlCarTwrongExecMatchesGhat{} of them solving that graph exactly (Table~\ref{tab:v4_transcription}).

\begin{table}[!t]\centering\scriptsize
\setlength{\tabcolsep}{2.6pt}
\caption{Extra-large grids ($7{\times}7$ to $10{\times}10$) and large rendered diagrams, primary model: grids transcribed exactly and (instance, task) pairs solved. One shot: one call reads and solves. Two-view: upright plus $90^\circ$-rotated reading, a crop check of each disputed pair, then code. \car{}: the routing prompt (a control). Rendered diagrams are solved by an exact solver from the transcription, so only exact transcriptions are shown. $^\dagger$Pre-registered. Tests are in the text; other models in Table~\ref{tab:xl_summary}.}
\label{tab:xlgrid}
\resizebox{\linewidth}{!}{%
\begin{tabular}{@{}llcc@{}}
\toprule
Set & Method & exact & solved \\
\midrule
24 grids & chain of thought (all configurations) & & 0/60 \\
 & Code, one shot & & 6/60 \\
 & Perceive+Code or \car{}, one shot & & 14/60 \\
 & exact text: Text$\to$Code / Text\,\car{} & & 57/60 / 54/60 \\
 & one view / Zoom-$k$ crops ($k{=}2$, $3$) & 13/24 / 1/24, 7/24 & \\
 & two-view ($90^\circ$): Text$\to$Code / Text\,\car{} & 22/24 & 49/60 / \VerifyNumbers \\
 & rerun: two-view / two upright ($T{=}0$) & \vFiveBTVExact/24 / \vFiveBBthreeExact/24 & \\
\midrule
96 grids$^\dagger$ & two-view ($90^\circ$), then Text\,\car{} & \vSixBSevenTVExact/96 & \vSixXEightHOneKa/\vSixXEightPairs \\
 & two-view ($180^\circ$) & \vSixBSevenTVExact/96 & \\
 & two upright readings ($T{=}0$) & \vSixBSevenBthreeExact/96 & \\
 & \car{} / Code, one shot & & \vSixXEightHTwoKa/\vSixXEightPairs{} / \vSixXEightHTwoKb/\vSixXEightPairs \\
 & exact text: Text$\to$Code & & \vSixXEightGThreeEightLowTcodeK/\vSixXEightPairs \\
\midrule
48 rendered$^\dagger$ & one view / quadrant zoom & \rNineOneVGraphK/48 / \rNineZoomGraphK/48 & \\
 & two-view, $90^\circ$ / $180^\circ$ & \rNineTwoVGraphK/48 / \rNineRotOneEightyGraphK/48 & \\
\bottomrule
\end{tabular}}
\end{table}

\paragraph{Why reading fails at scale.} Single-view misses are anisotropic: in whole-image reading 38 of 1,384 horizontal edges are missed against 6 of 1,393 vertical ones, and rotating the image by $90^\circ$ moves the misses to the edges that are now horizontal ($p={}$\vFourOrientMissHPimages{} on the core grids; Table~\ref{tab:orient}, Table~\ref{tab:v4_orient}); cutting the image into crops (Zoom-$k$, Appendix~\ref{app:codecond}) makes reading worse (Table~\ref{tab:zoom}). The misses are systematic for a given view, and a second, decorrelated reading repairs most of them. In our first run the union of an upright and a rotated reading missed none of 2,777 true edges, and a crop check of each disputed pair (8.6 per grid) answered 207/207 correctly (Table~\ref{tab:v41_xl}; other models in Table~\ref{tab:twoview_models}). On 96 fresh pre-registered grids, a second reading of a $90^\circ$ or of a $180^\circ$ copy with crop checks is exact on \vSixBSevenTVExact/\vSixBSevenTVNGrids{} grids either way ($b/c$ \vSixBSevenTVVsRotExactB/\vSixBSevenTVVsRotExactC), so the gain does not depend on turning horizontal edges vertical, whereas repeating the identical upright request at $T{=}0$ gives \vSixBSevenBthreeExact/\vSixBSevenTVNGrids{} (Holm $p\le{}$\vSixBSevenRotVsBthreeExactPHolm{} for either rotation; Appendix~\ref{app:v6}). The two upright readings miss the same edges (\vSixBSevenBthreeBothMissed{} missed by both, against \vSixBSevenTVBothMissed{} and \vSixBSevenRotBothMissed{} with a $90^\circ$ or $180^\circ$ copy). A same-session rerun on the first 24 grids is not significant ($p={}$\vFiveBBthreeExactP; Appendix~\ref{app:v5}), and two-view reading also beats the one-shot routing prompt (Holm $p={}$\vSixXEightHOnePHolm; Appendix~\ref{app:x8}).

\paragraph{Rendered diagrams and held-out tasks.} In a pre-registered test on \rNineTotalGraphs{} graphs rendered in four realistic styles (PCB traces, blueprint floorplans, transit maps, slanted desk scenes with cast shadows; Appendix~\ref{app:r9}), \rNineCritPass{} of the \rNineCritN{} registered criteria hold on the primary model: Direct accuracy falls with answer magnitude ($p=\rNineBDirectLogitP$), single views miss horizontal edges $\rNineAnisoRatio$ as often as vertical ones ($p=\rNineAnisoWilcoxonP$), and on the 48 large diagrams two-view reading beats one view (one-sided McNemar over graph--task pairs, $p=\rNineTwoVvsOneVP$) and quadrant zoom (Table~\ref{tab:xlgrid}). Four fail: the routing prompt gains less than registered, matched $\sST$/$\sAM$ twins differ by \rNineTwinDirectSTvsAMDiff{} points ($p=\rNineTwinDirectSTvsAMP$), $\sPM$ leads $\sAM$ by less than registered, and a $180^\circ$ second view does as well as $90^\circ$, so the repair comes from a second reading of a rotated copy with crop checks rather than from swapping axes. On three held-out counting tasks (independent sets, $\shP$-complete~\cite{vadhan2001complexity}; dominating sets, $\shP$-complete; shortest corner-to-corner paths, linear time; 198 instances each), Code and the routing prompt both reach 90\% against 56\% for Visual (Table~\ref{tab:heldout}).

\section{Discussion and Conclusion}
\label{sec:discussion}
At benchmark scale, chain-of-thought success tracks answer magnitude far more than class; the class adds $\approx 1$ decade of $m_{50}$ at equal magnitude in the primary model's pooled data (concentrated in one twin pair and not detectable on \texttt{3.6-flash} or \texttt{3.1-pro} in the pooled sample), so varying class without controlling for magnitude mostly measures answer size, a caution for benchmarks that grade difficulty by complexity class~\cite{fan2024nphardeval,fan2024nphardeval4v}.

\paragraph{Limitations.} Degraded renderings (Appendix~\ref{app:d8}) and four realistic rendering styles (Appendix~\ref{app:r9}) reproduce the magnitude law and the blind spot, but every image is rendered from a known graph; photographs and natural scenes with occlusion remain untested. The API runs cover \nConfigs{} configurations of \nModels{} models from one provider with coarse thinking levels, provider token counts, and unlogged resolved versions in the initial sweep (Appendix~\ref{app:v4}); our vLLM runs add self-hosted Gemma~4\ifincludeQW{} and Qwen3-VL\fi{} under exact token caps, though other families may differ. Algorithm-family labels use a hand-validated rule classifier (Appendix~\ref{app:code}). The magnitude regression is observational: answer size covaries with instance size within a task, and because class is nested in task, dropping $\sST$ or $\sPATH$ removes the class term, so we report $\approx 1$ decade of $m_{50}$ tied to specific twin pairs rather than a class constant. Finally, the classes are asymptotic (\S\ref{sec:predictions}), and at bounded grid width both members of a grid twin are polynomial-time (Proposition~\ref{prop:car}).

\paragraph{Conclusion.} In chain of thought, most Gemini API models read small diagrams correctly and then fail at counting, at a rate set mainly by the size of the answer rather than by its complexity class; given code and an exact transcription, the Gemini models we tested with code count correctly, and reading large diagrams becomes the failure. Complexity twins expose the two regimes, and magnitude-matched twins separate class from answer size. At scale, a second, decorrelated reading with pair-crop checks largely repairs an anisotropic horizontal-edge blind spot in plain and realistically styled diagrams.

\clearpage
\section*{LLM Usage Statement}
Apart from the models evaluated in this study, we used a non-public agentic coding and writing assistant built on commercial LLMs (Claude and Gemini model families).
We used it extensively: it helped brainstorm candidate research directions (we chose this one), wrote the code and ran the experiments under our direction, drafted and revised the text and figures, and searched for related work.
Results were checked with unit tests and recomputation from raw logs, every reference was checked against its source, and we take full responsibility for all content.

{\small
\bibliographystyle{ieeenat_fullname}
\bibliography{refs_v4}
}

\clearpage
\onecolumn
\appendix
\raggedbottom
\tolerance=200
\emergencystretch=0pt
\renewcommand{\topfraction}{0.99}
\renewcommand{\bottomfraction}{0.99}
\renewcommand{\textfraction}{0.01}
\renewcommand{\floatpagefraction}{0.99}
\renewcommand{\dbltopfraction}{0.99}
\renewcommand{\dblfloatpagefraction}{0.99}
\setcounter{topnumber}{99}
\setcounter{bottomnumber}{99}
\setcounter{totalnumber}{99}
\setcounter{dbltopnumber}{2}
\setlength{\textfloatsep}{20pt plus 2pt minus 4pt}
\setlength{\dbltextfloatsep}{20pt plus 2pt minus 4pt}
\setlength{\intextsep}{12pt plus 2pt minus 2pt}
\setlength{\abovecaptionskip}{10pt}
\setlength{\belowcaptionskip}{0pt}
\clearcaptionsetup{table}
\setcounter{table}{0}
\setcounter{figure}{0}
\renewcommand{\thetable}{S\arabic{table}}
\renewcommand{\thefigure}{S\arabic{figure}}
\renewcommand{\theHtable}{S\arabic{table}}
\renewcommand{\theHfigure}{S\arabic{figure}}
\begin{center}{\Large\bf Appendix\par}\end{center}

\noindent This appendix gives the proofs, the transformer predictions behind Section~\ref{sec:predictions}, the dataset statistics, the verbatim prompts, every result table referred to in the main paper with instance counts, the full-length summaries of each experiment (generated from the result files together with the tables), the code-classifier validation and examples, failure cases, compute and cost, the expanded-seed and rendering-style replication (6{,}856 additional evaluations), the sensitivity analyses with their provenance notes (\S\ref{app:v4}), the same-session rerun of Experiment B (\S\ref{app:v5}), its pre-registered replication on fresh grids (\S\ref{app:v6}), the one-shot conditions on the same fresh grids (\S\ref{app:x8}), the core images under ten degradations (\S\ref{app:d8}), magnitude-matched twins on eight thinking configurations with a copy and an explicit-algorithm control (\S\ref{app:t6}), the reruns of the calls that never returned (\S\ref{app:rerun}), the self-hosted Gemma~4 experiments under controlled thinking budgets (\S\ref{app:g9})\ifincludeQW{}, the self-hosted Qwen3-VL control family (\S\ref{app:qw}), the finished-calls truncation sensitivity check (\S\ref{app:trunc})\fi{}, the realistic engineering and physical diagram benchmark (\S\ref{app:r9}) and the code conditions, routing prompt and reading pipelines that the main paper summarises (\S\ref{app:codecond}). Table and figure numbers with the prefix S refer to this appendix; all other table and figure numbers refer to the main paper.

\section{Proofs and conventions}
\label{app:proofs}

\paragraph{Connectivity conventions.} Throughout, black (foreground) cells are 4-connected and white (background) cells are 8-connected; the image is padded by one ring of white cells so that the outer background is a single 8-connected region. $C_4(I)$ is the number of black 4-components and $H_8(I)$ the number of bounded white 8-components.

\begin{proposition}[bit-quad identity]
\label{prop:gray}
For every binary image $I$, $\chi(I)\triangleq C_4(I)-H_8(I)=N_{\bullet}(I)-N_{\bullet\bullet}(I)+N_{\boxplus}(I)$.
\end{proposition}
\begin{proof}
Let $\Gamma$ be the plane graph whose vertices are the centres of black cells and whose edges join 4-adjacent black cells by unit segments; $\Gamma$ is the induced subgraph of the lattice on the black cells, so $|V|=N_{\bullet}$ and $|E|=N_{\bullet\bullet}$. Euler's formula for a plane graph with $k$ connected components gives $|V|-|E|+|F|=1+k$, where $F$ includes the unbounded face. The bounded faces of $\Gamma$ are of two kinds. A unit square all of whose four corners are black is a face, because all four boundary edges are present; there are $N_{\boxplus}$ of them. Every other bounded face has area at least~2 and, since $\Gamma$ is induced, is not subdivided by a missing lattice edge; it therefore contains at least one white cell centre, and the white cell centres it contains form exactly one 8-connected white region: diagonal steps between white cells cross no edge of $\Gamma$ (edges lie on lattice lines, a diagonal crosses a lattice point only at the centre of a unit square that is not all black), and conversely a lattice path inside an open face can be discretised into 8-steps between white centres. Bounded white 8-regions touching the padded border lie in the unbounded face, and every bounded one lies in a bounded face. Hence the non-square bounded faces are in bijection with the holes, $|F|=1+N_{\boxplus}+H_8$, and $N_{\bullet}-N_{\bullet\bullet}+1+N_{\boxplus}+H_8=1+C_4$. (This is the classical bit-quad identity of Gray~\cite{gray1971local}; we include the argument to fix the conventions used by our generator, which asserts the identity on every image.)
\end{proof}

\begin{corollary}
$\chi\in\TC$ and $\chi\notin\mathsf{AC}^0$.
\end{corollary}
\begin{proof}
Each of the three counts is the number of ones in a vector of local predicates of at most four pixels, i.e.\ an iterated addition, which is in $\TC$~\cite{chandra1984constant}; the signed sum of three $O(\log n)$-bit numbers is in $\mathsf{AC}^0$. For the lower bound, an image whose black cells are pairwise non-adjacent has $\chi=N_{\bullet}$, so an $\mathsf{AC}^0$ circuit for the low-order bit of $\chi$ would compute parity, contradicting~\cite{furst1981parity,hastad1986almost}.
\end{proof}

\begin{proposition}[BEST with vanishing factorial]
\label{prop:best}
Let $D$ be a connected digraph with $d^-(v)=d^+(v)\in\{1,2\}$ for every vertex. The number of Eulerian circuits of $D$ (as cyclic sequences of arcs) equals $t_w(D)=\det L^{+}_{-w}$ for any vertex $w$, where $L^+$ is the out-degree Laplacian and $L^{+}_{-w}$ is $L^+$ with row and column $w$ removed.
\end{proposition}
\begin{proof}
By the BEST theorem~\cite{vanaardenne1951circuits,smith1941unicursal}, $\mathrm{ec}(D)=t_w(D)\prod_v(d^+(v)-1)!$ where $t_w$ is the number of spanning arborescences oriented towards $w$; the product is $1$ because every factor is $0!$ or $1!$. The directed matrix-tree theorem expresses $t_w$ as the stated cofactor. Our generator asserts the factorial product equals~1 on every instance and cross-checks $t_w$ by two determinant routines.
\end{proof}

\paragraph{What the ladder predicts for a transformer.} Section~\ref{sec:predictions} summarises the following facts and proposition. (T1)~A log-precision fixed-depth transformer answering \emph{without} chain of thought computes a function in uniform $\TC$~\cite{merrill2023parallelism}. (T2)~With $t(n)$ chain-of-thought tokens it computes functions in $\mathsf{TIME}(\mathrm{poly}(n+t(n)))$, and $t(n)=O(\log n)$ stays inside $\Lsp$~\cite{merrill2024expressive,li2024chain}. (T3)~Polynomially many tokens suffice to simulate any polynomial-size circuit~\cite{li2024chain}.
\begin{proposition}[predictions]
\label{prop:pred}
Let $\mathcal M$ be a family of fixed-depth log-precision transformers; parts (b) and (c) assume one set of weights for all $n$, and with weights that depend on $n$ they give $\NC{1}\subseteq\TC/\mathrm{poly}$ and $\GapL\subseteq\FL/\mathrm{poly}$ instead. (a)~$\chi$ lies in uniform $\TC$, so the known $\TC$ upper bound on $\mathcal M$ without chain of thought~\cite{merrill2023parallelism} does not exclude it; this is a statement about the class, not a guarantee that a trained model computes $\chi$. (b)~If $\mathcal M$ computes $\sCC$ on all grid images without chain of thought then $\TC=\NC{1}$. (c)~$\sST$, $\sPM$ and $\sECd$ are computable with $\mathrm{poly}(n)$ thinking tokens; conversely, if $\mathcal M$ computes integer determinants, to which all three reduce and to which $\sPM$ on weighted planar bipartite graphs is equivalent~\cite{datta2010planarity}, with $O(\log n)$ tokens for all $n$, then $\GapL\subseteq\FL$. (d)~If $\mathcal M$ computes $\sAM$, $\sPATH$ or $\sEC$ on the graph class on which Table~\ref{tab:ladder} lists it as $\shP$-complete, with $\mathrm{poly}(n)$ thinking tokens for all $n$, then $\shP\subseteq\FP/\mathrm{poly}$ and the polynomial hierarchy collapses~\cite{toda1991pp,karp1980some}.
\end{proposition}
The proposition does not say what a finite model does at $n=36$; it says how the \emph{required} budget must scale with $n$ for each rung, hence which comparisons are informative. No condition we ran is the no-chain-of-thought setting of T1: with thinking minimal or off a model emits no thinking tokens but still writes visible output before its answer (mean tokens per call in Table~\ref{tab:m50}).

\begin{proof}[Proof of Proposition~\ref{prop:pred}]
(a) By Proposition~\ref{prop:gray} and its corollary $\chi$ is in uniform $\TC$; T1 is an upper bound on the class, and the claim is that the class contains the target (existence of weights is a separate matter, which we do not claim). (b) If a no-chain-of-thought family computes $\sCC$, then by T1 $\sCC\in\TC$; deciding $C_4(I)=1$ is then in $\TC$, and undirected grid-graph reachability reduces to it by an $\mathsf{AC}^0$ reduction, so $\NC{1}\subseteq\TC$ by~\cite{allender2009planar}. (c) Upper bound: each of the three counts is a determinant of an integer matrix of size $O(n)$ built by a logspace transducer from the drawing, the determinant is in $\GapL\subseteq\NC{2}\subseteq\mathsf{P}$, and T3 gives a $\mathrm{poly}(n)$-token chain of thought that simulates the polynomial-size circuit. Lower bound: by T2, $O(\log n)$ tokens keep the computed function inside $\FL$; if that function is the integer determinant, then $\GapL\subseteq\FL$ because the determinant is $\GapL$-complete~\cite{mahajan1997determinant}, and the same conclusion follows for $\sPM$ on weighted planar bipartite graphs, which is logspace-equivalent to the determinant~\cite{datta2010planarity}. We do not claim that the unweighted lattice instances we actually use are $\GapL$-hard (no such theorem is known to us), which is why the lower bound in (c) is phrased for the determinant. (d) If a family with $\mathrm{poly}(n)$ tokens computes a $\shP$-complete function, T2 places it in $\mathsf{TIME}(\mathrm{poly})$ with the family's weights as advice, so $\shP\subseteq\FP/\mathrm{poly}$; by Toda's theorem $\PH\subseteq\mathsf{P}^{\shP}\subseteq\mathsf{P}/\mathrm{poly}$, which collapses the hierarchy by the Karp--Lipton theorem.
\end{proof}

\begin{proof}[Proof of Proposition~\ref{prop:car} (cost of a correct route)]
Kirchhoff's and the directed matrix-tree determinants are $(n-1)\times(n-1)$; Kasteleyn's is $(n/2)\times(n/2)$; Bareiss elimination computes each exactly in $O(n^3)$ ring operations on integers of $O(n\log n)$ bits. Orient the lattice so that its rows have $w$ cells. For $\sAM$ and $\sPM$, process cells in row-major order with a state recording, for the $w$ most recent cells, whether each is already matched; each cell is either left unmatched, matched to its right neighbour, or matched to the cell below, so there are at most $2^{w+1}$ states and $O(1)$ transitions per state per cell. For $\sPATH$ from $s$ to $t$, process the edges in row-major order and decide for each whether it lies on the path; the frontier, the processed vertices that still have an unprocessed edge, has at most $w+1$ vertices. The state records, for every frontier vertex, its degree in the chosen edges and, if that degree is~1, the other end of its path fragment (or that the fragment reaches $s$ or $t$): the mate array of frontier-based search, as in Knuth's SIMPATH~\cite{knuth2011taocp}. An edge that would close a cycle or give a vertex degree~3 is rejected, and so is a vertex other than $s$ and $t$ that leaves the frontier with degree~1. The fragments lie in the processed part of a plane graph and the frontier vertices lie in order on its boundary, so no two fragments join interleaved pairs of frontier vertices: the pairings do not cross, and there are $2^{O(w)}$ states. Each of the $O(n)$ edges updates every state in $O(w)$ steps, and every count is at most $2^{|E|}\le 4^n$, so the program performs $n\,2^{O(w)}$ additions of $O(n)$-bit integers. Enumeration visits every matching (or path) at least once, and the number of matchings of a subgraph of the lattice containing a perfect matching is at least $2^{n/2}$. Our Eulerian drawings have at most 26 edges, and the independent ground-truth check of Appendix~\ref{app:data} counts their circuits by direct edge-sequence search.
\end{proof}

\section{Dataset details}
\label{app:data}
\input{gen_v5r/tab_dataset.tex}

\paragraph{Batch labels.} Tables in this appendix label the core slice v2, and the seed expansion and the re-rendered style variants (Appendix~\ref{app:v3_ext}) v3, after the generation batches that produced them.

\paragraph{Ground truth.} All counts use exact arithmetic. The generator checks $\sAM$ and $\sPM$ against an edge-inclusion recursion when $n\le16$, undirected Eulerian circuits against an edge-sequence search when the drawing has at most 18 edges, and $\chi$ from the bit-quad formula against $C_4-H_8$ from flood fills. An independent re-implementation that shares no code with the generator computes spanning trees by a Bareiss determinant of the reduced Laplacian, all and perfect matchings by a row-major frontier dynamic program, simple corner-to-corner paths by depth-first search with reachability pruning, $C_4$, $H_8$ and $\chi$ by flood fills (with $\chi$ also from the bit-quad formula), directed Eulerian circuits by the BEST theorem and by arc-sequence search, and undirected ones by edge-sequence search. It reproduces every answer of the eight tasks on all \GTCheckN{} instances (\GTCheckNCore{} core, \GTCheckNSeed{} seed and \GTCheckNXL{} extra-large), with \GTCheckMismatch{} mismatches.

\section{Prompts}
\label{app:prompts}
\input{gen_v8p/gen_v5r/prompts.tex}

\section{Full result tables}
\label{app:tables}
\subsection{Visual condition, per task}
\PerceptSummary{} \AttribSummary{}

\begin{center}\scriptsize
\resizebox{\linewidth}{!}{\input{gen_v5r/tab_main_visual.tex}}
\end{center}
\captionof{table}{Exact match (\%) on the Visual condition for every model/thinking configuration, by task (columns ordered by complexity class) and averaged by class. $n$ per cell is 56--66 (full) or 23--30 (subset) instances; the per-size tables (\S\ref{app:sizes}) give the counts.}\label{tab:main_visual}

\begin{center}\scriptsize\renewcommand{\arraystretch}{0.88}
\input{gen_v5r/tab_attribution.tex}
\end{center}
\captionof{table}{Error attribution on the Visual condition for every configuration (Table~\ref{tab:attribution} shows five of them). P: transcription exact; C: answer exact. Columns are percentages of instances in each class.}\label{tab:attribution_all}
\subsection{Class averages and thinking budget}
\BudgetSummary{} \BudgetDecadesSummary{}
\begin{center}\scriptsize\input{gen_v5r/tab_budget_visual.tex}\end{center}
\captionof{table}{Class-average exact match (\%) and mean thinking + output tokens (thousands) on the Visual condition, for every model/thinking configuration (core benchmark).}\label{tab:budget_visual}
\begin{center}\scriptsize\input{gen_v5r/tab_m50.tex}\end{center}
\captionof{table}{The magnitude at which exact match crosses 50\%, $m_{50}$, from a logistic fit of the Visual outcome on $\log_{10}$(answer) per configuration (grid and Eulerian families). ``tokens'': mean thinking + output tokens per call. The last column splits the fit by class.}\label{tab:m50}

\paragraph{Calls that never returned.} Our loader keeps the last successful record of each call, and \vFiveLostCoreKeys{} of the \vFiveAllCoreKeys{} core calls never succeeded. Most belong to \texttt{gemini-3.8-flash} at high thinking, where the API answered HTTP 503 (``high demand'') to every retry for \vFiveHiVisLost/\vFiveHiVisKeys{} Visual, \vFiveHiTextLost/\vFiveHiTextKeys{} Text and \vFiveHiHintLost/\vFiveHiHintKeys{} Hint calls; medium thinking lost \vFiveMedVisLost/\vFiveMedVisKeys{} Visual calls, and the \texttt{3.6-flash} ladder lost \vFiveGThreeSixVisLost{} Visual call. The losses are not spread evenly: among the \vFiveHiVisFitTotal{} instances of the high-thinking $m_{50}$ fit, the \vFiveHiVisFitLost{} lost Visual calls have median answer $10^{\vFiveHiVisMedLost}$ against $10^{\vFiveHiVisMedKept}$ for the \vFiveHiVisFitReturned{} returned ones. Table~\ref{tab:m50} fits the \vFiveHiVisFitReturned{} returned calls ($10^{\vFiveHiVisMfiftyNow}$; $\GapL$ $10^{\vFiveHiVisMfiftyPolyNow}$, $\shP$ $10^{\vFiveHiVisMfiftyHardNow}$); scoring the lost calls as wrong gives $10^{\vFiveHiVisMfiftyLow}$ ($10^{\vFiveHiVisMfiftyPolyLow}$ and $10^{\vFiveHiVisMfiftyHardLow}$), and for Text $10^{\vFiveHiTextMfiftyNow}$ against $10^{\vFiveHiTextMfiftyLow}$. Across the \vFourBudgetK{} configurations of Table~\ref{tab:m50} the budget slope falls from \vFourBudgetSlope{} to \vFiveSlopeLow{} decades per tenfold tokens (95\% CI $[\vFiveSlopeLowCIlo, \vFiveSlopeLowCIhi]$) and within the \texttt{3.8-flash} ladder from \vFourBudgetFamGThreeEight{} to \vFiveSlopeLowFamGThreeEight{}, while the \texttt{3.6-flash} ladder stays at \vFourBudgetFamGThreeSix. A 503 carries no answer, so the two fits above only bracket the effect of the losses. We reran these calls several days later, together with the other calls of the paper's configurations that never returned (\S\ref{app:rerun}). All \vFiveCReturnedLost{} \texttt{3.8-flash} calls returned; merged, the high-thinking Visual $m_{50}$ is $10^{\vFiveCHiVisMfiftyMerged}$ and the budget slope \vFiveCSlopeMergedTwo{} (95\% CI $[\vFiveCSlopeMergedCIlo, \vFiveCSlopeMergedCIhi]$; \vFiveCSlopeMergedFamGThreeEight{} within the \texttt{3.8-flash} ladder), inside the range above, and a drift control on \vFiveCDriftCtlN{} re-issued calls finds no detectable change.

\subsection{Text condition}
\begin{center}\scriptsize\input{gen_v5r/tab_main_text.tex}\end{center}
\captionof{table}{Exact match (\%) on the Text condition (ground-truth structure given as text, no image).}\label{tab:main_text}
\subsection{Hint condition}
\HintSummary{}
\begin{center}\scriptsize\input{gen_v5r/tab_main_hint.tex}\end{center}
\captionof{table}{Exact match (\%) on the Hint condition.}\label{tab:main_hint}
\subsection{Per-size tables}
\label{app:sizes}
For \texttt{gemini-3.8-flash} (low), \SizeSummary{}
\begin{figure}[htb]\centering
\includegraphics[width=\linewidth]{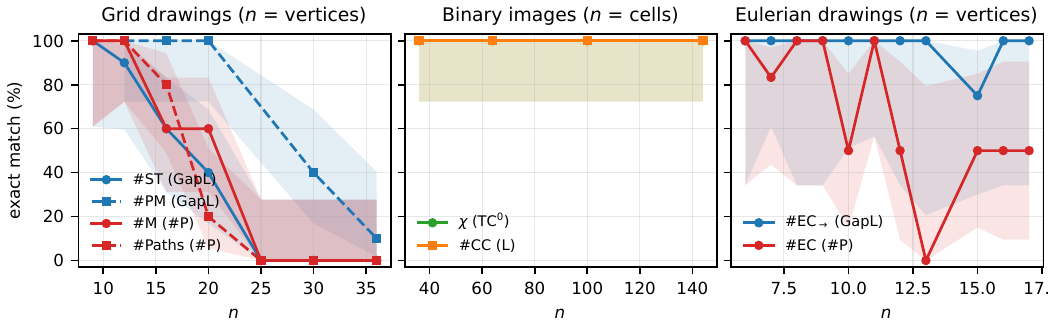}
\caption{Exact match against instance size for \texttt{gemini-3.8-flash} (low) on grid drawings (left) and binary images (right), Visual condition. Shaded bands are Wilson 95\% intervals; colours encode the class.}
\label{fig:acc_vs_n}
\end{figure}
\input{gen_v5r/tab_sizes_all.tex}
\subsection{Twin tests: all configurations, and the Text and Hint conditions}
\TwinSummary{} \MagnitudeSummary{}

\begin{center}\scriptsize\renewcommand{\arraystretch}{0.88}\input{gen_v5r/tab_twins_visual.tex}\end{center}
\captionof{table}{Paired twin tests, Visual condition, all configurations.}\label{tab:twins_all}
\begin{center}\scriptsize\renewcommand{\arraystretch}{0.88}\input{gen_v5r/tab_twins_text.tex}\end{center}
\captionof{table}{Paired twin tests, Text condition.}\label{tab:twins_text}
\begin{center}\scriptsize\input{gen_v5r/tab_twins_hint.tex}\end{center}
\captionof{table}{Paired twin tests, Hint condition.}\label{tab:twins_hint}
\subsection{Token tables}
\TokenSummary{}
\begin{center}\scriptsize
\resizebox{\linewidth}{!}{\input{gen_v5r/tab_tokens_visual.tex}}
\end{center}
\captionof{table}{Mean thinking + output tokens per call (thousands), Visual condition.}\label{tab:tokens}
\begin{center}\scriptsize\input{gen_v5r/tab_tokens_text.tex}\end{center}
\captionof{table}{Mean thinking + output tokens per call (thousands), Text condition.}\label{tab:tokens_text}
\begin{center}\scriptsize\input{gen_v5r/tab_tokens_hint.tex}\end{center}
\captionof{table}{Mean thinking + output tokens per call (thousands), Hint condition.}\label{tab:tokens_hint}
\subsection{Order-of-magnitude error}
\begin{center}\scriptsize\input{gen_v5r/tab_logerr_visual.tex}\end{center}
\captionof{table}{Median $|\log_{10}(\text{answer}/\text{truth})|$ over calls with a positive integer answer, Visual condition; $\emptyset k$ marks $k$ calls with no parsable answer.}\label{tab:logerr}
\subsection{Perception}
\label{app:perception}
{\scriptsize\input{gen_v5r/tab_perception.tex}}
\subsection{\car{} self-declared class}
\begin{center}\scriptsize\input{gen_v5r/tab_car_class.tex}\end{center}
\captionof{table}{Fraction of \car{} programs whose \texttt{\# CLASS} line is consistent with the task (\texttt{poly-*} for the $\TC$, $\Lsp$ and $\GapL$ tasks, \texttt{sharpP-*} for the $\shP$ tasks).}\label{tab:car_class}
\subsection{Extra-large grids}
\label{app:xl}
\CarSummary{}

\ZoomSummary{} \TwoViewOthers{}
\begin{figure}[htb]\centering
\includegraphics[width=0.8\linewidth]{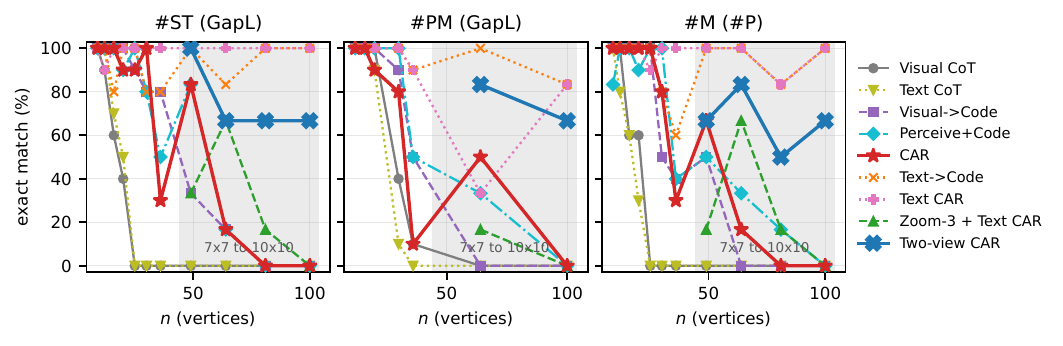}
\caption{Scaling to large grids: exact match against $n$ for Visual, Code, Perceive+Code, \car{}, Two-view \car{} and the text-input code conditions on $\sST$ and $\sAM$ ($\sPM$ at even $n$), \texttt{gemini-3.8-flash} (low). The shaded region marks the $7{\times}7$ to $10{\times}10$ extension.}
\label{fig:car_scaling}
\end{figure}
\begin{center}\scriptsize\resizebox{\linewidth}{!}{\input{gen_v8p/gen_v5r/tab_xl.tex}}\end{center}
\captionof{table}{Exact match (\%) with instance counts on the $7{\times}7$ to $10{\times}10$ grids, all configurations and conditions run on them.}\label{tab:xl}
\subsection{Where the transcription fails: orientation and continuity}
\label{app:orient}
\OrientSummary{}
\begin{center}\small\input{gen_v5r/tab_orient.tex}\end{center}
\captionof{table}{Whole-image transcriptions of the 24 extra-large grids by \texttt{gemini-3.8-flash} (low) under different readings: grids transcribed exactly, true horizontal (H) and vertical (V) edges missed, spurious edges added. \emph{Rotated} counts refer to the original orientation of the edge (an edge that is horizontal in the drawing is vertical on the rotated page). The last row is the two-view procedure with disputed-pair verification. Counts are over edges, and the edges of one grid are not independent; call- and image-level tests are in Table~\ref{tab:v4_orient}.}\label{tab:orient}
\begin{center}\includegraphics[width=0.95\linewidth]{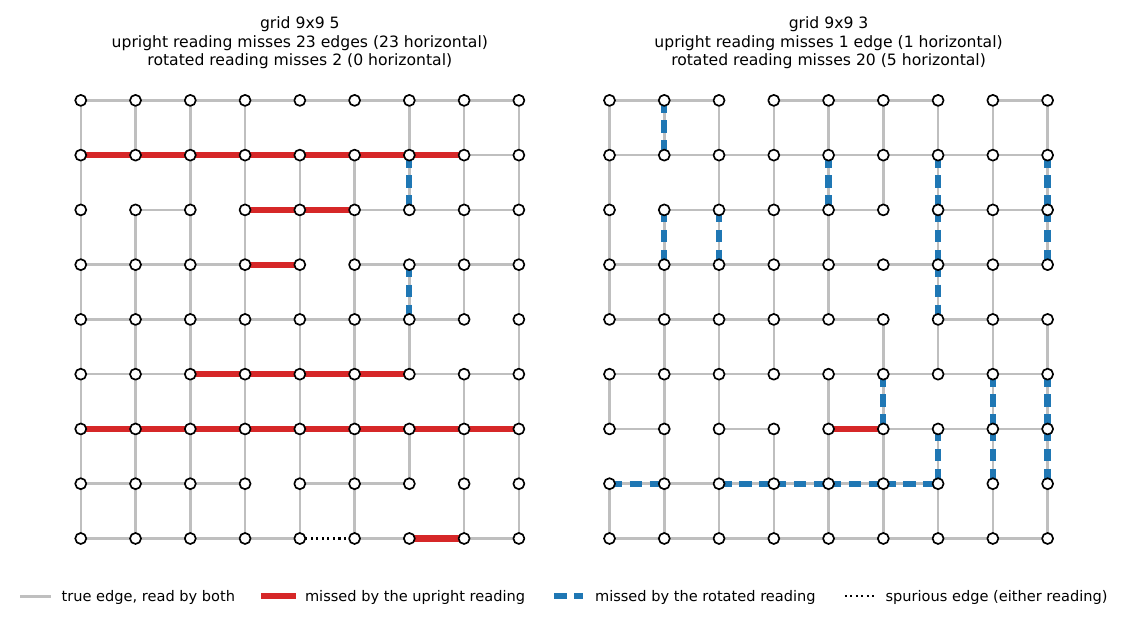}\end{center}
\captionof{figure}{The two extra-large grids on which the upright reading (left) and the rotated reading (right) of \texttt{gemini-3.8-flash} (low) miss the most edges. Missed edges come in runs along a line that continues straight through several circles, horizontal on the page for the upright reading and vertical in the drawing (hence horizontal on the rotated page) for the rotated reading; the two readings almost never miss the same edge, which is what the two-view procedure exploits.}\label{fig:orient_examples}
\begin{center}\small\input{gen_v5r/tab_runctx.tex}\end{center}
\captionof{table}{Miss rate of true edges by orientation and context, for the upright whole-image reading and inside the Zoom-2/3 crops (edges whose two circles are fully inside the crop). A segment is \emph{inside a run} when the two neighbouring segments along the same line are also present, i.e.\ the line continues straight through both of its endpoints.}\label{tab:runctx}
\begin{center}\small\resizebox{\linewidth}{!}{\input{gen_v5r/tab_twoview_models.tex}}\end{center}
\captionof{table}{The two-view pipeline on every configuration it was run on ($7{\times}7$ to $10{\times}10$ grids, 24 grids, 60 (instance, task) pairs). Grids transcribed exactly by the one-shot upright reading, by the union of the upright and rotated readings, and after verification of the disputed pairs; mean edge F1 after verification; verification calls per grid; pairs solved by one-shot \car{} (transcription embedded in the program), by Text\,\car{} and Text$\to$Code on the verified list, and by Text\,\car{} on the ground-truth edge list.}\label{tab:twoview_models}
\subsection{Algorithm choice by instance size}
\CodeSummary{} \AlgoScaleSummary{}
\begin{center}\scriptsize\input{gen_v5r/tab_algo_size.tex}\end{center}
\captionof{table}{Share (\%) of generated programs that use the polynomial algebraic method (det.: Kirchhoff or Kasteleyn determinant, for $\sST$ and $\sPM$) or a memoised/frontier dynamic program (DP, for $\sAM$ and $\sPATH$), by instance size, for \texttt{gemini-3.8-flash} (low); 6--10 programs per cell.}\label{tab:algo_size}
\subsection{Where the errors of the code conditions come from}
\CodeAttribSummary{}
\begin{center}\scriptsize\input{gen_v8p/gen_v5r/tab_attrib.tex}\end{center}
\captionof{table}{Attribution of wrong programs on grid instances (all sizes, including the extra-large grids and the held-out tasks). For every program we recover the edge list it embeds and recompute the exact answer for \emph{that} graph with our reference solvers. ``transcription'': the program's answer is correct for the graph it transcribed, which differs from the truth; ``algorithm'': the transcription is exact and the computation is wrong; ``both'': neither; ``unscorable'': the program embeds no explicit edge list (adjacency dictionaries, prose transcriptions). The last column counts correct programs whose recovered list nevertheless differs from the truth, an upper bound on extraction artefacts.}\label{tab:attrib}
\subsection{\car{} on the primary model: per size and held-out tasks}
\CarAblSummary{} \HeldoutSummary{}
\begin{center}\scriptsize
\resizebox{\linewidth}{!}{\input{gen_v8p/gen_v5r/tab_methods_g38_low.tex}}
\end{center}
\captionof{table}{\texttt{gemini-3.8-flash} (low): exact match per task and mean tokens for the direct, hinted, code and \car{} conditions on the full benchmark.}\label{tab:car}
\begin{center}\scriptsize
\resizebox{\linewidth}{!}{\input{gen_v8p/gen_v5r/tab_heldout.tex}}
\end{center}
\captionof{table}{Held-out tasks on the grid drawings (never used while designing the \car{} prompt): exact match (\%) with instance counts, the fraction of programs that use a dynamic program or polynomial routine, and mean tokens per call.}\label{tab:heldout}
\subsection{\car{} and ablations on other models}
\input{gen_v8p/gen_v5r/tab_methods_all.tex}
\subsection{Across model generations}
\GenSummary{}
\begin{figure}[htb]\centering
\includegraphics[width=0.7\linewidth]{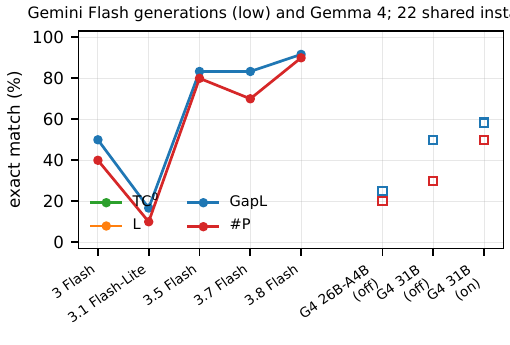}
\caption{Class-average exact match (Visual) across model generations and sizes. Each line is a complexity class.}
\label{fig:generations}
\end{figure}

\section{Generated code: classifier and examples}
\label{app:code}
\let\ClassifierNoteVOne\ClassifierNote
\renewcommand{\ClassifierNote}{\ClassifierNoteVOne{} The raw logs also store an \texttt{algo} field written by the earlier keyword classifier (v1); every share in the paper uses the validated classifier, which re-labels each program.}
\input{gen_v5r/code_examples.tex}

\section{Failure cases}
\label{app:failures}
\input{gen_v5r/failure_examples.tex}

\section{Compute, cost and reproducibility}
\label{app:compute}
\input{gen_v5r/compute.tex}

\section{Expanded Seeds and Rendering-Style Robustness}
\label{app:v3_ext}
To test sample-size stability and rendering-style invariance, we generated \vThreeSeedPairs{} additional (instance, task) pairs from fresh seeds (bringing each grid size $3{\times}4$ to $6{\times}6$ and each binary image size $6{\times}6$ to $12{\times}12$ from 10 to 25 instances per size, and each Eulerian size bucket from 8 to 16 pairs, for \vThreePooledPairs{} pooled pairs across 320 unique instances; $3{\times}3$ grids have only 6 non-isomorphic valid instances) and evaluated eight (model, condition) cells (\vThreeSeedRecords{} API calls). In addition, we re-rendered all 106 core grid and binary instances under four alternate visual styles (1\,px thin lines with smaller vertex circles, 8\,px thick lines, $90^\circ$ transposed orientation, and 60\% resolution on a gray background; \vThreeStylePairs{} styled pairs) and evaluated \texttt{gemini-3.8-flash} (low) under Visual CoT and Visual$\to$Code (\vThreeStyleRecords{} API calls, total spend of this extension \$\vThreeSpendUSD{}).

\begin{center}\scriptsize\input{gen_v3/tab_v3_dataset.tex}\end{center}
\captionof{table}{Instance and pair counts in the core slice, the seed expansion, and the pooled benchmark.}\label{tab:v3_dataset}

\begin{center}\scriptsize\resizebox{\linewidth}{!}{\input{gen_v8p/gen_v3/tab_v3_twins.tex}}\end{center}
\captionof{table}{Paired same-image twin tests comparing the 10-per-size core slice against the 25-per-size pooled benchmark ($N=\vThreePooledPairs{}$ pairs).}\label{tab:v3_twins}

\begin{center}\scriptsize\input{gen_v8p/gen_v3/tab_v3_magnitude.tex}\end{center}
\captionof{table}{Logistic regression of exact match on $\log_{10}(\text{answer})$ and complexity class ($\GapL$ vs.\ $\shP$) on the core vs.\ pooled samples.}\label{tab:v3_magnitude}

\begin{center}\scriptsize\resizebox{\linewidth}{!}{\input{gen_v3/tab_v3_styles.tex}}\end{center}
\captionof{table}{Rendering-style robustness across five visual styles on the core grid and binary instances ($n=328$ pairs per style).}\label{tab:v3_styles}

\begin{center}\scriptsize\input{gen_v3/tab_v3_styles_task.tex}\end{center}
\captionof{table}{Per-task exact match (\%) across the five rendering styles.}\label{tab:v3_styles_task}

\begin{center}\scriptsize\resizebox{\linewidth}{!}{\input{gen_v3/tab_v3_orient.tex}}\end{center}
\captionof{table}{Edge-orientation miss rates across the five rendering styles on the core grid drawings. Under the $90^\circ$ transposed style, the 3,941 originally horizontal edges drop from 56 misses to 0 ($b/c=56/0$, McNemar $p=2.8{\times}10^{-17}$), while the 3,748 originally vertical edges rise from 0 misses to 26 ($b/c=0/26$, $p=3.0{\times}10^{-8}$). These tests treat every edge as an observation although the edges of one image are read in one call; Table~\ref{tab:v4_orient} repeats them with the call and the image as units.}\label{tab:v3_orient}

\begin{center}\scriptsize\input{gen_v3/tab_v3_size.tex}\end{center}
\captionof{table}{Per-size exact match (\%) with Wilson 95\% intervals on the pooled 25-per-size benchmark.}\label{tab:v3_size}

{\scriptsize\input{gen_v8p/gen_v4/tab_v3_replication_lt.tex}}\addtocounter{table}{-1}%
\captionof{table}{Task-by-task replication check between the core slice and the fresh seed slice with two-sided Fisher exact $p$-values and pooled Wilson 95\% intervals.}\label{tab:v3_replication}

\section{Sensitivity analyses and provenance}
\label{app:v4}
\begin{figure}[htb]\centering
\includegraphics[width=0.6\linewidth]{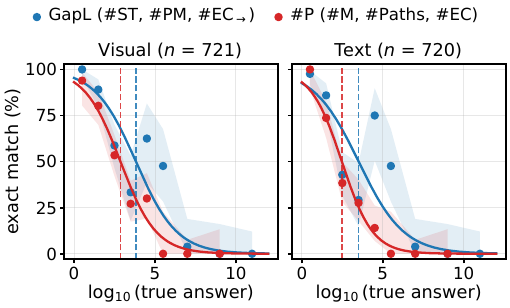}
\caption{Exact match against the magnitude of the true answer, \texttt{gemini-3.8-flash} (low), Visual (left, $n=\vFourMagPooledNoxlVisualN$) and Text (right, $n=\vFourMagPooledNoxlTextN$), grid and Eulerian families of the pooled sample, extra-large grids excluded. Points: binned rates (Wilson intervals); curves: per-class logistic fits; dashed: $m_{50}$. Panel (b) of Fig.~\ref{fig:teaser} shows the Visual half.}
\label{fig:magnitude}
\end{figure}

This appendix collects sensitivity analyses and provenance notes. Every number is recomputed from the raw result files by our analysis scripts, which also write the tables below, Fig.~\ref{fig:magnitude} and panel (b) of Fig.~\ref{fig:teaser}; apart from a seven-call check of the resolved model versions, none of them issues model calls. \emph{Core} denotes the 10-per-size slice of the benchmark, \emph{pooled} the 25-per-size benchmark that adds the fresh-seed slice of \S\ref{app:v3_ext}, and \emph{XL} the 24 extra-large grids ($7{\times}7$ to $10{\times}10$, 60 pairs). Unless stated otherwise the configuration is \texttt{gemini-3.8-flash} (low). Intervals on paired differences are percentile cluster-bootstrap intervals over images ($B=2000$), $p$-values for paired binary outcomes are exact McNemar tests, and families of tests are adjusted by Holm~\cite{holm1979simple} and by Benjamini--Hochberg~\cite{benjamini1995controlling}.

\paragraph{Call counts and provenance.} The core-slice result files hold 15,169 lines, of which 14,685 are successful records; 14,678 remain after dropping the aborted \texttt{gm26\_high} configuration, a rule our loader applies throughout. The seed-expansion and style files hold 6,874 lines, 18 of them failures, hence 6,856 records. The core-slice and seed-expansion results therefore rest on \vFourNUniqueCalls{} unique scored calls. The compute table of \S\ref{app:compute} counts the \nTotalCalls{} of these records that lie outside the three held-out tasks; the other 792 are the held-out calls. A pilot file of 281 calls served prompt development and enters no table. We keep a per-file manifest of the result files (sizes, hashes, time stamps, per-model counts). The runner did not log the \texttt{modelVersion} and \texttt{responseId} fields of the API response, so the resolved version behind each public model id during the runs (2 to 4 October 2026, PDT) cannot be recovered from the logs. Queried on 5 October 2026 (PDT) with the same request shape, the API returned a \texttt{modelVersion} equal to the requested id for \texttt{gemini-3.8-flash} (after one 503 retry), \texttt{gemini-3.6-flash} and \texttt{gemini-3.1-pro-preview}, and the model-listing endpoint reported the version strings \texttt{3.0}, \texttt{3.6-flash-07-2026} and \texttt{3.1-pro-preview-01-2026}; the runner now logs both fields.

\paragraph{Magnitude regression without the extra-large grids.} A core fit that includes the 60 extra-large pairs ($n=372$ under Visual) mixes the scaling suite into the benchmark proper, so we report both. Table~\ref{tab:v4_magnitude} repeats the logistic fit of exact match on $\log_{10}$(answer) and class with and without them, on the core and on the pooled sample, for the two other Gemini models fitted and for the code conditions of the primary model. Dropping the extra-large pairs moves the core class coefficient from $-0.56$ ($p=0.09$) to $\vFourMagCoreNoxlVisualBcls$ ($p=\vFourMagCoreNoxlVisualP$) under Visual and from $-0.65$ ($p=0.049$) to $\vFourMagCoreNoxlTextBcls$ ($p=\vFourMagCoreNoxlTextP$) under Text, and the $m_{50}$ gap from 0.5 to $\vFourMagCoreNoxlVisualMfiftyDiff$ and from 0.7 to $\vFourMagCoreNoxlTextMfiftyDiff$ decades. On the pooled sample the class term is estimated with a smaller standard error ($\beta=\vFourMagPooledNoxlVisualBcls\pm\vFourMagPooledNoxlVisualSEcls$, LR $p={}$\vFourMagPooledNoxlVisualP{} under Visual; $\vFourMagPooledNoxlTextBcls\pm\vFourMagPooledNoxlTextSEcls$, $p={}$\vFourMagPooledNoxlTextP{} under Text) and the gap is $\vFourMagPooledNoxlVisualMfiftyDiff$ decade in both conditions, with bootstrap intervals $[\vFourMagPooledNoxlVisualMfiftyDiffCIlo, \vFourMagPooledNoxlVisualMfiftyDiffCIhi]$ (Visual) and $[\vFourMagPooledNoxlTextMfiftyDiffCIlo, \vFourMagPooledNoxlTextMfiftyDiffCIhi]$ (Text) against $[\vFourMagCoreNoxlVisualMfiftyDiffCIlo, \vFourMagCoreNoxlVisualMfiftyDiffCIhi]$ and $[\vFourMagCoreNoxlTextMfiftyDiffCIlo, \vFourMagCoreNoxlTextMfiftyDiffCIhi]$ on the core. For \texttt{gemini-3.6-flash} and \texttt{gemini-3.1-pro-preview} the class term is not detectable ($p\ge0.10$), and for the code conditions both terms are small because accuracy is near the ceiling. The per-configuration $m_{50}$ values of Table~\ref{tab:m50} here and of Table~\ref{tab:budget} never included the extra-large pairs.

\begin{center}\scriptsize\resizebox{\linewidth}{!}{\input{gen_v8p/gen_v5/tab_v4_magnitude_v5.tex}}\end{center}
\captionof{table}{Logistic regression of exact match on $\log_{10}$(answer) and complexity class ($\shP$ against $\GapL$), with and without the 60 extra-large pairs, on the core and pooled samples and for other configurations and conditions. odds$/$decade: multiplicative change in the odds per decade of the answer; $\beta_{\shP}$: class coefficient with its standard error; LR $p$: likelihood-ratio test of the class term; $m_{50}$: answer magnitude at which the fitted probability crosses one half, per class; $\Delta$: their difference in decades.}\label{tab:v4_magnitude}

Fig.~\ref{fig:magnitude} plots the two pooled fits of Section~\ref{sec:res_twins}; panel (b) of Fig.~\ref{fig:teaser} repeats its Visual half.

Table~\ref{tab:v4_covariates} adds instance covariates to the pooled fit and removes one task at a time. The class term survives $n$, $m$, density and family as regressors (LR $p=1.3{\times}10^{-9}$ under Visual), but it rests on two of the six tasks: without $\sST$ or without $\sPATH$ the coefficient changes sign and loses significance ($p=0.37$ and $0.36$ under Visual; $0.049$ and $0.81$ under Text), whereas removing $\sPM$ or $\sAM$ makes it larger. Within the grid family alone it remains ($p=0.0017$ under Visual), within the Eulerian family alone it does not ($p=\vFourCovPooledNoxlVisualEulerP$). Class is nested within task, so a fit with task effects cannot estimate it at all. The pattern says that the class term depends on the joint presence of $\sST$, whose large answers anchor the magnitude slope, and $\sPATH$, whose small answers predict more success than the models achieve; it does not establish a class effect separate from the tasks that represent each class.

\begin{center}\scriptsize\input{gen_v4/tab_v4_covariates.tex}\end{center}
\captionof{table}{Pooled magnitude regression, \texttt{gemini-3.8-flash} (low): coefficients of the full model with instance covariates ($n$, $m$ and edge density standardised; family indicator), likelihood-ratio tests of the class and magnitude terms in the full model, and the class coefficient of the two-regressor model after leaving out one task or one family.}\label{tab:v4_covariates}

Table~\ref{tab:v4_discordant} looks inside the images. Among same-graph twin pairs of which exactly one member is solved, the fraction in which the solved member is the $\shP$ one is tested against one half (two-sided binomial), by the gap between the two answers. When the two answers lie within a decade of each other the split is even under Visual ($14$ of $30$ discordant pairs favour the $\shP$ member) and favours the $\GapL$ member under Text ($7$ of $31$); once the $\shP$ answer is at least a decade larger the $\GapL$ member wins almost always ($2$ of $47$ under Visual, $3$ of $51$ under Text); and when the $\shP$ answer is at least a decade \emph{smaller} the $\GapL$ member is still solved more often ($13$ of $21$ under Visual, $p=0.38$; $23$ of $33$ under Text, $p=0.035$). The first two facts are the magnitude effect seen within an image; the third is the $\sPATH$ pattern of the previous paragraph.

\begin{center}\scriptsize\input{gen_v4/tab_v4_discordant.tex}\end{center}
\captionof{table}{Discordant same-graph twin pairs (exactly one member solved) by the gap $\log_{10}(\text{answer}_{\shP})-\log_{10}(\text{answer}_{\GapL})$, pooled sample, \texttt{gemini-3.8-flash} (low). disc.: discordant pairs; $\shP$ right: pairs in which the solved member is the $\shP$ one; $p$: two-sided binomial test against one half.}\label{tab:v4_discordant}

\paragraph{Twin tests with multiple-comparison control.} Table~\ref{tab:v4_twins_pooled} lists the paired tests on the pooled sample for every configuration and condition run on it, with bootstrap intervals and Holm-adjusted $p$-values within the table. Of the \vFourTwinsMainTableVisualNTests{} tests in Table~\ref{tab:twins}, \vFourTwinsMainTableVisualSigRaw{} are significant at $0.05$ before and \vFourTwinsMainTableVisualSigHolm{} after Holm correction; of the \vFourTwinsAllCoreNTests{} core tests with at least ten pairs (Tables~\ref{tab:twins_all}, \ref{tab:twins_text} and~\ref{tab:twins_hint}), \vFourTwinsAllCoreSigRaw{}, \vFourTwinsAllCoreSigHolm{} after Holm and \vFourTwinsAllCoreSigBh{} at BH $q<0.05$; of the \vFourTwinsPooledNTests{} pooled tests, \vFourTwinsPooledSigRaw{}, \vFourTwinsPooledSigHolm{} and \vFourTwinsPooledSigBh{}; on the fresh-seed slice alone (\vFourTwinsPooledVThreeOnlyNTests{} tests) \vFourTwinsPooledVThreeOnlySigRaw{}, \vFourTwinsPooledVThreeOnlySigHolm{} and \vFourTwinsPooledVThreeOnlySigBh{}. Every comparison that survives Holm is $\sPM$ against $\sAM$ or $\sECd$ against $\sEC$. No $\sST$ comparison survives Holm in any sample; the four that pass BH point in both directions ($\sST$ behind $\sPATH$ on \texttt{gemini-3.6-flash} at minimal thinking and on \texttt{gemini-3.1-pro-preview}, ahead of $\sAM$ under Text for the primary model on the pooled sample and on its fresh-seed half).

\begin{center}\scriptsize\resizebox{\linewidth}{!}{\input{gen_v4/tab_v4_twins_pooled.tex}}\end{center}
\captionof{table}{Paired twin tests on the pooled 25-per-size sample, every configuration and condition run on it. acc$_A$/acc$_B$: exact match on the lower-class/higher-class member; $\Delta$ [95\% CI]: paired difference in points with a cluster-bootstrap interval over images; $b/c$: pairs solved only by A / only by B; $p$: exact McNemar; Holm $p$: adjusted over the \vFourTwinsPooledNTests{} tests of this table.}\label{tab:v4_twins_pooled}

\paragraph{\car{} against Code, paired.} Table~\ref{tab:v4_carcode} gives the paired comparison of \car{} with Visual$\to$Code, and of Text\,\car{} with Text$\to$Code, on the same instances for every configuration on which both were run. On the benchmark proper no Visual comparison on a Gemini model reaches significance, and the pooled estimate for the primary model is $\vFourCCPooledNoxlGThreeEightLowCarCodeDiff$ points $[\vFourCCPooledNoxlGThreeEightLowCarCodeCIlo, \vFourCCPooledNoxlGThreeEightLowCarCodeCIhi]$ over \vFourCCPooledNoxlGThreeEightLowCarCodeN{} pairs. The two significant differences go in opposite directions: Text\,\car{} is ahead of Text$\to$Code by $\vFourCCCoreNoxlGThreeEightLowTcarTcodeDiff$ points $[\vFourCCCoreNoxlGThreeEightLowTcarTcodeCIlo, \vFourCCCoreNoxlGThreeEightLowTcarTcodeCIhi]$, and \car{} is behind Code on \texttt{gemma-4-31b-it} without thinking by $\vFourCCCoreNoxlGmThreeOneMinCarCodeDiff$ points $[\vFourCCCoreNoxlGmThreeOneMinCarCodeCIlo, \vFourCCCoreNoxlGmThreeOneMinCarCodeCIhi]$. On the extra-large grids \car{} is ahead of Code by $\vFourCCXlGThreeEightLowCarCodeDiff$ points $[\vFourCCXlGThreeEightLowCarCodeCIlo, \vFourCCXlGThreeEightLowCarCodeCIhi]$ ($p=\vFourCCXlGThreeEightLowCarCodeP$) for the primary model; Perceive+Code, which has the transcription step but not the routing sentences, reaches the same 14/60, so the step rather than the routing is the plausible cause.

\begin{center}\scriptsize\resizebox{\linewidth}{!}{\input{gen_v4/tab_v4_carcode.tex}}\end{center}
\captionof{table}{Paired comparison of \car{} with the plain code condition on the same instances (acc$_A$: \car{}; acc$_B$: Code), with cluster-bootstrap intervals over images, discordant counts $b/c$ (solved only by \car{} / only by Code) and exact McNemar $p$-values.}\label{tab:v4_carcode}

\paragraph{Transcription against computation, re-solved offline.} For every program-writing call on a grid instance we extract the edge list the program embeds (the syntactic extractor of \S\ref{app:code}), compare it with the drawn graph $G$, and re-solve the task on the embedded graph $\hat G$ with the reference solver. Table~\ref{tab:v4_transcription} cross-tabulates transcription exactness with answer correctness on the pooled, core and extra-large grids; for Visual chain of thought the transcription is the one the model writes before counting, scored as in Table~\ref{tab:attribution}, and the corresponding pooled counts are \vFourDecVisPooledNoxlTOneAOne{} exact-and-right, \vFourDecVisPooledNoxlTOneAZero{} exact-and-wrong, \vFourDecVisPooledNoxlTZeroAOne{} inexact-and-right and \vFourDecVisPooledNoxlTZeroAZero{} inexact-and-wrong over \vFourDecVisPooledNoxlScorable{} scorable calls. Three facts follow. A \car{} program whose embedded graph matches the drawing is right in \vFourDecPooledNoxlCarPAOneGivenTOne\% of pooled scorable calls; when the embedded graph is wrong, the printed answer is exactly the reference answer for $\hat G$ in \vFourDecPooledNoxlCarTwrongExecMatchesGhat{} of the \vFourDecPooledNoxlCarTwrongCountAvailable{} cases in which the comparison is possible, and in the others the extracted list is not the graph the program used or the algorithm is wrong as well, which the extractor cannot tell apart. On the extra-large grids all \vFourDecXlCarTZeroAZero{} wrong \car{} programs embed a wrong graph, \vFourDecXlCarTwrongExecMatchesGhat{} of them printing the right answer for it, and none of the \vFourDecXlCarTOneAOne{} exact transcriptions is followed by a wrong answer. For Visual chain of thought the split is the opposite: \vFourDecVisPooledNoxlShareErrorsWithTOne\% of the pooled errors follow an exact transcription and $P(\text{right}\mid\hat G=G)$ is \vFourDecVisPooledNoxlPAOneGivenTOne\%. The decomposition has limits. The extractor leaves \vFourDecPooledNoxlCarUnscorable{} of \vFourDecPooledNoxlCarRecords{} pooled \car{} programs and \vFourDecPooledNoxlCodeUnscorable{} of \vFourDecPooledNoxlCodeRecords{} Code programs unscored (no literal edge list, for instance an adjacency built in a loop), and it flags \vFourDecPooledNoxlCarTZeroAOne{} \car{} programs as mismatched although they print the right answer, so mismatch rates are upper bounds. The transcription is produced inside a task-specific call rather than once per image; in the \vFourDecPooledNoxlCarImagesWithGeTwoScorable{} pooled images with at least two scorable \car{} transcriptions they are identical across tasks in \vFourDecPooledNoxlCarImagesAllTranscriptionsIdentical{} and differ in the others. Experiment~A (next paragraph) collected one task-independent transcription per image; the syntactic extractor exists only for the grid drawings, not for the binary images or the Eulerian drawings.

\begin{center}\scriptsize\resizebox{\linewidth}{!}{\input{gen_v4/tab_v4_transcription.tex}}\end{center}
\captionof{table}{Transcription exactness against answer correctness for the program-writing conditions on grid instances, \texttt{gemini-3.8-flash} (low). records: calls; no prog.: calls without a program; unscor.: programs without an extractable edge list; the four middle columns split the scorable calls by whether the embedded graph $\hat G$ equals the drawn graph $G$ and whether the answer is right; the two conditional probabilities follow; exec$=\hat G$: among the calls with $\hat G\neq G$ and a comparable answer, those whose printed answer equals the reference answer for $\hat G$.}\label{tab:v4_transcription}

\paragraph{Orientation, tested on calls and images.} The edge-level counts of Tables~\ref{tab:orient} and~\ref{tab:v3_orient} treat every edge as an observation, but the edges of one grid are read in one call and are not independent. Table~\ref{tab:v4_orient} repeats the comparison with the call and the image as units, pairing the upright reading with the transposed reading of the same image in the five-style replication of \S\ref{app:v3_ext} (\vFourOrientCalls{} paired calls on \vFourOrientImages{} images) and, on the extra-large grids, the one upright and the one rotated whole-image reading of each grid. The direction of the effect is the same at every level: originally horizontal edges are missed only by the upright reading (\vFourOrientMissHCallsUp{} calls on \vFourOrientMissHImgUp{} images against none; $p={}$\vFourOrientMissHPcalls{} over calls, \vFourOrientMissHPimages{} over images) and originally vertical edges only by the transposed reading (\vFourOrientMissVCallsRot{} calls on \vFourOrientMissVImgRot{} images; $p={}$\vFourOrientMissVPcalls{} and \vFourOrientMissVPimages). Overall exactness does not differ between the two readings on the core grids (\vFourOrientAnyCallsUp{} against \vFourOrientAnyCallsRot{} inexact calls, $p=\vFourOrientAnyPcalls$), and on the extra-large grids the rotated reading is worse (an edge missed on \vFourOrientXLMissingGridsRot{} of 24 grids against \vFourOrientXLMissingGridsUp{}, $p={}$\vFourOrientXLMissingP; exact on \vFourOrientXLExactRot{} grids against \vFourOrientXLExactUp). The second view helps because its misses are independent of the first view's, not because the rotated page is read better.

\begin{center}\scriptsize\resizebox{\linewidth}{!}{\input{gen_v4/tab_v4_orient.tex}}\end{center}
\captionof{table}{Orientation of missed edges with the call, the image or the grid as the unit. Core grids: upright against transposed rendering of the same image, \texttt{gemini-3.8-flash} (low), five-style replication; XL grids: one upright and one rotated whole-image reading per grid. $b/c$: units with the event under the upright reading only / under the transposed reading only; $p$: exact McNemar.}\label{tab:v4_orient}

\paragraph{Budget slope.} Across the \vFourBudgetK{} configurations of \vFourBudgetModels{} models for which Table~\ref{tab:m50} reports an $m_{50}$, a least-squares fit of $\log_{10} m_{50}$ on $\log_{10}$ of the mean thinking plus output tokens per call has slope \vFourBudgetSlope{} (SE \vFourBudgetSe; 95\% CI $[\vFourBudgetCIlo, \vFourBudgetCIhi]$; $p={}$\vFourBudgetP; $R^2=\vFourBudgetRTwo$); leaving out one configuration at a time keeps the slope between \vFourBudgetLOOlo{} and \vFourBudgetLOOhi, and within a family it is \vFourBudgetFamGThreeEight{} for \texttt{gemini-3.8-flash} and \vFourBudgetFamGThreeSix{} for \texttt{gemini-3.6-flash} (three thinking levels each). The slope is consistent with a procedure whose cost grows with the answer; it does not identify the procedure, and the points are not independent draws, several coming from the same family.

\paragraph{Extra-large grids: full tables.} Tables~\ref{tab:zoom} and~\ref{tab:xl_summary} give the full extra-large-grid results that the main paper summarises.

\begin{center}\scriptsize\resizebox{\linewidth}{!}{\input{gen_v8p/gen_v5r/tab_zoom.tex}}\end{center}
\captionof{table}{Transcription at scale on the $7{\times}7$ to $10{\times}10$ grids, \texttt{gemini-3.8-flash} (low). Left: grids transcribed exactly (count per size), mean edge F1 and transcription tokens; right: (instance, task) pairs solved when the transcribed list is handed to Text\,\car{} or Text$\to$Code. First rows: whole-image code conditions (transcription embedded in the program); last rows: ground-truth text input as the ceiling. Zoom-$k$ and Two-view are defined in \S\ref{app:codecond}.}\label{tab:zoom}

\begin{center}\scriptsize
\setlength{\tabcolsep}{4pt}
\begin{tabular}{lrrrrrrrr}
\toprule
 & \multicolumn{5}{c}{\texttt{gemini-3.8-flash} (low)} & 3.6-flash & Gemma & 3.1-pro \\
\cmidrule(lr){2-6}\cmidrule(lr){7-7}\cmidrule(lr){8-8}\cmidrule(lr){9-9}
Condition & $7{\times}7$ & $8{\times}8$ & $9{\times}9$ & $10{\times}10$ & all & (low) & 31B (off) & (low) \\
\midrule
Visual CoT & 0/12 & 0/18 & 0/12 & 0/18 & 0/60 & -- & 0/60 & 0/59 \\
Text CoT & 0/12 & 0/18 & 0/12 & 0/18 & 0/60 & -- & -- & -- \\
Code & 5/12 & 1/18 & 0/12 & 0/18 & 6/60 & 9/60 & 0/60 & 2/59 \\
Perceive+Code & 8/12 & 5/18 & 1/12 & 0/18 & 14/60 & -- & -- & -- \\
\car{} & 9/12 & 5/18 & 0/12 & 0/18 & 14/60 & 15/60 & 0/60 & 2/59 \\
Two-view + Text\,\car{} & 10/12 & 14/18 & 7/12 & 12/18 & 43/60 & 40/60 & -- & 36/60 \\
Two-view + Text$\to$Code & 11/12 & 15/18 & 11/12 & 12/18 & 49/60 & 39/60 & -- & 35/60 \\
\midrule
Text$\to$Code (GT list) & 12/12 & 17/18 & 11/12 & 17/18 & 57/60 & 53/60 & -- & -- \\
Text\,\car{} (GT list) & 12/12 & 14/18 & 11/12 & 17/18 & 54/60 & 51/60 & 51/60 & 59/59 \\
\bottomrule
\end{tabular}
\end{center}
\captionof{table}{Extra-large grids ($7{\times}7$ to $10{\times}10$; 24 grids, 60 (instance, task) pairs, answers $10^{5}$ to $10^{37}$): pairs solved exactly, by size for the primary configuration and in total for the others. CoT: chain of thought; GT list: ground-truth edge list as text input; --: not run; in the \texttt{3.1-pro} column the one-shot and GT-list rows use the 59 pairs on which all of them returned (one Visual call got HTTP 503 on every retry), and the two-view rows all 60. Per-task accuracies with instance counts in Table~\ref{tab:xl}.}\label{tab:xl_summary}

\paragraph{Task-independent shared transcription (Experiment A).} To separate reading from counting without the syntactic extractor's gaps, we ran one task-independent transcription call per unique pooled image (\vFourOneAImages{} images: 156 grid drawings, 100 binary images, 64 directed and 64 undirected Eulerian drawings) on \texttt{gemini-3.8-flash} (low) with no task mentioned in the prompt, then solved all \vFourOneAPairs{} pooled (instance, task) pairs from the transcribed structure $\hat G$ both with the reference solver ($f(\hat G)$) and with the model under the Text-condition prompt ($\text{Transcribe}\to\text{Text}$; analysis plan fixed before the run). Every API call of Experiments A and B logs \texttt{modelVersion} (\vFourOneModelVersions{}) and \texttt{responseId}. Table~\ref{tab:v41_transcription} gives the image-level transcription quality: \vFourOneAImagesExact{} of \vFourOneAImagesParsed{} parsed images (\vFourOneAImagesExactPct\%) are transcribed with zero structural error ($\hat G=G$), missing \vFourOneAImgAllMissing{} of \vFourOneAImgAllUnits{} edges or black cells overall. Table~\ref{tab:v41_decomposition} decomposes every pair into structural perception error ($S=\mathbb I[\hat G\neq G]$), answer-relevant perception error ($P=\mathbb I[f(\hat G)\neq f(G)]$), and downstream computation error ($C=\mathbb I[\hat y\neq f(\hat G)]$). Across the \vFourOneAPairsOk{} evaluated pairs, the shared transcription is structurally exact on \vFourOneAStructExactPairs{} pairs (\vFourOneAStructExactPct\%) and answer-exact ($f(\hat G)=f(G)$) on \vFourOneAAnswerExactPairs{} (\vFourOneAAnswerExactPct\%); \vFourOneAPOnly{} errors come from perception alone ($P=1,C=0$), \vFourOneACOnly{} from computation alone ($P=0,C=1$), \vFourOneABoth{} from both ($P=1,C=1$), and \vFourOneACancel{} exhibit error cancellation ($P=1$ yet $\hat y=f(G)$). Conditional on an answer-exact transcription ($P=0$), the model's downstream computation error rate is \vFourOneACGivenPzeroPct\% overall (\vFourOneAPolyCGivenPzeroPct\% on polynomial tasks vs.\ \vFourOneAHardCGivenPzeroPct\% on $\shP$-complete tasks). Table~\ref{tab:v41_paired} compares the two-call $\text{Transcribe}\to\text{Text}$ pipeline against the single-call Visual and Text conditions on the same instances.

\begin{center}\scriptsize\resizebox{\linewidth}{!}{\input{gen_v4/tab_v41_transcription.tex}}\end{center}
\captionof{table}{Task-independent transcription quality per unique image across the pooled benchmark (Experiment~A, \texttt{gemini-3.8-flash} low). Exact pairs compare the shared image transcription against the per-task Visual-condition transcription on the same pairs.}\label{tab:v41_transcription}

\begin{center}\scriptsize\resizebox{\linewidth}{!}{\input{gen_v4/tab_v41_decomposition.tex}}\end{center}
\captionof{table}{Decomposition of every pooled (instance, task) pair under the shared task-independent transcription $\hat G$ (Experiment~A, \texttt{gemini-3.8-flash} low). $\hat G{=}G$: structurally exact transcription; $f(\hat G){=}f(G)$: reference solver gets the true answer on $\hat G$; perc.\ only: $P{=}1, C{=}0$; comp.\ only: $P{=}0, C{=}1$; both: $P{=}1, C{=}1$; cancel: $P{=}1$ and $\hat y{=}f(G)$.}\label{tab:v41_decomposition}

\begin{center}\scriptsize\resizebox{\linewidth}{!}{\input{gen_v8p/gen_v4/tab_v41_paired.tex}}\end{center}
\captionof{table}{Paired comparisons between the two-call $\text{Transcribe}\to\text{Text}$ pipeline, single-call Visual CoT, and single-call Text CoT on the pooled benchmark, with cluster-bootstrap 95\% intervals over images and exact McNemar $p$-values.}\label{tab:v41_paired}

\paragraph{Equal-budget upright-twice controls on the extra-large grids (Experiment B, first run).} To test whether the two-view gain on the 24 extra-large grids ($7{\times}7$ to $10{\times}10$, 60 pairs) comes from taking two readings or specifically from rotating the second view by $90^\circ$, we ran two equal-structure controls that replace the upright-plus-rotated pair with two upright readings of each grid, followed by the identical disputed-edge crop verifier and the identical Text\,\car{} and Text$\to$Code solvers (analysis plan fixed before the run): (i)~\texttt{upright$\times$2} with independent sampling ($T=1.0$), and (ii)~\texttt{upright$\times$2} with repeated greedy decoding ($T=0$). Table~\ref{tab:v41_xl} and Table~\ref{tab:v41_xl_paired} report the results. The two upright readings miss the same horizontal edges in both passes (\vFourOneBBoneBothMissed{} edges missed by both readings at $T=1.0$ and \vFourOneBBthreeBothMissed{} at $T=0$, against \vFourOneBTVBothMissed{} for upright plus rotated) and flag far fewer disputed edges for the crop verifier (\vFourOneBBoneDisputedPerGrid{} and \vFourOneBBthreeDisputedPerGrid{} per grid vs.\ \vFourOneBTVDisputedPerGrid{}). After verification, \texttt{upright$\times$2} is exact on \vFourOneBBoneExact/\vFourOneBBoneGrids{} grids at $T=1.0$ and \vFourOneBBthreeExact/\vFourOneBBthreeGrids{} at $T=0$, against \vFourOneBTVExact/\vFourOneBTVGrids{} for upright plus rotated ($\Delta=\vFourOneBBoneExactDiff$ points, \vFourOneBBoneExactCI, $p={}$\vFourOneBBoneExactP{} for $T=1.0$; $\Delta=\vFourOneBBthreeExactDiff$ points, \vFourOneBBthreeExactCI, $p={}$\vFourOneBBthreeExactP{} for $T=0$), solving \vFourOneBBoneVcar/\vFourOneBBoneVcarN{} and \vFourOneBBthreeVcar/\vFourOneBBthreeVcarN{} pairs under Text\,\car{} (\vFourOneBBoneVcode/\vFourOneBBoneVcodeN{} and \vFourOneBBthreeVcode/\vFourOneBBthreeVcodeN{} under Text$\to$Code) against \vFourOneBTVVcar/\vFourOneBTVVcarN{} and \vFourOneBTVVcode/\vFourOneBTVVcodeN{} for two-view. This first run compared calls made on different days (two-view on 2026-10-03, controls on 2026-10-06, UTC), which confounds the comparison with the day; \S\ref{app:v5} reruns all three arms in one session, and the main paper reports that rerun. Table~\ref{tab:v41_cost} summarises the call, token and dollar accounting for Experiment A and the first run of Experiment B (\$\vFourOneCostTotalUsd{} across \vFourOneCostTotalCalls{} calls).

\begin{center}\scriptsize\resizebox{\linewidth}{!}{\input{gen_v4/tab_v41_xl.tex}}\end{center}
\captionof{table}{Two-view (upright + rotated $90^\circ$) against equal-structure \texttt{upright$\times$2} controls ($T=1.0$ and $T=0$) on the 24 extra-large grids ($7{\times}7$ to $10{\times}10$, 60 pairs), \texttt{gemini-3.8-flash} (low).}\label{tab:v41_xl}

\begin{center}\scriptsize\resizebox{\linewidth}{!}{\input{gen_v8p/gen_v4/tab_v41_xl_paired.tex}}\end{center}
\captionof{table}{Paired differences between each \texttt{upright$\times$2} control and the upright-plus-rotated two-view pipeline on the extra-large grids, with cluster-bootstrap 95\% intervals over grids and exact McNemar $p$-values.}\label{tab:v41_xl_paired}

\begin{center}\scriptsize\input{gen_v4/tab_v41_cost.tex}\end{center}
\captionof{table}{API call, token and list-price cost summary for Experiment~A and the first run of Experiment~B (\texttt{gemini-3.8-flash} low).}\label{tab:v41_cost}

\paragraph{Twin tests of Table~\ref{tab:twins}.} Table~\ref{tab:v4_twins_main} lists the \vFourTwinsMainTableVisualNTests{} paired tests that Table~\ref{tab:twins} summarises, with their intervals and raw and Holm-adjusted $p$-values.

\begin{keepblock}
\begin{center}\scriptsize\resizebox{\linewidth}{!}{\input{gen_v4/tab_v4_twins_main.tex}}\end{center}
\captionof{table}{Paired same-image twin tests of Table~\ref{tab:twins} (core slice, Visual, five configurations). acc$_A$/acc$_B$: exact match on the lower-class/higher-class member; $\Delta$ [95\% CI]: paired difference in points with a cluster-bootstrap interval over images; $b/c$: pairs solved only by A / only by B; $p$: exact McNemar; Holm $p$: adjusted over the \vFourTwinsMainTableVisualNTests{} tests of this table.}\label{tab:v4_twins_main}
\end{keepblock}

\paragraph{Remaining scope limits.} Evaluating a model family from another provider, running the full set of conditions on more than 24 extra-large grids for every configuration (\S\ref{app:v6} extends the two-view comparison to fresh grids, and \S\ref{app:x8} the one-shot conditions for four configurations), and isolating stroke width from vertex-circle radius and font size in the thin-line rendering remain for future work.

\section{Same-session rerun of Experiment B}
\label{app:v5}

\paragraph{Why a rerun.} The first run of Experiment B (\S\ref{app:v4}) compared arms whose calls were made on different days: every two-view call on 2026-10-03 (01:32 to 02:22 UTC) and every control call on 2026-10-06 (06:18 to 08:31 UTC). Identical requests do not give identical outcomes across such a gap, and the changes have a direction. Among the pooled pairs whose task-independent transcription reproduces the input exactly (for graphs, in the generator's edge order), the Text prompt sent in Experiment~A on 2026-10-06 is byte-identical to the one sent for the Text condition between 2026-10-02 and 2026-10-04, both at temperature~0. Exactness differs on 98 of these 797 pairs (12.3\%): 35 are right only on the later date and 63 only on the earlier one (exact McNemar $p=0.006$). The flips come from grid pairs (78 of 534; 25 vs.\ 53, $p=0.002$) and Eulerian pairs (20 of 63; 10 vs.\ 10); binary images never flip (0 of 200). A comparison of arms run on different days is therefore confounded with the day.

\paragraph{Design.} We reran all three arms of Experiment B on the same 24 grids and 60 pairs, interleaved in one session, with the analysis plan fixed before the first call. Every arm uses fresh calls; nothing is reused from earlier logs. The two-view arm reads the upright image with the row-wise prompt and the image rotated by $90^\circ$, both at $T=0$; B1 reads the upright image twice with the same row-wise prompt at $T=1$, and B3 twice at $T=0$, so the upright reading of the two-view arm and both B3 readings are byte-identical requests. Reads are queued grid by grid in a random order, with the six reading calls of a grid in a random order; the verification and solve jobs of the three arms are shuffled and merged round-robin; failed calls (mostly HTTP 503) are retried in later passes. The crop verifier, the merge rule and the Text\,\car{} and Text$\to$Code solvers are those of the first run. A fourth set of calls sends the two solve prompts built from the ground-truth edge list, twice; these are the prompts any arm sends when its transcription is exact, so they give the solver ceiling and the same-session test-retest rate of byte-identical requests.

\paragraph{Results.} The calls ran from \vFiveSpanUtc{} (\vFiveSpanHours{} hours, \vFiveCallsTotal{} responses, every one reporting the model version \texttt{\vFiveModelVersions}; \$\vFiveUsdTotal{} at list prices). The only deviation from the plan was a higher spending stop, set before the verification phase began. Table~\ref{tab:v5_xl} summarises each arm, Table~\ref{tab:v5_xl_paired} gives the paired tests against two-view with 95\% intervals from a bootstrap over the 24 grids, Table~\ref{tab:v5_xl_orient} the orientation of the missed edges, and Table~\ref{tab:v5_retest} the repeated ground-truth calls. After verification two-view is exact on \vFiveBTVExact/24 grids, against \vFiveBBoneExact/24 for upright twice at $T=1$ and \vFiveBBthreeExact/24 at $T=0$; it solves \vFiveBTVVcar/60 pairs with Text\,\car{} and \vFiveBTVVcode/60 with Text$\to$Code, against \vFiveBBoneVcar/60 and \vFiveBBoneVcode/60 at $T=1$ and \vFiveBBthreeVcar/60 and \vFiveBBthreeVcode/60 at $T=0$. Against $T=1$ all three intervals exclude zero; against $T=0$ none does, although the McNemar test, which treats the 60 pairs of 24 grids as independent, gives $p={}$\vFiveBBthreeVcarP{} for Text\,\car{}. The verifier answers every disputed pair correctly in every arm (216/216, 35/35 and 52/52), so the arms differ in what reaches it: the two readings of two-view disagree on 9.0 lattice pairs per grid and both miss \vFiveBTVBothMissed{} true edges, whereas the repeated upright readings disagree on 1.5 and 2.2 pairs per grid and both miss \vFiveBBoneBothMissed{} ($T=1$) and \vFiveBBthreeBothMissed{} ($T=0$, all horizontal) true edges, which the verifier never sees. Relative to its own first reading, verification adds 11 exact grids for two-view ($p={}$\vFiveBTVFinalVsROneP) and 4 for each upright-twice control ($p={}$\vFiveBBoneFinalVsROneP). On these 24 development grids and without a $180^\circ$ or mirrored second reading, the rerun shows that a rotated view catches misses which a repeated upright reading shares; it does not show that orthogonality itself is required. \S\ref{app:v6} adds that control on fresh grids.

\paragraph{Test-retest.} Sending each ground-truth solve prompt twice in this session changes exactness on \vFiveRTVcarFlips{} of 60 Text\,\car{} pairs (\vFiveRTVcarB{} vs.\ \vFiveRTVcarC, $p={}$\vFiveRTVcarP) and on \vFiveRTVcodeFlips{} of 60 Text$\to$Code pairs (\vFiveRTVcodeB{} vs.\ \vFiveRTVcodeC, $p={}$\vFiveRTVcodeP), \vFiveRTFlipsTotal{} of \vFiveRTPairsTotal{} repetitions in all (\vFiveRTFlipPct\%). The three byte-identical upright readings at $T=0$ (the first reading of two-view and both B3 readings) never return identical text and give the same cleaned edge set on \vFiveRAThreeIdentical{} of 24 grids, disagreeing on \vFiveRAThreeDisagreeMean{} lattice pairs per grid on average. Differences of a few pairs between two single runs, such as the two runs of the two-view pipeline (\VerifyNumbers{} and 49/60 pairs in the first run, \vFiveBTVVcar/60 and \vFiveBTVVcode/60 here, with Text\,\car{} and Text$\to$Code), are of the size of this run-to-run variation.

\begin{center}\scriptsize\resizebox{\linewidth}{!}{\input{gen_v5/tab_v5_xl.tex}}\end{center}
\captionof{table}{Same-session rerun of Experiment B on the 24 extra-large grids (60 pairs), \texttt{gemini-3.8-flash} (low), all arms interleaved. reads exact: single readings transcribed exactly (reading 1/reading 2); Jaccard: mean similarity of the two cleaned readings; disp./grid: disputed lattice pairs per grid; both miss: true edges missed by both readings; agree wrong: non-edges reported by both; verifier: correct crop answers; exact: grids transcribed exactly after verification; miss/extra: final missing and extra edges; calls and list-price USD per arm.}\label{tab:v5_xl}

\begin{center}\scriptsize\resizebox{\linewidth}{!}{\input{gen_v8p/gen_v5/tab_v5_xl_paired.tex}}\end{center}
\captionof{table}{Same-session rerun: paired differences (control minus two-view, in points) with exact McNemar $p$-values and 95\% bootstrap intervals clustered by grid ($B=2000$).}\label{tab:v5_xl_paired}

\begin{center}\scriptsize\resizebox{\linewidth}{!}{\input{gen_v5/tab_v5_xl_orient.tex}}\end{center}
\captionof{table}{Same-session rerun: missed and extra edges by their orientation in the upright image (H horizontal, V vertical). Reading 2 of two-view is the image rotated by $90^\circ$.}\label{tab:v5_xl_orient}

\begin{center}\scriptsize\input{gen_v5/tab_v5_retest.tex}\end{center}
\captionof{table}{Same-session test-retest: each solve prompt built from the ground-truth edge list, sent twice as byte-identical requests at temperature~0. flips $b/c$: pairs exact only in repetition 1 / only in repetition 2.}\label{tab:v5_retest}

\section{Two-view reading on fresh grids}
\label{app:v6}

\paragraph{Why.} Experiment B and its rerun (\S\ref{app:v5}) use the 24 extra-large development grids, on which the two-view pipeline was built, and their only controls read the upright image twice. They cannot tell whether the gain needs the second view turned by $90^\circ$, so that the edges horizontal in one picture are vertical in the other, or only a second view rendered differently from the first. We therefore ran two experiments on grids generated after the pipeline was fixed. B6 made the orientation question primary: upright plus $90^\circ$ against upright plus $180^\circ$. B7, four times larger, made the comparison behind the main-text claim primary: each rotated second view against a repeated upright reading. For both, the plan, the runner and the grid file were hash-recorded before the first call and the analysis script before any result record was read (the logs then held 23 and 75 records, unread); both scripts were run unchanged, and both experiments are reported in full.

\paragraph{Freeze times.} Each plan was hash-frozen before its first call. Each analysis script was frozen before any record content was read, in most experiments shortly after the batch had started, when the logs held the counts in Table~\ref{tab:prereg_timing} (only line counts and progress counters were seen, as the freeze files record; R9 was frozen before its first call).

\begin{keepblock}
\begin{center}\footnotesize\setlength{\tabcolsep}{3pt}\input{gen_v8/tab_prereg_timing.tex}\end{center}
\captionof{table}{Freeze and call times of the pre-registered experiments: B6 and B7 (this appendix), X8 (\S\ref{app:x8}), D8 (\S\ref{app:d8}), T6 (\S\ref{app:t6}), C503 and F6 (\S\ref{app:rerun}), G9 (\S\ref{app:g9}), QW (\S\ref{app:qw}) and R9 (\S\ref{app:r9}). Plan and analysis freeze times come from the freeze files, first and last call from the result logs; records at freeze: the count stated in the freeze file where it gives one, otherwise the records written by then.}\label{tab:prereg_timing}
\end{keepblock}

\paragraph{Grids, arms and analysis.} The fresh grids come from a copy of the development-grid generator with the same acceptance rules, renderer and doubly computed ground truth, and new seeds; an edge set identical to an earlier grid is rejected (none was). B6 has \vFiveBSixTVNGrids{} grids, 6 per size from $7{\times}7$ to $10{\times}10$ (\vFiveBSixTVVcarN{} pairs, the layout of the development set), and B7 has \vSixBSevenGrids, 24 per size (\vSixBSevenPairs{} pairs: $\sST$ and $\sAM$ on every grid, $\sPM$ on the grids with an even number of vertices). Every call goes to \texttt{gemini-3.8-flash} at the low thinking level. The arms are two-view as before (upright, then rotated $90^\circ$ clockwise with the rotated-image prompt); upright plus $180^\circ$, whose second picture shows the labels upside down while every segment keeps its orientation, with a prompt that differs from the $90^\circ$ one only in the sentence describing the rotation; and the upright reading twice at $T=0$ (B3 of \S\ref{app:v5}). B6 also repeats the $T=1$ control (B1); every other reading is at $T=0$, so the four upright $T=0$ readings of a grid are byte-identical requests. The crop verifier, the merge rule and the two solvers are those of \S\ref{app:v5}, and the two solve prompts are also sent once on the true edge list (the ceiling). Reads are queued grid by grid in random order, with each grid's reading calls in random order; the verification and solve jobs of the arms are shuffled and merged round-robin; failed calls are retried in later passes. The unit is the grid for transcriptions and the pair for solves; tests are exact two-sided McNemar tests, and intervals are percentile bootstraps over grids. In B6 the primary test is $90^\circ$ against $180^\circ$ on final exact transcriptions, with Holm over the four transcription-level comparisons. In B7 the two co-primary tests are $90^\circ$ against the repeat and $180^\circ$ against the repeat on final exact transcriptions, with Holm over the two at family-wise level $0.05$; $90^\circ$ against $180^\circ$ and the six solve-level comparisons (Holm over the six) are secondary. Counts pooled over B6 and B7 are descriptive only, since B6 had been seen when B7 was planned.

\paragraph{Results of B7.} The calls ran from \vSixBSevenSpanUtc{} (\vSixBSevenSpanHours{} hours, \vSixBSevenCallsTotal{} responses, every one reporting the model version \texttt{\vSixBSevenModelVersions}; \$\vSixBSevenUsdTotal{} at list prices), and every planned call returned. The only deviation was a correction, made before the analysis was frozen, to the order in which the plan text wrote B6's discordant counts; no rule changed. After verification the $90^\circ$ and $180^\circ$ arms are each exact on \vSixBSevenTVExact{} of \vSixBSevenGrids{} grids and the repeat on \vSixBSevenBthreeExact{} (Table~\ref{tab:v6_b7_xl}). Both co-primary tests are significant (Table~\ref{tab:v6_b7_paired}): for $90^\circ$ against the repeat $b/c=\vSixBSevenTVVsBthreeExactB/\vSixBSevenTVVsBthreeExactC$, a difference of \vSixBSevenTVVsBthreeExactDiff{} points \vSixBSevenTVVsBthreeExactCI{} and Holm $p={}$\vSixBSevenTVVsBthreeExactPHolm; for $180^\circ$ against the repeat $b/c=\vSixBSevenRotVsBthreeExactB/\vSixBSevenRotVsBthreeExactC$, \vSixBSevenRotVsBthreeExactDiff{} points \vSixBSevenRotVsBthreeExactCI{} and Holm $p={}$\vSixBSevenRotVsBthreeExactPHolm. The two rotations tie ($b/c=\vSixBSevenTVVsRotExactB/\vSixBSevenTVVsRotExactC$, $p={}$\vSixBSevenTVVsRotExactP, interval \vSixBSevenTVVsRotExactCI). The solves keep this order: Text\,\car{} solves \vSixBSevenTVVcar, \vSixBSevenRotVcar{} and \vSixBSevenBthreeVcar{} of \vSixBSevenPairs{} pairs and Text$\to$Code \vSixBSevenTVVcode, \vSixBSevenRotVcode{} and \vSixBSevenBthreeVcode, against a ceiling of \vSixBSevenGTVcar{} and \vSixBSevenGTVcode; both rotated arms beat the repeat with either solver (Holm $p<10^{-4}$) and do not differ from each other (Holm $p={}$\vSixBSevenTVVsRotVcarPHolm{} and \vSixBSevenTVVsRotVcodePHolm). No arm solves a pair from an inexact transcription; on the pairs whose transcription is exact, Text\,\car{} solves \vSixBSevenTVVcarGivenExact/\vSixBSevenTVVcarGivenExactN, \vSixBSevenRotVcarGivenExact/\vSixBSevenRotVcarGivenExactN{} and \vSixBSevenBthreeVcarGivenExact/\vSixBSevenBthreeVcarGivenExactN.

The verifier answers every disputed pair correctly in every arm (\vSixBSevenTVNVerifyCorrect/\vSixBSevenTVNChecked, \vSixBSevenRotNVerifyCorrect/\vSixBSevenRotNChecked{} and \vSixBSevenBthreeNVerifyCorrect/\vSixBSevenBthreeNChecked), so, as in \S\ref{app:v5}, the arms differ in what reaches it. On their own the rotated readings are worse than a second upright one (exact on \vSixBSevenTVRTwoExact{} and \vSixBSevenRotRTwoExact{} grids against \vSixBSevenBthreeRTwoExact), but they disagree with the upright reading on \vSixBSevenTVDisputedPerGrid{} and \vSixBSevenRotDisputedPerGrid{} lattice pairs per grid against \vSixBSevenBthreeDisputedPerGrid{} for the repeat, and the true edges missed by both readings, which the verifier never sees, number \vSixBSevenTVBothMissed{} and \vSixBSevenRotBothMissed{} against \vSixBSevenBthreeBothMissed. Single readings miss the edges that are horizontal in their own picture (Table~\ref{tab:v6_b7_orient}): upright readings miss \vSixBSevenORUpzeroMissHRatePct\% of the horizontal and \vSixBSevenORUpzeroMissVRatePct\% of the vertical edges, $90^\circ$ readings \vSixBSevenORRotninetyMissVRatePct\% of the vertical edges (horizontal in that picture) and \vSixBSevenORRotninetyMissHRatePct\% of the horizontal ones, and $180^\circ$ readings \vSixBSevenORRotoneeightyMissHRatePct\% and \vSixBSevenORRotoneeightyMissVRatePct\%. A $180^\circ$ view thus errs on the same lattice direction as the upright one, and \vSixBSevenRotBothMissedH{} of the \vSixBSevenRotBothMissed{} edges both of them miss are horizontal; it still errs on different edges often enough to end exact on as many grids. The gain sits on the two largest sizes (Table~\ref{tab:v6_b7_size}): at $9{\times}9$ and $10{\times}10$ the repeat is exact on \vSixBSevenBthreeSizeNineXNineExact{} and \vSixBSevenBthreeSizeOneZeroXOneZeroExact{} of \vSixBSevenBthreeSizeNineXNineN{} grids, the $90^\circ$ arm on \vSixBSevenTVSizeNineXNineExact{} and \vSixBSevenTVSizeOneZeroXOneZeroExact{} and the $180^\circ$ arm on \vSixBSevenRotSizeNineXNineExact{} and \vSixBSevenRotSizeOneZeroXOneZeroExact. The four byte-identical upright readings of a grid return identical text on \vSixBSevenRAFourRawIdentical{} grids and the same cleaned edge set on \vSixBSevenRAFourIdentical{} of \vSixBSevenRAFourN, disagreeing on \vSixBSevenRAFourDisagreeMean{} lattice pairs per grid on average.

\paragraph{Results of B6.} The calls ran from \vFiveBSixSpanUtc{} (\vFiveBSixSpanHours{} hours, \vFiveBSixCallsTotal{} responses, model version \texttt{\vFiveBSixModelVersions} in every one; \$\vFiveBSixUsdTotal{} at list prices). The only deviation was a higher spending stop for the verification and solve phases, set while the readings were running; grids, arms, prompts and analysis did not change. After verification the $90^\circ$ arm is exact on \vFiveBSixTVExact/\vFiveBSixTVNGrids{} grids, the $180^\circ$ arm on \vFiveBSixRotExact/\vFiveBSixRotNGrids, and the repeats on \vFiveBSixBoneExact/\vFiveBSixBoneNGrids{} ($T=1$) and \vFiveBSixBthreeExact/\vFiveBSixBthreeNGrids{} ($T=0$) (Table~\ref{tab:v6_b6_xl}). The primary test is null ($b/c=\vFiveBSixTVVsRotExactB/\vFiveBSixTVVsRotExactC$, $p={}$\vFiveBSixTVVsRotExactP; Table~\ref{tab:v6_b6_paired}). Against the $T=0$ repeat both rotated arms point the way B7 later confirmed but do not reach significance on \vFiveBSixTVNGrids{} grids ($b/c=\vFiveBSixTVVsBthreeExactB/\vFiveBSixTVVsBthreeExactC$ and $\vFiveBSixRotVsBthreeExactB/\vFiveBSixRotVsBthreeExactC$, Holm $p={}$\vFiveBSixTVVsBthreeExactPHolm{} and \vFiveBSixRotVsBthreeExactPHolm). Both readings miss \vFiveBSixTVBothMissed{} and \vFiveBSixRotBothMissed{} true edges in the rotated arms against \vFiveBSixBoneBothMissed{} and \vFiveBSixBthreeBothMissed{} in the repeats, single readings miss edges by orientation as in B7 (Table~\ref{tab:v6_b6_orient}), and the ceiling is \vFiveBSixGTVcar/\vFiveBSixGTVcarN{} with Text\,\car{} and \vFiveBSixGTVcode/\vFiveBSixGTVcodeN{} with Text$\to$Code.

\paragraph{Reading.} Over the \vSixBSevenPoolTVN{} fresh grids of both experiments (Table~\ref{tab:v6_pooled}; descriptive), the $90^\circ$ and $180^\circ$ arms are exact on \vSixBSevenPoolTVExact{} and \vSixBSevenPoolRotExact{} grids and the $T=0$ repeat on \vSixBSevenPoolBthreeExact. On grids the pipeline was never tuned on, a second reading rendered differently from the first catches most of what one reading misses, whereas a repeated reading misses many of the same edges again; turning the second view by $90^\circ$ rather than $180^\circ$ changes how many edges both readings miss, not how many grids end exact. This sharpens the development-grid result of \S\ref{app:v5}, whose controls could not separate the second view from its orientation.

\begin{center}\scriptsize\resizebox{\linewidth}{!}{\input{gen_v6/tab_v6_b7_xl.tex}}\end{center}
\captionof{table}{B7 on \vSixBSevenGrids{} fresh extra-large grids ($7{\times}7$ to $10{\times}10$, 24 per size, \vSixBSevenPairs{} pairs), \texttt{gemini-3.8-flash} (low), all arms interleaved in one session. reads exact: single readings transcribed exactly (reading 1/reading 2); Jaccard: mean similarity of the two cleaned readings; disp./grid: disputed lattice pairs per grid; both miss: true edges missed by both readings; agree wrong: non-edges reported by both; verifier: correct crop answers; exact: grids transcribed exactly after verification; mean F1: final edge F1; miss/extra: final missing and extra edges; Text\,\car{} and Text$\to$Code: pairs solved from the final transcription; calls and list-price USD per arm. True structure: the two solve prompts on the ground-truth edge list (ceiling).}\label{tab:v6_b7_xl}

\begin{center}\scriptsize\resizebox{\linewidth}{!}{\input{gen_v8p/gen_v6/tab_v6_b7_paired.tex}}\end{center}
\captionof{table}{B7: paired comparisons. first/second: count for the first/second arm; diff: first minus second, in points; $b/c$: grids or pairs right only for the first / only for the second arm; exact two-sided McNemar $p$, printed as $<10^{-4}$ when smaller (exact values for the primary tests are in the text); Holm over the two primary tests and, separately, over the six solve-level tests; 95\% percentile bootstrap intervals over grids ($B=4000$).}\label{tab:v6_b7_paired}

\begin{center}\scriptsize\input{gen_v6/tab_v6_b7_orient.tex}\end{center}
\captionof{table}{B7: single-reading errors by edge orientation in grid coordinates (H horizontal, V vertical in the upright image), summed over grids. upright T=0: the four byte-identical upright readings of every grid; rot.\ $90^\circ$ and rot.\ $180^\circ$: the second readings of the two rotated arms.}\label{tab:v6_b7_orient}

\begin{center}\scriptsize\input{gen_v6/tab_v6_b7_size.tex}\end{center}
\captionof{table}{B7: grids transcribed exactly after verification, by size.}\label{tab:v6_b7_size}

\begin{center}\scriptsize\input{gen_v6/tab_v6_b7_pooled.tex}\end{center}
\captionof{table}{Fresh grids of B6 and B7 together (descriptive; B6 had been seen when B7 was planned): exact transcriptions after verification for the three arms both experiments ran, and discordant counts $b/c$ (first arm right only / second arm right only).}\label{tab:v6_pooled}

\begin{center}\scriptsize\resizebox{\linewidth}{!}{\input{gen_v5/tab_v5_b6_xl.tex}}\end{center}
\captionof{table}{B6 on \vFiveBSixTVNGrids{} fresh extra-large grids (6 per size, \vFiveBSixTVVcarN{} pairs), \texttt{gemini-3.8-flash} (low), all arms interleaved in one session; columns as in Table~\ref{tab:v6_b7_xl}. The $T=1$ repeat is the B1 control of \S\ref{app:v5}.}\label{tab:v6_b6_xl}

\begin{center}\scriptsize\resizebox{\linewidth}{!}{\input{gen_v8p/gen_v5/tab_v5_b6_paired.tex}}\end{center}
\captionof{table}{B6: paired comparisons; the primary test is rot.\ $90^\circ$ against rot.\ $180^\circ$ on exact transcriptions. Columns as in Table~\ref{tab:v6_b7_paired}; Holm over the four transcription-level tests; 95\% percentile bootstrap intervals over grids ($B=2000$).}\label{tab:v6_b6_paired}

\begin{center}\scriptsize\input{gen_v5/tab_v5_b6_orient.tex}\end{center}
\captionof{table}{B6: single-reading errors by edge orientation, as in Table~\ref{tab:v6_b7_orient}; upright T=1: the two readings of the $T=1$ repeat.}\label{tab:v6_b6_orient}

\section{One-shot conditions on fresh extra-large grids}
\label{app:x8}

\paragraph{Why.} The extra-large results of Section~\ref{sec:car_results} (Table~\ref{tab:xl}) come from the 24 development grids and their 60 pairs, the grids on which the two-view pipeline was also built: chain of thought solves none of the pairs, programs written from the exact text solve almost all of them, and from pixels the explicit transcription step of Perceive+Code and \car{} solves more pairs than Visual$\to$Code. B7 (\S\ref{app:v6}) generated \vSixBSevenGrids{} fresh grids and measured on them the two-view pipeline and the solver ceiling, but no one-shot condition. X8 runs the seven one-shot conditions of Table~\ref{tab:xl} on the same \vSixBSevenPairs{} pairs. This replicates the extra-large rows on four times as many grids and turns the comparison between the two-view pipeline and a one-shot condition into a paired one, on grids that neither was tuned on. As for B6 and B7, the plan, the runner and the grid file were hash-recorded before the first call, and the analysis script before any result record was read (the logs then held 183 records, unread).

\paragraph{Grids, conditions and analysis.} X8 uses the B7 grids unchanged: \vSixBSevenGrids{} grids, 24 per size from $7{\times}7$ to $10{\times}10$, with $\sST$ and $\sAM$ asked on every grid and $\sPM$ on the grids with an even number of vertices, each shown as its upright B7 image. The conditions are Visual CoT, Text CoT, Visual$\to$Code, Perceive+Code, \car{}, Text$\to$Code and Text\,\car{}, with the prompts, scoring and code runner of the main experiments unchanged; the image is attached in the four conditions that read pixels. Four configurations run every condition: \texttt{gemini-3.8-flash} at the low thinking level, the primary model, and \texttt{gemini-3.6-flash} (low), \texttt{gemini-3.1-pro-preview} (low) and \texttt{gemma-4-31b-it} (no thinking) as secondary models. We left out the high thinking level of \texttt{gemini-3.8-flash} because many of its extra-large calls in the main runs never returned. The 1{,}680 calls of a configuration are queued with the seven conditions shuffled and merged round-robin, so that all conditions advance together in one session, and failed calls are retried in later passes. The Text$\to$Code and Text\,\car{} requests of \texttt{gemini-3.8-flash} are byte-identical to the two ceiling requests of B7, which send the same prompts on the true edge list (we checked this on all pairs before launch), so they also measure test-retest across sessions.

The unit is the pair for answers and the grid for transcriptions; paired tests are exact two-sided McNemar tests, and intervals are percentile bootstraps over grids ($B=4000$). The two co-primary tests, on \texttt{gemini-3.8-flash} with Holm over the two at family-wise level $0.05$, are the two-view pipeline of B7 (Text\,\car{} on the verified two-view transcription) against one-shot \car{} from pixels (H1), and one-shot \car{} against Visual$\to$Code (H2), which isolates the explicit transcription step. H1 compares calls from two sessions, B7's and X8's, so a drift between sessions of the kind described in \S\ref{app:v5} cannot be excluded; H2 compares calls from one session. Each test uses the pairs on which both arms have a record and is repeated with a missing record scored as not exact. The secondary analyses of \texttt{gemini-3.8-flash} are the exact counts by size and by task for all seven conditions; seven more paired comparisons with Holm over the seven (Perceive+Code against Visual$\to$Code, \car{} against Perceive+Code, Text$\to$Code against Visual$\to$Code, Text\,\car{} against \car{}, Text\,\car{} against Text$\to$Code, and the two-view pipeline against Text\,\car{} and against Perceive+Code); the exact answers from pixels at $9{\times}9$ and $10{\times}10$; where the errors of the pixel conditions come from, decomposed as in Table~\ref{tab:attrib} (the edge list embedded in a program, or the final edge list of a Visual CoT response, against the true one, and among wrong-graph programs those whose answer is the exact count of their own graph); the algorithm families of the programs; the retest against B7's ceiling; and a comparison of each condition on the fresh grids with the development grids (Fisher exact test, descriptive only). For each secondary model we report the counts, the decomposition and three paired comparisons, \car{} against Visual$\to$Code, Perceive+Code against Visual$\to$Code and Text\,\car{} against \car{}, with Holm over the three.

\paragraph{Results.} The calls ran from \vSixXEightSpanUtc{} (\vSixXEightSpanHours{} hours, \vSixXEightCallsOkTotal{} responses, model versions \texttt{\vSixXEightModelVersions}); after the retries no call is without a response, and the list price was \$\vSixXEightUsdTotal{}. The fresh grids reproduce the pattern of the extra-large rows of Table~\ref{tab:xl} (Table~\ref{tab:x8_main}). Chain of thought solves none of the \vSixXEightPairs{} pairs in any configuration, from pixels or from text. From the exact text \texttt{gemini-3.8-flash} solves \vSixXEightGThreeEightLowTcodeK{} pairs with Text$\to$Code and \vSixXEightGThreeEightLowTcarK{} with Text\,\car{}, and \texttt{gemini-3.1-pro-preview} \vSixXEightProLowTcodeK{} and \vSixXEightProLowTcarK{}. From pixels \texttt{gemini-3.8-flash} solves \vSixXEightGThreeEightLowCodeK{} with Visual$\to$Code, \vSixXEightGThreeEightLowPcodeK{} with Perceive+Code and \vSixXEightGThreeEightLowCarK{} with \car{}.

H1 is supported: the two-view pipeline of B7 is exact on \vSixXEightHOneKa{} pairs and one-shot \car{} on \vSixXEightHOneKb{} ($b/c={}$\vSixXEightHOneB/\vSixXEightHOneC, Holm $p={}$\vSixXEightHOnePHolm; Table~\ref{tab:x8_paired}). This test crosses sessions, so the retest matters: the Text\,\car{} and Text$\to$Code requests of X8, byte-identical to the two ceiling requests of B7, are exact on \vSixXEightRetestTcarKa{} and \vSixXEightRetestTcodeKa{} pairs against \vSixXEightRetestTcarKb{} and \vSixXEightRetestTcodeKb{} in the B7 session ($p={}$\vSixXEightRetestTcarP{} and \vSixXEightRetestTcodeP; Table~\ref{tab:x8_retest}, top), so where the two sessions can be compared they agree. H2 is not supported: one-shot \car{} is exact on \vSixXEightHTwoKa{} pairs and Visual$\to$Code on \vSixXEightHTwoKb{} ($b/c={}$\vSixXEightHTwoB/\vSixXEightHTwoC, Holm $p={}$\vSixXEightHTwoPHolm). On the fresh grids a single explicit transcription step does not beat writing the program directly, and the point estimate goes the other way. No record is missing, so the sensitivity rows repeat the primary ones.

Among the secondary comparisons of \texttt{gemini-3.8-flash}, Perceive+Code does not differ from Visual$\to$Code ($b/c={}$\vSixXEightPcodeVsCodeB/\vSixXEightPcodeVsCodeC, Holm $p={}$\vSixXEightPcodeVsCodePHolm), nor \car{} from Perceive+Code (\vSixXEightCarVsPcodeB/\vSixXEightCarVsPcodeC, Holm $p={}$\vSixXEightCarVsPcodePHolm). Giving the program the exact text instead of the picture raises the exact pairs from \vSixXEightGThreeEightLowCodeK{} to \vSixXEightGThreeEightLowTcodeK{} (\vSixXEightTcodeVsCodeB/\vSixXEightTcodeVsCodeC, Holm $p={}$\vSixXEightTcodeVsCodePHolm), and for \car{} from \vSixXEightGThreeEightLowCarK{} to \vSixXEightGThreeEightLowTcarK{} (\vSixXEightTcarVsCarB/\vSixXEightTcarVsCarC, Holm $p={}$\vSixXEightTcarVsCarPHolm). The two-view pipeline recovers most of this gap but not all of it: Text\,\car{} on the true edge list is exact on \vSixXEightTvVsTcarC{} pairs on which the pipeline is not, and the pipeline on \vSixXEightTvVsTcarB{} the other way round (Holm $p={}$\vSixXEightTvVsTcarPHolm). The decomposition (Table~\ref{tab:x8_decomp}) places the errors from pixels in the reading. Visual CoT ends with the exact edge list in \vSixXEightDecGThreeEightLowVisualTOneAZero{} of \vSixXEightDecGThreeEightLowVisualRecords{} responses and answers none of them exactly. Among the programs with a wrong embedded edge list whose own graph we could count, \vSixXEightDecGThreeEightLowCodeTwrongExecMatchesGhat{} of \vSixXEightDecGThreeEightLowCodeTwrongCountAvailable{} (Visual$\to$Code), \vSixXEightDecGThreeEightLowPcodeTwrongExecMatchesGhat{} of \vSixXEightDecGThreeEightLowPcodeTwrongCountAvailable{} (Perceive+Code) and \vSixXEightDecGThreeEightLowCarTwrongExecMatchesGhat{} of \vSixXEightDecGThreeEightLowCarTwrongCountAvailable{} (\car{}) return the exact count of the graph they wrote down. Only \vSixXEightDecGThreeEightLowCodeScorable{} of the \vSixXEightDecGThreeEightLowCodeRecords{} Visual$\to$Code programs embed an edge list the extractor can read, so the decomposition covers part of the responses.

The secondary models repeat the pattern. \texttt{gemini-3.6-flash} solves \vSixXEightGThreeSixLowCodeK{}, \vSixXEightGThreeSixLowPcodeK{} and \vSixXEightGThreeSixLowCarK{} pairs from pixels (Visual$\to$Code, Perceive+Code, \car{}) and \vSixXEightGThreeSixLowTcodeK{} and \vSixXEightGThreeSixLowTcarK{} from text. \texttt{gemini-3.1-pro-preview} at low thinking reads the fresh grids worst of the three Gemini models, \vSixXEightProLowCodeK{}, \vSixXEightProLowPcodeK{} and \vSixXEightProLowCarK{} pairs from pixels, while it solves \vSixXEightProLowTcarK{} of \vSixXEightPairs{} with Text\,\car{}. \texttt{gemma-4-31b-it} without thinking solves no pair from pixels and \vSixXEightGmThreeOneMinTcarK{} with Text\,\car{}. For \texttt{gemini-3.6-flash} and \texttt{gemini-3.1-pro-preview}, neither \car{} nor Perceive+Code differs from Visual$\to$Code (Holm $p={}$\vSixXEightGThreeSixLowCarVsCodePHolm{} and \vSixXEightGThreeSixLowPcodeVsCodePHolm{}; \vSixXEightProLowCarVsCodePHolm{} and \vSixXEightProLowPcodeVsCodePHolm). Against the development grids (Table~\ref{tab:x8_retest}, bottom; descriptive), one cell differs: Visual$\to$Code of \texttt{gemini-3.8-flash} is exact on \vSixXEightDevGThreeEightLowCodeK{} of \vSixXEightDevGThreeEightLowCodeN{} development pairs and on \vSixXEightGThreeEightLowCodeK{} of \vSixXEightPairs{} fresh ones (Fisher $p={}$\vSixXEightDevGThreeEightLowCodeP), while \car{} and Perceive+Code score as on the development grids (Fisher $p={}$\vSixXEightDevGThreeEightLowCarP{} and \vSixXEightDevGThreeEightLowPcodeP). The advantage of the explicit transcription step over Visual$\to$Code that Section~\ref{sec:car_results} reports on the development grids thus appears to come from a low Visual$\to$Code score there, and it does not carry over to the fresh grids.

\begin{keepblock}
\begin{center}\scriptsize\genfitsetinput{\input{gen_v6/tab_v6_x8_main.tex}}\end{center}
\captionof{table}{X8: exact answers on the \vSixBSevenPairs{} pairs of the \vSixBSevenGrids{} fresh extra-large grids by grid size, for the seven one-shot conditions and four configurations; the calls of each configuration ran interleaved in one session. 95\% CI: percentile bootstrap over grids ($B=4000$); tokens: mean output tokens per call (thoughts plus answer). The last three rows are B7 on the same pairs, from its own session: the two-view pipeline, and the two solve prompts on the true edge list (the ceiling).}\label{tab:x8_main}
\end{keepblock}

\begin{keepblock}
\begin{center}\scriptsize\genfitsetinput{\input{gen_v6/tab_v6_x8_task.tex}}\end{center}
\captionof{table}{X8: exact answers by task ($\sST$ and $\sAM$ on all \vSixBSevenGrids{} grids, $\sPM$ on the grids with an even number of vertices); rows as in Table~\ref{tab:x8_main}.}\label{tab:x8_task}
\end{keepblock}

\begin{keepblock}
\begin{center}\scriptsize\genfitsetinput{\input{gen_v8p/gen_v6/tab_v6_x8_paired.tex}}\end{center}
\captionof{table}{X8: paired comparisons on pairs. diff: first minus second, in points; $b/c$: pairs right only for the first / only for the second arm; exact two-sided McNemar $p$, printed as $<10^{-4}$ when smaller; 95\% percentile bootstrap intervals over grids ($B=4000$). role: primary, Holm over the two; sensitivity, the same test with a missing record scored as not exact; secondary, Holm over the seven listed in the text for \texttt{gemini-3.8-flash} and over the three within each other configuration; descriptive, no correction. Arms without a model name are \texttt{gemini-3.8-flash} (low); two-view (B7) and true structure (B7) come from the B7 session.}\label{tab:x8_paired}
\end{keepblock}

\begin{keepblock}
\begin{center}\scriptsize\genfitsetinput{\input{gen_v6/tab_v6_x8_decomp.tex}}\end{center}
\captionof{table}{X8: where the errors come from. records: responses with a record; no list: responses without a program or without an extractable edge list (not scored); $T$: the edge list embedded in the program, or the final edge list of a Visual CoT response, equals the true one; $A$: the answer is exact; last column: among scorable programs with $T=0$ whose own graph could be counted, those whose answer is the exact count of that graph. The Text$\to$Code and Text\,\car{} rows are descriptive.}\label{tab:x8_decomp}
\end{keepblock}

\begin{keepblock}
\begin{center}\scriptsize\genfitsetinput{\input{gen_v6/tab_v6_x8_retest.tex}}\end{center}
\captionof{table}{X8: descriptive comparisons. Top: the Text\,\car{} and Text$\to$Code requests of \texttt{gemini-3.8-flash} in X8 against the byte-identical ceiling requests of B7, sent in a different session ($b/c$ and $p$ as in Table~\ref{tab:x8_paired}). Bottom: exact answers on the 24 development grids (60 pairs, main runs) against the fresh grids (X8) for each configuration and condition run on both, with a two-sided Fisher exact test; not confirmatory.}\label{tab:x8_retest}
\end{keepblock}

\section{Degraded renderings of the core images}
\label{app:d8}

\paragraph{Why.} Section~\ref{sec:res_percept} reports that transcription of our schematic inputs is mostly exact, and the limitations in Section~\ref{sec:discussion} add that the images are clean and synthetic. D8 asks two questions about the same drawings rendered less cleanly. First, how far does exact transcription survive compression, blur, a slanted capture, a hand-drawn wobble of the strokes, half resolution and a photo-like combination of these? Second, when perception is made harder on purpose, does the attribution statistic of Table~\ref{tab:attribution}, the share of wrong answers that come with an exact transcription, move errors toward perception, as it should if it attributes them correctly? D8 tests degraded renderings of the same synthetic drawings; it does not test natural images, hand-drawn diagrams or other model families. The plan, the runner, the image generator and the hash of every image were recorded before the first call, and the analysis script before any result record was read (the logs then held 823 records, unread).

\paragraph{Images and arms.} The population is the \vSixDEightImages{} core images of the benchmark (66 grid drawings, 40 binary images and 64 Eulerian drawings, two per graph) and their \vSixDEightPairs{} (image, task) pairs; the extra-large grids are not included. There are twelve arms (Figure~\ref{fig:d8_arms}). Clean sends the original files, the bytes every earlier run sent, and clean, repeated sends the same requests a second time. The ten degraded arms are JPEG at quality 30 and 10; Gaussian blur with $\sigma$ of 1.5 and 3 pixels; perspective, in which each corner moves inward by up to 6\% or 12\% of the side and the drawing is mapped onto that quadrilateral, as when a printout is photographed at a slant; elastic, a smooth random displacement of at most 3 or 6 pixels that makes the strokes wobble as if drawn by hand; half resolution, a $2{\times}2$ box average; and photo-like, which chains a 4\% perspective, a 2-pixel elastic displacement, a blur with $\sigma=0.8$, a linear fall-off of the illumination, a colour cast, Gaussian noise ($\sigma=5$ on the 0 to 255 scale) and JPEG at quality 55. Sizes are in pixels of the source image, whose long side is 381 to 741 pixels. Each degraded image is a deterministic function of its source and a seed, and the runner checks every file against its recorded hash before a run. The prompts are the same in every arm, so the arms differ only in the image bytes; under the elastic and perspective arms the segments are slightly curved or projectively distorted while the prompts still call them straight, which we keep on purpose.

\paragraph{Calls and analysis.} \texttt{gemini-3.8-flash} at the low thinking level, the primary model, gets three kinds of call in every arm: the task-independent transcription prompt of Experiment~A (\S\ref{app:v4}), once per image, and Visual CoT and Visual$\to$Code, once per pair. \texttt{gemini-3.1-pro-preview} (low) and \texttt{gemini-3.6-flash} (low) get the transcription and Visual CoT calls, and \texttt{gemma-4-31b-it} (no thinking) the transcription call, 26{,}976 calls in all. Prompts, scoring and code runner are those of the earlier runs. Before launch we checked, without sending anything, that the clean transcription requests are byte-identical to those of Experiment~A, that the clean Visual CoT and Visual$\to$Code requests are byte-identical to those of the main runs, that every repeated request is byte-identical to its clean request, and that every degraded request differs from its clean request only in the image bytes. For each configuration and kind the items are shuffled once and the twelve arms of an item are sent in a shuffled order within minutes of each other, so every arm, the clean control included, is measured in the same session; failed calls are retried in later passes.

The unit is the image for transcriptions and the pair for answers, and the outcome is an exact transcription or answer (an unparsable response counts as not exact). Each arm is compared with clean by an exact two-sided McNemar test, and the effect is the difference in exact rate (arm minus clean) with Newcombe's hybrid-score interval for paired proportions and a percentile bootstrap interval over the \vSixDEightClusters{} clusters (a grid, a binary image, or an Eulerian graph with its two drawings; $B=4000$). The primary analysis is the transcription of \texttt{gemini-3.8-flash}, clean against each of the ten degraded arms on the \vSixDEightImages{} images, with Holm over the ten at family-wise level $0.05$. An arm whose Holm-adjusted $p$ is $0.05$ or more is described as showing no detectable change, always together with the lower end of its interval, never as showing no change. As sensitivity analyses each test is repeated with a missing record scored as not exact, and at the level of clusters, a cluster counting as exact in an arm when all its images are (Holm over the ten in each). Clean against its repeat is the negative control, outside every Holm family: it measures how far a byte-identical request moves within one session, and we also count the keys whose two stored responses are identical.

The secondary analyses of \texttt{gemini-3.8-flash} are: transcription errors per arm by family and size (mean F1; missed, extra and off-lattice edges of grids, with missed edges split into horizontal and vertical as rates of the 1{,}051 horizontal and 1{,}004 vertical drawn edges, an unparsed grid transcription missing all of its edges; wrong cells of binary images); Visual CoT and Visual$\to$Code answers, each degraded arm against clean with Holm over the ten for each kind; attribution under degradation, that is, per arm the two-by-two table of an exact transcription inside the Visual CoT response by an exact answer, the probability of an exact answer given an exact transcription, and the share of wrong answers that come with an exact transcription, each compared with clean by a cluster bootstrap, together with the Spearman correlation over the ten arms between the change in exact transcription and the change in that share; the exact rate of the transcription inside Visual CoT next to that of the separate transcription call; the decomposition of Visual$\to$Code as in \S\ref{app:x8}, scored on grid programs only, because the edge-list extractor of Table~\ref{tab:attrib} reads grid programs (binary-image and Eulerian programs are counted but not scored); the stronger level of each degradation against the milder one; and the clean calls against byte-identical requests of earlier sessions (Experiment~A for transcriptions, the main runs for answers). The last two are descriptive. The two other Gemini models get the transcription and Visual CoT tests, with Holm over the ten for each, and the descriptive analyses; \texttt{gemma-4-31b-it} gets the transcription tests.

\paragraph{Results.} The calls ran from \vSixDEightSpanUtc{} (\vSixDEightSpanHours{} hours, \vSixDEightCallsOkTotal{} responses, model versions \texttt{\vSixDEightModelVersions}); no key is without a response, and the list price was \$\vSixDEightUsdTotal{}. In the primary analysis (Table~\ref{tab:d8_primary}) \texttt{gemini-3.8-flash} transcribes \vSixDEightGThreeEightLowTransCleanK{} of the \vSixDEightImages{} clean images exactly, and one of the ten degraded arms shows a detectable loss: JPEG at quality 30, with \vSixDEightGThreeEightLowTransJpegThreeZeroK{} exact transcriptions ($b/c={}$\vSixDEightGThreeEightLowTransJpegThreeZeroVsCleanB/\vSixDEightGThreeEightLowTransJpegThreeZeroVsCleanC, Holm $p={}$\vSixDEightGThreeEightLowTransJpegThreeZeroVsCleanPHolm; difference $\vSixDEightGThreeEightLowTransJpegThreeZeroVsCleanDiff$ points, Newcombe interval \vSixDEightGThreeEightLowTransJpegThreeZeroVsCleanCI). The other nine arms show no detectable change, with \vSixDEightGThreeEightLowTransJpegOneZeroK{} to \vSixDEightGThreeEightLowTransElasticSixK{} exact transcriptions; the lower ends of their Newcombe intervals lie between $\vSixDEightGThreeEightLowTransJpegOneZeroVsCleanLower$ points (JPEG at quality 10, and photo-like) and $\vSixDEightGThreeEightLowTransElasticSixVsCleanLower$ points (elastic, 6 pixels). The stronger compression does not cost more than the milder one: quality 10 is exact on \vSixDEightDoseGThreeEightLowJpegC{} images on which quality 30 is not, and quality 30 on \vSixDEightDoseGThreeEightLowJpegB{} the other way round ($p={}$\vSixDEightDoseGThreeEightLowJpegP; Table~\ref{tab:d8_retest}, bottom). In the negative control the byte-identical repeat of the clean requests is exact on \vSixDEightGThreeEightLowTransCleanRepK{} images ($b/c={}$\vSixDEightGThreeEightLowTransRepB/\vSixDEightGThreeEightLowTransRepC, $p={}$\vSixDEightGThreeEightLowTransRepP), and the clean calls match the byte-identical calls of Experiment~A, \vSixDEightRetestGThreeEightLowTransExpAKa{} against \vSixDEightRetestGThreeEightLowTransExpAKb{} exact images. The two other Gemini models show no detectable change in any arm (Table~\ref{tab:d8_models}); \texttt{gemma-4-31b-it} loses under blur with $\sigma=3$, with \vSixDEightGmThreeOneMinTransBlurThreeZeroK{} exact transcriptions against \vSixDEightGmThreeOneMinTransCleanK{} under clean (Holm $p={}$\vSixDEightGmThreeOneMinTransBlurThreeZeroVsCleanPHolm).

The answers of Visual CoT move within the noise of the session (Table~\ref{tab:d8_answers}). Under clean \texttt{gemini-3.8-flash} answers \vSixDEightGThreeEightLowVisualCleanK{} of the \vSixDEightPairs{} pairs exactly and \vSixDEightGThreeEightLowVisualCleanRepK{} in the byte-identical repeat ($b/c={}$\vSixDEightGThreeEightLowVisualRepB/\vSixDEightGThreeEightLowVisualRepC, $p={}$\vSixDEightGThreeEightLowVisualRepP). The degraded arms give \vSixDEightGThreeEightLowVisualDegMinK{} to \vSixDEightGThreeEightLowVisualDegMaxK{}, and only elastic with 3 pixels passes Holm (\vSixDEightGThreeEightLowVisualElasticThreeVsCleanB/\vSixDEightGThreeEightLowVisualElasticThreeVsCleanC, Holm $p={}$\vSixDEightGThreeEightLowVisualElasticThreeVsCleanPHolm), by a margin of the same order as the difference between the clean calls and their repeat. Neither \texttt{gemini-3.1-pro-preview} nor \texttt{gemini-3.6-flash} changes detectably in Visual CoT under any arm. Visual$\to$Code is more sensitive: from \vSixDEightGThreeEightLowCodeCleanK{} exact answers under clean it falls to \vSixDEightGThreeEightLowCodeJpegThreeZeroK{} under JPEG at quality 30 (\vSixDEightGThreeEightLowCodeJpegThreeZeroVsCleanB/\vSixDEightGThreeEightLowCodeJpegThreeZeroVsCleanC, Holm $p={}$\vSixDEightGThreeEightLowCodeJpegThreeZeroVsCleanPHolm) and to \vSixDEightGThreeEightLowCodePhotoK{} under the photo-like arm (\vSixDEightGThreeEightLowCodePhotoVsCleanB/\vSixDEightGThreeEightLowCodePhotoVsCleanC, Holm $p={}$\vSixDEightGThreeEightLowCodePhotoVsCleanPHolm). The decomposition puts these losses on the reading (Table~\ref{tab:d8_code}): under JPEG at quality 30, \vSixDEightDecGThreeEightLowJpegThreeZeroTZeroAZero{} scorable grid programs embed a wrong edge list and return a wrong answer, against \vSixDEightDecGThreeEightLowCleanTZeroAZero{} under clean, and \vSixDEightDecGThreeEightLowJpegThreeZeroTwrongExecMatchesGhat{} of the \vSixDEightDecGThreeEightLowJpegThreeZeroTwrongCountAvailable{} wrong-graph programs we could check return the exact count of their own graph.

The attribution statistic moves the way it should when perception is made harder (Table~\ref{tab:d8_attrib}). For \texttt{gemini-3.8-flash}, the share of wrong Visual CoT answers that come with an exact transcription is \vSixDEightAttGThreeEightLowCleanShare\% under clean, \vSixDEightAttGThreeEightLowJpegThreeZeroShare\% under JPEG at quality 30 (change $\vSixDEightAttGThreeEightLowJpegThreeZeroShareDiff$ points, \vSixDEightAttGThreeEightLowJpegThreeZeroShareCI) and \vSixDEightAttGThreeEightLowPhotoShare\% under the photo-like arm ($\vSixDEightAttGThreeEightLowPhotoShareDiff$, \vSixDEightAttGThreeEightLowPhotoShareCI). Over the ten arms the change in this share follows the change in exact transcription (Spearman $\rho={}$\vSixDEightAttGThreeEightLowSpearmanRho, $p={}$\vSixDEightAttGThreeEightLowSpearmanP; \texttt{gemini-3.6-flash}: $\rho={}$\vSixDEightAttGThreeSixLowSpearmanRho, $p={}$\vSixDEightAttGThreeSixLowSpearmanP). For \texttt{gemini-3.1-pro-preview} it does not ($\rho={}$\vSixDEightAttProLowSpearmanRho, $p={}$\vSixDEightAttProLowSpearmanP), but neither its transcriptions nor its answers change detectably, so there is little for the share to follow. In short, for the three Gemini models the transcription of these drawings shows no detectable loss under blur, slanted capture, wobbly strokes and half resolution; heavy JPEG compression is the one degradation that costs the primary model exact transcriptions and, through them, programs; and where transcription gets worse, the attribution statistic of Table~\ref{tab:attribution} moves the errors toward perception.

\begin{keepblock}
\begin{center}\setlength{\tabcolsep}{2pt}\scriptsize
\begin{tabular}{cccccc}
\includegraphics[width=0.155\linewidth]{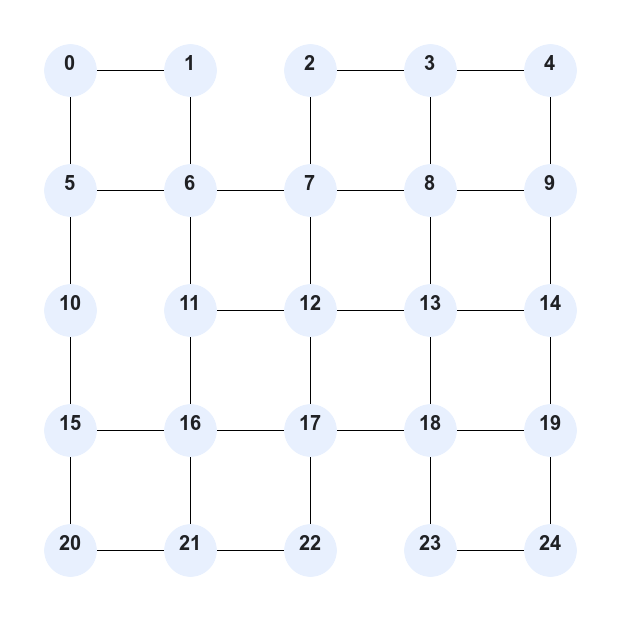} &
\includegraphics[width=0.155\linewidth]{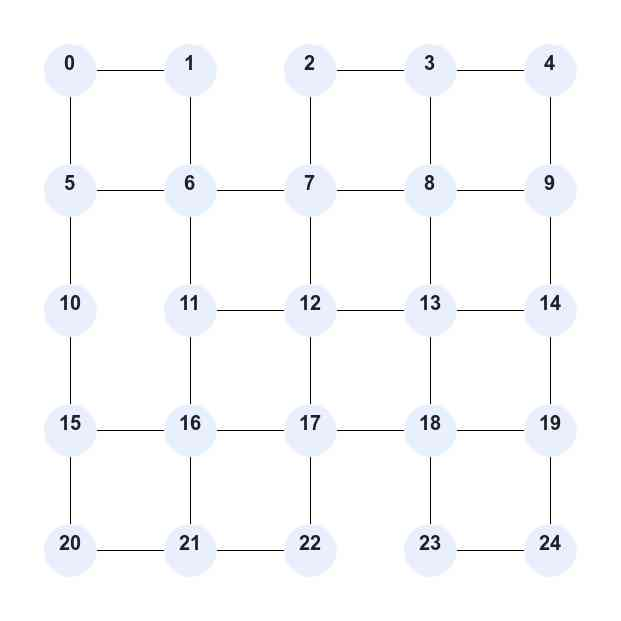} &
\includegraphics[width=0.155\linewidth]{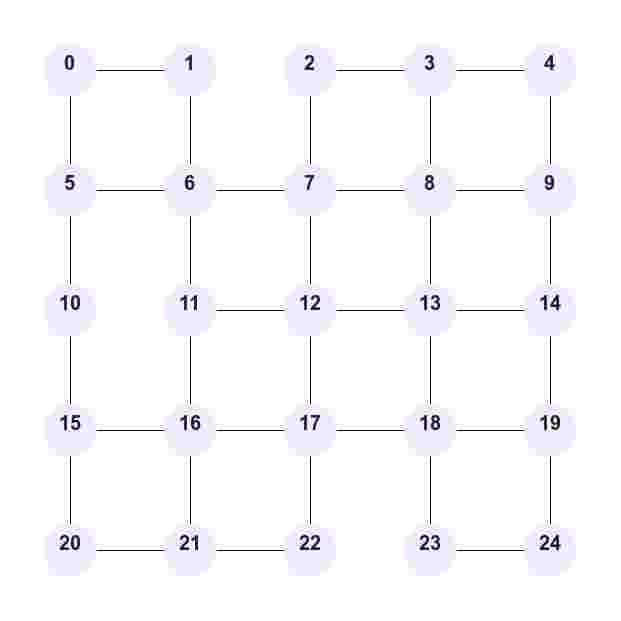} &
\includegraphics[width=0.155\linewidth]{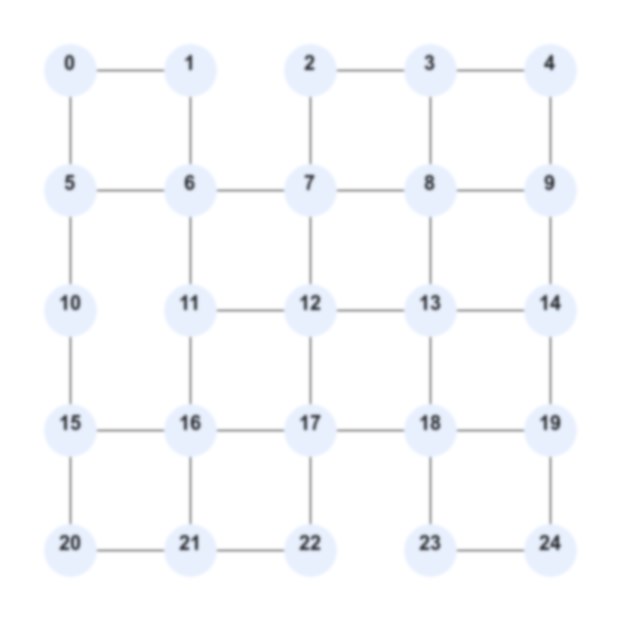} &
\includegraphics[width=0.155\linewidth]{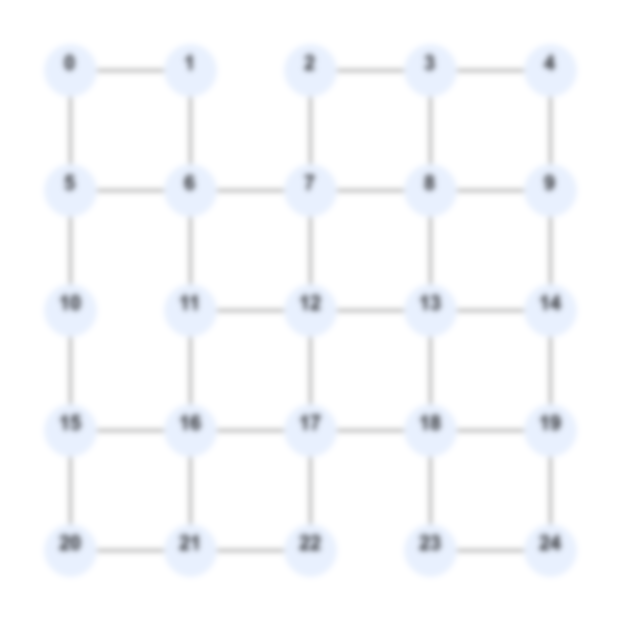} &
\includegraphics[width=0.155\linewidth]{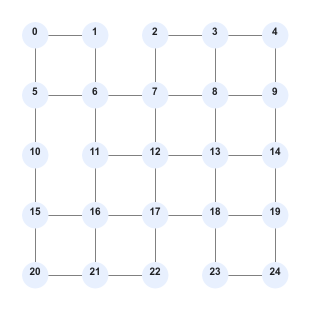} \\
clean & JPEG $q{=}30$ & JPEG $q{=}10$ & blur $\sigma{=}1.5$ & blur $\sigma{=}3$ & half resolution \\[4pt]
\includegraphics[width=0.155\linewidth]{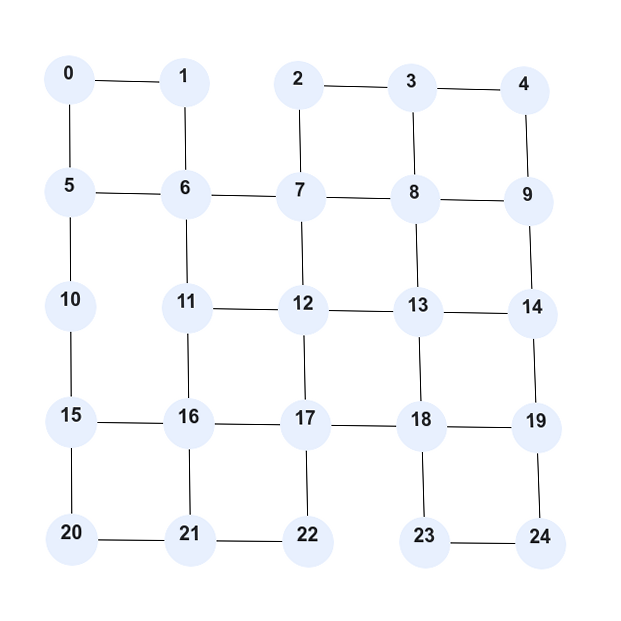} &
\includegraphics[width=0.155\linewidth]{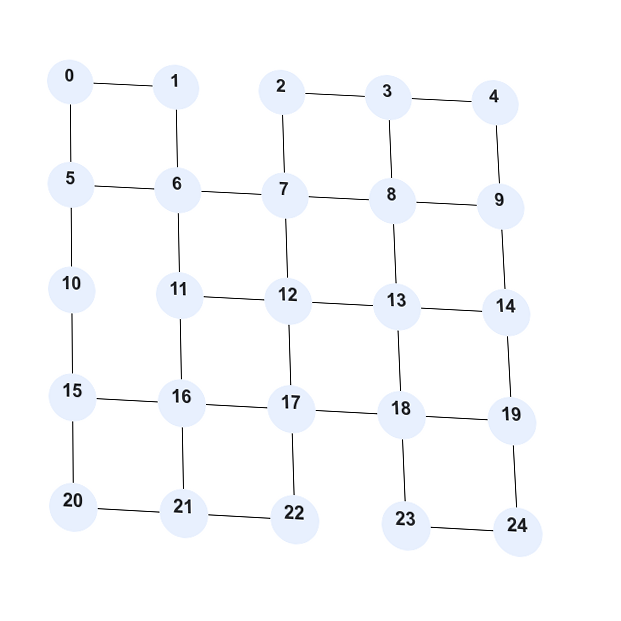} &
\includegraphics[width=0.155\linewidth]{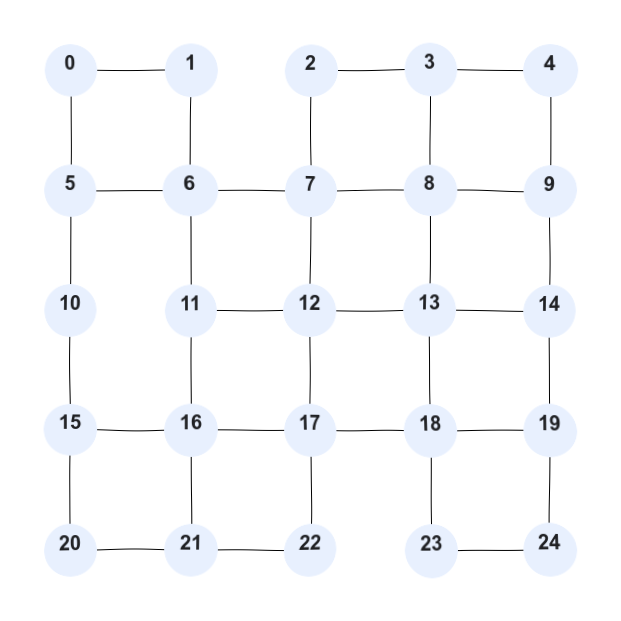} &
\includegraphics[width=0.155\linewidth]{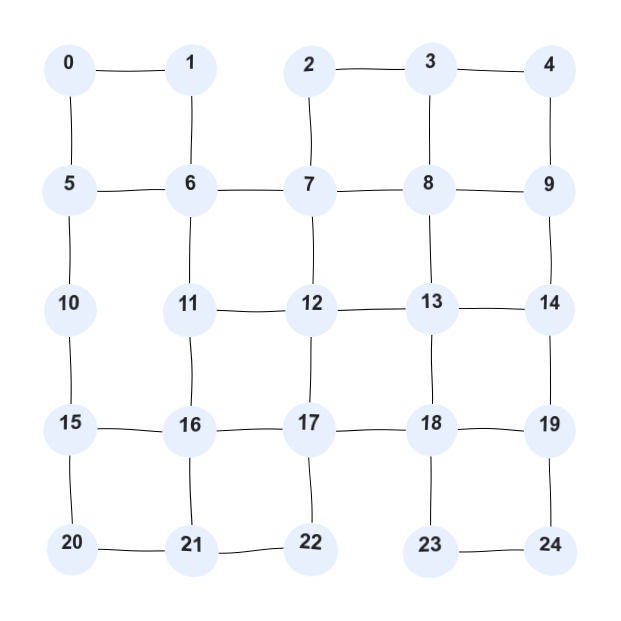} &
\includegraphics[width=0.155\linewidth]{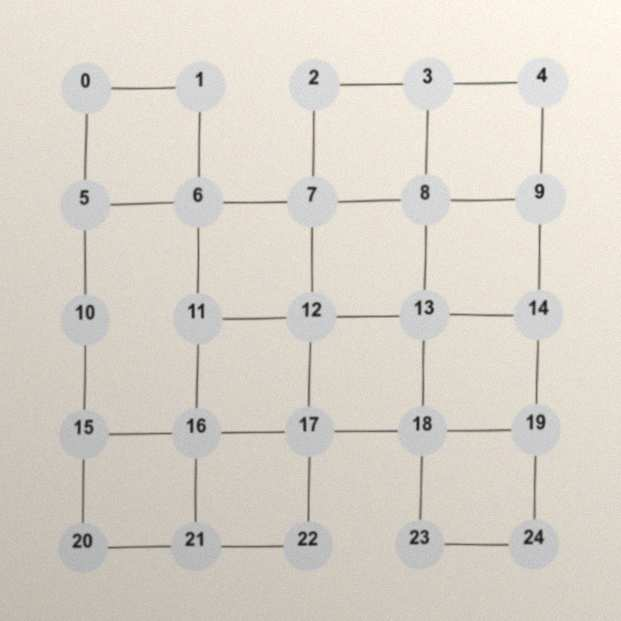} & \\
perspective 6\% & perspective 12\% & elastic 3\,px & elastic 6\,px & photo-like & \\
\end{tabular}\end{center}

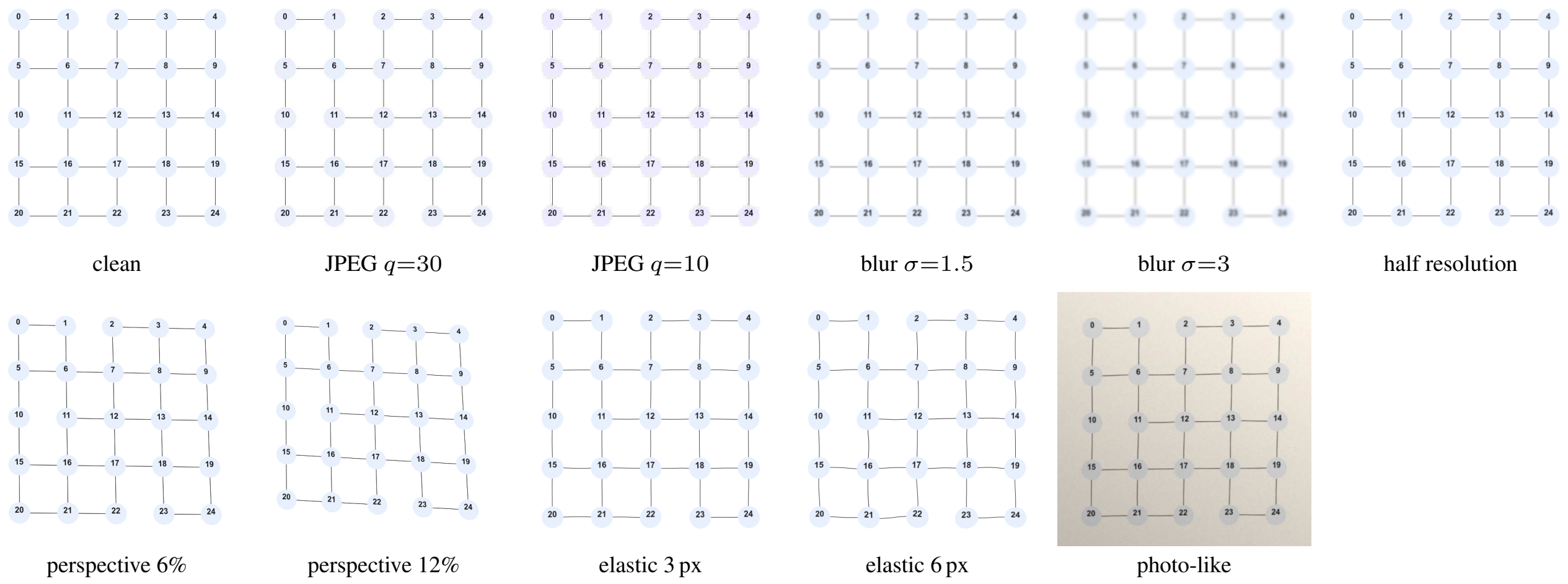
\captionof{figure}{D8: one core drawing in the clean arm and the ten degraded arms (the first $5{\times}5$ grid in sorted order, fixed before any result was read). Clean, repeated sends the clean bytes again and has no image of its own. The half-resolution image is shown at the same size as the others.}\label{fig:d8_arms}
\end{keepblock}

\begin{keepblock}
\begin{center}\scriptsize\genfitsetinput{\input{gen_v8p/gen_v6/tab_v6_d8_primary.tex}}\end{center}
\captionof{table}{D8, primary analysis: exact transcriptions of the \vSixDEightImages{} core images by \texttt{gemini-3.8-flash} (low), each arm against clean in the same session. diff: arm minus clean, in points; Newcombe: hybrid-score 95\% interval for paired proportions; bootstrap: 95\% percentile interval over the \vSixDEightClusters{} clusters ($B=4000$); $b/c$: images exact only under clean / only under the arm; exact two-sided McNemar $p$, printed as $<10^{-4}$ when smaller; Holm over the ten degraded arms. clean, repeated: the byte-identical repeat of the clean request, the negative control (no Holm family).}\label{tab:d8_primary}
\end{keepblock}

\begin{keepblock}
\begin{center}\scriptsize\genfitsetinput{\input{gen_v6/tab_v6_d8_sens.tex}}\end{center}
\captionof{table}{D8, sensitivity of the transcription tests for every configuration: with a missing record scored as not exact (left), and at the level of the \vSixDEightClusters{} clusters, a cluster counting as exact in an arm when all its images are (right); Holm over the ten degraded arms in each.}\label{tab:d8_sens}
\end{keepblock}

\begin{keepblock}
\begin{center}\scriptsize\genfitsetinput{\input{gen_v6/tab_v6_d8_models.tex}}\end{center}
\captionof{table}{D8: exact transcriptions per arm for the four configurations, with the Holm-adjusted $p$ against clean over the ten degraded arms of each configuration; $^\dagger$: unadjusted $p$ of the repeat against clean (negative control).}\label{tab:d8_models}
\end{keepblock}

\begin{keepblock}
\begin{center}\scriptsize\genfitsetinput{\input{gen_v6/tab_v6_d8_answers.tex}}\end{center}
\captionof{table}{D8: exact answers on the \vSixDEightPairs{} pairs per arm, Visual CoT for the three Gemini models and Visual$\to$Code for \texttt{gemini-3.8-flash}, with Holm over the ten degraded arms in each column; $^\dagger$ as in Table~\ref{tab:d8_models}.}\label{tab:d8_answers}
\end{keepblock}

\begin{keepblock}
\begin{center}\scriptsize\genfitsetinput{\input{gen_v8p/gen_v6/tab_v6_d8_attrib.tex}}\end{center}
\captionof{table}{D8: attribution under degradation (Visual CoT). $T$: the transcription inside the response is exact; $A$: the answer is exact; $P(A{\mid}T{=}1)$ in \%; errors with $T{=}1$: the share of wrong answers that come with an exact transcription, the statistic of Table~\ref{tab:attribution}; vs clean: its change against clean in points, with a 95\% percentile bootstrap interval over clusters ($B=4000$).}\label{tab:d8_attrib}
\end{keepblock}

\begin{keepblock}
\begin{center}\scriptsize\genfitsetinput{\input{gen_v6/tab_v6_d8_percept.tex}}\end{center}
\captionof{table}{D8: exact transcription inside the Visual CoT response (pairs) next to the separate transcription call (images), per arm and configuration.}\label{tab:d8_percept}
\end{keepblock}

\begin{keepblock}
\begin{center}\scriptsize\genfitsetinput{\input{gen_v6/tab_v6_d8_errors.tex}}\end{center}
\captionof{table}{D8: transcription errors per arm and configuration. Exact transcriptions per family (Eulerian drawings split into undirected and directed); mean F1 over parsed responses; missed horizontal and vertical grid edges as rates of the 1{,}051 / 1{,}004 drawn edges (an unparsed grid transcription misses all of its edges); extra edges (grids and Eulerian graphs) and off-lattice edges (grids), summed over images; wrong cells of binary images, summed; unparsed responses.}\label{tab:d8_errors}
\end{keepblock}

\begin{keepblock}
\begin{center}\scriptsize\genfitsetinput{\input{gen_v6/tab_v6_d8_code.tex}}\end{center}
\captionof{table}{D8: where the errors of Visual$\to$Code come from (\texttt{gemini-3.8-flash}), per arm. programs: responses with a record; binary/Euler: programs on binary images and Eulerian drawings, counted but not scored; grid, no list: grid responses without a program or without an extractable edge list; $T$: the edge list embedded in the program equals the true one; $A$: the answer is exact; last column: among programs with $T=0$ whose own graph could be counted, those whose answer is the exact count of that graph.}\label{tab:d8_code}
\end{keepblock}

\begin{keepblock}
\begin{center}\scriptsize\genfitsetinput{\input{gen_v6/tab_v6_d8_retest.tex}}\end{center}
\captionof{table}{D8: descriptive comparisons, no correction. Top: clean against its byte-identical repeat in the same session, per configuration and kind, with the number of keys whose two stored responses are identical. Middle: the D8 clean calls against byte-identical requests sent in earlier sessions (Experiment~A for transcriptions, the main runs for Visual CoT and Visual$\to$Code; $n$ counts the units with a record in both). Bottom: the milder against the stronger level of each degradation (transcription). $b/c$: units right only in the first / only in the second; exact two-sided McNemar $p$.}\label{tab:d8_retest}
\end{keepblock}

\begin{keepblock}
\begin{center}\scriptsize\genfitsetinput{\input{gen_v6/tab_v6_d8_cost.tex}}\end{center}
\captionof{table}{D8: calls with a record, mean thought tokens under clean and averaged over the ten degraded arms, list-price USD, time span and reported model version, per configuration and kind.}\label{tab:d8_cost}
\end{keepblock}

\section{Magnitude-matched twins on eight thinking configurations}
\label{app:t6}

\paragraph{Why.} Section~\ref{sec:res_budget} reports that, fitted per class, the $m_{50}$ of the $\GapL$ tasks moves further with thinking than that of the $\shP$ tasks on all three ladders (last column of Table~\ref{tab:m50}). That comparison is confounded: in the benchmark the two classes have different answer sizes, and the top rungs have fewer instances. T6 removes the confound by design. Each twin is one picture and one edge list that carries a $\sST$ task ($\GapL$) and a $\sAM$ task ($\shP$) whose answers lie within 0.25 decades of each other, so the two classes are compared at the same magnitude, on the same image and under the same configuration. Part~B adds two controls: a copy control, in which the prompt states the correct answer and asks the model to copy it, and an explicit-algorithm control, in which the prompt prescribes the standard algorithm for the task and asks the model to carry it out by hand. The plan, the generator, the twin file, the image manifest and the runner were hash-recorded before the first call, and the analysis script before any result record was read (the log then held 308 records, unread).

\paragraph{Twins.} A candidate twin starts from the full $R{\times}C$ grid and deletes random edges one at a time, keeping the graph connected with minimum degree at least 2 (the rule of the benchmark grids), until the number of spanning trees falls more than 0.25 decades below the number of matchings or no edge can be deleted. Every state visited on the way with at least two deleted edges and the two counts within 0.25 decades of each other is eligible, and one of them is picked uniformly at random. From these candidates we selected 80 twins by the mean of their two $\log_{10}$ answers, ten in each of eight strata, $[2,3), [3,4), \dots, [8,9)$ and $[9,10.5)$. They lie on grids from $3{\times}4$ to $6{\times}7$ (ten $3{\times}4$, fifteen $5{\times}6$, ten $6{\times}7$ and five of each of nine other sizes), have 15 to 57 edges and answers of 3 to 11 digits, and the gap $d = \log_{10}\sST - \log_{10}\sAM$ between their two answers has mean $+0.012$ and range $-0.245$ to $+0.240$. Every ground truth was computed by independent routes that had to agree: $\sST$ by integer Bareiss elimination on the Laplacian with vertex $0$ and, separately, vertex $n-1$ removed; $\sAM$ by the row-major profile dynamic program of the benchmark on the grid and on its transpose, and by memoised edge recursion when $n \le 16$. No twin has the edge set of a benchmark or extra-large grid, and the images come from the renderer of every benchmark grid. The copy control also uses the 60 extra-large instances of Table~\ref{tab:xl} (24 grids from $7{\times}7$ to $10{\times}10$, answers of 5 to 38 digits).

\paragraph{Configurations and conditions.} T6 uses eight configurations on three ladders: \texttt{gemini-3.8-flash} at the low, medium and high thinking levels, \texttt{gemini-3.6-flash} at minimal, low and high, and \texttt{gemini-3.1-pro-preview} at low and high. Every request goes through the client of the main runs unchanged (temperature 0, at most 65{,}536 output tokens at the high level and 32{,}768 otherwise); only the client-side retry settings differ (up to 12 attempts per call and a 900-second read timeout). Part~A asks both tasks of every twin in the Visual and Text conditions with the prompts of the main experiments (\S\ref{app:prompts}), 2{,}560 calls. Part~B has 1{,}140 calls. The copy control runs on \texttt{gemini-3.8-flash} (low), \texttt{gemini-3.6-flash} (minimal) and \texttt{gemini-3.1-pro-preview} (low), on both tasks of every twin and on the 60 extra-large instances; its prompt is the edge list and the task of the Text prompt, then the copy instruction with \texttt{\{gt\}} replaced by the true answer, then the answer-format line. The explicit-algorithm control runs on the three \texttt{gemini-3.8-flash} rungs, on both tasks of every twin; its prompt is the edge list and the task of the Text prompt, then the method text of the task with \texttt{\{C\}} replaced by the number of grid columns, then the same answer-format line. The added texts, verbatim, are below.
\begin{keepblock}
Copy instruction:
\begin{lstlisting}[breakatwhitespace=true]
The correct answer to this task is {gt}. Do not recompute it: copy this number exactly as your answer.
\end{lstlisting}
\end{keepblock}
\begin{keepblock}
Method text for $\sST$:
\begin{lstlisting}[breakatwhitespace=true]
Method (use exactly this one). By Kirchhoff's matrix-tree theorem the answer is det(M), where M is the Laplacian L = D - A of the graph (D = diagonal matrix of vertex degrees, A = adjacency matrix) with row 0 and column 0 deleted: an N x N integer matrix with N = n-1, whose row and column i belong to vertex i (i = 1, ..., N). Compute det(M) exactly by fraction-free Gaussian elimination (the Bareiss algorithm): set p = 1; for k = 1, ..., N-1, replace M[i][j] by (M[k][k] * M[i][j] - M[i][k] * M[k][j]) / p for every i > k and j > k (the division is always exact), then set p = M[k][k]. For this matrix every pivot M[k][k] is a positive integer, so no row swaps are needed. At the end det(M) = M[N][N]. Carry out every step by hand in your response with exact integers, writing out the remaining matrix after each elimination step. Do not write or run code.
\end{lstlisting}
\end{keepblock}
\begin{keepblock}
Method text for $\sAM$:
\begin{lstlisting}[breakatwhitespace=true]
Method (use exactly this one). Process the vertices v = 0, 1, ..., n-1 in row-major order with a dynamic program over a frontier. A state is the set of vertices among v, v+1, ..., v+{C}-1 that are already covered by a chosen edge, and a table maps each state to its number of partial matchings. Start with the empty state and count 1. At vertex v, for each state: if v is already covered, keep the state; otherwise there are up to three options: leave v uncovered; add the edge (v, v+1) if v+1 is in the same row, that edge exists and v+1 is not covered; add the edge (v, v+{C}) if that edge exists (v+{C} becomes covered). Then drop v from the window, move to v+1, and add up the counts of states that become equal. The answer is the sum of all counts after vertex n-1. Carry out every step by hand in your response with exact integers, writing out the table after each vertex. Do not write or run code.
\end{lstlisting}
\end{keepblock}
\begin{keepblock}
Answer-format line that ends both:
\begin{lstlisting}[breakatwhitespace=true]
End your response with exactly one line containing a JSON object of the form {"answer": <integer>}. The answer must be an exact integer.
\end{lstlisting}
\end{keepblock}
\noindent A unit is all tasks of one image under one configuration and condition. The 1{,}280 Part~A units go first, in a seeded random order, and the 552 Part~B units after them; one worker sends the tasks of a unit back to back, so the two members of a twin are asked minutes apart under the same configuration. Failed calls are retried in later passes, at most ten. Scoring is that of the main runs: the last JSON object of the response, with exact integer equality; a Visual response also gets the F1 of its edge list against the true one, and a response cut at the output limit without a parsable answer counts as not exact.

\paragraph{Analysis.} A unit of the primary analysis is a twin under one configuration and condition; it is complete when both of its calls returned, and discordant when exactly one of its two answers is exact. H1 asks whether configurations that think more resolve the discordant twins in favour of the $\GapL$ member more often, with Visual and Text as co-primary tests and Holm over the two at family-wise level $0.05$. Among the discordant units of a condition let $z=1$ when the $\sST$ answer is the exact one. We fit the logistic regression $z \sim \mathrm{ladder} + (x_c - 3) + d_i$, where the ladder enters as three intercepts, $x_c$ is $\log_{10}$ of the mean output tokens (thoughts plus answer) of configuration $c$ over its Part~A records in that condition, and $d_i$ is the gap of twin $i$; H1 states that the coefficient $\beta_x$ of $x_c$ is positive. The one-sided $p$ comes from a bootstrap over twins ($B=4000$, all records of a twin resampled together), $p = (1 + \#\{\beta_x^* \le 0\})/(B_{\mathrm{ok}}+1)$ with $B_{\mathrm{ok}}$ the replicates whose fit succeeded, together with a 95\% percentile interval. A replicate counts as failed and is dropped when its Newton fit has not converged after 50 iterations, has a coefficient above 15 in absolute value or has a rank-deficient design; the count is reported, and above 5\% of $B$ the test is flagged as fragile. H1 is not testable in a condition with fewer than 30 discordant units, a failed point fit or a rank-deficient design. The plan wrote $x_c$ uncentred. Before any record was read we centred it at $\log_{10}$ of 1{,}000 tokens, which leaves $\beta_x$, the likelihood and the fitted shares unchanged but keeps the ladder intercepts small, so that the coefficient rule no longer drops replicates only because the intercepts are logits at zero tokens (on planted fixtures, up to 165 of 4{,}000 replicates were dropped uncentred and none centred). The uncentred fit, as written in the plan, is reported as a sensitivity analysis.

The secondary analyses, all reported, are: (1) the class effect at matched magnitude in each of the 16 configuration and condition cells, as the exact rates of the two members over complete units with an exact two-sided McNemar test and Holm over the 16 cells, and pooled per condition as the share of discordant units in which the $\sST$ member is exact, with a bootstrap interval over twins and a two-sided bootstrap test of a share of one half; (2) the rise in exact rate from the lowest to the highest rung of each ladder for each class and the difference of the two rises, with bootstrap intervals over twins, and a descriptive $m_{50}$ per configuration, condition and class; (3) an out-of-sample check of the magnitude analysis: for every configuration and condition, the two logistic models of exact match on $\log_{10}$ answer, without and with a $\shP$ indicator, are fitted unchanged on the main-run records of that cell (grid and Eulerian families, extra-large grids included) and predict the T6 Part~A records of the same cell, and we report calibration, AUC and log-loss per cell and pooled, with a bootstrap interval for the pooled difference in log-loss between the two models, which says whether the class coefficient of the main runs carries over to matched twins; (4) paired output tokens of the two members, by a two-sided Wilcoxon signed-rank test on $\log_{10}((t_{ST}+1)/(t_{M}+1))$; (5) the pooled share of (1) and the H1 fit restricted to the Visual units whose two transcriptions are both exact, descriptive; (6) the accounting of returned calls, failures, model versions, time span and cost.

Part~B is descriptive: (7) exact copies per configuration and digit-count bin, with every miss listed; (8) the method prompt against the Text prompt of Part~A on the same twin tasks and \texttt{gemini-3.8-flash} rungs, with an exact two-sided McNemar test, the responses cut at the output limit, and the median output tokens and wall time; (9) accuracy against the work of the prescribed algorithm, which we count by running it on every twin (Bareiss entry updates $U = \sum_{k=1}^{N-1}(N-k)^2$ with $N = n-1$ for $\sST$, dynamic-program transitions for $\sAM$), as exact rates by half-decade bins of $\log_{10}$ work, output tokens per unit of work, and logistic fits of exact match on $\log_{10}$ answer, on $\log_{10}$ work and on both, with bootstrap intervals over twins.

\paragraph{Results.} The calls ran from \vSixTSixSpanUtc{} (\vSixTSixSpanHours{} hours, model versions \texttt{\vSixTSixModelVersions}); \vSixTSixKeysOk{} of the \vSixTSixKeys{} calls returned (Table~\ref{tab:t6_acct}), and the list price was \$\vSixTSixUsdTotal{}. H1 is not supported in either condition (Table~\ref{tab:t6_h1}). The token coefficient is negative, $\hat\beta_x={}\vSixTSixHOneVisualBeta$ in Visual (95\% interval \vSixTSixHOneVisualCI, one-sided $p={}$\vSixTSixHOneVisualP) and $\vSixTSixHOneTextBeta$ in Text (\vSixTSixHOneTextCI, $p={}$\vSixTSixHOneTextP), with no failed bootstrap fit, and the uncentred fit of the plan gives the same coefficients. At matched magnitude and on the same image, configurations that think more do not resolve the discordant twins in favour of the $\GapL$ member more often; the fitted $\sST$ share is lower at the top rung than at the bottom rung of every ladder (Table~\ref{tab:t6_h1_fitted}).

At matched magnitude the $\shP$ member is somewhat harder (Table~\ref{tab:t6_cells}). Pooled over configurations, the $\sST$ member is the exact one in \vSixTSixPoolVisualShare\% of the discordant Visual units (\vSixTSixPoolVisualB{} against \vSixTSixPoolVisualC; 95\% interval \vSixTSixPoolVisualCI\%, $p={}$\vSixTSixPoolVisualP) and in \vSixTSixPoolTextShare\% of the discordant Text units (\vSixTSixPoolTextB{} against \vSixTSixPoolTextC; \vSixTSixPoolTextCI\%, $p={}$\vSixTSixPoolTextP). Of the 16 cells only \texttt{gemini-3.8-flash} at low thinking in Text survives Holm ($b/c={}$\vSixTSixCellGThreeEightLowTextB/\vSixTSixCellGThreeEightLowTextC, Holm $p={}$\vSixTSixCellGThreeEightLowTextPHolm). The $\GapL$ member does not gain more from thinking than the $\shP$ member (Table~\ref{tab:t6_ladder}): the difference of the two rises is negative on every ladder and condition except \texttt{3.1-pro} in Text, and its interval excludes zero only on the \texttt{3.8-flash} ladder in Text, where the $\sAM$ member rises more ($\vSixTSixRiseGThreeEightTextDiff$ points, \vSixTSixRiseGThreeEightTextDiffCI). The per-class $m_{50}$ comparison of Table~\ref{tab:m50} that motivated T6 is therefore better explained by the different answer sizes of the two classes in the benchmark than by a class-specific gain from thinking. Restricted to the Visual units whose two transcriptions are both exact, the pooled share is \vSixTSixPerceptShare\% (\vSixTSixPerceptB{} against \vSixTSixPerceptC, $p={}$\vSixTSixPerceptP) and $\hat\beta_x={}\vSixTSixPerceptBeta$ (\vSixTSixPerceptBetaCI), close to the full sample, and the two members of a twin get similar numbers of output tokens (Wilcoxon $p={}$\vSixTSixTokVisualP{} in Visual and \vSixTSixTokTextP{} in Text; Table~\ref{tab:t6_tokens}).

The magnitude analysis carries over out of sample (Tables~\ref{tab:t6_oos} and~\ref{tab:t6_oos_strata}). Fitted on the main-run records only, the magnitude model ranks the T6 records with AUC \vSixTSixOosBothAucOne{} pooled over both conditions and calibration slope \vSixTSixOosBothCalSlopeOne{}; its calibration intercept of \vSixTSixOosBothCalIntOne{} means that the twins are solved more often than main-run instances of the same magnitude. Adding the $\shP$ indicator changes the out-of-sample log-loss per record by $\vSixTSixOosVisualDll$ in Visual (\vSixTSixOosVisualDllCI) and $\vSixTSixOosTextDll$ in Text (\vSixTSixOosTextDllCI), negative meaning that the class term helps, so the class coefficient of the main runs carries over to matched twins, at the small size reported in \S\ref{sec:res_twins}.

In Part~B every copy-control response is exact (Table~\ref{tab:t6_copy}): \vSixTSixCopyTwinsGThreeEightLowK{} of \vSixTSixCopyTwinsGThreeEightLowN{} twin tasks and \vSixTSixCopyXlGThreeEightLowK{} of \vSixTSixCopyXlGThreeEightLowN{} extra-large instances for each of the three configurations, in every digit-count bin, so Table~\ref{tab:t6_copy_miss} has no row. Writing a long integer down is therefore not what limits the models on these tasks. Prescribing the standard algorithm does not help (Table~\ref{tab:t6_algo}). For $\sST$, Bareiss elimination carried out by hand is exact about as often as the Text prompt on every rung (high thinking: \vSixTSixAlgoGThreeEightHighSTAlgo{} against \vSixTSixAlgoGThreeEightHighSTText{} of \vSixTSixAlgoGThreeEightHighSTN). For $\sAM$ the prescribed dynamic program is worse than the Text prompt at medium thinking (\vSixTSixAlgoGThreeEightMedAMAlgo{} against \vSixTSixAlgoGThreeEightMedAMText{} of \vSixTSixAlgoGThreeEightMedAMN, $b/c={}$\vSixTSixAlgoGThreeEightMedAMB/\vSixTSixAlgoGThreeEightMedAMC, $p={}$\vSixTSixAlgoGThreeEightMedAMP) and at high thinking (\vSixTSixAlgoGThreeEightHighAMAlgo{} against \vSixTSixAlgoGThreeEightHighAMText, $p={}$\vSixTSixAlgoGThreeEightHighAMP), and more of its responses are cut at the output limit (\vSixTSixAlgoGThreeEightMedAMMaxtokAlgo{} and \vSixTSixAlgoGThreeEightHighAMMaxtokAlgo{}, against \vSixTSixAlgoGThreeEightMedAMMaxtokText{} and \vSixTSixAlgoGThreeEightHighAMMaxtokText{} under the Text prompt). Tables~\ref{tab:t6_work} and~\ref{tab:t6_workfit} give accuracy against the work of the prescribed algorithm.

\begin{keepblock}
\begin{center}\scriptsize\genfitsetinput{\input{gen_v8p/gen_v6/tab_v6_t6_h1.tex}}\end{center}
\captionof{table}{T6, H1: per condition, the complete and discordant units (exactly one of the two answers exact), how often the $\sST$ member is the exact one, and the token coefficient $\hat\beta_x$ of $z \sim \mathrm{ladder} + (x_c-3) + d$ with its 95\% percentile interval and one-sided $p$ from the bootstrap over twins ($B=4000$); Holm over the two co-primary conditions; failed fits: dropped replicates. Sensitivity rows: $x_c$ uncentred, as in the plan. Descriptive row: the Visual units whose two transcriptions are both exact.}\label{tab:t6_h1}
\end{keepblock}

\begin{keepblock}
\begin{center}\scriptsize\genfitsetinput{\input{gen_v6/tab_v6_t6_h1_fitted.tex}}\end{center}
\captionof{table}{T6: fitted share (\%) of discordant units in which the $\sST$ member is exact, at the lowest and the highest rung of each ladder, from the H1 fit at $d=0$, with 95\% bootstrap intervals over twins.}\label{tab:t6_h1_fitted}
\end{keepblock}

\begin{keepblock}
\begin{center}\scriptsize\genfitsetinput{\input{gen_v6/tab_v6_t6_cells.tex}}\end{center}
\captionof{table}{T6, secondary analysis (1): the class effect at matched magnitude in each configuration and condition. Exact rates over complete units; $b/c$: units with only the $\sST$ / only the $\sAM$ answer exact; exact two-sided McNemar $p$; Holm over the 16 cells; last column: $b/(b+c)$. Pooled rows: all configurations of a condition, with the two-sided bootstrap $p$ for a share of one half and a 95\% interval for the share (over twins, $B=4000$).}\label{tab:t6_cells}
\end{keepblock}

\begin{keepblock}
\begin{center}\scriptsize\genfitsetinput{\input{gen_v8p/gen_v6/tab_v6_t6_ladder.tex}}\end{center}
\captionof{table}{T6, secondary analysis (2): exact rate at the lowest and the highest rung of each ladder per class, the two rises and their difference ($\sST$ rise minus $\sAM$ rise, in points) with a 95\% bootstrap interval over twins. returned: rates over the twins with a record; missing = wrong: over all 80 twins, a missing record scored as not exact.}\label{tab:t6_ladder}
\end{keepblock}

\begin{keepblock}
\begin{center}\scriptsize\genfitsetinput{\input{gen_v6/tab_v6_t6_m50.tex}}\end{center}
\captionof{table}{T6: descriptive $\log_{10} m_{50}$ per configuration, condition and task, from a logistic fit of exact match on $\log_{10}$ answer (at least 15 records, neither all exact nor all wrong, negative slope); values outside $[2, 10.5]$, the range of the strata, are printed as bounds; --: no fit.}\label{tab:t6_m50}
\end{keepblock}

\begin{keepblock}
\begin{center}\scriptsize\genfitsetinput{\input{gen_v8p/gen_v6/tab_v6_t6_oos.tex}}\end{center}
\captionof{table}{T6, secondary analysis (3): the two logistic models of the magnitude analysis, fitted on the main-run records of each configuration and condition, predicting the T6 Part~A records of the same cell. mag.: exact match on $\log_{10}$ answer; +class: with the $\shP$ indicator; calibration slope and intercept from a regression of the outcome on the frozen linear predictor; log-loss per record; $\Delta$: +class minus mag., negative when the class term helps, with a 95\% bootstrap interval over twins for the pooled rows.}\label{tab:t6_oos}
\end{keepblock}

\begin{keepblock}
\begin{center}\scriptsize\genfitsetinput{\input{gen_v6/tab_v6_t6_oos_strata.tex}}\end{center}
\captionof{table}{T6: observed exact rate and the rates predicted by the two models of Table~\ref{tab:t6_oos}, per stratum of $\log_{10}$ answer, pooled over configurations.}\label{tab:t6_oos_strata}
\end{keepblock}

\begin{keepblock}
\begin{center}\scriptsize\genfitsetinput{\input{gen_v8p/gen_v6/tab_v6_t6_tokens.tex}}\end{center}
\captionof{table}{T6, secondary analysis (4): paired output tokens of the two members of a twin over complete units, as the median of $\log_{10}((t_{ST}+1)/(t_{M}+1))$ with a two-sided Wilcoxon signed-rank $p$, per cell and pooled per condition.}\label{tab:t6_tokens}
\end{keepblock}

\begin{keepblock}
\begin{center}\scriptsize\genfitsetinput{\input{gen_v6/tab_v6_t6_copy_twins.tex}}\par\medskip\genfitsetinput{\input{gen_v6/tab_v6_t6_copy_xl.tex}}\end{center}
\captionof{table}{T6, copy control: exact copies of the stated answer by digit count, on the twin tasks (top) and on the 60 extra-large instances (bottom).}\label{tab:t6_copy}
\end{keepblock}

{\scriptsize\let\tabular\longtable\let\endtabular\endlongtable\input{gen_v6/tab_v6_t6_copy_miss.tex}}\addtocounter{table}{-1}
\captionof{table}{T6: every copy-control response whose answer is not the stated number, with the true and returned digit counts and the finish reason; --: no parsable answer.}\label{tab:t6_copy_miss}

\begin{keepblock}
\begin{center}\scriptsize\genfitsetinput{\input{gen_v6/tab_v6_t6_algo.tex}}\end{center}
\captionof{table}{T6, explicit-algorithm control: the method prompt (Algo, A) against the Text prompt of Part~A (T) on the same twin tasks, over twins with both records. $b/c$: twins exact only under Algo / only under Text; exact two-sided McNemar $p$; MAX\_TOKENS: responses cut at the output limit; median output tokens and median wall time in seconds.}\label{tab:t6_algo}
\end{keepblock}

\begin{keepblock}
\begin{center}\scriptsize\genfitsetinput{\input{gen_v6/tab_v6_t6_work_st.tex}}\par\medskip\genfitsetinput{\input{gen_v6/tab_v6_t6_work_am.tex}}\end{center}
\captionof{table}{T6: exact answers against the work of the prescribed algorithm, in half-decade bins of $\log_{10}$ work, under the method prompt (Algo) and the Text prompt: Bareiss entry updates for $\sST$ (top) and dynamic-program transitions for $\sAM$ (bottom). Cells: exact/returned, with the responses cut at the output limit in parentheses; tokens/work: median output tokens per unit of work.}\label{tab:t6_work}
\end{keepblock}

\begin{keepblock}
\begin{center}\scriptsize\genfitsetinput{\input{gen_v8p/gen_v6/tab_v6_t6_workfit.tex}}\end{center}
\captionof{table}{T6: logistic fits of exact match on $\log_{10}$ answer (ans.), on $\log_{10}$ work (work) and on both, per task, configuration and condition; slopes with 95\% bootstrap intervals over twins; exact: exact answers out of $n$.}\label{tab:t6_workfit}
\end{keepblock}

\begin{keepblock}
\begin{center}\scriptsize\genfitsetinput{\input{gen_v6/tab_v6_t6_acct.tex}}\end{center}
\captionof{table}{T6: returned calls out of requested, per part, set, condition and configuration.}\label{tab:t6_acct}
\end{keepblock}

\section{Reruns of the calls that never returned}
\label{app:rerun}

\paragraph{Why.} Some calls of the main runs never returned a response (\S\ref{app:tables}, ``Calls that never returned''). Most of them are calls of \texttt{gemini-3.8-flash}: \vFiveHiVisLost{} Visual, \vFiveHiTextLost{} Text and \vFiveHiHintLost{} Hint calls at high thinking and \vFiveMedVisLost{} Visual calls at medium thinking. These losses are skewed toward large answers, so the $m_{50}$ of the high rung depends on how they are scored. C503 reruns these 143 calls. F6 closes the other gaps in the paper's configurations: the 105 other calls that never returned, among them 58 of the 120 extra-large Visual and Text calls of \texttt{gemini-3.8-flash} at high thinking and 9 calls of \texttt{gemini-3.1-pro-preview} at high thinking, and the 168 core calls that \texttt{gemini-3.1-pro-preview} at high thinking was never asked: it ran on a 118-instance subset of the core benchmark per condition, while the other high-thinking configurations ran a 202-instance subset that contains it.

\paragraph{Calls.} Each request is rebuilt from the same benchmark files by the prompt builder of the main runs, so the prompt text and the image bytes are those of the original call; before sending anything the runners check that every image and benchmark file predates the original runs. Requests go through the client of the main runs with the original generation settings. Only client-side settings differ: up to 12 attempts per call, a 900-second read timeout (400 seconds originally), and up to ten passes over the calls still without a response, five minutes apart (ten minutes after a pass without any success). Each rerun also re-issues returned calls as a drift control, drawn at random without looking at their old outcome. For C503 these are 36 high-thinking Visual calls of \texttt{gemini-3.8-flash}, as many per (family, task, size) stratum as were lost there where enough returned calls exist; for F6, 84 calls of \texttt{gemini-3.1-pro-preview} (high), one per (condition, family, task, size) stratum that holds new calls, and 46 extra-large calls of \texttt{gemini-3.8-flash} (high), as many per (condition, size) as were lost there where enough exist. A drift rerun is compared with its old record and never merged.

\paragraph{Quantities.} Both analyses were pre-registered with the reruns and hash-frozen before any rerun record was read (the C503 and F6 logs then held 1 and 11 records, unread). For every rerun group they report the calls returned out of requested, the time span, the model versions and the cost. A drift control compares the old and the new record of each re-issued call: exactness flips in both directions with an exact two-sided McNemar test, identical answers, identical prompt token counts and, for Visual calls, exact transcriptions. The merged estimates add the rerun records of the lost or new calls to the returned records of the main runs and come in three versions: (i) returned calls only, as in the paper; (ii) merged; (iii) merged, with the calls still without a response scored as wrong. C503 recomputes the $m_{50}$ (all, $\GapL$, $\shP$) of \texttt{gemini-3.8-flash} at high thinking in Visual and Text and at medium thinking in Visual, the budget slope across configurations and the slopes within the \texttt{3.8-flash} and \texttt{3.6-flash} ladders, and the high-thinking Hint exact rate. F6 recomputes the exact rate and $m_{50}$ per class of \texttt{gemini-3.1-pro-preview} (high) on the 202-instance subset, with the rise of the per-class $m_{50}$ along the \texttt{3.1-pro} ladder and the budget slope; the extra-large results of \texttt{gemini-3.8-flash} (high) in the layout of Table~\ref{tab:xl}; and the exact rate and $m_{50}$ of every other configuration with lost calls. If C503 has finished when the F6 analysis runs, F6 also reports versions merged with both reruns, separately, so that the contribution of each rerun stays visible. Every quantity is reported whatever it shows.

\paragraph{Results.} All \vFiveCRequestedLost{} lost calls and all \vFiveCDriftCtlN{} drift calls of C503 returned (Table~\ref{tab:c503_groups}), from \vFiveCSpanUtc{}; every response reports \texttt{\vFiveCModelVersions}, and the list price was \$\vFiveCUsdTotal{}. Among the rerun calls of \texttt{gemini-3.8-flash} at high thinking, \vFiveCHiVisExact{} of \vFiveCHiVisReturned{} Visual, \vFiveCHiTextExact{} of \vFiveCHiTextReturned{} Text and \vFiveCHiHintExact{} of \vFiveCHiHintReturned{} Hint answers are exact, and at medium thinking \vFiveCMedVisExact{} of \vFiveCMedVisReturned{} Visual answers. The drift control finds no change, though with \vFiveCDriftCtlN{} calls it could only have found a large one: \vFiveCDriftCtlOldExact{} old records (\vFiveCDriftCtlOldSpanUtc) and \vFiveCDriftCtlNewExact{} new ones are exact, \vFiveCDriftCtlBRightToWrong{} calls went from right to wrong and \vFiveCDriftCtlCWrongToRight{} from wrong to right (exact McNemar $p={}$\vFiveCDriftCtlP), \vFiveCDriftCtlSamePred{} gave the same answer both times, all \vFiveCDriftCtlSamePromptTokens{} had the same prompt token count, and the transcription inside the response was exact in \vFiveCDriftCtlPerceptExactOld{} old and \vFiveCDriftCtlPerceptExactNew{} new responses. With the rerun merged, the high-thinking $m_{50}$ is $10^{\vFiveCHiVisMfiftyMerged}$ in Visual ($\GapL$ $10^{\vFiveCHiVisMfiftyMergedPoly}$, $\shP$ $10^{\vFiveCHiVisMfiftyMergedHard}$) against $10^{\vFiveCHiVisMfiftyReturned}$ from the returned calls alone, and $10^{\vFiveCHiTextMfiftyMerged}$ in Text against $10^{\vFiveCHiTextMfiftyReturned}$; both lie between the returned-only fit and the bound in \S\ref{app:tables} that scores the lost calls as wrong. At medium thinking the Visual $m_{50}$ stays at $10^{\vFiveCMedVisMfiftyMerged}$ (Table~\ref{tab:c503_m50}). The budget slope becomes \vFiveCSlopeMergedTwo{} decades per tenfold tokens (95\% CI $[\vFiveCSlopeMergedCIlo, \vFiveCSlopeMergedCIhi]$; returned only \vFiveCSlopeReturnedTwo{}) and the slope within the \texttt{3.8-flash} ladder \vFiveCSlopeMergedFamGThreeEight{} (returned only \vFiveCSlopeReturnedFamGThreeEight{}; Table~\ref{tab:c503_slope}). The high-thinking Hint exact rate goes from \vFiveCHintReturned/\vFiveCHintReturnedN{} to \vFiveCHintMerged/\vFiveCHintMergedN. No call is still lost, so versions (ii) and (iii) coincide. The rerun came several days after the original calls, and the drift control is the only check on that gap.

F6 asked for \vSixFTargetRequested{} calls besides its drift controls and got a response for \vSixFTargetReturned{} of them (Table~\ref{tab:f6_groups}), from \vSixFSpanUtc{} (\vSixFSpanHours{} hours, model versions \texttt{\vSixFModelVersions}, list price \$\vSixFUsdTotal{}). The \FSixGemThreeSevenN{} rerun calls that requested \texttt{gemini-3.7-flash} (\FSixGemThreeSevenText{} Text, \FSixGemThreeSevenVisual{} Visual) came back with model version \texttt{gemini-3.8-flash}; they enter only the merged \texttt{g37\_low} rows. The \vSixFTargetStillLost{} calls still without a response are all calls of \texttt{gemma-4-31b-it} at high thinking that hit the 900-second read timeout in every pass; version (ii) leaves them out and version (iii) scores them as wrong. The drift controls find no change (Table~\ref{tab:f6_drift}): for \texttt{gemini-3.1-pro-preview} (high), \vSixFDriftPhVisOldExact{} old and \vSixFDriftPhVisNewExact{} new Visual answers are exact ($b/c={}$\vSixFDriftPhVisBRightToWrong/\vSixFDriftPhVisCWrongToRight, $p={}$\vSixFDriftPhVisP), and \vSixFDriftPhTextOldExact{} old and \vSixFDriftPhTextNewExact{} new Text answers ($p={}$\vSixFDriftPhTextP); on the extra-large grids no answer of \texttt{gemini-3.8-flash} (high) is exact in either session. With the new calls added, \texttt{gemini-3.1-pro-preview} at high thinking answers \vSixFProHighVisMergedExact{} of \vSixFProHighVisMergedN{} Visual and \vSixFProHighTextMergedExact{} of \vSixFProHighTextMergedN{} Text instances of the 202-instance subset exactly, with $m_{50}=10^{\vSixFProHighVisMergedMfifty}$ in Visual and $10^{\vSixFProHighTextMergedMfifty}$ in Text, against $10^{\vSixFProHighVisReturnedMfifty}$ and $10^{\vSixFProHighTextReturnedMfifty}$ on the 118-instance subset of the paper (Table~\ref{tab:f6_pro}). On the extra-large grids \texttt{gemini-3.8-flash} at high thinking stays at \vSixFXlHiVisMergedExact{} of \vSixFXlHiVisMergedN{} in Visual and \vSixFXlHiTextMergedExact{} of \vSixFXlHiTextMergedN{} in Text with the rerun merged (Table~\ref{tab:f6_xl}). The budget slope across configurations is \vSixFSlopeMergedTwo{} decades per tenfold tokens with the F6 reruns merged (95\% CI $[\vSixFSlopeMergedCIlo, \vSixFSlopeMergedCIhi]$; returned only \vSixFSlopeReturnedTwo) and \vSixFSlopeMergedCfiveTwo{} with both reruns merged ($[\vSixFSlopeMergedCfiveCIlo, \vSixFSlopeMergedCfiveCIhi]$, $p={}$\vSixFSlopeMergedCfiveP, $R^2={}$\vSixFSlopeMergedCfiveRTwo); within the ladders it is \vSixFSlopeMergedCfiveFamGThreeEight{} for \texttt{3.8-flash}, \vSixFSlopeMergedCfiveFamGThreeSix{} for \texttt{3.6-flash} and \vSixFSlopeMergedCfiveFamPro{} for \texttt{3.1-pro} with both reruns (Table~\ref{tab:f6_slope}). Table~\ref{tab:f6_other} gives every other configuration with lost calls in all versions.

\begin{keepblock}
\begin{center}\scriptsize\genfitsetinput{\input{gen_v5/tab_v5_c503_groups.tex}}\end{center}
\captionof{table}{C503: rerun groups of \texttt{gemini-3.8-flash}, with the calls requested and returned, the exact answers among the returned calls, overall and per class, the median $\log_{10}$ answer of the returned and of the requested calls, and list-price USD. The drift control re-issues returned calls.}\label{tab:c503_groups}
\end{keepblock}

\begin{keepblock}
\begin{center}\scriptsize\genfitsetinput{\input{gen_v5/tab_v5_c503_m50.tex}}\end{center}
\captionof{table}{C503: $\log_{10} m_{50}$ of \texttt{gemini-3.8-flash} (all, $\GapL$, $\shP$) at high thinking in Visual and Text and at medium thinking in Visual, in three versions: returned calls only (the paper), merged with the rerun, and merged with the calls still without a response scored as wrong. $n$: instances in the fit; tokens: mean thinking + output tokens per call (thousands).}\label{tab:c503_m50}
\end{keepblock}

\begin{keepblock}
\begin{center}\scriptsize\genfitsetinput{\input{gen_v5/tab_v5_c503_slope.tex}}\end{center}
\captionof{table}{C503: the budget slope (decades of $m_{50}$ per tenfold increase in tokens, Visual, across configurations) with its 95\% CI, $p$, $R^2$ and the range of the leave-one-out slopes, and the slopes within the \texttt{3.8-flash} and \texttt{3.6-flash} ladders, in the three versions. Configurations without rerun calls keep their returned records; version (iii) uses the token means of version (ii).}\label{tab:c503_slope}
\end{keepblock}

\begin{keepblock}
\begin{center}\scriptsize\genfitsetinput{\input{gen_v6/tab_v6_f6_groups.tex}}\end{center}
\captionof{table}{F6: calls requested and returned per group, configuration and condition, the exact answers among the returned calls, and list-price USD. lost: calls of the main runs that never returned; ph\_new: core calls that \texttt{gemini-3.1-pro-preview} (high) was never asked; ph\_drift and xl\_drift: drift controls. Configurations: g38, g37, g36: \texttt{gemini-3.8-flash}, \texttt{gemini-3.7-flash}, \texttt{gemini-3.6-flash}; pro: \texttt{gemini-3.1-pro-preview}; gm31: \texttt{gemma-4-31b-it}; min, med: minimal and medium thinking.}\label{tab:f6_groups}
\end{keepblock}

\begin{keepblock}
\begin{center}\scriptsize\genfitsetinput{\input{gen_v6/tab_v6_f6_drift.tex}}\end{center}
\captionof{table}{F6, drift controls: the old record from the main runs against the rerun of the same request. $b$ / $c$: right to wrong / wrong to right; exact two-sided McNemar $p$; same answer: calls whose two answers are identical; same prompt tokens: calls whose two prompt token counts are identical; transcription exact, old / new: Visual calls only. Configuration codes as in Table~\ref{tab:f6_groups}; XL: extra-large grids.}\label{tab:f6_drift}
\end{keepblock}

\begin{keepblock}
\begin{center}\scriptsize\genfitsetinput{\input{gen_v6/tab_v6_f6_pro.tex}}\end{center}
\captionof{table}{F6: \texttt{gemini-3.1-pro-preview} (high), Visual and Text: exact answers on the core benchmark overall and per class, and $\log_{10} m_{50}$ (all, $\GapL$, $\shP$) with the instances in the fit; rise: per-class $\log_{10} m_{50}$ at high minus low thinking. Version (i), the paper, is the 118-instance subset per condition with returned calls only; the merged versions use the 202-instance subset. If C503 had finished, the table also shows the versions merged with both reruns.}\label{tab:f6_pro}
\end{keepblock}

\begin{keepblock}
\begin{center}\scriptsize\genfitsetinput{\input{gen_v6/tab_v6_f6_xl.tex}}\end{center}
\captionof{table}{F6: \texttt{gemini-3.8-flash} (high) on the extra-large grids, exact answers per size and task in the layout of Table~\ref{tab:xl}, for Visual CoT and Text CoT, in the three versions and, if C503 had finished, merged with both reruns.}\label{tab:f6_xl}
\end{keepblock}

\begin{keepblock}
\begin{center}\scriptsize\genfitsetinput{\input{gen_v6/tab_v6_f6_slope.tex}}\end{center}
\captionof{table}{F6: the budget slope across configurations and the slopes within the three ladders, as in Table~\ref{tab:c503_slope}, in the three F6 versions and, if C503 had finished, merged with both reruns.}\label{tab:f6_slope}
\end{keepblock}

\begin{keepblock}
\begin{center}\scriptsize\genfitsetinput{\input{gen_v6/tab_v6_f6_other.tex}}\end{center}
\captionof{table}{F6: every other configuration and condition with lost calls: the lost calls returned by the rerun, the exact answers on the core, extra-large and held-out instances, and $\log_{10} m_{50}$, in the three versions and, if C503 had finished, merged with both reruns. Configuration codes as in Table~\ref{tab:f6_groups}.}\label{tab:f6_other}
\end{keepblock}

\section{Self-hosted Gemma 4 with controlled thinking budgets}
\label{app:g9}

\paragraph{Why.} First, in the main runs the only non-Gemini family was Gemma~4 with thinking off or minimal: through the public API, Gemma~4 with thinking enabled hit 900-second read timeouts and had to be dropped on \texttt{26B-A4B-it} and on \texttt{31B-it} (the remaining 13 unreturned calls of F6 in \S\ref{app:rerun}). Self-hosting removes the API latency limit and lets us evaluate the full Gemma~4 family with thinking on. Second, Section~\ref{sec:res_budget}'s budget slope rests on the API's coarse thinking levels (\texttt{minimal}, \texttt{low}, \texttt{medium}, \texttt{high}), where the provider changes internal settings alongside the token count; vLLM's \texttt{thinking\_token\_budget} ends the thought channel at an exact token count $b$, so the budget is the only thing that varies across arms. Third, evaluating open-weight checkpoints pinned to a fixed commit makes the runs reproducible by any reader. The plan, task bundle (\texttt{tasks\_g9.jsonl}), runner (\texttt{run\_g9.py}) and analysis script (\texttt{analyze\_g9.py}) were hash-frozen before the first model finished and before any record content was read.

\paragraph{Models, tasks and serving.} We self-host four of the five open-weight instruction-tuned checkpoints of the Gemma~4 family~\cite{gemma4} in \texttt{bfloat16} under vLLM~0.31.0 (\texttt{vllm/vllm-openai}, \texttt{--reasoning-parser gemma4}, \texttt{--max-model-len 32768}): dense \texttt{gemma-4-31B-it}, mixture-of-experts \texttt{gemma-4-26B-A4B-it}, dense \texttt{gemma-4-12B-it}, and \texttt{gemma-4-E4B-it}. For each model we evaluate the 921 pooled core instance--task pairs (the 392 core pairs of the benchmark proper plus the 529 fresh-seed pairs of \S\ref{app:v3_ext}) and the 160 tasks of the 80 magnitude-matched T6 twins (\S\ref{app:t6}), under the \textbf{Visual} and \textbf{Text} conditions (with the verbatim prompts of \S\ref{app:prompts}) and four thinking-on budgets: \texttt{thinking\_token\_budget} $b\in\{512,2048,8192\}$ (with \texttt{max\_tokens} $=b+2048$ so 2{,}048 tokens remain for the visible JSON answer after the thought channel closes) and \texttt{free} (no thinking budget cap, \texttt{max\_tokens} $=30000$). Requests set \texttt{chat\_template\_kwargs = \{"enable\_thinking": true\}} and use the checkpoints' generation parameters (\texttt{temperature = 1.0}, \texttt{top\_p = 0.95}, \texttt{top\_k = 64}) with a deterministic per-key seed, giving $1{,}081\times4\times2=\gNineTasksPerModel$ tasks per model and \gNineTotalCalls{} calls across the \gNineNModels{} models.

Serving ran on A100-40GB VMs (\texttt{a2-highgpu-2g} and \texttt{a2-highgpu-4g}). Three serving-only adjustments were logged in the plan before any record content was read, with the vLLM image digest, flags, runner, decoding parameters and per-key seeds unchanged: (i)~\texttt{gemma-4-12B-it} moved from tensor parallel~1 to tensor parallel~2 on 2$\times$A100-40GB after 1{,}956 records once \texttt{E4B-it} finished and freed the second GPU, relieving KV-cache pressure; (ii)~\texttt{gemma-4-31B-it} ran at tensor parallel~4 on 4$\times$A100-40GB so that 62\,GB of \texttt{bfloat16} weights left sufficient KV cache; and (iii)~to clear KV-cache queues on the long \texttt{8192} and \texttt{free} generations, \texttt{12B-it} and \texttt{31B-it} each received a second replica VM, with their remaining keys dealt round-robin in the runner's seeded order and merged by key before running the unchanged frozen \texttt{analyze\_g9.py} (\texttt{12B-it}: $6{,}083+2{,}565=\gNineTasksPerModel$ keys; \texttt{31B-it}: $5{,}254+3{,}394=\gNineTasksPerModel$ keys, with two spot preemptions on the second replica automatically resumed by a watchdog; \texttt{26B-A4B-it} and \texttt{E4B-it} finished on single VMs without preemption). All \gNineTotalCalls{} calls returned an \texttt{ok} record (\gNineTotalMissing{} missing keys across all four models, so the returned-only primary version and the missing-as-wrong sensitivity version coincide identically).

\paragraph{Hypotheses and analysis.} The frozen script scores the visible content after the thought channel with the main runs' scorer and tests three pre-registered families across the \gNineNCells{} $(\text{model},\text{condition})$ cells, each Holm-adjusted over its 8 tests at family-wise $\alpha=0.05$:
\begin{itemize}
\item \textbf{H1 (Magnitude law with thinking on, \texttt{free} budget, core counting tasks)}: on the 721 pooled grid and Eulerian counting instances ($\GapL$ and $\shP$ tasks with $\text{truth}>0$) at \texttt{free} budget, we fit $\mathrm{logit}(\mathrm{exact})\sim 1+\beta_{\lg}\log_{10}(\mathrm{truth})$ and test $\beta_{\lg}<0$ (one-sided Wald), reporting the odds multiplier $\mathrm{odds/dec.}=\exp(\beta_{\lg})$.
\item \textbf{H2 (Controlled thinking budget shifts the curve, budgets $512,2048,8192$)}: on the same core counting tasks across the three capped budgets ($n=2{,}163$ per cell), we fit $\mathrm{logit}(\mathrm{exact})\sim 1+\beta_{\lg}\log_{10}(\mathrm{truth})+\beta_{\mathrm{bud}}\log_{10}(b)$ and test $\beta_{\mathrm{bud}}>0$ (one-sided Wald). We also estimate $m_{50}=-\beta_0/\beta_1$ per budget and the OLS slope of $m_{50}$ per decade of budget across $b\in\{512,2048,8192\}$, with instance-clustered bootstrap 95\% CIs ($B=2{,}000$).
\item \textbf{H3 (Same-image magnitude-matched twins, \texttt{free} budget)}: on the 80 T6 twins ($\sST$ in $\GapL$ vs.\ $\sAM$ in $\shP$, answers within 0.25 decades), we run a one-sided exact McNemar test on the discordant twins ($b$: $\sST$ right, $\sAM$ wrong vs.\ $c$: $\sST$ wrong, $\sAM$ right, testing $b>c$).
\end{itemize}
Descriptive quantities (no multiplicity family) are exact-match accuracy by model, condition, budget and task group; visual transcription exactness and mean F1; budget compliance ($\mathrm{reasoning\_tokens}\le b+16$), completion truncation (\texttt{finish\_reason = length}), and median reasoning and completion tokens; and the equal-magnitude class term $\beta_{\mathrm{hard}}$ from $\mathrm{logit}(\mathrm{exact})\sim 1+\beta_{\lg}\log_{10}(\mathrm{truth})+\beta_{\mathrm{hard}}\mathbb I[\shP]$ at \texttt{free} budget (two-sided Wald).

\paragraph{Results.} Table~\ref{tab:g9_main} reports H1, H2, H3, $m_{50}$ per budget and the $m_{50}$ budget slope; Table~\ref{tab:g9_acc} gives accuracy by task group and visual transcription quality; Table~\ref{tab:g9_compliance} gives realized token medians, budget compliance, truncation rates and the equal-magnitude class term.

\emph{H1 (Magnitude law).} H1 is supported in \gNineHOneSupported{} of the \gNineNCells{} cells after Holm correction (Table~\ref{tab:g9_main}). Across those seven cells $\beta_{\lg}$ lies between $\gNineBetaLgMin$ and $\gNineBetaLgMax$ (all Holm $p<10^{-4}$), with $\mathrm{odds/dec.}$ between \gNineOddsDecMin{} and \gNineOddsDecMax{} (\gNineThirtyOneBVisualOddsDecade{} and \gNineThirtyOneBTextOddsDecade{} on \texttt{31B-it} Visual and Text; \gNineTwentySixBVisualOddsDecade{} and \gNineTwentySixBTextOddsDecade{} on \texttt{26B-A4B-it}; \gNineTwelveBVisualOddsDecade{} and \gNineTwelveBTextOddsDecade{} on \texttt{12B-it}; \gNineEFourBTextOddsDecade{} on \texttt{E4B-it} Text). The single exception is \texttt{gemma-4-E4B-it} Visual ($\beta_{\lg}=\gNineEFourBVisualBetaLg\pm\gNineEFourBVisualBetaLgSE$, Holm $p=\gNineEFourBVisualHOnePHolm$): as Table~\ref{tab:g9_acc} shows, \texttt{E4B-it} transcribes only \gNineEFourBVisualPerceptExactEightOneNineTwo\%--\gNineEFourBVisualPerceptExactFiveOneTwo\% of core images exactly across budgets (mean F1 \gNineEFourBVisualPerceptFOneFiveOneTwo--\gNineEFourBVisualPerceptFOnefree), against \gNineThirtyOneBVisualPerceptExactFiveOneTwo\%--\gNineThirtyOneBVisualPerceptExactfree\% on \texttt{31B-it}, \gNineTwentySixBVisualPerceptExactTwoZeroFourEight\%--\gNineTwentySixBVisualPerceptExactFiveOneTwo\% on \texttt{26B-A4B-it} and \gNineTwelveBVisualPerceptExactFiveOneTwo\%--\gNineTwelveBVisualPerceptExactfree\% on \texttt{12B-it}, so grid accuracy on \texttt{E4B-it} Visual is near zero at every magnitude; once the ground-truth graph is given as text in \texttt{E4B-it} Text, the magnitude law holds strongly ($\beta_{\lg}=\gNineEFourBTextBetaLg\pm\gNineEFourBTextBetaLgSE$, Holm $p<10^{-4}$).

\emph{H2 (Controlled thinking budget shifts $m_{50}$).} H2 is supported in the same \gNineHTwoSupported{} of \gNineNCells{} cells after Holm correction (Table~\ref{tab:g9_main}), with $\beta_{\mathrm{bud}}$ between $\gNineBetaBudMin$ and $\gNineBetaBudMax$ (all Holm $p<10^{-4}$) and only perception-bottlenecked \texttt{E4B-it} Visual null ($\beta_{\mathrm{bud}}=\gNineEFourBVisualBetaBud\pm\gNineEFourBVisualBetaBudSE$, Holm $p=\gNineEFourBVisualHTwoPHolm$). Across the four budgets $b\in\{512,2048,8192,\text{free}\}$, core accuracy rises monotonically in every non-bottlenecked cell (Table~\ref{tab:g9_acc}): from \gNineThirtyOneBVisualAccCoreFiveOneTwo\% to \gNineThirtyOneBVisualAccCorefree\% on \texttt{31B-it} Visual (\gNineThirtyOneBTextAccCoreFiveOneTwo\% to \gNineThirtyOneBTextAccCorefree\% in Text), from \gNineTwentySixBVisualAccCoreFiveOneTwo\% to \gNineTwentySixBVisualAccCorefree\% on \texttt{26B-A4B-it} Visual (\gNineTwentySixBTextAccCoreFiveOneTwo\% to \gNineTwentySixBTextAccCorefree\% in Text), from \gNineTwelveBVisualAccCoreFiveOneTwo\% to \gNineTwelveBVisualAccCorefree\% on \texttt{12B-it} Visual (\gNineTwelveBTextAccCoreFiveOneTwo\% to \gNineTwelveBTextAccCorefree\% in Text), and from \gNineEFourBTextAccCoreFiveOneTwo\% to \gNineEFourBTextAccCorefree\% on \texttt{E4B-it} Text. Correspondingly, $\log_{10}m_{50}$ climbs from \gNineThirtyOneBVisualMFiftyTwoFiveOneTwo{} at $b=512$ to \gNineThirtyOneBVisualMFiftyTwofree{} at \texttt{free} on \texttt{31B-it} Visual (OLS slope across the three capped budgets: \gNineThirtyOneBVisualMSlope{} decades of $m_{50}$ per tenfold budget, 95\% CI $[\gNineThirtyOneBVisualMSlopeCILo, \gNineThirtyOneBVisualMSlopeCIHi]$) and from \gNineThirtyOneBTextMFiftyTwoFiveOneTwo{} to \gNineThirtyOneBTextMFiftyTwofree{} on \texttt{31B-it} Text (slope \gNineThirtyOneBTextMSlope, CI $[\gNineThirtyOneBTextMSlopeCILo, \gNineThirtyOneBTextMSlopeCIHi]$); the slopes are \gNineTwentySixBVisualMSlope{} $[\gNineTwentySixBVisualMSlopeCILo, \gNineTwentySixBVisualMSlopeCIHi]$ and \gNineTwentySixBTextMSlope{} $[\gNineTwentySixBTextMSlopeCILo, \gNineTwentySixBTextMSlopeCIHi]$ on \texttt{26B-A4B-it}, \gNineTwelveBVisualMSlope{} $[\gNineTwelveBVisualMSlopeCILo, \gNineTwelveBVisualMSlopeCIHi]$ and \gNineTwelveBTextMSlope{} $[\gNineTwelveBTextMSlopeCILo, \gNineTwelveBTextMSlopeCIHi]$ on \texttt{12B-it}, and \gNineEFourBTextMSlope{} $[\gNineEFourBTextMSlopeCILo, \gNineEFourBTextMSlopeCIHi]$ on \texttt{E4B-it} Text. Thus when only the thinking token cap is varied on open-weight models, a tenfold increase in thinking budget still buys roughly $0.5$ to $1.3$ decades of answer magnitude.

\emph{H3 (Matched twins) and descriptive checks.} H3 is supported in \gNineHThreeSupported{} of \gNineNCells{} cells (Table~\ref{tab:g9_main}): on the 80 magnitude-matched T6 twins at \texttt{free} budget, few pairs with 3- to 11-digit answers are solved by chain of thought (up to \gNineThirtyOneBTextAccTwinfree\% on \texttt{31B-it} Text, Table~\ref{tab:g9_acc}), and the discordant counts $b/c$ are \gNineThirtyOneBVisualHThreeB/\gNineThirtyOneBVisualHThreeC{} and \gNineThirtyOneBTextHThreeB/\gNineThirtyOneBTextHThreeC{} on \texttt{31B-it}, \gNineTwentySixBVisualHThreeB/\gNineTwentySixBVisualHThreeC{} and \gNineTwentySixBTextHThreeB/\gNineTwentySixBTextHThreeC{} on \texttt{26B-A4B-it}, \gNineTwelveBVisualHThreeB/\gNineTwelveBVisualHThreeC{} and \gNineTwelveBTextHThreeB/\gNineTwelveBTextHThreeC{} on \texttt{12B-it}, and \gNineEFourBVisualHThreeB/\gNineEFourBVisualHThreeC{} on \texttt{E4B-it} (all Holm $p=1.00$ where testable). Table~\ref{tab:g9_compliance} confirms that vLLM's \texttt{thinking\_token\_budget} enforces the caps tightly: across all 24 capped cells ($b\in\{512,2048,8192\}$), median realized reasoning tokens are \gNineThirtyOneBVisualMedRtokFiveOneTwo{} at $b=512$, \gNineThirtyOneBVisualMedRtokTwoZeroFourEight{} at $b=2048$ and \gNineThirtyOneBVisualMedRtokEightOneNineTwo{} at $b=8192$ (except on \texttt{E4B-it} at $8192$, which finishes thinking early at median \gNineEFourBVisualMedRtokEightOneNineTwo{} and \gNineEFourBTextMedRtokEightOneNineTwo{} tokens), with $\ge\gNineTwelveBTextCompFiveOneTwo\%$ compliance ($\mathrm{reasoning\_tokens}\le b+16$) in every cell and $100.0\%$ in 20 of the 24 cells. At low budgets, models frequently continue reasoning in the visible response after the thought channel closes and hit \texttt{max\_tokens} (\texttt{finish\_reason = length}: \gNineThirtyOneBVisualTruncFiveOneTwo\% on \texttt{31B-it} Visual at $b=512$, falling to \gNineThirtyOneBVisualTruncEightOneNineTwo\% at $b=8192$ and \gNineThirtyOneBVisualTruncfree\% at \texttt{free}). Finally, in the two-regressor fit on the core counting tasks at \texttt{free} budget (Table~\ref{tab:g9_compliance}, right), the $\shP$ indicator $\beta_{\mathrm{hard}}$ is small or positive ($\gNineTwentySixBTextClassBeta\pm\gNineTwentySixBTextClassSE$ on \texttt{26B-A4B-it} Text to $\gNineThirtyOneBVisualClassBeta\pm\gNineThirtyOneBVisualClassSE$ on \texttt{31B-it} Visual).

\begin{keepblock}
\begin{center}\scriptsize\genfitsetinput{\input{gen_v8p/gen_v8/tab_g9_main.tex}}\end{center}
\captionof{table}{G9: pre-registered hypothesis tests (H1, H2, H3) and $\log_{10}m_{50}$ scaling across the four thinking-on budgets $b\in\{512,2048,8192,\text{free}\}$ for all four self-hosted Gemma~4 models and both conditions (\gNineTotalCalls{} calls total, \gNineTotalMissing{} missing). H1 fits $\mathrm{logit}(\mathrm{exact})\sim 1+\beta_{\lg}\log_{10}(\mathrm{truth})$ on the 721 pooled core counting tasks ($\GapL$ and $\shP$) at \texttt{free} budget; $\mathrm{odds/dec.}=\exp(\beta_{\lg})$. H2 fits $\mathrm{logit}(\mathrm{exact})\sim 1+\beta_{\lg}\log_{10}(\mathrm{truth})+\beta_{\mathrm{bud}}\log_{10}(b)$ across $b\in\{512,2048,8192\}$ ($n=2{,}163$ per cell). H3 tests $b>c$ (exact one-sided McNemar) on the 80 magnitude-matched T6 twins at \texttt{free} budget ($b$: $\sST$ right only, $c$: $\sAM$ right only). Holm $p$-values are adjusted over the 8 cells in each family; slope/dec.\ is the OLS slope of $\log_{10}m_{50}$ per decade of budget across $b\in\{512,2048,8192\}$, with an instance-clustered bootstrap 95\% CI ($B=2{,}000$).}\label{tab:g9_main}
\end{keepblock}

\begin{keepblock}
\begin{center}\scriptsize\genfitsetinput{\input{gen_v8/tab_g9_acc.tex}}\end{center}
\captionof{table}{G9: exact-match accuracy (\%) by model, condition, thinking budget and task group, plus visual transcription exactness (\%) and mean edge/cell F1 on the core benchmark. Core ($921$): all pooled core pairs; Bin ($200$): $\chi$ ($\TC$) and $\sCC$ ($\Lsp$) on binary images; Grid $\GapL$ ($291$): $\sST$ and $\sPM$; Grid $\shP$ ($350$): $\sAM$ and $\sPATH$; Euler ($80$): $\sECd$ ($\GapL$) and $\sEC$ ($\shP$); Twins ($160$): the 80 magnitude-matched T6 twins, split into Twin $\sST$ ($80$) and Twin $\sAM$ ($80$).}\label{tab:g9_acc}
\end{keepblock}

\begin{keepblock}
\begin{center}\scriptsize\genfitsetinput{\input{gen_v8p/gen_v8/tab_g9_compliance.tex}}\end{center}
\captionof{table}{G9: thinking-budget compliance, realized token medians and equal-magnitude class term. $n=1{,}081$ calls per row (921 core + 160 twin tasks); Med.\ reason.\ tok.\ and Med.\ compl.\ tok.: median reasoning and completion tokens; Compliance: share (\%) of calls with $\mathrm{reasoning\_tokens}\le b+16$; Length-trunc.: share (\%) of calls stopping with \texttt{finish\_reason = length}; right columns: coefficient $\beta_{\mathrm{hard}}\pm\text{SE}$ and two-sided Wald $p$ of the $\shP$ indicator in $\mathrm{logit}(\mathrm{exact})\sim 1+\beta_{\lg}\log_{10}(\mathrm{truth})+\beta_{\mathrm{hard}}\mathbb I[\shP]$ on the 721 core counting tasks at \texttt{free} budget (sep.: quasi-complete separation on \texttt{E4B-it} Visual where grid accuracy is near zero).}\label{tab:g9_compliance}
\end{keepblock}

\ifincludeQW
\section{Self-hosted Qwen3-VL control family}
\label{app:qw}

\paragraph{Why.} Every model in the main paper and Appendix~\ref{app:g9} belongs to the Gemini or Gemma~4 families from a single provider. To check whether our three main conclusions (the steep drop in accuracy with answer magnitude, the positive shift of the counting curve with the exact thinking token budget once the input graph is perceived, and the absence of a $\GapL$ advantage on same-image magnitude-matched twins) hold beyond one vendor's vision encoder, tokenizer and reasoning post-training, we replicate the controlled-budget protocol of Appendix~\ref{app:g9} on a cross-vendor open-weight vision-language family: Alibaba's Qwen3-VL (\texttt{Qwen3-VL-32B-Thinking}, \texttt{Qwen3-VL-8B-Thinking} and \texttt{Qwen3-VL-8B-Instruct}). Evaluating the matched 8B pair (\texttt{Qwen3-VL-8B-Thinking} vs.\ \texttt{Qwen3-VL-8B-Instruct}, which share the same 8B backbone and vision encoder and differ only in thinking post-training) also gives a direct within-backbone comparison between chain-of-thought reasoning and non-thinking instruction following at \texttt{free} budget. As in Appendix~\ref{app:g9}, the experimental plan (\texttt{EXPERIMENT\_QW\_PLAN.md}), task bundles, runner (\texttt{run\_g9.py}) and derived analysis script (\texttt{analyze\_qw.py}) were hash-frozen before the first model finished and before any record content was read.

\paragraph{Models, tasks and serving.} We self-host three open-weight checkpoints in \texttt{bfloat16} under the same pinned vLLM~0.31.0 image (\texttt{--max-model-len 32768 --gpu-memory-utilization 0.92}), each pinned to its HuggingFace commit hash:
\begin{enumerate}
\item \texttt{Qwen/Qwen3-VL-32B-Thinking} (commit \texttt{7edd10ffd1196091948fb245ff63e406ccb2d4d1}), dense 32B thinking model served with \texttt{--reasoning-parser qwen3} at tensor parallel~4 on 4$\times$A100-40GB (\texttt{a2-highgpu-4g} spot VM replicas), using the checkpoint's generation parameters (\texttt{temperature = 0.8}, \texttt{top\_p = 0.95}, \texttt{top\_k = 20});
\item \texttt{Qwen/Qwen3-VL-8B-Thinking} (commit \texttt{92f3c4b4feadd3a016ef468d103bb5f58b2a2c6b}), dense 8B thinking model served with \texttt{--reasoning-parser qwen3} at tensor parallel~2 on 2$\times$A100-40GB (\texttt{a2-highgpu-2g} spot VM replicas), with \texttt{temperature = 1.0}, \texttt{top\_p = 0.95}, \texttt{top\_k = 20}; and
\item \texttt{Qwen/Qwen3-VL-8B-Instruct} (commit \texttt{0c351dd01ed87e9c1b53cbc748cba10e6187ff3b}), the matched dense 8B non-thinking instruction model served without a reasoning parser on A100-40GB VMs, with \texttt{temperature = 0.7}, \texttt{top\_p = 0.8}, \texttt{top\_k = 20}.
\end{enumerate}
The two thinking models (\texttt{32B-Thinking} and \texttt{8B-Thinking}) run on the exact same frozen task bundle as Appendix~\ref{app:g9} (\texttt{tasks\_g9.jsonl}, SHA-256 \texttt{a08e1b81...}): all 921 pooled core instance--task pairs plus the 160 tasks of the 80 magnitude-matched T6 twins ($1{,}081$ pairs) across \textbf{Visual} and \textbf{Text} and all four budgets $b\in\{512,2048,8192,\text{free}\}$ ($8{,}648$ calls per model, with \texttt{thinking\_token\_budget} $=b$ and \texttt{max\_tokens} $=b+2048$ for capped budgets, and no thinking cap with \texttt{max\_tokens} $=30{,}000$ for \texttt{free}). Because \texttt{Qwen3-VL-8B-Instruct} has no thinking channel to cap, it runs the $2{,}162$ \texttt{free}-budget rows (\texttt{max\_tokens} $=30{,}000$, bundle SHA-256 \texttt{e17fac6b...}) across all $1{,}081$ instances and both conditions, giving $8{,}648+8{,}648+2{,}162=\qWTotalCalls$ calls across the \qWNModels{} models.

Before launching the runs, format-only smoke checks (24 throw-away records on \texttt{8B-Thinking} and 11 on \texttt{8B-Instruct}) verified budget capping, \texttt{reasoning\_tokens = 0} on \texttt{8B-Instruct}, and JSON output formatting, and were discarded. Because the long \texttt{8192} and \texttt{free} generations were KV-cache bound, remaining keys were dealt round-robin in the runner's seeded order (\texttt{order\_seed = 20261009}) across data-parallel spot VM replicas, logged under Deviations in \texttt{EXPERIMENT\_QW\_PLAN.md} before any record content was read, with spot preemptions automatically resumed by a 5-minute watchdog and merged by key before running the unchanged frozen \texttt{analyze\_qw.py}: (i)~\texttt{Qwen3-VL-8B-Thinking} ran across 2 shards ($4{,}609+4{,}039=8{,}648$ \texttt{ok} keys on two \texttt{a2-highgpu-2g} VMs at tensor parallel~2); (ii)~\texttt{Qwen3-VL-32B-Thinking} ran across 4 shards ($3{,}891+3{,}574+658+525=8{,}648$ \texttt{ok} keys on four \texttt{a2-highgpu-4g} VMs at tensor parallel~4); and (iii)~\texttt{Qwen3-VL-8B-Instruct} ran across 6 shards ($763+546+590+123+78+62=2{,}162$ \texttt{ok} keys across three \texttt{a2-highgpu-2g} VMs at tensor parallel~2 and three repurposed \texttt{a2-highgpu-4g} VMs at tensor parallel~4 once \texttt{32B-Thinking} completed). All \qWTotalCalls{} calls returned an \texttt{ok} record (\qWTotalMissing{} missing keys across all three models, so the returned-only primary analysis and the missing-as-wrong sensitivity analysis coincide identically).

\paragraph{Hypotheses and analysis.} The frozen script (\texttt{analyze\_qw.py}) applies the same scorer and statistical models as Appendix~\ref{app:g9}, with Holm--Bonferroni adjustments matched to the active cells of each test family at $\alpha=0.05$:
\begin{itemize}
\item \textbf{H1 (Magnitude law at \texttt{free} budget, \qWHOneNCells{} cells)}: $\mathrm{logit}(\mathrm{exact})\sim 1+\beta_{\lg}\log_{10}(\mathrm{truth})$ on the 721 pooled core counting instances ($\GapL$ and $\shP$ tasks with $\text{truth}>0$) at \texttt{free} budget across all 3 models $\times$ 2 conditions, testing $\beta_{\lg}<0$ (one-sided Wald, Holm-adjusted over the \qWHOneNCells{} cells).
\item \textbf{H2 (Controlled thinking budget shifts the curve, \qWHTwoNCells{} cells)}: $\mathrm{logit}(\mathrm{exact})\sim 1+\beta_{\lg}\log_{10}(\mathrm{truth})+\beta_{\mathrm{bud}}\log_{10}(b)$ across the capped budgets $b\in\{512,2048,8192\}$ ($n=2{,}163$ per cell) on the two thinking models $\times$ 2 conditions, testing $\beta_{\mathrm{bud}}>0$ (one-sided Wald, Holm-adjusted over the \qWHTwoNCells{} cells), along with per-budget $m_{50}$ and the OLS slope of $\log_{10}m_{50}$ per decade of budget with instance-clustered bootstrap 95\% CIs ($B=2{,}000$).
\item \textbf{H3 (Same-image magnitude-matched twins at \texttt{free} budget, \qWHThreeNCells{} cells)}: one-sided exact McNemar test ($b>c$, where $b$: $\sST$ right only and $c$: $\sAM$ right only) on the 80 T6 twins at \texttt{free} budget across all 3 models $\times$ 2 conditions (Holm-adjusted over the testable cells).
\item \textbf{Thinking vs.\ Instruct within-backbone comparison (2 conditions)}: a secondary paired two-sided exact McNemar test comparing \texttt{Qwen3-VL-8B-Thinking} (\texttt{free}) against \texttt{Qwen3-VL-8B-Instruct} (\texttt{free}) on the \qWThinkVsInstrVisualNPairs{} pooled core counting instances in Visual and Text (Holm-adjusted over the 2 conditions).
\end{itemize}

\paragraph{Results.} Table~\ref{tab:qw_main} reports H1, H2, H3, $m_{50}$ per budget and the $m_{50}$ budget slope; Table~\ref{tab:qw_acc} gives exact-match accuracy by task group and visual transcription quality; Table~\ref{tab:qw_compliance} gives realized token medians, budget compliance, truncation rates and the equal-magnitude class term.

\emph{H1 (Magnitude law).} H1 is supported in all \qWHOneSupported{} of the \qWHOneNCells{} $(\text{model},\text{condition})$ cells after Holm correction (Table~\ref{tab:qw_main}), with $\beta_{\lg}$ between $\qWBetaLgMin$ and $\qWBetaLgMax$ (all Holm $p\le\qWEightBInstrVisualHOnePTex$) and odds multipliers $\mathrm{odds/dec.}=\exp(\beta_{\lg})$ between \qWOddsDecMin{} and \qWOddsDecMax{}: \qWThirtyTwoBThinkVisualOddsDecade{} ($\beta_{\lg}=\qWThirtyTwoBThinkVisualBetaLg\pm\qWThirtyTwoBThinkVisualBetaLgSE$, Holm $p<10^{-4}$) and \qWThirtyTwoBThinkTextOddsDecade{} ($\beta_{\lg}=\qWThirtyTwoBThinkTextBetaLg\pm\qWThirtyTwoBThinkTextBetaLgSE$, $p<10^{-4}$) on \texttt{32B-Thinking} Visual and Text; \qWEightBThinkVisualOddsDecade{} ($\beta_{\lg}=\qWEightBThinkVisualBetaLg\pm\qWEightBThinkVisualBetaLgSE$, $p=\qWEightBThinkVisualHOnePTex$) and \qWEightBThinkTextOddsDecade{} ($\beta_{\lg}=\qWEightBThinkTextBetaLg\pm\qWEightBThinkTextBetaLgSE$, $p<10^{-4}$) on \texttt{8B-Thinking} Visual and Text; and \qWEightBInstrVisualOddsDecade{} ($\beta_{\lg}=\qWEightBInstrVisualBetaLg\pm\qWEightBInstrVisualBetaLgSE$, $p=\qWEightBInstrVisualHOnePTex$) and \qWEightBInstrTextOddsDecade{} ($\beta_{\lg}=\qWEightBInstrTextBetaLg\pm\qWEightBInstrTextBetaLgSE$, $p<10^{-4}$) on \texttt{8B-Instruct} Visual and Text.

\emph{H2 (Controlled thinking budget) and Thinking vs.\ Instruct.} H2 is supported in \qWHTwoSupported{} of the \qWHTwoNCells{} thinking cells after Holm correction (Table~\ref{tab:qw_main}), namely both \textbf{Text} cells (\texttt{32B-Thinking} Text: $\beta_{\mathrm{bud}}=\qWThirtyTwoBThinkTextBetaBud\pm\qWThirtyTwoBThinkTextBetaBudSE$, Holm $p<10^{-4}$; \texttt{8B-Thinking} Text: $\beta_{\mathrm{bud}}=\qWEightBThinkTextBetaBud\pm\qWEightBThinkTextBetaBudSE$, Holm $p<10^{-4}$). In \textbf{Text}, where the ground-truth graph is given symbolically, core accuracy (Table~\ref{tab:qw_acc}) rises monotonically across $b\in\{512,2048,8192,\text{free}\}$ from \qWThirtyTwoBThinkTextAccCoreFiveOneTwo\% to \qWThirtyTwoBThinkTextAccCoreTwoZeroFourEight\%, \qWThirtyTwoBThinkTextAccCoreEightOneNineTwo\% and \qWThirtyTwoBThinkTextAccCorefree\% on \texttt{32B-Thinking} (Grid $\GapL$: \qWThirtyTwoBThinkTextAccGridPolyFiveOneTwo\% to \qWThirtyTwoBThinkTextAccGridPolyfree\%; Grid $\shP$: \qWThirtyTwoBThinkTextAccGridHardFiveOneTwo\% to \qWThirtyTwoBThinkTextAccGridHardfree\%; Bin: \qWThirtyTwoBThinkTextAccBinFiveOneTwo\% to \qWThirtyTwoBThinkTextAccBinfree\%) and from \qWEightBThinkTextAccCoreFiveOneTwo\% to \qWEightBThinkTextAccCoreTwoZeroFourEight\%, \qWEightBThinkTextAccCoreEightOneNineTwo\% and \qWEightBThinkTextAccCorefree\% on \texttt{8B-Thinking} (Grid $\GapL$: \qWEightBThinkTextAccGridPolyFiveOneTwo\% to \qWEightBThinkTextAccGridPolyfree\%; Grid $\shP$: \qWEightBThinkTextAccGridHardFiveOneTwo\% to \qWEightBThinkTextAccGridHardfree\%; Bin: \qWEightBThinkTextAccBinFiveOneTwo\% to \qWEightBThinkTextAccBinfree\%), shifting $\log_{10}m_{50}$ from \qWThirtyTwoBThinkTextMFiftyTwoFiveOneTwo{} at $b=512$ to \qWThirtyTwoBThinkTextMFiftyTwoEightOneNineTwo{} at $b=8192$ and \qWThirtyTwoBThinkTextMFiftyTwofree{} at \texttt{free} on \texttt{32B-Thinking} Text (OLS slope \qWThirtyTwoBThinkTextMSlope{} decades/decade, 95\% CI $[\qWThirtyTwoBThinkTextMSlopeCILo, \qWThirtyTwoBThinkTextMSlopeCIHi]$) and from \qWEightBThinkTextMFiftyTwoFiveOneTwo{} to \qWEightBThinkTextMFiftyTwoEightOneNineTwo{} and \qWEightBThinkTextMFiftyTwofree{} on \texttt{8B-Thinking} Text (slope \qWEightBThinkTextMSlope, CI $[\qWEightBThinkTextMSlopeCILo, \qWEightBThinkTextMSlopeCIHi]$). In \textbf{Visual}, by contrast, Qwen3-VL is severely bottlenecked by schematic diagram transcription: although overall core Visual accuracy rises with budget from \qWThirtyTwoBThinkVisualAccCoreFiveOneTwo\% to \qWThirtyTwoBThinkVisualAccCorefree\% on \texttt{32B-Thinking} and from \qWEightBThinkVisualAccCoreFiveOneTwo\% to \qWEightBThinkVisualAccCorefree\% on \texttt{8B-Thinking}, Table~\ref{tab:qw_acc} shows that this gain comes almost entirely from the binary-image tasks (Bin: \qWThirtyTwoBThinkVisualAccBinFiveOneTwo\% to \qWThirtyTwoBThinkVisualAccBinfree\% on \texttt{32B-Thinking}; \qWEightBThinkVisualAccBinFiveOneTwo\% to \qWEightBThinkVisualAccBinfree\% on \texttt{8B-Thinking}). On the core grid and Eulerian graphs that enter H2, exact visual transcription is near zero (overall core transcription exactness is only \qWThirtyTwoBThinkVisualPerceptExactFiveOneTwo\%--\qWThirtyTwoBThinkVisualPerceptExactfree\% on \texttt{32B-Thinking}, \qWEightBThinkVisualPerceptExactFiveOneTwo\%--\qWEightBThinkVisualPerceptExactTwoZeroFourEight\% on \texttt{8B-Thinking}, and \qWEightBInstrVisualPerceptExactfree\% on \texttt{8B-Instruct}, concentrated on the binary matrices), keeping Grid $\GapL$ and Grid $\shP$ accuracy at the floor across all budgets ($\le\qWThirtyTwoBThinkVisualAccGridHardEightOneNineTwo\%$ on \texttt{32B-Thinking} Visual and $\le\qWEightBThinkVisualAccGridPolyFiveOneTwo\%$ on \texttt{8B-Thinking} Visual). Consequently, fitted Visual $\log_{10}m_{50}$ is negative at the floor ($\qWThirtyTwoBThinkVisualMFiftyTwoFiveOneTwo$ to $\qWThirtyTwoBThinkVisualMFiftyTwoEightOneNineTwo$ on \texttt{32B-Thinking}; $\qWEightBThinkVisualMFiftyTwofree$ to $\qWEightBThinkVisualMFiftyTwoEightOneNineTwo$ on \texttt{8B-Thinking}) and H2 is null in Visual (\texttt{32B-Thinking} Visual: $\beta_{\mathrm{bud}}=\qWThirtyTwoBThinkVisualBetaBud\pm\qWThirtyTwoBThinkVisualBetaBudSE$, raw $p=\qWThirtyTwoBThinkVisualHTwoP$, Holm $p=\qWThirtyTwoBThinkVisualHTwoPTex$; \texttt{8B-Thinking} Visual: $\beta_{\mathrm{bud}}=\qWEightBThinkVisualBetaBud\pm\qWEightBThinkVisualBetaBudSE$, Holm $p=\qWEightBThinkVisualHTwoPTex$). Just like \texttt{gemma-4-E4B-it} Visual in Appendix~\ref{app:g9}, this contrast directly reinforces the paper's Seeing vs.\ Counting separation: exact thinking token budget shifts the counting curve strongly once the graph is given as text, whereas perception failure caps accuracy near the floor regardless of thinking budget. The paired within-backbone comparison at \texttt{free} budget on the \qWThinkVsInstrTextNPairs{} core counting instances exhibits the same dissociation: in \textbf{Text}, \texttt{Qwen3-VL-8B-Thinking} beats \texttt{Qwen3-VL-8B-Instruct} by $\qWThinkVsInstrTextAccDiff$ percentage points (\qWThinkVsInstrTextAccThink\% vs.\ \qWThinkVsInstrTextAccInstr\%, discordant counts $b/c=\qWThinkVsInstrTextB/\qWThinkVsInstrTextC$, two-sided McNemar Holm $p={}$\qWThinkVsInstrTextPHolmTex, shifting $\log_{10}m_{50}$ from \qWThinkVsInstrTextMFiftyInstr{} to \qWThinkVsInstrTextMFiftyThink{}, $\Delta m_{50}=\qWThinkVsInstrTextMFiftyDiff$ decades), whereas in \textbf{Visual} both are tied at the visual floor (\qWThinkVsInstrVisualAccThink\% vs.\ \qWThinkVsInstrVisualAccInstr\%, $b/c=\qWThinkVsInstrVisualB/\qWThinkVsInstrVisualC$, Holm $p={}$\qWThinkVsInstrVisualPHolmTex). All accuracies count a call that reaches the token cap before emitting an answer as wrong, and in Text such calls are frequent: on \texttt{32B-Thinking}, \qWThirtyTwoBThinkTextTruncFiveOneTwo\% of calls end at the cap at $b=512$, \qWThirtyTwoBThinkTextTruncEightOneNineTwo\% at $b=8192$ and \qWThirtyTwoBThinkTextTruncfree\% at \texttt{free} (\texttt{8B-Thinking}: \qWEightBThinkTextTruncFiveOneTwo\%, \qWEightBThinkTextTruncEightOneNineTwo\% and \qWEightBThinkTextTruncfree\%; Table~\ref{tab:qw_compliance}), so part of the budget effect in Text is the share of calls that finish. \texttt{8B-Instruct} reaches its 30{,}000-token cap on \qWEightBInstrTextTruncfree\% of Text calls, against \qWEightBThinkTextTruncfree\% for \texttt{8B-Thinking} at \texttt{free}.

\emph{H3 (Matched twins) and descriptive checks.} On the 80 magnitude-matched T6 twins at \texttt{free} budget (H3, Table~\ref{tab:qw_main}), \qWHThreeSupported{} of the \qWHThreeNCells{} cells pass Holm correction: because T6 twin answers span $10^2$--$10^{11}$, well above Qwen3-VL's $m_{50}\le 10^{\qWThirtyTwoBThinkTextMFiftyTwofree}$, twin accuracy is at the floor, with discordant counts $b/c=\qWThirtyTwoBThinkVisualHThreeB/\qWThirtyTwoBThinkVisualHThreeC$ (Holm $p=\qWThirtyTwoBThinkVisualHThreePTex$) on \texttt{32B-Thinking} Visual and $0/0$ in the other five cells. Table~\ref{tab:qw_compliance} confirms $100.0\%$ budget compliance ($\mathrm{reasoning\_tokens}\le b+16$) in all 12 capped cells (median reasoning tokens \qWThirtyTwoBThinkVisualMedRtokFiveOneTwo{}, \qWThirtyTwoBThinkVisualMedRtokTwoZeroFourEight{} and \qWThirtyTwoBThinkVisualMedRtokEightOneNineTwo{} at $b=512,2048,8192$). At \texttt{free} budget, median reasoning tokens reach \qWThirtyTwoBThinkVisualMedRtokfree{} (Visual) and \qWThirtyTwoBThinkTextMedRtokfree{} (Text) on \texttt{32B-Thinking} and \qWEightBThinkVisualMedRtokfree{} (Visual) and \qWEightBThinkTextMedRtokfree{} (Text) on \texttt{8B-Thinking}, while \texttt{8B-Instruct} emits \qWEightBInstrVisualMedRtokfree{} reasoning tokens and a median of \qWEightBInstrVisualMedCtokfree{} (Visual) and \qWEightBInstrTextMedCtokfree{} (Text, hitting \texttt{finish\_reason = length} on \qWEightBInstrTextTruncfree\% of calls) completion tokens. Finally, the equal-magnitude class term $\beta_{\mathrm{hard}}$ at \texttt{free} budget (Table~\ref{tab:qw_compliance}, right) is non-negative or within noise ($\qWThirtyTwoBThinkVisualClassBeta\pm\qWThirtyTwoBThinkVisualClassSE$, $p=\qWThirtyTwoBThinkVisualClassPTex$ on \texttt{32B-Thinking} Visual; $\qWThirtyTwoBThinkTextClassBeta\pm\qWThirtyTwoBThinkTextClassSE$, $p=\qWThirtyTwoBThinkTextClassPTex$ on \texttt{32B-Thinking} Text; $\qWEightBThinkVisualClassBeta\pm\qWEightBThinkVisualClassSE$, $p=\qWEightBThinkVisualClassPTex$ on \texttt{8B-Thinking} Visual; $\qWEightBThinkTextClassBeta\pm\qWEightBThinkTextClassSE$, $p=\qWEightBThinkTextClassPTex$ on \texttt{8B-Thinking} Text; quasi-complete separation on \texttt{8B-Instruct} Visual; and $\qWEightBInstrTextClassBeta\pm\qWEightBInstrTextClassSE$, $p=\qWEightBInstrTextClassPTex$ on \texttt{8B-Instruct} Text).

\begin{keepblock}
\begin{center}\scriptsize\genfitsetinput{\input{gen_v8p/gen_v8/tab_qw_main.tex}}\end{center}
\captionof{table}{QW: pre-registered hypothesis tests (H1, H2, H3) and $\log_{10}m_{50}$ scaling across the self-hosted Qwen3-VL control family (\qWTotalCalls{} calls total, \qWTotalMissing{} missing). H1 fits $\mathrm{logit}(\mathrm{exact})\sim 1+\beta_{\lg}\log_{10}(\mathrm{truth})$ on the 721 pooled core counting tasks ($\GapL$ and $\shP$) at \texttt{free} budget (Holm-adjusted over \qWHOneNCells{} cells); $\mathrm{odds/dec.}=\exp(\beta_{\lg})$. H2 fits $\mathrm{logit}(\mathrm{exact})\sim 1+\beta_{\lg}\log_{10}(\mathrm{truth})+\beta_{\mathrm{bud}}\log_{10}(b)$ across $b\in\{512,2048,8192\}$ ($n=2{,}163$ per cell) on the two thinking models (Holm-adjusted over \qWHTwoNCells{} cells). H3 tests $b>c$ (exact one-sided McNemar) on the 80 magnitude-matched T6 twins at \texttt{free} budget ($b$: $\sST$ right only, $c$: $\sAM$ right only; Holm-adjusted over testable cells). slope/dec.\ is the OLS slope of $\log_{10}m_{50}$ per decade of budget across $b\in\{512,2048,8192\}$, with an instance-clustered bootstrap 95\% CI ($B=2{,}000$).}\label{tab:qw_main}
\end{keepblock}

\begin{keepblock}
\begin{center}\scriptsize\genfitsetinput{\input{gen_v8/tab_qw_acc.tex}}\end{center}
\captionof{table}{QW: exact-match accuracy (\%) by model, condition, thinking budget and task group, plus visual transcription exactness (\%) and mean edge/cell F1 on the core benchmark for the Qwen3-VL control family. Column definitions match Table~\ref{tab:g9_acc}.}\label{tab:qw_acc}
\end{keepblock}

\begin{keepblock}
\begin{center}\scriptsize\genfitsetinput{\input{gen_v8p/gen_v8/tab_qw_compliance.tex}}\end{center}
\captionof{table}{QW: thinking-budget compliance, realized token medians and equal-magnitude class term for the Qwen3-VL control family ($n=1{,}081$ calls per row: 921 core + 160 twin tasks). Column definitions match Table~\ref{tab:g9_compliance} (sep.: quasi-complete separation on \texttt{8B-Instruct} Visual where grid accuracy is near zero).}\label{tab:qw_compliance}
\end{keepblock}
\fi

\ifincludeQW
\section{Exploratory: how much of the budget and magnitude effects is truncation?}
\label{app:trunc}
This analysis was not pre-registered and was run after the results of Appendices~\ref{app:g9} and~\ref{app:qw} were known. The frozen scorer counts a call that stops at the token cap (\texttt{finish\_reason = length}) as wrong unless its partial output already contains the right integer (\truncExactMin{} to \truncExactMax{} such calls per model), and capped calls are common on the smaller models at low budgets (Tables~\ref{tab:g9_compliance} and~\ref{tab:qw_compliance}). We first rebuilt both result files from the raw records with the frozen code path (\truncValidGNine{} G9 and \truncValidQW{} QW values identical) and then refit H1 and H2 on finished calls only, with the same models, test directions and Holm families (Table~\ref{tab:trunc_finished}). On finished calls the magnitude law holds in \truncHOneSurv{} of \truncHOneN{} cells (frozen analyses: \truncHOneFrozen{} of \truncHOneN{}), with coefficients close to the frozen ones (median finished/frozen ratio \truncHOneRatioMed{} over the \truncHOneFrozen{} cells that pass in the frozen analyses). The budget effect holds in \truncHTwoSurv{} of \truncHTwoN{} cells (frozen: \truncHTwoFrozen{} of \truncHTwoN{}) but shrinks (median ratio \truncHTwoRatioMed{} over the \truncHTwoFrozen{} cells that pass in the frozen analyses; \texttt{gemma-4-12B-it} Text goes from \truncBTwelveTextFrozen{} to \truncBTwelveTextFin{}), so a substantial part of the budget effect on these models comes from whether a call finishes. Restricted to pairs in which both calls finished, \texttt{Qwen3-VL-8B-Thinking} still beats \texttt{Qwen3-VL-8B-Instruct} in Text (\truncTviTextThink{}\% vs.\ \truncTviTextInstr{}\%, $b/c=\truncTviTextB/\truncTviTextC$, exact McNemar $p=\truncTviTextP$, $n=\truncTviTextN$ pairs). Finishing itself depends on budget and answer size, so the finished-only rows are a selected sample and serve as a robustness check, not as an estimate of a separate effect.

\begin{keepblock}
\begin{center}\scriptsize\genfitsetinput{\input{gen_v8p/gen_v8/tab_trunc_finished.tex}}\end{center}
\captionof{table}{Exploratory, not pre-registered. H1 and H2 of Appendices~\ref{app:g9} and~\ref{app:qw} refit on finished calls only, that is, without the calls that stopped at the token cap (\texttt{finish\_reason = length}). H1 rows give $\beta_{\lg}$ of $\mathrm{logit}(\mathrm{exact})\sim 1+\beta_{\lg}\log_{10}(\mathrm{truth})$ on the core counting tasks at \texttt{free} budget; H2 rows give $\beta_{\mathrm{bud}}$ of $\mathrm{logit}(\mathrm{exact})\sim 1+\beta_{\lg}\log_{10}(\mathrm{truth})+\beta_{\mathrm{bud}}\log_{10}(b)$ across the capped budgets. Exp.: experiment (G9 or QW); Cond.: input condition; frozen: the coefficient of the pre-registered analysis on all calls (Tables~\ref{tab:g9_main} and~\ref{tab:qw_main}); finished: the same coefficient refit on finished calls only; $p$: one-sided Wald $p$ of the finished-only fit in the pre-registered direction ($\beta_{\lg}<0$ for H1, $\beta_{\mathrm{bud}}>0$ for H2); $p_{\mathrm{Holm}}$: exploratory Holm adjustment of that $p$ over the cells of the same experiment and hypothesis, mirroring the frozen families; $n_{\mathrm{fin}}$: finished calls in the fit, out of \truncHOneNRows{} per H1 cell and \truncHTwoNRows{} per H2 cell; survives: whether the finished-only fit meets the frozen rule (Holm-adjusted $p$ at or below the pre-registered $\alpha$, with the pre-registered sign); $^\dagger$: the verdict differs from the frozen analysis.}\label{tab:trunc_finished}
\end{keepblock}
\fi

\section{Realistic-style rendered diagrams (Experiment R9)}
\label{app:r9}

\paragraph{Why.} Appendices~\ref{app:v3_ext} and~\ref{app:d8} test whether our conclusions survive five rendering styles and ten visual degradations applied to synthetic node-link drawings. A natural computer-vision question is whether the three core phenomena of the paper (the steep drop of chain-of-thought accuracy with answer magnitude rather than complexity class on magnitude-matched twins, the large accuracy gain from writing code once the structure is perceived, and the anisotropic horizontal-over-vertical edge false-negative blind spot on large grids repaired by rotated multi-view plus local pair-crop verification) also hold when the same kind of graphs is rendered in realistic, domain-textured styles rather than as clean SVG drawings on white backgrounds. Every R9 image is procedurally rendered from a known graph; none is a photograph.

\paragraph{Domains, slices and pre-registration.} Experiment R9 (\texttt{EXPERIMENT\_R9\_PLAN.md} and \texttt{analyze\_r9.py}, hash-frozen in \texttt{PLAN\_FREEZE\_r9.txt} and \texttt{ANALYSIS\_FREEZE\_r9.txt} before any API call was issued) evaluates $\rNineTotalGraphs$ procedural instances across $\rNineDomains$ realistic visual styles (Fig.~\ref{fig:r9_domains}):
\begin{enumerate}
\item \textbf{FR-4 printed circuit boards (\texttt{pcb\_trace})}: dark solder-mask substrate with silkscreen grid ticks, component reference designators (\texttt{U0}--\texttt{U63}), gold-ringed copper solder pads, and angled copper traces with specular highlights and board grain;
\item \textbf{Architectural cyanotype floorplans (\texttt{blueprint\_floorplan})}: Prussian-blue blueprint paper with fold creases, CAD title blocks, room partition boxes, and double-line architectural corridors;
\item \textbf{Metro transit schematics (\texttt{transit\_map})}: off-white parchment transit map with river/park background geography, line-colored route segments, and white-rimmed transfer station interchanges; and
\item \textbf{Rendered desk scenes (\texttt{camera\_desk\_shadow})}: a wood-grain tabletop rendered under a projective homography slant, vignette lighting, sensor noise, 3D-shaded physical pucks casting directional drop shadows, and taut connecting cords.
\end{enumerate}
The benchmark comprises two pre-registered slices:
\begin{itemize}
\item \textbf{Slice R9-A (Large realistic diagrams, $\rNineLargeGrids$ images, $\rNineANPairs$ task pairs)}: $6$ grids at $7{\times}7$ and $6$ at $8{\times}8$ per domain ($12\times 4=\rNineLargeGrids$ images), each evaluated on $\sST$ ($\GapL$) and $\sAM$ ($\shP$-complete) under six conditions: \textbf{Direct} (chain-of-thought JSON transcription and count), \textbf{One-shot Code} (single-turn Python), \textbf{TwoStage-1v} (one upright transcription call solved by exact Kirchhoff/profile-DP code), \textbf{TwoStage-Zoom} ($2{\times}2$ overlapping quadrant crops merged and solved by exact code), \textbf{TwoStage-rot180} ($0^\circ{+}180^\circ$ readings plus local pair-crop adjudication on disagreements), and \textbf{TwoStage-2v} ($0^\circ{+}90^\circ$ axis-swapping readings plus local pair-crop adjudication on disagreements).
\item \textbf{Slice R9-B (Magnitude-matched realistic twins, $\rNineTwinGraphs$ images, $\rNineBNTasks$ task pairs)}: $10$ grids per domain across $4{\times}4$, $4{\times}5$, $5{\times}5$ and $5{\times}6$ ($10\times 4=\rNineTwinGraphs$ images) filtered so that every graph has $|\log_{10}\sST(G)-\log_{10}\sAM(G)|\le 0.25$ decades and $\sPM(G)\ge 2$, evaluated on $\sST$, $\sAM$ and the smaller-magnitude control $\sPM$ under \textbf{Direct} and \textbf{Code (\car)}.
\end{itemize}
We evaluate both \texttt{gemini-3.8-flash} (low, primary) and \texttt{gemini-3.1-pro-preview} (low), totalling $\rNinePerModelRecords$ standardized evaluation records per model ($\rNineTotalRecords$ records overall from $2{,}761$ raw API calls, with zero missing records).

\begin{keepblock}
\begin{center}
\setlength{\tabcolsep}{4pt}
\resizebox{\linewidth}{!}{%
\begin{tabular}{cccc}
\includegraphics[height=3.2cm]{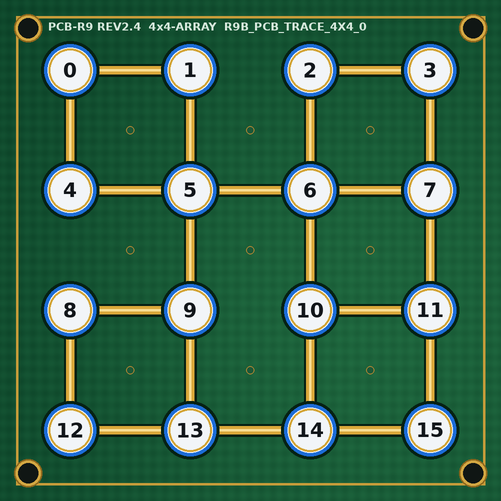} &
\includegraphics[height=3.2cm]{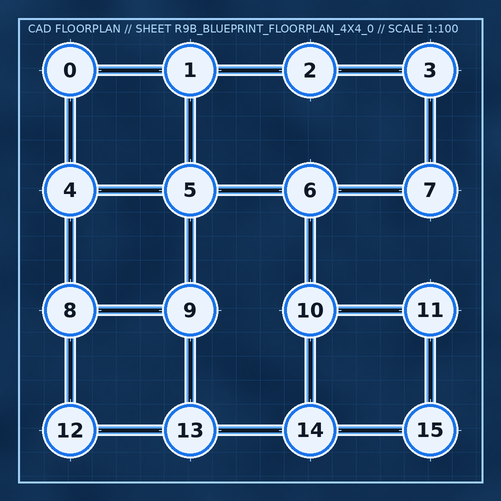} &
\includegraphics[height=3.2cm]{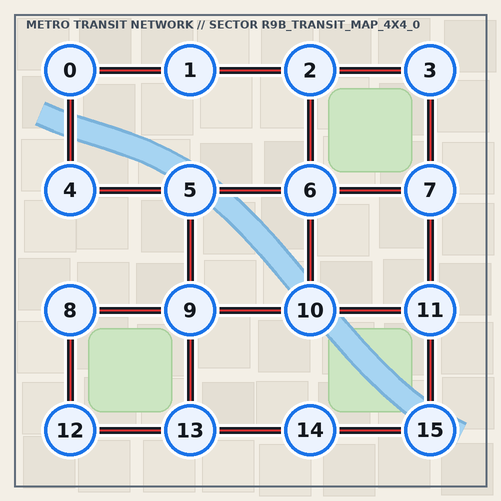} &
\includegraphics[height=3.2cm]{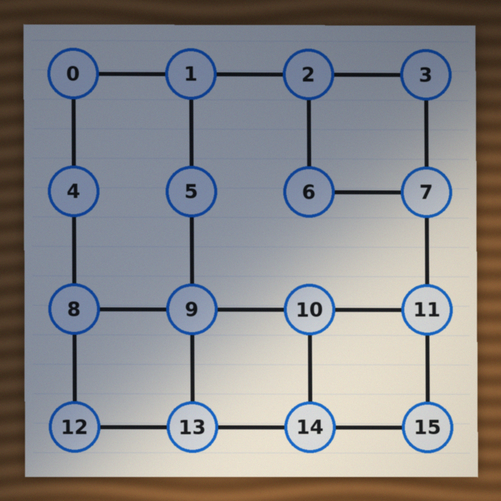} \\
{\footnotesize (a) \texttt{pcb\_trace} (FR-4 copper PCB)} &
{\footnotesize (b) \texttt{blueprint\_floorplan} (CAD blueprint)} &
{\footnotesize (c) \texttt{transit\_map} (metro schematic)} &
{\footnotesize (d) \texttt{camera\_desk\_shadow} (rendered desk scene)}
\end{tabular}}
\end{center}

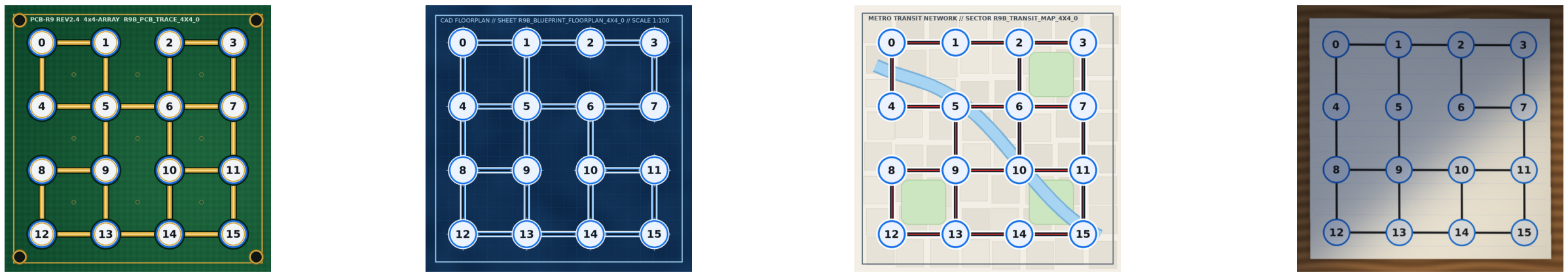
\captionof{figure}{Representative instances from the four realistic rendering styles of Experiment R9 (\S\ref{app:r9}); all images are rendered from known graphs. Each style is evaluated on both magnitude-matched complexity twins ($4{\times}4$--$5{\times}6$, Slice R9-B) and large $7{\times}7$--$8{\times}8$ diagrams (Slice R9-A).}\label{fig:r9_domains}
\end{keepblock}

\paragraph{Registered criteria.} The plan fixes a pass threshold for every hypothesis. \texttt{analyze\_r9.py} computes the statistics; Table~\ref{tab:r9_verdicts} applies the registered thresholds to them. On \texttt{gemini-3.8-flash}, \rNineCritPass{} of the \rNineCritN{} criteria pass: the magnitude decay (H1b), the horizontal-edge anisotropy (H3a), and the gains of the two-view pipeline over one view and over quadrant zoom (H3b-i, H3b-ii). Four fail: code gains less than registered (H1a), matched $\sST$/$\sAM$ twins differ by \rNineTwinDirectSTvsAMDiff{} points rather than at most 15 (H2a), $\sPM$ leads $\sAM$ by less than 35 points (H2b), and a $180^\circ$ second view does as well as a $90^\circ$ one (H3b-iii). On \texttt{gemini-3.1-pro-preview}, \rNineProCritPass{} pass. Because H3b-iii fails on both models, R9 supports the second, differently rendered reading with crop adjudication, not the axis swap, as the source of the repair, in line with Appendix~\ref{app:v6}. The plan's text calls $\sPM$ $\shP$-complete; on planar grids it is in $\GapL$ (Kasteleyn), which does not change the registered H2b threshold.

\begin{keepblock}
\begin{center}\scriptsize\genfitsetinput{\input{gen_v8p/gen_v8/tab_r9_verdicts.tex}}\end{center}
\captionof{table}{Experiment R9: registered criteria from \texttt{EXPERIMENT\_R9\_PLAN.md} (Section 3, frozen before the first API call) and verdicts for both models, computed from the frozen analysis output. ``style'': one of the four rendering styles.}\label{tab:r9_verdicts}
\end{keepblock}

\paragraph{Results.} Table~\ref{tab:r9} reports all pre-registered metrics for both models.
\begin{enumerate}
\item \textbf{H1 and H2 on Slice R9-B (Magnitude law, matched twins, and code routing)}: On the $\rNineBNTasks$ realistic twin evaluations, \textbf{Direct} chain-of-thought accuracy decays monotonically across the four pre-registered $\log_{10}(\text{truth})$ bins ($[0,2.5)$, $[2.5,4.5)$, $[4.5,6.5)$, $[6.5,\infty)$) from \rNineBDirectBinOneAcc{} to \rNineBDirectBinTwoAcc{}, \rNineBDirectBinThreeAcc{} and \rNineBDirectBinFourAcc{} on \texttt{gemini-3.8-flash} ($\beta_{\lg}=\rNineBDirectBetaLg\pm\rNineBDirectBetaLgSE$, odds multiplier \rNineBDirectOddsDec{}/decade, Wald $p=\rNineBDirectLogitP$) and from \rNineProBDirectBinOneAcc{} to \rNineProBDirectBinTwoAcc{}, \rNineProBDirectBinThreeAcc{} and \rNineProBDirectBinFourAcc{} on \texttt{gemini-3.1-pro-preview} ($\beta_{\lg}=\rNineProBDirectBetaLg\pm\rNineProBDirectBetaLgSE$, $p=\rNineProBDirectLogitP$). On the $\rNineTwinGraphs$ magnitude-matched twins ($|\log_{10}\sST-\log_{10}\sAM|\le 0.25$), \textbf{Direct} accuracy on $\sST$ ($\GapL$) is higher than on $\sAM$ ($\shP$-complete) but not significantly so (\rNineTwinDirectSTAcc{} vs.\ \rNineTwinDirectAMAcc{}, two-sided McNemar $p=\rNineTwinDirectSTvsAMP$ on \texttt{3.8-flash}; \rNineProTwinDirectSTAcc{} vs.\ \rNineProTwinDirectAMAcc{}, $p=\rNineProTwinDirectSTvsAMP$ on \texttt{3.1-pro}); the \texttt{3.8-flash} gap exceeds the registered 15-point bound, so H2a fails. The $\GapL$ task $\sPM$, whose answers are smaller, reaches \rNineTwinDirectPMAcc{} on the same images (\rNineTwinDirectPMvsAMGain{} over $\sAM$, one-sided McNemar $p=\rNineTwinDirectPMvsAMP$) on \texttt{3.8-flash} and \rNineProTwinDirectPMAcc{} (\rNineProTwinDirectPMvsAMGain{}, $p=\rNineProTwinDirectPMvsAMP$) on \texttt{3.1-pro}; this contrast changes class and magnitude together. Routing to code (\car) lifts overall R9-B accuracy from \rNineBDirectAcc{} to \rNineBCodeAcc{} (\rNineBCodeGain{}, two-sided McNemar $p=\rNineBCodeMcnemarP$) on \texttt{3.8-flash} and from \rNineProBDirectAcc{} to \rNineProBCodeAcc{} (\rNineProBCodeGain{}, $p=\rNineProBCodeMcnemarP$) on \texttt{3.1-pro}, with gains in every style on \texttt{3.8-flash} (\texttt{pcb\_trace} \rNineBPcbDirAcc{}$\to$\rNineBPcbCodeAcc{}, \texttt{blueprint\_floorplan} \rNineBBlueDirAcc{}$\to$\rNineBBlueCodeAcc{}, \texttt{transit\_map} \rNineBTransDirAcc{}$\to$\rNineBTransCodeAcc{}, \texttt{camera\_desk\_shadow} \rNineBCamDirAcc{}$\to$\rNineBCamCodeAcc{}), though below the registered size (H1a).
\item \textbf{H3a on Slice R9-A (Anisotropic horizontal-edge blind spot)}: Across the $\rNineLargeGrids$ large $7{\times}7$ and $8{\times}8$ diagrams, upright single-view transcription (\textbf{TwoStage-1v}) misses horizontal segments at $\rNineAnisoRatio$ the rate of vertical segments on \texttt{gemini-3.8-flash} (mean $\mathrm{FNR}_H=\rNineAnisoFnrH$ vs.\ $\mathrm{FNR}_V=\rNineAnisoFnrV$, one-sided Wilcoxon $p=\rNineAnisoWilcoxonP$, sign-permutation $p=\rNineAnisoSignP$), holding in all four styles (\texttt{pcb\_trace} $\rNineAnisoPcbFnrH$ vs.\ $\rNineAnisoPcbFnrV$, \texttt{blueprint\_floorplan} $\rNineAnisoBlueFnrH$ vs.\ $\rNineAnisoBlueFnrV$, \texttt{transit\_map} $\rNineAnisoTransFnrH$ vs.\ $\rNineAnisoTransFnrV$, and \texttt{camera\_desk\_shadow} $\rNineAnisoCamFnrH$ vs.\ $\rNineAnisoCamFnrV$). On \texttt{gemini-3.1-pro-preview}, $\mathrm{FNR}_H=\rNineProAnisoFnrH$ vs.\ $\mathrm{FNR}_V=\rNineProAnisoFnrV$ ($\rNineProAnisoRatio$, one-sided Wilcoxon $p=\rNineProAnisoWilcoxonP$), short of the registered $p<10^{-4}$.
\item \textbf{H3b on Slice R9-A (Multi-view plus local pair-crop verification repair)}: On the $\rNineANPairs$ large-diagram task evaluations, chain of thought (\textbf{Direct}) scores \rNineDirTaskAcc{} on both models, and single-turn \textbf{One-shot Code} solves only \rNineCodeTaskAcc{} on \texttt{3.8-flash} (\rNineProCodeTaskAcc{} on \texttt{3.1-pro}). Separating transcription in a single upright view (\textbf{TwoStage-1v}) reaches \rNineOneVTaskAcc{} (\rNineOneVGraphK/48 exact graphs, mean edge $F_1=\rNineOneVFOne$) on \texttt{3.8-flash} and \rNineProOneVTaskAcc{} (\rNineProOneVGraphK/48 exact graphs) on \texttt{3.1-pro}, while unrotated $2{\times}2$ quadrant cropping (\textbf{TwoStage-Zoom}) reaches only \rNineZoomTaskAcc{} on \texttt{3.8-flash} and \rNineProZoomTaskAcc{} on \texttt{3.1-pro}. Pairing the upright reading with a rotated second reading and adjudicating only the disagreeing adjacent vertex pairs via local pair crops raises exact graph recovery and end-to-end solving to \rNineTwoVTaskAcc{} (\rNineTwoVGraphK/48 exact graphs, $F_1=\rNineTwoVFOne$, \rNineTwoVvsOneVGain{} over \textbf{1v}, $b/c=\rNineTwoVvsOneVBC$, one-sided exact McNemar $p=\rNineTwoVvsOneVP$; \rNineTwoVvsZoomGain{} over \textbf{Zoom}, $p=\rNineTwoVvsZoomP$) for $90^\circ$ (\textbf{TwoStage-2v}) and \rNineRotOneEightyTaskAcc{} (\rNineRotOneEightyGraphK/48 exact graphs) for $180^\circ$ (\textbf{TwoStage-rot180}) on \texttt{3.8-flash}, and to \rNineProTwoVTaskAcc{} (\rNineProTwoVGraphK/48 exact graphs, \rNineProTwoVvsOneVGain{} over \textbf{1v}, $b/c=\rNineProTwoVvsOneVBC$, $p=\rNineProTwoVvsOneVP$) for $90^\circ$ and \rNineProRotOneEightyTaskAcc{} (\rNineProRotOneEightyGraphK/48 exact graphs) for $180^\circ$ on \texttt{3.1-pro}. The $180^\circ$ view, which keeps every edge's orientation, is at least as good as the $90^\circ$ view on both models (H3b-iii fails).
\end{enumerate}

\begin{keepblock}
\begin{center}\scriptsize\genfitsetinput{\input{gen_v8/tab_r9.tex}}\end{center}
\captionof{table}{Experiment R9: pre-registered evaluation across four realistic engineering and physical diagram domains (\texttt{pcb\_trace}, \texttt{blueprint\_floorplan}, \texttt{transit\_map}, \texttt{camera\_desk\_shadow}; Fig.~\ref{fig:r9_domains}) on \texttt{gemini-3.8-flash} (low) and \texttt{gemini-3.1-pro-preview} (low) ($\rNineTotalRecords$ standardized evaluation records across $2{,}761$ API calls). Top: Slice R9-A ($48$ large $7{\times}7$ and $8{\times}8$ diagrams, $96$ task pairs across $\sST$ and $\sAM$), reporting exact graph transcription rate, end-to-end exact task accuracy, and mean edge $F_1$ across six conditions. Middle: single-view horizontal ($\mathrm{FNR}_H$) vs.\ vertical ($\mathrm{FNR}_V$) edge false-negative rates and anisotropy ratios $\mathrm{FNR}_H/\mathrm{FNR}_V$ with one-sided paired Wilcoxon $p$-values. Bottom: Slice R9-B ($40$ magnitude-matched twins with $|\log_{10}\sST-\log_{10}\sAM|\le 0.25$, $120$ evaluations across $\sST$, $\sAM$ and smaller-magnitude $\sPM$), comparing Direct (CoT) against Code (\car) overall, by visual domain, and by task.}\label{tab:r9}
\end{keepblock}

\section{Code conditions, the routing prompt and the reading pipelines}
\label{app:codecond}
This appendix gives the details that Section~\ref{sec:car} summarises (its paragraph ``Code and multi-view reading'') and the full version of Table~\ref{tab:code}.

\paragraph{The routing prompt (\car).} A model that writes code to count typically transcribes the structure and writes the first algorithm that comes to mind, usually enumeration. \car{} (Complexity-Aware Routing) inserts one step. The prompt (\S\ref{app:prompts}) is generic, never names a task, and asks for three things: \emph{Perceive}, transcribe the structure exactly; \emph{Classify}, decide whether the counting problem admits a known polynomial-time exact algorithm or is $\shP$-complete in general so that only exponential methods exist, and name the theorem or algorithm; \emph{Solve by code}, write one program that hard-codes the transcription and prints one integer, with exact integer arithmetic for algebraic routes and a frontier dynamic program rather than naive enumeration for $\shP$-complete problems. The main paper uses \car{} only as a control: if routing mattered, \car{} should beat plain Code where enumeration is too slow and tie elsewhere (Proposition~\ref{prop:car}).

\paragraph{Four controls.} Four controls isolate the ingredients: \textbf{Perceive+Code} keeps the transcription step and drops the classification and routing sentences; \textbf{Route-only} keeps the first two steps but asks the model to carry out the chosen algorithm by hand; \textbf{Text\,\car{}} runs \car{} on the text input; \textbf{Hint$\to$Code} gives the task-specific theorem and asks for code.

\paragraph{Zoom-$k$ and two-view reading.} Because the extra-large grids fail through single-view transcription (Section~\ref{sec:car_results}), where horizontal segments are missed far more often than vertical ones, two further pipelines separate the Perceive step from the rest; no ground truth enters either. \textbf{Zoom-$k$} cuts the image into $k{\times}k$ overlapping crops, each wide enough to contain any lattice edge with both endpoint circles, asks for the edges in each crop with the same transcription prompt, merges the lists and solves the merged list with Text\,\car{} or Text$\to$Code; Zoom-1 is a single-shot transcription of the whole image (Table~\ref{tab:zoom}). \textbf{Two-view} transcribes the image upright and as a copy rotated by $90^\circ$; the design meant to turn horizontal edges vertical, with a $180^\circ$ rotation, which keeps them horizontal, as the control, but the two rotations tie (\S\ref{app:v6}), so we read the gain as coming from a second, decorrelated reading. Pairs on which the two readings agree are accepted or rejected accordingly, and each disagreement is settled by one further call showing a local crop of the two endpoint vertices alone. The verified list is then solved by code.

\paragraph{Code conditions on every configuration.} Table~\ref{tab:code_all} lists every code prompt run on the benchmark proper; Table~\ref{tab:code} keeps the Visual$\to$Code, Text$\to$Code and \car{} rows of \texttt{gemini-3.8-flash} (low).

\begin{keepblock}
\begin{center}\scriptsize\genfitsetinput{\input{gen_v8/tab_code_all_v84.tex}}\end{center}
\captionof{table}{Code conditions on the benchmark proper (extra-large grids excluded). ``libs'': programs that import a library; ``algebraic'': fraction of programs for $\sST$, $\sPM$, $\sECd$ that compute a determinant/Pfaffian; ``DP'': fraction of programs for $\sAM$, $\sPATH$, $\sEC$ that use memoised/frontier dynamic programming rather than enumeration.}\label{tab:code_all}
\end{keepblock}

\end{document}

%% file: gen_v8p/gen_v5r/numbers.tex
\newcommand{\nTotalCalls}{13,886}
\newcommand{\nModels}{9}
\newcommand{\nConfigs}{15}

\newcommand{\PerceptSummary}{Over the 3,600 Visual calls with a scorable transcription, the mean edge or cell F1 is 0.994 and the transcription is exact in 85\% of calls. For \texttt{gemini-3.8-flash} (low) it is exact in 88\% of grid drawings, 100\% of binary images and 100\% of Eulerian drawings, and the weakest configuration, \texttt{gemini-3.1-flash-lite} (low), is still exact in 42\%.}
\newcommand{\PerceptPooledN}{3,600}
\newcommand{\PerceptPooledF}{0.994}
\newcommand{\PerceptPooledExact}{85}

\newcommand{\AttribSummary}{For \texttt{gemini-3.8-flash} (low), 115 of its 144 Visual errors (80\%) occur with an exactly transcribed input and are therefore computational by our definition. Of the remaining 29, 29 are also wrong in the Text control on the same instance, which leaves at most 0 errors (0\% of all Visual calls) that could be attributed to perception. Pooled over all configurations, 68\% of Visual errors come with an exact transcription.}
\newcommand{\AttribPooledPct}{68}
\newcommand{\TwinSummary}{For \texttt{gemini-3.8-flash} (low): $\sST$ 38\% vs.\ $\sAM$ 42\% ($b/c=4/7$, $p=0.55$); $\sST$ 38\% vs.\ $\sPATH$ 39\% ($b/c=4/5$, $p=1.00$); $\sPM$ 70\% vs.\ $\sAM$ 44\% ($b/c=13/0$, $p=0.00024$); $\chi$ 100\% vs.\ $\sCC$ 100\% ($b/c=0/0$, $p=1.00$); $\sECd$ 97\% vs.\ $\sEC$ 72\% ($b/c=8/0$, $p=0.0078$). The one twin that runs the other way is $\sST$ vs.\ $\sPATH$: the path counts are small (median 268 against 540011 for spanning trees), so enumeration, whose cost is proportional to the answer, is feasible for the $\shP$ member and the determinant is not what the model computes for the $\GapL$ member. This is the magnitude confound the matched twins are designed to remove, and it is why we treat $\sST$/$\sAM$, $\sPM$/$\sAM$ and $\sECd$/$\sEC$ as the primary tests. Over all 69 (configuration, twin) pairs with at least ten paired instances, the tractable member has the higher accuracy in 34, the two tie in 13, and the intractable member is higher in 22 (\texttt{gemini-3-flash-preview} (low): $\sST$ vs.\ $\sPATH$; \texttt{gemini-3-flash-preview} (low): $\chi$ vs.\ $\sCC$; \texttt{gemini-3-flash-preview} (low): $\sECd$ vs.\ $\sEC$; \texttt{gemini-3.1-flash-lite} (low): $\sST$ vs.\ $\sPATH$; \texttt{gemini-3.1-flash-lite} (low): $\chi$ vs.\ $\sCC$; \texttt{gemini-3.5-flash} (low): $\sST$ vs.\ $\sAM$). The exact McNemar test rejects equality at $p<0.05$ in 12 pairs in the predicted direction and in 3 in the opposite direction.}
\newcommand{\SizeSummary}{on grid drawings, exact match falls between $3{\times}3$ and $6{\times}6$ from 100\% to 0\% for $\sST$, from 100\% ($3{\times}4$) to 10\% for $\sPM$, from 100\% to 0\% for $\sAM$ and from 100\% to 0\% for $\sPATH$; on binary images $\chi$ goes from 100\% to 100\% and $\sCC$ from 100\% to 100\%; on Eulerian drawings $\sECd$ goes from 100\% (S) to 88\% (XL) and $\sEC$ from 88\% to 50\%.}
\newcommand{\BudgetSummary}{Class averages ($\TC$/$\Lsp$/$\GapL$/$\shP$, exact match \%) on the instances shared by all levels of a family: \texttt{gemini-3.8-flash} low/medium/high gives 100/100/69/67 $\to$ 100/100/86/78 $\to$ 100/100/92/83 (152 common instances); \texttt{gemini-3.6-flash} minimal/low/high gives 100/95/29/26 $\to$ 100/100/47/41 $\to$ 100/100/68/57 (202 common instances); Gemma~4 31B thinking off/on gives --/--/50/27 $\to$ --/--/58/45 (23 common instances); \texttt{gemini-3.1-pro-preview} low/high gives 100/100/43/45 $\to$ 100/100/69/63 (115 common instances). For \texttt{gemini-3.8-flash} the gap between the bottom two rungs and the $\shP$ rung is 33 points at low and 17 points at high thinking, so the ordering is sharpest when the budget is smallest.}
\newcommand{\TokenSummary}{For \texttt{gemini-3.8-flash} (low) the mean thinking plus output tokens per call are 1.4k ($\TC$), 1.1k ($\Lsp$), 5.2k ($\GapL$) and 6.1k ($\shP$); restricted to correct answers they are 1.4k, 1.1k, 5.2k and 7.3k. At the high setting the same ordering holds with 8.8k, 3.4k, 25.2k and 29.8k.}
\newcommand{\HintSummary}{For \texttt{gemini-3.6-flash} (high) (202 instances) the hint moves exact match by $-$6 points on $\sST$ (51$\to$46), +0 points on $\sPM$ (72$\to$72), +0 points on $\sECd$ (100$\to$100), and by $-$14 on $\sAM$, $-$3 on $\sPATH$, +6 on $\sEC$ (+0 on $\chi$, +0 on $\sCC$). For \texttt{gemini-3.8-flash} (low) (391 instances) the hint moves exact match by +8 points on $\sST$ (37$\to$45), $-$6 points on $\sPM$ (70$\to$64), +3 points on $\sECd$ (97$\to$100), and by $-$2 on $\sAM$, +2 on $\sPATH$, +22 on $\sEC$ (+0 on $\chi$, +0 on $\sCC$). Averaged over the 4 hinted configurations and their tasks, the hint adds $-$3 points on the $\GapL$ tasks and +0 points on the $\shP$ tasks.}
\newcommand{\CodeSummary}{With \texttt{gemini-3.6-flash} (low), Code reaches 94\% exact against 55\% for Visual on the same 202 instances; 54\% of its programs for $\sST$, $\sPM$, $\sECd$ compute a determinant or Pfaffian and 22\% of its programs for the $\shP$ tasks use dynamic programming, the rest enumerate; 3\% of programs time out at 60\,s (5 of the 6 time-outs on $\shP$ tasks). With \texttt{gemini-3.8-flash} (low), Code reaches 93\% exact against 63\% for Visual on the same 392 instances; 62\% of its programs for $\sST$, $\sPM$, $\sECd$ compute a determinant or Pfaffian and 27\% of its programs for the $\shP$ tasks use dynamic programming, the rest enumerate; 0\% of programs time out at 60\,s (0 of the 0 time-outs on $\shP$ tasks); with text input (Text$\to$Code) the figures are 96\%, 64\% algebraic and 32\% DP, so the algorithm choice is not a perception artefact. With Gemma~4 31B (thinking off), Code reaches 67\% exact against 33\% for Visual on the same 392 instances; 67\% of its programs for $\sST$, $\sPM$, $\sECd$ compute a determinant or Pfaffian and 33\% of its programs for the $\shP$ tasks use dynamic programming, the rest enumerate; 0\% of programs time out at 60\,s (0 of the 0 time-outs on $\shP$ tasks). With \texttt{gemini-3.1-pro-preview} (low), Code reaches 86\% exact against 58\% for Visual on the same 202 instances; 62\% of its programs for $\sST$, $\sPM$, $\sECd$ compute a determinant or Pfaffian and 27\% of its programs for the $\shP$ tasks use dynamic programming, the rest enumerate; 0\% of programs time out at 60\,s (0 of the 0 time-outs on $\shP$ tasks).}
\newcommand{\AlgoScaleSummary}{For \texttt{gemini-3.8-flash} (low) the share of Code programs that use a dynamic program for $\sAM$ rises with the instance: 0/6 at $3{\times}3$, 1/10 at $3{\times}4$, 4/10 at $4{\times}4$, 9/10 at $4{\times}5$, 10/10 at $5{\times}6$ and 10/10 at $6{\times}6$, and every program from $7{\times}7$ on is a dynamic program (Text$\to$Code: 0/6, 2/10, then 10/10 from $4{\times}4$ on). For $\sST$ the programs compute the Kirchhoff determinant at every size (6/6 at $3{\times}3$, 4/6 at $10{\times}10$), whereas for $\sPM$ the Kasteleyn determinant, which needs a planar orientation, is almost never written below $8{\times}8$ (0/10 at $6{\times}6$) and appears only on the largest grids (3/6 at $10{\times}10$); $\sPATH$ programs enumerate by depth-first search at every size (1/10 use a DP at $6{\times}6$), which is harmless because path counts stay small.}
\newcommand{\CarSummary}{On the core benchmark the code-producing conditions are within a few points of one another: on \texttt{gemini-3.6-flash} (low) (202 instances) \car{} 95\% at 1.9k tokens, Code 94\% at 1.6k tokens; on \texttt{gemini-3.8-flash} (low) (392 instances) \car{} 91\% at 1.5k tokens, Code 93\% at 1.7k tokens, Perceive+Code 93\% at 1.5k tokens; with the image replaced by text (392 instances) Text\,\car{} 98\% against Text$\to$Code 96\%; on Gemma~4 31B (thinking off) (202 instances) \car{} 65\% at 1.3k tokens, Code 69\% at 0.9k tokens, Perceive+Code 64\% at 1.1k tokens; on Gemma~4 31B (thinking on) (17 instances) \car{} 94\% at 7.6k tokens, Code 94\% at 8.1k tokens; on \texttt{gemini-3.1-pro-preview} (low) (202 instances) \car{} 88\% at 2.4k tokens, Code 86\% at 2.5k tokens, Perceive+Code 91\% at 2.2k tokens. On the $7{\times}7$ to $10{\times}10$ grids (answers $10^{5}$ to $10^{37}$; instances shared by all conditions of a configuration) the solved counts are: \texttt{gemini-3.6-flash} (low): Code 9/60, \car{} 15/60, Text$\to$Code 53/60, Text\,\car{} 51/60. \texttt{gemini-3.8-flash} (low) (60 instances per condition): $7{\times}7$: Visual 0/12, Text 0/12, Code 5/12, Perceive+Code 8/12, \car{} 9/12, Text$\to$Code 12/12, Text\,\car{} 12/12; $8{\times}8$: Visual 0/18, Text 0/18, Code 1/18, Perceive+Code 5/18, \car{} 5/18, Text$\to$Code 17/18, Text\,\car{} 14/18; $9{\times}9$: Visual 0/12, Text 0/12, Code 0/12, Perceive+Code 1/12, \car{} 0/12, Text$\to$Code 11/12, Text\,\car{} 11/12; $10{\times}10$: Visual 0/18, Text 0/18, Code 0/18, Perceive+Code 0/18, \car{} 0/18, Text$\to$Code 17/18, Text\,\car{} 17/18; in total Visual 0/60, Text 0/60, Code 6/60, Perceive+Code 14/60, \car{} 14/60, Text$\to$Code 57/60, Text\,\car{} 54/60. Gemma~4 31B (thinking off): Visual 0/60, Code 0/60, \car{} 0/60, Text\,\car{} 51/60. \texttt{gemini-3.1-pro-preview} (low): Visual 0/59, Code 2/59, \car{} 2/59, Text\,\car{} 59/59.}
\newcommand{\CarClassAcc}{96\%}

\newcommand{\CarAblSummary}{On \texttt{gemini-3.8-flash} (low), Perceive+Code, which keeps the transcription step but drops the classification and routing sentences, reaches 93\% on the 392 instances shared with \car{} (91\%), with an algebraic rate of 66\% and a DP rate of 23\% against 72\% and 41\% for \car{}; plain Code is at 93\%. Text\,\car{} reaches 98\% against 91\% for \car{} on the same 392 instances, so replacing the image by the exact edge list changes the result by +8 points: what separates the conditions is the input the program is run on, not the routing sentences.}
\newcommand{\GenSummary}{On the 194 instances shared by all generations (\texttt{gemini-3-flash-preview} (low), \texttt{gemini-3.1-flash-lite} (low), \texttt{gemini-3.5-flash} (low), \texttt{gemini-3.7-flash} (low), \texttt{gemini-3.8-flash} (low)), the $\TC$ average goes from 95\% to 100\%; the $\Lsp$ average goes from 100\% to 100\%; the $\GapL$ average goes from 19\% to 62\%; the $\shP$ average goes from 26\% to 47\%. The answer magnitude at which exact match crosses 50\% (Visual, pooled over the $\GapL$ and $\shP$ tasks of the grid and Eulerian families) is \texttt{gemini-3-flash-preview} (low) $10^{1.0}$, \texttt{gemini-3.1-flash-lite} (low) $10^{0.4}$, \texttt{gemini-3.5-flash} (low) $10^{2.5}$, \texttt{gemini-3.7-flash} (low) $10^{3.2}$, \texttt{gemini-3.8-flash} (low) $10^{3.2}$. Gemma~4 on its 23 shared instances: Gemma~4 26B-A4B (thinking off) --/--/25/18; Gemma~4 31B (thinking off) --/--/50/27; Gemma~4 31B (thinking on) --/--/58/45 ($\TC$/$\Lsp$/$\GapL$/$\shP$).}
\newcommand{\HeldoutSummary}{\texttt{gemini-3.8-flash} (low) (198 instances per condition): \car{} 90\% overall: 88\% \#IS, 88\% \#Dom, 94\% \#ShortestPaths (DP in 81\% of programs); Visual 56\% overall: 62\% \#IS, 15\% \#Dom, 91\% \#ShortestPaths; Code 90\% overall: 89\% \#IS, 86\% \#Dom, 94\% \#ShortestPaths (DP in 65\% of programs). Gemma~4 31B (thinking off) (198 instances per condition): \car{} 43\% overall: 48\% \#IS, 36\% \#Dom, 44\% \#ShortestPaths (DP in 88\% of programs).}
\newcommand{\ClassifierNote}{We validated the labeller by hand on three stratified samples of 60 programs each, drawn across configurations, conditions and tasks. The first sample was used to design the rules and ended at 58/60 agreement; a second sample held out from rule design gave 55/60, with all five disagreements on the held-out tasks ($\sSHORT$ and $\sDOM$), which led to a per-task rule for $\sSHORT$ and an all-subsets-enumeration signal; a third, fresh sample labelled with the final rules gave 60/60. On that third sample a first version of the labeller that matched keywords anywhere in the program, comments included, disagreed with the hand labels on 11 of 60 programs, mostly bitmask enumerations labelled DP and memoised DPs labelled algebraic because a comment mentioned Kasteleyn. A recurring pattern is that a program \emph{states} a polynomial method (Kasteleyn, BEST, a frontier DP) and \emph{implements} memoised bitmask recursion or plain enumeration, which is why we report the implemented algorithm, not the stated one.}
\newcommand{\CodeAttribSummary}{For \texttt{gemini-3.8-flash} (low) on all grid instances including the large ones, \car{}: 101 wrong programs out of 506, of which 53 compute the right answer for a wrongly transcribed graph, 9 compute a wrong answer for the right graph, 13 do both and 26 embed no explicit edge list; Code: 101 wrong programs out of 506, of which 37 compute the right answer for a wrongly transcribed graph, 3 compute a wrong answer for the right graph, 10 do both and 51 embed no explicit edge list; Perceive+Code: 71 wrong programs out of 308, of which 32 compute the right answer for a wrongly transcribed graph, 6 compute a wrong answer for the right graph, 12 do both and 21 embed no explicit edge list. With text input every error is an algorithm error (Text\,\car{} 11 of 12, Text$\to$Code 17 of 17), as it must be. The extractor is conservative: a program whose embedded list disagrees with the truth while its answer is correct is counted as a mismatch, and this happens in 11 of 296 correct Text\,\car{} programs, which bounds the rate at which the ``transcription'' label could be an artefact of extraction.}
\newcommand{\MagnitudeSummary}{For \texttt{gemini-3.8-flash} (low) on the grid and Eulerian families, under Visual (372 calls) the odds of an exact answer fall by a factor 2.81 per decade of the answer, magnitude alone explains a deviance of 268 against 3 for the class alone, and adding the class to the magnitude model changes the deviance by 2.9 ($p=0.09$; coefficient $-$0.56 for $\shP$); the answer size at which accuracy crosses 50\% is $10^{3.5}$ for the $\GapL$ tasks and $10^{3.0}$ for the $\shP$ tasks; under Text (371 calls) the odds of an exact answer fall by a factor 3.02 per decade of the answer, magnitude alone explains a deviance of 267 against 4 for the class alone, and adding the class to the magnitude model changes the deviance by 3.9 ($p=0.049$; coefficient $-$0.65 for $\shP$); the answer size at which accuracy crosses 50\% is $10^{3.4}$ for the $\GapL$ tasks and $10^{2.7}$ for the $\shP$ tasks.}

\newcommand{\BudgetDecadesSummary}{Within a family the crossing magnitude moves with the budget: \texttt{gemini-3.6-flash} (minimal) $10^{1.2}$ at 1.8k tokens, \texttt{gemini-3.6-flash} (low) $10^{2.2}$ at 7.0k tokens, \texttt{gemini-3.6-flash} (high) $10^{3.9}$ at 25.4k tokens; \texttt{gemini-3.8-flash} (low) $10^{3.2}$ at 5.7k tokens, \texttt{gemini-3.8-flash} (medium) $10^{4.5}$ at 13.2k tokens, \texttt{gemini-3.8-flash} (high) $10^{7.1}$ at 27.4k tokens. Across the 14 configurations, $m_{50}$ grows by 3.54 decades per tenfold increase in tokens.}
\newcommand{\ZoomSummary}{Table~\ref{tab:zoom} reports, for \texttt{gemini-3.8-flash} (low) on the 24 extra-large grids, how often the transcription is exact and how many of the 60 pairs the program then solves. Cutting the image into crops does not help: Zoom-1 transcribes 13/24 grids exactly and solves 31/60 pairs; Zoom-2 transcribes 1/24 grids exactly and solves 2/60 pairs; Zoom-3 transcribes 7/24 grids exactly and solves 14/60 pairs, against 14/60 for \car{} from the whole image. The errors are not a matter of resolution. In the whole-image reading 38 of 1384 horizontal edges are missed against 6 of 1393 vertical ones (with 4+3 spurious edges); a horizontal segment whose two row-neighbours are also present is missed 3.8\% of the time, other horizontal segments 1.3\%, and the corresponding vertical segments 0.8\%. Inside the Zoom-2 and Zoom-3 crops the same interior horizontal segments are missed 13.1\% of the time (274/2087) and the vertical ones 0.09\% (2/2127): zooming in makes the failure worse, not better. Asking for a column-by-column reading keeps the asymmetry (81 horizontal against 15 vertical edges missed), whereas rotating the image by $90^\circ$ moves it to the edges that are now horizontal on the page (101 of the original vertical edges missed against 27 of the horizontal ones). The failure follows the orientation of the line on the page, not the reading order: it is a property of the visual front end. The two readings are complementary: their union misses 0 of the 2777 true edges and adds 37 spurious ones, so every residual error is a disagreement between the two views. Checking each disputed pair with a crop of the two circles (8.6 checks per grid, 66 tokens each) answers 207/207 of them correctly and makes the transcription exact on 22/24 grids; the programs then solve 43/60 pairs with Text\,\car{} ($7{\times}7$ 10/12, $8{\times}8$ 14/18, $9{\times}9$ 7/12, $10{\times}10$ 12/18) and 49/60 with Text$\to$Code, against 14/60 from a one-shot reading and 54/60 from the ground-truth text.}

\newcommand{\OrientSummary}{The errors are not a matter of resolution. In the whole-image reading 38 of 1384 horizontal edges are missed against 6 of 1393 vertical ones (with 4+3 spurious edges); a horizontal segment whose two row-neighbours are also present is missed 3.8\% of the time, other horizontal segments 1.3\%, and the corresponding vertical segments 0.8\%. Inside the Zoom-2 and Zoom-3 crops the same interior horizontal segments are missed 13.1\% of the time (274/2087) and the vertical ones 0.09\% (2/2127): zooming in makes the failure worse, not better. Asking for a column-by-column reading keeps the asymmetry (81 horizontal against 15 vertical edges missed), whereas rotating the image by $90^\circ$ moves it to the edges that are now horizontal on the page (101 of the original vertical edges missed against 27 of the horizontal ones). The failure follows the orientation of the line on the page, not the reading order: it is a property of the visual front end. The two readings are complementary: their union misses 0 of the 2777 true edges and adds 37 spurious ones, so every residual error is a disagreement between the two views. Checking each disputed pair with a crop of the two circles (8.6 checks per grid, 66 tokens each) answers 207/207 of them correctly and makes the transcription exact on 22/24 grids; the programs then solve 43/60 pairs with Text\,\car{} ($7{\times}7$ 10/12, $8{\times}8$ 14/18, $9{\times}9$ 7/12, $10{\times}10$ 12/18) and 49/60 with Text$\to$Code, against 14/60 from a one-shot reading and 54/60 from the ground-truth text.}
\newcommand{\VerifyNumbers}{43/60}
\newcommand{\TwoViewOthers}{The same pipeline, unchanged, transfers to other configurations (Table~\ref{tab:twoview_models}): \texttt{gemini-3.6-flash} (low) transcribes 18/24 grids exactly after verification (11/24 one-shot) and solves 40/60 pairs with Text\,\car{} against 15/60 for one-shot \car{} and 51/60 from the ground-truth text; \texttt{gemini-3.1-pro-preview} (low) transcribes 15/24 grids exactly after verification (7/24 one-shot) and solves 36/60 pairs with Text\,\car{} against 2/60 for one-shot \car{} and 60/60 from the ground-truth text.}
\newcommand{\nBenchPairs}{392}
\newcommand{\nXLPairs}{60}

%% file: gen_v3/numbers_v3.tex
\newcommand{\vThreePooledPairs}{921}

\newcommand{\vThreeSeedPairs}{529}
\newcommand{\vThreeSeedRecords}{4232}
\newcommand{\vThreeSpendUSD}{130}

\newcommand{\vThreeStylePairs}{1312}
\newcommand{\vThreeStyleRecords}{2624}

%% file: gen_v4/numbers_v4.tex
\newcommand{\vFourBudgetCIhi}{5.0}
\newcommand{\vFourBudgetCIlo}{2.1}
\newcommand{\vFourBudgetFamGThreeEight}{5.6}
\newcommand{\vFourBudgetFamGThreeSix}{2.4}
\newcommand{\vFourBudgetK}{14}
\newcommand{\vFourBudgetLOOhi}{3.8}
\newcommand{\vFourBudgetLOOlo}{2.8}
\newcommand{\vFourBudgetModels}{9}
\newcommand{\vFourBudgetP}{$1.5\times 10^{-4}$}
\newcommand{\vFourBudgetRTwo}{0.71}
\newcommand{\vFourBudgetSe}{0.65}
\newcommand{\vFourBudgetSlope}{3.54}

\newcommand{\vFourCCCoreNoxlGThreeEightLowCarCodeAccA}{90.6}

\newcommand{\vFourCCCoreNoxlGThreeEightLowCarCodeAccB}{92.9}

\newcommand{\vFourCCCoreNoxlGThreeEightLowCarCodeN}{392}

\newcommand{\vFourCCCoreNoxlGThreeEightLowCarCodeP}{0.18}

\newcommand{\vFourCCCoreNoxlGThreeEightLowTcarTcodeCIhi}{+5.4}
\newcommand{\vFourCCCoreNoxlGThreeEightLowTcarTcodeCIlo}{+0.3}
\newcommand{\vFourCCCoreNoxlGThreeEightLowTcarTcodeDiff}{+2.8}

\newcommand{\vFourCCCoreNoxlGmThreeOneMinCarCodeCIhi}{-1.1}
\newcommand{\vFourCCCoreNoxlGmThreeOneMinCarCodeCIlo}{-10.4}
\newcommand{\vFourCCCoreNoxlGmThreeOneMinCarCodeDiff}{-5.6}

\newcommand{\vFourCCPooledNoxlGThreeEightLowCarCodeCIhi}{+1.3}
\newcommand{\vFourCCPooledNoxlGThreeEightLowCarCodeCIlo}{-2.6}
\newcommand{\vFourCCPooledNoxlGThreeEightLowCarCodeDiff}{-0.7}

\newcommand{\vFourCCPooledNoxlGThreeEightLowCarCodeN}{921}

\newcommand{\vFourCCPooledNoxlGThreeEightLowCarCodeP}{0.59}

\newcommand{\vFourCCXlGThreeEightLowCarCodeCIhi}{+27.9}
\newcommand{\vFourCCXlGThreeEightLowCarCodeCIlo}{+1.5}
\newcommand{\vFourCCXlGThreeEightLowCarCodeDiff}{+13.3}

\newcommand{\vFourCCXlGThreeEightLowCarCodeP}{0.039}

\newcommand{\vFourCovPooledNoxlVisualEulerP}{0.23}

\newcommand{\vFourDecPooledNoxlCarImagesAllTranscriptionsIdentical}{95}
\newcommand{\vFourDecPooledNoxlCarImagesWithGeTwoScorable}{153}

\newcommand{\vFourDecPooledNoxlCarPAOneGivenTOne}{99}

\newcommand{\vFourDecPooledNoxlCarRecords}{593}

\newcommand{\vFourDecPooledNoxlCarTZeroAOne}{44}

\newcommand{\vFourDecPooledNoxlCarTwrongCountAvailable}{94}

\newcommand{\vFourDecPooledNoxlCarTwrongExecMatchesGhat}{43}
\newcommand{\vFourDecPooledNoxlCarUnscorable}{80}

\newcommand{\vFourDecPooledNoxlCodeRecords}{593}

\newcommand{\vFourDecPooledNoxlCodeUnscorable}{127}

\newcommand{\vFourDecVisPooledNoxlPAOneGivenTOne}{50}

\newcommand{\vFourDecVisPooledNoxlScorable}{589}
\newcommand{\vFourDecVisPooledNoxlShareErrorsWithTOne}{79}
\newcommand{\vFourDecVisPooledNoxlTOneAOne}{256}
\newcommand{\vFourDecVisPooledNoxlTOneAZero}{261}
\newcommand{\vFourDecVisPooledNoxlTZeroAOne}{1}
\newcommand{\vFourDecVisPooledNoxlTZeroAZero}{71}

\newcommand{\vFourDecXlCarTOneAOne}{10}

\newcommand{\vFourDecXlCarTZeroAZero}{28}

\newcommand{\vFourDecXlCarTwrongExecMatchesGhat}{22}

\newcommand{\vFourMagCoreNoxlTextBcls}{-0.70}

\newcommand{\vFourMagCoreNoxlTextMfiftyDiff}{0.8}
\newcommand{\vFourMagCoreNoxlTextMfiftyDiffCIhi}{1.6}
\newcommand{\vFourMagCoreNoxlTextMfiftyDiffCIlo}{0.0}

\newcommand{\vFourMagCoreNoxlTextP}{0.037}

\newcommand{\vFourMagCoreNoxlVisualBcls}{-0.62}

\newcommand{\vFourMagCoreNoxlVisualMfiftyDiff}{0.6}
\newcommand{\vFourMagCoreNoxlVisualMfiftyDiffCIhi}{1.4}
\newcommand{\vFourMagCoreNoxlVisualMfiftyDiffCIlo}{-0.1}

\newcommand{\vFourMagCoreNoxlVisualN}{312}

\newcommand{\vFourMagCoreNoxlVisualORinv}{2.74}
\newcommand{\vFourMagCoreNoxlVisualP}{0.062}

\newcommand{\vFourMagPooledNoxlTextBcls}{-0.79}

\newcommand{\vFourMagPooledNoxlTextMfiftyDiffCIhi}{1.6}
\newcommand{\vFourMagPooledNoxlTextMfiftyDiffCIlo}{0.5}
\newcommand{\vFourMagPooledNoxlTextMfiftyHard}{2.5}
\newcommand{\vFourMagPooledNoxlTextMfiftyPoly}{3.5}
\newcommand{\vFourMagPooledNoxlTextN}{720}

\newcommand{\vFourMagPooledNoxlTextORinv}{2.41}
\newcommand{\vFourMagPooledNoxlTextP}{$8.1\times 10^{-5}$}

\newcommand{\vFourMagPooledNoxlTextSEcls}{0.20}

\newcommand{\vFourMagPooledNoxlVisualBcls}{-0.75}

\newcommand{\vFourMagPooledNoxlVisualMfiftyDiff}{1.0}
\newcommand{\vFourMagPooledNoxlVisualMfiftyDiffCIhi}{1.5}
\newcommand{\vFourMagPooledNoxlVisualMfiftyDiffCIlo}{0.4}
\newcommand{\vFourMagPooledNoxlVisualMfiftyHard}{2.9}
\newcommand{\vFourMagPooledNoxlVisualMfiftyPoly}{3.8}
\newcommand{\vFourMagPooledNoxlVisualN}{721}

\newcommand{\vFourMagPooledNoxlVisualORinv}{2.38}
\newcommand{\vFourMagPooledNoxlVisualP}{$2.4\times 10^{-4}$}

\newcommand{\vFourMagPooledNoxlVisualSEcls}{0.21}

\newcommand{\vFourNUniqueCalls}{21,534}

\newcommand{\vFourOrientAnyCallsRot}{18}
\newcommand{\vFourOrientAnyCallsUp}{27}

\newcommand{\vFourOrientAnyPcalls}{0.16}

\newcommand{\vFourOrientCalls}{246}

\newcommand{\vFourOrientImages}{66}

\newcommand{\vFourOrientMissHCallsUp}{27}

\newcommand{\vFourOrientMissHImgUp}{11}

\newcommand{\vFourOrientMissHPcalls}{$1.5\times 10^{-8}$}
\newcommand{\vFourOrientMissHPimages}{$9.8\times 10^{-4}$}

\newcommand{\vFourOrientMissVCallsRot}{17}

\newcommand{\vFourOrientMissVImgRot}{8}

\newcommand{\vFourOrientMissVPcalls}{$1.5\times 10^{-5}$}
\newcommand{\vFourOrientMissVPimages}{0.0078}

\newcommand{\vFourOrientXLExactRot}{2}

\newcommand{\vFourOrientXLExactUp}{13}

\newcommand{\vFourOrientXLMissingGridsRot}{22}
\newcommand{\vFourOrientXLMissingGridsUp}{9}

\newcommand{\vFourOrientXLMissingP}{$2.4\times 10^{-4}$}

\newcommand{\vFourTwinsAllCoreNTests}{153}

\newcommand{\vFourTwinsAllCoreSigBh}{14}
\newcommand{\vFourTwinsAllCoreSigHolm}{4}
\newcommand{\vFourTwinsAllCoreSigRaw}{32}

\newcommand{\vFourTwinsMainTableVisualNTests}{24}

\newcommand{\vFourTwinsMainTableVisualSigHolm}{2}
\newcommand{\vFourTwinsMainTableVisualSigRaw}{5}

\newcommand{\vFourTwinsPooledNTests}{40}

\newcommand{\vFourTwinsPooledSigBh}{13}
\newcommand{\vFourTwinsPooledSigHolm}{10}
\newcommand{\vFourTwinsPooledSigRaw}{16}

\newcommand{\vFourTwinsPooledVThreeOnlyNTests}{40}

\newcommand{\vFourTwinsPooledVThreeOnlySigBh}{10}
\newcommand{\vFourTwinsPooledVThreeOnlySigHolm}{6}
\newcommand{\vFourTwinsPooledVThreeOnlySigRaw}{10}

%% file: gen_v4/numbers_v41.tex
\newcommand{\vFourOneAAnswerExactPairs}{897}
\newcommand{\vFourOneAAnswerExactPct}{97}
\newcommand{\vFourOneABoth}{23}

\newcommand{\vFourOneACGivenPzeroPct}{43}
\newcommand{\vFourOneACOnly}{389}
\newcommand{\vFourOneACancel}{0}

\newcommand{\vFourOneAHardCGivenPzeroPct}{66}

\newcommand{\vFourOneAImages}{384}

\newcommand{\vFourOneAImagesExact}{378}
\newcommand{\vFourOneAImagesExactPct}{98}
\newcommand{\vFourOneAImagesParsed}{384}

\newcommand{\vFourOneAImgAllMissing}{11}
\newcommand{\vFourOneAImgAllUnits}{9003}

\newcommand{\vFourOneAPOnly}{1}
\newcommand{\vFourOneAPairs}{921}

\newcommand{\vFourOneAPairsOk}{921}

\newcommand{\vFourOneAPolyCGivenPzeroPct}{28}

\newcommand{\vFourOneAStructExactPairs}{897}
\newcommand{\vFourOneAStructExactPct}{97}

\newcommand{\vFourOneBBoneBothMissed}{10}

\newcommand{\vFourOneBBoneDisputedPerGrid}{2.6}
\newcommand{\vFourOneBBoneExact}{15}

\newcommand{\vFourOneBBoneExactCI}{$[-50.0, -8.3]$}
\newcommand{\vFourOneBBoneExactDiff}{-29.2}

\newcommand{\vFourOneBBoneExactP}{$0.0391$}

\newcommand{\vFourOneBBoneGrids}{24}

\newcommand{\vFourOneBBoneVcar}{35}

\newcommand{\vFourOneBBoneVcarN}{60}

\newcommand{\vFourOneBBoneVcode}{33}

\newcommand{\vFourOneBBoneVcodeN}{60}

\newcommand{\vFourOneBBthreeBothMissed}{12}

\newcommand{\vFourOneBBthreeDisputedPerGrid}{1.5}
\newcommand{\vFourOneBBthreeExact}{13}

\newcommand{\vFourOneBBthreeExactCI}{$[-58.3, -12.5]$}
\newcommand{\vFourOneBBthreeExactDiff}{-37.5}

\newcommand{\vFourOneBBthreeExactP}{$0.0117$}

\newcommand{\vFourOneBBthreeGrids}{24}

\newcommand{\vFourOneBBthreeVcar}{31}

\newcommand{\vFourOneBBthreeVcarN}{60}

\newcommand{\vFourOneBBthreeVcode}{28}

\newcommand{\vFourOneBBthreeVcodeN}{60}

\newcommand{\vFourOneBTVBothMissed}{0}

\newcommand{\vFourOneBTVDisputedPerGrid}{8.6}
\newcommand{\vFourOneBTVExact}{22}

\newcommand{\vFourOneBTVGrids}{24}

\newcommand{\vFourOneBTVVcar}{43}
\newcommand{\vFourOneBTVVcarN}{60}
\newcommand{\vFourOneBTVVcode}{49}
\newcommand{\vFourOneBTVVcodeN}{60}

\newcommand{\vFourOneCostTotalCalls}{1,815}
\newcommand{\vFourOneCostTotalUsd}{15.73}
\newcommand{\vFourOneModelVersions}{\texttt{gemini-3.8-flash} (1741)}

%% file: gen_v5/numbers_v5.tex
\newcommand{\vFiveBTVExact}{21}

\newcommand{\vFiveBTVBothMissed}{3}

\newcommand{\vFiveBTVVcar}{50}

\newcommand{\vFiveBTVVcode}{42}

\newcommand{\vFiveBBoneExact}{13}

\newcommand{\vFiveBBoneBothMissed}{11}

\newcommand{\vFiveBBoneVcar}{30}

\newcommand{\vFiveBBoneVcode}{26}

\newcommand{\vFiveBBthreeExact}{15}

\newcommand{\vFiveBBthreeBothMissed}{26}

\newcommand{\vFiveBBthreeVcar}{38}

\newcommand{\vFiveBBthreeVcode}{31}

\newcommand{\vFiveBBthreeExactP}{$0.109$}

\newcommand{\vFiveBBthreeVcarP}{$0.0227$}

\newcommand{\vFiveBTVFinalVsROneP}{$0.000977$}

\newcommand{\vFiveBBoneFinalVsROneP}{$0.125$}

\newcommand{\vFiveRTVcarB}{3}
\newcommand{\vFiveRTVcarC}{4}
\newcommand{\vFiveRTVcarFlips}{7}
\newcommand{\vFiveRTVcarP}{$1$}

\newcommand{\vFiveRTVcodeB}{4}
\newcommand{\vFiveRTVcodeC}{9}
\newcommand{\vFiveRTVcodeFlips}{13}
\newcommand{\vFiveRTVcodeP}{$0.267$}

\newcommand{\vFiveRTPairsTotal}{120}
\newcommand{\vFiveRTFlipsTotal}{20}
\newcommand{\vFiveRTFlipPct}{16.7}

\newcommand{\vFiveRAThreeIdentical}{11}

\newcommand{\vFiveRAThreeDisagreeMean}{4.2}

\newcommand{\vFiveSpanUtc}{2026-10-07 07:56 UTC to 19:09 UTC}
\newcommand{\vFiveSpanHours}{11.2}
\newcommand{\vFiveModelVersions}{gemini-3.8-flash}
\newcommand{\vFiveCallsTotal}{1047}
\newcommand{\vFiveUsdTotal}{8.24}

%% file: gen_v5/numbers_v5_503.tex
\newcommand{\vFiveHiVisKeys}{202}
\newcommand{\vFiveHiVisLost}{46}
\newcommand{\vFiveHiVisLostPct}{23}
\newcommand{\vFiveHiVisFitReturned}{117}
\newcommand{\vFiveHiVisFitLost}{45}
\newcommand{\vFiveHiVisMedKept}{1.8}
\newcommand{\vFiveHiVisMedLost}{3.8}
\newcommand{\vFiveHiVisMfiftyNow}{7.1}
\newcommand{\vFiveHiVisMfiftyPolyNow}{8.4}
\newcommand{\vFiveHiVisMfiftyHardNow}{5.9}
\newcommand{\vFiveHiVisFitTotal}{162}
\newcommand{\vFiveHiVisMfiftyLow}{4.4}
\newcommand{\vFiveHiVisMfiftyPolyLow}{6.0}
\newcommand{\vFiveHiVisMfiftyHardLow}{3.4}
\newcommand{\vFiveHiTextKeys}{202}
\newcommand{\vFiveHiTextLost}{44}

\newcommand{\vFiveHiTextMfiftyNow}{6.9}

\newcommand{\vFiveHiTextMfiftyLow}{4.3}

\newcommand{\vFiveMedVisLost}{15}
\newcommand{\vFiveMedVisKeys}{392}

\newcommand{\vFiveHiHintKeys}{202}
\newcommand{\vFiveHiHintLost}{38}
\newcommand{\vFiveGThreeSixVisLost}{1}
\newcommand{\vFiveLostCoreKeys}{189}
\newcommand{\vFiveAllCoreKeys}{12{,}874}

%% file: gen_v5/numbers_v5_503_slope.tex
\newcommand{\vFiveSlopeLow}{2.8}
\newcommand{\vFiveSlopeLowCIlo}{1.8}
\newcommand{\vFiveSlopeLowCIhi}{3.8}
\newcommand{\vFiveSlopeLowFamGThreeEight}{1.7}

%% file: gen_v6/numbers_v6_b7.tex
\newcommand{\vSixBSevenTVExact}{86}
\newcommand{\vSixBSevenTVNGrids}{96}

\newcommand{\vSixBSevenTVRTwoExact}{8}

\newcommand{\vSixBSevenTVBothMissed}{7}

\newcommand{\vSixBSevenTVNVerifyCorrect}{968}
\newcommand{\vSixBSevenTVNChecked}{968}

\newcommand{\vSixBSevenTVDisputedPerGrid}{10.1}

\newcommand{\vSixBSevenTVVcar}{195}

\newcommand{\vSixBSevenTVVcarGivenExact}{195}
\newcommand{\vSixBSevenTVVcarGivenExactN}{215}
\newcommand{\vSixBSevenTVVcode}{190}

\newcommand{\vSixBSevenTVSizeNineXNineExact}{19}

\newcommand{\vSixBSevenTVSizeOneZeroXOneZeroExact}{19}

\newcommand{\vSixBSevenRotExact}{86}

\newcommand{\vSixBSevenRotRTwoExact}{20}

\newcommand{\vSixBSevenRotBothMissed}{23}
\newcommand{\vSixBSevenRotBothMissedH}{23}

\newcommand{\vSixBSevenRotNVerifyCorrect}{852}
\newcommand{\vSixBSevenRotNChecked}{852}

\newcommand{\vSixBSevenRotDisputedPerGrid}{8.9}

\newcommand{\vSixBSevenRotVcar}{203}

\newcommand{\vSixBSevenRotVcarGivenExact}{203}
\newcommand{\vSixBSevenRotVcarGivenExactN}{217}
\newcommand{\vSixBSevenRotVcode}{194}

\newcommand{\vSixBSevenRotSizeNineXNineExact}{18}

\newcommand{\vSixBSevenRotSizeOneZeroXOneZeroExact}{22}

\newcommand{\vSixBSevenBthreeExact}{57}

\newcommand{\vSixBSevenBthreeRTwoExact}{45}

\newcommand{\vSixBSevenBthreeBothMissed}{113}

\newcommand{\vSixBSevenBthreeNVerifyCorrect}{273}
\newcommand{\vSixBSevenBthreeNChecked}{273}

\newcommand{\vSixBSevenBthreeDisputedPerGrid}{2.8}

\newcommand{\vSixBSevenBthreeVcar}{133}

\newcommand{\vSixBSevenBthreeVcarGivenExact}{133}
\newcommand{\vSixBSevenBthreeVcarGivenExactN}{143}
\newcommand{\vSixBSevenBthreeVcode}{128}

\newcommand{\vSixBSevenBthreeSizeNineXNineExact}{7}
\newcommand{\vSixBSevenBthreeSizeNineXNineN}{24}
\newcommand{\vSixBSevenBthreeSizeOneZeroXOneZeroExact}{7}

\newcommand{\vSixBSevenGTVcar}{222}

\newcommand{\vSixBSevenGTVcode}{211}

\newcommand{\vSixBSevenTVVsBthreeExactDiff}{+30.2}
\newcommand{\vSixBSevenTVVsBthreeExactB}{30}
\newcommand{\vSixBSevenTVVsBthreeExactC}{1}

\newcommand{\vSixBSevenTVVsBthreeExactPHolm}{$6.0{\times}10^{-8}$}
\newcommand{\vSixBSevenTVVsBthreeExactCI}{$[+20.8, +39.6]$}

\newcommand{\vSixBSevenRotVsBthreeExactDiff}{+30.2}
\newcommand{\vSixBSevenRotVsBthreeExactB}{32}
\newcommand{\vSixBSevenRotVsBthreeExactC}{3}

\newcommand{\vSixBSevenRotVsBthreeExactPHolm}{$4.2{\times}10^{-7}$}
\newcommand{\vSixBSevenRotVsBthreeExactCI}{$[+19.8, +40.6]$}

\newcommand{\vSixBSevenTVVsRotExactB}{6}
\newcommand{\vSixBSevenTVVsRotExactC}{6}
\newcommand{\vSixBSevenTVVsRotExactP}{$1$}
\newcommand{\vSixBSevenTVVsRotExactCI}{$[-7.3, +7.3]$}

\newcommand{\vSixBSevenTVVsRotVcarPHolm}{$0.699$}

\newcommand{\vSixBSevenTVVsRotVcodePHolm}{$0.724$}

\newcommand{\vSixBSevenORUpzeroMissHRatePct}{2.8}
\newcommand{\vSixBSevenORUpzeroMissVRatePct}{0.8}

\newcommand{\vSixBSevenORRotninetyMissHRatePct}{2.9}
\newcommand{\vSixBSevenORRotninetyMissVRatePct}{9.0}

\newcommand{\vSixBSevenORRotoneeightyMissHRatePct}{8.6}
\newcommand{\vSixBSevenORRotoneeightyMissVRatePct}{3.4}
\newcommand{\vSixBSevenRAFourN}{96}
\newcommand{\vSixBSevenRAFourIdentical}{26}
\newcommand{\vSixBSevenRAFourRawIdentical}{0}

\newcommand{\vSixBSevenRAFourDisagreeMean}{4.7}
\newcommand{\vSixBSevenPoolTVExact}{107}
\newcommand{\vSixBSevenPoolTVN}{120}
\newcommand{\vSixBSevenPoolRotExact}{108}

\newcommand{\vSixBSevenPoolBthreeExact}{74}

\newcommand{\vSixBSevenSpanUtc}{2026-10-07 23:28 UTC to 2026-10-08 03:41 UTC}

\newcommand{\vSixBSevenSpanHours}{4.2}
\newcommand{\vSixBSevenModelVersions}{gemini-3.8-flash}
\newcommand{\vSixBSevenCallsTotal}{4589}
\newcommand{\vSixBSevenUsdTotal}{29.81}
\newcommand{\vSixBSevenGrids}{96}
\newcommand{\vSixBSevenPairs}{240}

%% file: gen_v6/numbers_v6_x8.tex
\providecommand{\vSixXEightPartial}{}

\newcommand{\vSixXEightGThreeEightLowCodeK}{66}

\newcommand{\vSixXEightGThreeEightLowPcodeK}{55}

\newcommand{\vSixXEightGThreeEightLowCarK}{53}

\newcommand{\vSixXEightGThreeEightLowTcodeK}{216}

\newcommand{\vSixXEightGThreeEightLowTcarK}{224}

\newcommand{\vSixXEightGThreeSixLowCodeK}{41}

\newcommand{\vSixXEightGThreeSixLowPcodeK}{34}

\newcommand{\vSixXEightGThreeSixLowCarK}{43}

\newcommand{\vSixXEightGThreeSixLowTcodeK}{212}

\newcommand{\vSixXEightGThreeSixLowTcarK}{209}

\newcommand{\vSixXEightProLowCodeK}{13}

\newcommand{\vSixXEightProLowPcodeK}{12}

\newcommand{\vSixXEightProLowCarK}{14}

\newcommand{\vSixXEightProLowTcodeK}{238}

\newcommand{\vSixXEightProLowTcarK}{240}

\newcommand{\vSixXEightGmThreeOneMinTcarK}{202}

\newcommand{\vSixXEightHOneB}{147}
\newcommand{\vSixXEightHOneC}{5}
\newcommand{\vSixXEightHOneKa}{195}
\newcommand{\vSixXEightHOneKb}{53}

\newcommand{\vSixXEightHOnePHolm}{$4.6{\times}10^{-37}$}

\newcommand{\vSixXEightHTwoB}{21}
\newcommand{\vSixXEightHTwoC}{34}
\newcommand{\vSixXEightHTwoKa}{53}
\newcommand{\vSixXEightHTwoKb}{66}

\newcommand{\vSixXEightHTwoPHolm}{$0.105$}

\newcommand{\vSixXEightPcodeVsCodeB}{18}
\newcommand{\vSixXEightPcodeVsCodeC}{29}

\newcommand{\vSixXEightPcodeVsCodePHolm}{$0.432$}

\newcommand{\vSixXEightCarVsPcodeB}{22}
\newcommand{\vSixXEightCarVsPcodeC}{24}

\newcommand{\vSixXEightCarVsPcodePHolm}{$0.883$}

\newcommand{\vSixXEightTcodeVsCodeB}{154}
\newcommand{\vSixXEightTcodeVsCodeC}{4}

\newcommand{\vSixXEightTcodeVsCodePHolm}{$8.4{\times}10^{-40}$}

\newcommand{\vSixXEightTcarVsCarB}{177}
\newcommand{\vSixXEightTcarVsCarC}{6}

\newcommand{\vSixXEightTcarVsCarPHolm}{$5.7{\times}10^{-44}$}

\newcommand{\vSixXEightTvVsTcarB}{12}
\newcommand{\vSixXEightTvVsTcarC}{41}

\newcommand{\vSixXEightTvVsTcarPHolm}{$0.000327$}

\newcommand{\vSixXEightGThreeSixLowCarVsCodePHolm}{$0.883$}

\newcommand{\vSixXEightGThreeSixLowPcodeVsCodePHolm}{$0.698$}

\newcommand{\vSixXEightProLowCarVsCodePHolm}{$1$}

\newcommand{\vSixXEightProLowPcodeVsCodePHolm}{$1$}

\newcommand{\vSixXEightRetestTcarKa}{224}
\newcommand{\vSixXEightRetestTcarKb}{222}
\newcommand{\vSixXEightRetestTcarP}{$0.851$}

\newcommand{\vSixXEightRetestTcodeKa}{216}
\newcommand{\vSixXEightRetestTcodeKb}{211}
\newcommand{\vSixXEightRetestTcodeP}{$0.56$}

\newcommand{\vSixXEightDecGThreeEightLowVisualRecords}{240}

\newcommand{\vSixXEightDecGThreeEightLowVisualTOneAZero}{75}

\newcommand{\vSixXEightDecGThreeEightLowCodeRecords}{240}

\newcommand{\vSixXEightDecGThreeEightLowCodeScorable}{95}

\newcommand{\vSixXEightDecGThreeEightLowCodeTwrongCountAvailable}{62}
\newcommand{\vSixXEightDecGThreeEightLowCodeTwrongExecMatchesGhat}{42}

\newcommand{\vSixXEightDecGThreeEightLowPcodeTwrongCountAvailable}{96}
\newcommand{\vSixXEightDecGThreeEightLowPcodeTwrongExecMatchesGhat}{71}

\newcommand{\vSixXEightDecGThreeEightLowCarTwrongCountAvailable}{85}
\newcommand{\vSixXEightDecGThreeEightLowCarTwrongExecMatchesGhat}{71}

\newcommand{\vSixXEightDevGThreeEightLowCarP}{$0.863$}
\newcommand{\vSixXEightDevGThreeEightLowCodeK}{6}
\newcommand{\vSixXEightDevGThreeEightLowCodeN}{60}
\newcommand{\vSixXEightDevGThreeEightLowCodeP}{$0.00383$}

\newcommand{\vSixXEightDevGThreeEightLowPcodeP}{$1$}

\newcommand{\vSixXEightSpanUtc}{2026-10-08 18:11 UTC to 2026-10-09 07:37 UTC}

\newcommand{\vSixXEightSpanHours}{13.4}
\newcommand{\vSixXEightModelVersions}{gemini-3.1-pro-preview, gemini-3.6-flash, gemini-3.8-flash, gemma-4-31b-it}
\newcommand{\vSixXEightUsdTotal}{126.20}
\newcommand{\vSixXEightCallsOkTotal}{6720}

\newcommand{\vSixXEightPairs}{240}

%% file: gen_v6/numbers_v6_t6.tex
\providecommand{\vSixTSixPartial}{}

\newcommand{\vSixTSixHOneVisualBeta}{-0.61}
\newcommand{\vSixTSixHOneVisualCI}{$[-1.88, +0.38]$}
\newcommand{\vSixTSixHOneVisualP}{$0.908$}

\newcommand{\vSixTSixPoolVisualB}{73}
\newcommand{\vSixTSixPoolVisualC}{49}
\newcommand{\vSixTSixPoolVisualShare}{60}
\newcommand{\vSixTSixPoolVisualCI}{$[50, 69]$}
\newcommand{\vSixTSixPoolVisualP}{$0.0555$}

\newcommand{\vSixTSixTokVisualP}{$0.132$}

\newcommand{\vSixTSixHOneTextBeta}{-0.51}
\newcommand{\vSixTSixHOneTextCI}{$[-1.60, +0.35]$}
\newcommand{\vSixTSixHOneTextP}{$0.88$}

\newcommand{\vSixTSixPoolTextB}{86}
\newcommand{\vSixTSixPoolTextC}{58}
\newcommand{\vSixTSixPoolTextShare}{60}
\newcommand{\vSixTSixPoolTextCI}{$[51, 69]$}
\newcommand{\vSixTSixPoolTextP}{$0.037$}
\newcommand{\vSixTSixCellGThreeEightLowTextB}{22}
\newcommand{\vSixTSixCellGThreeEightLowTextC}{4}

\newcommand{\vSixTSixCellGThreeEightLowTextPHolm}{$0.00854$}

\newcommand{\vSixTSixTokTextP}{$0.708$}

\newcommand{\vSixTSixRiseGThreeEightTextDiff}{-18}
\newcommand{\vSixTSixRiseGThreeEightTextDiffCI}{$[-34, -3]$}

\newcommand{\vSixTSixOosVisualDll}{-0.008}
\newcommand{\vSixTSixOosVisualDllCI}{$[-0.016, -0.001]$}

\newcommand{\vSixTSixOosTextDll}{-0.014}
\newcommand{\vSixTSixOosTextDllCI}{$[-0.021, -0.007]$}

\newcommand{\vSixTSixOosBothAucOne}{0.92}

\newcommand{\vSixTSixOosBothCalSlopeOne}{0.99}
\newcommand{\vSixTSixOosBothCalIntOne}{1.56}

\newcommand{\vSixTSixPerceptB}{62}
\newcommand{\vSixTSixPerceptC}{43}
\newcommand{\vSixTSixPerceptShare}{59}

\newcommand{\vSixTSixPerceptP}{$0.112$}
\newcommand{\vSixTSixPerceptBeta}{-0.57}
\newcommand{\vSixTSixPerceptBetaCI}{$[-1.97, +0.51]$}
\newcommand{\vSixTSixCopyTwinsGThreeEightLowK}{160}
\newcommand{\vSixTSixCopyTwinsGThreeEightLowN}{160}

\newcommand{\vSixTSixCopyXlGThreeEightLowK}{60}
\newcommand{\vSixTSixCopyXlGThreeEightLowN}{60}

\newcommand{\vSixTSixAlgoGThreeEightMedAMN}{80}
\newcommand{\vSixTSixAlgoGThreeEightMedAMAlgo}{31}
\newcommand{\vSixTSixAlgoGThreeEightMedAMText}{47}
\newcommand{\vSixTSixAlgoGThreeEightMedAMB}{3}
\newcommand{\vSixTSixAlgoGThreeEightMedAMC}{19}
\newcommand{\vSixTSixAlgoGThreeEightMedAMP}{$0.000855$}
\newcommand{\vSixTSixAlgoGThreeEightMedAMMaxtokAlgo}{18}
\newcommand{\vSixTSixAlgoGThreeEightMedAMMaxtokText}{9}
\newcommand{\vSixTSixAlgoGThreeEightHighSTN}{79}
\newcommand{\vSixTSixAlgoGThreeEightHighSTAlgo}{69}
\newcommand{\vSixTSixAlgoGThreeEightHighSTText}{69}

\newcommand{\vSixTSixAlgoGThreeEightHighAMAlgo}{54}
\newcommand{\vSixTSixAlgoGThreeEightHighAMText}{66}

\newcommand{\vSixTSixAlgoGThreeEightHighAMP}{$0.00183$}
\newcommand{\vSixTSixAlgoGThreeEightHighAMMaxtokAlgo}{17}
\newcommand{\vSixTSixAlgoGThreeEightHighAMMaxtokText}{0}

\newcommand{\vSixTSixKeysOk}{3699}
\newcommand{\vSixTSixKeys}{3700}

\newcommand{\vSixTSixUsdTotal}{337.00}

\newcommand{\vSixTSixSpanUtc}{2026-10-07 23:09--2026-10-09 11:46 UTC}

\newcommand{\vSixTSixSpanHours}{36.6}
\newcommand{\vSixTSixModelVersions}{gemini-3.1-pro-preview, gemini-3.6-flash, gemini-3.8-flash}

%% file: gen_v6/numbers_v6_f6.tex
\providecommand{\vSixFPartial}{}

\newcommand{\vSixFDriftPhVisOldExact}{30}
\newcommand{\vSixFDriftPhVisNewExact}{31}
\newcommand{\vSixFDriftPhVisBRightToWrong}{1}
\newcommand{\vSixFDriftPhVisCWrongToRight}{2}

\newcommand{\vSixFDriftPhVisP}{1.00}

\newcommand{\vSixFDriftPhTextOldExact}{31}
\newcommand{\vSixFDriftPhTextNewExact}{31}

\newcommand{\vSixFDriftPhTextP}{1.00}

\newcommand{\vSixFProHighVisReturnedMfifty}{4.3}

\newcommand{\vSixFProHighVisMergedExact}{145}
\newcommand{\vSixFProHighVisMergedN}{202}
\newcommand{\vSixFProHighVisMergedMfifty}{4.1}

\newcommand{\vSixFProHighTextReturnedMfifty}{5.1}

\newcommand{\vSixFProHighTextMergedExact}{155}
\newcommand{\vSixFProHighTextMergedN}{202}
\newcommand{\vSixFProHighTextMergedMfifty}{5.0}

\newcommand{\vSixFSlopeReturnedTwo}{3.54}

\newcommand{\vSixFSlopeMergedTwo}{3.51}
\newcommand{\vSixFSlopeMergedCIlo}{2.1}
\newcommand{\vSixFSlopeMergedCIhi}{4.9}

\newcommand{\vSixFSlopeMergedCfiveTwo}{3.24}
\newcommand{\vSixFSlopeMergedCfiveCIlo}{2.1}
\newcommand{\vSixFSlopeMergedCfiveCIhi}{4.4}
\newcommand{\vSixFSlopeMergedCfiveP}{$5.1\times 10^{-5}$}
\newcommand{\vSixFSlopeMergedCfiveRTwo}{0.76}
\newcommand{\vSixFSlopeMergedCfiveK}{14}
\newcommand{\vSixFSlopeMergedCfiveFamGThreeEight}{3.9}
\newcommand{\vSixFSlopeMergedCfiveFamGThreeSix}{2.4}
\newcommand{\vSixFSlopeMergedCfiveFamPro}{1.9}

\newcommand{\vSixFXlHiVisMergedExact}{0}
\newcommand{\vSixFXlHiVisMergedN}{60}

\newcommand{\vSixFXlHiTextMergedExact}{0}
\newcommand{\vSixFXlHiTextMergedN}{60}

\newcommand{\vSixFTargetRequested}{273}
\newcommand{\vSixFTargetReturned}{260}
\newcommand{\vSixFTargetStillLost}{13}
\newcommand{\vSixFSpanUtc}{2026-10-07 23:00 to 2026-10-09 07:59 UTC}

\newcommand{\vSixFSpanHours}{33.0}
\newcommand{\vSixFUsdTotal}{95.32}

\newcommand{\vSixFModelVersions}{gemini-3.1-pro-preview, gemini-3.6-flash, gemini-3.8-flash, gemma-4-31b-it}

%% file: gen_v5/numbers_v5_c503.tex
\providecommand{\vFiveCPartial}{}

\newcommand{\vFiveCHiVisReturned}{46}
\newcommand{\vFiveCHiVisExact}{23}

\newcommand{\vFiveCHiTextReturned}{44}
\newcommand{\vFiveCHiTextExact}{26}

\newcommand{\vFiveCMedVisReturned}{15}
\newcommand{\vFiveCMedVisExact}{8}

\newcommand{\vFiveCHiHintReturned}{38}
\newcommand{\vFiveCHiHintExact}{18}

\newcommand{\vFiveCDriftCtlOldExact}{22}
\newcommand{\vFiveCDriftCtlNewExact}{21}
\newcommand{\vFiveCDriftCtlBRightToWrong}{2}
\newcommand{\vFiveCDriftCtlCWrongToRight}{1}
\newcommand{\vFiveCDriftCtlSamePred}{20}
\newcommand{\vFiveCDriftCtlSamePromptTokens}{36}
\newcommand{\vFiveCDriftCtlPerceptExactOld}{29}
\newcommand{\vFiveCDriftCtlPerceptExactNew}{31}

\newcommand{\vFiveCDriftCtlP}{1.00}

\newcommand{\vFiveCDriftCtlN}{36}
\newcommand{\vFiveCDriftCtlOldSpanUtc}{2026-10-03 00:14 to 12:28 UTC}
\newcommand{\vFiveCHiVisMfiftyReturned}{7.1}

\newcommand{\vFiveCHiVisMfiftyMerged}{6.1}
\newcommand{\vFiveCHiVisMfiftyMergedPoly}{7.2}
\newcommand{\vFiveCHiVisMfiftyMergedHard}{5.3}

\newcommand{\vFiveCHiTextMfiftyReturned}{6.9}

\newcommand{\vFiveCHiTextMfiftyMerged}{6.4}

\newcommand{\vFiveCMedVisMfiftyMerged}{4.5}

\newcommand{\vFiveCSlopeReturnedTwo}{3.54}

\newcommand{\vFiveCSlopeReturnedFamGThreeEight}{5.6}

\newcommand{\vFiveCHintReturned}{143}
\newcommand{\vFiveCHintReturnedN}{164}

\newcommand{\vFiveCSlopeMergedTwo}{3.27}
\newcommand{\vFiveCSlopeMergedCIlo}{2.1}
\newcommand{\vFiveCSlopeMergedCIhi}{4.4}

\newcommand{\vFiveCSlopeMergedFamGThreeEight}{3.9}

\newcommand{\vFiveCHintMerged}{161}
\newcommand{\vFiveCHintMergedN}{202}

\newcommand{\vFiveCReturnedLost}{143}
\newcommand{\vFiveCRequestedLost}{143}
\newcommand{\vFiveCSpanUtc}{2026-10-07 21:00 to 2026-10-08 21:30 UTC}

\newcommand{\vFiveCUsdTotal}{26.72}

\newcommand{\vFiveCModelVersions}{gemini-3.8-flash}

%% file: gen_v8/numbers_g9.tex
\providecommand{\gNinePartial}{}
\newcommand{\gNineTotalCalls}{34{,}592}
\newcommand{\gNineTotalMissing}{0}
\newcommand{\gNineTasksPerModel}{8{,}648}
\newcommand{\gNineNModels}{4}
\newcommand{\gNineNCells}{8}
\newcommand{\gNineHOneSupported}{7}
\newcommand{\gNineHTwoSupported}{7}
\newcommand{\gNineHThreeSupported}{0}

\newcommand{\gNineBetaLgMin}{-1.72}
\newcommand{\gNineBetaLgMax}{-1.22}
\newcommand{\gNineOddsDecMin}{0.18}
\newcommand{\gNineOddsDecMax}{0.30}
\newcommand{\gNineBetaBudMin}{1.18}
\newcommand{\gNineBetaBudMax}{1.98}
\newcommand{\gNineThirtyOneBVisualOddsDecade}{0.30}

\newcommand{\gNineThirtyOneBVisualHThreeB}{1}
\newcommand{\gNineThirtyOneBVisualHThreeC}{5}

\newcommand{\gNineThirtyOneBVisualMSlope}{0.79}

\newcommand{\gNineThirtyOneBVisualAccCoreFiveOneTwo}{24.5}

\newcommand{\gNineThirtyOneBVisualAccCorefree}{38.1}

\newcommand{\gNineThirtyOneBVisualMSlopeCILo}{0.55}
\newcommand{\gNineThirtyOneBVisualMSlopeCIHi}{1.06}
\newcommand{\gNineThirtyOneBVisualMFiftyTwoFiveOneTwo}{0.04}

\newcommand{\gNineThirtyOneBVisualTruncFiveOneTwo}{33.9}

\newcommand{\gNineThirtyOneBVisualMedRtokFiveOneTwo}{510}

\newcommand{\gNineThirtyOneBVisualPerceptExactFiveOneTwo}{34.5}

\newcommand{\gNineThirtyOneBVisualMedRtokTwoZeroFourEight}{2046}

\newcommand{\gNineThirtyOneBVisualTruncEightOneNineTwo}{1.0}

\newcommand{\gNineThirtyOneBVisualMedRtokEightOneNineTwo}{8190}

\newcommand{\gNineThirtyOneBVisualMFiftyTwofree}{1.15}

\newcommand{\gNineThirtyOneBVisualTruncfree}{0.1}

\newcommand{\gNineThirtyOneBVisualPerceptExactfree}{40.3}

\newcommand{\gNineThirtyOneBVisualClassBeta}{0.91}
\newcommand{\gNineThirtyOneBVisualClassSE}{0.24}

\newcommand{\gNineThirtyOneBTextOddsDecade}{0.23}

\newcommand{\gNineThirtyOneBTextHThreeB}{0}
\newcommand{\gNineThirtyOneBTextHThreeC}{8}

\newcommand{\gNineThirtyOneBTextMSlope}{0.88}

\newcommand{\gNineThirtyOneBTextAccCoreFiveOneTwo}{27.7}

\newcommand{\gNineThirtyOneBTextAccCorefree}{47.1}

\newcommand{\gNineThirtyOneBTextMSlopeCILo}{0.70}
\newcommand{\gNineThirtyOneBTextMSlopeCIHi}{1.12}
\newcommand{\gNineThirtyOneBTextMFiftyTwoFiveOneTwo}{0.43}

\newcommand{\gNineThirtyOneBTextMFiftyTwofree}{1.84}
\newcommand{\gNineThirtyOneBTextAccTwinfree}{10.0}

\newcommand{\gNineTwentySixBVisualOddsDecade}{0.18}

\newcommand{\gNineTwentySixBVisualHThreeB}{1}
\newcommand{\gNineTwentySixBVisualHThreeC}{1}

\newcommand{\gNineTwentySixBVisualMSlope}{1.03}

\newcommand{\gNineTwentySixBVisualAccCoreFiveOneTwo}{14.7}

\newcommand{\gNineTwentySixBVisualAccCorefree}{28.1}

\newcommand{\gNineTwentySixBVisualMSlopeCILo}{0.18}
\newcommand{\gNineTwentySixBVisualMSlopeCIHi}{2.73}

\newcommand{\gNineTwentySixBVisualPerceptExactFiveOneTwo}{35.5}

\newcommand{\gNineTwentySixBVisualPerceptExactTwoZeroFourEight}{27.6}

\newcommand{\gNineTwentySixBTextOddsDecade}{0.19}

\newcommand{\gNineTwentySixBTextHThreeB}{2}
\newcommand{\gNineTwentySixBTextHThreeC}{3}

\newcommand{\gNineTwentySixBTextMSlope}{0.62}

\newcommand{\gNineTwentySixBTextAccCoreFiveOneTwo}{13.7}

\newcommand{\gNineTwentySixBTextAccCorefree}{43.2}

\newcommand{\gNineTwentySixBTextMSlopeCILo}{0.38}
\newcommand{\gNineTwentySixBTextMSlopeCIHi}{0.90}

\newcommand{\gNineTwentySixBTextClassBeta}{-0.06}
\newcommand{\gNineTwentySixBTextClassSE}{0.24}

\newcommand{\gNineTwelveBVisualOddsDecade}{0.23}

\newcommand{\gNineTwelveBVisualHThreeB}{1}
\newcommand{\gNineTwelveBVisualHThreeC}{2}

\newcommand{\gNineTwelveBVisualMSlope}{1.07}

\newcommand{\gNineTwelveBVisualAccCoreFiveOneTwo}{14.9}

\newcommand{\gNineTwelveBVisualAccCorefree}{28.6}

\newcommand{\gNineTwelveBVisualMSlopeCILo}{0.51}
\newcommand{\gNineTwelveBVisualMSlopeCIHi}{2.07}

\newcommand{\gNineTwelveBVisualPerceptExactFiveOneTwo}{25.7}

\newcommand{\gNineTwelveBVisualPerceptExactfree}{33.3}

\newcommand{\gNineTwelveBTextOddsDecade}{0.22}

\newcommand{\gNineTwelveBTextHThreeB}{1}
\newcommand{\gNineTwelveBTextHThreeC}{1}

\newcommand{\gNineTwelveBTextMSlope}{1.28}

\newcommand{\gNineTwelveBTextAccCoreFiveOneTwo}{14.8}

\newcommand{\gNineTwelveBTextAccCorefree}{36.7}

\newcommand{\gNineTwelveBTextMSlopeCILo}{0.88}
\newcommand{\gNineTwelveBTextMSlopeCIHi}{1.79}

\newcommand{\gNineTwelveBTextCompFiveOneTwo}{99.2}

\newcommand{\gNineEFourBVisualHOnePHolm}{0.3716}
\newcommand{\gNineEFourBVisualHTwoPHolm}{0.5000}
\newcommand{\gNineEFourBVisualHThreeB}{0}
\newcommand{\gNineEFourBVisualHThreeC}{0}

\newcommand{\gNineEFourBVisualBetaLg}{-0.14}
\newcommand{\gNineEFourBVisualBetaLgSE}{0.44}
\newcommand{\gNineEFourBVisualBetaBud}{0.00}
\newcommand{\gNineEFourBVisualBetaBudSE}{0.77}

\newcommand{\gNineEFourBVisualPerceptExactFiveOneTwo}{2.6}
\newcommand{\gNineEFourBVisualPerceptFOneFiveOneTwo}{0.786}

\newcommand{\gNineEFourBVisualMedRtokEightOneNineTwo}{3364}

\newcommand{\gNineEFourBVisualPerceptExactEightOneNineTwo}{2.0}

\newcommand{\gNineEFourBVisualPerceptFOnefree}{0.817}

\newcommand{\gNineEFourBTextOddsDecade}{0.18}

\newcommand{\gNineEFourBTextMSlope}{0.49}

\newcommand{\gNineEFourBTextAccCoreFiveOneTwo}{14.0}

\newcommand{\gNineEFourBTextAccCorefree}{25.6}
\newcommand{\gNineEFourBTextBetaLg}{-1.72}
\newcommand{\gNineEFourBTextBetaLgSE}{0.21}

\newcommand{\gNineEFourBTextMSlopeCILo}{0.22}
\newcommand{\gNineEFourBTextMSlopeCIHi}{0.82}

\newcommand{\gNineEFourBTextMedRtokEightOneNineTwo}{3672}

%% file: gen_v8/qw_ready.tex
\includeQWtrue

%% file: gen_v8/numbers_qw.tex
\newcommand{\qWTotalCalls}{19{,}458}
\newcommand{\qWTotalMissing}{0}
\newcommand{\qWNModels}{3}
\newcommand{\qWHOneNCells}{6}
\newcommand{\qWHTwoNCells}{4}
\newcommand{\qWHThreeNCells}{6}
\newcommand{\qWHOneSupported}{6}
\newcommand{\qWHTwoSupported}{2}
\newcommand{\qWHThreeSupported}{0}
\newcommand{\qWBetaLgMin}{-2.90}
\newcommand{\qWBetaLgMax}{-1.08}
\newcommand{\qWOddsDecMin}{0.05}
\newcommand{\qWOddsDecMax}{0.34}

\newcommand{\qWThirtyTwoBThinkVisualOddsDecade}{0.26}

\newcommand{\qWThirtyTwoBThinkVisualHTwoP}{0.0616}

\newcommand{\qWThirtyTwoBThinkVisualHTwoPTex}{0.123}
\newcommand{\qWThirtyTwoBThinkVisualHThreeB}{1}
\newcommand{\qWThirtyTwoBThinkVisualHThreeC}{0}

\newcommand{\qWThirtyTwoBThinkVisualHThreePTex}{0.500}

\newcommand{\qWThirtyTwoBThinkVisualMFiftyTwoFiveOneTwo}{-0.63}
\newcommand{\qWThirtyTwoBThinkVisualAccCoreFiveOneTwo}{6.3}

\newcommand{\qWThirtyTwoBThinkVisualMedRtokFiveOneTwo}{511}

\newcommand{\qWThirtyTwoBThinkVisualAccBinFiveOneTwo}{24.5}

\newcommand{\qWThirtyTwoBThinkVisualPerceptExactFiveOneTwo}{11.7}

\newcommand{\qWThirtyTwoBThinkVisualMedRtokTwoZeroFourEight}{2047}

\newcommand{\qWThirtyTwoBThinkVisualMFiftyTwoEightOneNineTwo}{-1.16}

\newcommand{\qWThirtyTwoBThinkVisualMedRtokEightOneNineTwo}{8191}

\newcommand{\qWThirtyTwoBThinkVisualAccGridHardEightOneNineTwo}{2.2}

\newcommand{\qWThirtyTwoBThinkVisualAccCorefree}{20.0}

\newcommand{\qWThirtyTwoBThinkVisualMedRtokfree}{17935}

\newcommand{\qWThirtyTwoBThinkVisualAccBinfree}{83.0}

\newcommand{\qWThirtyTwoBThinkVisualPerceptExactfree}{19.6}

\newcommand{\qWThirtyTwoBThinkVisualBetaLg}{-1.34}
\newcommand{\qWThirtyTwoBThinkVisualBetaLgSE}{0.33}
\newcommand{\qWThirtyTwoBThinkVisualBetaBud}{0.58}
\newcommand{\qWThirtyTwoBThinkVisualBetaBudSE}{0.38}

\newcommand{\qWThirtyTwoBThinkVisualClassBeta}{-0.25}
\newcommand{\qWThirtyTwoBThinkVisualClassSE}{0.58}

\newcommand{\qWThirtyTwoBThinkVisualClassPTex}{0.662}
\newcommand{\qWThirtyTwoBThinkTextOddsDecade}{0.20}

\newcommand{\qWThirtyTwoBThinkTextMSlope}{0.23}

\newcommand{\qWThirtyTwoBThinkTextMFiftyTwoFiveOneTwo}{-0.07}
\newcommand{\qWThirtyTwoBThinkTextAccCoreFiveOneTwo}{4.8}

\newcommand{\qWThirtyTwoBThinkTextTruncFiveOneTwo}{78.8}

\newcommand{\qWThirtyTwoBThinkTextAccBinFiveOneTwo}{11.5}
\newcommand{\qWThirtyTwoBThinkTextAccGridPolyFiveOneTwo}{0.4}
\newcommand{\qWThirtyTwoBThinkTextAccGridHardFiveOneTwo}{1.6}

\newcommand{\qWThirtyTwoBThinkTextAccCoreTwoZeroFourEight}{8.1}

\newcommand{\qWThirtyTwoBThinkTextMFiftyTwoEightOneNineTwo}{0.21}
\newcommand{\qWThirtyTwoBThinkTextAccCoreEightOneNineTwo}{22.8}

\newcommand{\qWThirtyTwoBThinkTextTruncEightOneNineTwo}{22.5}

\newcommand{\qWThirtyTwoBThinkTextMFiftyTwofree}{0.81}
\newcommand{\qWThirtyTwoBThinkTextAccCorefree}{34.2}

\newcommand{\qWThirtyTwoBThinkTextTruncfree}{39.3}

\newcommand{\qWThirtyTwoBThinkTextMedRtokfree}{24376}

\newcommand{\qWThirtyTwoBThinkTextAccBinfree}{98.5}
\newcommand{\qWThirtyTwoBThinkTextAccGridPolyfree}{16.0}
\newcommand{\qWThirtyTwoBThinkTextAccGridHardfree}{15.7}

\newcommand{\qWThirtyTwoBThinkTextBetaLg}{-1.61}
\newcommand{\qWThirtyTwoBThinkTextBetaLgSE}{0.17}
\newcommand{\qWThirtyTwoBThinkTextBetaBud}{1.45}
\newcommand{\qWThirtyTwoBThinkTextBetaBudSE}{0.25}
\newcommand{\qWThirtyTwoBThinkTextMSlopeCILo}{-0.07}
\newcommand{\qWThirtyTwoBThinkTextMSlopeCIHi}{0.59}
\newcommand{\qWThirtyTwoBThinkTextClassBeta}{0.39}
\newcommand{\qWThirtyTwoBThinkTextClassSE}{0.27}

\newcommand{\qWThirtyTwoBThinkTextClassPTex}{0.148}
\newcommand{\qWEightBThinkVisualOddsDecade}{0.06}

\newcommand{\qWEightBThinkVisualHOnePTex}{0.018}

\newcommand{\qWEightBThinkVisualHTwoPTex}{0.994}

\newcommand{\qWEightBThinkVisualAccCoreFiveOneTwo}{3.9}

\newcommand{\qWEightBThinkVisualAccBinFiveOneTwo}{12.5}
\newcommand{\qWEightBThinkVisualAccGridPolyFiveOneTwo}{1.1}

\newcommand{\qWEightBThinkVisualPerceptExactFiveOneTwo}{5.2}

\newcommand{\qWEightBThinkVisualPerceptExactTwoZeroFourEight}{5.8}

\newcommand{\qWEightBThinkVisualMFiftyTwoEightOneNineTwo}{-4.30}

\newcommand{\qWEightBThinkVisualMFiftyTwofree}{-0.84}
\newcommand{\qWEightBThinkVisualAccCorefree}{12.6}

\newcommand{\qWEightBThinkVisualMedRtokfree}{16104}

\newcommand{\qWEightBThinkVisualAccBinfree}{56.0}

\newcommand{\qWEightBThinkVisualBetaLg}{-2.77}
\newcommand{\qWEightBThinkVisualBetaLgSE}{1.17}
\newcommand{\qWEightBThinkVisualBetaBud}{-1.43}
\newcommand{\qWEightBThinkVisualBetaBudSE}{0.57}

\newcommand{\qWEightBThinkVisualClassBeta}{0.84}
\newcommand{\qWEightBThinkVisualClassSE}{1.52}

\newcommand{\qWEightBThinkVisualClassPTex}{0.581}
\newcommand{\qWEightBThinkTextOddsDecade}{0.22}

\newcommand{\qWEightBThinkTextMSlope}{0.34}

\newcommand{\qWEightBThinkTextMFiftyTwoFiveOneTwo}{-0.29}
\newcommand{\qWEightBThinkTextAccCoreFiveOneTwo}{4.8}

\newcommand{\qWEightBThinkTextTruncFiveOneTwo}{46.6}

\newcommand{\qWEightBThinkTextAccBinFiveOneTwo}{15.5}
\newcommand{\qWEightBThinkTextAccGridPolyFiveOneTwo}{1.8}
\newcommand{\qWEightBThinkTextAccGridHardFiveOneTwo}{0.0}

\newcommand{\qWEightBThinkTextAccCoreTwoZeroFourEight}{7.9}

\newcommand{\qWEightBThinkTextMFiftyTwoEightOneNineTwo}{0.12}
\newcommand{\qWEightBThinkTextAccCoreEightOneNineTwo}{19.9}

\newcommand{\qWEightBThinkTextTruncEightOneNineTwo}{13.7}

\newcommand{\qWEightBThinkTextMFiftyTwofree}{0.32}
\newcommand{\qWEightBThinkTextAccCorefree}{29.3}

\newcommand{\qWEightBThinkTextTruncfree}{46.1}

\newcommand{\qWEightBThinkTextMedRtokfree}{28778}

\newcommand{\qWEightBThinkTextAccBinfree}{96.0}
\newcommand{\qWEightBThinkTextAccGridPolyfree}{11.7}
\newcommand{\qWEightBThinkTextAccGridHardfree}{11.5}

\newcommand{\qWEightBThinkTextBetaLg}{-1.50}
\newcommand{\qWEightBThinkTextBetaLgSE}{0.19}
\newcommand{\qWEightBThinkTextBetaBud}{1.78}
\newcommand{\qWEightBThinkTextBetaBudSE}{0.30}
\newcommand{\qWEightBThinkTextMSlopeCILo}{-0.04}
\newcommand{\qWEightBThinkTextMSlopeCIHi}{0.84}
\newcommand{\qWEightBThinkTextClassBeta}{0.79}
\newcommand{\qWEightBThinkTextClassSE}{0.31}

\newcommand{\qWEightBThinkTextClassPTex}{0.010}
\newcommand{\qWEightBInstrVisualOddsDecade}{0.34}

\newcommand{\qWEightBInstrVisualHOnePTex}{0.036}

\newcommand{\qWEightBInstrVisualMedRtokfree}{0}
\newcommand{\qWEightBInstrVisualMedCtokfree}{1647}

\newcommand{\qWEightBInstrVisualPerceptExactfree}{3.6}

\newcommand{\qWEightBInstrVisualBetaLg}{-1.08}
\newcommand{\qWEightBInstrVisualBetaLgSE}{0.60}

\newcommand{\qWEightBInstrTextOddsDecade}{0.05}

\newcommand{\qWEightBInstrTextTruncfree}{52.1}

\newcommand{\qWEightBInstrTextMedCtokfree}{30000}

\newcommand{\qWEightBInstrTextBetaLg}{-2.90}
\newcommand{\qWEightBInstrTextBetaLgSE}{0.56}

\newcommand{\qWEightBInstrTextClassBeta}{0.61}
\newcommand{\qWEightBInstrTextClassSE}{0.67}

\newcommand{\qWEightBInstrTextClassPTex}{0.361}
\newcommand{\qWThinkVsInstrVisualAccThink}{0.6}
\newcommand{\qWThinkVsInstrVisualAccInstr}{0.6}

\newcommand{\qWThinkVsInstrVisualB}{4}
\newcommand{\qWThinkVsInstrVisualC}{4}

\newcommand{\qWThinkVsInstrVisualPHolmTex}{1.00}
\newcommand{\qWThinkVsInstrVisualNPairs}{721}

\newcommand{\qWThinkVsInstrTextAccThink}{10.8}
\newcommand{\qWThinkVsInstrTextAccInstr}{2.8}
\newcommand{\qWThinkVsInstrTextAccDiff}{+8.0}
\newcommand{\qWThinkVsInstrTextB}{68}
\newcommand{\qWThinkVsInstrTextC}{10}

\newcommand{\qWThinkVsInstrTextPHolmTex}{$<10^{-4}$}
\newcommand{\qWThinkVsInstrTextNPairs}{721}
\newcommand{\qWThinkVsInstrTextMFiftyThink}{0.32}
\newcommand{\qWThinkVsInstrTextMFiftyInstr}{-0.15}
\newcommand{\qWThinkVsInstrTextMFiftyDiff}{+0.47}

%% file: gen_v8/numbers_trunc.tex
\providecommand{\truncValidGNine}{685}
\providecommand{\truncValidQW}{499}
\providecommand{\truncExactMin}{17}
\providecommand{\truncExactMax}{120}
\providecommand{\truncHOneN}{14}
\providecommand{\truncHOneFrozen}{13}
\providecommand{\truncHOneSurv}{12}
\providecommand{\truncHOneRatioMed}{1.00}

\providecommand{\truncHTwoN}{12}
\providecommand{\truncHTwoFrozen}{9}
\providecommand{\truncHTwoSurv}{9}
\providecommand{\truncHTwoRatioMed}{0.66}

\providecommand{\truncBTwelveTextFrozen}{1.98}
\providecommand{\truncBTwelveTextFin}{0.72}

\providecommand{\truncTviTextN}{133}
\providecommand{\truncTviTextThink}{9.8}
\providecommand{\truncTviTextInstr}{3.0}
\providecommand{\truncTviTextB}{9}
\providecommand{\truncTviTextC}{0}
\providecommand{\truncTviTextP}{0.0039}

\providecommand{\truncHOneNRows}{721}
\providecommand{\truncHTwoNRows}{2{,}163}

%% file: gen_v8/numbers_r9.tex
\newcommand{\rNinePartial}{}
\newcommand{\rNineTotalGraphs}{88}
\newcommand{\rNineLargeGrids}{48}
\newcommand{\rNineTwinGraphs}{40}
\newcommand{\rNineDomains}{4}
\newcommand{\rNineANPairs}{96}
\newcommand{\rNineBNTasks}{120}
\newcommand{\rNinePerModelRecords}{816}

\newcommand{\rNineBDirectAcc}{60.0\%}

\newcommand{\rNineBCodeAcc}{83.3\%}

\newcommand{\rNineBCodeGain}{+23.3\text{ pp}}

\newcommand{\rNineBCodeMcnemarP}{4.3\times 10^{-6}}
\newcommand{\rNineBDirectBetaLg}{-0.36}
\newcommand{\rNineBDirectBetaLgSE}{0.10}
\newcommand{\rNineBDirectOddsDec}{0.69}
\newcommand{\rNineBDirectLogitP}{3.0\times 10^{-4}}
\newcommand{\rNineBDirectBinOneAcc}{75.0\%}

\newcommand{\rNineBDirectBinTwoAcc}{64.7\%}

\newcommand{\rNineBDirectBinThreeAcc}{61.5\%}

\newcommand{\rNineBDirectBinFourAcc}{29.2\%}

\newcommand{\rNineBPcbDirAcc}{53.3\%}

\newcommand{\rNineBPcbCodeAcc}{70.0\%}

\newcommand{\rNineBBlueDirAcc}{50.0\%}

\newcommand{\rNineBBlueCodeAcc}{76.7\%}

\newcommand{\rNineBTransDirAcc}{60.0\%}

\newcommand{\rNineBTransCodeAcc}{93.3\%}

\newcommand{\rNineBCamDirAcc}{76.7\%}

\newcommand{\rNineBCamCodeAcc}{93.3\%}

\newcommand{\rNineTwinDirectSTAcc}{62.5\%}

\newcommand{\rNineTwinDirectAMAcc}{42.5\%}

\newcommand{\rNineTwinDirectPMAcc}{75.0\%}

\newcommand{\rNineTwinDirectSTvsAMP}{0.077}
\newcommand{\rNineTwinDirectPMvsAMGain}{+32.5\text{ pp}}
\newcommand{\rNineTwinDirectPMvsAMP}{0.0012}
\newcommand{\rNineAnisoFnrH}{7.3\%}

\newcommand{\rNineAnisoFnrV}{1.6\%}

\newcommand{\rNineAnisoRatio}{4.7\times}

\newcommand{\rNineAnisoWilcoxonP}{7.6\times 10^{-7}}
\newcommand{\rNineAnisoSignP}{5.0\times 10^{-5}}
\newcommand{\rNineAnisoPcbFnrH}{10.9\%}

\newcommand{\rNineAnisoPcbFnrV}{2.4\%}

\newcommand{\rNineAnisoBlueFnrH}{9.5\%}

\newcommand{\rNineAnisoBlueFnrV}{1.9\%}

\newcommand{\rNineAnisoTransFnrH}{6.5\%}

\newcommand{\rNineAnisoTransFnrV}{1.9\%}

\newcommand{\rNineAnisoCamFnrH}{2.4\%}

\newcommand{\rNineAnisoCamFnrV}{0.0\%}

\newcommand{\rNineDirTaskAcc}{0.0\%}

\newcommand{\rNineCodeTaskAcc}{14.6\%}

\newcommand{\rNineOneVGraphK}{13}

\newcommand{\rNineOneVTaskAcc}{27.1\%}

\newcommand{\rNineOneVFOne}{0.977}
\newcommand{\rNineZoomGraphK}{18}

\newcommand{\rNineZoomTaskAcc}{37.5\%}

\newcommand{\rNineRotOneEightyGraphK}{37}

\newcommand{\rNineRotOneEightyTaskAcc}{77.1\%}

\newcommand{\rNineTwoVGraphK}{35}

\newcommand{\rNineTwoVTaskAcc}{72.9\%}

\newcommand{\rNineTwoVFOne}{0.998}
\newcommand{\rNineTwoVvsOneVGain}{+45.8\text{ pp}}

\newcommand{\rNineTwoVvsOneVBC}{44/0}
\newcommand{\rNineTwoVvsOneVP}{5.7\times 10^{-14}}
\newcommand{\rNineTwoVvsZoomGain}{+35.4\text{ pp}}

\newcommand{\rNineTwoVvsZoomP}{4.1\times 10^{-6}}

\newcommand{\rNineTotalRecords}{1{,}632}

\newcommand{\rNineProBDirectAcc}{26.7\%}

\newcommand{\rNineProBCodeAcc}{70.0\%}

\newcommand{\rNineProBCodeGain}{+43.3\text{ pp}}

\newcommand{\rNineProBCodeMcnemarP}{9.1\times 10^{-13}}
\newcommand{\rNineProBDirectBetaLg}{-0.60}
\newcommand{\rNineProBDirectBetaLgSE}{0.13}

\newcommand{\rNineProBDirectLogitP}{5.1\times 10^{-6}}
\newcommand{\rNineProBDirectBinOneAcc}{50.0\%}

\newcommand{\rNineProBDirectBinTwoAcc}{35.3\%}

\newcommand{\rNineProBDirectBinThreeAcc}{15.4\%}

\newcommand{\rNineProBDirectBinFourAcc}{0.0\%}

\newcommand{\rNineProTwinDirectSTAcc}{22.5\%}

\newcommand{\rNineProTwinDirectAMAcc}{7.5\%}

\newcommand{\rNineProTwinDirectPMAcc}{50.0\%}

\newcommand{\rNineProTwinDirectSTvsAMP}{0.11}
\newcommand{\rNineProTwinDirectPMvsAMGain}{+42.5\text{ pp}}
\newcommand{\rNineProTwinDirectPMvsAMP}{1.1\times 10^{-4}}
\newcommand{\rNineProAnisoFnrH}{1.8\%}

\newcommand{\rNineProAnisoFnrV}{0.6\%}

\newcommand{\rNineProAnisoRatio}{3.0\times}

\newcommand{\rNineProAnisoWilcoxonP}{0.0083}

\newcommand{\rNineProCodeTaskAcc}{7.3\%}

\newcommand{\rNineProOneVGraphK}{18}

\newcommand{\rNineProOneVTaskAcc}{37.5\%}

\newcommand{\rNineProZoomTaskAcc}{16.7\%}

\newcommand{\rNineProRotOneEightyGraphK}{38}

\newcommand{\rNineProRotOneEightyTaskAcc}{79.2\%}

\newcommand{\rNineProTwoVGraphK}{34}

\newcommand{\rNineProTwoVTaskAcc}{70.8\%}

\newcommand{\rNineProTwoVvsOneVGain}{+33.3\text{ pp}}

\newcommand{\rNineProTwoVvsOneVBC}{32/0}
\newcommand{\rNineProTwoVvsOneVP}{2.3\times 10^{-10}}

\newcommand{\rNineCritN}{8}
\newcommand{\rNineCritPass}{4}

\newcommand{\rNineTwinDirectSTvsAMDiff}{20.0}
\newcommand{\rNineProCritPass}{3}

%% file: gen_v8/numbers_review.tex
\providecommand{\TwinMagSmallTwo}{0.63}
\providecommand{\TwinMagFive}{1.1}
\providecommand{\TwinMagSix}{2.6}
\providecommand{\BinDensMin}{0.08}
\providecommand{\BinDensMax}{0.47}
\providecommand{\GTCheckN}{344}
\providecommand{\GTCheckMismatch}{0}
\providecommand{\GTCheckNCore}{138}
\providecommand{\GTCheckNSeed}{182}
\providecommand{\GTCheckNXL}{24}
\providecommand{\FSixGemThreeSevenN}{11}
\providecommand{\FSixGemThreeSevenText}{3}
\providecommand{\FSixGemThreeSevenVisual}{8}
\providecommand{\LargeAnsN}{1{,}596}
\providecommand{\LargeAnsExact}{41}
\providecommand{\LargeAnsExactMedHigh}{35}
\providecommand{\LargeAnsPrimN}{292}
\providecommand{\LargeAnsPrimExact}{2}
\providecommand{\LargeAnsMaxTok}{63{,}842}

%% file: gen_v5/numbers_v5_b6.tex
\newcommand{\vFiveBSixTVExact}{21}
\newcommand{\vFiveBSixTVNGrids}{24}

\newcommand{\vFiveBSixTVBothMissed}{3}

\newcommand{\vFiveBSixTVVcarN}{60}

\newcommand{\vFiveBSixRotExact}{22}
\newcommand{\vFiveBSixRotNGrids}{24}

\newcommand{\vFiveBSixRotBothMissed}{1}

\newcommand{\vFiveBSixBoneExact}{17}
\newcommand{\vFiveBSixBoneNGrids}{24}

\newcommand{\vFiveBSixBoneBothMissed}{14}

\newcommand{\vFiveBSixBthreeExact}{17}
\newcommand{\vFiveBSixBthreeNGrids}{24}

\newcommand{\vFiveBSixBthreeBothMissed}{15}

\newcommand{\vFiveBSixGTVcar}{55}
\newcommand{\vFiveBSixGTVcarN}{60}
\newcommand{\vFiveBSixGTVcode}{49}
\newcommand{\vFiveBSixGTVcodeN}{60}

\newcommand{\vFiveBSixTVVsRotExactB}{2}
\newcommand{\vFiveBSixTVVsRotExactC}{3}
\newcommand{\vFiveBSixTVVsRotExactP}{$1$}

\newcommand{\vFiveBSixTVVsBthreeExactB}{5}
\newcommand{\vFiveBSixTVVsBthreeExactC}{1}

\newcommand{\vFiveBSixTVVsBthreeExactPHolm}{$0.656$}

\newcommand{\vFiveBSixRotVsBthreeExactB}{6}
\newcommand{\vFiveBSixRotVsBthreeExactC}{1}

\newcommand{\vFiveBSixRotVsBthreeExactPHolm}{$0.5$}

\newcommand{\vFiveBSixSpanUtc}{2026-10-07 21:00 UTC to 22:47 UTC}
\newcommand{\vFiveBSixSpanHours}{1.8}
\newcommand{\vFiveBSixModelVersions}{gemini-3.8-flash}
\newcommand{\vFiveBSixCallsTotal}{1335}
\newcommand{\vFiveBSixUsdTotal}{9.32}

%% file: gen_v6/numbers_v6_d8.tex
\providecommand{\vSixDEightPartial}{}
\newcommand{\vSixDEightGThreeEightLowTransCleanK}{167}

\newcommand{\vSixDEightGThreeEightLowTransCleanRepK}{170}

\newcommand{\vSixDEightGThreeEightLowTransJpegThreeZeroK}{153}

\newcommand{\vSixDEightGThreeEightLowTransJpegOneZeroK}{160}

\newcommand{\vSixDEightGThreeEightLowTransElasticSixK}{170}

\newcommand{\vSixDEightGThreeEightLowTransJpegThreeZeroVsCleanDiff}{-8.2}
\newcommand{\vSixDEightGThreeEightLowTransJpegThreeZeroVsCleanB}{15}
\newcommand{\vSixDEightGThreeEightLowTransJpegThreeZeroVsCleanC}{1}

\newcommand{\vSixDEightGThreeEightLowTransJpegThreeZeroVsCleanPHolm}{$0.00519$}
\newcommand{\vSixDEightGThreeEightLowTransJpegThreeZeroVsCleanCI}{$[-13.5, -4.0]$}

\newcommand{\vSixDEightGThreeEightLowTransJpegOneZeroVsCleanLower}{-8.7}

\newcommand{\vSixDEightGThreeEightLowTransElasticSixVsCleanLower}{-0.7}

\newcommand{\vSixDEightGThreeEightLowTransRepB}{0}
\newcommand{\vSixDEightGThreeEightLowTransRepC}{3}

\newcommand{\vSixDEightGThreeEightLowTransRepP}{$0.25$}

\newcommand{\vSixDEightGThreeEightLowVisualCleanK}{257}

\newcommand{\vSixDEightGThreeEightLowVisualCleanRepK}{244}

\newcommand{\vSixDEightGThreeEightLowVisualElasticThreeVsCleanB}{30}
\newcommand{\vSixDEightGThreeEightLowVisualElasticThreeVsCleanC}{11}

\newcommand{\vSixDEightGThreeEightLowVisualElasticThreeVsCleanPHolm}{$0.0432$}

\newcommand{\vSixDEightGThreeEightLowVisualDegMinK}{238}
\newcommand{\vSixDEightGThreeEightLowVisualDegMaxK}{254}

\newcommand{\vSixDEightGThreeEightLowVisualRepB}{28}
\newcommand{\vSixDEightGThreeEightLowVisualRepC}{15}

\newcommand{\vSixDEightGThreeEightLowVisualRepP}{$0.066$}

\newcommand{\vSixDEightGThreeEightLowCodeCleanK}{364}

\newcommand{\vSixDEightGThreeEightLowCodeJpegThreeZeroK}{309}

\newcommand{\vSixDEightGThreeEightLowCodePhotoK}{330}

\newcommand{\vSixDEightGThreeEightLowCodeJpegThreeZeroVsCleanB}{64}
\newcommand{\vSixDEightGThreeEightLowCodeJpegThreeZeroVsCleanC}{9}

\newcommand{\vSixDEightGThreeEightLowCodeJpegThreeZeroVsCleanPHolm}{$2.4{\times}10^{-10}$}

\newcommand{\vSixDEightGThreeEightLowCodePhotoVsCleanB}{44}
\newcommand{\vSixDEightGThreeEightLowCodePhotoVsCleanC}{10}

\newcommand{\vSixDEightGThreeEightLowCodePhotoVsCleanPHolm}{$3.0{\times}10^{-5}$}

\newcommand{\vSixDEightGmThreeOneMinTransCleanK}{139}

\newcommand{\vSixDEightGmThreeOneMinTransBlurThreeZeroK}{125}

\newcommand{\vSixDEightGmThreeOneMinTransBlurThreeZeroVsCleanPHolm}{$0.0434$}

\newcommand{\vSixDEightAttGThreeEightLowCleanShare}{85}

\newcommand{\vSixDEightAttGThreeEightLowJpegThreeZeroShare}{51}

\newcommand{\vSixDEightAttGThreeEightLowJpegThreeZeroShareDiff}{-35}
\newcommand{\vSixDEightAttGThreeEightLowJpegThreeZeroShareCI}{$[-47, -22]$}

\newcommand{\vSixDEightAttGThreeEightLowPhotoShare}{62}

\newcommand{\vSixDEightAttGThreeEightLowPhotoShareDiff}{-24}
\newcommand{\vSixDEightAttGThreeEightLowPhotoShareCI}{$[-35, -12]$}

\newcommand{\vSixDEightAttGThreeEightLowSpearmanRho}{+0.89}
\newcommand{\vSixDEightAttGThreeEightLowSpearmanP}{$0.000623$}

\newcommand{\vSixDEightAttProLowSpearmanRho}{+0.15}
\newcommand{\vSixDEightAttProLowSpearmanP}{$0.686$}

\newcommand{\vSixDEightAttGThreeSixLowSpearmanRho}{+0.74}
\newcommand{\vSixDEightAttGThreeSixLowSpearmanP}{$0.0146$}

\newcommand{\vSixDEightDecGThreeEightLowCleanTZeroAZero}{18}

\newcommand{\vSixDEightDecGThreeEightLowJpegThreeZeroTZeroAZero}{65}
\newcommand{\vSixDEightDecGThreeEightLowJpegThreeZeroTwrongCountAvailable}{66}
\newcommand{\vSixDEightDecGThreeEightLowJpegThreeZeroTwrongExecMatchesGhat}{62}

\newcommand{\vSixDEightDoseGThreeEightLowJpegB}{4}
\newcommand{\vSixDEightDoseGThreeEightLowJpegC}{11}
\newcommand{\vSixDEightDoseGThreeEightLowJpegP}{$0.118$}

\newcommand{\vSixDEightRetestGThreeEightLowTransExpAKa}{167}
\newcommand{\vSixDEightRetestGThreeEightLowTransExpAKb}{167}

\newcommand{\vSixDEightSpanUtc}{2026-10-08 19:01 UTC to 2026-10-09 17:51 UTC}

\newcommand{\vSixDEightSpanHours}{22.8}
\newcommand{\vSixDEightModelVersions}{gemini-3.1-pro-preview, gemini-3.6-flash, gemini-3.8-flash, gemma-4-31b-it}
\newcommand{\vSixDEightUsdTotal}{528.28}
\newcommand{\vSixDEightCallsOkTotal}{26976}

\newcommand{\vSixDEightImages}{170}
\newcommand{\vSixDEightPairs}{392}
\newcommand{\vSixDEightClusters}{138}

%% file: gen_v8/tab_attribution_main_v82.tex
\begin{tabular}{llrrrrr}
\toprule
Model & Class & $n$ & P\checkmark C\checkmark & P\checkmark C$\times$ & P$\times$ C\checkmark & P$\times$ C$\times$ \\
\midrule
3.8 Flash (low) & $\mathsf{TC}^0$ & 40 & 100 & 0 & 0 & 0 \\
 & $\mathsf{L}$ & 40 & 100 & 0 & 0 & 0 \\
 & $\mathsf{GapL}$ & 148 & 61 & 30 & 0 & 9 \\
 & $\#\mathsf{P}$ & 164 & 47 & 43 & 0 & 10 \\
\midrule
3.8 Flash (high) & $\mathsf{TC}^0$ & 19 & 100 & 0 & 0 & 0 \\
 & $\mathsf{L}$ & 20 & 100 & 0 & 0 & 0 \\
 & $\mathsf{GapL}$ & 61 & 90 & 5 & 0 & 5 \\
 & $\#\mathsf{P}$ & 56 & 82 & 11 & 0 & 7 \\
\midrule
3.6 Flash (low) & $\mathsf{TC}^0$ & 40 & 98 & 0 & 2 & 0 \\
 & $\mathsf{L}$ & 40 & 98 & 0 & 2 & 0 \\
 & $\mathsf{GapL}$ & 148 & 45 & 53 & 0 & 2 \\
 & $\#\mathsf{P}$ & 163 & 36 & 61 & 0 & 3 \\
\midrule
3.1 Pro (high) & $\mathsf{TC}^0$ & 12 & 100 & 0 & 0 & 0 \\
 & $\mathsf{L}$ & 12 & 100 & 0 & 0 & 0 \\
 & $\mathsf{GapL}$ & 42 & 69 & 17 & 0 & 14 \\
 & $\#\mathsf{P}$ & 49 & 63 & 24 & 0 & 12 \\
\midrule
Gemma 31B (off) & $\mathsf{TC}^0$ & 40 & 88 & 8 & 2 & 2 \\
 & $\mathsf{L}$ & 40 & 92 & 2 & 2 & 2 \\
 & $\mathsf{GapL}$ & 147 & 18 & 40 & 0 & 42 \\
 & $\#\mathsf{P}$ & 162 & 17 & 39 & 1 & 43 \\
\bottomrule
\end{tabular}

%% file: gen_v8/tab_twins_matrix.tex
\begin{tabular}{@{}lrrrrrrrrrr@{}}
\toprule
 & \multicolumn{2}{c}{3.8 Flash (low)} & \multicolumn{2}{c}{3.8 Flash (high)} & \multicolumn{2}{c}{3.6 Flash (high)} & \multicolumn{2}{c}{3.1 Pro (high)} & \multicolumn{2}{c}{Gemma 31B (off)} \\
\cmidrule(lr){2-3}\cmidrule(lr){4-5}\cmidrule(lr){6-7}\cmidrule(lr){8-9}\cmidrule(lr){10-11}
Twin pair (A vs.\ B) & acc & $b/c$ & acc & $b/c$ & acc & $b/c$ & acc & $b/c$ & acc & $b/c$ \\
\midrule
$\sST$ vs.\ $\sAM$ & 38/42 & 4/7 & 90/90 & 0/0 & 51/51 & 3/3 & 50/60 & 1/3 & 8/3 & 4/1 \\
$\sST$ vs.\ $\sPATH$ & 38/39 & 4/5 & 86/81 & 1/0 & 51/49 & 4/3 & 53/58 & 1/2 & 8/21 & 1/10 \\
$\sPM$ vs.\ $\sAM$ & 70/44 & 13/0$^\dagger$ & 93/86 & 1/0 & 72/48 & 6/0 & 80/53 & 4/0 & 20/0 & 10/0$^\dagger$ \\
$\chi$ vs.\ $\sCC$ & 100/100 & 0/0 & 100/100 & 0/0 & 100/100 & 0/0 & 100/100 & 0/0 & 90/95 & 1/3 \\
$\sECd$ vs.\ $\sEC$ & 97/72 & 8/0 & 100/91 & 1/0 & 100/88 & 2/0 & n/a & n/a & 34/41 & 1/3 \\
\bottomrule
\end{tabular}

%% file: gen_v8/tab_code_main_v84.tex
\begin{tabular}{@{}lrrrrrrrr@{}}
\toprule
Condition & $n$ & acc & exec & t/o & libs & alg. & DP & tokens \\
\midrule
Visual$\to$Code & 392 & 93 & 100 & 0 & 6 & 62 & 27 & 1.7k \\
Text$\to$Code & 392 & 96 & 98 & 0 & 3 & 64 & 32 & 0.9k \\
Routing prompt (control) & 392 & 91 & 100 & 0 & 0 & 72 & 41 & 1.5k \\
\bottomrule
\end{tabular}

%% file: gen_v5r/tab_dataset.tex
\paragraph{Grid drawings.} Per size: number of instances, vertices $n$, mean number of edges $m$, and for each task the median $\log_{10}$ of the ground truth with its range (min--max of $\log_{10}$). $\sPM$ is defined only when $n$ is even.
\begin{center}\small\begin{tabular}{lrrrrrrr}\toprule
Size & \# & $n$ & $\bar m$ & $\log_{10}\sST$ & $\log_{10}\sPM$ & $\log_{10}\sAM$ & $\log_{10}\sPATH$ \\ \midrule
3x3 & 6 & 9 & 10.0 & 1.4 [1.4,1.5] & -- & 1.9 [1.9,1.9] & 0.6 [0.5,0.6] \\
3x4 & 10 & 12 & 14.9 & 2.6 [2.2,2.6] & 0.8 [0.6,1.0] & 2.8 [2.7,2.8] & 1.1 [0.8,1.1] \\
4x4 & 10 & 16 & 21.1 & 3.9 [2.9,4.3] & 1.2 [0.8,1.4] & 3.8 [3.6,3.8] & 1.6 [1.1,1.8] \\
4x5 & 10 & 20 & 27.8 & 5.5 [4.2,5.8] & 1.6 [1.2,1.8] & 4.8 [4.7,4.9] & 2.3 [1.6,2.6] \\
5x5 & 10 & 25 & 36.1 & 7.2 [6.8,7.9] & -- & 6.1 [6.0,6.3] & 3.0 [2.6,3.4] \\
5x6 & 10 & 30 & 44.4 & 9.4 [7.5,10.1] & 2.5 [1.9,2.8] & 7.5 [7.1,7.6] & 3.9 [2.8,4.4] \\
6x6 & 10 & 36 & 55.2 & 11.7 [10.6,12.8] & 3.3 [2.5,3.5] & 9.1 [8.8,9.3] & 4.9 [4.2,5.6] \\
7x7 & 6 & 49 & 74.8 & 15.8 [14.8,16.7] & -- & 12.3 [12.1,12.5] & -- \\
8x8 & 6 & 64 & 100.2 & 21.4 [20.2,23.7] & 5.3 [4.9,6.2] & 16.2 [16.0,16.7] & -- \\
9x9 & 6 & 81 & 129.3 & 28.0 [27.1,31.1] & -- & 20.7 [20.4,21.4] & -- \\
10x10 & 6 & 100 & 158.5 & 34.5 [33.3,37.1] & 8.3 [7.7,9.3] & 25.5 [25.3,26.1] & -- \\
\bottomrule\end{tabular}\end{center}
\paragraph{Binary images.} Per size: instances, cells, mean number of black cells, and ranges of $C_4$ (components), $H_8$ (holes) and $\chi$.
\begin{center}\small\begin{tabular}{lrrrrrr}\toprule
Size & \# & cells & $\bar N_\bullet$ & $C_4$ & $H_8$ & $\chi$ \\ \midrule
6x6 & 10 & 36 & 13.6 & 2--3 & 1--1 & 1--2 \\
8x8 & 10 & 64 & 19.0 & 2--4 & 1--2 & 0--2 \\
10x10 & 10 & 100 & 20.4 & 2--4 & 1--2 & 1--3 \\
12x12 & 10 & 144 & 24.0 & 2--5 & 1--3 & 0--3 \\
\bottomrule\end{tabular}\end{center}
\paragraph{Eulerian drawings.} Per size bucket: pairs, vertices, arcs, number of degree-4 vertices, and ranges of the two counts.
\begin{center}\small\begin{tabular}{lrrrrrr}\toprule
Bucket & \# & $n$ & $m$ & deg-4 & $\sECd$ & $\sEC$ \\ \midrule
S & 8 & 6--7 & 9--12 & 2--5 & 1--9 & 4--120 \\
M & 8 & 8--10 & 10--15 & 2--5 & 1--3 & 4--64 \\
L & 8 & 11--13 & 15--21 & 4--8 & 1--15 & 16--1472 \\
XL & 8 & 15--17 & 22--25 & 7--10 & 1--48 & 128--11520 \\
\bottomrule\end{tabular}\end{center}
\paragraph{Magnitude matching.} Median $\log_{10}$ answers of $\sST$ and $\sAM$ on the same grids differ by 0.50 (3x3), 0.18 (3x4), 0.16 (4x4), 0.63 (4x5), 1.07 (5x5), 1.90 (5x6), 2.64 (6x6), 3.52 (7x7), 5.14 (8x8), 7.36 (9x9), 8.97 (10x10) decades.

%% file: gen_v8p/gen_v5r/prompts.tex
\paragraph{Task statements.} Each prompt contains an input description (per family), the task statement below, and condition-specific instructions.
\begin{description}\small
\item[$\chi$ (Euler characteristic).] Compute the exact 2D Euler characteristic chi = C4 - H8 of the black cells, where C4 is the number of 4-connected black components (black cells that share an edge are connected) and H8 is the number of holes, i.e. 8-connected white regions (white cells that share an edge or a corner are connected) that do not touch the outer boundary of the grid.
\item[$\sCC$ (\#Connected Components).] Count the exact number of 4-connected black components (two black cells belong to the same component if they are connected by a chain of black cells that share edges, not just corners).
\item[$\sST$ (\#Spanning Trees).] Count the exact number of spanning trees of this graph (a spanning tree is a set of n-1 edges that connects all n vertices without forming a cycle).
\item[$\sPM$ (\#Perfect Matchings).] Count the exact number of perfect matchings of this graph (sets of edges in which every vertex is covered by exactly one edge).
\item[$\sECd$ (\#Directed Eulerian Circuits).] Count the exact number of directed Eulerian circuits of this graph: closed walks that traverse every arc exactly once, in the direction of its arrow. Two circuits that differ only in the starting point (cyclic rotation) count as one.
\item[$\sAM$ (\#All Matchings).] Count the exact number of matchings of this graph, i.e., subsets of edges in which no two edges share a vertex, of any size, including the empty set.
\item[$\sPATH$ (\#Simple s-t Paths).] Count the exact number of simple paths (paths that visit no vertex more than once) from vertex 0 to vertex \{last\}.
\item[$\sEC$ (\#Undirected Eulerian Circuits).] Count the exact number of Eulerian circuits of this graph: closed walks that traverse every edge exactly once. Two circuits that differ only in the starting point or in the direction of traversal (cyclic rotation or reversal) count as one.
\end{description}
\paragraph{Hints (Hint condition).}
\begin{description}\small
\item[$\chi$.] Hint (Gray's local formula): chi = (number of black cells) - (number of horizontally or vertically adjacent black-black cell pairs) + (number of 2x2 all-black blocks). Transcribe the grid first, then count these three local quantities.
\item[$\sCC$.] Hint: transcribe the grid, then label components by flood fill or union-find over horizontally and vertically adjacent black cells.
\item[$\sST$.] Hint (Kirchhoff's matrix-tree theorem): the number of spanning trees equals the determinant of the Laplacian matrix L = D - A (D = diagonal degree matrix, A = adjacency matrix) with any one row and the same column removed. Transcribe the edges first, then compute this (n-1)x(n-1) determinant exactly.
\item[$\sPM$.] Hint (Kasteleyn / FKT theorem): because the graph is planar, the number of perfect matchings equals the Pfaffian of a suitably oriented (Kasteleyn-oriented) signed adjacency matrix, i.e. the square root of its determinant; for a bipartite planar graph this is the absolute value of the determinant of a Kasteleyn-signed biadjacency matrix. Transcribe the edges first.
\item[$\sECd$.] Hint (BEST theorem): the number of Eulerian circuits of a connected balanced digraph equals t\_w * prod\_v (outdeg(v) - 1)!, where t\_w is the number of spanning arborescences rooted at any fixed vertex w, which equals the determinant of the out-degree Laplacian (L\_ii = outdeg(i), L\_ij = -1 if i -\textgreater{} j is an arc, 0 otherwise) with row and column w removed. Every vertex here has out-degree 1 or 2, so the factorial product is 1. Transcribe the arcs first.
\item[$\sAM$.] Hint: use an exact dynamic program that processes the vertices in row-major order and whose state is the set of already-matched vertices in the current frontier (the next \{C\} vertices), or the recurrence M(G) = M(G - e) + M(G - u - v) for an edge e = (u, v). Transcribe the edges first.
\item[$\sPATH$.] Hint: enumerate simple paths from 0 to \{last\} by depth-first search, extending the current path only to unvisited vertices, and count each completed path once. Transcribe the edges first.
\item[$\sEC$.] Hint: at each degree-4 vertex choose one of the 3 ways to pair up its four incident edges (degree-2 vertices have a unique pairing); Eulerian circuits correspond exactly to the pairing systems whose induced closed trails form a single trail. Transcribe the edges first, then count such systems.
\end{description}
\paragraph{Input descriptions.}
\begin{lstlisting}[caption={grid family}]
The image shows an undirected graph with 12 vertices drawn as blue-outlined circles labeled 0 to 11, placed on a 3 x 4 lattice and labeled in row-major order (vertex r*4+c is in row r, column c). An edge exists exactly where a straight black segment joins two circles; some lattice positions have no segment.
\end{lstlisting}

\begin{lstlisting}[caption={bin family}]
The image shows a 6 x 6 grid of square cells; row indices 0..5 are printed on the left and column indices 0..5 on top. Dark cells are black (foreground, value 1) and light cells are white (background, value 0).
\end{lstlisting}

\begin{lstlisting}[caption={euler family}]
The image shows a directed graph with 7 vertices drawn as blue-outlined circles labeled 0 to 6. Each arc is a straight black segment with a red arrowhead at its head, pointing into the target vertex.
\end{lstlisting}

\paragraph{Full prompts for one instance (\texttt{grid\_3x4\_0}, $\sST$).}
\begin{lstlisting}[caption={Condition: Visual CoT}]
The image shows an undirected graph with 12 vertices drawn as blue-outlined circles labeled 0 to 11, placed on a 3 x 4 lattice and labeled in row-major order (vertex r*4+c is in row r, column c). An edge exists exactly where a straight black segment joins two circles; some lattice positions have no segment.

Task: Count the exact number of spanning trees of this graph (a spanning tree is a set of n-1 edges that connects all n vertices without forming a cycle).

First transcribe exactly what you see, then solve. End your response with exactly one line containing a JSON object of the form {"edges": [[u, v], ...], "answer": <integer>}. The answer must be an exact integer.
\end{lstlisting}

\begin{lstlisting}[caption={Condition: Text CoT}]
Vertices 0..11. Undirected edges as [u, v] pairs: [[0, 1], [0, 4], [1, 5], [2, 3], [2, 6], [3, 7], [4, 5], [4, 8], [5, 6], [5, 9], [6, 7], [7, 11], [8, 9], [9, 10], [10, 11]]

Task: Count the exact number of spanning trees of this graph (a spanning tree is a set of n-1 edges that connects all n vertices without forming a cycle).

End your response with exactly one line containing a JSON object of the form {"answer": <integer>}. The answer must be an exact integer.
\end{lstlisting}

\begin{lstlisting}[caption={Condition: Visual+Hint}]
The image shows an undirected graph with 12 vertices drawn as blue-outlined circles labeled 0 to 11, placed on a 3 x 4 lattice and labeled in row-major order (vertex r*4+c is in row r, column c). An edge exists exactly where a straight black segment joins two circles; some lattice positions have no segment.

Task: Count the exact number of spanning trees of this graph (a spanning tree is a set of n-1 edges that connects all n vertices without forming a cycle).

Hint (Kirchhoff's matrix-tree theorem): the number of spanning trees equals the determinant of the Laplacian matrix L = D - A (D = diagonal degree matrix, A = adjacency matrix) with any one row and the same column removed. Transcribe the edges first, then compute this (n-1)x(n-1) determinant exactly.

First transcribe exactly what you see, then solve. End your response with exactly one line containing a JSON object of the form {"edges": [[u, v], ...], "answer": <integer>}. The answer must be an exact integer.
\end{lstlisting}

\begin{lstlisting}[caption={Condition: Visual$\to$Code}]
The image shows an undirected graph with 12 vertices drawn as blue-outlined circles labeled 0 to 11, placed on a 3 x 4 lattice and labeled in row-major order (vertex r*4+c is in row r, column c). An edge exists exactly where a straight black segment joins two circles; some lattice positions have no segment.

Task: Count the exact number of spanning trees of this graph (a spanning tree is a set of n-1 edges that connects all n vertices without forming a cycle).

Write one complete, self-contained Python 3 program (standard library only, no external packages) that hard-codes the structure you see in the image and prints exactly one integer: the answer. Reply with only the program in a single ```python code block.
\end{lstlisting}

\begin{lstlisting}[caption={Condition: Text$\to$Code}]
Vertices 0..11. Undirected edges as [u, v] pairs: [[0, 1], [0, 4], [1, 5], [2, 3], [2, 6], [3, 7], [4, 5], [4, 8], [5, 6], [5, 9], [6, 7], [7, 11], [8, 9], [9, 10], [10, 11]]

Task: Count the exact number of spanning trees of this graph (a spanning tree is a set of n-1 edges that connects all n vertices without forming a cycle).

Write one complete, self-contained Python 3 program (standard library only, no external packages) that hard-codes this structure and prints exactly one integer: the answer. Reply with only the program in a single ```python code block.
\end{lstlisting}

\begin{lstlisting}[caption={Condition: CAR}]
The image shows an undirected graph with 12 vertices drawn as blue-outlined circles labeled 0 to 11, placed on a 3 x 4 lattice and labeled in row-major order (vertex r*4+c is in row r, column c). An edge exists exactly where a straight black segment joins two circles; some lattice positions have no segment.

Task: Count the exact number of spanning trees of this graph (a spanning tree is a set of n-1 edges that connects all n vertices without forming a cycle).

Solve the following visual counting problem in three explicit steps.
Step 1 (Perceive): transcribe the structure shown in the image exactly (the full edge list with the vertex labels shown, or the full 0/1 cell matrix).
Step 2 (Classify): identify the counting problem and its computational complexity. Decide whether it admits a known polynomial-time EXACT algorithm (for example via a determinant or Pfaffian identity, a matrix-tree-type theorem, a local counting formula, or a linear-time graph traversal), or whether it is #P-complete in general so that only exponential-time exact methods (search or dynamic programming over subsets/frontiers) are available. Name the theorem or algorithm you will use.
Step 3 (Solve by code): write one complete, self-contained Python 3 program (standard library only) that hard-codes the transcribed structure and prints exactly one integer, the answer. If a polynomial-time algebraic algorithm exists, implement it with exact integer arithmetic (fractions.Fraction or Bareiss elimination), never floating-point determinants. If the problem is #P-complete in general, implement an exact dynamic program that exploits the grid-like structure (row-by-row frontier/profile DP with memoization) instead of naive enumeration. The program must start with two comment lines: '# CLASS: <poly-algebraic | poly-other | sharpP-general>' and '# METHOD: <short name of the theorem or algorithm>'.
Reply with the program in a single ```python code block (you may put Steps 1 and 2 as comments inside the program).
\end{lstlisting}

\begin{lstlisting}[caption={Condition: Perceive+Code}]
The image shows an undirected graph with 12 vertices drawn as blue-outlined circles labeled 0 to 11, placed on a 3 x 4 lattice and labeled in row-major order (vertex r*4+c is in row r, column c). An edge exists exactly where a straight black segment joins two circles; some lattice positions have no segment.

Task: Count the exact number of spanning trees of this graph (a spanning tree is a set of n-1 edges that connects all n vertices without forming a cycle).

Solve the following visual counting problem in two explicit steps.
Step 1 (Perceive): transcribe the structure shown in the image exactly (the full edge list with the vertex labels shown, or the full 0/1 cell matrix).
Step 2 (Solve by code): write one complete, self-contained Python 3 program (standard library only) that hard-codes the transcribed structure and prints exactly one integer, the answer.
Reply with the program in a single ```python code block (you may put Step 1 as comments inside the program).
\end{lstlisting}

\begin{lstlisting}[caption={Condition: Text CAR}]
Vertices 0..11. Undirected edges as [u, v] pairs: [[0, 1], [0, 4], [1, 5], [2, 3], [2, 6], [3, 7], [4, 5], [4, 8], [5, 6], [5, 9], [6, 7], [7, 11], [8, 9], [9, 10], [10, 11]]

Task: Count the exact number of spanning trees of this graph (a spanning tree is a set of n-1 edges that connects all n vertices without forming a cycle).

Solve the following visual counting problem in three explicit steps.
Step 1 (Perceive): restate the structure given above exactly (the full edge list with the vertex labels shown, or the full 0/1 cell matrix).
Step 2 (Classify): identify the counting problem and its computational complexity. Decide whether it admits a known polynomial-time EXACT algorithm (for example via a determinant or Pfaffian identity, a matrix-tree-type theorem, a local counting formula, or a linear-time graph traversal), or whether it is #P-complete in general so that only exponential-time exact methods (search or dynamic programming over subsets/frontiers) are available. Name the theorem or algorithm you will use.
Step 3 (Solve by code): write one complete, self-contained Python 3 program (standard library only) that hard-codes the transcribed structure and prints exactly one integer, the answer. If a polynomial-time algebraic algorithm exists, implement it with exact integer arithmetic (fractions.Fraction or Bareiss elimination), never floating-point determinants. If the problem is #P-complete in general, implement an exact dynamic program that exploits the grid-like structure (row-by-row frontier/profile DP with memoization) instead of naive enumeration. The program must start with two comment lines: '# CLASS: <poly-algebraic | poly-other | sharpP-general>' and '# METHOD: <short name of the theorem or algorithm>'.
Reply with the program in a single ```python code block (you may put Steps 1 and 2 as comments inside the program).
\end{lstlisting}

\paragraph{Zoomed transcription (Zoom-$k$).} The prompt sent with every crop (here for a $10{\times}10$ grid; Zoom-1 omits the crop sentences):
\begin{lstlisting}[caption={Zoom-k: per-crop transcription prompt}]
This image is a rectangular crop of a larger drawing. The full drawing shows an undirected graph with 100 vertices drawn as blue-outlined circles labeled 0 to 99, placed on a 10 x 10 lattice and labeled in row-major order (vertex r*10+c is in row r, column c). An edge exists exactly where a straight black segment joins two horizontally or vertically neighbouring circles; some lattice positions have no segment. The crop shows only part of the lattice; circles near the crop boundary may be cut off. List ONLY edges for which BOTH endpoint circles, including their printed labels, are fully visible inside this crop. Do not guess edges whose endpoints are cut off or whose labels you cannot read.

Transcribe the edges exactly. Go row by row: for each visible vertex, state whether it is joined to its right neighbour and to its lower neighbour. End your response with exactly one line containing a JSON object of the form {"edges": [[u, v], ...]} listing every edge once.
\end{lstlisting}

\paragraph{Text input for the other families.}
\begin{lstlisting}[caption={binary image as text}]
Grid rows from top to bottom ('1' = black, '0' = white):
row 0: 000001
row 1: 000001
row 2: 111101
row 3: 110101
row 4: 011100
row 5: 001000
\end{lstlisting}

\begin{lstlisting}[caption={directed Eulerian drawing as text}]
Vertices 0..6. Arcs as [u, v] meaning u -> v: [[0, 3], [3, 2], [2, 0], [1, 4], [4, 3], [3, 1], [5, 6], [6, 2], [2, 5]]
\end{lstlisting}

%% file: gen_v5r/tab_main_visual.tex
\begin{tabular}{lrrrrrrrrrrrr}
\toprule
 & \multicolumn{1}{c}{$\chi$} & \multicolumn{1}{c}{\#CC} & \multicolumn{1}{c}{\#ST} & \multicolumn{1}{c}{\#PM} & \multicolumn{1}{c}{\#EC$_{\to}$} & \multicolumn{1}{c}{\#M} & \multicolumn{1}{c}{\#Paths} & \multicolumn{1}{c}{\#EC} & $\mathsf{TC}^0$ & $\mathsf{L}$ & $\mathsf{GapL}$ & $\#\mathsf{P}$ \\
Model (thinking) & \scriptsize $\mathsf{TC}^0$ & \scriptsize $\mathsf{L}$ & \scriptsize $\mathsf{GapL}$ & \scriptsize $\mathsf{GapL}$ & \scriptsize $\mathsf{GapL}$ & \scriptsize $\#\mathsf{P}$ & \scriptsize $\#\mathsf{P}$ & \scriptsize $\#\mathsf{P}$ & \multicolumn{4}{c}{\scriptsize class average} \\
\midrule
Gemini 3 Flash (low) & 95 & 100 & 17 & 8 & 38 & 9 & 31 & 56 & 95 & 100 & 18 & 27 \\
Gemini 3.1 Flash-Lite (low) & 80 & 90 & 0 & 0 & 50 & 0 & 14 & 6 & 80 & 90 & 11 & 7 \\
Gemini 3.5 Flash (low) & 100 & 100 & 26 & 60 & 100 & 34 & 40 & 62 & 100 & 100 & 53 & 42 \\
Gemini 3.7 Flash (low) & 100 & 100 & 41 & 57 & 100 & 46 & 47 & 67 & 100 & 100 & 59 & 49 \\
Gemini 3.6 Flash (minimal) & 100 & 98 & 9 & 28 & 62 & 6 & 27 & 28 & 100 & 98 & 27 & 19 \\
Gemini 3.6 Flash (low) & 100 & 100 & 26 & 40 & 94 & 25 & 36 & 56 & 100 & 100 & 45 & 36 \\
Gemini 3.6 Flash (high) & 100 & 100 & 51 & 72 & 100 & 51 & 49 & 88 & 100 & 100 & 68 & 57 \\
Gemini 3.8 Flash (low) & 100 & 100 & 38 & 70 & 97 & 42 & 39 & 72 & 100 & 100 & 61 & 47 \\
Gemini 3.8 Flash (medium) & 100 & 100 & 59 & 81 & 100 & 51 & 53 & 83 & 100 & 100 & 76 & 58 \\
Gemini 3.8 Flash (high) & 100 & 100 & 85 & 89 & 100 & 83 & 77 & 91 & 100 & 100 & 90 & 82 \\
Gemma 4 26B-A4B (no thinking) & 85 & 95 & 11 & 12 & 56 & 9 & 20 & 44 & 85 & 95 & 21 & 20 \\
Gemma 4 31B (no thinking) & 90 & 95 & 8 & 20 & 34 & 3 & 21 & 41 & 90 & 95 & 18 & 18 \\
Gemma 4 31B (thinking) & -- & -- & 40 & 67 & 75 & 50 & 67 & 25 & -- & -- & 58 & 45 \\
Gemini 3.1 Pro (low) & 100 & 100 & 31 & 56 & 88 & 37 & 43 & 69 & 100 & 100 & 51 & 45 \\
Gemini 3.1 Pro (high) & 100 & 100 & 50 & 80 & 100 & 57 & 55 & 100 & 100 & 100 & 69 & 63 \\
\bottomrule
\end{tabular}

%% file: gen_v5r/tab_attribution.tex
\begin{tabular}{llrrrrr}
\toprule
Model & Class & $n$ & P\checkmark C\checkmark & P\checkmark C$\times$ & P$\times$ C\checkmark & P$\times$ C$\times$ \\
\midrule
Gemini 3 Flash (low) & $\mathsf{TC}^0$ & 20 & 95 & 5 & 0 & 0 \\
 & $\mathsf{L}$ & 20 & 100 & 0 & 0 & 0 \\
 & $\mathsf{GapL}$ & 76 & 18 & 51 & 0 & 30 \\
 & $\#\mathsf{P}$ & 86 & 27 & 55 & 0 & 19 \\
\midrule
Gemini 3.1 Flash-Lite (low) & $\mathsf{TC}^0$ & 20 & 50 & 5 & 30 & 15 \\
 & $\mathsf{L}$ & 20 & 65 & 0 & 25 & 10 \\
 & $\mathsf{GapL}$ & 76 & 11 & 24 & 0 & 66 \\
 & $\#\mathsf{P}$ & 86 & 7 & 34 & 0 & 59 \\
\midrule
Gemini 3.5 Flash (low) & $\mathsf{TC}^0$ & 20 & 100 & 0 & 0 & 0 \\
 & $\mathsf{L}$ & 20 & 100 & 0 & 0 & 0 \\
 & $\mathsf{GapL}$ & 76 & 53 & 39 & 0 & 8 \\
 & $\#\mathsf{P}$ & 86 & 42 & 48 & 0 & 10 \\
\midrule
Gemini 3.7 Flash (low) & $\mathsf{TC}^0$ & 20 & 100 & 0 & 0 & 0 \\
 & $\mathsf{L}$ & 20 & 100 & 0 & 0 & 0 \\
 & $\mathsf{GapL}$ & 73 & 59 & 37 & 0 & 4 \\
 & $\#\mathsf{P}$ & 81 & 49 & 44 & 0 & 6 \\
\midrule
Gemini 3.6 Flash (minimal) & $\mathsf{TC}^0$ & 40 & 95 & 0 & 5 & 0 \\
 & $\mathsf{L}$ & 40 & 90 & 2 & 8 & 0 \\
 & $\mathsf{GapL}$ & 148 & 27 & 64 & 0 & 9 \\
 & $\#\mathsf{P}$ & 164 & 19 & 73 & 0 & 8 \\
\midrule
Gemini 3.6 Flash (low) & $\mathsf{TC}^0$ & 40 & 98 & 0 & 2 & 0 \\
 & $\mathsf{L}$ & 40 & 98 & 0 & 2 & 0 \\
 & $\mathsf{GapL}$ & 148 & 45 & 53 & 0 & 2 \\
 & $\#\mathsf{P}$ & 163 & 36 & 61 & 0 & 3 \\
\midrule
Gemini 3.6 Flash (high) & $\mathsf{TC}^0$ & 20 & 100 & 0 & 0 & 0 \\
 & $\mathsf{L}$ & 20 & 100 & 0 & 0 & 0 \\
 & $\mathsf{GapL}$ & 76 & 68 & 26 & 0 & 5 \\
 & $\#\mathsf{P}$ & 86 & 57 & 38 & 0 & 5 \\
\midrule
Gemini 3.8 Flash (low) & $\mathsf{TC}^0$ & 40 & 100 & 0 & 0 & 0 \\
 & $\mathsf{L}$ & 40 & 100 & 0 & 0 & 0 \\
 & $\mathsf{GapL}$ & 148 & 61 & 30 & 0 & 9 \\
 & $\#\mathsf{P}$ & 164 & 47 & 43 & 0 & 10 \\
\midrule
Gemini 3.8 Flash (medium) & $\mathsf{TC}^0$ & 40 & 100 & 0 & 0 & 0 \\
 & $\mathsf{L}$ & 40 & 100 & 0 & 0 & 0 \\
 & $\mathsf{GapL}$ & 142 & 76 & 15 & 0 & 8 \\
 & $\#\mathsf{P}$ & 147 & 60 & 31 & 1 & 9 \\
\midrule
Gemini 3.8 Flash (high) & $\mathsf{TC}^0$ & 19 & 100 & 0 & 0 & 0 \\
 & $\mathsf{L}$ & 20 & 100 & 0 & 0 & 0 \\
 & $\mathsf{GapL}$ & 61 & 90 & 5 & 0 & 5 \\
 & $\#\mathsf{P}$ & 56 & 82 & 11 & 0 & 7 \\
\midrule
Gemma 4 26B-A4B (no thinking) & $\mathsf{TC}^0$ & 20 & 85 & 15 & 0 & 0 \\
 & $\mathsf{L}$ & 20 & 95 & 5 & 0 & 0 \\
 & $\mathsf{GapL}$ & 58 & 24 & 5 & 2 & 69 \\
 & $\#\mathsf{P}$ & 71 & 24 & 4 & 0 & 72 \\
\midrule
Gemma 4 31B (no thinking) & $\mathsf{TC}^0$ & 40 & 88 & 8 & 2 & 2 \\
 & $\mathsf{L}$ & 40 & 92 & 2 & 2 & 2 \\
 & $\mathsf{GapL}$ & 147 & 18 & 40 & 0 & 42 \\
 & $\#\mathsf{P}$ & 162 & 17 & 39 & 1 & 43 \\
\midrule
Gemma 4 31B (thinking) & $\mathsf{GapL}$ & 12 & 58 & 42 & 0 & 0 \\
 & $\#\mathsf{P}$ & 11 & 45 & 45 & 0 & 9 \\
\midrule
Gemini 3.1 Pro (low) & $\mathsf{TC}^0$ & 20 & 100 & 0 & 0 & 0 \\
 & $\mathsf{L}$ & 20 & 100 & 0 & 0 & 0 \\
 & $\mathsf{GapL}$ & 76 & 50 & 37 & 1 & 12 \\
 & $\#\mathsf{P}$ & 86 & 45 & 43 & 0 & 12 \\
\midrule
Gemini 3.1 Pro (high) & $\mathsf{TC}^0$ & 12 & 100 & 0 & 0 & 0 \\
 & $\mathsf{L}$ & 12 & 100 & 0 & 0 & 0 \\
 & $\mathsf{GapL}$ & 42 & 69 & 17 & 0 & 14 \\
 & $\#\mathsf{P}$ & 49 & 63 & 24 & 0 & 12 \\
\bottomrule
\end{tabular}

%% file: gen_v5r/tab_budget_visual.tex
\begin{tabular}{lrrrrrrrr}
\toprule
Model (thinking) & \multicolumn{2}{c}{$\mathsf{TC}^0$} & \multicolumn{2}{c}{$\mathsf{L}$} & \multicolumn{2}{c}{$\mathsf{GapL}$} & \multicolumn{2}{c}{$\#\mathsf{P}$} \\
 & acc & tok & acc & tok & acc & tok & acc & tok \\
\midrule
Gemini 3 Flash (low) & 95 & 2.8k & 100 & 2.6k & 18 & 2.3k & 27 & 3.0k \\
Gemini 3.1 Flash-Lite (low) & 80 & 1.1k & 90 & 1.0k & 11 & 1.4k & 7 & 1.3k \\
Gemini 3.5 Flash (low) & 100 & 2.5k & 100 & 1.8k & 53 & 8.0k & 42 & 7.8k \\
Gemini 3.7 Flash (low) & 100 & 1.4k & 100 & 0.9k & 59 & 4.2k & 49 & 4.2k \\
Gemini 3.6 Flash (minimal) & 100 & 1.0k & 98 & 0.7k & 27 & 1.6k & 19 & 2.0k \\
Gemini 3.6 Flash (low) & 100 & 1.9k & 100 & 1.5k & 45 & 6.6k & 36 & 7.4k \\
Gemini 3.6 Flash (high) & 100 & 8.0k & 100 & 4.3k & 68 & 23.9k & 57 & 26.8k \\
Gemini 3.8 Flash (low) & 100 & 1.4k & 100 & 1.1k & 61 & 5.2k & 47 & 6.1k \\
Gemini 3.8 Flash (medium) & 100 & 2.2k & 100 & 1.6k & 76 & 11.9k & 58 & 14.4k \\
Gemini 3.8 Flash (high) & 100 & 8.8k & 100 & 3.4k & 90 & 25.2k & 82 & 29.8k \\
Gemma 4 26B-A4B (no thinking) & 85 & 1.4k & 95 & 1.0k & 21 & 7.5k & 20 & 5.9k \\
Gemma 4 31B (no thinking) & 90 & 0.9k & 95 & 0.8k & 18 & 2.1k & 18 & 2.0k \\
Gemma 4 31B (thinking) & -- & -- & -- & -- & 58 & 13.7k & 45 & 13.3k \\
Gemini 3.1 Pro (low) & 100 & 2.0k & 100 & 1.6k & 51 & 4.3k & 45 & 4.6k \\
Gemini 3.1 Pro (high) & 100 & 8.5k & 100 & 4.0k & 69 & 28.1k & 63 & 30.8k \\
\bottomrule
\end{tabular}

%% file: gen_v5r/tab_m50.tex
\begin{tabular}{lrrrrr}
\toprule
Configuration & $n$ & acc.\ (\%) & tokens & $m_{50}$ all & $m_{50}$ $\GapL$ / $\shP$ \\
\midrule
Gemini 3 Flash (low) & 162 & 23 & 2.6k & $10^{1.0}$ & $10^{0.0}$ / $10^{1.6}$ \\
Gemini 3.1 Flash-Lite (low) & 162 & 9 & 1.3k & $10^{0.4}$ & $10^{0.1}$ / $10^{0.5}$ \\
Gemini 3.5 Flash (low) & 162 & 47 & 7.9k & $10^{2.5}$ & $10^{2.3}$ / $10^{2.5}$ \\
Gemini 3.7 Flash (low) & 154 & 54 & 4.2k & $10^{3.2}$ & $10^{3.6}$ / $10^{3.0}$ \\
Gemini 3.6 Flash (minimal) & 312 & 23 & 1.8k & $10^{1.2}$ & $10^{1.0}$ / $10^{1.3}$ \\
Gemini 3.6 Flash (low) & 311 & 40 & 7.0k & $10^{2.2}$ & $10^{2.1}$ / $10^{2.2}$ \\
Gemini 3.6 Flash (high) & 162 & 62 & 25.4k & $10^{3.9}$ & $10^{4.7}$ / $10^{3.5}$ \\
Gemini 3.8 Flash (low) & 312 & 54 & 5.7k & $10^{3.2}$ & $10^{3.6}$ / $10^{3.0}$ \\
Gemini 3.8 Flash (medium) & 297 & 66 & 13.2k & $10^{4.5}$ & $10^{5.9}$ / $10^{3.7}$ \\
Gemini 3.8 Flash (high) & 117 & 86 & 27.4k & $10^{7.1}$ & $10^{8.4}$ / $10^{5.9}$ \\
Gemma 4 26B-A4B (no thinking) & 162 & 20 & 6.6k & $10^{0.9}$ & $10^{0.7}$ / $10^{1.2}$ \\
Gemma 4 31B (no thinking) & 312 & 18 & 2.1k & $10^{0.8}$ & $10^{0.4}$ / $10^{1.3}$ \\
Gemini 3.1 Pro (low) & 162 & 48 & 4.4k & $10^{2.6}$ & $10^{2.5}$ / $10^{2.7}$ \\
Gemini 3.1 Pro (high) & 91 & 66 & 29.5k & $10^{4.3}$ & $10^{4.5}$ / $10^{4.1}$ \\
\bottomrule
\end{tabular}

%% file: gen_v5r/tab_main_text.tex
\begin{tabular}{lrrrrrrrrrrrr}
\toprule
 & \multicolumn{1}{c}{$\chi$} & \multicolumn{1}{c}{\#CC} & \multicolumn{1}{c}{\#ST} & \multicolumn{1}{c}{\#PM} & \multicolumn{1}{c}{\#EC$_{\to}$} & \multicolumn{1}{c}{\#M} & \multicolumn{1}{c}{\#Paths} & \multicolumn{1}{c}{\#EC} & $\mathsf{TC}^0$ & $\mathsf{L}$ & $\mathsf{GapL}$ & $\#\mathsf{P}$ \\
Model (thinking) & \scriptsize $\mathsf{TC}^0$ & \scriptsize $\mathsf{L}$ & \scriptsize $\mathsf{GapL}$ & \scriptsize $\mathsf{GapL}$ & \scriptsize $\mathsf{GapL}$ & \scriptsize $\#\mathsf{P}$ & \scriptsize $\#\mathsf{P}$ & \scriptsize $\#\mathsf{P}$ & \multicolumn{4}{c}{\scriptsize class average} \\
\midrule
Gemini 3 Flash (low) & 95 & 100 & 20 & 16 & 31 & 9 & 31 & 50 & 95 & 100 & 21 & 26 \\
Gemini 3.1 Flash-Lite (low) & 75 & 85 & 3 & 12 & 38 & 0 & 20 & 0 & 75 & 85 & 13 & 8 \\
Gemini 3.5 Flash (low) & 95 & 100 & 26 & 60 & 100 & 29 & 43 & 69 & 95 & 100 & 53 & 42 \\
Gemini 3.7 Flash (low) & 100 & 100 & 38 & 62 & 94 & 40 & 38 & 56 & 100 & 100 & 58 & 42 \\
Gemini 3.6 Flash (minimal) & 88 & 100 & 21 & 36 & 66 & 17 & 27 & 31 & 88 & 100 & 36 & 24 \\
Gemini 3.6 Flash (low) & 98 & 100 & 29 & 60 & 91 & 38 & 29 & 62 & 98 & 100 & 53 & 39 \\
Gemini 3.6 Flash (high) & 100 & 100 & 43 & 80 & 100 & 54 & 49 & 69 & 100 & 100 & 67 & 55 \\
Gemini 3.8 Flash (low) & 100 & 100 & 42 & 60 & 94 & 35 & 38 & 71 & 100 & 100 & 59 & 43 \\
Gemini 3.8 Flash (medium) & 100 & 100 & 58 & 84 & 97 & 56 & 47 & 88 & 100 & 100 & 75 & 59 \\
Gemini 3.8 Flash (high) & 100 & 100 & 77 & 91 & 100 & 80 & 70 & 100 & 100 & 100 & 88 & 80 \\
Gemma 4 26B-A4B (no thinking) & 100 & 100 & 17 & 28 & 56 & 11 & 31 & 19 & 100 & 100 & 29 & 21 \\
Gemma 4 31B (no thinking) & 90 & 100 & 11 & 28 & 19 & 5 & 20 & 38 & 90 & 100 & 18 & 17 \\
Gemma 4 31B (thinking) & -- & -- & 75 & 100 & 67 & 50 & 60 & 50 & -- & -- & 80 & 54 \\
Gemini 3.1 Pro (low) & 100 & 100 & 26 & 60 & 100 & 31 & 40 & 75 & 100 & 100 & 53 & 43 \\
Gemini 3.1 Pro (high) & 100 & 100 & 63 & 86 & 100 & 70 & 53 & 88 & 100 & 100 & 78 & 66 \\
\bottomrule
\end{tabular}

%% file: gen_v5r/tab_main_hint.tex
\begin{tabular}{lrrrrrrrrrrrr}
\toprule
 & \multicolumn{1}{c}{$\chi$} & \multicolumn{1}{c}{\#CC} & \multicolumn{1}{c}{\#ST} & \multicolumn{1}{c}{\#PM} & \multicolumn{1}{c}{\#EC$_{\to}$} & \multicolumn{1}{c}{\#M} & \multicolumn{1}{c}{\#Paths} & \multicolumn{1}{c}{\#EC} & $\mathsf{TC}^0$ & $\mathsf{L}$ & $\mathsf{GapL}$ & $\#\mathsf{P}$ \\
Model (thinking) & \scriptsize $\mathsf{TC}^0$ & \scriptsize $\mathsf{L}$ & \scriptsize $\mathsf{GapL}$ & \scriptsize $\mathsf{GapL}$ & \scriptsize $\mathsf{GapL}$ & \scriptsize $\#\mathsf{P}$ & \scriptsize $\#\mathsf{P}$ & \scriptsize $\#\mathsf{P}$ & \multicolumn{4}{c}{\scriptsize class average} \\
\midrule
Gemini 3.6 Flash (high) & 100 & 100 & 46 & 72 & 100 & 37 & 46 & 94 & 100 & 100 & 66 & 51 \\
Gemini 3.8 Flash (low) & 100 & 100 & 45 & 64 & 100 & 41 & 41 & 94 & 100 & 100 & 63 & 51 \\
Gemini 3.8 Flash (high) & 100 & 100 & 75 & 94 & 100 & 77 & 73 & 93 & 100 & 100 & 87 & 79 \\
Gemma 4 31B (thinking) & -- & -- & 40 & 67 & 50 & 0 & 100 & 75 & -- & -- & 50 & 50 \\
\bottomrule
\end{tabular}

%% file: gen_v5r/tab_sizes_all.tex
\paragraph{\texttt{gemini-3-flash-preview} (low), Visual CoT.} Exact match (\%) with the number of instances in parentheses.
\begin{center}\scriptsize
\begin{tabular}{lrrrr}
\toprule
Size & \#ST ($\mathsf{GapL}$) & \#PM ($\mathsf{GapL}$) & \#M ($\#\mathsf{P}$) & \#Paths ($\#\mathsf{P}$) \\
\midrule
3x3 & 80\,\tiny(5) & --\,\tiny(0) & 60\,\tiny(5) & 100\,\tiny(5) \\
3x4 & 20\,\tiny(5) & 40\,\tiny(5) & 0\,\tiny(5) & 100\,\tiny(5) \\
4x4 & 20\,\tiny(5) & 0\,\tiny(5) & 0\,\tiny(5) & 20\,\tiny(5) \\
4x5 & 0\,\tiny(5) & 0\,\tiny(5) & 0\,\tiny(5) & 0\,\tiny(5) \\
5x5 & 0\,\tiny(5) & --\,\tiny(0) & 0\,\tiny(5) & 0\,\tiny(5) \\
5x6 & 0\,\tiny(5) & 0\,\tiny(5) & 0\,\tiny(5) & 0\,\tiny(5) \\
6x6 & 0\,\tiny(5) & 0\,\tiny(5) & 0\,\tiny(5) & 0\,\tiny(5) \\
\bottomrule
\end{tabular}

\begin{tabular}{lrr}
\toprule
Size & $\chi$ ($\mathsf{TC}^0$) & \#CC ($\mathsf{L}$) \\
\midrule
6x6 & 100\,\tiny(5) & 100\,\tiny(5) \\
8x8 & 100\,\tiny(5) & 100\,\tiny(5) \\
10x10 & 100\,\tiny(5) & 100\,\tiny(5) \\
12x12 & 80\,\tiny(5) & 100\,\tiny(5) \\
\bottomrule
\end{tabular}

\begin{tabular}{lrr}
\toprule
Size & \#EC$_{\to}$ ($\mathsf{GapL}$) & \#EC ($\#\mathsf{P}$) \\
\midrule
S & 50\,\tiny(4) & 100\,\tiny(4) \\
M & 50\,\tiny(4) & 75\,\tiny(4) \\
L & 50\,\tiny(4) & 50\,\tiny(4) \\
XL & 0\,\tiny(4) & 0\,\tiny(4) \\
\bottomrule
\end{tabular}

\end{center}
\paragraph{\texttt{gemini-3-flash-preview} (low), Text CoT.} Exact match (\%) with the number of instances in parentheses.
\begin{center}\scriptsize
\begin{tabular}{lrrrr}
\toprule
Size & \#ST ($\mathsf{GapL}$) & \#PM ($\mathsf{GapL}$) & \#M ($\#\mathsf{P}$) & \#Paths ($\#\mathsf{P}$) \\
\midrule
3x3 & 80\,\tiny(5) & --\,\tiny(0) & 60\,\tiny(5) & 100\,\tiny(5) \\
3x4 & 40\,\tiny(5) & 40\,\tiny(5) & 0\,\tiny(5) & 100\,\tiny(5) \\
4x4 & 20\,\tiny(5) & 40\,\tiny(5) & 0\,\tiny(5) & 20\,\tiny(5) \\
4x5 & 0\,\tiny(5) & 0\,\tiny(5) & 0\,\tiny(5) & 0\,\tiny(5) \\
5x5 & 0\,\tiny(5) & --\,\tiny(0) & 0\,\tiny(5) & 0\,\tiny(5) \\
5x6 & 0\,\tiny(5) & 0\,\tiny(5) & 0\,\tiny(5) & 0\,\tiny(5) \\
6x6 & 0\,\tiny(5) & 0\,\tiny(5) & 0\,\tiny(5) & 0\,\tiny(5) \\
\bottomrule
\end{tabular}

\begin{tabular}{lrr}
\toprule
Size & $\chi$ ($\mathsf{TC}^0$) & \#CC ($\mathsf{L}$) \\
\midrule
6x6 & 80\,\tiny(5) & 100\,\tiny(5) \\
8x8 & 100\,\tiny(5) & 100\,\tiny(5) \\
10x10 & 100\,\tiny(5) & 100\,\tiny(5) \\
12x12 & 100\,\tiny(5) & 100\,\tiny(5) \\
\bottomrule
\end{tabular}

\begin{tabular}{lrr}
\toprule
Size & \#EC$_{\to}$ ($\mathsf{GapL}$) & \#EC ($\#\mathsf{P}$) \\
\midrule
S & 50\,\tiny(4) & 100\,\tiny(4) \\
M & 25\,\tiny(4) & 50\,\tiny(4) \\
L & 25\,\tiny(4) & 50\,\tiny(4) \\
XL & 25\,\tiny(4) & 0\,\tiny(4) \\
\bottomrule
\end{tabular}

\end{center}
\paragraph{\texttt{gemini-3.1-flash-lite} (low), Visual CoT.} Exact match (\%) with the number of instances in parentheses.
\begin{center}\scriptsize
\begin{tabular}{lrrrr}
\toprule
Size & \#ST ($\mathsf{GapL}$) & \#PM ($\mathsf{GapL}$) & \#M ($\#\mathsf{P}$) & \#Paths ($\#\mathsf{P}$) \\
\midrule
3x3 & 0\,\tiny(5) & --\,\tiny(0) & 0\,\tiny(5) & 60\,\tiny(5) \\
3x4 & 0\,\tiny(5) & 0\,\tiny(5) & 0\,\tiny(5) & 40\,\tiny(5) \\
4x4 & 0\,\tiny(5) & 0\,\tiny(5) & 0\,\tiny(5) & 0\,\tiny(5) \\
4x5 & 0\,\tiny(5) & 0\,\tiny(5) & 0\,\tiny(5) & 0\,\tiny(5) \\
5x5 & 0\,\tiny(5) & --\,\tiny(0) & 0\,\tiny(5) & 0\,\tiny(5) \\
5x6 & 0\,\tiny(5) & 0\,\tiny(5) & 0\,\tiny(5) & 0\,\tiny(5) \\
6x6 & 0\,\tiny(5) & 0\,\tiny(5) & 0\,\tiny(5) & 0\,\tiny(5) \\
\bottomrule
\end{tabular}

\begin{tabular}{lrr}
\toprule
Size & $\chi$ ($\mathsf{TC}^0$) & \#CC ($\mathsf{L}$) \\
\midrule
6x6 & 100\,\tiny(5) & 100\,\tiny(5) \\
8x8 & 80\,\tiny(5) & 100\,\tiny(5) \\
10x10 & 80\,\tiny(5) & 100\,\tiny(5) \\
12x12 & 60\,\tiny(5) & 60\,\tiny(5) \\
\bottomrule
\end{tabular}

\begin{tabular}{lrr}
\toprule
Size & \#EC$_{\to}$ ($\mathsf{GapL}$) & \#EC ($\#\mathsf{P}$) \\
\midrule
S & 75\,\tiny(4) & 0\,\tiny(4) \\
M & 75\,\tiny(4) & 25\,\tiny(4) \\
L & 50\,\tiny(4) & 0\,\tiny(4) \\
XL & 0\,\tiny(4) & 0\,\tiny(4) \\
\bottomrule
\end{tabular}

\end{center}
\paragraph{\texttt{gemini-3.1-flash-lite} (low), Text CoT.} Exact match (\%) with the number of instances in parentheses.
\begin{center}\scriptsize
\begin{tabular}{lrrrr}
\toprule
Size & \#ST ($\mathsf{GapL}$) & \#PM ($\mathsf{GapL}$) & \#M ($\#\mathsf{P}$) & \#Paths ($\#\mathsf{P}$) \\
\midrule
3x3 & 0\,\tiny(5) & --\,\tiny(0) & 0\,\tiny(5) & 100\,\tiny(5) \\
3x4 & 20\,\tiny(5) & 40\,\tiny(5) & 0\,\tiny(5) & 20\,\tiny(5) \\
4x4 & 0\,\tiny(5) & 0\,\tiny(5) & 0\,\tiny(5) & 20\,\tiny(5) \\
4x5 & 0\,\tiny(5) & 20\,\tiny(5) & 0\,\tiny(5) & 0\,\tiny(5) \\
5x5 & 0\,\tiny(5) & --\,\tiny(0) & 0\,\tiny(5) & 0\,\tiny(5) \\
5x6 & 0\,\tiny(5) & 0\,\tiny(5) & 0\,\tiny(5) & 0\,\tiny(5) \\
6x6 & 0\,\tiny(5) & 0\,\tiny(5) & 0\,\tiny(5) & 0\,\tiny(5) \\
\bottomrule
\end{tabular}

\begin{tabular}{lrr}
\toprule
Size & $\chi$ ($\mathsf{TC}^0$) & \#CC ($\mathsf{L}$) \\
\midrule
6x6 & 80\,\tiny(5) & 100\,\tiny(5) \\
8x8 & 80\,\tiny(5) & 80\,\tiny(5) \\
10x10 & 60\,\tiny(5) & 80\,\tiny(5) \\
12x12 & 80\,\tiny(5) & 80\,\tiny(5) \\
\bottomrule
\end{tabular}

\begin{tabular}{lrr}
\toprule
Size & \#EC$_{\to}$ ($\mathsf{GapL}$) & \#EC ($\#\mathsf{P}$) \\
\midrule
S & 100\,\tiny(4) & 0\,\tiny(4) \\
M & 25\,\tiny(4) & 0\,\tiny(4) \\
L & 25\,\tiny(4) & 0\,\tiny(4) \\
XL & 0\,\tiny(4) & 0\,\tiny(4) \\
\bottomrule
\end{tabular}

\end{center}
\paragraph{\texttt{gemini-3.5-flash} (low), Visual CoT.} Exact match (\%) with the number of instances in parentheses.
\begin{center}\scriptsize
\begin{tabular}{lrrrr}
\toprule
Size & \#ST ($\mathsf{GapL}$) & \#PM ($\mathsf{GapL}$) & \#M ($\#\mathsf{P}$) & \#Paths ($\#\mathsf{P}$) \\
\midrule
3x3 & 100\,\tiny(5) & --\,\tiny(0) & 80\,\tiny(5) & 100\,\tiny(5) \\
3x4 & 60\,\tiny(5) & 100\,\tiny(5) & 80\,\tiny(5) & 100\,\tiny(5) \\
4x4 & 20\,\tiny(5) & 100\,\tiny(5) & 40\,\tiny(5) & 40\,\tiny(5) \\
4x5 & 0\,\tiny(5) & 100\,\tiny(5) & 40\,\tiny(5) & 40\,\tiny(5) \\
5x5 & 0\,\tiny(5) & --\,\tiny(0) & 0\,\tiny(5) & 0\,\tiny(5) \\
5x6 & 0\,\tiny(5) & 0\,\tiny(5) & 0\,\tiny(5) & 0\,\tiny(5) \\
6x6 & 0\,\tiny(5) & 0\,\tiny(5) & 0\,\tiny(5) & 0\,\tiny(5) \\
\bottomrule
\end{tabular}

\begin{tabular}{lrr}
\toprule
Size & $\chi$ ($\mathsf{TC}^0$) & \#CC ($\mathsf{L}$) \\
\midrule
6x6 & 100\,\tiny(5) & 100\,\tiny(5) \\
8x8 & 100\,\tiny(5) & 100\,\tiny(5) \\
10x10 & 100\,\tiny(5) & 100\,\tiny(5) \\
12x12 & 100\,\tiny(5) & 100\,\tiny(5) \\
\bottomrule
\end{tabular}

\begin{tabular}{lrr}
\toprule
Size & \#EC$_{\to}$ ($\mathsf{GapL}$) & \#EC ($\#\mathsf{P}$) \\
\midrule
S & 100\,\tiny(4) & 100\,\tiny(4) \\
M & 100\,\tiny(4) & 50\,\tiny(4) \\
L & 100\,\tiny(4) & 75\,\tiny(4) \\
XL & 100\,\tiny(4) & 25\,\tiny(4) \\
\bottomrule
\end{tabular}

\end{center}
\paragraph{\texttt{gemini-3.5-flash} (low), Text CoT.} Exact match (\%) with the number of instances in parentheses.
\begin{center}\scriptsize
\begin{tabular}{lrrrr}
\toprule
Size & \#ST ($\mathsf{GapL}$) & \#PM ($\mathsf{GapL}$) & \#M ($\#\mathsf{P}$) & \#Paths ($\#\mathsf{P}$) \\
\midrule
3x3 & 100\,\tiny(5) & --\,\tiny(0) & 100\,\tiny(5) & 100\,\tiny(5) \\
3x4 & 40\,\tiny(5) & 100\,\tiny(5) & 20\,\tiny(5) & 100\,\tiny(5) \\
4x4 & 20\,\tiny(5) & 100\,\tiny(5) & 40\,\tiny(5) & 80\,\tiny(5) \\
4x5 & 20\,\tiny(5) & 60\,\tiny(5) & 40\,\tiny(5) & 20\,\tiny(5) \\
5x5 & 0\,\tiny(5) & --\,\tiny(0) & 0\,\tiny(5) & 0\,\tiny(5) \\
5x6 & 0\,\tiny(5) & 40\,\tiny(5) & 0\,\tiny(5) & 0\,\tiny(5) \\
6x6 & 0\,\tiny(5) & 0\,\tiny(5) & 0\,\tiny(5) & 0\,\tiny(5) \\
\bottomrule
\end{tabular}

\begin{tabular}{lrr}
\toprule
Size & $\chi$ ($\mathsf{TC}^0$) & \#CC ($\mathsf{L}$) \\
\midrule
6x6 & 100\,\tiny(5) & 100\,\tiny(5) \\
8x8 & 80\,\tiny(5) & 100\,\tiny(5) \\
10x10 & 100\,\tiny(5) & 100\,\tiny(5) \\
12x12 & 100\,\tiny(5) & 100\,\tiny(5) \\
\bottomrule
\end{tabular}

\begin{tabular}{lrr}
\toprule
Size & \#EC$_{\to}$ ($\mathsf{GapL}$) & \#EC ($\#\mathsf{P}$) \\
\midrule
S & 100\,\tiny(4) & 100\,\tiny(4) \\
M & 100\,\tiny(4) & 100\,\tiny(4) \\
L & 100\,\tiny(4) & 50\,\tiny(4) \\
XL & 100\,\tiny(4) & 25\,\tiny(4) \\
\bottomrule
\end{tabular}

\end{center}
\paragraph{\texttt{gemini-3.7-flash} (low), Visual CoT.} Exact match (\%) with the number of instances in parentheses.
\begin{center}\scriptsize
\begin{tabular}{lrrrr}
\toprule
Size & \#ST ($\mathsf{GapL}$) & \#PM ($\mathsf{GapL}$) & \#M ($\#\mathsf{P}$) & \#Paths ($\#\mathsf{P}$) \\
\midrule
3x3 & 100\,\tiny(5) & --\,\tiny(0) & 100\,\tiny(5) & 100\,\tiny(5) \\
3x4 & 80\,\tiny(5) & 100\,\tiny(5) & 100\,\tiny(5) & 100\,\tiny(4) \\
4x4 & 25\,\tiny(4) & 100\,\tiny(5) & 80\,\tiny(5) & 100\,\tiny(5) \\
4x5 & 60\,\tiny(5) & 60\,\tiny(5) & 40\,\tiny(5) & 40\,\tiny(5) \\
5x5 & 20\,\tiny(5) & --\,\tiny(0) & 0\,\tiny(5) & 0\,\tiny(5) \\
5x6 & 0\,\tiny(5) & 0\,\tiny(3) & 0\,\tiny(5) & 0\,\tiny(5) \\
6x6 & 0\,\tiny(5) & 0\,\tiny(5) & 0\,\tiny(5) & 0\,\tiny(5) \\
\bottomrule
\end{tabular}

\begin{tabular}{lrr}
\toprule
Size & $\chi$ ($\mathsf{TC}^0$) & \#CC ($\mathsf{L}$) \\
\midrule
6x6 & 100\,\tiny(5) & 100\,\tiny(5) \\
8x8 & 100\,\tiny(5) & 100\,\tiny(5) \\
10x10 & 100\,\tiny(5) & 100\,\tiny(5) \\
12x12 & 100\,\tiny(5) & 100\,\tiny(5) \\
\bottomrule
\end{tabular}

\begin{tabular}{lrr}
\toprule
Size & \#EC$_{\to}$ ($\mathsf{GapL}$) & \#EC ($\#\mathsf{P}$) \\
\midrule
S & 100\,\tiny(4) & 100\,\tiny(4) \\
M & 100\,\tiny(4) & 100\,\tiny(3) \\
L & 100\,\tiny(4) & 33\,\tiny(3) \\
XL & 100\,\tiny(4) & 0\,\tiny(2) \\
\bottomrule
\end{tabular}

\end{center}
\paragraph{\texttt{gemini-3.7-flash} (low), Text CoT.} Exact match (\%) with the number of instances in parentheses.
\begin{center}\scriptsize
\begin{tabular}{lrrrr}
\toprule
Size & \#ST ($\mathsf{GapL}$) & \#PM ($\mathsf{GapL}$) & \#M ($\#\mathsf{P}$) & \#Paths ($\#\mathsf{P}$) \\
\midrule
3x3 & 100\,\tiny(5) & --\,\tiny(0) & 80\,\tiny(5) & 100\,\tiny(5) \\
3x4 & 40\,\tiny(5) & 100\,\tiny(5) & 60\,\tiny(5) & 100\,\tiny(5) \\
4x4 & 100\,\tiny(4) & 100\,\tiny(5) & 80\,\tiny(5) & 40\,\tiny(5) \\
4x5 & 40\,\tiny(5) & 60\,\tiny(5) & 60\,\tiny(5) & 25\,\tiny(4) \\
5x5 & 0\,\tiny(5) & --\,\tiny(0) & 0\,\tiny(5) & 0\,\tiny(5) \\
5x6 & 0\,\tiny(5) & 40\,\tiny(5) & 0\,\tiny(5) & 0\,\tiny(5) \\
6x6 & 0\,\tiny(5) & 0\,\tiny(4) & 0\,\tiny(5) & 0\,\tiny(5) \\
\bottomrule
\end{tabular}

\begin{tabular}{lrr}
\toprule
Size & $\chi$ ($\mathsf{TC}^0$) & \#CC ($\mathsf{L}$) \\
\midrule
6x6 & 100\,\tiny(5) & 100\,\tiny(5) \\
8x8 & 100\,\tiny(5) & 100\,\tiny(5) \\
10x10 & 100\,\tiny(5) & 100\,\tiny(5) \\
12x12 & 100\,\tiny(5) & 100\,\tiny(5) \\
\bottomrule
\end{tabular}

\begin{tabular}{lrr}
\toprule
Size & \#EC$_{\to}$ ($\mathsf{GapL}$) & \#EC ($\#\mathsf{P}$) \\
\midrule
S & 100\,\tiny(4) & 75\,\tiny(4) \\
M & 100\,\tiny(4) & 50\,\tiny(4) \\
L & 100\,\tiny(4) & 75\,\tiny(4) \\
XL & 75\,\tiny(4) & 25\,\tiny(4) \\
\bottomrule
\end{tabular}

\end{center}
\paragraph{\texttt{gemini-3.6-flash} (minimal), Visual CoT.} Exact match (\%) with the number of instances in parentheses.
\begin{center}\scriptsize
\begin{tabular}{lrrrr}
\toprule
Size & \#ST ($\mathsf{GapL}$) & \#PM ($\mathsf{GapL}$) & \#M ($\#\mathsf{P}$) & \#Paths ($\#\mathsf{P}$) \\
\midrule
3x3 & 67\,\tiny(6) & --\,\tiny(0) & 50\,\tiny(6) & 100\,\tiny(6) \\
3x4 & 10\,\tiny(10) & 90\,\tiny(10) & 10\,\tiny(10) & 90\,\tiny(10) \\
4x4 & 10\,\tiny(10) & 30\,\tiny(10) & 0\,\tiny(10) & 30\,\tiny(10) \\
4x5 & 0\,\tiny(10) & 20\,\tiny(10) & 0\,\tiny(10) & 0\,\tiny(10) \\
5x5 & 0\,\tiny(10) & --\,\tiny(0) & 0\,\tiny(10) & 0\,\tiny(10) \\
5x6 & 0\,\tiny(10) & 0\,\tiny(10) & 0\,\tiny(10) & 0\,\tiny(10) \\
6x6 & 0\,\tiny(10) & 0\,\tiny(10) & 0\,\tiny(10) & 0\,\tiny(10) \\
\bottomrule
\end{tabular}

\begin{tabular}{lrr}
\toprule
Size & $\chi$ ($\mathsf{TC}^0$) & \#CC ($\mathsf{L}$) \\
\midrule
6x6 & 100\,\tiny(10) & 100\,\tiny(10) \\
8x8 & 100\,\tiny(10) & 100\,\tiny(10) \\
10x10 & 100\,\tiny(10) & 100\,\tiny(10) \\
12x12 & 100\,\tiny(10) & 90\,\tiny(10) \\
\bottomrule
\end{tabular}

\begin{tabular}{lrr}
\toprule
Size & \#EC$_{\to}$ ($\mathsf{GapL}$) & \#EC ($\#\mathsf{P}$) \\
\midrule
S & 62\,\tiny(8) & 50\,\tiny(8) \\
M & 88\,\tiny(8) & 38\,\tiny(8) \\
L & 50\,\tiny(8) & 12\,\tiny(8) \\
XL & 50\,\tiny(8) & 12\,\tiny(8) \\
\bottomrule
\end{tabular}

\end{center}
\paragraph{\texttt{gemini-3.6-flash} (minimal), Text CoT.} Exact match (\%) with the number of instances in parentheses.
\begin{center}\scriptsize
\begin{tabular}{lrrrr}
\toprule
Size & \#ST ($\mathsf{GapL}$) & \#PM ($\mathsf{GapL}$) & \#M ($\#\mathsf{P}$) & \#Paths ($\#\mathsf{P}$) \\
\midrule
3x3 & 67\,\tiny(6) & --\,\tiny(0) & 67\,\tiny(6) & 100\,\tiny(6) \\
3x4 & 80\,\tiny(10) & 100\,\tiny(10) & 70\,\tiny(10) & 100\,\tiny(10) \\
4x4 & 20\,\tiny(10) & 60\,\tiny(10) & 0\,\tiny(10) & 20\,\tiny(10) \\
4x5 & 0\,\tiny(10) & 20\,\tiny(10) & 0\,\tiny(10) & 0\,\tiny(10) \\
5x5 & 0\,\tiny(10) & --\,\tiny(0) & 0\,\tiny(10) & 0\,\tiny(10) \\
5x6 & 0\,\tiny(10) & 0\,\tiny(10) & 0\,\tiny(10) & 0\,\tiny(10) \\
6x6 & 0\,\tiny(10) & 0\,\tiny(10) & 0\,\tiny(10) & 0\,\tiny(10) \\
\bottomrule
\end{tabular}

\begin{tabular}{lrr}
\toprule
Size & $\chi$ ($\mathsf{TC}^0$) & \#CC ($\mathsf{L}$) \\
\midrule
6x6 & 100\,\tiny(10) & 100\,\tiny(10) \\
8x8 & 90\,\tiny(10) & 100\,\tiny(10) \\
10x10 & 80\,\tiny(10) & 100\,\tiny(10) \\
12x12 & 80\,\tiny(10) & 100\,\tiny(10) \\
\bottomrule
\end{tabular}

\begin{tabular}{lrr}
\toprule
Size & \#EC$_{\to}$ ($\mathsf{GapL}$) & \#EC ($\#\mathsf{P}$) \\
\midrule
S & 88\,\tiny(8) & 38\,\tiny(8) \\
M & 88\,\tiny(8) & 50\,\tiny(8) \\
L & 62\,\tiny(8) & 25\,\tiny(8) \\
XL & 25\,\tiny(8) & 12\,\tiny(8) \\
\bottomrule
\end{tabular}

\end{center}
\paragraph{\texttt{gemini-3.6-flash} (low), Visual CoT.} Exact match (\%) with the number of instances in parentheses.
\begin{center}\scriptsize
\begin{tabular}{lrrrr}
\toprule
Size & \#ST ($\mathsf{GapL}$) & \#PM ($\mathsf{GapL}$) & \#M ($\#\mathsf{P}$) & \#Paths ($\#\mathsf{P}$) \\
\midrule
3x3 & 83\,\tiny(6) & --\,\tiny(0) & 83\,\tiny(6) & 100\,\tiny(6) \\
3x4 & 80\,\tiny(10) & 90\,\tiny(10) & 30\,\tiny(10) & 90\,\tiny(10) \\
4x4 & 20\,\tiny(10) & 100\,\tiny(10) & 50\,\tiny(10) & 60\,\tiny(10) \\
4x5 & 20\,\tiny(10) & 10\,\tiny(10) & 22\,\tiny(9) & 30\,\tiny(10) \\
5x5 & 0\,\tiny(10) & --\,\tiny(0) & 10\,\tiny(10) & 0\,\tiny(10) \\
5x6 & 0\,\tiny(10) & 0\,\tiny(10) & 0\,\tiny(10) & 0\,\tiny(10) \\
6x6 & 0\,\tiny(10) & 0\,\tiny(10) & 0\,\tiny(10) & 0\,\tiny(10) \\
\bottomrule
\end{tabular}

\begin{tabular}{lrr}
\toprule
Size & $\chi$ ($\mathsf{TC}^0$) & \#CC ($\mathsf{L}$) \\
\midrule
6x6 & 100\,\tiny(10) & 100\,\tiny(10) \\
8x8 & 100\,\tiny(10) & 100\,\tiny(10) \\
10x10 & 100\,\tiny(10) & 100\,\tiny(10) \\
12x12 & 100\,\tiny(10) & 100\,\tiny(10) \\
\bottomrule
\end{tabular}

\begin{tabular}{lrr}
\toprule
Size & \#EC$_{\to}$ ($\mathsf{GapL}$) & \#EC ($\#\mathsf{P}$) \\
\midrule
S & 100\,\tiny(8) & 75\,\tiny(8) \\
M & 100\,\tiny(8) & 100\,\tiny(8) \\
L & 100\,\tiny(8) & 38\,\tiny(8) \\
XL & 75\,\tiny(8) & 12\,\tiny(8) \\
\bottomrule
\end{tabular}

\end{center}
\paragraph{\texttt{gemini-3.6-flash} (low), Text CoT.} Exact match (\%) with the number of instances in parentheses.
\begin{center}\scriptsize
\begin{tabular}{lrrrr}
\toprule
Size & \#ST ($\mathsf{GapL}$) & \#PM ($\mathsf{GapL}$) & \#M ($\#\mathsf{P}$) & \#Paths ($\#\mathsf{P}$) \\
\midrule
3x3 & 100\,\tiny(6) & --\,\tiny(0) & 83\,\tiny(6) & 100\,\tiny(6) \\
3x4 & 70\,\tiny(10) & 90\,\tiny(10) & 90\,\tiny(10) & 80\,\tiny(10) \\
4x4 & 40\,\tiny(10) & 100\,\tiny(10) & 60\,\tiny(10) & 30\,\tiny(10) \\
4x5 & 10\,\tiny(10) & 60\,\tiny(10) & 40\,\tiny(10) & 20\,\tiny(10) \\
5x5 & 10\,\tiny(10) & --\,\tiny(0) & 10\,\tiny(10) & 0\,\tiny(10) \\
5x6 & 0\,\tiny(10) & 20\,\tiny(10) & 0\,\tiny(10) & 0\,\tiny(10) \\
6x6 & 0\,\tiny(10) & 30\,\tiny(10) & 0\,\tiny(10) & 0\,\tiny(10) \\
\bottomrule
\end{tabular}

\begin{tabular}{lrr}
\toprule
Size & $\chi$ ($\mathsf{TC}^0$) & \#CC ($\mathsf{L}$) \\
\midrule
6x6 & 100\,\tiny(10) & 100\,\tiny(10) \\
8x8 & 100\,\tiny(10) & 100\,\tiny(10) \\
10x10 & 100\,\tiny(10) & 100\,\tiny(10) \\
12x12 & 90\,\tiny(10) & 100\,\tiny(10) \\
\bottomrule
\end{tabular}

\begin{tabular}{lrr}
\toprule
Size & \#EC$_{\to}$ ($\mathsf{GapL}$) & \#EC ($\#\mathsf{P}$) \\
\midrule
S & 75\,\tiny(8) & 62\,\tiny(8) \\
M & 100\,\tiny(8) & 100\,\tiny(8) \\
L & 88\,\tiny(8) & 50\,\tiny(8) \\
XL & 100\,\tiny(8) & 38\,\tiny(8) \\
\bottomrule
\end{tabular}

\end{center}
\paragraph{\texttt{gemini-3.6-flash} (high), Visual CoT.} Exact match (\%) with the number of instances in parentheses.
\begin{center}\scriptsize
\begin{tabular}{lrrrr}
\toprule
Size & \#ST ($\mathsf{GapL}$) & \#PM ($\mathsf{GapL}$) & \#M ($\#\mathsf{P}$) & \#Paths ($\#\mathsf{P}$) \\
\midrule
3x3 & 100\,\tiny(5) & --\,\tiny(0) & 100\,\tiny(5) & 100\,\tiny(5) \\
3x4 & 100\,\tiny(5) & 100\,\tiny(5) & 100\,\tiny(5) & 100\,\tiny(5) \\
4x4 & 80\,\tiny(5) & 100\,\tiny(5) & 100\,\tiny(5) & 100\,\tiny(5) \\
4x5 & 40\,\tiny(5) & 100\,\tiny(5) & 40\,\tiny(5) & 40\,\tiny(5) \\
5x5 & 20\,\tiny(5) & --\,\tiny(0) & 20\,\tiny(5) & 0\,\tiny(5) \\
5x6 & 20\,\tiny(5) & 60\,\tiny(5) & 0\,\tiny(5) & 0\,\tiny(5) \\
6x6 & 0\,\tiny(5) & 0\,\tiny(5) & 0\,\tiny(5) & 0\,\tiny(5) \\
\bottomrule
\end{tabular}

\begin{tabular}{lrr}
\toprule
Size & $\chi$ ($\mathsf{TC}^0$) & \#CC ($\mathsf{L}$) \\
\midrule
6x6 & 100\,\tiny(5) & 100\,\tiny(5) \\
8x8 & 100\,\tiny(5) & 100\,\tiny(5) \\
10x10 & 100\,\tiny(5) & 100\,\tiny(5) \\
12x12 & 100\,\tiny(5) & 100\,\tiny(5) \\
\bottomrule
\end{tabular}

\begin{tabular}{lrr}
\toprule
Size & \#EC$_{\to}$ ($\mathsf{GapL}$) & \#EC ($\#\mathsf{P}$) \\
\midrule
S & 100\,\tiny(4) & 100\,\tiny(4) \\
M & 100\,\tiny(4) & 100\,\tiny(4) \\
L & 100\,\tiny(4) & 100\,\tiny(4) \\
XL & 100\,\tiny(4) & 50\,\tiny(4) \\
\bottomrule
\end{tabular}

\end{center}
\paragraph{\texttt{gemini-3.6-flash} (high), Text CoT.} Exact match (\%) with the number of instances in parentheses.
\begin{center}\scriptsize
\begin{tabular}{lrrrr}
\toprule
Size & \#ST ($\mathsf{GapL}$) & \#PM ($\mathsf{GapL}$) & \#M ($\#\mathsf{P}$) & \#Paths ($\#\mathsf{P}$) \\
\midrule
3x3 & 100\,\tiny(5) & --\,\tiny(0) & 100\,\tiny(5) & 100\,\tiny(5) \\
3x4 & 100\,\tiny(5) & 100\,\tiny(5) & 100\,\tiny(5) & 100\,\tiny(5) \\
4x4 & 80\,\tiny(5) & 100\,\tiny(5) & 80\,\tiny(5) & 100\,\tiny(5) \\
4x5 & 20\,\tiny(5) & 100\,\tiny(5) & 60\,\tiny(5) & 40\,\tiny(5) \\
5x5 & 0\,\tiny(5) & --\,\tiny(0) & 20\,\tiny(5) & 0\,\tiny(5) \\
5x6 & 0\,\tiny(5) & 60\,\tiny(5) & 20\,\tiny(5) & 0\,\tiny(5) \\
6x6 & 0\,\tiny(5) & 40\,\tiny(5) & 0\,\tiny(5) & 0\,\tiny(5) \\
\bottomrule
\end{tabular}

\begin{tabular}{lrr}
\toprule
Size & $\chi$ ($\mathsf{TC}^0$) & \#CC ($\mathsf{L}$) \\
\midrule
6x6 & 100\,\tiny(5) & 100\,\tiny(5) \\
8x8 & 100\,\tiny(4) & 100\,\tiny(5) \\
10x10 & 100\,\tiny(5) & 100\,\tiny(5) \\
12x12 & 100\,\tiny(5) & 100\,\tiny(5) \\
\bottomrule
\end{tabular}

\begin{tabular}{lrr}
\toprule
Size & \#EC$_{\to}$ ($\mathsf{GapL}$) & \#EC ($\#\mathsf{P}$) \\
\midrule
S & 100\,\tiny(4) & 100\,\tiny(4) \\
M & 100\,\tiny(4) & 100\,\tiny(4) \\
L & 100\,\tiny(4) & 75\,\tiny(4) \\
XL & 100\,\tiny(4) & 0\,\tiny(4) \\
\bottomrule
\end{tabular}

\end{center}
\paragraph{\texttt{gemini-3.8-flash} (low), Visual CoT.} Exact match (\%) with the number of instances in parentheses.
\begin{center}\scriptsize
\begin{tabular}{lrrrr}
\toprule
Size & \#ST ($\mathsf{GapL}$) & \#PM ($\mathsf{GapL}$) & \#M ($\#\mathsf{P}$) & \#Paths ($\#\mathsf{P}$) \\
\midrule
3x3 & 100\,\tiny(6) & --\,\tiny(0) & 100\,\tiny(6) & 100\,\tiny(6) \\
3x4 & 90\,\tiny(10) & 100\,\tiny(10) & 100\,\tiny(10) & 100\,\tiny(10) \\
4x4 & 60\,\tiny(10) & 100\,\tiny(10) & 60\,\tiny(10) & 80\,\tiny(10) \\
4x5 & 40\,\tiny(10) & 100\,\tiny(10) & 60\,\tiny(10) & 20\,\tiny(10) \\
5x5 & 0\,\tiny(10) & --\,\tiny(0) & 0\,\tiny(10) & 0\,\tiny(10) \\
5x6 & 0\,\tiny(10) & 40\,\tiny(10) & 0\,\tiny(10) & 0\,\tiny(10) \\
6x6 & 0\,\tiny(10) & 10\,\tiny(10) & 0\,\tiny(10) & 0\,\tiny(10) \\
7x7 & 0\,\tiny(6) & --\,\tiny(0) & 0\,\tiny(6) & --\,\tiny(0) \\
8x8 & 0\,\tiny(6) & 0\,\tiny(6) & 0\,\tiny(6) & --\,\tiny(0) \\
9x9 & 0\,\tiny(6) & --\,\tiny(0) & 0\,\tiny(6) & --\,\tiny(0) \\
10x10 & 0\,\tiny(6) & 0\,\tiny(6) & 0\,\tiny(6) & --\,\tiny(0) \\
\bottomrule
\end{tabular}

\begin{tabular}{lrr}
\toprule
Size & $\chi$ ($\mathsf{TC}^0$) & \#CC ($\mathsf{L}$) \\
\midrule
6x6 & 100\,\tiny(10) & 100\,\tiny(10) \\
8x8 & 100\,\tiny(10) & 100\,\tiny(10) \\
10x10 & 100\,\tiny(10) & 100\,\tiny(10) \\
12x12 & 100\,\tiny(10) & 100\,\tiny(10) \\
\bottomrule
\end{tabular}

\begin{tabular}{lrr}
\toprule
Size & \#EC$_{\to}$ ($\mathsf{GapL}$) & \#EC ($\#\mathsf{P}$) \\
\midrule
S & 100\,\tiny(8) & 88\,\tiny(8) \\
M & 100\,\tiny(8) & 75\,\tiny(8) \\
L & 100\,\tiny(8) & 75\,\tiny(8) \\
XL & 88\,\tiny(8) & 50\,\tiny(8) \\
\bottomrule
\end{tabular}

\end{center}
\paragraph{\texttt{gemini-3.8-flash} (low), Text CoT.} Exact match (\%) with the number of instances in parentheses.
\begin{center}\scriptsize
\begin{tabular}{lrrrr}
\toprule
Size & \#ST ($\mathsf{GapL}$) & \#PM ($\mathsf{GapL}$) & \#M ($\#\mathsf{P}$) & \#Paths ($\#\mathsf{P}$) \\
\midrule
3x3 & 100\,\tiny(6) & --\,\tiny(0) & 100\,\tiny(6) & 100\,\tiny(6) \\
3x4 & 100\,\tiny(10) & 100\,\tiny(10) & 80\,\tiny(10) & 100\,\tiny(10) \\
4x4 & 70\,\tiny(10) & 100\,\tiny(10) & 60\,\tiny(10) & 50\,\tiny(10) \\
4x5 & 50\,\tiny(10) & 90\,\tiny(10) & 30\,\tiny(10) & 30\,\tiny(10) \\
5x5 & 0\,\tiny(10) & --\,\tiny(0) & 0\,\tiny(10) & 10\,\tiny(10) \\
5x6 & 0\,\tiny(10) & 10\,\tiny(10) & 0\,\tiny(10) & 0\,\tiny(10) \\
6x6 & 0\,\tiny(10) & 0\,\tiny(10) & 0\,\tiny(10) & 0\,\tiny(10) \\
7x7 & 0\,\tiny(6) & --\,\tiny(0) & 0\,\tiny(6) & --\,\tiny(0) \\
8x8 & 0\,\tiny(6) & 0\,\tiny(6) & 0\,\tiny(6) & --\,\tiny(0) \\
9x9 & 0\,\tiny(6) & --\,\tiny(0) & 0\,\tiny(6) & --\,\tiny(0) \\
10x10 & 0\,\tiny(6) & 0\,\tiny(6) & 0\,\tiny(6) & --\,\tiny(0) \\
\bottomrule
\end{tabular}

\begin{tabular}{lrr}
\toprule
Size & $\chi$ ($\mathsf{TC}^0$) & \#CC ($\mathsf{L}$) \\
\midrule
6x6 & 100\,\tiny(10) & 100\,\tiny(10) \\
8x8 & 100\,\tiny(10) & 100\,\tiny(10) \\
10x10 & 100\,\tiny(10) & 100\,\tiny(10) \\
12x12 & 100\,\tiny(10) & 100\,\tiny(10) \\
\bottomrule
\end{tabular}

\begin{tabular}{lrr}
\toprule
Size & \#EC$_{\to}$ ($\mathsf{GapL}$) & \#EC ($\#\mathsf{P}$) \\
\midrule
S & 100\,\tiny(8) & 88\,\tiny(8) \\
M & 100\,\tiny(8) & 100\,\tiny(8) \\
L & 88\,\tiny(8) & 62\,\tiny(8) \\
XL & 88\,\tiny(8) & 29\,\tiny(7) \\
\bottomrule
\end{tabular}

\end{center}
\paragraph{\texttt{gemini-3.8-flash} (medium), Visual CoT.} Exact match (\%) with the number of instances in parentheses.
\begin{center}\scriptsize
\begin{tabular}{lrrrr}
\toprule
Size & \#ST ($\mathsf{GapL}$) & \#PM ($\mathsf{GapL}$) & \#M ($\#\mathsf{P}$) & \#Paths ($\#\mathsf{P}$) \\
\midrule
3x3 & 100\,\tiny(6) & --\,\tiny(0) & 100\,\tiny(6) & 100\,\tiny(6) \\
3x4 & 100\,\tiny(10) & 100\,\tiny(10) & 100\,\tiny(10) & 100\,\tiny(10) \\
4x4 & 100\,\tiny(10) & 100\,\tiny(10) & 89\,\tiny(9) & 100\,\tiny(10) \\
4x5 & 88\,\tiny(8) & 100\,\tiny(10) & 78\,\tiny(9) & 75\,\tiny(8) \\
5x5 & 30\,\tiny(10) & --\,\tiny(0) & 11\,\tiny(9) & 11\,\tiny(9) \\
5x6 & 11\,\tiny(9) & 67\,\tiny(9) & 0\,\tiny(10) & 0\,\tiny(10) \\
6x6 & 0\,\tiny(10) & 33\,\tiny(9) & 0\,\tiny(10) & 0\,\tiny(9) \\
\bottomrule
\end{tabular}

\begin{tabular}{lrr}
\toprule
Size & $\chi$ ($\mathsf{TC}^0$) & \#CC ($\mathsf{L}$) \\
\midrule
6x6 & 100\,\tiny(10) & 100\,\tiny(10) \\
8x8 & 100\,\tiny(10) & 100\,\tiny(10) \\
10x10 & 100\,\tiny(10) & 100\,\tiny(10) \\
12x12 & 100\,\tiny(10) & 100\,\tiny(10) \\
\bottomrule
\end{tabular}

\begin{tabular}{lrr}
\toprule
Size & \#EC$_{\to}$ ($\mathsf{GapL}$) & \#EC ($\#\mathsf{P}$) \\
\midrule
S & 100\,\tiny(8) & 88\,\tiny(8) \\
M & 100\,\tiny(8) & 100\,\tiny(7) \\
L & 100\,\tiny(8) & 88\,\tiny(8) \\
XL & 100\,\tiny(8) & 50\,\tiny(6) \\
\bottomrule
\end{tabular}

\end{center}
\paragraph{\texttt{gemini-3.8-flash} (medium), Text CoT.} Exact match (\%) with the number of instances in parentheses.
\begin{center}\scriptsize
\begin{tabular}{lrrrr}
\toprule
Size & \#ST ($\mathsf{GapL}$) & \#PM ($\mathsf{GapL}$) & \#M ($\#\mathsf{P}$) & \#Paths ($\#\mathsf{P}$) \\
\midrule
3x3 & 100\,\tiny(6) & --\,\tiny(0) & 100\,\tiny(6) & 100\,\tiny(6) \\
3x4 & 100\,\tiny(10) & 100\,\tiny(10) & 100\,\tiny(10) & 100\,\tiny(10) \\
4x4 & 100\,\tiny(9) & 100\,\tiny(10) & 100\,\tiny(10) & 100\,\tiny(9) \\
4x5 & 70\,\tiny(10) & 100\,\tiny(10) & 90\,\tiny(10) & 50\,\tiny(8) \\
5x5 & 44\,\tiny(9) & --\,\tiny(0) & 10\,\tiny(10) & 0\,\tiny(9) \\
5x6 & 10\,\tiny(10) & 80\,\tiny(10) & 10\,\tiny(10) & 0\,\tiny(10) \\
6x6 & 0\,\tiny(10) & 33\,\tiny(9) & 0\,\tiny(10) & 0\,\tiny(10) \\
\bottomrule
\end{tabular}

\begin{tabular}{lrr}
\toprule
Size & $\chi$ ($\mathsf{TC}^0$) & \#CC ($\mathsf{L}$) \\
\midrule
6x6 & 100\,\tiny(10) & 100\,\tiny(10) \\
8x8 & 100\,\tiny(10) & 100\,\tiny(10) \\
10x10 & 100\,\tiny(10) & 100\,\tiny(10) \\
12x12 & 100\,\tiny(10) & 100\,\tiny(10) \\
\bottomrule
\end{tabular}

\begin{tabular}{lrr}
\toprule
Size & \#EC$_{\to}$ ($\mathsf{GapL}$) & \#EC ($\#\mathsf{P}$) \\
\midrule
S & 100\,\tiny(8) & 100\,\tiny(8) \\
M & 100\,\tiny(8) & 100\,\tiny(8) \\
L & 100\,\tiny(8) & 100\,\tiny(8) \\
XL & 88\,\tiny(8) & 50\,\tiny(8) \\
\bottomrule
\end{tabular}

\end{center}
\paragraph{\texttt{gemini-3.8-flash} (high), Visual CoT.} Exact match (\%) with the number of instances in parentheses.
\begin{center}\scriptsize
\begin{tabular}{lrrrr}
\toprule
Size & \#ST ($\mathsf{GapL}$) & \#PM ($\mathsf{GapL}$) & \#M ($\#\mathsf{P}$) & \#Paths ($\#\mathsf{P}$) \\
\midrule
3x3 & 100\,\tiny(5) & --\,\tiny(0) & 100\,\tiny(5) & 100\,\tiny(5) \\
3x4 & 100\,\tiny(5) & 100\,\tiny(5) & 100\,\tiny(5) & 100\,\tiny(5) \\
4x4 & 100\,\tiny(5) & 100\,\tiny(5) & 100\,\tiny(3) & 100\,\tiny(4) \\
4x5 & 100\,\tiny(4) & 100\,\tiny(5) & 100\,\tiny(3) & 100\,\tiny(3) \\
5x5 & 67\,\tiny(3) & --\,\tiny(0) & 100\,\tiny(2) & 0\,\tiny(2) \\
5x6 & 67\,\tiny(3) & 100\,\tiny(2) & 50\,\tiny(2) & 0\,\tiny(1) \\
6x6 & 0\,\tiny(2) & 0\,\tiny(2) & 0\,\tiny(3) & 0\,\tiny(2) \\
7x7 & 0\,\tiny(3) & --\,\tiny(0) & 0\,\tiny(5) & --\,\tiny(0) \\
8x8 & 0\,\tiny(4) & 0\,\tiny(2) & 0\,\tiny(4) & --\,\tiny(0) \\
9x9 & 0\,\tiny(4) & --\,\tiny(0) & 0\,\tiny(3) & --\,\tiny(0) \\
10x10 & 0\,\tiny(4) & 0\,\tiny(2) & 0\,\tiny(6) & --\,\tiny(0) \\
\bottomrule
\end{tabular}

\begin{tabular}{lrr}
\toprule
Size & $\chi$ ($\mathsf{TC}^0$) & \#CC ($\mathsf{L}$) \\
\midrule
6x6 & 100\,\tiny(5) & 100\,\tiny(5) \\
8x8 & 100\,\tiny(4) & 100\,\tiny(5) \\
10x10 & 100\,\tiny(5) & 100\,\tiny(5) \\
12x12 & 100\,\tiny(5) & 100\,\tiny(5) \\
\bottomrule
\end{tabular}

\begin{tabular}{lrr}
\toprule
Size & \#EC$_{\to}$ ($\mathsf{GapL}$) & \#EC ($\#\mathsf{P}$) \\
\midrule
S & 100\,\tiny(4) & 100\,\tiny(4) \\
M & 100\,\tiny(4) & 100\,\tiny(2) \\
L & 100\,\tiny(4) & 100\,\tiny(3) \\
XL & 100\,\tiny(3) & 50\,\tiny(2) \\
\bottomrule
\end{tabular}

\end{center}
\paragraph{\texttt{gemini-3.8-flash} (high), Text CoT.} Exact match (\%) with the number of instances in parentheses.
\begin{center}\scriptsize
\begin{tabular}{lrrrr}
\toprule
Size & \#ST ($\mathsf{GapL}$) & \#PM ($\mathsf{GapL}$) & \#M ($\#\mathsf{P}$) & \#Paths ($\#\mathsf{P}$) \\
\midrule
3x3 & 100\,\tiny(5) & --\,\tiny(0) & 100\,\tiny(5) & 100\,\tiny(5) \\
3x4 & 100\,\tiny(4) & 100\,\tiny(5) & 100\,\tiny(5) & 100\,\tiny(5) \\
4x4 & 100\,\tiny(3) & 100\,\tiny(5) & 100\,\tiny(4) & 75\,\tiny(4) \\
4x5 & 100\,\tiny(2) & 100\,\tiny(5) & 100\,\tiny(3) & 100\,\tiny(2) \\
5x5 & 100\,\tiny(2) & --\,\tiny(0) & 67\,\tiny(3) & 25\,\tiny(4) \\
5x6 & 33\,\tiny(3) & 100\,\tiny(4) & 0\,\tiny(2) & 0\,\tiny(2) \\
6x6 & 0\,\tiny(3) & 33\,\tiny(3) & 33\,\tiny(3) & 0\,\tiny(1) \\
7x7 & 0\,\tiny(2) & --\,\tiny(0) & 0\,\tiny(5) & --\,\tiny(0) \\
8x8 & 0\,\tiny(3) & 0\,\tiny(3) & 0\,\tiny(3) & --\,\tiny(0) \\
9x9 & 0\,\tiny(2) & --\,\tiny(0) & 0\,\tiny(3) & --\,\tiny(0) \\
10x10 & 0\,\tiny(1) & 0\,\tiny(1) & 0\,\tiny(2) & --\,\tiny(0) \\
\bottomrule
\end{tabular}

\begin{tabular}{lrr}
\toprule
Size & $\chi$ ($\mathsf{TC}^0$) & \#CC ($\mathsf{L}$) \\
\midrule
6x6 & 100\,\tiny(5) & 100\,\tiny(5) \\
8x8 & 100\,\tiny(5) & 100\,\tiny(5) \\
10x10 & 100\,\tiny(5) & 100\,\tiny(5) \\
12x12 & 100\,\tiny(5) & 100\,\tiny(5) \\
\bottomrule
\end{tabular}

\begin{tabular}{lrr}
\toprule
Size & \#EC$_{\to}$ ($\mathsf{GapL}$) & \#EC ($\#\mathsf{P}$) \\
\midrule
S & 100\,\tiny(4) & 100\,\tiny(4) \\
M & 100\,\tiny(4) & 100\,\tiny(2) \\
L & 100\,\tiny(3) & 100\,\tiny(2) \\
XL & 100\,\tiny(3) & 100\,\tiny(4) \\
\bottomrule
\end{tabular}

\end{center}
\paragraph{Gemma~4 26B-A4B (thinking off), Visual CoT.} Exact match (\%) with the number of instances in parentheses.
\begin{center}\scriptsize
\begin{tabular}{lrrrr}
\toprule
Size & \#ST ($\mathsf{GapL}$) & \#PM ($\mathsf{GapL}$) & \#M ($\#\mathsf{P}$) & \#Paths ($\#\mathsf{P}$) \\
\midrule
3x3 & 80\,\tiny(5) & --\,\tiny(0) & 60\,\tiny(5) & 100\,\tiny(5) \\
3x4 & 0\,\tiny(5) & 40\,\tiny(5) & 0\,\tiny(5) & 40\,\tiny(5) \\
4x4 & 0\,\tiny(5) & 20\,\tiny(5) & 0\,\tiny(5) & 0\,\tiny(5) \\
4x5 & 0\,\tiny(5) & 0\,\tiny(5) & 0\,\tiny(5) & 0\,\tiny(5) \\
5x5 & 0\,\tiny(5) & --\,\tiny(0) & 0\,\tiny(5) & 0\,\tiny(5) \\
5x6 & 0\,\tiny(5) & 0\,\tiny(5) & 0\,\tiny(5) & 0\,\tiny(5) \\
6x6 & 0\,\tiny(5) & 0\,\tiny(5) & 0\,\tiny(5) & 0\,\tiny(5) \\
\bottomrule
\end{tabular}

\begin{tabular}{lrr}
\toprule
Size & $\chi$ ($\mathsf{TC}^0$) & \#CC ($\mathsf{L}$) \\
\midrule
6x6 & 100\,\tiny(5) & 100\,\tiny(5) \\
8x8 & 80\,\tiny(5) & 80\,\tiny(5) \\
10x10 & 100\,\tiny(5) & 100\,\tiny(5) \\
12x12 & 60\,\tiny(5) & 100\,\tiny(5) \\
\bottomrule
\end{tabular}

\begin{tabular}{lrr}
\toprule
Size & \#EC$_{\to}$ ($\mathsf{GapL}$) & \#EC ($\#\mathsf{P}$) \\
\midrule
S & 100\,\tiny(4) & 75\,\tiny(4) \\
M & 50\,\tiny(4) & 50\,\tiny(4) \\
L & 75\,\tiny(4) & 50\,\tiny(4) \\
XL & 0\,\tiny(4) & 0\,\tiny(4) \\
\bottomrule
\end{tabular}

\end{center}
\paragraph{Gemma~4 26B-A4B (thinking off), Text CoT.} Exact match (\%) with the number of instances in parentheses.
\begin{center}\scriptsize
\begin{tabular}{lrrrr}
\toprule
Size & \#ST ($\mathsf{GapL}$) & \#PM ($\mathsf{GapL}$) & \#M ($\#\mathsf{P}$) & \#Paths ($\#\mathsf{P}$) \\
\midrule
3x3 & 100\,\tiny(5) & --\,\tiny(0) & 80\,\tiny(5) & 100\,\tiny(5) \\
3x4 & 0\,\tiny(5) & 100\,\tiny(5) & 0\,\tiny(5) & 100\,\tiny(5) \\
4x4 & 20\,\tiny(5) & 20\,\tiny(5) & 0\,\tiny(5) & 20\,\tiny(5) \\
4x5 & 0\,\tiny(5) & 20\,\tiny(5) & 0\,\tiny(5) & 0\,\tiny(5) \\
5x5 & 0\,\tiny(5) & --\,\tiny(0) & 0\,\tiny(5) & 0\,\tiny(5) \\
5x6 & 0\,\tiny(5) & 0\,\tiny(5) & 0\,\tiny(5) & 0\,\tiny(5) \\
6x6 & 0\,\tiny(5) & 0\,\tiny(5) & 0\,\tiny(5) & 0\,\tiny(5) \\
\bottomrule
\end{tabular}

\begin{tabular}{lrr}
\toprule
Size & $\chi$ ($\mathsf{TC}^0$) & \#CC ($\mathsf{L}$) \\
\midrule
6x6 & 100\,\tiny(5) & 100\,\tiny(5) \\
8x8 & 100\,\tiny(5) & 100\,\tiny(5) \\
10x10 & 100\,\tiny(5) & 100\,\tiny(5) \\
12x12 & 100\,\tiny(5) & 100\,\tiny(5) \\
\bottomrule
\end{tabular}

\begin{tabular}{lrr}
\toprule
Size & \#EC$_{\to}$ ($\mathsf{GapL}$) & \#EC ($\#\mathsf{P}$) \\
\midrule
S & 75\,\tiny(4) & 25\,\tiny(4) \\
M & 100\,\tiny(4) & 25\,\tiny(4) \\
L & 50\,\tiny(4) & 25\,\tiny(4) \\
XL & 0\,\tiny(4) & 0\,\tiny(4) \\
\bottomrule
\end{tabular}

\end{center}
\paragraph{Gemma~4 31B (thinking off), Visual CoT.} Exact match (\%) with the number of instances in parentheses.
\begin{center}\scriptsize
\begin{tabular}{lrrrr}
\toprule
Size & \#ST ($\mathsf{GapL}$) & \#PM ($\mathsf{GapL}$) & \#M ($\#\mathsf{P}$) & \#Paths ($\#\mathsf{P}$) \\
\midrule
3x3 & 50\,\tiny(6) & --\,\tiny(0) & 33\,\tiny(6) & 100\,\tiny(6) \\
3x4 & 10\,\tiny(10) & 80\,\tiny(10) & 0\,\tiny(10) & 70\,\tiny(10) \\
4x4 & 10\,\tiny(10) & 20\,\tiny(10) & 0\,\tiny(10) & 10\,\tiny(10) \\
4x5 & 0\,\tiny(10) & 0\,\tiny(10) & 0\,\tiny(10) & 0\,\tiny(10) \\
5x5 & 0\,\tiny(10) & --\,\tiny(0) & 0\,\tiny(10) & 0\,\tiny(10) \\
5x6 & 0\,\tiny(10) & 0\,\tiny(10) & 0\,\tiny(10) & 0\,\tiny(10) \\
6x6 & 0\,\tiny(10) & 0\,\tiny(10) & 0\,\tiny(10) & 0\,\tiny(10) \\
7x7 & 0\,\tiny(6) & --\,\tiny(0) & 0\,\tiny(6) & --\,\tiny(0) \\
8x8 & 0\,\tiny(6) & 0\,\tiny(6) & 0\,\tiny(6) & --\,\tiny(0) \\
9x9 & 0\,\tiny(6) & --\,\tiny(0) & 0\,\tiny(6) & --\,\tiny(0) \\
10x10 & 0\,\tiny(6) & 0\,\tiny(6) & 0\,\tiny(6) & --\,\tiny(0) \\
\bottomrule
\end{tabular}

\begin{tabular}{lrr}
\toprule
Size & $\chi$ ($\mathsf{TC}^0$) & \#CC ($\mathsf{L}$) \\
\midrule
6x6 & 90\,\tiny(10) & 100\,\tiny(10) \\
8x8 & 90\,\tiny(10) & 100\,\tiny(10) \\
10x10 & 100\,\tiny(10) & 100\,\tiny(10) \\
12x12 & 80\,\tiny(10) & 80\,\tiny(10) \\
\bottomrule
\end{tabular}

\begin{tabular}{lrr}
\toprule
Size & \#EC$_{\to}$ ($\mathsf{GapL}$) & \#EC ($\#\mathsf{P}$) \\
\midrule
S & 50\,\tiny(8) & 50\,\tiny(8) \\
M & 50\,\tiny(8) & 75\,\tiny(8) \\
L & 38\,\tiny(8) & 25\,\tiny(8) \\
XL & 0\,\tiny(8) & 12\,\tiny(8) \\
\bottomrule
\end{tabular}

\end{center}
\paragraph{Gemma~4 31B (thinking off), Text CoT.} Exact match (\%) with the number of instances in parentheses.
\begin{center}\scriptsize
\begin{tabular}{lrrrr}
\toprule
Size & \#ST ($\mathsf{GapL}$) & \#PM ($\mathsf{GapL}$) & \#M ($\#\mathsf{P}$) & \#Paths ($\#\mathsf{P}$) \\
\midrule
3x3 & 100\,\tiny(6) & --\,\tiny(0) & 33\,\tiny(6) & 83\,\tiny(6) \\
3x4 & 0\,\tiny(10) & 100\,\tiny(10) & 10\,\tiny(10) & 80\,\tiny(10) \\
4x4 & 10\,\tiny(10) & 40\,\tiny(10) & 0\,\tiny(10) & 0\,\tiny(10) \\
4x5 & 0\,\tiny(10) & 0\,\tiny(10) & 0\,\tiny(10) & 0\,\tiny(10) \\
5x5 & 0\,\tiny(9) & --\,\tiny(0) & 0\,\tiny(10) & 0\,\tiny(10) \\
5x6 & 0\,\tiny(10) & 0\,\tiny(10) & 0\,\tiny(10) & 0\,\tiny(10) \\
6x6 & 0\,\tiny(10) & 0\,\tiny(10) & 0\,\tiny(10) & 0\,\tiny(10) \\
\bottomrule
\end{tabular}

\begin{tabular}{lrr}
\toprule
Size & $\chi$ ($\mathsf{TC}^0$) & \#CC ($\mathsf{L}$) \\
\midrule
6x6 & 90\,\tiny(10) & 100\,\tiny(10) \\
8x8 & 80\,\tiny(10) & 100\,\tiny(10) \\
10x10 & 100\,\tiny(10) & 100\,\tiny(10) \\
12x12 & 90\,\tiny(10) & 100\,\tiny(10) \\
\bottomrule
\end{tabular}

\begin{tabular}{lrr}
\toprule
Size & \#EC$_{\to}$ ($\mathsf{GapL}$) & \#EC ($\#\mathsf{P}$) \\
\midrule
S & 38\,\tiny(8) & 50\,\tiny(8) \\
M & 25\,\tiny(8) & 62\,\tiny(8) \\
L & 12\,\tiny(8) & 25\,\tiny(8) \\
XL & 0\,\tiny(8) & 12\,\tiny(8) \\
\bottomrule
\end{tabular}

\end{center}
\paragraph{Gemma~4 31B (thinking on), Visual CoT.} Exact match (\%) with the number of instances in parentheses.
\begin{center}\scriptsize
\begin{tabular}{lrrrr}
\toprule
Size & \#ST ($\mathsf{GapL}$) & \#PM ($\mathsf{GapL}$) & \#M ($\#\mathsf{P}$) & \#Paths ($\#\mathsf{P}$) \\
\midrule
3x3 & 100\,\tiny(1) & --\,\tiny(0) & 100\,\tiny(1) & 100\,\tiny(1) \\
3x4 & 0\,\tiny(1) & 100\,\tiny(1) & 100\,\tiny(1) & --\,\tiny(0) \\
4x4 & 100\,\tiny(1) & 100\,\tiny(1) & 0\,\tiny(1) & 100\,\tiny(1) \\
4x5 & 0\,\tiny(1) & 0\,\tiny(1) & 0\,\tiny(1) & 0\,\tiny(1) \\
5x5 & 0\,\tiny(1) & --\,\tiny(0) & --\,\tiny(0) & --\,\tiny(0) \\
\bottomrule
\end{tabular}

\begin{tabular}{lrr}
\toprule
Size & \#EC$_{\to}$ ($\mathsf{GapL}$) & \#EC ($\#\mathsf{P}$) \\
\midrule
S & 100\,\tiny(1) & 100\,\tiny(1) \\
M & 100\,\tiny(1) & 0\,\tiny(1) \\
L & 100\,\tiny(1) & 0\,\tiny(1) \\
XL & 0\,\tiny(1) & 0\,\tiny(1) \\
\bottomrule
\end{tabular}

\end{center}
\paragraph{Gemma~4 31B (thinking on), Text CoT.} Exact match (\%) with the number of instances in parentheses.
\begin{center}\scriptsize
\begin{tabular}{lrrrr}
\toprule
Size & \#ST ($\mathsf{GapL}$) & \#PM ($\mathsf{GapL}$) & \#M ($\#\mathsf{P}$) & \#Paths ($\#\mathsf{P}$) \\
\midrule
3x3 & 100\,\tiny(1) & --\,\tiny(0) & 100\,\tiny(1) & 100\,\tiny(1) \\
3x4 & 100\,\tiny(1) & 100\,\tiny(1) & 0\,\tiny(1) & 100\,\tiny(1) \\
4x4 & 100\,\tiny(1) & 100\,\tiny(1) & 100\,\tiny(1) & 100\,\tiny(1) \\
4x5 & 0\,\tiny(1) & 100\,\tiny(1) & 0\,\tiny(1) & 0\,\tiny(1) \\
5x5 & --\,\tiny(0) & --\,\tiny(0) & --\,\tiny(0) & 0\,\tiny(1) \\
\bottomrule
\end{tabular}

\begin{tabular}{lrr}
\toprule
Size & \#EC$_{\to}$ ($\mathsf{GapL}$) & \#EC ($\#\mathsf{P}$) \\
\midrule
S & --\,\tiny(0) & 100\,\tiny(1) \\
M & 100\,\tiny(1) & 100\,\tiny(1) \\
L & 100\,\tiny(1) & 0\,\tiny(1) \\
XL & 0\,\tiny(1) & 0\,\tiny(1) \\
\bottomrule
\end{tabular}

\end{center}
\paragraph{\texttt{gemini-3.1-pro-preview} (low), Visual CoT.} Exact match (\%) with the number of instances in parentheses.
\begin{center}\scriptsize
\begin{tabular}{lrrrr}
\toprule
Size & \#ST ($\mathsf{GapL}$) & \#PM ($\mathsf{GapL}$) & \#M ($\#\mathsf{P}$) & \#Paths ($\#\mathsf{P}$) \\
\midrule
3x3 & 100\,\tiny(5) & --\,\tiny(0) & 100\,\tiny(5) & 100\,\tiny(5) \\
3x4 & 60\,\tiny(5) & 100\,\tiny(5) & 100\,\tiny(5) & 80\,\tiny(5) \\
4x4 & 60\,\tiny(5) & 80\,\tiny(5) & 40\,\tiny(5) & 80\,\tiny(5) \\
4x5 & 0\,\tiny(5) & 80\,\tiny(5) & 20\,\tiny(5) & 40\,\tiny(5) \\
5x5 & 0\,\tiny(5) & --\,\tiny(0) & 0\,\tiny(5) & 0\,\tiny(5) \\
5x6 & 0\,\tiny(5) & 20\,\tiny(5) & 0\,\tiny(5) & 0\,\tiny(5) \\
6x6 & 0\,\tiny(5) & 0\,\tiny(5) & 0\,\tiny(5) & 0\,\tiny(5) \\
7x7 & 0\,\tiny(6) & --\,\tiny(0) & 0\,\tiny(6) & --\,\tiny(0) \\
8x8 & 0\,\tiny(6) & 0\,\tiny(6) & 0\,\tiny(6) & --\,\tiny(0) \\
9x9 & 0\,\tiny(6) & --\,\tiny(0) & 0\,\tiny(6) & --\,\tiny(0) \\
10x10 & 0\,\tiny(6) & 0\,\tiny(6) & 0\,\tiny(5) & --\,\tiny(0) \\
\bottomrule
\end{tabular}

\begin{tabular}{lrr}
\toprule
Size & $\chi$ ($\mathsf{TC}^0$) & \#CC ($\mathsf{L}$) \\
\midrule
6x6 & 100\,\tiny(5) & 100\,\tiny(5) \\
8x8 & 100\,\tiny(5) & 100\,\tiny(5) \\
10x10 & 100\,\tiny(5) & 100\,\tiny(5) \\
12x12 & 100\,\tiny(5) & 100\,\tiny(5) \\
\bottomrule
\end{tabular}

\begin{tabular}{lrr}
\toprule
Size & \#EC$_{\to}$ ($\mathsf{GapL}$) & \#EC ($\#\mathsf{P}$) \\
\midrule
S & 100\,\tiny(4) & 100\,\tiny(4) \\
M & 75\,\tiny(4) & 100\,\tiny(4) \\
L & 75\,\tiny(4) & 75\,\tiny(4) \\
XL & 100\,\tiny(4) & 0\,\tiny(4) \\
\bottomrule
\end{tabular}

\end{center}
\paragraph{\texttt{gemini-3.1-pro-preview} (low), Text CoT.} Exact match (\%) with the number of instances in parentheses.
\begin{center}\scriptsize
\begin{tabular}{lrrrr}
\toprule
Size & \#ST ($\mathsf{GapL}$) & \#PM ($\mathsf{GapL}$) & \#M ($\#\mathsf{P}$) & \#Paths ($\#\mathsf{P}$) \\
\midrule
3x3 & 100\,\tiny(5) & --\,\tiny(0) & 100\,\tiny(5) & 100\,\tiny(5) \\
3x4 & 60\,\tiny(5) & 100\,\tiny(5) & 100\,\tiny(5) & 100\,\tiny(5) \\
4x4 & 20\,\tiny(5) & 100\,\tiny(5) & 20\,\tiny(5) & 60\,\tiny(5) \\
4x5 & 0\,\tiny(5) & 80\,\tiny(5) & 0\,\tiny(5) & 20\,\tiny(5) \\
5x5 & 0\,\tiny(5) & --\,\tiny(0) & 0\,\tiny(5) & 0\,\tiny(5) \\
5x6 & 0\,\tiny(5) & 20\,\tiny(5) & 0\,\tiny(5) & 0\,\tiny(5) \\
6x6 & 0\,\tiny(5) & 0\,\tiny(5) & 0\,\tiny(5) & 0\,\tiny(5) \\
\bottomrule
\end{tabular}

\begin{tabular}{lrr}
\toprule
Size & $\chi$ ($\mathsf{TC}^0$) & \#CC ($\mathsf{L}$) \\
\midrule
6x6 & 100\,\tiny(5) & 100\,\tiny(5) \\
8x8 & 100\,\tiny(5) & 100\,\tiny(5) \\
10x10 & 100\,\tiny(5) & 100\,\tiny(5) \\
12x12 & 100\,\tiny(5) & 100\,\tiny(5) \\
\bottomrule
\end{tabular}

\begin{tabular}{lrr}
\toprule
Size & \#EC$_{\to}$ ($\mathsf{GapL}$) & \#EC ($\#\mathsf{P}$) \\
\midrule
S & 100\,\tiny(4) & 100\,\tiny(4) \\
M & 100\,\tiny(4) & 100\,\tiny(4) \\
L & 100\,\tiny(4) & 75\,\tiny(4) \\
XL & 100\,\tiny(4) & 25\,\tiny(4) \\
\bottomrule
\end{tabular}

\end{center}
\paragraph{\texttt{gemini-3.1-pro-preview} (high), Visual CoT.} Exact match (\%) with the number of instances in parentheses.
\begin{center}\scriptsize
\begin{tabular}{lrrrr}
\toprule
Size & \#ST ($\mathsf{GapL}$) & \#PM ($\mathsf{GapL}$) & \#M ($\#\mathsf{P}$) & \#Paths ($\#\mathsf{P}$) \\
\midrule
3x3 & 100\,\tiny(3) & --\,\tiny(0) & 100\,\tiny(3) & 100\,\tiny(3) \\
3x4 & 100\,\tiny(3) & 100\,\tiny(3) & 100\,\tiny(3) & 100\,\tiny(3) \\
4x4 & 67\,\tiny(3) & 100\,\tiny(3) & 67\,\tiny(3) & 100\,\tiny(3) \\
4x5 & 67\,\tiny(3) & 100\,\tiny(3) & 100\,\tiny(3) & 67\,\tiny(3) \\
5x5 & 0\,\tiny(3) & --\,\tiny(0) & 33\,\tiny(3) & 0\,\tiny(3) \\
5x6 & 0\,\tiny(2) & 67\,\tiny(3) & 0\,\tiny(3) & 0\,\tiny(2) \\
6x6 & 0\,\tiny(3) & 33\,\tiny(3) & 0\,\tiny(3) & 0\,\tiny(3) \\
\bottomrule
\end{tabular}

\begin{tabular}{lrr}
\toprule
Size & $\chi$ ($\mathsf{TC}^0$) & \#CC ($\mathsf{L}$) \\
\midrule
6x6 & 100\,\tiny(3) & 100\,\tiny(3) \\
8x8 & 100\,\tiny(3) & 100\,\tiny(3) \\
10x10 & 100\,\tiny(3) & 100\,\tiny(3) \\
12x12 & 100\,\tiny(3) & 100\,\tiny(3) \\
\bottomrule
\end{tabular}

\begin{tabular}{lrr}
\toprule
Size & \#EC$_{\to}$ ($\mathsf{GapL}$) & \#EC ($\#\mathsf{P}$) \\
\midrule
S & 100\,\tiny(2) & 100\,\tiny(2) \\
M & 100\,\tiny(2) & 100\,\tiny(2) \\
L & 100\,\tiny(2) & 100\,\tiny(2) \\
XL & 100\,\tiny(1) & 100\,\tiny(2) \\
\bottomrule
\end{tabular}

\end{center}
\paragraph{\texttt{gemini-3.1-pro-preview} (high), Text CoT.} Exact match (\%) with the number of instances in parentheses.
\begin{center}\scriptsize
\begin{tabular}{lrrrr}
\toprule
Size & \#ST ($\mathsf{GapL}$) & \#PM ($\mathsf{GapL}$) & \#M ($\#\mathsf{P}$) & \#Paths ($\#\mathsf{P}$) \\
\midrule
3x3 & 100\,\tiny(3) & --\,\tiny(0) & 100\,\tiny(3) & 100\,\tiny(3) \\
3x4 & 100\,\tiny(3) & 100\,\tiny(3) & 100\,\tiny(3) & 100\,\tiny(3) \\
4x4 & 100\,\tiny(3) & 100\,\tiny(3) & 100\,\tiny(3) & 100\,\tiny(3) \\
4x5 & 67\,\tiny(3) & 100\,\tiny(3) & 67\,\tiny(3) & 50\,\tiny(2) \\
5x5 & 33\,\tiny(3) & --\,\tiny(0) & 67\,\tiny(3) & 0\,\tiny(2) \\
5x6 & 0\,\tiny(2) & 50\,\tiny(2) & 50\,\tiny(2) & 0\,\tiny(3) \\
6x6 & 0\,\tiny(2) & 67\,\tiny(3) & 0\,\tiny(3) & 0\,\tiny(3) \\
\bottomrule
\end{tabular}

\begin{tabular}{lrr}
\toprule
Size & $\chi$ ($\mathsf{TC}^0$) & \#CC ($\mathsf{L}$) \\
\midrule
6x6 & 100\,\tiny(3) & 100\,\tiny(3) \\
8x8 & 100\,\tiny(3) & 100\,\tiny(3) \\
10x10 & 100\,\tiny(3) & 100\,\tiny(3) \\
12x12 & 100\,\tiny(3) & 100\,\tiny(3) \\
\bottomrule
\end{tabular}

\begin{tabular}{lrr}
\toprule
Size & \#EC$_{\to}$ ($\mathsf{GapL}$) & \#EC ($\#\mathsf{P}$) \\
\midrule
S & 100\,\tiny(2) & 100\,\tiny(2) \\
M & 100\,\tiny(2) & 100\,\tiny(2) \\
L & 100\,\tiny(2) & 100\,\tiny(2) \\
XL & 100\,\tiny(2) & 50\,\tiny(2) \\
\bottomrule
\end{tabular}

\end{center}

%% file: gen_v5r/tab_twins_visual.tex
\begin{tabular}{llrrrrl}
\toprule
Model & Twin pair & $n$ & acc$_A$ & acc$_B$ & $b/c$ & McNemar $p$ \\
\midrule
Gemini 3 Flash (low) & same graph: \#ST (GapL) vs \#M (\#P) & 35 & 17 & 9 & 4/1 & 0.375 \\
 & same graph: \#ST (GapL) vs \#Paths (\#P) & 35 & 17 & 31 & 0/5 & 0.0625 \\
 & same graph: \#PM (GapL) vs \#M (\#P) & 25 & 8 & 0 & 2/0 & 0.5 \\
 & same image: $\chi$ (TC0) vs \#CC (L) & 20 & 95 & 100 & 0/1 & 1 \\
 & same drawing $\pm$ arrows: \#EC$_{\to}$ (GapL) vs \#EC (\#P) & 16 & 38 & 56 & 1/4 & 0.375 \\
\midrule
Gemini 3.1 Flash-Lite (low) & same graph: \#ST (GapL) vs \#M (\#P) & 35 & 0 & 0 & 0/0 & 1 \\
 & same graph: \#ST (GapL) vs \#Paths (\#P) & 35 & 0 & 14 & 0/5 & 0.0625 \\
\bottomrule
\end{tabular}

%% file: gen_v5r/tab_twins_text.tex
\begin{tabular}{llrrrrl}
\toprule
Model & Twin pair & $n$ & acc$_A$ & acc$_B$ & $b/c$ & McNemar $p$ \\
\midrule
Gemini 3 Flash (low) & same graph: \#ST (GapL) vs \#M (\#P) & 35 & 20 & 9 & 4/0 & 0.125 \\
 & same graph: \#ST (GapL) vs \#Paths (\#P) & 35 & 20 & 31 & 0/4 & 0.125 \\
 & same graph: \#PM (GapL) vs \#M (\#P) & 25 & 16 & 0 & 4/0 & 0.125 \\
 & same image: $\chi$ (TC0) vs \#CC (L) & 20 & 95 & 100 & 0/1 & 1 \\
 & same drawing $\pm$ arrows: \#EC$_{\to}$ (GapL) vs \#EC (\#P) & 16 & 31 & 50 & 1/4 & 0.375 \\
\midrule
Gemini 3.1 Flash-Lite (low) & same graph: \#ST (GapL) vs \#M (\#P) & 35 & 3 & 0 & 1/0 & 1 \\
 & same graph: \#ST (GapL) vs \#Paths (\#P) & 35 & 3 & 20 & 1/7 & 0.0703 \\
\bottomrule
\end{tabular}

%% file: gen_v5r/tab_twins_hint.tex
\begin{tabular}{llrrrrl}
\toprule
Model & Twin pair & $n$ & acc$_A$ & acc$_B$ & $b/c$ & McNemar $p$ \\
\midrule
Gemini 3.6 Flash (high) & same graph: \#ST (GapL) vs \#M (\#P) & 35 & 46 & 37 & 4/1 & 0.375 \\
 & same graph: \#ST (GapL) vs \#Paths (\#P) & 35 & 46 & 46 & 2/2 & 1 \\
 & same graph: \#PM (GapL) vs \#M (\#P) & 25 & 72 & 32 & 10/0 & 0.00195 \\
 & same image: $\chi$ (TC0) vs \#CC (L) & 20 & 100 & 100 & 0/0 & 1 \\
 & same drawing $\pm$ arrows: \#EC$_{\to}$ (GapL) vs \#EC (\#P) & 16 & 100 & 94 & 1/0 & 1 \\
\midrule
Gemini 3.8 Flash (low) & same graph: \#ST (GapL) vs \#M (\#P) & 65 & 45 & 42 & 7/5 & 0.774 \\
 & same graph: \#ST (GapL) vs \#Paths (\#P) & 65 & 45 & 40 & 6/3 & 0.508 \\
 & same graph: \#PM (GapL) vs \#M (\#P) & 50 & 64 & 40 & 13/1 & 0.00183 \\
 & same image: $\chi$ (TC0) vs \#CC (L) & 40 & 100 & 100 & 0/0 & 1 \\
 & same drawing $\pm$ arrows: \#EC$_{\to}$ (GapL) vs \#EC (\#P) & 32 & 100 & 94 & 2/0 & 0.5 \\
\midrule
Gemini 3.8 Flash (high) & same graph: \#ST (GapL) vs \#M (\#P) & 22 & 82 & 91 & 0/2 & 0.5 \\
 & same graph: \#ST (GapL) vs \#Paths (\#P) & 19 & 84 & 84 & 1/1 & 1 \\
 & same graph: \#PM (GapL) vs \#M (\#P) & 13 & 100 & 85 & 2/0 & 0.5 \\
 & same image: $\chi$ (TC0) vs \#CC (L) & 20 & 100 & 100 & 0/0 & 1 \\
 & same drawing $\pm$ arrows: \#EC$_{\to}$ (GapL) vs \#EC (\#P) & 14 & 100 & 93 & 1/0 & 1 \\
\midrule
Gemma 4 31B (thinking) & same graph: \#ST (GapL) vs \#M (\#P) & 5 & 40 & 0 & 2/0 & 0.5 \\
 & same graph: \#ST (GapL) vs \#Paths (\#P) & 3 & 67 & 100 & 0/1 & 1 \\
 & same graph: \#PM (GapL) vs \#M (\#P) & 3 & 67 & 0 & 2/0 & 0.5 \\
 & same drawing $\pm$ arrows: \#EC$_{\to}$ (GapL) vs \#EC (\#P) & 4 & 50 & 75 & 0/1 & 1 \\
\bottomrule
\end{tabular}

%% file: gen_v5r/tab_tokens_visual.tex
\begin{tabular}{lrrrrrrrr}
\toprule
Model & $\chi$ & \#CC & \#ST & \#PM & \#EC$_{\to}$ & \#M & \#Paths & \#EC \\
 & \scriptsize $\mathsf{TC}^0$ & \scriptsize $\mathsf{L}$ & \scriptsize $\mathsf{GapL}$ & \scriptsize $\mathsf{GapL}$ & \scriptsize $\mathsf{GapL}$ & \scriptsize $\#\mathsf{P}$ & \scriptsize $\#\mathsf{P}$ & \scriptsize $\#\mathsf{P}$ \\
\midrule
Gemini 3 Flash (low) & 2.8k & 2.6k & 2.5k & 1.8k & 2.4k & 2.5k & 3.0k & 3.9k \\
Gemini 3.1 Flash-Lite (low) & 1.1k & 1.0k & 1.3k & 1.5k & 1.4k & 1.3k & 1.3k & 1.2k \\
Gemini 3.5 Flash (low) & 2.5k & 1.8k & 9.5k & 7.5k & 5.5k & 8.9k & 6.3k & 8.7k \\
Gemini 3.7 Flash (low) & 1.4k & 0.9k & 4.6k & 4.3k & 3.2k & 5.2k & 3.6k & 3.2k \\
Gemini 3.6 Flash (minimal) & 1.0k & 0.7k & 1.1k & 2.0k & 2.1k & 2.4k & 1.4k & 2.6k \\
Gemini 3.6 Flash (low) & 1.9k & 1.5k & 7.8k & 6.3k & 4.4k & 9.7k & 5.2k & 7.1k \\
Gemini 3.6 Flash (high) & 8.0k & 4.3k & 28.4k & 21.7k & 17.7k & 30.9k & 24.3k & 23.1k \\
Gemini 3.8 Flash (low) & 1.4k & 1.1k & 5.3k & 5.9k & 4.1k & 6.1k & 4.7k & 8.9k \\
Gemini 3.8 Flash (medium) & 2.2k & 1.6k & 13.9k & 12.3k & 7.1k & 14.8k & 13.7k & 15.1k \\
Gemini 3.8 Flash (high) & 8.8k & 3.4k & 28.6k & 25.0k & 19.3k & 31.7k & 29.8k & 25.9k \\
Gemma 4 26B-A4B (no thinking) & 1.4k & 1.0k & 8.5k & 5.7k & 7.9k & 5.5k & 6.0k & 6.2k \\
Gemma 4 31B (no thinking) & 0.9k & 0.8k & 2.1k & 1.8k & 2.7k & 2.0k & 2.1k & 1.8k \\
Gemma 4 31B (thinking) & -- & -- & 14.1k & 11.0k & 15.2k & 16.4k & 12.6k & 10.9k \\
Gemini 3.1 Pro (low) & 2.0k & 1.6k & 3.7k & 4.7k & 4.8k & 4.7k & 4.0k & 5.5k \\
Gemini 3.1 Pro (high) & 8.5k & 4.0k & 32.2k & 26.4k & 20.1k & 34.6k & 29.4k & 24.2k \\
\bottomrule
\end{tabular}

%% file: gen_v5r/tab_tokens_text.tex
\begin{tabular}{lrrrrrrrr}
\toprule
Model & $\chi$ & \#CC & \#ST & \#PM & \#EC$_{\to}$ & \#M & \#Paths & \#EC \\
 & \scriptsize $\mathsf{TC}^0$ & \scriptsize $\mathsf{L}$ & \scriptsize $\mathsf{GapL}$ & \scriptsize $\mathsf{GapL}$ & \scriptsize $\mathsf{GapL}$ & \scriptsize $\#\mathsf{P}$ & \scriptsize $\#\mathsf{P}$ & \scriptsize $\#\mathsf{P}$ \\
\midrule
Gemini 3 Flash (low) & 2.2k & 1.8k & 3.7k & 2.8k & 2.4k & 2.9k & 2.8k & 2.5k \\
Gemini 3.1 Flash-Lite (low) & 1.0k & 0.9k & 1.0k & 0.9k & 1.3k & 0.9k & 0.8k & 1.0k \\
Gemini 3.5 Flash (low) & 2.4k & 1.2k & 9.4k & 7.5k & 5.0k & 9.7k & 5.4k & 8.2k \\
Gemini 3.7 Flash (low) & 1.0k & 0.6k & 3.5k & 3.7k & 2.3k & 4.1k & 2.5k & 3.3k \\
Gemini 3.6 Flash (minimal) & 1.0k & 0.7k & 1.9k & 2.1k & 1.9k & 2.5k & 1.5k & 2.4k \\
Gemini 3.6 Flash (low) & 1.6k & 1.1k & 9.6k & 7.0k & 4.0k & 10.7k & 6.2k & 10.0k \\
Gemini 3.6 Flash (high) & 8.5k & 3.4k & 27.6k & 22.0k & 16.6k & 29.1k & 24.3k & 22.2k \\
Gemini 3.8 Flash (low) & 1.2k & 0.7k & 4.5k & 3.6k & 3.5k & 4.4k & 3.4k & 4.5k \\
Gemini 3.8 Flash (medium) & 1.9k & 1.0k & 16.1k & 12.0k & 5.6k & 13.8k & 13.6k & 12.7k \\
Gemini 3.8 Flash (high) & 10.3k & 2.3k & 28.0k & 23.3k & 16.8k & 30.8k & 28.7k & 26.6k \\
Gemma 4 26B-A4B (no thinking) & 1.7k & 1.1k & 13.3k & 12.4k & 11.3k & 8.1k & 7.6k & 11.5k \\
Gemma 4 31B (no thinking) & 1.1k & 0.9k & 2.9k & 3.4k & 2.8k & 3.2k & 3.8k & 1.6k \\
Gemma 4 31B (thinking) & -- & -- & 12.9k & 13.9k & 16.6k & 16.0k & 13.1k & 12.5k \\
Gemini 3.1 Pro (low) & 1.7k & 1.3k & 2.7k & 4.5k & 3.6k & 3.6k & 2.9k & 4.2k \\
Gemini 3.1 Pro (high) & 7.2k & 2.8k & 28.0k & 24.0k & 13.8k & 30.6k & 25.0k & 24.5k \\
\bottomrule
\end{tabular}

%% file: gen_v5r/tab_tokens_hint.tex
\begin{tabular}{lrrrrrrrr}
\toprule
Model & $\chi$ & \#CC & \#ST & \#PM & \#EC$_{\to}$ & \#M & \#Paths & \#EC \\
 & \scriptsize $\mathsf{TC}^0$ & \scriptsize $\mathsf{L}$ & \scriptsize $\mathsf{GapL}$ & \scriptsize $\mathsf{GapL}$ & \scriptsize $\mathsf{GapL}$ & \scriptsize $\#\mathsf{P}$ & \scriptsize $\#\mathsf{P}$ & \scriptsize $\#\mathsf{P}$ \\
\midrule
Gemini 3.6 Flash (high) & 7.0k & 4.8k & 28.3k & 20.3k & 15.8k & 29.2k & 23.6k & 18.3k \\
Gemini 3.8 Flash (low) & 2.0k & 1.1k & 6.3k & 7.0k & 4.4k & 7.0k & 5.7k & 7.9k \\
Gemini 3.8 Flash (high) & 6.3k & 3.6k & 30.3k & 25.9k & 20.4k & 34.0k & 27.8k & 24.7k \\
Gemma 4 31B (thinking) & -- & -- & 14.4k & 14.1k & 16.7k & 16.1k & 11.6k & 16.2k \\
\bottomrule
\end{tabular}

%% file: gen_v5r/tab_logerr_visual.tex
\begin{tabular}{lrrrrrrrr}
\toprule
Model & $\chi$ & \#CC & \#ST & \#PM & \#EC$_{\to}$ & \#M & \#Paths & \#EC \\
\midrule
Gemini 3 Flash (low) & 0.00 & 0.00 & 0.49 & 0.40 & 0.30 & 0.29 & 0.22 & 0.00 \\
Gemini 3.1 Flash-Lite (low) & 0.00 & 0.00 & 1.01 & 0.36 & 0.00 & 0.43 & 0.48 & 0.60 \\
Gemini 3.5 Flash (low) & 0.00 & 0.00 & 0.19 & 0.00 & 0.00 & 0.08 & 0.08 & 0.00 \\
Gemini 3.7 Flash (low) & 0.00 & 0.00 & 0.05 & 0.00 & 0.00 & 0.01 & 0.03 & 0.00 \\
Gemini 3.6 Flash (minimal) & 0.00 & 0.00 & 0.89 & 0.22 & 0.00 & 0.43 & 0.16 & 0.51 \\
Gemini 3.6 Flash (low) & 0.00 & 0.00 & 0.46 & 0.05 & 0.00 & 0.07 & 0.09 & 0.00 \\
Gemini 3.6 Flash (high) & 0.00 & 0.00 & 0.00 & 0.00 & 0.00 & 0.00 & 0.01 & 0.00 \\
Gemini 3.8 Flash (low) & 0.00 & 0.00 & 0.14 & 0.00 & 0.00 & 0.06 & 0.06 & 0.00 \\
Gemini 3.8 Flash (medium) & 0.00 & 0.00 & 0.00 & 0.00 & 0.00 & 0.00 & 0.00 & 0.00 \\
Gemini 3.8 Flash (high) & 0.00 & 0.00 & 0.00 & 0.00 & 0.00 & 0.00 & 0.00 & 0.00 \\
Gemma 4 26B-A4B (no thinking) & 0.00 & 0.00 & 0.89 & 0.33 & 0.00 & 0.68 & 0.70 & 0.00 \\
Gemma 4 31B (no thinking) & 0.00 & 0.00 & 0.92 & 0.18 & 0.18 & 0.69 & 0.49 & 0.12 \\
Gemma 4 31B (thinking) & -- & -- & 0.01 & 0.00 & 0.00 & 0.00 & 0.00 & 0.30 \\
Gemini 3.1 Pro (low) & 0.00 & 0.00 & 0.60 & 0.00 & 0.00 & 0.14 & 0.08 & 0.00 \\
Gemini 3.1 Pro (high) & 0.00 & 0.00 & 0.00 & 0.00 & 0.00 & 0.00 & 0.00 & 0.00 \\
\bottomrule
\end{tabular}

%% file: gen_v5r/tab_perception.tex
\begin{longtable}{llrrr}
\caption{Transcription quality on the Visual, Hint and Route-only conditions: mean edge F1 (grids, Eulerian drawings) or black-cell F1 (binary images) and the fraction of exact transcriptions, by configuration and size.}\label{tab:perception}\\
\toprule
Model & Family / size & $n$ & mean F1 & exact (\%) \\
\midrule
\endfirsthead
\toprule
Model & Family / size & $n$ & mean F1 & exact (\%) \\
\midrule
\endhead
\midrule
\multicolumn{5}{r}{\emph{continued on next page}}\\
\endfoot
\bottomrule
\endlastfoot
Gemini 3 Flash (low) & grid 3x3 & 15 & 1.000 & 100 \\
 & grid 3x4 & 20 & 1.000 & 100 \\
 & grid 4x4 & 20 & 0.996 & 80 \\
 & grid 4x5 & 20 & 0.996 & 80 \\
 & grid 5x5 & 15 & 0.999 & 93 \\
 & grid 5x6 & 20 & 0.986 & 15 \\
 & grid 6x6 & 20 & 0.987 & 40 \\
 & bin 6x6 & 10 & 1.000 & 100 \\
\end{longtable}

%% file: gen_v5r/tab_car_class.tex
\begin{tabular}{lrrrrrrrr}
\toprule
Model & $\chi$ & \#CC & \#ST & \#PM & \#EC$_{\to}$ & \#M & \#Paths & \#EC \\
 & \scriptsize $\mathsf{TC}^0$ & \scriptsize $\mathsf{L}$ & \scriptsize $\mathsf{GapL}$ & \scriptsize $\mathsf{GapL}$ & \scriptsize $\mathsf{GapL}$ & \scriptsize $\#\mathsf{P}$ & \scriptsize $\#\mathsf{P}$ & \scriptsize $\#\mathsf{P}$ \\
\midrule
Gemini 3.6 Flash (low) & 100 & 100 & 100 & 63 & 100 & 100 & 100 & 100 \\
Gemini 3.8 Flash (low) & 100 & 100 & 100 & 98 & 100 & 100 & 100 & 66 \\
Gemma 4 31B (no thinking) & 100 & 100 & 100 & 95 & 100 & 100 & 100 & 91 \\
Gemma 4 31B (thinking) & -- & -- & 100 & 0 & 100 & 100 & 100 & 100 \\
Gemini 3.1 Pro (low) & 100 & 100 & 100 & 57 & 100 & 100 & 100 & 100 \\
\bottomrule
\end{tabular}

%% file: gen_v8p/gen_v5r/tab_xl.tex
\begin{tabular}{llrrrrrrrrrrr}
\toprule
Model & Condition & 7x7 \#ST & 7x7 \#M & 8x8 \#ST & 8x8 \#PM & 8x8 \#M & 9x9 \#ST & 9x9 \#M & 10x10 \#ST & 10x10 \#PM & 10x10 \#M & tokens \\
 & & \scriptsize $\mathsf{GapL}$ & \scriptsize $\#\mathsf{P}$ & \scriptsize $\mathsf{GapL}$ & \scriptsize $\mathsf{GapL}$ & \scriptsize $\#\mathsf{P}$ & \scriptsize $\mathsf{GapL}$ & \scriptsize $\#\mathsf{P}$ & \scriptsize $\mathsf{GapL}$ & \scriptsize $\mathsf{GapL}$ & \scriptsize $\#\mathsf{P}$ & \\
\midrule
Gemini 3.6 Flash (low) & Visual$\to$Code & 17\,\tiny(6) & 67\,\tiny(6) & 17\,\tiny(6) & 17\,\tiny(6) & 0\,\tiny(6) & 17\,\tiny(6) & 17\,\tiny(6) & 0\,\tiny(6) & 0\,\tiny(6) & 0\,\tiny(6) & 4.2k \\
 & Text$\to$Code & 100\,\tiny(6) & 83\,\tiny(6) & 83\,\tiny(6) & 67\,\tiny(6) & 100\,\tiny(6) & 67\,\tiny(6) & 100\,\tiny(6) & 83\,\tiny(6) & 100\,\tiny(6) & 100\,\tiny(6) & 3.1k \\
 & CAR & 100\,\tiny(6) & 67\,\tiny(6) & 33\,\tiny(6) & 33\,\tiny(6) & 0\,\tiny(6) & 17\,\tiny(6) & 0\,\tiny(6) & 0\,\tiny(6) & 0\,\tiny(6) & 0\,\tiny(6) & 3.6k \\
 & Text CAR & 100\,\tiny(6) & 100\,\tiny(6) & 100\,\tiny(6) & 50\,\tiny(6) & 100\,\tiny(6) & 83\,\tiny(6) & 83\,\tiny(6) & 100\,\tiny(6) & 67\,\tiny(6) & 67\,\tiny(6) & 2.7k \\
\midrule
Gemini 3.8 Flash (low) & Visual CoT & 0\,\tiny(6) & 0\,\tiny(6) & 0\,\tiny(6) & 0\,\tiny(6) & 0\,\tiny(6) & 0\,\tiny(6) & 0\,\tiny(6) & 0\,\tiny(6) & 0\,\tiny(6) & 0\,\tiny(6) & 5.6k \\
 & Text CoT & 0\,\tiny(6) & 0\,\tiny(6) & 0\,\tiny(6) & 0\,\tiny(6) & 0\,\tiny(6) & 0\,\tiny(6) & 0\,\tiny(6) & 0\,\tiny(6) & 0\,\tiny(6) & 0\,\tiny(6) & 4.4k \\
 & Visual$\to$Code & 33\,\tiny(6) & 50\,\tiny(6) & 17\,\tiny(6) & 0\,\tiny(6) & 0\,\tiny(6) & 0\,\tiny(6) & 0\,\tiny(6) & 0\,\tiny(6) & 0\,\tiny(6) & 0\,\tiny(6) & 4.7k \\
 & Text$\to$Code & 100\,\tiny(6) & 100\,\tiny(6) & 83\,\tiny(6) & 100\,\tiny(6) & 100\,\tiny(6) & 100\,\tiny(6) & 83\,\tiny(6) & 100\,\tiny(6) & 83\,\tiny(6) & 100\,\tiny(6) & 2.2k \\
 & Perceive+Code & 83\,\tiny(6) & 50\,\tiny(6) & 17\,\tiny(6) & 33\,\tiny(6) & 33\,\tiny(6) & 0\,\tiny(6) & 17\,\tiny(6) & 0\,\tiny(6) & 0\,\tiny(6) & 0\,\tiny(6) & 3.9k \\
 & CAR & 83\,\tiny(6) & 67\,\tiny(6) & 17\,\tiny(6) & 50\,\tiny(6) & 17\,\tiny(6) & 0\,\tiny(6) & 0\,\tiny(6) & 0\,\tiny(6) & 0\,\tiny(6) & 0\,\tiny(6) & 3.6k \\
 & Text CAR & 100\,\tiny(6) & 100\,\tiny(6) & 100\,\tiny(6) & 33\,\tiny(6) & 100\,\tiny(6) & 100\,\tiny(6) & 83\,\tiny(6) & 100\,\tiny(6) & 83\,\tiny(6) & 100\,\tiny(6) & 2.5k \\
\midrule
Gemini 3.8 Flash (high) & Visual CoT & 0\,\tiny(3) & 0\,\tiny(5) & 0\,\tiny(4) & 0\,\tiny(2) & 0\,\tiny(4) & 0\,\tiny(4) & 0\,\tiny(3) & 0\,\tiny(4) & 0\,\tiny(2) & 0\,\tiny(6) & 40.8k \\
 & Text CoT & 0\,\tiny(2) & 0\,\tiny(5) & 0\,\tiny(3) & 0\,\tiny(3) & 0\,\tiny(3) & 0\,\tiny(2) & 0\,\tiny(3) & 0\,\tiny(1) & 0\,\tiny(1) & 0\,\tiny(2) & 45.9k \\
\midrule
Gemma 4 31B (no thinking) & Visual CoT & 0\,\tiny(6) & 0\,\tiny(6) & 0\,\tiny(6) & 0\,\tiny(6) & 0\,\tiny(6) & 0\,\tiny(6) & 0\,\tiny(6) & 0\,\tiny(6) & 0\,\tiny(6) & 0\,\tiny(6) & 3.5k \\
 & Visual$\to$Code & 0\,\tiny(6) & 0\,\tiny(6) & 0\,\tiny(6) & 0\,\tiny(6) & 0\,\tiny(6) & 0\,\tiny(6) & 0\,\tiny(6) & 0\,\tiny(6) & 0\,\tiny(6) & 0\,\tiny(6) & 3.5k \\
 & CAR & 0\,\tiny(6) & 0\,\tiny(6) & 0\,\tiny(6) & 0\,\tiny(6) & 0\,\tiny(6) & 0\,\tiny(6) & 0\,\tiny(6) & 0\,\tiny(6) & 0\,\tiny(6) & 0\,\tiny(6) & 2.5k \\
 & Text CAR & 100\,\tiny(6) & 67\,\tiny(6) & 100\,\tiny(6) & 100\,\tiny(6) & 100\,\tiny(6) & 100\,\tiny(6) & 83\,\tiny(6) & 100\,\tiny(6) & 33\,\tiny(6) & 67\,\tiny(6) & 3.3k \\
\midrule
Gemini 3.1 Pro (low) & Visual CoT & 0\,\tiny(6) & 0\,\tiny(6) & 0\,\tiny(6) & 0\,\tiny(6) & 0\,\tiny(6) & 0\,\tiny(6) & 0\,\tiny(6) & 0\,\tiny(6) & 0\,\tiny(6) & 0\,\tiny(5) & 4.5k \\
 & Visual$\to$Code & 0\,\tiny(6) & 17\,\tiny(6) & 0\,\tiny(6) & 0\,\tiny(6) & 17\,\tiny(6) & 0\,\tiny(6) & 0\,\tiny(6) & 0\,\tiny(6) & 0\,\tiny(6) & 0\,\tiny(6) & 4.5k \\
 & CAR & 17\,\tiny(6) & 17\,\tiny(6) & 0\,\tiny(6) & 0\,\tiny(6) & 0\,\tiny(6) & 0\,\tiny(6) & 0\,\tiny(6) & 0\,\tiny(6) & 0\,\tiny(6) & 0\,\tiny(6) & 4.1k \\
 & Text CAR & 100\,\tiny(6) & 100\,\tiny(6) & 100\,\tiny(6) & 100\,\tiny(6) & 100\,\tiny(6) & 100\,\tiny(6) & 100\,\tiny(6) & 100\,\tiny(6) & 100\,\tiny(6) & 100\,\tiny(6) & 3.5k \\
\bottomrule
\end{tabular}

%% file: gen_v5r/tab_orient.tex
\begin{tabular}{lrrrrr}
\toprule
Reading & exact grids & missed H & missed V & spurious H & spurious V \\
\midrule
upright, row by row (Zoom-1) & 13/24 & 38/1384 & 6/1393 & 4 & 3 \\
upright, column by column & 9/24 & 81/1384 & 15/1393 & 1 & 4 \\
rotated $90^\circ$, row by row & 2/24 & 27/1384 & 101/1393 & 24 & 8 \\
union of upright and rotated & 12/24 & 0/1384 & 0/1393 & 26 & 11 \\
two views + verification & 22/24 & \multicolumn{2}{c}{0 missed} & \multicolumn{2}{c}{2 spurious} \\
\bottomrule
\end{tabular}

%% file: gen_v5r/tab_runctx.tex
\begin{tabular}{llrr}
\toprule
Reading & segment & missed & rate \\
\midrule
whole image & horizontal, inside a run & 30/783 & 3.8\% \\
whole image & horizontal, other & 8/601 & 1.3\% \\
whole image & vertical, inside a run & 6/787 & 0.8\% \\
whole image & vertical, other & 0/606 & 0.0\% \\
\midrule
Zoom-2/3 crops & horizontal, inside a run & 274/2087 & 13.1\% \\
Zoom-2/3 crops & horizontal, other & 25/1554 & 1.6\% \\
Zoom-2/3 crops & vertical, inside a run & 2/2127 & 0.1\% \\
Zoom-2/3 crops & vertical, other & 0/1541 & 0.0\% \\
\bottomrule
\end{tabular}

%% file: gen_v5r/tab_twoview_models.tex
\begin{tabular}{lrrrrrrrrr}
\toprule
Configuration & \multicolumn{3}{c}{grids transcribed exactly} & F1 & checks & \multicolumn{4}{c}{pairs solved} \\
\cmidrule(lr){2-4}\cmidrule(lr){7-10}
 & one shot & two views & + verification & & per grid & \car{} one shot & Two-view \car{} & Two-view Code & GT text \car{} \\
\midrule
\texttt{gemini-3.8-flash} (low) & 13/24 & 12/24 & 22/24 & 1.000 & 8.6 & 14/60 & 43/60 & 49/60 & 54/60 \\
\texttt{gemini-3.6-flash} (low) & 11/24 & 6/24 & 18/24 & 0.998 & 11.2 & 15/60 & 40/60 & 39/60 & 51/60 \\
\texttt{gemini-3.1-pro-preview} (low) & 7/24 & 1/24 & 15/24 & 0.997 & 8.7 & 2/60 & 36/60 & 35/60 & 60/60 \\
\bottomrule
\end{tabular}

%% file: gen_v5r/tab_algo_size.tex
\begin{tabular}{lllrrrrrrrrrrr}
\toprule
Condition & Task & method & $3{\times}3$ & $3{\times}4$ & $4{\times}4$ & $4{\times}5$ & $5{\times}5$ & $5{\times}6$ & $6{\times}6$ & $7{\times}7$ & $8{\times}8$ & $9{\times}9$ & $10{\times}10$ \\
\midrule
Code & $\sST$ & det. & 100 & 100 & 100 & 100 & 100 & 100 & 100 & 100 & 100 & 100 & 67 \\
Code & $\sPM$ & det. & -- & 0 & 0 & 0 & -- & 0 & 0 & -- & 17 & -- & 50 \\
Code & $\sAM$ & DP & 0 & 10 & 40 & 90 & 90 & 100 & 100 & 100 & 100 & 100 & 100 \\
Code & $\sPATH$ & DP & 0 & 0 & 0 & 0 & 0 & 0 & 10 & -- & -- & -- & -- \\
\midrule
Text$\to$Code & $\sST$ & det. & 83 & 100 & 100 & 100 & 100 & 100 & 100 & 100 & 100 & 100 & 100 \\
Text$\to$Code & $\sPM$ & det. & -- & 0 & 0 & 0 & -- & 0 & 10 & -- & 67 & -- & 83 \\
Text$\to$Code & $\sAM$ & DP & 0 & 20 & 100 & 100 & 100 & 100 & 100 & 100 & 100 & 100 & 100 \\
Text$\to$Code & $\sPATH$ & DP & 0 & 0 & 0 & 0 & 0 & 0 & 10 & -- & -- & -- & -- \\
\midrule
\car{} & $\sST$ & det. & 100 & 100 & 100 & 100 & 100 & 100 & 100 & 100 & 100 & 100 & 100 \\
\car{} & $\sPM$ & det. & -- & 10 & 0 & 20 & -- & 10 & 40 & -- & 67 & -- & 83 \\
\car{} & $\sAM$ & DP & 17 & 90 & 80 & 100 & 100 & 100 & 100 & 100 & 100 & 100 & 100 \\
\car{} & $\sPATH$ & DP & 17 & 30 & 20 & 20 & 0 & 0 & 10 & -- & -- & -- & -- \\
\bottomrule
\end{tabular}

%% file: gen_v8p/gen_v5r/tab_attrib.tex
\begin{tabular}{llrrrrrrr}
\toprule
Model & Condition & programs & wrong & transcription & algorithm & both & unscorable & mismatch on correct \\
\midrule
Gemini 3.6 Flash (low) & Visual$\to$Code & 190 & 62 & 11 & 10 & 7 & 34 & 1/128 \\
 & Text$\to$Code & 60 & 7 & 0 & 6 & 0 & 1 & 0/53 \\
 & CAR & 190 & 54 & 18 & 4 & 4 & 28 & 3/136 \\
 & Text CAR & 60 & 9 & 1 & 8 & 0 & 0 & 0/51 \\
Gemini 3.8 Flash (low) & Visual$\to$Code & 506 & 101 & 37 & 3 & 10 & 51 & 6/405 \\
 & Text$\to$Code & 308 & 17 & 0 & 17 & 0 & 0 & 0/291 \\
 & Hint$\to$Code & 130 & 16 & 6 & 5 & 1 & 4 & 0/114 \\
 & Perceive+Code & 308 & 71 & 32 & 6 & 12 & 21 & 2/237 \\
 & CAR & 506 & 101 & 53 & 9 & 13 & 26 & 40/405 \\
 & Text CAR & 308 & 12 & 0 & 11 & 1 & 0 & 11/296 \\
Gemma 4 31B (no thinking) & Visual$\to$Code & 308 & 178 & 49 & 1 & 5 & 123 & 2/130 \\
 & Hint$\to$Code & 130 & 65 & 26 & 5 & 6 & 28 & 0/65 \\
 & Perceive+Code & 130 & 70 & 49 & 0 & 7 & 14 & 3/60 \\
 & CAR & 506 & 315 & 137 & 7 & 23 & 148 & 0/191 \\
 & Text CAR & 60 & 9 & 0 & 7 & 0 & 2 & 0/51 \\
Gemini 3.1 Pro (low) & Visual$\to$Code & 190 & 86 & 54 & 0 & 6 & 26 & 1/104 \\
 & Perceive+Code & 130 & 19 & 12 & 1 & 0 & 6 & 0/111 \\
 & CAR & 190 & 83 & 59 & 0 & 4 & 20 & 1/107 \\
 & Text CAR & 60 & 0 & 0 & 0 & 0 & 0 & 0/60 \\
\bottomrule
\end{tabular}

%% file: gen_v8p/gen_v5r/tab_methods_g38_low.tex
\begin{tabular}{lrrrrrrrrr}
\toprule
Condition & $\chi$ & \#CC & \#ST & \#PM & \#EC$_{\to}$ & \#M & \#Paths & \#EC & tokens \\
\midrule
Visual CoT & 100 & 100 & 38 & 70 & 97 & 42 & 39 & 72 & 4.8k \\
Text CoT & 100 & 100 & 42 & 60 & 94 & 35 & 38 & 71 & 3.4k \\
Visual+Hint & 100 & 100 & 45 & 64 & 100 & 41 & 41 & 94 & 5.4k \\
Route, no code & 100 & 100 & 54 & 68 & 94 & 40 & 49 & 69 & 5.0k \\
Visual$\to$Code & 100 & 100 & 92 & 88 & 97 & 83 & 92 & 100 & 1.7k \\
Text$\to$Code & 98 & 100 & 91 & 98 & 94 & 91 & 98 & 100 & 0.9k \\
Hint$\to$Code & 100 & 100 & 86 & 84 & 81 & 89 & 91 & 81 & 1.8k \\
Perceive+Code & 100 & 100 & 88 & 90 & 91 & 88 & 94 & 100 & 1.5k \\
CAR & 95 & 100 & 86 & 76 & 100 & 86 & 92 & 100 & 1.5k \\
Text CAR & 100 & 100 & 98 & 98 & 100 & 98 & 95 & 100 & 1.2k \\
\bottomrule
\end{tabular}

%% file: gen_v8p/gen_v5r/tab_heldout.tex
\begin{tabular}{llrrrrr}
\toprule
Model & Condition & \#IS & \#Dom & \#ShortestPaths & DP/poly route (\%) & tokens \\
 & & \scriptsize $\#\mathsf{P}$-c & \scriptsize $\#\mathsf{P}$-c & \scriptsize $\mathsf{FP}$ & & \\
\midrule
Gemini 3.8 Flash (low) & Visual CoT & 62\,\tiny(66) & 15\,\tiny(66) & 91\,\tiny(66) & -- & 4.6k \\
 & Visual$\to$Code & 89\,\tiny(66) & 86\,\tiny(66) & 94\,\tiny(66) & 65 & 1.8k \\
 & CAR & 88\,\tiny(66) & 88\,\tiny(66) & 94\,\tiny(66) & 81 & 1.6k \\
\midrule
Gemma 4 31B (no thinking) & CAR & 48\,\tiny(66) & 36\,\tiny(66) & 44\,\tiny(66) & 88 & 1.9k \\
\bottomrule
\end{tabular}

%% file: gen_v8p/gen_v5r/tab_methods_all.tex
\paragraph{\texttt{gemini-3.6-flash} (low).} Exact match (\%) per task and mean tokens per call for every condition run on this configuration (core benchmark only).
\begin{center}\scriptsize
\begin{tabular}{lrrrrrrrrr}
\toprule
Condition & $\chi$ & \#CC & \#ST & \#PM & \#EC$_{\to}$ & \#M & \#Paths & \#EC & tokens \\
\midrule
Visual CoT & 100 & 100 & 26 & 40 & 94 & 25 & 36 & 56 & 5.9k \\
Text CoT & 98 & 100 & 29 & 60 & 91 & 38 & 29 & 62 & 6.8k \\
Visual$\to$Code & 100 & 100 & 86 & 100 & 94 & 86 & 97 & 100 & 1.6k \\
CAR & 100 & 100 & 94 & 92 & 94 & 86 & 100 & 94 & 1.9k \\
\bottomrule
\end{tabular}

\end{center}
\paragraph{\texttt{gemini-3.8-flash} (low).} Exact match (\%) per task and mean tokens per call for every condition run on this configuration (core benchmark only).
\begin{center}\scriptsize
\begin{tabular}{lrrrrrrrrr}
\toprule
Condition & $\chi$ & \#CC & \#ST & \#PM & \#EC$_{\to}$ & \#M & \#Paths & \#EC & tokens \\
\midrule
Visual CoT & 100 & 100 & 38 & 70 & 97 & 42 & 39 & 72 & 4.8k \\
Text CoT & 100 & 100 & 42 & 60 & 94 & 35 & 38 & 71 & 3.4k \\
Visual+Hint & 100 & 100 & 45 & 64 & 100 & 41 & 41 & 94 & 5.4k \\
Route, no code & 100 & 100 & 54 & 68 & 94 & 40 & 49 & 69 & 5.0k \\
Visual$\to$Code & 100 & 100 & 92 & 88 & 97 & 83 & 92 & 100 & 1.7k \\
Text$\to$Code & 98 & 100 & 91 & 98 & 94 & 91 & 98 & 100 & 0.9k \\
Hint$\to$Code & 100 & 100 & 86 & 84 & 81 & 89 & 91 & 81 & 1.8k \\
Perceive+Code & 100 & 100 & 88 & 90 & 91 & 88 & 94 & 100 & 1.5k \\
CAR & 95 & 100 & 86 & 76 & 100 & 86 & 92 & 100 & 1.5k \\
Text CAR & 100 & 100 & 98 & 98 & 100 & 98 & 95 & 100 & 1.2k \\
\bottomrule
\end{tabular}

\end{center}
\paragraph{Gemma~4 31B (thinking off).} Exact match (\%) per task and mean tokens per call for every condition run on this configuration (core benchmark only).
\begin{center}\scriptsize
\begin{tabular}{lrrrrrrrrr}
\toprule
Condition & $\chi$ & \#CC & \#ST & \#PM & \#EC$_{\to}$ & \#M & \#Paths & \#EC & tokens \\
\midrule
Visual CoT & 90 & 95 & 8 & 20 & 34 & 3 & 21 & 41 & 1.8k \\
Text CoT & 90 & 100 & 11 & 28 & 19 & 5 & 20 & 38 & 2.7k \\
Visual$\to$Code & 100 & 98 & 48 & 54 & 91 & 61 & 47 & 81 & 0.9k \\
Hint$\to$Code & 100 & 100 & 49 & 44 & 75 & 51 & 54 & 62 & 1.0k \\
Perceive+Code & 100 & 100 & 49 & 48 & 94 & 46 & 43 & 94 & 1.1k \\
CAR & 100 & 100 & 45 & 40 & 88 & 44 & 41 & 88 & 1.2k \\
\bottomrule
\end{tabular}

\end{center}
\paragraph{Gemma~4 31B (thinking on).} Exact match (\%) per task and mean tokens per call for every condition run on this configuration (core benchmark only).
\begin{center}\scriptsize
\begin{tabular}{lrrrrrrr}
\toprule
Condition & \#ST & \#PM & \#EC$_{\to}$ & \#M & \#Paths & \#EC & tokens \\
\midrule
Visual CoT & 40 & 67 & 75 & 50 & 67 & 25 & 13.5k \\
Text CoT & 75 & 100 & 67 & 50 & 60 & 50 & 14.0k \\
Visual+Hint & 40 & 67 & 50 & 0 & 100 & 75 & 15.0k \\
Visual$\to$Code & 100 & 100 & 75 & 100 & 100 & 100 & 8.0k \\
CAR & 100 & 100 & 75 & 100 & 80 & 100 & 7.3k \\
\bottomrule
\end{tabular}

\end{center}
\paragraph{\texttt{gemini-3.1-pro-preview} (low).} Exact match (\%) per task and mean tokens per call for every condition run on this configuration (core benchmark only).
\begin{center}\scriptsize
\begin{tabular}{lrrrrrrrrr}
\toprule
Condition & $\chi$ & \#CC & \#ST & \#PM & \#EC$_{\to}$ & \#M & \#Paths & \#EC & tokens \\
\midrule
Visual CoT & 100 & 100 & 31 & 56 & 88 & 37 & 43 & 69 & 3.9k \\
Text CoT & 100 & 100 & 26 & 60 & 100 & 31 & 40 & 75 & 3.1k \\
Visual$\to$Code & 100 & 100 & 77 & 80 & 100 & 83 & 74 & 100 & 2.5k \\
Perceive+Code & 100 & 100 & 83 & 84 & 100 & 83 & 91 & 100 & 2.2k \\
CAR & 100 & 100 & 86 & 72 & 100 & 86 & 77 & 100 & 2.4k \\
\bottomrule
\end{tabular}

\end{center}

%% file: gen_v5r/code_examples.tex
\paragraph{Algorithm-family classifier.} Programs are labelled by a two-pass rule-based classifier (\texttt{codeclass.py}, released with the code). Pass~1 inspects only executable code: comments and string literals are blanked out with the Python tokenizer, so that a model's commentary about methods it does \emph{not} use cannot trigger a label. It looks for structural evidence of linear algebra (identifiers or calls such as \texttt{det}, \texttt{slogdet}, \texttt{linalg}, \texttt{Fraction}, Laplacian/Kirchhoff/Bareiss/Pfaffian/Kasteleyn/cofactor, Gaussian elimination or pivoting, the construction \texttt{M[u][v] -= 1} or \texttt{= -1}, NumPy/SymPy), of dynamic programming (\texttt{dp}, \texttt{memo}, \texttt{lru\_cache}/\texttt{cache}, profile, frontier, transfer, states, layer, table, or a \texttt{for mask, cnt in cur.items()} transfer loop), and of explicit enumeration (recursive \texttt{dfs}/\texttt{backtrack}/\texttt{search} functions, \texttt{visited}/\texttt{used} bookkeeping, \texttt{itertools}, bitmask recursion). On the tractable tasks ($\sST$, $\sPM$, $\sECd$) algebraic evidence takes precedence over dp; on the $\#\mathsf{P}$ tasks dp takes precedence over algebraic; enumeration evidence alone gives \emph{backtracking}. Pass~2 is consulted only when pass~1 finds nothing (typically a hand-written elimination with anonymous loop variables): it applies the same key sets to comments and docstrings, ignoring every sentence that contains a contrast word (\emph{unlike}, \emph{not}, \emph{instead}, \emph{rather than}, \emph{cannot}, \emph{\#P}, \ldots). $\sCC$ programs are labelled \emph{traversal}; $\chi$ programs are \emph{local formula} if they reference $2{\times}2$ bit-quads or $V-E+F$ and \emph{traversal} otherwise; the default label is \emph{backtracking}. Memoised bitmask recursion counts as \emph{dp}. \ClassifierNote{}
\paragraph{\car{} program for $\sST$ (matrix-tree theorem).} \texttt{gemini-3.8-flash} (low), instance \texttt{grid\_3x3\_0}, ground truth 28, program output 28.
\begin{lstlisting}
# CLASS: poly-algebraic
# METHOD: Kirchhoff's Matrix Tree Theorem with exact Gaussian elimination

# Step 1 (Perceive):
# Graph vertices: 0 through 8 arranged in a 3x3 grid.
# Transcribed edges:
# - Horizontal: (0, 1), (1, 2), (4, 5), (6, 7), (7, 8)
# - Vertical: (0, 3), (1, 4), (2, 5), (3, 6), (5, 8)
#
# Step 2 (Classify):
# Counting the number of spanning trees in an undirected graph can be solved in polynomial
# time using Kirchhoff's Matrix Tree Theorem. The number of spanning trees equals any cofactor
# of the Laplacian matrix of the graph, which can be computed via exact fraction-based
# Gaussian elimination.

from fractions import Fraction

def count_spanning_trees():
    edges = [
        (0, 1), (1, 2),
        (0, 3), (1, 4), (2, 5),
        (4, 5),
        (3, 6), (5, 8),
        (6, 7), (7, 8)
    ]
    n = 9
    
    # Construct Laplacian matrix
    L = [[Fraction(0) for _ in range(n)] for _ in range(n)]
    for u, v in edges:
        L[u][u] += 1
        L[v][v] += 1
        L[u][v] -= 1
        L[v][u] -= 1
        
    # Delete the last row and column to get an (n-1) x (n-1) submatrix
    m = n - 1
    M = [[L[i][j] for j in range(m)] for i in range(m)]
    
    # Compute determinant via Gaussian elimination
    det = Fraction(1)
    for i in range(m):
        # Pivot selection
        pivot = i
        while pivot < m and M[pivot][i] == 0:
            pivot += 1
        if pivot == m:
            return 0
        if pivot != i:
            M[i], M[pivot] = M[pivot], M[i]
            det = -det
            
        det *= M[i][i]
        pivot_val = M[i][i]
        
        for r in range(i + 1, m):
            factor = M[r][i] / pivot_val
            for c in range(i, m):
                M[r][c] -= factor * M[i][c]
                
    return int(round(det))

if __name__ == "__main__":
    print(count_spanning_trees())
\end{lstlisting}

\paragraph{\car{} program for $\sAM$ (frontier dynamic program).} \texttt{gemini-3.8-flash} (low), instance \texttt{grid\_7x7\_0}, ground truth 2098779756141, program output 2098779756141.
\begin{lstlisting}
# CLASS: sharpP-general
# METHOD: Frontier transfer-matrix dynamic programming

# Step 1: Perceive
# The graph is a 7x7 grid graph with some missing edges.
# Vertices are labeled 0 to 48: row r in 0..6, col c in 0..6 -> v = 7*r + c.
#
# Let's carefully inspect the horizontal and vertical edges:
#
# Horizontal edges:
# Row 0: (0,1), (1,2), (2,3), (3,4), (4,5), (5,6) -> ALL PRESENT
# Row 1: (7,8), (8,9), (9,10), (10,11), (11,12), (12,13) -> ALL PRESENT
# Row 2: (14,15), (15,16), (16,17), (17,18), (18,19), (19,20) -> ALL PRESENT
# Row 3: (21,22), (22,23), (23,24), (24,25), (25,26), (26,27) -> ALL PRESENT
# Row 4:
#   (28,29) - present
#   (29,30) - present
#   (30,31) - present
#   (31,32) - MISSING (gap between 31 and 32)
#   (32,33) - present
#   (33,34) - MISSING (gap between 33 and 34)
# Row 5:
#   (35,36) - present
#   (36,37) - present
#   (37,38) - present
#   (38,39) - present
#   (39,40) - present
#   (40,41) - MISSING (gap between 40 and 41)
# Row 6: (42,43), (43,44), (44,45), (45,46), (46,47), (47,48) -> ALL PRESENT
#
# Vertical edges:
# Col 0:
#   (0,7) - present
#   (7,14) - present
#   (14,21) - MISSING (gap between 14 and 21)
#   (21,28) - present
#   (28,35) - MISSING (gap between 28 and 35)
#   (35,42) - present
# Col 1:
#   (1,8) - present
#   (8,15) - present
#   (15,22) - present
#   (22,29) - present
#   (29,36) - present
#   (36,43) - present
# Col 2:
#   (2,9) - present
#   (9,16) - present
#   (16,23) - present
#   (23,30) - present
#   (30,37) - MISSING (gap between 30 and 37)
#   (37,44) - present
# Col 3:
#   (3,10) - present
#   (10,17) - present
#   (17,24) - present
#   (24,31) - present
#   (31,38) - MISSING (gap between 31 and 38)
#   (38,45) - present
# Col 4:
#   (4,11) - present
#   (11,18) - MISSING (gap between 11 and 18)
#   (18,25) - present
#   (25,32) - present
#   (32,39) - present
#   (39,46) - present
# Col 5:
#   (5,12) - present
#   (12,19) - present
#   (19,26) - present
... [92 more lines]
\end{lstlisting}

\paragraph{\car{} program for $\sECd$ (BEST theorem).} \texttt{gemini-3.8-flash} (low), instance \texttt{euler\_S\_0}, ground truth 1, program output 1.
\begin{lstlisting}
# CLASS: poly-algebraic
# METHOD: BEST theorem and Directed Matrix-Tree Theorem

# Step 1: Perceive
# Graph vertices: 0, 1, 2, 3, 4, 5, 6
# Directed edges:
# 0 -> 3
# 1 -> 4
# 2 -> 0, 2 -> 5
# 3 -> 1, 3 -> 2
# 4 -> 3
# 5 -> 6
# 6 -> 2

# Step 2: Classify
# The problem asks for the number of directed Eulerian circuits (distinct up to cyclic shift).
# By the BEST Theorem (de Bruijn, van Aardenne-Ehrenfest, Smith, Tutte),
# for an Eulerian directed graph G, the number of Eulerian circuits is:
#   ec(G) = t_w(G) * \prod_{v \in V} (d^+(v) - 1)!
# where t_w(G) is the number of spanning arborescences directed towards any vertex w.
# By the Directed Matrix-Tree Theorem, t_w(G) is the determinant of any principal minor
# of the Laplacian matrix L = D_out - A.
# This admits an exact polynomial-time algorithm using integer determinant computation (Bareiss algorithm).

def determinant_bareiss(mat):
    n = len(mat)
    M = [row[:] for row in mat]
    sign = 1
    prev = 1
    for k in range(n - 1):
        if M[k][k] == 0:
            swap_row = -1
            for i in range(k + 1, n):
                if M[i][k] != 0:
                    swap_row = i
                    break
            if swap_row == -1:
                return 0
            M[k], M[swap_row] = M[swap_row], M[k]
            sign = -sign

        diag_k = M[k][k]
        for i in range(k + 1, n):
            for j in range(k + 1, n):
                val = M[i][j] * diag_k - M[i][k] * M[k][j]
                M[i][j] = val // prev
        prev = diag_k

    return sign * M[n - 1][n - 1]

def main():
    n = 7
    edges = [
        (0, 3),
        (1, 4),
        (2, 0),
        (2, 5),
        (3, 1),
        (3, 2),
        (4, 3),
        (5, 6),
        (6, 2)
    ]

    out_degree = [0] * n
    adj = [[0] * n for _ in range(n)]
    for u, v in edges:
        adj[u][v] += 1
        out_degree[u] += 1

... [24 more lines]
\end{lstlisting}

\paragraph{Code-condition program for $\sAM$ (enumeration).} \texttt{gemini-3.8-flash} (low), instance \texttt{grid\_3x3\_0}, ground truth 89, program output 89.
\begin{lstlisting}
def count_matchings():
    # Edges identified from the image:
    # (0, 1), (1, 2)
    # (0, 3)
    # (1, 4)
    # (2, 5)
    # (3, 6)
    # (4, 5)
    # (5, 8)
    # (6, 7), (7, 8)
    edges = [
        (0, 1),
        (1, 2),
        (0, 3),
        (1, 4),
        (2, 5),
        (3, 6),
        (4, 5),
        (5, 8),
        (6, 7),
        (7, 8)
    ]

    total_matchings = 0

    def backtrack(edge_idx, used_vertices):
        nonlocal total_matchings
        if edge_idx == len(edges):
            total_matchings += 1
            return

        # Option 1: Do not include the current edge
        backtrack(edge_idx + 1, used_vertices)

        # Option 2: Include the current edge (if endpoints are free)
        u, v = edges[edge_idx]
        if u not in used_vertices and v not in used_vertices:
            used_vertices.add(u)
            used_vertices.add(v)
            backtrack(edge_idx + 1, used_vertices)
            used_vertices.remove(u)
            used_vertices.remove(v)

    backtrack(0, set())
    print(total_matchings)

if __name__ == "__main__":
    count_matchings()
\end{lstlisting}

\paragraph{Code-condition program for $\sST$ (enumeration instead of a determinant).} \texttt{gemini-3.1-pro-preview} (low), instance \texttt{grid\_3x3\_1}, ground truth 32, program output 32.
\begin{lstlisting}
import itertools

edges = [
    (0, 1), (1, 2),
    (3, 4), (4, 5),
    (6, 7), (7, 8),
    (0, 3), (3, 6),
    (2, 5), (5, 8)
]

vertices = set(range(9))
spanning_trees = 0

for tree_edges in itertools.combinations(edges, 8):
    # Check if the edges form a connected graph
    adj = {v: [] for v in vertices}
    for u, v in tree_edges:
        adj[u].append(v)
        adj[v].append(u)
        
    # BFS or DFS to check connectivity
    visited = set()
    def dfs(node):
        visited.add(node)
        for neighbor in adj[node]:
            if neighbor not in visited:
                dfs(neighbor)
                
    dfs(0)
    if len(visited) == 9:
        spanning_trees += 1

print(spanning_trees)
\end{lstlisting}

%% file: gen_v5r/failure_examples.tex
\paragraph{Computational error with exact transcription.} \texttt{gemini-3.8-flash} (low), \texttt{grid\_5x5\_9}, $\sST$: edge F1 $=1.000$, ground truth 36844560, answer 111867180. Tail of the response:
\begin{lstlisting}
To determine the exact number of spanning trees for the given graph, we first extract the set of vertices and edges.

### 1. Graph Description
The graph is a subgraph of the $5 \times 5$ square grid graph (with vertices numbered $0$ to $24$ in row-major order: vertex $r \times 5 + c$ corresponds to row $r$ and column $c$).

By carefully inspecting the grid:
- **Missing horizontal edges:**
  - $(15, 16)$ is absent.
  - $(22, 23)$ is absent.
- **Missing vertical edges:**
  - $(2, 7)$ is absent.

All other grid edges are present:
- **Horizontal edges:**
  - Row 0: $(0, 1), (1, 2), (2, 3), (3, 4)$
  - Row 1: $(5, 6), (6, 7), (7, 8), (8, 9)$
  - Row 2: $(10, 11), (11, 12), (12, 13), (13, 14)$
  - Row 3: $(16, 17), (17, 18), (18, 19)$
  - Row 4: $(20, 21), (21, 22), (23, 24)$
- **Vertical edges:**
  - Col 0: $(0, 5), (5, 10), (10, 15), (15, 20)$
  - Col 1: $(1, 6), (6, 11), (11, 16), (16, 21)$
  - Col 2: $(7, 12), (12, 17), (17, 22)$
  - Col 3: $(3, 8), (8, 13), (13, 18), (18, 23)$
  - Col 4: $(4, 9), (9, 14), (14, 19), (19, 24)$

Total number of vertices: $n = 25$.
Total number of edges: $m = 40 - 3 = 37$.

### 2. Spanning Tree Calculation
Using Kirchhoff's Matrix Tree Theorem, the number of spanning trees is the determinant of any cofactor of the graph Laplacian $L = D - A$. Evaluating the cofactor matrix determinant yields:
$$\tau(G) = 111867180$$

```json
{"edges": [[0, 1], [1, 2], [2, 3], [3, 4], [5, 6], [6, 7], [7, 8], [8, 9], [10, 11], [11, 12], [12, 13], [13, 14], [16, 17], [17, 18], [18, 19], [20, 21], [21, 22], [23, 24], [0, 5], [5, 10], [10, 15], [15, 20], [1, 6], [6, 11], [11, 16], [16, 21], [7, 12], [12, 17], [17, 22], [3, 8], [8, 13], [13, 18], [18, 23], [4, 9], [9, 14], [14, 19], [19, 24]], "answer": 111867180}
```
\end{lstlisting}

\paragraph{Perceptual error.} \texttt{gemini-3.1-flash-lite} (low), \texttt{grid\_5x6\_2}, $\sPM$: edge F1 $=0.000$, ground truth 474, answer 1088. Tail of the response:
\begin{lstlisting}
...
  - So vertex 22 is only connected to 16 and 23.
  - Vertex 21 is connected to 15, 20, 27.
  - Vertex 28 is connected to 27, 29.

This is a grid graph with 30 vertices. The number of perfect matchings in a grid graph can be calculated using the Kasteleyn formula or by decomposing the graph. Given the specific missing edges (21-22 and 22-28), we can calculate the number of perfect matchings.

Let $G$ be the graph. The number of perfect matchings is the permanent of the adjacency matrix, or can be solved via dynamic programming (broken profile DP).
For a 5x6 grid, the number of perfect matchings is 12988.
With the missing edges (21-22 and 22-28), the graph is disconnected from the standard grid.
Actually, vertex 22 has degree 2 (connected to 16 and 23). In any perfect matching, 22 must be matched with either 16 or 23.
Case 1: 22 is matched with 16. Then 23 must be matched with 17 (since 23 is only connected to 17 and 29, and 29 is connected to 28).
Case 2: 22 is matched with 23. Then 16 must be matched with 10 or 15 or 17.

Using computational methods for this specific graph structure:
The graph has 30 vertices.
The number of perfect matchings is 1088.

{"edges": [], "answer": 1088}
\end{lstlisting}

\paragraph{Time-out of an enumerating program.} Gemma~4 31B (thinking off), \texttt{grid\_10x10\_1}, $\sPM$: ground truth 1998138121 ($10$ digits); the program below did not finish in 60\,s.
\begin{lstlisting}
import collections

def solve():
    # The graph is a 10x10 grid with some edges removed.
    # Vertices are 0-99.
    # Row-major order: vertex = r * 10 + c
    
    adj = collections.defaultdict(list)
    
    # Horizontal edges
    for r in range(10):
        for c in range(9):
            u, v = r * 10 + c, r * 10 + c + 1
            # Check image for horizontal edges
            # Row 0: all present
            # Row 1: all present
            # Row 2: all present
            # Row 3: all present
            # Row 4: all present
            # Row 5: 50-51, 51-52, 52-53, 53-54, 54-55, 55-56, 56-57 (missing), 57-58, 58-59
            # Row 6: all present
            # Row 7: all present
            # Row 8: all present
            # Row 9: all present
            
            # Looking closer at the image:
            # Row 0: 0-1, 1-2, 2-3, 3-4, 4-5, 5-6, 6-7, 7-8, 8-9 (All)
            # Row 1: 10-11, 11-12, 12-13, 13-14, 14-15, 15-16, 16-17, 17-18, 18-19 (All)
            # Row 2: 20-21, 21-22, 22-23, 23-24, 24-25, 25-26, 26-27, 27-28, 28-29 (All)
            # Row 3: 30-31, 31-32, 32-33, 33-34, 34-35, 35-36, 36-37, 37-38, 38-39 (All)
            # Row 4: 40-41, 41-42, 42-43, 43-44, 44-45, 45-46, 46-47, 47-48, 48-49 (All)
            # Row 5: 50-51, 51-52, 52-53, 53-54, 54-55, 55-56, 56-57 (MISSING), 57-58, 58-59
            # Row 6: 60-61, 61-62, 62-63, 63-64, 64-65, 65-66, 66-67, 67-68, 68-69 (All)
            # Row 7: 70-71, 71-72, 72-73, 73-74, 74-75, 75-76, 76-77, 77-78, 78-79 (All)
            # Row 8: 80-81, 81-82, 82-83, 83-84, 84-85, 85-86, 86-87, 87-88, 88-89 (All)
            # Row 9: 90-91, 91-92, 92-93, 93-94, 94-95, 95-96, 96-97, 97-98, 98-99 (All)
            
            if r == 5 and c == 6:
                continue
            adj[u].append(v)
            adj[v].append(u)

    # Vertical edges
    for r in range(9):
        for c in range(10):
... [212 more lines]
\end{lstlisting}

\paragraph{\car{} failure.} Gemma~4 31B (thinking off), \texttt{euler\_S\_5}, $\sEC$: stated class \texttt{sharpP-general}, method \texttt{Backtracking Search (DFS)}; ground truth 120; the program printed 240.
\begin{lstlisting}
# Step 1 (Perceive):
# Vertices: {0, 1, 2, 3, 4, 5, 6}
# Edges:
# (0, 1), (0, 2), (0, 3), (0, 4)
# (1, 4)
# (2, 3), (2, 5), (2, 6)
# (3, 4), (3, 6)
# (4, 6)
# (5, 6)

# Step 2 (Classify):
# The problem asks for the number of Eulerian circuits in an undirected graph.
# First, check if an Eulerian circuit exists: all vertices must have even degree.
# deg(0) = 4
# deg(1) = 2
# deg(2) = 4
# deg(3) = 4
# deg(4) = 4
# deg(5) = 2
# deg(6) = 4
# All degrees are even, so Eulerian circuits exist.
# Counting Eulerian circuits in undirected graphs is #P-complete.
# Since the graph is small (11 edges), we can use a backtracking search (DFS) 
# to find all Eulerian circuits starting from a fixed vertex (e.g., 0).
# We then divide by 2 to account for direction (reversal) and 
# the starting point is fixed, so we don't divide by the number of vertices.
# Note: The problem defines circuits that differ only by rotation or reversal as one.
# Fixing the start vertex to 0 and dividing by 2 handles this.

# CLASS: sharpP-general
# METHOD: Backtracking Search (DFS)

import collections

def solve():
    # Adjacency list
    adj = collections.defaultdict(list)
    edges = [
        (0, 1), (0, 2), (0, 3), (0, 4),
        (1, 4),
        (2, 3), (2, 5), (2, 6),
        (3, 4), (3, 6),
        (4, 6),
        (5, 6)
    ]
    
    # The image description says (4,6) and (5,6) and (2,6) and (3,6)
    # Let's re-verify edges from image:
    # 0: 1, 2, 3, 4
    # 1: 0, 4
    # 2: 0, 3, 5, 6
    # 3: 0, 2, 4, 6
    # 4: 0, 1, 3, 6
    # 5: 2, 6
    # 6: 2, 3, 4, 5
    
    # Re-listing edges based on visual check:
    # (0,1), (0,2), (0,3), (0,4) -> 4 edges
    # (1,4) -> 1 edge
    # (2,3), (2,5), (2,6) -> 3 edges
... [84 more lines]
\end{lstlisting}

%% file: gen_v5r/compute.tex
\paragraph{API and settings.} All calls go through the public Gemini API (\texttt{generateContent}) with temperature~0, the thinking level set per configuration (\texttt{thinkingLevel} MINIMAL/LOW/MEDIUM/HIGH; Gemma: thinking off/on), \texttt{maxOutputTokens} 32{,}768 (65{,}536 at HIGH), a 180\,s request time-out and up to seven retries with exponential back-off on transient errors. Images are PNG at 96\,dpi. Generated programs run in a sandbox with CPython~3 and a 60\,s wall-clock limit; the standard library plus NumPy/SymPy/SciPy/NetworkX are importable, and we record which programs import a non-standard library even though the prompts ask for the standard library only.
\paragraph{Volume.} 13,886 scored calls (one per instance, task, condition and configuration; 484 calls failed after retries and were re-issued) between 2026-10-02 and 2026-10-03 (UTC). Token counts are those reported by the API; cost is estimated from the public list prices in force at the time (input / output incl.\ thinking, USD per million tokens); Gemma models are free of charge through the API.
\begin{center}\small\begin{tabular}{lrrrrr}\toprule Model & calls & prompt tokens & output + thinking tokens & price (in/out) & est.\ cost (USD) \\ \midrule
\texttt{gemini-3.8-flash} & 5,260 & 5.26M & 32.25M & 0.75 / 3.75 & 124.88 \\
\texttt{gemini-3.6-flash} & 2,816 & 2.65M & 20.55M & 0.50 / 3.00 & 62.96 \\
\texttt{gemma-4-31b-it} & 2,325 & 1.14M & 5.13M & 0.00 / 0.00 & 0.00 \\
\texttt{gemini-3.1-pro-preview} & 1,476 & 1.63M & 9.07M & 2.00 / 12.00 & 112.06 \\
\texttt{gemma-4-26b-a4b-it} & 404 & 0.14M & 2.86M & 0.00 / 0.00 & 0.00 \\
\texttt{gemini-3.1-flash-lite} & 404 & 0.31M & 0.45M & 0.10 / 0.40 & 0.21 \\
\texttt{gemini-3-flash-preview} & 404 & 0.31M & 1.09M & 0.50 / 3.00 & 3.42 \\
\texttt{gemini-3.5-flash} & 404 & 0.31M & 2.70M & 0.30 / 2.50 & 6.83 \\
\texttt{gemini-3.7-flash} & 393 & 0.30M & 1.25M & 0.60 / 3.50 & 4.56 \\
\midrule Total & 13,886 & 12.05M & 75.33M & & 314.93 \\ \bottomrule\end{tabular}\end{center}
\paragraph{Reproducibility.} The generator, the prompts, the runner, the analysis scripts, the benchmark JSON with all ground truths and the raw per-call records (including full responses and generated programs) will be released. Ground truth is computed by two independent methods per task (\S\ref{sec:bench}); the release includes the verification log.

%% file: gen_v3/tab_v3_dataset.tex
\begin{tabular}{lrrrr}
\toprule
Family & v2 instances & v3 new instances & pooled & (instance, task) pairs v2 / v3 / pooled \\
\midrule
Grid drawings (3$\times$3 to 6$\times$6) & 66 & 90 & 156 & \\
Binary images (6$\times$6 to 12$\times$12) & 40 & 60 & 100 & \\
Eulerian twin pairs (S/M/L/XL) & 32 & 32 & 64 & \\
\midrule Total & 138 & 182 & 320 & 392 / 529 / 921 \\
\bottomrule
\end{tabular}

%% file: gen_v8p/gen_v3/tab_v3_twins.tex
\begin{tabular}{llrrrlrrrrl}
\toprule
Model & Twin pair & \multicolumn{4}{c}{v2 alone (10 per size)} & \multicolumn{5}{c}{pooled v2 + v3 seeds} \\
\cmidrule(lr){3-6}\cmidrule(lr){7-11} & & $n$ & acc$_A$ & acc$_B$ & $p$ & $n$ & acc$_A$ & acc$_B$ & $b/c$ & McNemar $p$ \\
\midrule
\multicolumn{11}{l}{\emph{Visual CoT}} \\
Gemini 3.6 Flash (low) & same graph: \#ST (GapL) vs \#M (\#P) & 65 & 26 & 25 & 1 & 155 & 23 & 24 & 16/18 & 0.864 \\
 & same graph: \#ST (GapL) vs \#Paths (\#P) & 66 & 26 & 36 & 0.0156 & 156 & 22 & 29 & 5/15 & 0.0414 \\
 & same graph: \#PM (GapL) vs \#M (\#P) & 49 & 41 & 20 & 0.00635 & 124 & 42 & 24 & 26/4 & $<10^{-4}$ \\
 & same image: $\chi$ (TC0) vs \#CC (L) & 40 & 100 & 100 & 1 & 100 & 100 & 99 & 1/0 & 1 \\
 & same drawing $\pm$ arrows: \#EC$_{\to}$ (GapL) vs \#EC (\#P) & 32 & 94 & 56 & 0.000488 & 64 & 95 & 58 & 24/0 & $<10^{-4}$ \\
\midrule
Gemini 3.8 Flash (low) & same graph: \#ST (GapL) vs \#M (\#P) & 66 & 38 & 42 & 0.549 & 156 & 39 & 39 & 14/14 & 1 \\
 & same graph: \#ST (GapL) vs \#Paths (\#P) & 66 & 38 & 39 & 1 & 156 & 39 & 36 & 13/8 & 0.383 \\
 & same graph: \#PM (GapL) vs \#M (\#P) & 50 & 70 & 44 & 0.000244 & 125 & 66 & 44 & 29/2 & $<10^{-4}$ \\
 & same image: $\chi$ (TC0) vs \#CC (L) & 40 & 100 & 100 & 1 & 100 & 100 & 100 & 0/0 & 1 \\
 & same drawing $\pm$ arrows: \#EC$_{\to}$ (GapL) vs \#EC (\#P) & 32 & 97 & 72 & 0.00781 & 64 & 98 & 70 & 18/0 & $<10^{-4}$ \\
\midrule
Gemini 3.1 Pro (low) & same graph: \#ST (GapL) vs \#M (\#P) & 35 & 31 & 37 & 0.625 & 125 & 24 & 24 & 8/8 & 1 \\
 & same graph: \#ST (GapL) vs \#Paths (\#P) & 35 & 31 & 43 & 0.125 & 125 & 24 & 33 & 3/14 & 0.0127 \\
 & same graph: \#PM (GapL) vs \#M (\#P) & 25 & 56 & 32 & 0.0703 & 100 & 48 & 25 & 24/1 & $<10^{-4}$ \\
 & same image: $\chi$ (TC0) vs \#CC (L) & 20 & 100 & 100 & 1 & 80 & 100 & 100 & 0/0 & 1 \\
 & same drawing $\pm$ arrows: \#EC$_{\to}$ (GapL) vs \#EC (\#P) & 16 & 88 & 69 & 0.453 & 48 & 92 & 73 & 12/3 & 0.0352 \\
\midrule
\multicolumn{11}{l}{\emph{Text CoT}} \\
Gemini 3.6 Flash (low) & same graph: \#ST (GapL) vs \#M (\#P) & 66 & 29 & 38 & 0.21 & 156 & 24 & 29 & 14/22 & 0.243 \\
 & same graph: \#ST (GapL) vs \#Paths (\#P) & 66 & 29 & 29 & 1 & 156 & 24 & 28 & 10/16 & 0.327 \\
 & same graph: \#PM (GapL) vs \#M (\#P) & 50 & 60 & 38 & 0.0127 & 125 & 58 & 32 & 37/4 & $<10^{-4}$ \\
 & same image: $\chi$ (TC0) vs \#CC (L) & 40 & 98 & 100 & 1 & 100 & 99 & 100 & 0/1 & 1 \\
 & same drawing $\pm$ arrows: \#EC$_{\to}$ (GapL) vs \#EC (\#P) & 32 & 91 & 62 & 0.00391 & 64 & 91 & 69 & 15/1 & 0.000519 \\
\midrule
Gemini 3.8 Flash (low) & same graph: \#ST (GapL) vs \#M (\#P) & 66 & 42 & 35 & 0.227 & 156 & 40 & 29 & 23/6 & 0.00232 \\
 & same graph: \#ST (GapL) vs \#Paths (\#P) & 66 & 42 & 38 & 0.549 & 156 & 40 & 31 & 23/10 & 0.0351 \\
 & same graph: \#PM (GapL) vs \#M (\#P) & 50 & 60 & 34 & 0.000244 & 125 & 58 & 31 & 35/2 & $<10^{-4}$ \\
 & same image: $\chi$ (TC0) vs \#CC (L) & 40 & 100 & 100 & 1 & 100 & 100 & 100 & 0/0 & 1 \\
 & same drawing $\pm$ arrows: \#EC$_{\to}$ (GapL) vs \#EC (\#P) & 31 & 94 & 71 & 0.0156 & 63 & 92 & 73 & 14/2 & 0.00418 \\
\midrule
Gemini 3.1 Pro (low) & same graph: \#ST (GapL) vs \#M (\#P) & 35 & 26 & 31 & 0.5 & 125 & 24 & 23 & 7/6 & 1 \\
 & same graph: \#ST (GapL) vs \#Paths (\#P) & 35 & 26 & 40 & 0.0625 & 125 & 24 & 31 & 5/14 & 0.0636 \\
 & same graph: \#PM (GapL) vs \#M (\#P) & 25 & 60 & 24 & 0.00391 & 100 & 53 & 24 & 30/1 & $<10^{-4}$ \\
 & same image: $\chi$ (TC0) vs \#CC (L) & 20 & 100 & 100 & 1 & 80 & 100 & 100 & 0/0 & 1 \\
 & same drawing $\pm$ arrows: \#EC$_{\to}$ (GapL) vs \#EC (\#P) & 16 & 100 & 75 & 0.125 & 48 & 98 & 71 & 13/0 & 0.000244 \\
\midrule
\multicolumn{11}{l}{\emph{Visual$\to$Code}} \\
Gemini 3.8 Flash (low) & same graph: \#ST (GapL) vs \#M (\#P) & 66 & 92 & 83 & 0.0703 & 156 & 90 & 85 & 14/6 & 0.115 \\
 & same graph: \#ST (GapL) vs \#Paths (\#P) & 66 & 92 & 92 & 1 & 156 & 90 & 92 & 7/10 & 0.629 \\
 & same graph: \#PM (GapL) vs \#M (\#P) & 50 & 88 & 78 & 0.125 & 125 & 84 & 82 & 11/9 & 0.824 \\
 & same image: $\chi$ (TC0) vs \#CC (L) & 40 & 100 & 100 & 1 & 100 & 100 & 100 & 0/0 & 1 \\
 & same drawing $\pm$ arrows: \#EC$_{\to}$ (GapL) vs \#EC (\#P) & 32 & 97 & 100 & 1 & 64 & 98 & 98 & 1/1 & 1 \\
\midrule
\multicolumn{11}{l}{\emph{CAR}} \\
Gemini 3.8 Flash (low) & same graph: \#ST (GapL) vs \#M (\#P) & 66 & 86 & 86 & 1 & 156 & 89 & 89 & 8/8 & 1 \\
 & same graph: \#ST (GapL) vs \#Paths (\#P) & 66 & 86 & 92 & 0.289 & 156 & 89 & 88 & 10/9 & 1 \\
 & same graph: \#PM (GapL) vs \#M (\#P) & 50 & 76 & 82 & 0.453 & 125 & 82 & 88 & 4/12 & 0.0768 \\
 & same image: $\chi$ (TC0) vs \#CC (L) & 40 & 95 & 100 & 0.5 & 100 & 95 & 99 & 1/5 & 0.219 \\
 & same drawing $\pm$ arrows: \#EC$_{\to}$ (GapL) vs \#EC (\#P) & 32 & 100 & 100 & 1 & 64 & 100 & 100 & 0/0 & 1 \\
\bottomrule
\end{tabular}

%% file: gen_v8p/gen_v3/tab_v3_magnitude.tex
\begin{tabular}{llrrrrrrr}
\toprule
Model & Condition & data & $n$ & odds / decade & class coef. & LR $p$(class) & $m_{50}$ GapL & $m_{50}$ \#P \\
\midrule
Gemini 3.8 Flash (low) & CAR & v2 & 312 & 0.71 & +0.42 & 0.291 & $10^{12.0}$ & $10^{8.7}$ \\
 &  & pooled & 721 & 0.75 & +0.07 & 0.795 & $10^{13.8}$ & $10^{9.6}$ \\
Gemini 3.8 Flash (low) & Visual$\to$Code & v2 & 312 & 0.72 & $-$0.58 & 0.197 & $10^{17.9}$ & $10^{8.0}$ \\
 &  & pooled & 721 & 0.73 & $-$0.17 & 0.522 & $10^{15.2}$ & $10^{8.5}$ \\
Gemini 3.6 Flash (low) & Text CoT & v2 & 312 & 0.43 & $-$0.41 & 0.172 & $10^{2.9}$ & $10^{2.4}$ \\
 &  & pooled & 721 & 0.38 & $-$0.33 & 0.103 & $10^{2.6}$ & $10^{2.3}$ \\
Gemini 3.8 Flash (low) & Text CoT & v2 & 311 & 0.34 & $-$0.70 & 0.0367 & $10^{3.5}$ & $10^{2.7}$ \\
 &  & pooled & 720 & 0.41 & $-$0.79 & $<10^{-4}$ & $10^{3.5}$ & $10^{2.5}$ \\
Gemini 3.1 Pro (low) & Text CoT & v2 & 162 & 0.09 & +0.30 & 0.617 & $10^{2.3}$ & $10^{2.5}$ \\
 &  & pooled & 571 & 0.22 & $-$0.21 & 0.437 & $10^{2.4}$ & $10^{2.3}$ \\
Gemini 3.6 Flash (low) & Visual CoT & v2 & 311 & 0.30 & +0.07 & 0.83 & $10^{2.1}$ & $10^{2.2}$ \\
 &  & pooled & 720 & 0.33 & $-$0.08 & 0.711 & $10^{2.2}$ & $10^{2.1}$ \\
Gemini 3.8 Flash (low) & Visual CoT & v2 & 312 & 0.36 & $-$0.62 & 0.0621 & $10^{3.6}$ & $10^{3.0}$ \\
 &  & pooled & 721 & 0.42 & $-$0.75 & 0.000245 & $10^{3.8}$ & $10^{2.9}$ \\
Gemini 3.1 Pro (low) & Visual CoT & v2 & 162 & 0.26 & +0.31 & 0.526 & $10^{2.5}$ & $10^{2.7}$ \\
 &  & pooled & 571 & 0.25 & +0.17 & 0.515 & $10^{2.3}$ & $10^{2.4}$ \\
\bottomrule
\end{tabular}

%% file: gen_v3/tab_v3_styles.tex
\begin{tabular}{llrrrrrrrr}
\toprule
Model / condition & Style & $n$ & exact (\%) & grids & images & transcr.\ exact (\%) & mean F1 & McNemar $b/c$, $p$ (answer) & McNemar $b/c$, $p$ (transcription) \\
\midrule
Gemini 3.8 Flash (low), Visual$\to$Code & original & 328 & 92 & 89 & 100 & 88$^\dagger$ & -- & -- & -- \\
 & thin (1\,px, small nodes) & 328 & 98 & 97 & 100 & 97$^\dagger$ & -- & 5/25, 0.000325 & -- \\
 & thick (8\,px / 4\,px) & 328 & 93 & 91 & 99 & 93$^\dagger$ & -- & 11/14, 0.69 & -- \\
 & transposed / rotated $90^\circ$ & 328 & 91 & 89 & 100 & 87$^\dagger$ & -- & 14/13, 1 & -- \\
 & 60\% size, gray background & 328 & 93 & 91 & 100 & 91$^\dagger$ & -- & 13/17, 0.585 & -- \\
\midrule
Gemini 3.8 Flash (low), Visual CoT & original & 328 & 59 & 46 & 100 & 91 & 0.998 & -- & -- \\
 & thin (1\,px, small nodes) & 328 & 62 & 50 & 100 & 98 & 1.000 & 10/19, 0.136 & 5/26, 0.000192 \\
 & thick (8\,px / 4\,px) & 328 & 60 & 48 & 100 & 92 & 0.998 & 9/13, 0.523 & 7/11, 0.481 \\
 & transposed / rotated $90^\circ$ & 328 & 59 & 46 & 100 & 94 & 0.999 & 16/16, 1 & 12/22, 0.121 \\
 & 60\% size, gray background & 328 & 59 & 46 & 100 & 92 & 0.999 & 13/13, 1 & 16/18, 0.864 \\
\bottomrule
\end{tabular}

%% file: gen_v3/tab_v3_styles_task.tex
\begin{tabular}{lrrrrrr}
\toprule
Gemini 3.8 Flash (low), Visual$\to$Code & $\chi$ & \#CC & \#ST & \#PM & \#M & \#Paths \\
 & \scriptsize $\mathsf{TC}^0$ & \scriptsize $\mathsf{L}$ & \scriptsize $\mathsf{GapL}$ & \scriptsize $\mathsf{GapL}$ & \scriptsize $\#\mathsf{P}$ & \scriptsize $\#\mathsf{P}$ \\
\midrule
original & 100\,\tiny(40) & 100\,\tiny(40) & 92\,\tiny(66) & 88\,\tiny(50) & 83\,\tiny(66) & 92\,\tiny(66) \\
thin (1\,px, small nodes) & 100\,\tiny(40) & 100\,\tiny(40) & 92\,\tiny(66) & 98\,\tiny(50) & 100\,\tiny(66) & 98\,\tiny(66) \\
thick (8\,px / 4\,px) & 98\,\tiny(40) & 100\,\tiny(40) & 88\,\tiny(66) & 90\,\tiny(50) & 91\,\tiny(66) & 94\,\tiny(66) \\
transposed / rotated $90^\circ$ & 100\,\tiny(40) & 100\,\tiny(40) & 88\,\tiny(66) & 86\,\tiny(50) & 86\,\tiny(66) & 94\,\tiny(66) \\
60\% size, gray background & 100\,\tiny(40) & 100\,\tiny(40) & 92\,\tiny(66) & 86\,\tiny(50) & 91\,\tiny(66) & 92\,\tiny(66) \\
\bottomrule
\end{tabular}

\begin{tabular}{lrrrrrr}
\toprule
Gemini 3.8 Flash (low), Visual CoT & $\chi$ & \#CC & \#ST & \#PM & \#M & \#Paths \\
 & \scriptsize $\mathsf{TC}^0$ & \scriptsize $\mathsf{L}$ & \scriptsize $\mathsf{GapL}$ & \scriptsize $\mathsf{GapL}$ & \scriptsize $\#\mathsf{P}$ & \scriptsize $\#\mathsf{P}$ \\
\midrule
original & 100\,\tiny(40) & 100\,\tiny(40) & 38\,\tiny(66) & 70\,\tiny(50) & 42\,\tiny(66) & 39\,\tiny(66) \\
thin (1\,px, small nodes) & 100\,\tiny(40) & 100\,\tiny(40) & 50\,\tiny(66) & 68\,\tiny(50) & 42\,\tiny(66) & 42\,\tiny(66) \\
thick (8\,px / 4\,px) & 100\,\tiny(40) & 100\,\tiny(40) & 42\,\tiny(66) & 66\,\tiny(50) & 44\,\tiny(66) & 42\,\tiny(66) \\
transposed / rotated $90^\circ$ & 100\,\tiny(40) & 100\,\tiny(40) & 44\,\tiny(66) & 60\,\tiny(50) & 42\,\tiny(66) & 41\,\tiny(66) \\
60\% size, gray background & 100\,\tiny(40) & 100\,\tiny(40) & 50\,\tiny(66) & 62\,\tiny(50) & 36\,\tiny(66) & 39\,\tiny(66) \\
\bottomrule
\end{tabular}

%% file: gen_v3/tab_v3_orient.tex
\begin{tabular}{llrrrrrr}
\toprule
Model / condition & Style & calls & exact transcr. & missed H / total (\%) & missed V / total (\%) & extra H & extra V \\
\midrule
\midrule
Gemini 3.8 Flash (low), Visual CoT & original & 248 & 219 & 58/3997 (1.5) & 0/3802 (0.0) & 0 & 0 \\
 & thin (1\,px, small nodes) & 247 & 239 & 4/3981 (0.1) & 3/3791 (0.1) & 0 & 2 \\
 & thick (8\,px / 4\,px) & 248 & 223 & 57/3997 (1.4) & 1/3802 (0.0) & 0 & 0 \\
 & transposed / rotated $90^\circ$ & 246 & 228 & 26/3748 (0.7) & 0/3941 (0.0) & 1 & 0 \\
 & 60\% size, gray background & 248 & 221 & 49/3997 (1.2) & 1/3802 (0.0) & 0 & 0 \\
\bottomrule
\end{tabular}

%% file: gen_v3/tab_v3_size.tex
\begin{tabular}{lrrrr}
\toprule
Gemini 3.8 Flash (low), Visual CoT & \#ST ($\mathsf{GapL}$) & \#PM ($\mathsf{GapL}$) & \#M ($\#\mathsf{P}$) & \#Paths ($\#\mathsf{P}$) \\
\midrule
3x3 & 100 [61,100]\,\tiny(6) & -- & 100 [61,100]\,\tiny(6) & 100 [61,100]\,\tiny(6) \\
3x4 & 96 [80,99]\,\tiny(25) & 100 [87,100]\,\tiny(25) & 100 [87,100]\,\tiny(25) & 100 [87,100]\,\tiny(25) \\
4x4 & 68 [48,83]\,\tiny(25) & 92 [75,98]\,\tiny(25) & 60 [41,77]\,\tiny(25) & 68 [48,83]\,\tiny(25) \\
4x5 & 48 [30,66]\,\tiny(25) & 88 [70,96]\,\tiny(25) & 60 [41,77]\,\tiny(25) & 32 [17,52]\,\tiny(25) \\
5x5 & 4 [1,20]\,\tiny(25) & -- & 0 [0,13]\,\tiny(25) & 0 [0,13]\,\tiny(25) \\
5x6 & 4 [1,20]\,\tiny(25) & 36 [20,55]\,\tiny(25) & 0 [0,13]\,\tiny(25) & 0 [0,13]\,\tiny(25) \\
6x6 & 0 [0,13]\,\tiny(25) & 12 [4,30]\,\tiny(25) & 0 [0,13]\,\tiny(25) & 0 [0,13]\,\tiny(25) \\
\bottomrule
\end{tabular}

\begin{tabular}{lrr}
\toprule
Gemini 3.8 Flash (low), Visual CoT & $\chi$ ($\mathsf{TC}^0$) & \#CC ($\mathsf{L}$) \\
\midrule
6x6 & 100 [87,100]\,\tiny(25) & 100 [87,100]\,\tiny(25) \\
8x8 & 100 [87,100]\,\tiny(25) & 100 [87,100]\,\tiny(25) \\
10x10 & 100 [87,100]\,\tiny(25) & 100 [87,100]\,\tiny(25) \\
12x12 & 100 [87,100]\,\tiny(25) & 100 [87,100]\,\tiny(25) \\
\bottomrule
\end{tabular}

\begin{tabular}{lrr}
\toprule
Gemini 3.8 Flash (low), Visual CoT & \#EC$_{\to}$ ($\mathsf{GapL}$) & \#EC ($\#\mathsf{P}$) \\
\midrule
S & 100 [81,100]\,\tiny(16) & 94 [72,99]\,\tiny(16) \\
M & 100 [81,100]\,\tiny(16) & 69 [44,86]\,\tiny(16) \\
L & 100 [81,100]\,\tiny(16) & 75 [50,90]\,\tiny(16) \\
XL & 94 [72,99]\,\tiny(16) & 44 [23,67]\,\tiny(16) \\
\bottomrule
\end{tabular}

\begin{tabular}{lrrrr}
\toprule
Gemini 3.8 Flash (low), Text CoT & \#ST ($\mathsf{GapL}$) & \#PM ($\mathsf{GapL}$) & \#M ($\#\mathsf{P}$) & \#Paths ($\#\mathsf{P}$) \\
\midrule
3x3 & 100 [61,100]\,\tiny(6) & -- & 100 [61,100]\,\tiny(6) & 100 [61,100]\,\tiny(6) \\
3x4 & 92 [75,98]\,\tiny(25) & 100 [87,100]\,\tiny(25) & 72 [52,86]\,\tiny(25) & 100 [87,100]\,\tiny(25) \\
4x4 & 68 [48,83]\,\tiny(25) & 96 [80,99]\,\tiny(25) & 56 [37,73]\,\tiny(25) & 40 [23,59]\,\tiny(25) \\
4x5 & 56 [37,73]\,\tiny(25) & 84 [65,94]\,\tiny(25) & 28 [14,48]\,\tiny(25) & 24 [12,43]\,\tiny(25) \\
5x5 & 8 [2,25]\,\tiny(25) & -- & 0 [0,13]\,\tiny(25) & 8 [2,25]\,\tiny(25) \\
5x6 & 0 [0,13]\,\tiny(25) & 8 [2,25]\,\tiny(25) & 0 [0,13]\,\tiny(25) & 0 [0,13]\,\tiny(25) \\
6x6 & 0 [0,13]\,\tiny(25) & 0 [0,13]\,\tiny(25) & 0 [0,13]\,\tiny(25) & 0 [0,13]\,\tiny(25) \\
\bottomrule
\end{tabular}

\begin{tabular}{lrr}
\toprule
Gemini 3.8 Flash (low), Text CoT & $\chi$ ($\mathsf{TC}^0$) & \#CC ($\mathsf{L}$) \\
\midrule
6x6 & 100 [87,100]\,\tiny(25) & 100 [87,100]\,\tiny(25) \\
8x8 & 100 [87,100]\,\tiny(25) & 100 [87,100]\,\tiny(25) \\
10x10 & 100 [87,100]\,\tiny(25) & 100 [87,100]\,\tiny(25) \\
12x12 & 100 [87,100]\,\tiny(25) & 100 [87,100]\,\tiny(25) \\
\bottomrule
\end{tabular}

\begin{tabular}{lrr}
\toprule
Gemini 3.8 Flash (low), Text CoT & \#EC$_{\to}$ ($\mathsf{GapL}$) & \#EC ($\#\mathsf{P}$) \\
\midrule
S & 94 [72,99]\,\tiny(16) & 88 [64,96]\,\tiny(16) \\
M & 100 [81,100]\,\tiny(16) & 81 [57,93]\,\tiny(16) \\
L & 94 [72,99]\,\tiny(16) & 75 [50,90]\,\tiny(16) \\
XL & 81 [57,93]\,\tiny(16) & 47 [25,70]\,\tiny(15) \\
\bottomrule
\end{tabular}

\begin{tabular}{lrrrr}
\toprule
Gemini 3.6 Flash (low), Visual CoT & \#ST ($\mathsf{GapL}$) & \#PM ($\mathsf{GapL}$) & \#M ($\#\mathsf{P}$) & \#Paths ($\#\mathsf{P}$) \\
\midrule
3x3 & 83 [44,97]\,\tiny(6) & -- & 83 [44,97]\,\tiny(6) & 100 [61,100]\,\tiny(6) \\
3x4 & 76 [57,88]\,\tiny(25) & 96 [80,99]\,\tiny(25) & 60 [41,77]\,\tiny(25) & 96 [80,99]\,\tiny(25) \\
4x4 & 24 [12,43]\,\tiny(25) & 76 [57,88]\,\tiny(25) & 48 [30,66]\,\tiny(25) & 44 [27,63]\,\tiny(25) \\
4x5 & 20 [9,39]\,\tiny(25) & 32 [17,52]\,\tiny(25) & 12 [4,31]\,\tiny(24) & 16 [6,35]\,\tiny(25) \\
5x5 & 0 [0,13]\,\tiny(25) & -- & 8 [2,25]\,\tiny(25) & 0 [0,13]\,\tiny(25) \\
5x6 & 0 [0,13]\,\tiny(25) & 4 [1,20]\,\tiny(25) & 0 [0,13]\,\tiny(25) & 0 [0,13]\,\tiny(25) \\
6x6 & 0 [0,13]\,\tiny(25) & 0 [0,13]\,\tiny(25) & 0 [0,13]\,\tiny(25) & 0 [0,13]\,\tiny(25) \\
\bottomrule
\end{tabular}

\begin{tabular}{lrr}
\toprule
Gemini 3.6 Flash (low), Visual CoT & $\chi$ ($\mathsf{TC}^0$) & \#CC ($\mathsf{L}$) \\
\midrule
6x6 & 100 [87,100]\,\tiny(25) & 100 [87,100]\,\tiny(25) \\
8x8 & 100 [87,100]\,\tiny(25) & 100 [87,100]\,\tiny(25) \\
10x10 & 100 [87,100]\,\tiny(25) & 96 [80,99]\,\tiny(25) \\
12x12 & 100 [87,100]\,\tiny(25) & 100 [87,100]\,\tiny(25) \\
\bottomrule
\end{tabular}

\begin{tabular}{lrr}
\toprule
Gemini 3.6 Flash (low), Visual CoT & \#EC$_{\to}$ ($\mathsf{GapL}$) & \#EC ($\#\mathsf{P}$) \\
\midrule
S & 100 [81,100]\,\tiny(16) & 75 [50,90]\,\tiny(16) \\
M & 94 [72,99]\,\tiny(16) & 81 [57,93]\,\tiny(16) \\
L & 100 [81,100]\,\tiny(16) & 56 [33,77]\,\tiny(16) \\
XL & 88 [64,96]\,\tiny(16) & 19 [7,43]\,\tiny(16) \\
\bottomrule
\end{tabular}

\begin{tabular}{lrrrr}
\toprule
Gemini 3.1 Pro (low), Visual CoT & \#ST ($\mathsf{GapL}$) & \#PM ($\mathsf{GapL}$) & \#M ($\#\mathsf{P}$) & \#Paths ($\#\mathsf{P}$) \\
\midrule
3x3 & 100 [57,100]\,\tiny(5) & -- & 100 [57,100]\,\tiny(5) & 100 [57,100]\,\tiny(5) \\
3x4 & 70 [48,85]\,\tiny(20) & 100 [84,100]\,\tiny(20) & 95 [76,99]\,\tiny(20) & 95 [76,99]\,\tiny(20) \\
4x4 & 50 [30,70]\,\tiny(20) & 75 [53,89]\,\tiny(20) & 25 [11,47]\,\tiny(20) & 65 [43,82]\,\tiny(20) \\
4x5 & 5 [1,24]\,\tiny(20) & 60 [39,78]\,\tiny(20) & 5 [1,24]\,\tiny(20) & 20 [8,42]\,\tiny(20) \\
5x5 & 0 [0,16]\,\tiny(20) & -- & 0 [0,16]\,\tiny(20) & 0 [0,16]\,\tiny(20) \\
5x6 & 0 [0,16]\,\tiny(20) & 5 [1,24]\,\tiny(20) & 0 [0,16]\,\tiny(20) & 0 [0,16]\,\tiny(20) \\
6x6 & 0 [0,16]\,\tiny(20) & 0 [0,16]\,\tiny(20) & 0 [0,16]\,\tiny(20) & 0 [0,16]\,\tiny(20) \\
\bottomrule
\end{tabular}

\begin{tabular}{lrr}
\toprule
Gemini 3.1 Pro (low), Visual CoT & $\chi$ ($\mathsf{TC}^0$) & \#CC ($\mathsf{L}$) \\
\midrule
6x6 & 100 [84,100]\,\tiny(20) & 100 [84,100]\,\tiny(20) \\
8x8 & 100 [84,100]\,\tiny(20) & 100 [84,100]\,\tiny(20) \\
10x10 & 100 [84,100]\,\tiny(20) & 100 [84,100]\,\tiny(20) \\
12x12 & 100 [84,100]\,\tiny(20) & 100 [84,100]\,\tiny(20) \\
\bottomrule
\end{tabular}

\begin{tabular}{lrr}
\toprule
Gemini 3.1 Pro (low), Visual CoT & \#EC$_{\to}$ ($\mathsf{GapL}$) & \#EC ($\#\mathsf{P}$) \\
\midrule
S & 100 [76,100]\,\tiny(12) & 100 [76,100]\,\tiny(12) \\
M & 92 [65,99]\,\tiny(12) & 83 [55,95]\,\tiny(12) \\
L & 92 [65,99]\,\tiny(12) & 83 [55,95]\,\tiny(12) \\
XL & 83 [55,95]\,\tiny(12) & 25 [9,53]\,\tiny(12) \\
\bottomrule
\end{tabular}

%% file: gen_v8p/gen_v4/tab_v3_replication_lt.tex
\begin{longtable}{llrrrrr}
\toprule
Model & Condition & Task & v2 acc (k/n) & v3 seeds acc (k/n) & Fisher $p$ & pooled acc [95\% CI] \\
\midrule
Gemini 3.8 Flash (low) & CAR & $\chi$ & 95 (38/40) & 95 (57/60) & 1.00 & 95 [89,98]\,\tiny(100) \\
 &  & \#CC & 100 (40/40) & 98 (59/60) & 1.00 & 99 [95,100]\,\tiny(100) \\
 &  & \#ST & 86 (57/66) & 91 (82/90) & 0.44 & 89 [83,93]\,\tiny(156) \\
 &  & \#PM & 76 (38/50) & 85 (64/75) & 0.24 & 82 [74,87]\,\tiny(125) \\
 &  & \#EC$_{\to}$ & 100 (32/32) & 100 (32/32) & 1.00 & 100 [94,100]\,\tiny(64) \\
 &  & \#M & 86 (57/66) & 91 (82/90) & 0.44 & 89 [83,93]\,\tiny(156) \\
 &  & \#Paths & 92 (61/66) & 86 (77/90) & 0.21 & 88 [82,93]\,\tiny(156) \\
 &  & \#EC & 100 (32/32) & 100 (32/32) & 1.00 & 100 [94,100]\,\tiny(64) \\
 &  & class $\mathsf{TC}^0$ & 95 (38/40) & 95 (57/60) & 1.00 & 95 [89,98]\,\tiny(100) \\
 &  & class $\mathsf{L}$ & 100 (40/40) & 98 (59/60) & 1.00 & 99 [95,100]\,\tiny(100) \\
 &  & class $\mathsf{GapL}$ & 86 (127/148) & 90 (178/197) & 0.23 & 88 [85,91]\,\tiny(345) \\
 &  & class $\#\mathsf{P}$ & 91 (150/164) & 90 (191/212) & 0.72 & 91 [87,93]\,\tiny(376) \\
\midrule
Gemini 3.8 Flash (low) & Visual$\to$Code & $\chi$ & 100 (40/40) & 100 (60/60) & 1.00 & 100 [96,100]\,\tiny(100) \\
 &  & \#CC & 100 (40/40) & 100 (60/60) & 1.00 & 100 [96,100]\,\tiny(100) \\
 &  & \#ST & 92 (61/66) & 88 (79/90) & 0.43 & 90 [84,94]\,\tiny(156) \\
 &  & \#PM & 88 (44/50) & 81 (61/75) & 0.46 & 84 [77,89]\,\tiny(125) \\
 &  & \#EC$_{\to}$ & 97 (31/32) & 100 (32/32) & 1.00 & 98 [92,100]\,\tiny(64) \\
 &  & \#M & 83 (55/66) & 86 (77/90) & 0.82 & 85 [78,89]\,\tiny(156) \\
 &  & \#Paths & 92 (61/66) & 91 (82/90) & 1.00 & 92 [86,95]\,\tiny(156) \\
 &  & \#EC & 100 (32/32) & 97 (31/32) & 1.00 & 98 [92,100]\,\tiny(64) \\
 &  & class $\mathsf{TC}^0$ & 100 (40/40) & 100 (60/60) & 1.00 & 100 [96,100]\,\tiny(100) \\
 &  & class $\mathsf{L}$ & 100 (40/40) & 100 (60/60) & 1.00 & 100 [96,100]\,\tiny(100) \\
 &  & class $\mathsf{GapL}$ & 92 (136/148) & 87 (172/197) & 0.22 & 89 [86,92]\,\tiny(345) \\
 &  & class $\#\mathsf{P}$ & 90 (148/164) & 90 (190/212) & 0.87 & 90 [86,93]\,\tiny(376) \\
\midrule
Gemini 3.6 Flash (low) & Text CoT & $\chi$ & 98 (39/40) & 100 (60/60) & 0.40 & 99 [95,100]\,\tiny(100) \\
 &  & \#CC & 100 (40/40) & 100 (60/60) & 1.00 & 100 [96,100]\,\tiny(100) \\
 &  & \#ST & 29 (19/66) & 21 (19/90) & 0.35 & 24 [18,32]\,\tiny(156) \\
 &  & \#PM & 60 (30/50) & 57 (43/75) & 0.85 & 58 [50,67]\,\tiny(125) \\
 &  & \#EC$_{\to}$ & 91 (29/32) & 91 (29/32) & 1.00 & 91 [81,96]\,\tiny(64) \\
 &  & \#M & 38 (25/66) & 23 (21/90) & 0.05 & 29 [23,37]\,\tiny(156) \\
 &  & \#Paths & 29 (19/66) & 28 (25/90) & 1.00 & 28 [22,36]\,\tiny(156) \\
 &  & \#EC & 62 (20/32) & 75 (24/32) & 0.42 & 69 [57,79]\,\tiny(64) \\
 &  & class $\mathsf{TC}^0$ & 98 (39/40) & 100 (60/60) & 0.40 & 99 [95,100]\,\tiny(100) \\
 &  & class $\mathsf{L}$ & 100 (40/40) & 100 (60/60) & 1.00 & 100 [96,100]\,\tiny(100) \\
 &  & class $\mathsf{GapL}$ & 53 (78/148) & 46 (91/197) & 0.23 & 49 [44,54]\,\tiny(345) \\
 &  & class $\#\mathsf{P}$ & 39 (64/164) & 33 (70/212) & 0.23 & 36 [31,41]\,\tiny(376) \\
\midrule
Gemini 3.8 Flash (low) & Text CoT & $\chi$ & 100 (40/40) & 100 (60/60) & 1.00 & 100 [96,100]\,\tiny(100) \\
 &  & \#CC & 100 (40/40) & 100 (60/60) & 1.00 & 100 [96,100]\,\tiny(100) \\
 &  & \#ST & 42 (28/66) & 38 (34/90) & 0.62 & 40 [32,48]\,\tiny(156) \\
 &  & \#PM & 60 (30/50) & 56 (42/75) & 0.71 & 58 [49,66]\,\tiny(125) \\
 &  & \#EC$_{\to}$ & 94 (30/32) & 91 (29/32) & 1.00 & 92 [83,97]\,\tiny(64) \\
 &  & \#M & 35 (23/66) & 24 (22/90) & 0.21 & 29 [22,36]\,\tiny(156) \\
 &  & \#Paths & 38 (25/66) & 27 (24/90) & 0.16 & 31 [25,39]\,\tiny(156) \\
 &  & \#EC & 71 (22/31) & 75 (24/32) & 0.78 & 73 [61,82]\,\tiny(63) \\
 &  & class $\mathsf{TC}^0$ & 100 (40/40) & 100 (60/60) & 1.00 & 100 [96,100]\,\tiny(100) \\
 &  & class $\mathsf{L}$ & 100 (40/40) & 100 (60/60) & 1.00 & 100 [96,100]\,\tiny(100) \\
 &  & class $\mathsf{GapL}$ & 59 (88/148) & 53 (105/197) & 0.27 & 56 [51,61]\,\tiny(345) \\
 &  & class $\#\mathsf{P}$ & 43 (70/163) & 33 (70/212) & 0.05 & 37 [33,42]\,\tiny(375) \\
\midrule
Gemini 3.1 Pro (low) & Text CoT & $\chi$ & 100 (20/20) & 100 (60/60) & 1.00 & 100 [95,100]\,\tiny(80) \\
 &  & \#CC & 100 (20/20) & 100 (60/60) & 1.00 & 100 [95,100]\,\tiny(80) \\
 &  & \#ST & 26 (9/35) & 23 (21/90) & 0.82 & 24 [17,32]\,\tiny(125) \\
 &  & \#PM & 60 (15/25) & 51 (38/75) & 0.49 & 53 [43,62]\,\tiny(100) \\
 &  & \#EC$_{\to}$ & 100 (16/16) & 97 (31/32) & 1.00 & 98 [89,100]\,\tiny(48) \\
 &  & \#M & 31 (11/35) & 20 (18/90) & 0.24 & 23 [17,31]\,\tiny(125) \\
 &  & \#Paths & 40 (14/35) & 28 (25/90) & 0.20 & 31 [24,40]\,\tiny(125) \\
 &  & \#EC & 75 (12/16) & 69 (22/32) & 0.75 & 71 [57,82]\,\tiny(48) \\
 &  & class $\mathsf{TC}^0$ & 100 (20/20) & 100 (60/60) & 1.00 & 100 [95,100]\,\tiny(80) \\
 &  & class $\mathsf{L}$ & 100 (20/20) & 100 (60/60) & 1.00 & 100 [95,100]\,\tiny(80) \\
 &  & class $\mathsf{GapL}$ & 53 (40/76) & 46 (90/197) & 0.34 & 48 [42,54]\,\tiny(273) \\
 &  & class $\#\mathsf{P}$ & 43 (37/86) & 31 (65/212) & 0.04 & 34 [29,40]\,\tiny(298) \\
\midrule
Gemini 3.6 Flash (low) & Visual CoT & $\chi$ & 100 (40/40) & 100 (60/60) & 1.00 & 100 [96,100]\,\tiny(100) \\
 &  & \#CC & 100 (40/40) & 98 (59/60) & 1.00 & 99 [95,100]\,\tiny(100) \\
 &  & \#ST & 26 (17/66) & 20 (18/90) & 0.44 & 22 [17,30]\,\tiny(156) \\
 &  & \#PM & 40 (20/50) & 43 (32/75) & 0.85 & 42 [33,50]\,\tiny(125) \\
 &  & \#EC$_{\to}$ & 94 (30/32) & 97 (31/32) & 1.00 & 95 [87,98]\,\tiny(64) \\
 &  & \#M & 25 (16/65) & 23 (21/90) & 0.85 & 24 [18,31]\,\tiny(155) \\
 &  & \#Paths & 36 (24/66) & 23 (21/90) & 0.11 & 29 [22,36]\,\tiny(156) \\
 &  & \#EC & 56 (18/32) & 59 (19/32) & 1.00 & 58 [46,69]\,\tiny(64) \\
 &  & class $\mathsf{TC}^0$ & 100 (40/40) & 100 (60/60) & 1.00 & 100 [96,100]\,\tiny(100) \\
 &  & class $\mathsf{L}$ & 100 (40/40) & 98 (59/60) & 1.00 & 99 [95,100]\,\tiny(100) \\
 &  & class $\mathsf{GapL}$ & 45 (67/148) & 41 (81/197) & 0.44 & 43 [38,48]\,\tiny(345) \\
 &  & class $\#\mathsf{P}$ & 36 (58/163) & 29 (61/212) & 0.18 & 32 [27,37]\,\tiny(375) \\
\midrule
Gemini 3.8 Flash (low) & Visual CoT & $\chi$ & 100 (40/40) & 100 (60/60) & 1.00 & 100 [96,100]\,\tiny(100) \\
 &  & \#CC & 100 (40/40) & 100 (60/60) & 1.00 & 100 [96,100]\,\tiny(100) \\
 &  & \#ST & 38 (25/66) & 40 (36/90) & 0.87 & 39 [32,47]\,\tiny(156) \\
 &  & \#PM & 70 (35/50) & 63 (47/75) & 0.45 & 66 [57,73]\,\tiny(125) \\
 &  & \#EC$_{\to}$ & 97 (31/32) & 100 (32/32) & 1.00 & 98 [92,100]\,\tiny(64) \\
 &  & \#M & 42 (28/66) & 37 (33/90) & 0.51 & 39 [32,47]\,\tiny(156) \\
 &  & \#Paths & 39 (26/66) & 33 (30/90) & 0.50 & 36 [29,44]\,\tiny(156) \\
 &  & \#EC & 72 (23/32) & 69 (22/32) & 1.00 & 70 [58,80]\,\tiny(64) \\
 &  & class $\mathsf{TC}^0$ & 100 (40/40) & 100 (60/60) & 1.00 & 100 [96,100]\,\tiny(100) \\
 &  & class $\mathsf{L}$ & 100 (40/40) & 100 (60/60) & 1.00 & 100 [96,100]\,\tiny(100) \\
 &  & class $\mathsf{GapL}$ & 61 (91/148) & 58 (115/197) & 0.58 & 60 [54,65]\,\tiny(345) \\
 &  & class $\#\mathsf{P}$ & 47 (77/164) & 40 (85/212) & 0.21 & 43 [38,48]\,\tiny(376) \\
\midrule
Gemini 3.1 Pro (low) & Visual CoT & $\chi$ & 100 (20/20) & 100 (60/60) & 1.00 & 100 [95,100]\,\tiny(80) \\
 &  & \#CC & 100 (20/20) & 100 (60/60) & 1.00 & 100 [95,100]\,\tiny(80) \\
 &  & \#ST & 31 (11/35) & 21 (19/90) & 0.25 & 24 [17,32]\,\tiny(125) \\
 &  & \#PM & 56 (14/25) & 45 (34/75) & 0.37 & 48 [38,58]\,\tiny(100) \\
 &  & \#EC$_{\to}$ & 88 (14/16) & 94 (30/32) & 0.59 & 92 [80,97]\,\tiny(48) \\
 &  & \#M & 37 (13/35) & 19 (17/90) & 0.04 & 24 [17,32]\,\tiny(125) \\
 &  & \#Paths & 43 (15/35) & 29 (26/90) & 0.14 & 33 [25,41]\,\tiny(125) \\
 &  & \#EC & 69 (11/16) & 75 (24/32) & 0.74 & 73 [59,83]\,\tiny(48) \\
 &  & class $\mathsf{TC}^0$ & 100 (20/20) & 100 (60/60) & 1.00 & 100 [95,100]\,\tiny(80) \\
 &  & class $\mathsf{L}$ & 100 (20/20) & 100 (60/60) & 1.00 & 100 [95,100]\,\tiny(80) \\
 &  & class $\mathsf{GapL}$ & 51 (39/76) & 42 (83/197) & 0.18 & 45 [39,51]\,\tiny(273) \\
 &  & class $\#\mathsf{P}$ & 45 (39/86) & 32 (67/212) & 0.03 & 36 [30,41]\,\tiny(298) \\
\bottomrule
\end{longtable}

%% file: gen_v8p/gen_v5/tab_v4_magnitude_v5.tex
\begin{tabular}{llrrrrrrr}
\toprule
Data & Cond. & $n$ & odds$/$decade & $\beta_{\shP}$ (SE) & LR $p$ & $m_{50}$ $\GapL$ & $m_{50}$ $\shP$ & $\Delta$ \\
\midrule
core, no XL & Visual & 312 & 0.36 & $-0.62$ (0.33) & 0.062 & $10^{3.6}$ & $10^{3.0}$ & 0.6 \\
 & Text & 311 & 0.34 & $-0.70$ (0.34) & 0.037 & $10^{3.5}$ & $10^{2.7}$ & 0.8 \\
core + XL & Visual & 372 & 0.36 & $-0.56$ (0.33) & 0.09 & $10^{3.5}$ & $10^{3.0}$ & 0.5 \\
 & Text & 371 & 0.33 & $-0.65$ (0.33) & 0.049 & $10^{3.4}$ & $10^{2.7}$ & 0.7 \\
pooled, no XL & Visual & 721 & 0.42 & $-0.75$ (0.21) & $2.4\times 10^{-4}$ & $10^{3.8}$ & $10^{2.9}$ & 1.0 \\
 & Text & 720 & 0.41 & $-0.79$ (0.20) & $8.1\times 10^{-5}$ & $10^{3.5}$ & $10^{2.5}$ & 1.0 \\
pooled + XL & Visual & 781 & 0.42 & $-0.72$ (0.21) & $4.3\times 10^{-4}$ & $10^{3.7}$ & $10^{2.9}$ & 0.9 \\
 & Text & 780 & 0.41 & $-0.77$ (0.20) & $1.3\times 10^{-4}$ & $10^{3.4}$ & $10^{2.5}$ & 1.0 \\
\midrule
Gemini 3.6 Flash (low), pooled & Visual & 720 & 0.33 & $-0.08$ (0.21) & 0.71 & $10^{2.2}$ & $10^{2.1}$ & 0.1 \\
Gemini 3.6 Flash (low), pooled & Text & 721 & 0.38 & $-0.33$ (0.20) & 0.10 & $10^{2.6}$ & $10^{2.3}$ & 0.3 \\
Gemini 3.1 Pro (low), pooled & Visual & 571 & 0.25 & $+0.17$ (0.26) & 0.52 & $10^{2.3}$ & $10^{2.4}$ & $-$0.1 \\
Gemini 3.1 Pro (low), pooled & Text & 571 & 0.22 & $-0.21$ (0.26) & 0.44 & $10^{2.4}$ & $10^{2.3}$ & 0.1 \\
Gemini 3.8 Flash (low), pooled & Visual$\to$Code & 721 & 0.73 & $-0.17$ (0.27) & 0.52 & $10^{15.2}$ & $10^{8.5}$ & 6.7 \\
Gemini 3.8 Flash (low), pooled & CAR & 721 & 0.75 & $+0.07$ (0.26) & 0.79 & $10^{13.8}$ & $10^{9.6}$ & 4.3 \\
\bottomrule
\end{tabular}

%% file: gen_v4/tab_v4_covariates.tex
\begin{tabular}{lrrrr}
\toprule
 & \multicolumn{2}{c}{Visual ($n$=721)} & \multicolumn{2}{c}{Text ($n$=720)} \\
\cmidrule(lr){2-3}\cmidrule(lr){4-5}
Term & $\beta$ (SE) & $p$ & $\beta$ (SE) & $p$ \\
\midrule
$\log_{10}$(answer) & $-0.48$ (0.08) & $5.9\times 10^{-9}$ & $-0.36$ (0.09) & $8.2\times 10^{-5}$ \\
$\shP$ class & $-1.54$ (0.27) & $1.3\times 10^{-8}$ & $-1.84$ (0.28) & $9.9\times 10^{-11}$ \\
$n$ (standardised) & $+1.22$ (1.76) & 0.49 & $+6.48$ (2.07) & 0.0018 \\
$m$ (standardised) & $-3.39$ (1.67) & 0.042 & $-9.87$ (2.09) & $2.5\times 10^{-6}$ \\
density (standardised) & $+0.10$ (0.38) & 0.80 & $-0.60$ (0.28) & 0.033 \\
Eulerian family & $-0.54$ (0.53) & 0.30 & $+1.57$ (0.58) & 0.0071 \\
\midrule
class, LR test in full model & LR $=36.7$ & $1.3\times 10^{-9}$ & LR $=49.2$ & $2.3\times 10^{-12}$ \\
magnitude, LR test in full model & LR $=39.8$ & $2.9\times 10^{-10}$ & LR $=16.3$ & $5.4\times 10^{-5}$ \\
\midrule
class, without \#ST & $+0.23$ (0.26) & 0.37 & $+0.52$ (0.26) & 0.049 \\
class, without \#PM & $-1.96$ (0.31) & $4.5\times 10^{-12}$ & $-2.14$ (0.31) & $1.3\times 10^{-14}$ \\
class, without \#EC$_{\to}$ & $-0.66$ (0.21) & 0.0014 & $-0.79$ (0.21) & $1.5\times 10^{-4}$ \\
class, without \#M & $-1.65$ (0.25) & $4.9\times 10^{-12}$ & $-1.35$ (0.23) & $2.3\times 10^{-9}$ \\
class, without \#Paths & $+0.23$ (0.25) & 0.36 & $+0.06$ (0.25) & 0.81 \\
class, without \#EC & $-0.77$ (0.22) & $3.7\times 10^{-4}$ & $-0.94$ (0.21) & $8.9\times 10^{-6}$ \\
class, grid family only & $-0.68$ (0.22) & 0.0017 & $-0.91$ (0.22) & $2.0\times 10^{-5}$ \\
class, Eulerian family only & $-1.29$ (1.17) & 0.23 & $+0.92$ (0.82) & 0.25 \\
\bottomrule
\end{tabular}

%% file: gen_v4/tab_v4_discordant.tex
\begin{tabular}{lrrrrrrrr}
\toprule
gap $\log_{10}$ & \multicolumn{4}{c}{Visual} & \multicolumn{4}{c}{Text} \\
\cmidrule(lr){2-5}\cmidrule(lr){6-9}
$(\shP)-(\GapL)$ & pairs & disc. & $\shP$ right & $p$ & pairs & disc. & $\shP$ right & $p$ \\
\midrule
$<-1$ & 206 & 21 & 8 & 0.38 & 206 & 33 & 10 & 0.035 \\
$[-1,0)$ & 62 & 21 & 10 & 1.00 & 62 & 17 & 4 & 0.049 \\
$[0,1)$ & 69 & 9 & 4 & 1.00 & 69 & 14 & 3 & 0.057 \\
$[1,2)$ & 42 & 10 & 0 & 0.002 & 42 & 12 & 0 & $4.9\times 10^{-4}$ \\
$[2,4)$ & 72 & 25 & 2 & $1.9\times 10^{-5}$ & 71 & 37 & 3 & $1.2\times 10^{-7}$ \\
$[4,\infty)$ & 50 & 12 & 0 & $4.9\times 10^{-4}$ & 50 & 2 & 0 & 0.50 \\
all & 501 & 98 & 24 & $4.2\times 10^{-7}$ & 500 & 115 & 20 & $7.1\times 10^{-13}$ \\
\bottomrule
\end{tabular}

%% file: gen_v4/tab_v4_twins_pooled.tex
\begin{tabular}{lllrrrlrll}
\toprule
Model & Cond. & Twin pair & $n$ & acc$_A$ & acc$_B$ & $\Delta$ [95\% CI] & $b/c$ & $p$ & Holm $p$ \\
\midrule
Gemini 3.6 Flash (low) & Text & same graph: \#ST ($\GapL$) vs \#M ($\shP$) & 156 & 24 & 29 & $-5$ $[-13, +3]$ & 14/22 & 0.24 & 1.00 \\
 &  & same graph: \#ST ($\GapL$) vs \#Paths ($\shP$) & 156 & 24 & 28 & $-4$ $[-10, +2]$ & 10/16 & 0.33 & 1.00 \\
 &  & same graph: \#PM ($\GapL$) vs \#M ($\shP$) & 125 & 58 & 32 & $+26$ $[+18, +35]$ & 37/4 & $1.0\times 10^{-7}$ & $3.9\times 10^{-6}$ \\
 &  & same image: $\chi$ ($\TC$) vs \#CC ($\Lsp$) & 100 & 99 & 100 & $-1$ $[-3, +0]$ & 0/1 & 1.00 & 1.00 \\
 &  & same drawing $\pm$ arrows: \#EC$_{\to}$ ($\GapL$) vs \#EC ($\shP$) & 64 & 91 & 69 & $+22$ $[+11, +33]$ & 15/1 & $5.2\times 10^{-4}$ & 0.016 \\
\midrule
Gemini 3.6 Flash (low) & Visual & same graph: \#ST ($\GapL$) vs \#M ($\shP$) & 155 & 23 & 24 & $-1$ $[-9, +6]$ & 16/18 & 0.86 & 1.00 \\
 &  & same graph: \#ST ($\GapL$) vs \#Paths ($\shP$) & 156 & 22 & 29 & $-6$ $[-12, -1]$ & 5/15 & 0.041 & 1.00 \\
 &  & same graph: \#PM ($\GapL$) vs \#M ($\shP$) & 124 & 42 & 24 & $+18$ $[+10, +26]$ & 26/4 & $5.9\times 10^{-5}$ & 0.002 \\
 &  & same image: $\chi$ ($\TC$) vs \#CC ($\Lsp$) & 100 & 100 & 99 & $+1$ $[+0, +3]$ & 1/0 & 1.00 & 1.00 \\
 &  & same drawing $\pm$ arrows: \#EC$_{\to}$ ($\GapL$) vs \#EC ($\shP$) & 64 & 95 & 58 & $+38$ $[+27, +50]$ & 24/0 & $1.2\times 10^{-7}$ & $4.4\times 10^{-6}$ \\
\midrule
Gemini 3.8 Flash (low) & CAR & same graph: \#ST ($\GapL$) vs \#M ($\shP$) & 156 & 89 & 89 & $+0$ $[-5, +5]$ & 8/8 & 1.00 & 1.00 \\
 &  & same graph: \#ST ($\GapL$) vs \#Paths ($\shP$) & 156 & 89 & 88 & $+1$ $[-5, +6]$ & 10/9 & 1.00 & 1.00 \\
 &  & same graph: \#PM ($\GapL$) vs \#M ($\shP$) & 125 & 82 & 88 & $-6$ $[-13, -1]$ & 4/12 & 0.077 & 1.00 \\
 &  & same image: $\chi$ ($\TC$) vs \#CC ($\Lsp$) & 100 & 95 & 99 & $-4$ $[-9, +1]$ & 1/5 & 0.22 & 1.00 \\
 &  & same drawing $\pm$ arrows: \#EC$_{\to}$ ($\GapL$) vs \#EC ($\shP$) & 64 & 100 & 100 & $+0$ $[+0, +0]$ & 0/0 & 1.00 & 1.00 \\
\midrule
Gemini 3.8 Flash (low) & Visual$\to$Code & same graph: \#ST ($\GapL$) vs \#M ($\shP$) & 156 & 90 & 85 & $+5$ $[-1, +11]$ & 14/6 & 0.12 & 1.00 \\
 &  & same graph: \#ST ($\GapL$) vs \#Paths ($\shP$) & 156 & 90 & 92 & $-2$ $[-7, +3]$ & 7/10 & 0.63 & 1.00 \\
 &  & same graph: \#PM ($\GapL$) vs \#M ($\shP$) & 125 & 84 & 82 & $+2$ $[-6, +9]$ & 11/9 & 0.82 & 1.00 \\
 &  & same image: $\chi$ ($\TC$) vs \#CC ($\Lsp$) & 100 & 100 & 100 & $+0$ $[+0, +0]$ & 0/0 & 1.00 & 1.00 \\
 &  & same drawing $\pm$ arrows: \#EC$_{\to}$ ($\GapL$) vs \#EC ($\shP$) & 64 & 98 & 98 & $+0$ $[-5, +5]$ & 1/1 & 1.00 & 1.00 \\
\midrule
Gemini 3.8 Flash (low) & Text & same graph: \#ST ($\GapL$) vs \#M ($\shP$) & 156 & 40 & 29 & $+11$ $[+4, +17]$ & 23/6 & 0.0023 & 0.069 \\
 &  & same graph: \#ST ($\GapL$) vs \#Paths ($\shP$) & 156 & 40 & 31 & $+8$ $[+1, +15]$ & 23/10 & 0.035 & 0.95 \\
 &  & same graph: \#PM ($\GapL$) vs \#M ($\shP$) & 125 & 58 & 31 & $+26$ $[+18, +34]$ & 35/2 & $1.0\times 10^{-8}$ & $4.1\times 10^{-7}$ \\
 &  & same image: $\chi$ ($\TC$) vs \#CC ($\Lsp$) & 100 & 100 & 100 & $+0$ $[+0, +0]$ & 0/0 & 1.00 & 1.00 \\
 &  & same drawing $\pm$ arrows: \#EC$_{\to}$ ($\GapL$) vs \#EC ($\shP$) & 63 & 92 & 73 & $+19$ $[+8, +32]$ & 14/2 & 0.0042 & 0.12 \\
\midrule
Gemini 3.8 Flash (low) & Visual & same graph: \#ST ($\GapL$) vs \#M ($\shP$) & 156 & 39 & 39 & $+0$ $[-6, +6]$ & 14/14 & 1.00 & 1.00 \\
 &  & same graph: \#ST ($\GapL$) vs \#Paths ($\shP$) & 156 & 39 & 36 & $+3$ $[-3, +9]$ & 13/8 & 0.38 & 1.00 \\
 &  & same graph: \#PM ($\GapL$) vs \#M ($\shP$) & 125 & 66 & 44 & $+22$ $[+14, +30]$ & 29/2 & $4.6\times 10^{-7}$ & $1.7\times 10^{-5}$ \\
 &  & same image: $\chi$ ($\TC$) vs \#CC ($\Lsp$) & 100 & 100 & 100 & $+0$ $[+0, +0]$ & 0/0 & 1.00 & 1.00 \\
 &  & same drawing $\pm$ arrows: \#EC$_{\to}$ ($\GapL$) vs \#EC ($\shP$) & 64 & 98 & 70 & $+28$ $[+17, +39]$ & 18/0 & $7.6\times 10^{-6}$ & $2.6\times 10^{-4}$ \\
\midrule
Gemini 3.1 Pro (low) & Text & same graph: \#ST ($\GapL$) vs \#M ($\shP$) & 125 & 24 & 23 & $+1$ $[-5, +6]$ & 7/6 & 1.00 & 1.00 \\
 &  & same graph: \#ST ($\GapL$) vs \#Paths ($\shP$) & 125 & 24 & 31 & $-7$ $[-14, -1]$ & 5/14 & 0.064 & 1.00 \\
 &  & same graph: \#PM ($\GapL$) vs \#M ($\shP$) & 100 & 53 & 24 & $+29$ $[+20, +38]$ & 30/1 & $3.0\times 10^{-8}$ & $1.2\times 10^{-6}$ \\
 &  & same image: $\chi$ ($\TC$) vs \#CC ($\Lsp$) & 80 & 100 & 100 & $+0$ $[+0, +0]$ & 0/0 & 1.00 & 1.00 \\
 &  & same drawing $\pm$ arrows: \#EC$_{\to}$ ($\GapL$) vs \#EC ($\shP$) & 48 & 98 & 71 & $+27$ $[+15, +40]$ & 13/0 & $2.4\times 10^{-4}$ & 0.0078 \\
\midrule
Gemini 3.1 Pro (low) & Visual & same graph: \#ST ($\GapL$) vs \#M ($\shP$) & 125 & 24 & 24 & $+0$ $[-6, +6]$ & 8/8 & 1.00 & 1.00 \\
 &  & same graph: \#ST ($\GapL$) vs \#Paths ($\shP$) & 125 & 24 & 33 & $-9$ $[-15, -2]$ & 3/14 & 0.013 & 0.36 \\
 &  & same graph: \#PM ($\GapL$) vs \#M ($\shP$) & 100 & 48 & 25 & $+23$ $[+15, +32]$ & 24/1 & $1.5\times 10^{-6}$ & $5.4\times 10^{-5}$ \\
 &  & same image: $\chi$ ($\TC$) vs \#CC ($\Lsp$) & 80 & 100 & 100 & $+0$ $[+0, +0]$ & 0/0 & 1.00 & 1.00 \\
 &  & same drawing $\pm$ arrows: \#EC$_{\to}$ ($\GapL$) vs \#EC ($\shP$) & 48 & 92 & 73 & $+19$ $[+4, +33]$ & 12/3 & 0.035 & 0.95 \\
\bottomrule
\end{tabular}

%% file: gen_v4/tab_v4_carcode.tex
\begin{tabular}{lllrrrlrl}
\toprule
Data & Model & Pair & $n$ & acc$_A$ & acc$_B$ & $\Delta$ [95\% CI] & $b/c$ & $p$ \\
\midrule
core & Gemini 3.6 Flash (low) & \car{} vs Visual$\to$Code & 202 & 94.6 & 94.1 & $+0.5$ $[-4.0, +4.6]$ & 10/9 & 1.00 \\
core & Gemini 3.8 Flash (low) & \car{} vs Visual$\to$Code & 392 & 90.6 & 92.9 & $-2.3$ $[-5.6, +0.8]$ & 13/22 & 0.18 \\
core & Gemini 3.8 Flash (low) & Text \car{} vs Text$\to$Code & 392 & 98.5 & 95.7 & $+2.8$ $[+0.3, +5.4]$ & 16/5 & 0.027 \\
core & Gemma 4 31B (no thinking) & \car{} vs Visual$\to$Code & 392 & 61.7 & 67.3 & $-5.6$ $[-10.4, -1.1]$ & 18/40 & 0.0054 \\
core & Gemma 4 31B (thinking) & \car{} vs Visual$\to$Code & 17 & 94.1 & 94.1 & $+0.0$ $[-17.6, +16.7]$ & 1/1 & 1.00 \\
core & Gemini 3.1 Pro (low) & \car{} vs Visual$\to$Code & 202 & 87.6 & 86.1 & $+1.5$ $[-2.0, +5.9]$ & 7/4 & 0.55 \\
pooled & Gemini 3.6 Flash (low) & \car{} vs Visual$\to$Code & 202 & 94.6 & 94.1 & $+0.5$ $[-4.0, +4.9]$ & 10/9 & 1.00 \\
pooled & Gemini 3.8 Flash (low) & \car{} vs Visual$\to$Code & 921 & 91.2 & 91.9 & $-0.7$ $[-2.6, +1.3]$ & 40/46 & 0.59 \\
pooled & Gemini 3.8 Flash (low) & Text \car{} vs Text$\to$Code & 392 & 98.5 & 95.7 & $+2.8$ $[+0.5, +5.3]$ & 16/5 & 0.027 \\
pooled & Gemma 4 31B (no thinking) & \car{} vs Visual$\to$Code & 392 & 61.7 & 67.3 & $-5.6$ $[-10.0, -1.3]$ & 18/40 & 0.0054 \\
pooled & Gemma 4 31B (thinking) & \car{} vs Visual$\to$Code & 17 & 94.1 & 94.1 & $+0.0$ $[-16.7, +16.7]$ & 1/1 & 1.00 \\
pooled & Gemini 3.1 Pro (low) & \car{} vs Visual$\to$Code & 202 & 87.6 & 86.1 & $+1.5$ $[-2.0, +5.9]$ & 7/4 & 0.55 \\
XL grids & Gemini 3.6 Flash (low) & \car{} vs Visual$\to$Code & 60 & 25.0 & 15.0 & $+10.0$ $[-1.6, +22.0]$ & 10/4 & 0.18 \\
XL grids & Gemini 3.6 Flash (low) & Text \car{} vs Text$\to$Code & 60 & 85.0 & 88.3 & $-3.3$ $[-15.0, +8.8]$ & 5/7 & 0.77 \\
XL grids & Gemini 3.8 Flash (low) & \car{} vs Visual$\to$Code & 60 & 23.3 & 10.0 & $+13.3$ $[+1.5, +27.9]$ & 10/2 & 0.039 \\
XL grids & Gemini 3.8 Flash (low) & Text \car{} vs Text$\to$Code & 60 & 90.0 & 95.0 & $-5.0$ $[-11.5, +1.8]$ & 2/5 & 0.45 \\
XL grids & Gemma 4 31B (no thinking) & \car{} vs Visual$\to$Code & 60 & 0.0 & 0.0 & $+0.0$ $[+0.0, +0.0]$ & 0/0 & 1.00 \\
XL grids & Gemini 3.1 Pro (low) & \car{} vs Visual$\to$Code & 60 & 3.3 & 3.3 & $+0.0$ $[-6.8, +8.8]$ & 2/2 & 1.00 \\
\bottomrule
\end{tabular}

%% file: gen_v4/tab_v4_transcription.tex
\begin{tabular}{llrrrrrrrrrr}
\toprule
Data & Cond. & records & no prog. & unscor. & $\hat G{=}G$, right & $\hat G{=}G$, wrong & $\hat G{\neq}G$, right & $\hat G{\neq}G$, wrong & $P(\text{right}\mid\hat G{=}G)$ & $P(\text{right}\mid\hat G{\neq}G)$ & exec$=\hat G$ \\
\midrule
pooled & Visual$\to$Code & 593 & 0 & 127 & 409 & 7 & 8 & 42 & 98\% & 16\% & 39/50 \\
pooled & Perceive+Code & 248 & 0 & 11 & 214 & 3 & 1 & 19 & 99\% & 5\% & 17/20 \\
pooled & \car{} & 593 & 0 & 80 & 414 & 5 & 44 & 50 & 99\% & 47\% & 43/94 \\
pooled & Text \car{} & 248 & 0 & 0 & 233 & 5 & 9 & 1 & 98\% & 90\% & 0/10 \\
pooled & Text$\to$Code & 248 & 0 & 0 & 234 & 14 & 0 & 0 & 94\% & 0\% & 0/0 \\
core & Visual$\to$Code & 248 & 0 & 53 & 171 & 0 & 4 & 20 & 100\% & 17\% & 19/24 \\
core & Perceive+Code & 248 & 0 & 11 & 214 & 3 & 1 & 19 & 99\% & 5\% & 17/20 \\
core & \car{} & 248 & 0 & 32 & 170 & 3 & 18 & 25 & 98\% & 42\% & 22/43 \\
core & Text \car{} & 248 & 0 & 0 & 233 & 5 & 9 & 1 & 98\% & 90\% & 0/10 \\
core & Text$\to$Code & 248 & 0 & 0 & 234 & 14 & 0 & 0 & 94\% & 0\% & 0/0 \\
XL grids & Visual$\to$Code & 60 & 0 & 42 & 3 & 1 & 1 & 13 & 75\% & 7\% & 8/14 \\
XL grids & Perceive+Code & 60 & 0 & 21 & 10 & 3 & 1 & 25 & 77\% & 4\% & 15/26 \\
XL grids & \car{} & 60 & 0 & 22 & 10 & 0 & 0 & 28 & 100\% & 0\% & 22/28 \\
XL grids & Text \car{} & 60 & 0 & 0 & 52 & 6 & 2 & 0 & 90\% & 100\% & 0/2 \\
XL grids & Text$\to$Code & 60 & 0 & 0 & 57 & 3 & 0 & 0 & 95\% & 0\% & 0/0 \\
\bottomrule
\end{tabular}

%% file: gen_v4/tab_v4_orient.tex
\begin{tabular}{llrrrl}
\toprule
Data & Event & unit & upright & transposed & $b/c$, $p$ \\
\midrule
core grids (v3 styles), 246 calls on 66 images & originally horizontal edge missed & call & 27 & 0 & 27/0, $1.5\times 10^{-8}$ \\
 & & image (any call) & 11 & 0 & 11/0, $9.8\times 10^{-4}$ \\
core grids (v3 styles), 246 calls on 66 images & originally vertical edge missed & call & 0 & 17 & 0/17, $1.5\times 10^{-5}$ \\
 & & image (any call) & 0 & 8 & 0/8, 0.0078 \\
core grids (v3 styles), 246 calls on 66 images & transcription not exact & call & 27 & 18 & 21/12, 0.16 \\
 & & image (any call) & 11 & 8 & 7/4, 0.55 \\
\midrule
XL grids, one call per grid (24 grids) & horizontal edge missed & grid & 8 & 9 & 5/6, 1.00 \\
XL grids, one call per grid (24 grids) & vertical edge missed & grid & 4 & 22 & 0/18, $7.6\times 10^{-6}$ \\
XL grids, one call per grid (24 grids) & any edge missed & grid & 9 & 22 & 0/13, $2.4\times 10^{-4}$ \\
\bottomrule
\end{tabular}

%% file: gen_v8p/gen_v5r/tab_zoom.tex
\begin{tabular}{lrrrrrrlrrrrr}
\toprule
Transcription & $7{\times}7$ & $8{\times}8$ & $9{\times}9$ & $10{\times}10$ & F1 & tokens & Solved by & $7{\times}7$ & $8{\times}8$ & $9{\times}9$ & $10{\times}10$ & total \\
\midrule
one shot, embedded in the program (Visual$\to$Code) & -- & -- & -- & -- & -- & -- & -- & 5/12 & 1/18 & 0/12 & 0/18 & 6/60 \\
one shot, embedded in the program (\car{}) & -- & -- & -- & -- & -- & -- & -- & 9/12 & 5/18 & 0/12 & 0/18 & 14/60 \\
\midrule
Zoom-1 (1 crop) & 5/6 & 5/6 & 1/6 & 2/6 & 0.991 & 4.6k & Text\,\car{} & 10/12 & 14/18 & 2/12 & 5/18 & 31/60 \\
 &  &  &  &  & &  & Text$\to$Code & 8/12 & 11/18 & 1/12 & 5/18 & 25/60 \\
Zoom-2 (4 crops) & 0/6 & 0/6 & 1/6 & 0/6 & 0.973 & 8.3k & Text\,\car{} & 0/12 & 0/18 & 2/12 & 0/18 & 2/60 \\
 &  &  &  &  & &  & Text$\to$Code & 0/12 & 0/18 & 2/12 & 0/18 & 2/60 \\
Zoom-3 (9 crops) & 2/6 & 4/6 & 1/6 & 0/6 & 0.992 & 14.8k & Text\,\car{} & 3/12 & 9/18 & 2/12 & 0/18 & 14/60 \\
 &  &  &  &  & &  & Text$\to$Code & 4/12 & 10/18 & 2/12 & 0/18 & 16/60 \\
two views, no verification & 6/6 & 3/6 & 0/6 & 0/6 & 0.991 & 9.6k & Text\,\car{} & 9/12 & 8/18 & 0/12 & 0/18 & 17/60 \\
 &  &  &  &  & &  & Text$\to$Code & 11/12 & 9/18 & 0/12 & 0/18 & 20/60 \\
two views + verification & 6/6 & 6/6 & 6/6 & 4/6 & 1.000 & 10.1k & Text\,\car{} & 10/12 & 14/18 & 7/12 & 12/18 & 43/60 \\
 &  &  &  &  & &  & Text$\to$Code & 11/12 & 15/18 & 11/12 & 12/18 & 49/60 \\
\midrule
ground-truth edge list (reference) & -- & -- & -- & -- & 1.000 & 0 & Text\,\car{} & 12/12 & 14/18 & 11/12 & 17/18 & 54/60 \\
 & -- & -- & -- & -- & 1.000 & 0 & Text$\to$Code & 12/12 & 17/18 & 11/12 & 17/18 & 57/60 \\
\bottomrule
\end{tabular}

%% file: gen_v4/tab_v41_transcription.tex
\begin{tabular}{lrrrrrrrrr}
\toprule
Images & $N$ & parsed & exact & mean F1 & missed & extra & non-lattice & exact pairs: shared / Visual & $p$ \\
\midrule
grid drawings & 156 & 156 & 150 (96\%) & 0.9993 & 11/5021 & 0 & 0 & 569/521 of 593 & $<10^{-4}$ \\
binary images & 100 & 100 & 100 (100\%) & 1.0000 & 0/1962 & 0 & -- & 200/200 of 200 & 1 \\
directed Eulerian & 64 & 64 & 64 (100\%) & 1.0000 & 0/1010 & 0 & -- & 64/64 of 64 & 1 \\
undirected Eulerian & 64 & 64 & 64 (100\%) & 1.0000 & 0/1010 & 0 & -- & 64/64 of 64 & 1 \\
all images & 384 & 384 & 378 (98\%) & 0.9997 & 11/9003 & 0 & 0 & 897/849 of 921 & $<10^{-4}$ \\
\bottomrule
\end{tabular}

%% file: gen_v4/tab_v41_decomposition.tex
\begin{tabular}{lrrrrrrrrrrrr}
\toprule
Task & pairs & excl. & $\hat G{=}G$ & $f(\hat G){=}f(G)$ & neither & perc.\ only & comp.\ only & both & cancel & Transcribe$\to$Text & Visual & Text \\
\midrule
$\sST$ & 156 & 0 & 150 & 150 & 59 & 0 & 91 & 6 & 0 & 38\% & 39\% & 40\% \\
$\sPM$ & 125 & 0 & 119 & 119 & 64 & 1 & 55 & 5 & 0 & 51\% & 66\% & 58\% \\
$\sECd$ & 64 & 0 & 64 & 64 & 61 & 0 & 3 & 0 & 0 & 95\% & 98\% & 92\% \\
$\chi$ & 100 & 0 & 100 & 100 & 100 & 0 & 0 & 0 & 0 & 100\% & 100\% & 100\% \\
$\sCC$ & 100 & 0 & 100 & 100 & 100 & 0 & 0 & 0 & 0 & 100\% & 100\% & 100\% \\
$\sAM$ & 156 & 0 & 150 & 150 & 35 & 0 & 115 & 6 & 0 & 22\% & 39\% & 29\% \\
$\sPATH$ & 156 & 0 & 150 & 150 & 43 & 0 & 107 & 6 & 0 & 28\% & 36\% & 31\% \\
$\sEC$ & 64 & 0 & 64 & 64 & 46 & 0 & 18 & 0 & 0 & 72\% & 70\% & 73\% \\
\midrule
polynomial & 545 & 0 & 533 & 533 & 384 & 1 & 149 & 11 & 0 & 70\% & 74\% & 72\% \\
$\shP$-complete & 376 & 0 & 364 & 364 & 124 & 0 & 240 & 12 & 0 & 33\% & 43\% & 37\% \\
all pairs & 921 & 0 & 897 & 897 & 508 & 1 & 389 & 23 & 0 & 55\% & 62\% & 58\% \\
\bottomrule
\end{tabular}

%% file: gen_v8p/gen_v4/tab_v41_paired.tex
\begin{tabular}{llrrrrrrr}
\toprule
Pairs & Comparison & $n$ & acc.\ A & acc.\ B & diff & $b/c$ & $p$ & 95\% CI \\
\midrule
all pairs & Transcribe$\to$Text vs.\ Visual & 921 & 55.2 & 61.7 & $-$6.5 & 28/88 & $<10^{-4}$ & $[-8.8, -4.3]$ \\
all pairs & Transcribe$\to$Text vs.\ Text & 920 & 55.2 & 57.9 & $-$2.7 & 41/66 & 0.0199 & $[-4.9, -0.4]$ \\
all pairs & Visual vs.\ Text & 920 & 61.7 & 57.9 & +3.8 & 77/42 & 0.00171 & $[+1.6, +6.0]$ \\
\midrule
polynomial & Transcribe$\to$Text vs.\ Visual & 545 & 70.5 & 74.5 & $-$4.0 & 16/38 & 0.00384 & $[-6.5, -1.7]$ \\
polynomial & Transcribe$\to$Text vs.\ Text & 545 & 70.5 & 72.1 & $-$1.7 & 21/30 & 0.262 & $[-4.3, +0.9]$ \\
polynomial & Visual vs.\ Text & 545 & 74.5 & 72.1 & +2.4 & 37/24 & 0.124 & $[-0.4, +5.2]$ \\
\midrule
$\shP$-complete & Transcribe$\to$Text vs.\ Visual & 376 & 33.0 & 43.1 & $-$10.1 & 12/50 & $<10^{-4}$ & $[-14.6, -5.9]$ \\
$\shP$-complete & Transcribe$\to$Text vs.\ Text & 375 & 33.1 & 37.3 & $-$4.3 & 20/36 & 0.044 & $[-8.4, -0.5]$ \\
$\shP$-complete & Visual vs.\ Text & 375 & 43.2 & 37.3 & +5.9 & 40/18 & 0.00535 & $[+1.9, +10.1]$ \\
\bottomrule
\end{tabular}

%% file: gen_v4/tab_v41_xl.tex
\begin{tabular}{lrrrrrrrrrrrrr}
\toprule
Scheme & reads exact & Jaccard & disp./grid & both miss & agree wrong & verifier & exact & mean F1 & miss/extra & Text \car{} & Text$\to$Code & calls & USD \\
\midrule
two-view (upright + rot.\ $90^\circ$, T=0) & 13/2 & 0.934 & 8.6 & 0 & 2 & 207/207 & 22/24 & 0.9997 & 0/2 & 43/60 & 49/60 & 375 & 2.16 \\
upright$\times$2, independent (T=1) & 11/8 & 0.980 & 2.6 & 10 & 6 & 63/63 & 15/24 & 0.9974 & 10/6 & 35/60 & 33/60 & 231 & 1.97 \\
upright$\times$2, repeated (T=0) & 9/8 & 0.988 & 1.5 & 12 & 7 & 37/37 & 13/24 & 0.9968 & 12/7 & 31/60 & 28/60 & 205 & 1.85 \\
\bottomrule
\end{tabular}

%% file: gen_v8p/gen_v4/tab_v41_xl_paired.tex
\begin{tabular}{llrrrrrrr}
\toprule
Control & Quantity & $n$ & control & two-view & diff (pts) & $b/c$ & $p$ & 95\% CI \\
\midrule
upright$\times$2, independent (T=1) & exact transcription (24 grids) & 24 & 15 & 22 & $-$29.2 & 1/8 & 0.0391 & $[-50.0, -8.3]$ \\
upright$\times$2, independent (T=1) & Text \car{} solved (60 pairs) & 60 & 35 & 43 & $-$13.3 & 9/17 & 0.169 & $[-37.7, +10.9]$ \\
upright$\times$2, independent (T=1) & Text$\to$Code solved (60 pairs) & 60 & 33 & 49 & $-$26.7 & 6/22 & 0.00372 & $[-51.6, -1.7]$ \\
upright$\times$2, repeated (T=0) & exact transcription (24 grids) & 24 & 13 & 22 & $-$37.5 & 1/10 & 0.0117 & $[-58.3, -12.5]$ \\
upright$\times$2, repeated (T=0) & Text \car{} solved (60 pairs) & 60 & 31 & 43 & $-$20.0 & 6/18 & 0.0227 & $[-41.3, +1.5]$ \\
upright$\times$2, repeated (T=0) & Text$\to$Code solved (60 pairs) & 60 & 28 & 49 & $-$35.0 & 5/26 & 0.000192 & $[-59.6, -10.0]$ \\
\bottomrule
\end{tabular}

%% file: gen_v4/tab_v41_cost.tex
\begin{tabular}{lrrrrr}
\toprule
Phase & calls & ok & input tokens & output tokens (incl.\ thinking) & USD \\
\midrule
A: transcription (one call per image) & 391 & 384 & 460,183 & 422,075 & 1.93 \\
A: task on $\hat G$ as text & 976 & 921 & 261,654 & 2,609,202 & 9.98 \\
B: upright readings & 105 & 96 & 118,272 & 380,464 & 1.52 \\
B: disputed-pair verification & 100 & 100 & 118,026 & 6,428 & 0.11 \\
B: solving from the verified list & 243 & 240 & 279,720 & 528,022 & 2.19 \\
\midrule
total & 1815 & 1741 & 1,237,855 & 3,946,191 & 15.73 \\
\bottomrule
\end{tabular}

%% file: gen_v4/tab_v4_twins_main.tex
\begin{tabular}{llrrrlrll}
\toprule
Model & Twin pair & $n$ & acc$_A$ & acc$_B$ & $\Delta$ [95\% CI] & $b/c$ & $p$ & Holm $p$ \\
\midrule
Gemini 3.8 Flash (low) & same graph: \#ST ($\GapL$) vs \#M ($\shP$) & 66 & 38 & 42 & $-5$ $[-14, +5]$ & 4/7 & 0.55 & 1.00 \\
 & same graph: \#ST ($\GapL$) vs \#Paths ($\shP$) & 66 & 38 & 39 & $-2$ $[-11, +8]$ & 4/5 & 1.00 & 1.00 \\
 & same graph: \#PM ($\GapL$) vs \#M ($\shP$) & 50 & 70 & 44 & $+26$ $[+14, +38]$ & 13/0 & $2.4\times 10^{-4}$ & 0.0059 \\
 & same image: $\chi$ ($\TC$) vs \#CC ($\Lsp$) & 40 & 100 & 100 & $+0$ $[+0, +0]$ & 0/0 & 1.00 & 1.00 \\
 & same drawing $\pm$ arrows: \#EC$_{\to}$ ($\GapL$) vs \#EC ($\shP$) & 32 & 97 & 72 & $+25$ $[+12, +41]$ & 8/0 & 0.0078 & 0.17 \\
\midrule
Gemini 3.8 Flash (high) & same graph: \#ST ($\GapL$) vs \#M ($\shP$) & 20 & 90 & 90 & $+0$ $[+0, +0]$ & 0/0 & 1.00 & 1.00 \\
 & same graph: \#ST ($\GapL$) vs \#Paths ($\shP$) & 21 & 86 & 81 & $+5$ $[+0, +14]$ & 1/0 & 1.00 & 1.00 \\
 & same graph: \#PM ($\GapL$) vs \#M ($\shP$) & 14 & 93 & 86 & $+7$ $[+0, +21]$ & 1/0 & 1.00 & 1.00 \\
 & same image: $\chi$ ($\TC$) vs \#CC ($\Lsp$) & 19 & 100 & 100 & $+0$ $[+0, +0]$ & 0/0 & 1.00 & 1.00 \\
 & same drawing $\pm$ arrows: \#EC$_{\to}$ ($\GapL$) vs \#EC ($\shP$) & 11 & 100 & 91 & $+9$ $[+0, +27]$ & 1/0 & 1.00 & 1.00 \\
\midrule
Gemini 3.6 Flash (high) & same graph: \#ST ($\GapL$) vs \#M ($\shP$) & 35 & 51 & 51 & $+0$ $[-14, +14]$ & 3/3 & 1.00 & 1.00 \\
 & same graph: \#ST ($\GapL$) vs \#Paths ($\shP$) & 35 & 51 & 49 & $+3$ $[-11, +17]$ & 4/3 & 1.00 & 1.00 \\
 & same graph: \#PM ($\GapL$) vs \#M ($\shP$) & 25 & 72 & 48 & $+24$ $[+8, +40]$ & 6/0 & 0.031 & 0.62 \\
 & same image: $\chi$ ($\TC$) vs \#CC ($\Lsp$) & 20 & 100 & 100 & $+0$ $[+0, +0]$ & 0/0 & 1.00 & 1.00 \\
 & same drawing $\pm$ arrows: \#EC$_{\to}$ ($\GapL$) vs \#EC ($\shP$) & 16 & 100 & 88 & $+12$ $[+0, +31]$ & 2/0 & 0.50 & 1.00 \\
\midrule
Gemini 3.1 Pro (high) & same graph: \#ST ($\GapL$) vs \#M ($\shP$) & 20 & 50 & 60 & $-10$ $[-30, +10]$ & 1/3 & 0.62 & 1.00 \\
 & same graph: \#ST ($\GapL$) vs \#Paths ($\shP$) & 19 & 53 & 58 & $-5$ $[-21, +11]$ & 1/2 & 1.00 & 1.00 \\
 & same graph: \#PM ($\GapL$) vs \#M ($\shP$) & 15 & 80 & 53 & $+27$ $[+7, +47]$ & 4/0 & 0.12 & 1.00 \\
 & same image: $\chi$ ($\TC$) vs \#CC ($\Lsp$) & 12 & 100 & 100 & $+0$ $[+0, +0]$ & 0/0 & 1.00 & 1.00 \\
\midrule
Gemma 4 31B (no thinking) & same graph: \#ST ($\GapL$) vs \#M ($\shP$) & 66 & 8 & 3 & $+5$ $[-2, +11]$ & 4/1 & 0.38 & 1.00 \\
 & same graph: \#ST ($\GapL$) vs \#Paths ($\shP$) & 66 & 8 & 21 & $-14$ $[-23, -5]$ & 1/10 & 0.012 & 0.25 \\
 & same graph: \#PM ($\GapL$) vs \#M ($\shP$) & 50 & 20 & 0 & $+20$ $[+10, +32]$ & 10/0 & 0.002 & 0.045 \\
 & same image: $\chi$ ($\TC$) vs \#CC ($\Lsp$) & 40 & 90 & 95 & $-5$ $[-15, +5]$ & 1/3 & 0.62 & 1.00 \\
 & same drawing $\pm$ arrows: \#EC$_{\to}$ ($\GapL$) vs \#EC ($\shP$) & 32 & 34 & 41 & $-6$ $[-19, +6]$ & 1/3 & 0.62 & 1.00 \\
\bottomrule
\end{tabular}

%% file: gen_v5/tab_v5_xl.tex
\begin{tabular}{lrrrrrrrrrrrrr}
\toprule
Scheme & reads exact & Jaccard & disp./grid & both miss & agree wrong & verifier & exact & mean F1 & miss/extra & Text \car{} & Text$\to$Code & calls & USD \\
\midrule
two-view (upright + rot.\ $90^\circ$, T=0) & 10/2 & 0.931 & 9.0 & 3 & 2 & 216/216 & 21/24 & 0.9994 & 3/2 & 50/60 & 42/60 & 384 & 2.16 \\
upright$\times$2, sampled (T=1) & 9/10 & 0.989 & 1.5 & 11 & 6 & 35/35 & 13/24 & 0.9974 & 11/6 & 30/60 & 26/60 & 203 & 1.89 \\
upright$\times$2, repeated (T=0) & 11/12 & 0.981 & 2.2 & 26 & 5 & 52/52 & 15/24 & 0.9946 & 26/5 & 38/60 & 31/60 & 220 & 1.98 \\
\bottomrule
\end{tabular}

%% file: gen_v8p/gen_v5/tab_v5_xl_paired.tex
\begin{tabular}{llrrrrrrr}
\toprule
Control & Quantity & $n$ & control & two-view & diff (pts) & $b/c$ & $p$ & 95\% CI \\
\midrule
upright$\times$2, sampled (T=1) & exact transcription (24 grids) & 24 & 13 & 21 & $-$33.3 & 0/8 & 0.00781 & $[-54.2, -16.7]$ \\
upright$\times$2, sampled (T=1) & Text \car{} solved (60 pairs) & 60 & 30 & 50 & $-$33.3 & 0/20 & $<10^{-4}$ & $[-52.5, -15.0]$ \\
upright$\times$2, sampled (T=1) & Text$\to$Code solved (60 pairs) & 60 & 26 & 42 & $-$26.7 & 5/21 & 0.00249 & $[-46.7, -5.5]$ \\
upright$\times$2, repeated (T=0) & exact transcription (24 grids) & 24 & 15 & 21 & $-$25.0 & 2/8 & 0.109 & $[-50.0, +0.0]$ \\
upright$\times$2, repeated (T=0) & Text \car{} solved (60 pairs) & 60 & 38 & 50 & $-$20.0 & 6/18 & 0.0227 & $[-44.4, +4.8]$ \\
upright$\times$2, repeated (T=0) & Text$\to$Code solved (60 pairs) & 60 & 31 & 42 & $-$18.3 & 9/20 & 0.0614 & $[-40.7, +4.8]$ \\
\bottomrule
\end{tabular}

%% file: gen_v5/tab_v5_xl_orient.tex
\begin{tabular}{lrrrrrr}
\toprule
Scheme & reading 1 miss H/V & reading 2 miss H/V & both miss H/V & final miss H/V & final extra H/V & unresolved \\
\midrule
two-view (upright + rot.\ $90^\circ$, T=0) & 40/15 & 36/109 & 2/1 & 2/1 & 2/0 & 0 \\
upright$\times$2, sampled (T=1) & 11/7 & 15/17 & 7/4 & 7/4 & 4/2 & 0 \\
upright$\times$2, repeated (T=0) & 32/4 & 59/4 & 26/0 & 26/0 & 3/2 & 0 \\
\bottomrule
\end{tabular}

%% file: gen_v5/tab_v5_retest.tex
\begin{tabular}{lrrrrrr}
\toprule
Prompt (same bytes twice) & $n$ & rep.\ 1 exact & rep.\ 2 exact & flips $b/c$ & $p$ & different answer \\
\midrule
Text \car{} on the true structure & 60 & 56 & 57 & 3/4 & 1 & 7 \\
Text$\to$Code on the true structure & 60 & 50 & 55 & 4/9 & 0.267 & 14 \\
\bottomrule
\end{tabular}

%% file: gen_v8/tab_prereg_timing.tex
\begin{tabular}{@{}llllrl@{}}
\toprule
 & plan & first & analysis & records & last \\
Exp. & frozen & call & frozen & at freeze & call \\
\midrule
B6 & 10-07 20:59:51 & 21:00:00 & 10-07 21:11 & 23 & 10-07 22:47 \\
C503 & 10-07 20:59:51 & 21:00:00 & 10-07 21:11 & 1 & 10-08 21:30 \\
F6 & 10-07 22:43:58 & 23:00:00 & 10-07 23:22 & 11 & 10-09 07:59 \\
T6 & 10-07 23:06:04 & 23:09:30 & 10-08 00:45 & 308 & 10-09 11:46 \\
B7 & 10-07 23:28:10 & 23:28:19 & 10-07 23:35 & 75 & 10-08 03:41 \\
X8 & 10-08 18:11:44 & 18:11:53 & 10-08 18:28 & 183 & 10-09 07:37 \\
D8 & 10-08 19:01:45 & 19:01:47 & 10-08 19:31 & 823 & 10-09 17:51 \\
G9 & 10-09 09:00:34 & 09:01:35 & 10-09 09:12 & 745 & 10-10 05:07 \\
QW & 10-09 22:43:04 & 22:44:31 & 10-09 22:45 & 33 & 10-10 12:19 \\
R9 & 10-11 10:23:39 & 10:24:04 & 10-11 10:23 & 0 & 10-11 11:15 \\
\bottomrule
\multicolumn{6}{@{}l}{Times UTC (2026); first call on the plan-freeze date.} \\
\end{tabular}

%% file: gen_v6/tab_v6_b7_xl.tex
\begin{tabular}{lrrrrrrrrrrrrr}
\toprule
Scheme & reads exact & Jaccard & disp./grid & both miss & agree wrong & verifier & exact & mean F1 & miss/extra & Text \car{} & Text$\to$Code & calls & USD \\
\midrule
upright + rot.\ $90^\circ$ & 47/8 & 0.924 & 10.1 & 7 & 4 & 968/968 & 86/96 & 0.9996 & 7/4 & 195/240 & 190/240 & 1640 & 9.01 \\
upright + rot.\ $180^\circ$ & 44/20 & 0.931 & 8.9 & 23 & 2 & 852/852 & 86/96 & 0.9990 & 23/2 & 203/240 & 194/240 & 1524 & 8.59 \\
upright$\times$2 (repeat) & 38/45 & 0.979 & 2.8 & 113 & 7 & 273/273 & 57/96 & 0.9951 & 113/7 & 133/240 & 128/240 & 945 & 7.82 \\
true structure (ceiling) & -- & -- & -- & -- & -- & -- & -- & -- & -- & 222/240 & 211/240 & -- & 4.39 \\
\bottomrule
\end{tabular}

%% file: gen_v8p/gen_v6/tab_v6_b7_paired.tex
\begin{tabular}{lllrrrrrrrr}
\toprule
Comparison & Quantity & role & $n$ & first & second & diff (pts) & $b/c$ & $p$ & Holm $p$ & 95\% CI \\
\midrule
rot.\ $90^\circ$ vs repeat & exact transcription (96 grids) & primary & 96 & 86 & 57 & +30.2 & 30/1 & $<10^{-4}$ & $<10^{-4}$ & $[+20.8, +39.6]$ \\
rot.\ $90^\circ$ vs repeat & Text \car{} solved (240 pairs) & secondary & 240 & 195 & 133 & +25.8 & 74/12 & $<10^{-4}$ & $<10^{-4}$ & $[+16.3, +35.9]$ \\
rot.\ $90^\circ$ vs repeat & Text$\to$Code solved (240 pairs) & secondary & 240 & 190 & 128 & +25.8 & 77/15 & $<10^{-4}$ & $<10^{-4}$ & $[+16.7, +35.3]$ \\
rot.\ $180^\circ$ vs repeat & exact transcription (96 grids) & primary & 96 & 86 & 57 & +30.2 & 32/3 & $<10^{-4}$ & $<10^{-4}$ & $[+19.8, +40.6]$ \\
rot.\ $180^\circ$ vs repeat & Text \car{} solved (240 pairs) & secondary & 240 & 203 & 133 & +29.2 & 88/18 & $<10^{-4}$ & $<10^{-4}$ & $[+18.2, +40.7]$ \\
rot.\ $180^\circ$ vs repeat & Text$\to$Code solved (240 pairs) & secondary & 240 & 194 & 128 & +27.5 & 87/21 & $<10^{-4}$ & $<10^{-4}$ & $[+16.9, +38.5]$ \\
rot.\ $90^\circ$ vs rot.\ $180^\circ$ & exact transcription (96 grids) & secondary & 96 & 86 & 86 & +0.0 & 6/6 & 1 & -- & $[-7.3, +7.3]$ \\
rot.\ $90^\circ$ vs rot.\ $180^\circ$ & Text \car{} solved (240 pairs) & secondary & 240 & 195 & 203 & $-$3.3 & 24/32 & 0.35 & 0.699 & $[-11.2, +4.9]$ \\
rot.\ $90^\circ$ vs rot.\ $180^\circ$ & Text$\to$Code solved (240 pairs) & secondary & 240 & 190 & 194 & $-$1.7 & 34/38 & 0.724 & 0.724 & $[-10.6, +7.5]$ \\
\bottomrule
\end{tabular}

%% file: gen_v6/tab_v6_b7_orient.tex
\begin{tabular}{lrrrrr}
\toprule
Single reading & readings & exact & missed H / true H & missed V / true V & extra H/V \\
\midrule
upright T=0 & 384 & 174 & 635/22376 & 181/22304 & 29/38 \\
rot.\ $90^\circ$ & 96 & 8 & 162/5594 & 500/5576 & 92/51 \\
rot.\ $180^\circ$ & 96 & 20 & 483/5594 & 190/5576 & 21/23 \\
\bottomrule
\end{tabular}

%% file: gen_v6/tab_v6_b7_size.tex
\begin{tabular}{lrrrrr}
\toprule
Scheme & 7$\times$7 & 8$\times$8 & 9$\times$9 & 10$\times$10 & all \\
\midrule
upright + rot.\ $90^\circ$ & 24/24 & 24/24 & 19/24 & 19/24 & 86/96 \\
upright + rot.\ $180^\circ$ & 23/24 & 23/24 & 18/24 & 22/24 & 86/96 \\
upright$\times$2 (repeat) & 21/24 & 22/24 & 7/24 & 7/24 & 57/96 \\
\bottomrule
\end{tabular}

%% file: gen_v6/tab_v6_b7_pooled.tex
\begin{tabular}{lrrr}
\toprule
Exact transcriptions (descriptive) & B6 & B7 & B6+B7 \\
\midrule
upright + rot.\ $90^\circ$ & 21/24 & 86/96 & 107/120 \\
upright + rot.\ $180^\circ$ & 22/24 & 86/96 & 108/120 \\
upright$\times$2 (repeat) & 17/24 & 57/96 & 74/120 \\
\midrule
rot.\ $90^\circ$ vs repeat: $b/c$ & 5/1 & 30/1 & 35/2 \\
rot.\ $180^\circ$ vs repeat: $b/c$ & 6/1 & 32/3 & 38/4 \\
rot.\ $90^\circ$ vs rot.\ $180^\circ$: $b/c$ & 2/3 & 6/6 & 8/9 \\
\bottomrule
\end{tabular}

%% file: gen_v5/tab_v5_b6_xl.tex
\begin{tabular}{lrrrrrrrrrrrrr}
\toprule
Scheme & reads exact & Jaccard & disp./grid & both miss & agree wrong & verifier & exact & mean F1 & miss/extra & Text \car{} & Text$\to$Code & calls & USD \\
\midrule
upright + rot.\ $90^\circ$ (T=0) & 12/3 & 0.925 & 9.3 & 3 & 1 & 224/224 & 21/24 & 0.9994 & 3/1 & 47/60 & 49/60 & 392 & 2.12 \\
upright + rot.\ $180^\circ$ (T=0) & 9/5 & 0.931 & 9.1 & 1 & 1 & 218/218 & 22/24 & 0.9997 & 1/1 & 55/60 & 45/60 & 386 & 2.16 \\
upright$\times$2, sampled (T=1) & 11/11 & 0.984 & 2.2 & 14 & 1 & 52/52 & 17/24 & 0.9974 & 14/1 & 42/60 & 38/60 & 220 & 1.99 \\
upright$\times$2, repeated (T=0) & 12/11 & 0.985 & 2.0 & 15 & 2 & 49/49 & 17/24 & 0.9971 & 15/2 & 41/60 & 37/60 & 217 & 1.93 \\
true structure (ceiling) & -- & -- & -- & -- & -- & -- & -- & -- & -- & 55/60 & 49/60 & -- & 1.12 \\
\bottomrule
\end{tabular}

%% file: gen_v8p/gen_v5/tab_v5_b6_paired.tex
\begin{tabular}{llrrrrrrrr}
\toprule
Comparison & Quantity & $n$ & first & second & diff (pts) & $b/c$ & $p$ & Holm $p$ & 95\% CI \\
\midrule
rot.\ $90^\circ$ vs rot.\ $180^\circ$ & exact transcription (24 grids) & 24 & 21 & 22 & $-$4.2 & 2/3 & 1 & 1 & $[-20.8, +12.5]$ \\
rot.\ $90^\circ$ vs rot.\ $180^\circ$ & Text \car{} solved (60 pairs) & 60 & 47 & 55 & $-$13.3 & 5/13 & 0.0963 & -- & $[-32.8, +6.6]$ \\
rot.\ $90^\circ$ vs rot.\ $180^\circ$ & Text$\to$Code solved (60 pairs) & 60 & 49 & 45 & +6.7 & 13/9 & 0.523 & -- & $[-13.6, +26.7]$ \\
rot.\ $90^\circ$ vs T=0 repeat & exact transcription (24 grids) & 24 & 21 & 17 & +16.7 & 5/1 & 0.219 & 0.656 & $[+0.0, +37.5]$ \\
rot.\ $90^\circ$ vs T=0 repeat & Text \car{} solved (60 pairs) & 60 & 47 & 41 & +10.0 & 14/8 & 0.286 & -- & $[-11.7, +31.3]$ \\
rot.\ $90^\circ$ vs T=0 repeat & Text$\to$Code solved (60 pairs) & 60 & 49 & 37 & +20.0 & 16/4 & 0.0118 & -- & $[+0.0, +39.0]$ \\
rot.\ $90^\circ$ vs T=1 repeat & exact transcription (24 grids) & 24 & 21 & 17 & +16.7 & 6/2 & 0.289 & 0.656 & $[-4.2, +37.5]$ \\
rot.\ $90^\circ$ vs T=1 repeat & Text \car{} solved (60 pairs) & 60 & 47 & 42 & +8.3 & 16/11 & 0.442 & -- & $[-16.1, +34.9]$ \\
rot.\ $90^\circ$ vs T=1 repeat & Text$\to$Code solved (60 pairs) & 60 & 49 & 38 & +18.3 & 17/6 & 0.0347 & -- & $[-1.9, +39.1]$ \\
rot.\ $180^\circ$ vs T=0 repeat & exact transcription (24 grids) & 24 & 22 & 17 & +20.8 & 6/1 & 0.125 & 0.5 & $[+0.0, +41.7]$ \\
rot.\ $180^\circ$ vs T=0 repeat & Text \car{} solved (60 pairs) & 60 & 55 & 41 & +23.3 & 16/2 & 0.00131 & -- & $[+5.1, +42.9]$ \\
rot.\ $180^\circ$ vs T=0 repeat & Text$\to$Code solved (60 pairs) & 60 & 45 & 37 & +13.3 & 17/9 & 0.169 & -- & $[-8.5, +35.7]$ \\
\bottomrule
\end{tabular}

%% file: gen_v5/tab_v5_b6_orient.tex
\begin{tabular}{lrrrrr}
\toprule
Single reading & readings & exact & missed H / true H & missed V / true V & extra H/V \\
\midrule
upright T=0 & 96 & 44 & 140/5616 & 31/5520 & 10/8 \\
upright T=1 & 48 & 22 & 58/2808 & 16/2760 & 6/2 \\
rot.\ $90^\circ$ & 24 & 3 & 30/1404 & 127/1380 & 14/11 \\
rot.\ $180^\circ$ & 24 & 5 & 124/1404 & 31/1380 & 5/6 \\
\bottomrule
\end{tabular}

%% file: gen_v6/tab_v6_x8_main.tex
\begin{tabular}{llrrrrrrr}
\toprule
Model & Condition & 7$\times$7 & 8$\times$8 & 9$\times$9 & 10$\times$10 & all & 95\% CI (\%) & tokens \\
\midrule
Gemini 3.8 Flash (low) & Visual CoT & 0/48 & 0/72 & 0/48 & 0/72 & 0/240 & [0, 0] & 5.4k \\
 & Text CoT & 0/48 & 0/72 & 0/48 & 0/72 & 0/240 & [0, 0] & 4.2k \\
 & Visual$\to$Code & 21/48 & 43/72 & 2/48 & 0/72 & 66/240 & [20, 35] & 4.6k \\
 & Perceive+Code & 20/48 & 33/72 & 2/48 & 0/72 & 55/240 & [16, 30] & 3.7k \\
 & \car{} & 18/48 & 32/72 & 3/48 & 0/72 & 53/240 & [16, 29] & 3.3k \\
 & Text$\to$Code & 45/48 & 66/72 & 42/48 & 63/72 & 216/240 & [86, 94] & 2.1k \\
 & Text\,\car{} & 46/48 & 65/72 & 45/48 & 68/72 & 224/240 & [90, 96] & 2.3k \\
\midrule
Gemini 3.6 Flash (low) & Visual CoT & 0/48 & 0/72 & 0/48 & 0/72 & 0/240 & [0, 0] & 6.9k \\
 & Text CoT & 0/48 & 0/72 & 0/48 & 0/72 & 0/240 & [0, 0] & 4.4k \\
 & Visual$\to$Code & 20/48 & 20/72 & 1/48 & 0/72 & 41/240 & [12, 23] & 3.8k \\
 & Perceive+Code & 12/48 & 20/72 & 2/48 & 0/72 & 34/240 & [9, 20] & 3.7k \\
 & \car{} & 21/48 & 18/72 & 4/48 & 0/72 & 43/240 & [12, 24] & 3.5k \\
 & Text$\to$Code & 45/48 & 65/72 & 42/48 & 60/72 & 212/240 & [85, 92] & 3.0k \\
 & Text\,\car{} & 43/48 & 63/72 & 43/48 & 60/72 & 209/240 & [83, 91] & 2.6k \\
\midrule
Gemini 3.1 Pro (low) & Visual CoT & 0/48 & 0/72 & 0/48 & 0/72 & 0/240 & [0, 0] & 5.0k \\
 & Text CoT & 0/48 & 0/72 & 0/48 & 0/72 & 0/240 & [0, 0] & 1.7k \\
 & Visual$\to$Code & 12/48 & 0/72 & 1/48 & 0/72 & 13/240 & [2, 10] & 4.6k \\
 & Perceive+Code & 10/48 & 2/72 & 0/48 & 0/72 & 12/240 & [2, 9] & 4.0k \\
 & \car{} & 10/48 & 4/72 & 0/48 & 0/72 & 14/240 & [2, 10] & 4.2k \\
 & Text$\to$Code & 48/48 & 71/72 & 48/48 & 71/72 & 238/240 & [98, 100] & 3.5k \\
 & Text\,\car{} & 48/48 & 72/72 & 48/48 & 72/72 & 240/240 & [100, 100] & 3.6k \\
\midrule
Gemma 4 31B (no thinking) & Visual CoT & 0/48 & 0/72 & 0/48 & 0/72 & 0/240 & [0, 0] & 3.2k \\
 & Text CoT & 0/48 & 0/72 & 0/48 & 0/72 & 0/240 & [0, 0] & 5.1k \\
 & Visual$\to$Code & 0/48 & 0/72 & 0/48 & 0/72 & 0/240 & [0, 0] & 3.0k \\
 & Perceive+Code & 0/48 & 0/72 & 0/48 & 0/72 & 0/240 & [0, 0] & 3.0k \\
 & \car{} & 0/48 & 0/72 & 0/48 & 0/72 & 0/240 & [0, 0] & 2.5k \\
 & Text$\to$Code & 29/48 & 39/72 & 13/48 & 30/72 & 111/240 & [40, 52] & 2.5k \\
 & Text\,\car{} & 42/48 & 62/72 & 44/48 & 54/72 & 202/240 & [80, 88] & 3.7k \\
\midrule
 & B7 two-view + Text\,\car{} & 45/48 & 63/72 & 35/48 & 52/72 & 195/240 & [74, 88] & -- \\
 & B7 true structure + Text\,\car{} & 46/48 & 65/72 & 47/48 & 64/72 & 222/240 & [89, 96] & -- \\
 & B7 true structure + Text$\to$Code & 41/48 & 65/72 & 43/48 & 62/72 & 211/240 & [84, 92] & -- \\
\bottomrule
\end{tabular}

%% file: gen_v6/tab_v6_x8_task.tex
\begin{tabular}{llrrrr}
\toprule
Model & Condition & \#ST & \#M & \#PM & all \\
\midrule
Gemini 3.8 Flash (low) & Visual CoT & 0/96 & 0/96 & 0/48 & 0/240 \\
 & Text CoT & 0/96 & 0/96 & 0/48 & 0/240 \\
 & Visual$\to$Code & 19/96 & 30/96 & 17/48 & 66/240 \\
 & Perceive+Code & 14/96 & 22/96 & 19/48 & 55/240 \\
 & \car{} & 24/96 & 19/96 & 10/48 & 53/240 \\
 & Text$\to$Code & 85/96 & 87/96 & 44/48 & 216/240 \\
 & Text\,\car{} & 96/96 & 91/96 & 37/48 & 224/240 \\
\midrule
Gemini 3.6 Flash (low) & Visual CoT & 0/96 & 0/96 & 0/48 & 0/240 \\
 & Text CoT & 0/96 & 0/96 & 0/48 & 0/240 \\
 & Visual$\to$Code & 12/96 & 19/96 & 10/48 & 41/240 \\
 & Perceive+Code & 10/96 & 13/96 & 11/48 & 34/240 \\
 & \car{} & 22/96 & 15/96 & 6/48 & 43/240 \\
 & Text$\to$Code & 78/96 & 88/96 & 46/48 & 212/240 \\
 & Text\,\car{} & 86/96 & 84/96 & 39/48 & 209/240 \\
\midrule
Gemini 3.1 Pro (low) & Visual CoT & 0/96 & 0/96 & 0/48 & 0/240 \\
 & Text CoT & 0/96 & 0/96 & 0/48 & 0/240 \\
 & Visual$\to$Code & 6/96 & 7/96 & 0/48 & 13/240 \\
 & Perceive+Code & 3/96 & 8/96 & 1/48 & 12/240 \\
 & \car{} & 9/96 & 4/96 & 1/48 & 14/240 \\
 & Text$\to$Code & 96/96 & 96/96 & 46/48 & 238/240 \\
 & Text\,\car{} & 96/96 & 96/96 & 48/48 & 240/240 \\
\midrule
Gemma 4 31B (no thinking) & Visual CoT & 0/96 & 0/96 & 0/48 & 0/240 \\
 & Text CoT & 0/96 & 0/96 & 0/48 & 0/240 \\
 & Visual$\to$Code & 0/96 & 0/96 & 0/48 & 0/240 \\
 & Perceive+Code & 0/96 & 0/96 & 0/48 & 0/240 \\
 & \car{} & 0/96 & 0/96 & 0/48 & 0/240 \\
 & Text$\to$Code & 37/96 & 59/96 & 15/48 & 111/240 \\
 & Text\,\car{} & 95/96 & 79/96 & 28/48 & 202/240 \\
\midrule
 & B7 two-view + Text\,\car{} & 85/96 & 76/96 & 34/48 & 195/240 \\
 & B7 true structure + Text\,\car{} & 96/96 & 87/96 & 39/48 & 222/240 \\
 & B7 true structure + Text$\to$Code & 82/96 & 87/96 & 42/48 & 211/240 \\
\bottomrule
\end{tabular}

%% file: gen_v8p/gen_v6/tab_v6_x8_paired.tex
\begin{tabular}{llrrrrrrrr}
\toprule
Comparison (pairs) & role & $n$ & first & second & diff (pts) & $b/c$ & $p$ & Holm $p$ & 95\% CI \\
\midrule
two-view (B7) Text\,\car{} vs \car{} & primary & 240 & 195 & 53 & +59.2 & 147/5 & $<10^{-4}$ & $<10^{-4}$ & $[+50.6, +67.8]$ \\
two-view (B7) Text\,\car{} vs \car{} & sensitivity & 240 & 195 & 53 & +59.2 & 147/5 & $<10^{-4}$ & -- & $[+50.4, +67.2]$ \\
\car{} vs Visual$\to$Code & primary & 240 & 53 & 66 & $-$5.4 & 21/34 & 0.105 & 0.105 & $[-11.5, +0.4]$ \\
\car{} vs Visual$\to$Code & sensitivity & 240 & 53 & 66 & $-$5.4 & 21/34 & 0.105 & -- & $[-11.5, +0.4]$ \\
Perceive+Code vs Visual$\to$Code & secondary & 240 & 55 & 66 & $-$4.6 & 18/29 & 0.144 & 0.432 & $[-10.6, +1.7]$ \\
\car{} vs Perceive+Code & secondary & 240 & 53 & 55 & $-$0.8 & 22/24 & 0.883 & 0.883 & $[-5.4, +3.8]$ \\
Text$\to$Code vs Visual$\to$Code & secondary & 240 & 216 & 66 & +62.5 & 154/4 & $<10^{-4}$ & $<10^{-4}$ & $[+54.4, +70.6]$ \\
Text\,\car{} vs \car{} & secondary & 240 & 224 & 53 & +71.2 & 177/6 & $<10^{-4}$ & $<10^{-4}$ & $[+63.2, +78.6]$ \\
Text\,\car{} vs Text$\to$Code & secondary & 240 & 224 & 216 & +3.3 & 24/16 & 0.268 & 0.536 & $[-1.3, +8.7]$ \\
two-view (B7) Text\,\car{} vs Text\,\car{} & secondary & 240 & 195 & 224 & $-$12.1 & 12/41 & $<10^{-4}$ & 0.000327 & $[-19.4, -5.0]$ \\
two-view (B7) Text\,\car{} vs Perceive+Code & secondary & 240 & 195 & 55 & +58.3 & 146/6 & $<10^{-4}$ & $<10^{-4}$ & $[+50.0, +67.1]$ \\
\car{} (Gemini 3.6 Flash (low)) vs Visual$\to$Code (Gemini 3.6 Flash (low)) & secondary & 240 & 43 & 41 & +0.8 & 24/22 & 0.883 & 0.883 & $[-4.2, +6.2]$ \\
Perceive+Code (Gemini 3.6 Flash (low)) vs Visual$\to$Code (Gemini 3.6 Flash (low)) & secondary & 240 & 34 & 41 & $-$2.9 & 17/24 & 0.349 & 0.698 & $[-8.5, +2.9]$ \\
Text\,\car{} (Gemini 3.6 Flash (low)) vs \car{} (Gemini 3.6 Flash (low)) & secondary & 240 & 209 & 43 & +69.2 & 173/7 & $<10^{-4}$ & $<10^{-4}$ & $[+61.7, +76.3]$ \\
\car{} (Gemini 3.1 Pro (low)) vs Visual$\to$Code (Gemini 3.1 Pro (low)) & secondary & 240 & 14 & 13 & +0.4 & 8/7 & 1 & 1 & $[-3.5, +4.5]$ \\
Perceive+Code (Gemini 3.1 Pro (low)) vs Visual$\to$Code (Gemini 3.1 Pro (low)) & secondary & 240 & 12 & 13 & $-$0.4 & 4/5 & 1 & 1 & $[-3.4, +2.5]$ \\
Text\,\car{} (Gemini 3.1 Pro (low)) vs \car{} (Gemini 3.1 Pro (low)) & secondary & 240 & 240 & 14 & +94.2 & 226/0 & $<10^{-4}$ & $<10^{-4}$ & $[+90.0, +97.6]$ \\
\car{} (Gemma 4 31B (no thinking)) vs Visual$\to$Code (Gemma 4 31B (no thinking)) & secondary & 240 & 0 & 0 & +0.0 & 0/0 & 1 & 1 & $[+0.0, +0.0]$ \\
Perceive+Code (Gemma 4 31B (no thinking)) vs Visual$\to$Code (Gemma 4 31B (no thinking)) & secondary & 240 & 0 & 0 & +0.0 & 0/0 & 1 & 1 & $[+0.0, +0.0]$ \\
Text\,\car{} (Gemma 4 31B (no thinking)) vs \car{} (Gemma 4 31B (no thinking)) & secondary & 240 & 202 & 0 & +84.2 & 202/0 & $<10^{-4}$ & $<10^{-4}$ & $[+79.6, +88.7]$ \\
Text\,\car{} vs true structure (B7) Text\,\car{} & descriptive & 240 & 224 & 222 & +0.8 & 15/13 & 0.851 & -- & $[-3.8, +5.7]$ \\
Text$\to$Code vs true structure (B7) Text$\to$Code & descriptive & 240 & 216 & 211 & +2.1 & 26/21 & 0.56 & -- & $[-2.9, +7.3]$ \\
\bottomrule
\end{tabular}

%% file: gen_v6/tab_v6_x8_decomp.tex
\begin{tabular}{llrrrrrrrr}
\toprule
Model & Condition & records & no list & scorable & $T{=}1,A{=}1$ & $T{=}1,A{=}0$ & $T{=}0,A{=}1$ & $T{=}0,A{=}0$ & $A=\#(\hat G)$ | $T{=}0$ \\
\midrule
Gemini 3.8 Flash (low) & Visual CoT & 240 & 0 & 240 & 0 & 75 & 0 & 165 & -- \\
 & Visual$\to$Code & 240 & 145 & 95 & 27 & 6 & 3 & 59 & 42/62 \\
 & Perceive+Code & 240 & 88 & 152 & 39 & 17 & 0 & 96 & 71/96 \\
 & \car{} & 240 & 121 & 119 & 28 & 6 & 2 & 83 & 71/85 \\
 & Text$\to$Code & 240 & 0 & 240 & 216 & 24 & 0 & 0 & 0/0 \\
 & Text\,\car{} & 240 & 0 & 240 & 217 & 16 & 7 & 0 & 0/7 \\
\midrule
Gemini 3.6 Flash (low) & Visual CoT & 240 & 0 & 240 & 0 & 69 & 0 & 171 & -- \\
 & Visual$\to$Code & 240 & 132 & 108 & 20 & 12 & 3 & 73 & 42/76 \\
 & Perceive+Code & 240 & 93 & 147 & 19 & 16 & 4 & 108 & 54/112 \\
 & \car{} & 240 & 144 & 96 & 24 & 1 & 5 & 66 & 54/71 \\
 & Text$\to$Code & 240 & 1 & 239 & 212 & 22 & 0 & 5 & 5/5 \\
 & Text\,\car{} & 240 & 3 & 237 & 209 & 19 & 0 & 9 & 6/9 \\
\midrule
Gemini 3.1 Pro (low) & Visual CoT & 240 & 0 & 240 & 0 & 17 & 0 & 223 & -- \\
 & Visual$\to$Code & 240 & 63 & 177 & 12 & 2 & 0 & 163 & 141/163 \\
 & Perceive+Code & 240 & 38 & 202 & 12 & 6 & 0 & 184 & 117/184 \\
 & \car{} & 240 & 59 & 181 & 11 & 0 & 0 & 170 & 148/170 \\
 & Text$\to$Code & 240 & 0 & 240 & 237 & 2 & 1 & 0 & 0/1 \\
 & Text\,\car{} & 240 & 0 & 240 & 238 & 0 & 2 & 0 & 0/2 \\
\midrule
Gemma 4 31B (no thinking) & Visual CoT & 240 & 5 & 235 & 0 & 0 & 0 & 235 & -- \\
 & Visual$\to$Code & 240 & 215 & 25 & 0 & 0 & 0 & 25 & 8/25 \\
 & Perceive+Code & 240 & 194 & 46 & 0 & 0 & 0 & 46 & 14/46 \\
 & \car{} & 240 & 210 & 30 & 0 & 0 & 0 & 30 & 20/30 \\
 & Text$\to$Code & 240 & 16 & 224 & 111 & 113 & 0 & 0 & 0/0 \\
 & Text\,\car{} & 240 & 0 & 240 & 202 & 38 & 0 & 0 & 0/0 \\
\bottomrule
\end{tabular}

%% file: gen_v6/tab_v6_x8_retest.tex
\begin{tabular}{lrrrrr}
\toprule
Descriptive comparison & $n$ & first & second & $b/c$ & $p$ \\
\midrule
Text\,\car{} (X8) vs true structure (B7) Text\,\car{} & 240 & 224 & 222 & 15/13 & 0.851 \\
Text$\to$Code (X8) vs true structure (B7) Text$\to$Code & 240 & 216 & 211 & 26/21 & 0.56 \\
\midrule
Gemini 3.6 Flash (low), \car{}: 24 development vs 96 fresh grids & 60 / 240 & 15 & 43 & -- & 0.272 \\
Gemini 3.6 Flash (low), Visual$\to$Code: 24 development vs 96 fresh grids & 60 / 240 & 9 & 41 & -- & 0.847 \\
Gemini 3.6 Flash (low), Text\,\car{}: 24 development vs 96 fresh grids & 60 / 240 & 51 & 209 & -- & 0.673 \\
Gemini 3.6 Flash (low), Text$\to$Code: 24 development vs 96 fresh grids & 60 / 240 & 53 & 212 & -- & 1 \\
Gemini 3.8 Flash (low), \car{}: 24 development vs 96 fresh grids & 60 / 240 & 14 & 53 & -- & 0.863 \\
Gemini 3.8 Flash (low), Visual$\to$Code: 24 development vs 96 fresh grids & 60 / 240 & 6 & 66 & -- & 0.00383 \\
Gemini 3.8 Flash (low), Perceive+Code: 24 development vs 96 fresh grids & 60 / 240 & 14 & 55 & -- & 1 \\
Gemini 3.8 Flash (low), Text\,\car{}: 24 development vs 96 fresh grids & 60 / 240 & 54 & 224 & -- & 0.406 \\
Gemini 3.8 Flash (low), Text$\to$Code: 24 development vs 96 fresh grids & 60 / 240 & 57 & 216 & -- & 0.315 \\
Gemini 3.8 Flash (low), Text CoT: 24 development vs 96 fresh grids & 60 / 240 & 0 & 0 & -- & 1 \\
Gemini 3.8 Flash (low), Visual CoT: 24 development vs 96 fresh grids & 60 / 240 & 0 & 0 & -- & 1 \\
Gemma 4 31B (no thinking), \car{}: 24 development vs 96 fresh grids & 60 / 240 & 0 & 0 & -- & 1 \\
Gemma 4 31B (no thinking), Visual$\to$Code: 24 development vs 96 fresh grids & 60 / 240 & 0 & 0 & -- & 1 \\
Gemma 4 31B (no thinking), Text\,\car{}: 24 development vs 96 fresh grids & 60 / 240 & 51 & 202 & -- & 1 \\
Gemma 4 31B (no thinking), Visual CoT: 24 development vs 96 fresh grids & 60 / 240 & 0 & 0 & -- & 1 \\
Gemini 3.1 Pro (low), \car{}: 24 development vs 96 fresh grids & 60 / 240 & 2 & 14 & -- & 0.747 \\
Gemini 3.1 Pro (low), Visual$\to$Code: 24 development vs 96 fresh grids & 60 / 240 & 2 & 13 & -- & 0.743 \\
Gemini 3.1 Pro (low), Text\,\car{}: 24 development vs 96 fresh grids & 60 / 240 & 60 & 240 & -- & 1 \\
Gemini 3.1 Pro (low), Visual CoT: 24 development vs 96 fresh grids & 59 / 240 & 0 & 0 & -- & 1 \\
\bottomrule
\end{tabular}

%% file: gen_v8p/gen_v6/tab_v6_d8_primary.tex
\begin{tabular}{lrrrrrrr}
\toprule
Arm & exact & diff (pts) & Newcombe 95\% CI & bootstrap 95\% CI & $b/c$ & $p$ & Holm $p$ \\
\midrule
clean & 167/170 &  &  &  &  &  &  \\
clean, repeated & 170/170 & +1.8 & $[-0.7, +5.1]$ & $[+0.0, +4.1]$ & 0/3 & 0.25 & -- \\
\midrule
JPEG $q{=}30$ & 153/170 & $-$8.2 & $[-13.5, -4.0]$ & $[-12.8, -4.1]$ & 15/1 & 0.000519 & 0.00519 \\
JPEG $q{=}10$ & 160/170 & $-$4.1 & $[-8.7, -0.2]$ & $[-8.1, -0.6]$ & 9/2 & 0.0654 & 0.589 \\
blur $\sigma{=}1.5$ & 165/170 & $-$1.2 & $[-4.8, +2.1]$ & $[-4.1, +1.7]$ & 4/2 & 0.688 & 1 \\
blur $\sigma{=}3$ & 167/170 & +0.0 & $[-3.5, +3.5]$ & $[-2.9, +2.9]$ & 3/3 & 1 & 1 \\
perspective 6\% & 167/170 & +0.0 & $[-3.5, +3.5]$ & $[-2.9, +2.9]$ & 3/3 & 1 & 1 \\
perspective 12\% & 163/170 & $-$2.4 & $[-6.7, +1.6]$ & $[-6.1, +1.2]$ & 7/3 & 0.344 & 1 \\
elastic 3\,px & 169/170 & +1.2 & $[-1.7, +4.5]$ & $[-1.2, +3.6]$ & 1/3 & 0.625 & 1 \\
elastic 6\,px & 170/170 & +1.8 & $[-0.7, +5.1]$ & $[+0.0, +4.0]$ & 0/3 & 0.25 & 1 \\
half resolution & 165/170 & $-$1.2 & $[-4.8, +2.1]$ & $[-4.2, +1.7]$ & 4/2 & 0.688 & 1 \\
photo-like & 160/170 & $-$4.1 & $[-8.7, -0.2]$ & $[-7.9, -0.6]$ & 9/2 & 0.0654 & 0.589 \\
\bottomrule
\end{tabular}

%% file: gen_v6/tab_v6_d8_sens.tex
\begin{tabular}{llrrrrrrr}
\toprule
Model & Arm & $n$ & $b/c$ (missing = wrong) & Holm $p$ & clusters & $b/c$ & $p$ & Holm $p$ \\
\midrule
3.8 Flash & JPEG $q{=}30$ & 170 & 15/1 & 0.00519 & 138 & 15/1 & 0.000519 & 0.00519 \\
 & JPEG $q{=}10$ & 170 & 9/2 & 0.589 & 138 & 9/2 & 0.0654 & 0.589 \\
 & blur $\sigma{=}1.5$ & 170 & 4/2 & 1 & 138 & 4/2 & 0.688 & 1 \\
 & blur $\sigma{=}3$ & 170 & 3/3 & 1 & 138 & 3/3 & 1 & 1 \\
 & perspective 6\% & 170 & 3/3 & 1 & 138 & 3/3 & 1 & 1 \\
 & perspective 12\% & 170 & 7/3 & 1 & 138 & 7/3 & 0.344 & 1 \\
 & elastic 3\,px & 170 & 1/3 & 1 & 138 & 1/3 & 0.625 & 1 \\
 & elastic 6\,px & 170 & 0/3 & 1 & 138 & 0/3 & 0.25 & 1 \\
 & half resolution & 170 & 4/2 & 1 & 138 & 4/2 & 0.688 & 1 \\
 & photo-like & 170 & 9/2 & 0.589 & 138 & 9/2 & 0.0654 & 0.589 \\
\midrule
3.1 Pro & JPEG $q{=}30$ & 170 & 6/2 & 1 & 138 & 6/2 & 0.289 & 1 \\
 & JPEG $q{=}10$ & 170 & 7/5 & 1 & 138 & 7/5 & 0.774 & 1 \\
 & blur $\sigma{=}1.5$ & 170 & 1/7 & 0.703 & 138 & 1/7 & 0.0703 & 0.703 \\
 & blur $\sigma{=}3$ & 170 & 7/6 & 1 & 138 & 7/6 & 1 & 1 \\
 & perspective 6\% & 170 & 0/1 & 1 & 138 & 0/1 & 1 & 1 \\
 & perspective 12\% & 170 & 7/4 & 1 & 138 & 7/4 & 0.549 & 1 \\
 & elastic 3\,px & 170 & 4/1 & 1 & 138 & 4/1 & 0.375 & 1 \\
 & elastic 6\,px & 170 & 3/3 & 1 & 138 & 3/3 & 1 & 1 \\
 & half resolution & 170 & 5/5 & 1 & 138 & 5/5 & 1 & 1 \\
 & photo-like & 170 & 10/6 & 1 & 138 & 10/6 & 0.454 & 1 \\
\midrule
3.6 Flash & JPEG $q{=}30$ & 170 & 2/2 & 1 & 138 & 2/2 & 1 & 1 \\
 & JPEG $q{=}10$ & 170 & 5/2 & 1 & 138 & 5/2 & 0.453 & 1 \\
 & blur $\sigma{=}1.5$ & 170 & 2/2 & 1 & 138 & 2/2 & 1 & 1 \\
 & blur $\sigma{=}3$ & 170 & 2/2 & 1 & 138 & 2/2 & 1 & 1 \\
 & perspective 6\% & 170 & 1/1 & 1 & 138 & 1/1 & 1 & 1 \\
 & perspective 12\% & 170 & 7/1 & 0.703 & 138 & 7/1 & 0.0703 & 0.703 \\
 & elastic 3\,px & 170 & 2/2 & 1 & 138 & 2/2 & 1 & 1 \\
 & elastic 6\,px & 170 & 3/1 & 1 & 138 & 3/1 & 0.625 & 1 \\
 & half resolution & 170 & 1/2 & 1 & 138 & 1/2 & 1 & 1 \\
 & photo-like & 170 & 3/2 & 1 & 138 & 3/2 & 1 & 1 \\
\midrule
Gemma 4 31B & JPEG $q{=}30$ & 170 & 9/7 & 1 & 138 & 7/7 & 1 & 1 \\
 & JPEG $q{=}10$ & 170 & 13/7 & 1 & 138 & 12/6 & 0.238 & 1 \\
 & blur $\sigma{=}1.5$ & 170 & 10/9 & 1 & 138 & 8/8 & 1 & 1 \\
 & blur $\sigma{=}3$ & 170 & 18/4 & 0.0434 & 138 & 17/4 & 0.0072 & 0.072 \\
 & perspective 6\% & 170 & 10/12 & 1 & 138 & 10/12 & 0.832 & 1 \\
 & perspective 12\% & 170 & 16/14 & 1 & 138 & 15/13 & 0.851 & 1 \\
 & elastic 3\,px & 170 & 4/10 & 1 & 138 & 2/9 & 0.0654 & 0.589 \\
 & elastic 6\,px & 170 & 7/7 & 1 & 138 & 5/6 & 1 & 1 \\
 & half resolution & 170 & 14/5 & 0.572 & 138 & 12/5 & 0.143 & 1 \\
 & photo-like & 170 & 12/11 & 1 & 138 & 11/10 & 1 & 1 \\
\bottomrule
\end{tabular}

%% file: gen_v6/tab_v6_d8_models.tex
\begin{tabular}{lrrrrrrrr}
\toprule
Arm & 3.8 Flash & Holm $p$ & 3.1 Pro & Holm $p$ & 3.6 Flash & Holm $p$ & Gemma 4 31B & Holm $p$ \\
\midrule
clean & 167/170 & -- & 161/170 & -- & 168/170 & -- & 139/170 & -- \\
clean, repeated & 170/170 & 0.25$^\dagger$ & 159/170 & 0.5$^\dagger$ & 168/170 & 1$^\dagger$ & 139/170 & 1$^\dagger$ \\
\midrule
JPEG $q{=}30$ & 153/170 & 0.00519 & 157/170 & 1 & 168/170 & 1 & 137/170 & 1 \\
JPEG $q{=}10$ & 160/170 & 0.589 & 159/170 & 1 & 165/170 & 1 & 133/170 & 1 \\
blur $\sigma{=}1.5$ & 165/170 & 1 & 167/170 & 0.703 & 168/170 & 1 & 138/170 & 1 \\
blur $\sigma{=}3$ & 167/170 & 1 & 160/170 & 1 & 168/170 & 1 & 125/170 & 0.0434 \\
perspective 6\% & 167/170 & 1 & 162/170 & 1 & 168/170 & 1 & 141/170 & 1 \\
perspective 12\% & 163/170 & 1 & 158/170 & 1 & 162/170 & 0.703 & 137/170 & 1 \\
elastic 3\,px & 169/170 & 1 & 158/170 & 1 & 168/170 & 1 & 145/170 & 1 \\
elastic 6\,px & 170/170 & 1 & 161/170 & 1 & 166/170 & 1 & 139/170 & 1 \\
half resolution & 165/170 & 1 & 161/170 & 1 & 169/170 & 1 & 130/170 & 0.572 \\
photo-like & 160/170 & 0.589 & 157/170 & 1 & 167/170 & 1 & 138/170 & 1 \\
\bottomrule
\end{tabular}

%% file: gen_v6/tab_v6_d8_answers.tex
\begin{tabular}{lrrrrrrrr}
\toprule
Arm & 3.8 Flash Visual CoT & Holm $p$ & 3.8 Flash Visual$\to$Code & Holm $p$ & 3.1 Pro Visual CoT & Holm $p$ & 3.6 Flash Visual CoT & Holm $p$ \\
\midrule
clean & 257/392 & -- & 364/392 & -- & 214/392 & -- & 210/392 & -- \\
clean, repeated & 244/392 & 0.066$^\dagger$ & 361/392 & 0.736$^\dagger$ & 223/392 & 0.136$^\dagger$ & 214/392 & 0.652$^\dagger$ \\
\midrule
JPEG $q{=}30$ & 242/392 & 0.122 & 309/392 & $<10^{-4}$ & 222/392 & 1 & 202/392 & 1 \\
JPEG $q{=}10$ & 247/392 & 0.61 & 346/392 & 0.106 & 221/392 & 1 & 211/392 & 1 \\
blur $\sigma{=}1.5$ & 246/392 & 0.704 & 357/392 & 1 & 217/392 & 1 & 207/392 & 1 \\
blur $\sigma{=}3$ & 246/392 & 0.704 & 360/392 & 1 & 210/392 & 1 & 196/392 & 0.595 \\
perspective 6\% & 248/392 & 0.704 & 367/392 & 1 & 221/392 & 1 & 215/392 & 1 \\
perspective 12\% & 250/392 & 0.721 & 370/392 & 1 & 220/392 & 1 & 208/392 & 1 \\
elastic 3\,px & 238/392 & 0.0432 & 365/392 & 1 & 226/392 & 0.73 & 199/392 & 1 \\
elastic 6\,px & 244/392 & 0.426 & 372/392 & 1 & 224/392 & 1 & 207/392 & 1 \\
half resolution & 254/392 & 0.755 & 364/392 & 1 & 217/392 & 1 & 207/392 & 1 \\
photo-like & 246/392 & 0.704 & 330/392 & $<10^{-4}$ & 221/392 & 1 & 208/392 & 1 \\
\bottomrule
\end{tabular}

%% file: gen_v8p/gen_v6/tab_v6_d8_attrib.tex
\begin{tabular}{llrrrrrrrr}
\toprule
Model & Arm & $n$ & $T{=}1,A{=}1$ & $T{=}1,A{=}0$ & $T{=}0,A{=}1$ & $T{=}0,A{=}0$ & $P(A{\mid}T{=}1)$ & errors with $T{=}1$ (\%) & vs clean (pts) \\
\midrule
3.8 Flash & clean & 392 & 257 & 115 & 0 & 20 & 69 & 85 & -- \\
 & clean, repeated & 392 & 244 & 118 & 0 & 30 & 67 & 80 & $-$5 $[-13, +3]$ \\
 & JPEG $q{=}30$ & 392 & 242 & 76 & 0 & 74 & 76 & 51 & $-$35 $[-47, -22]$ \\
 & JPEG $q{=}10$ & 392 & 247 & 113 & 0 & 32 & 69 & 78 & $-$7 $[-15, +1]$ \\
 & blur $\sigma{=}1.5$ & 392 & 246 & 114 & 0 & 32 & 68 & 78 & $-$7 $[-17, +2]$ \\
 & blur $\sigma{=}3$ & 392 & 246 & 117 & 0 & 29 & 68 & 80 & $-$5 $[-17, +7]$ \\
 & perspective 6\% & 392 & 247 & 127 & 1 & 17 & 66 & 88 & +3 $[-5, +12]$ \\
 & perspective 12\% & 392 & 243 & 123 & 7 & 19 & 66 & 87 & +1 $[-7, +9]$ \\
 & elastic 3\,px & 392 & 237 & 138 & 1 & 16 & 63 & 90 & +4 $[-4, +13]$ \\
 & elastic 6\,px & 392 & 244 & 141 & 0 & 7 & 63 & 95 & +10 $[+1, +20]$ \\
 & half resolution & 392 & 254 & 114 & 0 & 24 & 69 & 83 & $-$3 $[-11, +6]$ \\
 & photo-like & 392 & 244 & 90 & 2 & 56 & 73 & 62 & $-$24 $[-35, -12]$ \\
\midrule
3.1 Pro & clean & 392 & 214 & 119 & 0 & 59 & 64 & 67 & -- \\
 & clean, repeated & 392 & 223 & 112 & 0 & 57 & 67 & 66 & $-$1 $[-4, +3]$ \\
 & JPEG $q{=}30$ & 392 & 221 & 124 & 1 & 46 & 64 & 73 & +6 $[-5, +18]$ \\
 & JPEG $q{=}10$ & 392 & 221 & 124 & 0 & 47 & 64 & 73 & +6 $[-3, +15]$ \\
 & blur $\sigma{=}1.5$ & 392 & 216 & 153 & 1 & 22 & 59 & 87 & +21 $[+9, +32]$ \\
 & blur $\sigma{=}3$ & 392 & 210 & 155 & 0 & 27 & 58 & 85 & +18 $[+6, +30]$ \\
 & perspective 6\% & 392 & 219 & 123 & 2 & 48 & 64 & 72 & +5 $[-6, +16]$ \\
 & perspective 12\% & 392 & 214 & 114 & 6 & 58 & 65 & 66 & $-$1 $[-11, +10]$ \\
 & elastic 3\,px & 392 & 226 & 107 & 0 & 59 & 68 & 64 & $-$2 $[-11, +7]$ \\
 & elastic 6\,px & 392 & 224 & 114 & 0 & 54 & 66 & 68 & +1 $[-8, +10]$ \\
 & half resolution & 392 & 216 & 125 & 1 & 50 & 63 & 71 & +5 $[-3, +13]$ \\
 & photo-like & 392 & 219 & 130 & 2 & 41 & 63 & 76 & +9 $[-2, +21]$ \\
\midrule
3.6 Flash & clean & 392 & 210 & 173 & 0 & 9 & 55 & 95 & -- \\
 & clean, repeated & 392 & 211 & 171 & 3 & 7 & 55 & 96 & +1 $[-2, +4]$ \\
 & JPEG $q{=}30$ & 392 & 201 & 183 & 1 & 7 & 52 & 96 & +1 $[-5, +6]$ \\
 & JPEG $q{=}10$ & 392 & 208 & 158 & 3 & 23 & 57 & 87 & $-$8 $[-18, +1]$ \\
 & blur $\sigma{=}1.5$ & 392 & 205 & 180 & 2 & 5 & 53 & 97 & +2 $[-2, +7]$ \\
 & blur $\sigma{=}3$ & 392 & 194 & 186 & 2 & 10 & 51 & 95 & $-$0 $[-7, +6]$ \\
 & perspective 6\% & 392 & 214 & 167 & 1 & 10 & 56 & 94 & $-$1 $[-6, +4]$ \\
 & perspective 12\% & 392 & 197 & 167 & 11 & 17 & 54 & 91 & $-$4 $[-10, +1]$ \\
 & elastic 3\,px & 392 & 198 & 178 & 1 & 15 & 53 & 92 & $-$3 $[-6, +0]$ \\
 & elastic 6\,px & 392 & 206 & 176 & 1 & 9 & 54 & 95 & +0 $[-4, +4]$ \\
 & half resolution & 392 & 206 & 181 & 1 & 4 & 53 & 98 & +3 $[-1, +6]$ \\
 & photo-like & 392 & 207 & 167 & 1 & 17 & 55 & 91 & $-$4 $[-12, +3]$ \\
\bottomrule
\end{tabular}

%% file: gen_v6/tab_v6_d8_percept.tex
\begin{tabular}{lrrrrrr}
\toprule
Arm & 3.8 Flash: in Visual CoT (pairs) & transcription call (images) & 3.1 Pro: in Visual CoT (pairs) & transcription call (images) & 3.6 Flash: in Visual CoT (pairs) & transcription call (images) \\
\midrule
clean & 372/392 & 167/170 & 333/392 & 161/170 & 383/392 & 168/170 \\
clean, repeated & 362/392 & 170/170 & 335/392 & 159/170 & 382/392 & 168/170 \\
\midrule
JPEG $q{=}30$ & 318/392 & 153/170 & 345/392 & 157/170 & 384/392 & 168/170 \\
JPEG $q{=}10$ & 360/392 & 160/170 & 345/392 & 159/170 & 366/392 & 165/170 \\
blur $\sigma{=}1.5$ & 360/392 & 165/170 & 369/392 & 167/170 & 385/392 & 168/170 \\
blur $\sigma{=}3$ & 363/392 & 167/170 & 365/392 & 160/170 & 380/392 & 168/170 \\
perspective 6\% & 374/392 & 167/170 & 342/392 & 162/170 & 381/392 & 168/170 \\
perspective 12\% & 366/392 & 163/170 & 328/392 & 158/170 & 364/392 & 162/170 \\
elastic 3\,px & 375/392 & 169/170 & 333/392 & 158/170 & 376/392 & 168/170 \\
elastic 6\,px & 385/392 & 170/170 & 338/392 & 161/170 & 382/392 & 166/170 \\
half resolution & 368/392 & 165/170 & 341/392 & 161/170 & 387/392 & 169/170 \\
photo-like & 334/392 & 160/170 & 349/392 & 157/170 & 374/392 & 167/170 \\
\bottomrule
\end{tabular}

%% file: gen_v6/tab_v6_d8_errors.tex
\begin{tabular}{llrrrrrrrrrrr}
\toprule
Model & Arm & grid & binary & Euler undir. & Euler dir. & F1 & missed H (\%) & missed V (\%) & extra edges & non-lattice & wrong cells & unparsed \\
\midrule
3.8 Flash & clean & 63/66 & 40/40 & 32/32 & 32/32 & 1.000 & 0.5 & 0.0 & 0 & 0 & 0 & 0 \\
 & clean, repeated & 66/66 & 40/40 & 32/32 & 32/32 & 1.000 & 0.0 & 0.0 & 0 & 0 & 0 & 0 \\
 & JPEG $q{=}30$ & 50/66 & 39/40 & 32/32 & 32/32 & 0.996 & 4.2 & 0.1 & 0 & 0 & 4 & 0 \\
 & JPEG $q{=}10$ & 56/66 & 40/40 & 32/32 & 32/32 & 0.999 & 1.4 & 0.1 & 0 & 0 & 0 & 0 \\
 & blur $\sigma{=}1.5$ & 61/66 & 40/40 & 32/32 & 32/32 & 0.999 & 1.0 & 0.0 & 0 & 0 & 0 & 0 \\
 & blur $\sigma{=}3$ & 63/66 & 40/40 & 32/32 & 32/32 & 1.000 & 0.3 & 0.0 & 0 & 0 & 0 & 0 \\
 & perspective 6\% & 64/66 & 39/40 & 32/32 & 32/32 & 1.000 & 0.3 & 0.0 & 0 & 0 & 2 & 0 \\
 & perspective 12\% & 62/66 & 37/40 & 32/32 & 32/32 & 0.995 & 0.4 & 0.1 & 0 & 0 & 24 & 0 \\
 & elastic 3\,px & 65/66 & 40/40 & 32/32 & 32/32 & 1.000 & 0.1 & 0.0 & 0 & 0 & 0 & 0 \\
 & elastic 6\,px & 66/66 & 40/40 & 32/32 & 32/32 & 1.000 & 0.0 & 0.0 & 0 & 0 & 0 & 0 \\
 & half resolution & 61/66 & 40/40 & 32/32 & 32/32 & 0.999 & 1.5 & 0.0 & 0 & 0 & 0 & 0 \\
 & photo-like & 57/66 & 39/40 & 32/32 & 32/32 & 0.998 & 2.2 & 0.0 & 0 & 0 & 8 & 0 \\
\midrule
3.1 Pro & clean & 57/66 & 40/40 & 32/32 & 32/32 & 0.999 & 0.2 & 0.0 & 12 & 0 & 0 & 0 \\
 & clean, repeated & 56/66 & 40/40 & 32/32 & 31/32 & 0.999 & 0.2 & 0.1 & 12 & 0 & 0 & 0 \\
 & JPEG $q{=}30$ & 56/66 & 39/40 & 32/32 & 30/32 & 0.998 & 0.4 & 0.3 & 7 & 0 & 6 & 0 \\
 & JPEG $q{=}10$ & 57/66 & 39/40 & 32/32 & 31/32 & 0.999 & 0.1 & 0.3 & 13 & 0 & 1 & 0 \\
 & blur $\sigma{=}1.5$ & 63/66 & 40/40 & 32/32 & 32/32 & 1.000 & 0.1 & 0.0 & 2 & 0 & 0 & 0 \\
 & blur $\sigma{=}3$ & 59/66 & 38/40 & 32/32 & 31/32 & 0.999 & 0.8 & 0.1 & 1 & 0 & 4 & 0 \\
 & perspective 6\% & 58/66 & 40/40 & 32/32 & 32/32 & 0.999 & 0.2 & 0.0 & 9 & 0 & 0 & 0 \\
 & perspective 12\% & 59/66 & 36/40 & 32/32 & 31/32 & 0.997 & 0.1 & 0.0 & 8 & 0 & 17 & 0 \\
 & elastic 3\,px & 55/66 & 40/40 & 32/32 & 31/32 & 0.999 & 0.3 & 0.0 & 16 & 0 & 0 & 0 \\
 & elastic 6\,px & 57/66 & 40/40 & 32/32 & 32/32 & 0.999 & 0.2 & 0.1 & 10 & 0 & 0 & 0 \\
 & half resolution & 59/66 & 40/40 & 32/32 & 30/32 & 0.999 & 0.1 & 0.1 & 9 & 0 & 0 & 0 \\
 & photo-like & 56/66 & 38/40 & 32/32 & 31/32 & 0.998 & 1.4 & 0.7 & 1 & 0 & 2 & 0 \\
\midrule
3.6 Flash & clean & 64/66 & 40/40 & 32/32 & 32/32 & 1.000 & 0.0 & 0.1 & 1 & 0 & 0 & 0 \\
 & clean, repeated & 64/66 & 40/40 & 32/32 & 32/32 & 1.000 & 0.0 & 0.0 & 2 & 0 & 0 & 0 \\
 & JPEG $q{=}30$ & 64/66 & 40/40 & 32/32 & 32/32 & 1.000 & 0.1 & 0.0 & 2 & 0 & 0 & 0 \\
 & JPEG $q{=}10$ & 62/66 & 39/40 & 32/32 & 32/32 & 0.999 & 0.0 & 0.1 & 4 & 0 & 8 & 0 \\
 & blur $\sigma{=}1.5$ & 64/66 & 40/40 & 32/32 & 32/32 & 1.000 & 0.0 & 0.1 & 1 & 0 & 0 & 0 \\
 & blur $\sigma{=}3$ & 64/66 & 40/40 & 32/32 & 32/32 & 1.000 & 0.2 & 0.0 & 0 & 0 & 0 & 0 \\
 & perspective 6\% & 64/66 & 40/40 & 32/32 & 32/32 & 1.000 & 0.1 & 0.1 & 0 & 0 & 0 & 0 \\
 & perspective 12\% & 63/66 & 35/40 & 32/32 & 32/32 & 0.995 & 0.1 & 0.1 & 1 & 0 & 32 & 0 \\
 & elastic 3\,px & 64/66 & 40/40 & 32/32 & 32/32 & 1.000 & 0.1 & 0.0 & 1 & 0 & 0 & 0 \\
 & elastic 6\,px & 62/66 & 40/40 & 32/32 & 32/32 & 1.000 & 0.2 & 0.1 & 1 & 0 & 0 & 0 \\
 & half resolution & 65/66 & 40/40 & 32/32 & 32/32 & 1.000 & 0.1 & 0.0 & 0 & 0 & 0 & 0 \\
 & photo-like & 64/66 & 40/40 & 32/32 & 31/32 & 1.000 & 0.2 & 0.0 & 2 & 0 & 0 & 0 \\
\midrule
Gemma 4 31B & clean & 41/66 & 40/40 & 30/32 & 28/32 & 0.995 & 0.0 & 0.0 & 58 & 0 & 0 & 0 \\
 & clean, repeated & 41/66 & 40/40 & 30/32 & 28/32 & 0.995 & 0.0 & 0.0 & 58 & 0 & 0 & 0 \\
 & JPEG $q{=}30$ & 42/66 & 38/40 & 30/32 & 27/32 & 0.995 & 0.2 & 0.2 & 46 & 0 & 2 & 0 \\
 & JPEG $q{=}10$ & 40/66 & 37/40 & 32/32 & 24/32 & 0.992 & 0.2 & 0.1 & 62 & 0 & 6 & 0 \\
 & blur $\sigma{=}1.5$ & 42/66 & 36/40 & 31/32 & 29/32 & 0.996 & 0.0 & 0.1 & 49 & 0 & 4 & 0 \\
 & blur $\sigma{=}3$ & 40/66 & 34/40 & 31/32 & 20/32 & 0.990 & 0.0 & 0.3 & 61 & 0 & 9 & 0 \\
 & perspective 6\% & 45/66 & 33/40 & 32/32 & 31/32 & 0.995 & 0.2 & 0.3 & 33 & 0 & 14 & 0 \\
 & perspective 12\% & 50/66 & 27/40 & 32/32 & 28/32 & 0.988 & 0.0 & 0.3 & 25 & 0 & 70 & 0 \\
 & elastic 3\,px & 47/66 & 39/40 & 31/32 & 28/32 & 0.997 & 0.0 & 0.0 & 41 & 0 & 1 & 0 \\
 & elastic 6\,px & 43/66 & 37/40 & 31/32 & 28/32 & 0.996 & 0.0 & 0.3 & 39 & 0 & 3 & 0 \\
 & half resolution & 39/66 & 35/40 & 30/32 & 26/32 & 0.994 & 0.0 & 0.0 & 58 & 0 & 6 & 0 \\
 & photo-like & 46/66 & 34/40 & 30/32 & 28/32 & 0.995 & 0.0 & 0.3 & 34 & 0 & 9 & 0 \\
\bottomrule
\end{tabular}

%% file: gen_v6/tab_v6_d8_code.tex
\begin{tabular}{lrrrrrrrrr}
\toprule
Arm & programs & binary/Euler (not scored) & grid, no list & scorable & $T{=}1,A{=}1$ & $T{=}1,A{=}0$ & $T{=}0,A{=}1$ & $T{=}0,A{=}0$ & $A{=}\#(\hat G)$ among $T{=}0$ \\
\midrule
clean & 392 & 144 & 42 & 206 & 182 & 2 & 4 & 18 & 18/22 \\
clean, repeated & 392 & 144 & 44 & 204 & 179 & 3 & 1 & 21 & 19/22 \\
JPEG $q{=}30$ & 392 & 144 & 42 & 206 & 138 & 2 & 1 & 65 & 62/66 \\
JPEG $q{=}10$ & 392 & 144 & 43 & 205 & 170 & 1 & 2 & 32 & 30/34 \\
blur $\sigma{=}1.5$ & 392 & 144 & 45 & 203 & 175 & 1 & 2 & 25 & 22/27 \\
blur $\sigma{=}3$ & 392 & 144 & 58 & 190 & 167 & 3 & 2 & 18 & 16/20 \\
perspective 6\% & 392 & 144 & 46 & 202 & 185 & 0 & 1 & 16 & 15/17 \\
perspective 12\% & 392 & 144 & 40 & 208 & 188 & 4 & 3 & 13 & 11/16 \\
elastic 3\,px & 392 & 144 & 43 & 205 & 180 & 5 & 4 & 16 & 14/20 \\
elastic 6\,px & 392 & 144 & 45 & 203 & 186 & 7 & 2 & 8 & 7/10 \\
half resolution & 392 & 144 & 52 & 196 & 177 & 2 & 0 & 17 & 17/17 \\
photo-like & 392 & 144 & 45 & 203 & 153 & 3 & 3 & 44 & 43/47 \\
\bottomrule
\end{tabular}

%% file: gen_v6/tab_v6_d8_retest.tex
\begin{tabular}{lrrrrrr}
\toprule
Descriptive comparison & $n$ & first & second & $b/c$ & $p$ & identical text \\
\midrule
3.8 Flash Transcription: clean vs repeated clean & 170 & 167 & 170 & 0/3 & 0.25 & 30/170 \\
3.8 Flash Visual CoT: clean vs repeated clean & 392 & 257 & 244 & 28/15 & 0.066 & 20/392 \\
3.8 Flash Visual$\to$Code: clean vs repeated clean & 392 & 364 & 361 & 19/16 & 0.736 & 2/392 \\
3.1 Pro Transcription: clean vs repeated clean & 170 & 161 & 159 & 2/0 & 0.5 & 100/170 \\
3.1 Pro Visual CoT: clean vs repeated clean & 392 & 214 & 223 & 10/19 & 0.136 & 134/392 \\
3.6 Flash Transcription: clean vs repeated clean & 170 & 168 & 168 & 1/1 & 1 & 16/170 \\
3.6 Flash Visual CoT: clean vs repeated clean & 392 & 210 & 214 & 20/24 & 0.652 & 2/392 \\
Gemma 4 31B Transcription: clean vs repeated clean & 170 & 139 & 139 & 0/0 & 1 & 170/170 \\
\midrule
3.8 Flash Transcription: Exp A vs D8 clean & 170 & 167 & 167 & 1/1 & 1 & -- \\
3.8 Flash Visual CoT: p2\_v2 vs D8 clean & 392 & 248 & 257 & 16/25 & 0.211 & -- \\
3.8 Flash Visual$\to$Code: p2\_v2 vs D8 clean & 392 & 364 & 364 & 17/17 & 1 & -- \\
3.1 Pro Visual CoT: p2\_v2 vs D8 clean & 202 & 118 & 115 & 9/6 & 0.607 & -- \\
3.6 Flash Visual CoT: p2\_v2 vs D8 clean & 391 & 205 & 210 & 21/26 & 0.56 & -- \\
\midrule
3.8 Flash Transcription: JPEG $q{=}30$ vs JPEG $q{=}10$ & 170 & 153 & 160 & 4/11 & 0.118 & -- \\
3.8 Flash Transcription: blur $\sigma{=}1.5$ vs blur $\sigma{=}3$ & 170 & 165 & 167 & 2/4 & 0.688 & -- \\
3.8 Flash Transcription: perspective 6\% vs perspective 12\% & 170 & 167 & 163 & 6/2 & 0.289 & -- \\
3.8 Flash Transcription: elastic 3\,px vs elastic 6\,px & 170 & 169 & 170 & 0/1 & 1 & -- \\
3.1 Pro Transcription: JPEG $q{=}30$ vs JPEG $q{=}10$ & 170 & 157 & 159 & 6/8 & 0.791 & -- \\
3.1 Pro Transcription: blur $\sigma{=}1.5$ vs blur $\sigma{=}3$ & 170 & 167 & 160 & 10/3 & 0.0923 & -- \\
3.1 Pro Transcription: perspective 6\% vs perspective 12\% & 170 & 162 & 158 & 7/3 & 0.344 & -- \\
3.1 Pro Transcription: elastic 3\,px vs elastic 6\,px & 170 & 158 & 161 & 4/7 & 0.549 & -- \\
3.6 Flash Transcription: JPEG $q{=}30$ vs JPEG $q{=}10$ & 170 & 168 & 165 & 3/0 & 0.25 & -- \\
3.6 Flash Transcription: blur $\sigma{=}1.5$ vs blur $\sigma{=}3$ & 170 & 168 & 168 & 2/2 & 1 & -- \\
3.6 Flash Transcription: perspective 6\% vs perspective 12\% & 170 & 168 & 162 & 8/2 & 0.109 & -- \\
3.6 Flash Transcription: elastic 3\,px vs elastic 6\,px & 170 & 168 & 166 & 4/2 & 0.688 & -- \\
Gemma 4 31B Transcription: JPEG $q{=}30$ vs JPEG $q{=}10$ & 170 & 137 & 133 & 11/7 & 0.481 & -- \\
Gemma 4 31B Transcription: blur $\sigma{=}1.5$ vs blur $\sigma{=}3$ & 170 & 138 & 125 & 21/8 & 0.0241 & -- \\
Gemma 4 31B Transcription: perspective 6\% vs perspective 12\% & 170 & 141 & 137 & 14/10 & 0.541 & -- \\
Gemma 4 31B Transcription: elastic 3\,px vs elastic 6\,px & 170 & 145 & 139 & 10/4 & 0.18 & -- \\
\bottomrule
\end{tabular}

%% file: gen_v6/tab_v6_d8_cost.tex
\begin{tabular}{llrrrrll}
\toprule
Model & Kind & ok keys & thoughts (clean) & thoughts (degraded) & USD & UTC span & model version \\
\midrule
3.8 Flash & Transcription & 2040 & 793 & 853 & 10.74 & 2026-10-08 19:01 UTC to 2026-10-09 04:10 UTC & gemini-3.8-flash \\
3.8 Flash & Visual CoT & 4704 & 4323 & 4329 & 93.54 & 2026-10-08 19:02 UTC to 2026-10-09 17:51 UTC & gemini-3.8-flash \\
3.8 Flash & Visual$\to$Code & 4704 & 1039 & 1177 & 35.74 & 2026-10-08 19:04 UTC to 2026-10-09 17:48 UTC & gemini-3.8-flash \\
3.1 Pro & Transcription & 2040 & 782 & 695 & 28.80 & 2026-10-08 19:01 UTC to 2026-10-09 00:19 UTC & gemini-3.1-pro-preview \\
3.1 Pro & Visual CoT & 4704 & 3851 & 3977 & 262.36 & 2026-10-08 19:03 UTC to 2026-10-09 05:36 UTC & gemini-3.1-pro-preview \\
3.6 Flash & Transcription & 2040 & 924 & 968 & 9.37 & 2026-10-08 19:01 UTC to 2026-10-09 00:17 UTC & gemini-3.6-flash \\
3.6 Flash & Visual CoT & 4704 & 5585 & 5259 & 87.73 & 2026-10-08 19:02 UTC to 2026-10-09 05:43 UTC & gemini-3.6-flash \\
Gemma 4 31B & Transcription & 2040 & 0 & 0 & 0.00 & 2026-10-08 19:01 UTC to 22:33 UTC & gemma-4-31b-it \\
\bottomrule
\end{tabular}

%% file: gen_v8p/gen_v6/tab_v6_t6_h1.tex
\begin{tabular}{llrrrrrrrrl}
\toprule
Condition & role & complete units & discordant & ST is the exact one & $\hat\beta_x$ & 95\% CI & one-sided $p$ & Holm $p$ & failed fits & H1 \\
\midrule
Visual & co-primary & 640 & 122 & 73/122 & $-$0.61 & $[-1.88, +0.38]$ & 0.908 & 1 & 0/4000 & not supported \\
Text & co-primary & 639 & 144 & 86/144 & $-$0.51 & $[-1.60, +0.35]$ & 0.88 & 1 & 0/4000 & not supported \\
Visual, $x_c$ uncentred (plan text) & sensitivity & 640 & 122 & 73/122 & $-$0.61 & $[-1.88, +0.38]$ & 0.908 & 1 & 1/4000 & not supported \\
Text, $x_c$ uncentred (plan text) & sensitivity & 639 & 144 & 86/144 & $-$0.51 & $[-1.60, +0.35]$ & 0.88 & 1 & 0/4000 & not supported \\
Visual, both transcriptions exact & descriptive & 543 & 105 & 62/105 & $-$0.57 & $[-1.97, +0.51]$ & 0.878 & -- & 1/4000 & -- \\
\bottomrule
\end{tabular}

%% file: gen_v6/tab_v6_t6_h1_fitted.tex
\begin{tabular}{llllrlrr}
\toprule
Condition & Ladder & lowest rung & fitted ST share (\%) & 95\% CI & highest rung & fitted ST share (\%) & 95\% CI \\
\midrule
Visual & 3.8 Flash & Gemini 3.8 Flash (low) & 77 & $[65, 89]$ & Gemini 3.8 Flash (high) & 69 & $[52, 84]$ \\
Visual & 3.6 Flash & Gemini 3.6 Flash (minimal) & 47 & $[19, 84]$ & Gemini 3.6 Flash (high) & 29 & $[12, 46]$ \\
Visual & 3.1 Pro & Gemini 3.1 Pro (low) & 72 & $[53, 89]$ & Gemini 3.1 Pro (high) & 61 & $[42, 79]$ \\
Text & 3.8 Flash & Gemini 3.8 Flash (low) & 80 & $[67, 92]$ & Gemini 3.8 Flash (high) & 72 & $[56, 86]$ \\
Text & 3.6 Flash & Gemini 3.6 Flash (minimal) & 53 & $[30, 79]$ & Gemini 3.6 Flash (high) & 38 & $[22, 55]$ \\
Text & 3.1 Pro & Gemini 3.1 Pro (low) & 66 & $[45, 86]$ & Gemini 3.1 Pro (high) & 55 & $[37, 71]$ \\
\bottomrule
\end{tabular}

%% file: gen_v6/tab_v6_t6_cells.tex
\begin{tabular}{llrrrrrrl}
\toprule
Configuration & Cond. & complete units & ST exact (\%) & \#M exact (\%) & $b/c$ & $p$ & Holm $p$ & ST share of discordant (\%) \\
\midrule
Gemini 3.8 Flash (low) & Visual & 80 & 45 & 31 & 16/5 & 0.0266 & 0.399 & 76 \\
Gemini 3.8 Flash (medium) & Visual & 80 & 59 & 46 & 14/4 & 0.0309 & 0.432 & 78 \\
Gemini 3.8 Flash (high) & Visual & 80 & 79 & 72 & 11/6 & 0.332 & 1 & 65 \\
Gemini 3.6 Flash (minimal) & Visual & 80 & 1 & 4 & 1/3 & 0.625 & 1 & 25 \\
Gemini 3.6 Flash (low) & Visual & 80 & 18 & 22 & 2/6 & 0.289 & 1 & 25 \\
Gemini 3.6 Flash (high) & Visual & 80 & 35 & 42 & 8/14 & 0.286 & 1 & 36 \\
Gemini 3.1 Pro (low) & Visual & 80 & 26 & 16 & 10/2 & 0.0386 & 0.501 & 83 \\
Gemini 3.1 Pro (high) & Visual & 80 & 55 & 52 & 11/9 & 0.824 & 1 & 55 \\
pooled (Visual) & Visual & 640 & 40 & 36 & 73/49 & 0.0555 & -- & 60 $[50, 69]$ \\
\midrule
Gemini 3.8 Flash (low) & Text & 80 & 41 & 19 & 22/4 & 0.000534 & 0.00854 & 85 \\
Gemini 3.8 Flash (medium) & Text & 80 & 68 & 59 & 9/2 & 0.0654 & 0.785 & 82 \\
Gemini 3.8 Flash (high) & Text & 79 & 87 & 84 & 9/6 & 0.607 & 1 & 60 \\
Gemini 3.6 Flash (minimal) & Text & 80 & 10 & 9 & 6/5 & 1 & 1 & 55 \\
Gemini 3.6 Flash (low) & Text & 80 & 24 & 26 & 9/11 & 0.824 & 1 & 45 \\
Gemini 3.6 Flash (high) & Text & 80 & 35 & 42 & 8/14 & 0.286 & 1 & 36 \\
Gemini 3.1 Pro (low) & Text & 80 & 21 & 20 & 7/6 & 1 & 1 & 54 \\
Gemini 3.1 Pro (high) & Text & 80 & 56 & 49 & 16/10 & 0.327 & 1 & 62 \\
pooled (Text) & Text & 639 & 43 & 38 & 86/58 & 0.037 & -- & 60 $[51, 69]$ \\
\bottomrule
\end{tabular}

%% file: gen_v8p/gen_v6/tab_v6_t6_ladder.tex
\begin{tabular}{lllrrrrrr}
\toprule
Ladder & Cond. & rates over & ST lowest$\to$highest (\%) & ST rise & \#M lowest$\to$highest (\%) & \#M rise & ST rise $-$ \#M rise & 95\% CI \\
\midrule
3.8 Flash & Visual & returned & 45$\to$79 & +34 & 31$\to$72 & +41 & $-$8 & $[-22, +6]$ \\
3.8 Flash & Visual & missing = wrong & 45$\to$79 & +34 & 31$\to$72 & +41 & $-$8 & $[-22, +6]$ \\
3.8 Flash & Text & returned & 41$\to$87 & +46 & 19$\to$82 & +64 & $-$18 & $[-34, -3]$ \\
3.8 Flash & Text & missing = wrong & 41$\to$86 & +45 & 19$\to$82 & +64 & $-$19 & $[-34, -4]$ \\
3.6 Flash & Visual & returned & 1$\to$35 & +34 & 4$\to$42 & +39 & $-$5 & $[-18, +8]$ \\
3.6 Flash & Visual & missing = wrong & 1$\to$35 & +34 & 4$\to$42 & +39 & $-$5 & $[-18, +8]$ \\
3.6 Flash & Text & returned & 10$\to$35 & +25 & 9$\to$42 & +34 & $-$9 & $[-21, +5]$ \\
3.6 Flash & Text & missing = wrong & 10$\to$35 & +25 & 9$\to$42 & +34 & $-$9 & $[-21, +5]$ \\
3.1 Pro & Visual & returned & 26$\to$55 & +29 & 16$\to$52 & +36 & $-$8 & $[-21, +6]$ \\
3.1 Pro & Visual & missing = wrong & 26$\to$55 & +29 & 16$\to$52 & +36 & $-$8 & $[-21, +6]$ \\
3.1 Pro & Text & returned & 21$\to$56 & +35 & 20$\to$49 & +29 & +6 & $[-10, +21]$ \\
3.1 Pro & Text & missing = wrong & 21$\to$56 & +35 & 20$\to$49 & +29 & +6 & $[-10, +21]$ \\
\bottomrule
\end{tabular}

%% file: gen_v6/tab_v6_t6_m50.tex
\begin{tabular}{lrrrr}
\toprule
Configuration & Visual \#ST & Visual \#M & Text \#ST & Text \#M \\
\midrule
Gemini 3.8 Flash (low) & 5.6 & 4.6 & 5.4 & 3.5 \\
Gemini 3.8 Flash (medium) & 6.8 & 5.8 & 7.4 & 6.7 \\
Gemini 3.8 Flash (high) & 9.4 & 8.0 & 10.3 & 8.7 \\
Gemini 3.6 Flash (minimal) & $<2$ & $<2$ & 2.7 & 2.6 \\
Gemini 3.6 Flash (low) & 3.4 & 3.8 & 3.5 & 4.1 \\
Gemini 3.6 Flash (high) & 4.8 & 5.5 & 4.9 & 5.5 \\
Gemini 3.1 Pro (low) & 4.2 & 3.3 & 3.8 & 3.7 \\
Gemini 3.1 Pro (high) & 6.5 & 6.3 & 6.6 & 6.0 \\
\bottomrule
\end{tabular}

%% file: gen_v8p/gen_v6/tab_v6_t6_oos.tex
\begin{tabular}{llrrrrrrrrrrr}
\toprule
Configuration & Cond. & $n$ & AUC mag. & AUC +class & cal. slope mag. & cal. int. mag. & cal. slope +class & cal. int. +class & log-loss mag. & log-loss +class & $\Delta$ & 95\% CI \\
\midrule
Gemini 3.8 Flash (low) & Visual & 160 & 0.85 & 0.87 & 0.76 & 1.46 & 0.80 & 1.49 & 0.678 & 0.647 & $-$0.032 & -- \\
Gemini 3.8 Flash (low) & Text & 160 & 0.87 & 0.90 & 0.80 & 1.28 & 0.90 & 1.40 & 0.536 & 0.491 & $-$0.045 & -- \\
Gemini 3.8 Flash (medium) & Visual & 160 & 0.90 & 0.90 & 1.21 & 1.84 & 1.15 & 1.55 & 0.564 & 0.513 & $-$0.050 & -- \\
Gemini 3.8 Flash (medium) & Text & 160 & 0.94 & 0.93 & 2.01 & 3.71 & 1.69 & 2.93 & 0.627 & 0.578 & $-$0.049 & -- \\
Gemini 3.8 Flash (high) & Visual & 160 & 0.82 & 0.82 & 1.01 & 1.33 & 0.96 & 1.38 & 0.551 & 0.560 & +0.009 & -- \\
Gemini 3.8 Flash (high) & Text & 159 & 0.88 & 0.88 & 1.51 & 2.69 & 1.48 & 2.66 & 0.620 & 0.612 & $-$0.008 & -- \\
Gemini 3.6 Flash (minimal) & Visual & 160 & 0.88 & 0.91 & 0.59 & $-$0.23 & 0.61 & $-$0.07 & 0.105 & 0.105 & $-$0.000 & -- \\
Gemini 3.6 Flash (minimal) & Text & 160 & 0.91 & 0.91 & 1.05 & 1.53 & 1.05 & 1.53 & 0.240 & 0.239 & $-$0.001 & -- \\
Gemini 3.6 Flash (low) & Visual & 160 & 0.93 & 0.93 & 1.09 & 1.92 & 1.09 & 1.93 & 0.384 & 0.385 & +0.001 & -- \\
Gemini 3.6 Flash (low) & Text & 160 & 0.86 & 0.86 & 0.97 & 1.03 & 0.97 & 0.95 & 0.447 & 0.445 & $-$0.003 & -- \\
Gemini 3.6 Flash (high) & Visual & 160 & 0.90 & 0.89 & 1.19 & 1.34 & 1.10 & 1.08 & 0.451 & 0.458 & +0.007 & -- \\
Gemini 3.6 Flash (high) & Text & 160 & 0.89 & 0.88 & 0.94 & 1.40 & 0.88 & 1.18 & 0.537 & 0.538 & +0.001 & -- \\
Gemini 3.1 Pro (low) & Visual & 160 & 0.92 & 0.91 & 0.87 & 1.35 & 0.81 & 1.29 & 0.394 & 0.410 & +0.016 & -- \\
Gemini 3.1 Pro (low) & Text & 160 & 0.96 & 0.95 & 0.79 & 2.38 & 0.77 & 2.37 & 0.460 & 0.470 & +0.009 & -- \\
Gemini 3.1 Pro (high) & Visual & 160 & 0.90 & 0.89 & 1.11 & 2.05 & 1.11 & 2.00 & 0.640 & 0.626 & $-$0.014 & -- \\
Gemini 3.1 Pro (high) & Text & 160 & 0.86 & 0.86 & 1.17 & 0.93 & 1.11 & 0.78 & 0.523 & 0.510 & $-$0.014 & -- \\
\midrule
pooled & Visual & 1280 & 0.92 & 0.92 & 0.97 & 1.52 & 0.96 & 1.45 & 0.471 & 0.463 & $-$0.008 & $[-0.016, -0.001]$ \\
pooled & Text & 1279 & 0.92 & 0.92 & 1.00 & 1.59 & 0.99 & 1.50 & 0.499 & 0.485 & $-$0.014 & $[-0.021, -0.007]$ \\
pooled & both & 2559 & 0.92 & 0.92 & 0.99 & 1.56 & 0.97 & 1.47 & 0.485 & 0.474 & $-$0.011 & $[-0.016, -0.005]$ \\
\bottomrule
\end{tabular}

%% file: gen_v6/tab_v6_t6_oos_strata.tex
\begin{tabular}{llrrrr}
\toprule
Cond. & stratum ($\log_{10}$ answer) & $n$ & observed exact (\%) & predicted, mag. (\%) & predicted, +class (\%) \\
\midrule
Visual & [2,3) & 160 & 78 & 60 & 60 \\
Visual & [3,4) & 160 & 71 & 43 & 44 \\
Visual & [4,5) & 160 & 56 & 31 & 32 \\
Visual & [5,6) & 160 & 32 & 16 & 17 \\
Visual & [6,7) & 160 & 29 & 10 & 11 \\
Visual & [7,8) & 160 & 21 & 8 & 9 \\
Visual & [8,9) & 160 & 11 & 4 & 4 \\
Visual & [9,10.5) & 160 & 4 & 2 & 2 \\
Text & [2,3) & 160 & 85 & 61 & 61 \\
Text & [3,4) & 160 & 68 & 43 & 44 \\
Text & [4,5) & 160 & 62 & 32 & 33 \\
Text & [5,6) & 160 & 38 & 18 & 20 \\
Text & [6,7) & 160 & 24 & 12 & 13 \\
Text & [7,8) & 160 & 25 & 10 & 11 \\
Text & [8,9) & 160 & 17 & 5 & 5 \\
Text & [9,10.5) & 159 & 6 & 2 & 2 \\
\bottomrule
\end{tabular}

%% file: gen_v8p/gen_v6/tab_v6_t6_tokens.tex
\begin{tabular}{llrrr}
\toprule
Configuration & Cond. & complete units & median $\log_{10}\frac{t_{ST}+1}{t_{M}+1}$ & Wilcoxon $p$ \\
\midrule
Gemini 3.8 Flash (low) & Visual & 80 & +0.197 & 0.000463 \\
Gemini 3.8 Flash (medium) & Visual & 80 & $-$0.001 & 0.274 \\
Gemini 3.8 Flash (high) & Visual & 80 & $-$0.013 & 0.106 \\
Gemini 3.6 Flash (minimal) & Visual & 80 & $-$0.070 & $<10^{-4}$ \\
Gemini 3.6 Flash (low) & Visual & 80 & $-$0.003 & 0.481 \\
Gemini 3.6 Flash (high) & Visual & 80 & $-$0.012 & 0.256 \\
Gemini 3.1 Pro (low) & Visual & 80 & +0.127 & $<10^{-4}$ \\
Gemini 3.1 Pro (high) & Visual & 80 & $-$0.001 & 0.645 \\
pooled & Visual & 640 & $-$0.000 & 0.132 \\
\midrule
Gemini 3.8 Flash (low) & Text & 80 & +0.096 & 0.0538 \\
Gemini 3.8 Flash (medium) & Text & 80 & $-$0.002 & 0.659 \\
Gemini 3.8 Flash (high) & Text & 79 & $-$0.048 & 0.00031 \\
Gemini 3.6 Flash (minimal) & Text & 80 & $-$0.063 & 0.0369 \\
Gemini 3.6 Flash (low) & Text & 80 & $-$0.054 & 0.625 \\
Gemini 3.6 Flash (high) & Text & 80 & $-$0.030 & 0.00211 \\
Gemini 3.1 Pro (low) & Text & 80 & +0.141 & $<10^{-4}$ \\
Gemini 3.1 Pro (high) & Text & 80 & $-$0.014 & 0.0365 \\
pooled & Text & 639 & $-$0.014 & 0.708 \\
\bottomrule
\end{tabular}

%% file: gen_v6/tab_v6_t6_copy_twins.tex
\begin{tabular}{lrrrr}
\toprule
Configuration & 3-5 digits & 6-8 digits & 9-11 digits & all \\
\midrule
Gemini 3.8 Flash (low) & 59/59 & 61/61 & 40/40 & 160/160 \\
Gemini 3.6 Flash (minimal) & 59/59 & 61/61 & 40/40 & 160/160 \\
Gemini 3.1 Pro (low) & 59/59 & 61/61 & 40/40 & 160/160 \\
\bottomrule
\end{tabular}

%% file: gen_v6/tab_v6_t6_copy_xl.tex
\begin{tabular}{lrrrrr}
\toprule
Configuration & 5-10 digits & 13-20 digits & 21-30 digits & 31-38 digits & all \\
\midrule
Gemini 3.8 Flash (low) & 12/12 & 18/18 & 23/23 & 7/7 & 60/60 \\
Gemini 3.6 Flash (minimal) & 12/12 & 18/18 & 23/23 & 7/7 & 60/60 \\
Gemini 3.1 Pro (low) & 12/12 & 18/18 & 23/23 & 7/7 & 60/60 \\
\bottomrule
\end{tabular}

%% file: gen_v6/tab_v6_t6_copy_miss.tex
\begin{tabular}{lllrrrl}
\toprule
Set & Configuration & instance & task & true digits & returned digits & finish \\
\midrule
none &  &  &  &  &  &  \\
\bottomrule
\end{tabular}

%% file: gen_v6/tab_v6_t6_algo.tex
\begin{tabular}{llrrrrrrrr}
\toprule
Configuration & Task & pairs & Algo exact (\%) & Text exact (\%) & $b/c$ & $p$ & MAX\_TOKENS A/T & median tokens A/T & median s A/T \\
\midrule
Gemini 3.8 Flash (low) & \#ST & 80 & 36 & 41 & 8/12 & 0.503 & 0/0 & 3896/3978 & 37/35 \\
Gemini 3.8 Flash (low) & \#M & 80 & 19 & 19 & 7/7 & 1 & 0/0 & 3613/2982 & 28/33 \\
Gemini 3.8 Flash (medium) & \#ST & 80 & 70 & 68 & 9/7 & 0.804 & 11/9 & 21606/22573 & 154/164 \\
Gemini 3.8 Flash (medium) & \#M & 80 & 39 & 59 & 3/19 & 0.000855 & 18/9 & 22295/17650 & 141/149 \\
Gemini 3.8 Flash (high) & \#ST & 79 & 87 & 87 & 7/7 & 1 & 4/2 & 42587/35547 & 307/247 \\
Gemini 3.8 Flash (high) & \#M & 80 & 68 & 82 & 1/13 & 0.00183 & 17/0 & 55246/41018 & 313/257 \\
\bottomrule
\end{tabular}

%% file: gen_v6/tab_v6_t6_work_st.tex
\begin{tabular}{llrrrrr}
\toprule
Cond. & Configuration & $[2.5, 3.0)$ & $[3.0, 3.5)$ & $[3.5, 4.0)$ & $[4.0, 4.5)$ & tokens/work \\
\midrule
Algo & Gemini 3.8 Flash (low) & 12/15 (0) & 9/15 (0) & 8/30 (0) & 0/20 (0) & 1.5 \\
Algo & Gemini 3.8 Flash (medium) & 15/15 (0) & 14/15 (0) & 24/30 (5) & 3/20 (6) & 4.5 \\
Algo & Gemini 3.8 Flash (high) & 14/15 (1) & 14/15 (1) & 28/30 (0) & 13/20 (2) & 8.3 \\
Text & Gemini 3.8 Flash (low) & 13/15 (0) & 13/15 (0) & 5/30 (0) & 2/20 (0) & 1.2 \\
Text & Gemini 3.8 Flash (medium) & 15/15 (0) & 15/15 (0) & 22/30 (4) & 2/20 (5) & 4.6 \\
Text & Gemini 3.8 Flash (high) & 15/15 (0) & 14/15 (0) & 28/30 (0) & 12/19 (2) & 7.2 \\
\bottomrule
\end{tabular}

%% file: gen_v6/tab_v6_t6_work_am.tex
\begin{tabular}{llrrrrr}
\toprule
Cond. & Configuration & $[2.0, 2.5)$ & $[2.5, 3.0)$ & $[3.0, 3.5)$ & $[3.5, 4.0)$ & tokens/work \\
\midrule
Algo & Gemini 3.8 Flash (low) & 11/16 (0) & 2/23 (0) & 2/37 (0) & 0/4 (0) & 3.6 \\
Algo & Gemini 3.8 Flash (medium) & 13/16 (0) & 13/23 (5) & 5/37 (13) & 0/4 (0) & 26.4 \\
Algo & Gemini 3.8 Flash (high) & 16/16 (0) & 22/23 (3) & 15/37 (13) & 1/4 (1) & 55.6 \\
Text & Gemini 3.8 Flash (low) & 7/16 (0) & 8/23 (0) & 0/37 (0) & 0/4 (0) & 3.0 \\
Text & Gemini 3.8 Flash (medium) & 16/16 (0) & 18/23 (3) & 12/37 (5) & 1/4 (1) & 23.5 \\
Text & Gemini 3.8 Flash (high) & 16/16 (0) & 22/23 (0) & 27/37 (0) & 1/4 (0) & 43.3 \\
\bottomrule
\end{tabular}

%% file: gen_v8p/gen_v6/tab_v6_t6_workfit.tex
\begin{tabular}{lllrrllll}
\toprule
Task & Configuration & Cond. & $n$ & exact & ans. & work & both: ans. & both: work \\
\midrule
\#ST & Gemini 3.8 Flash (low) & Algo & 80 & 29 & $-$0.86 $[-1.39, -0.58]$ & $-$3.10 $[-4.28, -1.99]$ & $-$2.00 $[-3.64, -0.20]$ & +4.17 $[-3.17, +9.53]$ \\
\#ST & Gemini 3.8 Flash (low) & Text & 80 & 33 & $-$0.95 $[-1.74, -0.62]$ & $-$3.66 $[-4.28, -2.26]$ & $-$0.71 $[-3.73, +0.21]$ & $-$0.93 $[-4.23, +9.17]$ \\
\#ST & Gemini 3.8 Flash (medium) & Algo & 80 & 56 & $-$1.14 $[-1.86, -0.72]$ & -- & $-$2.38 $[-3.87, -0.73]$ & +5.81 $[-1.55, +10.55]$ \\
\#ST & Gemini 3.8 Flash (medium) & Text & 80 & 54 & $-$1.62 $[-2.02, -1.19]$ & -- & -- & -- \\
\#ST & Gemini 3.8 Flash (high) & Algo & 80 & 69 & $-$0.44 $[-1.00, -0.12]$ & $-$1.88 $[-3.22, -0.39]$ & $-$0.96 $[-2.31, +0.37]$ & +2.40 $[-3.36, +7.93]$ \\
\#ST & Gemini 3.8 Flash (high) & Text & 79 & 69 & $-$0.59 $[-1.23, -0.26]$ & $-$3.09 $[-3.37, -1.07]$ & -- & -- \\
\#M & Gemini 3.8 Flash (low) & Algo & 80 & 15 & $-$1.21 $[-3.55, -0.68]$ & $-$4.99 $[-6.13, -3.01]$ & $-$0.66 $[-4.18, +0.23]$ & $-$2.43 $[-5.97, +4.47]$ \\
\#M & Gemini 3.8 Flash (low) & Text & 80 & 15 & $-$1.34 $[-2.38, -0.94]$ & $-$3.62 $[-5.54, -2.34]$ & $-$3.82 $[-6.90, -2.02]$ & +7.99 $[+2.45, +14.04]$ \\
\#M & Gemini 3.8 Flash (medium) & Algo & 80 & 31 & $-$0.80 $[-1.31, -0.53]$ & $-$3.79 $[-5.15, -2.53]$ & $-$0.72 $[-1.67, -0.10]$ & $-$0.47 $[-3.65, +3.02]$ \\
\#M & Gemini 3.8 Flash (medium) & Text & 80 & 47 & $-$1.57 $[-2.17, -1.15]$ & $-$4.85 $[-4.91, -3.07]$ & $-$2.86 $[-4.66, -1.99]$ & +7.15 $[+1.91, +13.71]$ \\
\#M & Gemini 3.8 Flash (high) & Algo & 80 & 54 & $-$1.07 $[-1.81, -0.70]$ & -- & $-$1.19 $[-2.49, -0.56]$ & +0.82 $[-2.51, +5.45]$ \\
\#M & Gemini 3.8 Flash (high) & Text & 80 & 66 & $-$1.34 $[-1.71, -0.85]$ & -- & $-$1.89 $[-3.71, -1.01]$ & +4.04 $[-0.30, +11.60]$ \\
\bottomrule
\end{tabular}

%% file: gen_v6/tab_v6_t6_acct.tex
\begin{tabular}{llllr}
\toprule
Part & Set & Cond. & Configuration & returned \\
\midrule
A & twins & Text & Gemini 3.6 Flash (high) & 160/160 \\
A & twins & Text & Gemini 3.6 Flash (low) & 160/160 \\
A & twins & Text & Gemini 3.6 Flash (minimal) & 160/160 \\
A & twins & Text & Gemini 3.8 Flash (high) & 159/160 \\
A & twins & Text & Gemini 3.8 Flash (low) & 160/160 \\
A & twins & Text & Gemini 3.8 Flash (medium) & 160/160 \\
A & twins & Text & Gemini 3.1 Pro (high) & 160/160 \\
A & twins & Text & Gemini 3.1 Pro (low) & 160/160 \\
A & twins & Visual & Gemini 3.6 Flash (high) & 160/160 \\
A & twins & Visual & Gemini 3.6 Flash (low) & 160/160 \\
A & twins & Visual & Gemini 3.6 Flash (minimal) & 160/160 \\
A & twins & Visual & Gemini 3.8 Flash (high) & 160/160 \\
A & twins & Visual & Gemini 3.8 Flash (low) & 160/160 \\
A & twins & Visual & Gemini 3.8 Flash (medium) & 160/160 \\
A & twins & Visual & Gemini 3.1 Pro (high) & 160/160 \\
A & twins & Visual & Gemini 3.1 Pro (low) & 160/160 \\
B & twins & Algo & Gemini 3.8 Flash (high) & 160/160 \\
B & twins & Algo & Gemini 3.8 Flash (low) & 160/160 \\
B & twins & Algo & Gemini 3.8 Flash (medium) & 160/160 \\
B & twins & Copy & Gemini 3.6 Flash (minimal) & 160/160 \\
B & twins & Copy & Gemini 3.8 Flash (low) & 160/160 \\
B & twins & Copy & Gemini 3.1 Pro (low) & 160/160 \\
B & xl & Copy & Gemini 3.6 Flash (minimal) & 60/60 \\
B & xl & Copy & Gemini 3.8 Flash (low) & 60/60 \\
B & xl & Copy & Gemini 3.1 Pro (low) & 60/60 \\
\bottomrule
\end{tabular}

%% file: gen_v5/tab_v5_c503_groups.tex
\begin{tabular}{lrrrrrrr}
\toprule
Rerun group (\texttt{3.8-flash}) & requested & returned & exact & $\GapL$ & $\shP$ & median $\log_{10}$ answer (ret./req.) & USD \\
\midrule
high, Visual (lost) & 46 & 46 & 23/46 & 6/15 & 16/30 & 3.8 / 3.8 & 7.44 \\
high, Visual (drift control) & 36 & 36 & 21/36 & 8/12 & 12/23 & 4.0 / 4.0 & 5.36 \\
high, Text (lost) & 44 & 44 & 26/44 & 12/18 & 14/26 & 4.3 / 4.3 & 6.69 \\
medium, Visual (lost) & 15 & 15 & 8/15 & 3/5 & 5/10 & 3.1 / 3.1 & 1.22 \\
high, Visual+Hint (lost) & 38 & 38 & 18/38 & 9/14 & 9/24 & 4.0 / 4.0 & 6.02 \\
\bottomrule
\end{tabular}

%% file: gen_v5/tab_v5_c503_m50.tex
\begin{tabular}{llrrrrr}
\toprule
\texttt{3.8-flash} & version & $n$ & $\log_{10} m_{50}$ & $\GapL$ & $\shP$ & tokens (k) \\
\midrule
high, Visual & returned only (paper) & 117 & 7.1 & 8.4 & 5.9 & 27.4 \\
 & merged with the rerun & 162 & 6.1 & 7.2 & 5.3 & 31.9 \\
 & merged, still lost = wrong & 162 & 6.1 & 7.2 & 5.3 & 31.9 \\
\midrule
high, Text & returned only (paper) & 118 & 6.9 & 8.0 & 6.1 & 26.4 \\
 & merged with the rerun & 162 & 6.4 & 7.5 & 5.6 & 30.2 \\
 & merged, still lost = wrong & 162 & 6.4 & 7.5 & 5.6 & 30.2 \\
\midrule
medium, Visual & returned only (paper) & 297 & 4.5 & 5.9 & 3.7 & 13.2 \\
 & merged with the rerun & 312 & 4.5 & 5.9 & 3.7 & 13.6 \\
 & merged, still lost = wrong & 312 & 4.5 & 5.9 & 3.7 & 13.6 \\
\bottomrule
\end{tabular}

%% file: gen_v5/tab_v5_c503_slope.tex
\begin{tabular}{lrrrrrrr}
\toprule
version & slope & 95\% CI & $p$ & $R^2$ & leave-one-out & \texttt{3.8-flash} & \texttt{3.6-flash} \\
\midrule
returned only (paper) & 3.5 & $[2.1, 5.0]$ & $1.5\times 10^{-4}$ & 0.71 & 2.8--3.8 & 5.6 & 2.4 \\
merged with the rerun & 3.3 & $[2.1, 4.4]$ & $4.5\times 10^{-5}$ & 0.76 & 2.8--3.5 & 3.9 & 2.4 \\
merged, still lost = wrong & 3.3 & $[2.1, 4.4]$ & $4.5\times 10^{-5}$ & 0.76 & 2.8--3.5 & 3.9 & 2.4 \\
\bottomrule
\end{tabular}

%% file: gen_v6/tab_v6_f6_groups.tex
\begin{tabular}{lrrrr}
\toprule
group / configuration / condition & requested & returned & exact & USD \\
\midrule
lost / g36\_high / text & 1 & 1 & 1/1 & 0.04 \\
lost / g36\_low / visual & 1 & 1 & 0/1 & 0.05 \\
lost / g37\_low / text & 3 & 3 & 2/3 & 0.08 \\
lost / g37\_low / visual & 8 & 8 & 5/8 & 0.23 \\
lost / g38\_high / text & 35 & 35 & 0/35 & 6.41 \\
lost / g38\_high / visual & 23 & 23 & 0/23 & 3.88 \\
lost / g38\_low / hint & 1 & 1 & 1/1 & 0.05 \\
lost / g38\_low / text & 1 & 1 & 0/1 & 0.06 \\
lost / g38\_med / text & 7 & 7 & 2/7 & 0.64 \\
lost / gm31\_high / car & 4 & 0 & -- & 0.00 \\
lost / gm31\_high / code & 7 & 0 & -- & 0.00 \\
lost / gm31\_high / hint & 1 & 0 & -- & 0.00 \\
lost / gm31\_high / text & 1 & 1 & 0/1 & 0.00 \\
lost / gm31\_high / visual & 1 & 0 & -- & 0.00 \\
lost / gm31\_min / text & 1 & 1 & 0/1 & 0.00 \\
lost / pro\_high / text & 6 & 6 & 2/6 & 2.53 \\
lost / pro\_high / visual & 3 & 3 & 1/3 & 1.00 \\
lost / pro\_low / visual & 1 & 1 & 0/1 & 0.06 \\
ph\_drift / pro\_high / text & 42 & 42 & 31/42 & 11.44 \\
ph\_drift / pro\_high / visual & 42 & 42 & 31/42 & 13.49 \\
ph\_new / pro\_high / text & 84 & 84 & 66/84 & 21.83 \\
ph\_new / pro\_high / visual & 84 & 84 & 60/84 & 26.29 \\
xl\_drift / g38\_high / text & 23 & 23 & 0/23 & 3.80 \\
xl\_drift / g38\_high / visual & 23 & 23 & 0/23 & 3.44 \\
\bottomrule
\end{tabular}

%% file: gen_v6/tab_v6_f6_drift.tex
\begin{tabular}{lrrrrrrrr}
\toprule
drift control & $n$ & exact old & exact new & $b$ / $c$ & $p$ & same answer & same prompt tokens & transcr.\ exact old / new \\
\midrule
pro\_high Visual & 42 & 30 & 31 & 1 / 2 & 1.00 & 30 & 42 & 39 / 38 \\
pro\_high Text & 42 & 31 & 31 & 2 / 2 & 1.00 & 31 & 42 & -- \\
g38\_high XL Visual & 23 & 0 & 0 & 0 / 0 & 1.00 & 2 & 23 & 6 / 5 \\
g38\_high XL Text & 23 & 0 & 0 & 0 / 0 & 1.00 & 4 & 23 & -- \\
\bottomrule
\end{tabular}

%% file: gen_v6/tab_v6_f6_pro.tex
\begin{tabular}{llrrrrrrrrr}
\toprule
\texttt{3.1-pro} high & version & core exact & $\GapL$ & $\shP$ & $n$ (fit) & $\log_{10} m_{50}$ & $\GapL$ & $\shP$ & rise $\GapL$ & rise $\shP$ \\
\midrule
Visual & returned only (paper) & 84/115 & 29/42 & 31/49 & 91 & 4.3 & 4.5 & 4.1 & 2.0 & 1.5 \\
 & merged with F6 & 145/202 & 52/76 & 53/86 & 162 & 4.1 & 4.5 & 3.9 & 2.0 & 1.3 \\
 & merged, still lost = wrong & 145/202 & 52/76 & 53/86 & 162 & 4.1 & 4.5 & 3.9 & 2.0 & 1.3 \\
 & merged with F6 and C503 & 145/202 & 52/76 & 53/86 & 162 & 4.1 & 4.5 & 3.9 & 2.0 & 1.3 \\
 & merged with F6 and C503, still lost = wrong & 145/202 & 52/76 & 53/86 & 162 & 4.1 & 4.5 & 3.9 & 2.0 & 1.3 \\
\midrule
Text & returned only (paper) & 87/112 & 32/41 & 31/47 & 88 & 5.1 & 5.8 & 4.7 & 3.5 & 2.2 \\
 & merged with F6 & 155/202 & 58/76 & 57/86 & 162 & 5.0 & 5.8 & 4.6 & 3.5 & 2.1 \\
 & merged, still lost = wrong & 155/202 & 58/76 & 57/86 & 162 & 5.0 & 5.8 & 4.6 & 3.5 & 2.1 \\
 & merged with F6 and C503 & 155/202 & 58/76 & 57/86 & 162 & 5.0 & 5.8 & 4.6 & 3.5 & 2.1 \\
 & merged with F6 and C503, still lost = wrong & 155/202 & 58/76 & 57/86 & 162 & 5.0 & 5.8 & 4.6 & 3.5 & 2.1 \\
\bottomrule
\end{tabular}

%% file: gen_v6/tab_v6_f6_xl.tex
\begin{tabular}{llrrrrrrrrrr}
\toprule
\texttt{3.8-flash} high & version & 7x7 \#ST & 7x7 \#M & 8x8 \#ST & 8x8 \#PM & 8x8 \#M & 9x9 \#ST & 9x9 \#M & 10x10 \#ST & 10x10 \#PM & 10x10 \#M \\
\midrule
Visual CoT & returned only (paper) & 0/3 & 0/5 & 0/4 & 0/2 & 0/4 & 0/4 & 0/3 & 0/4 & 0/2 & 0/6 \\
 & merged with F6 & 0/6 & 0/6 & 0/6 & 0/6 & 0/6 & 0/6 & 0/6 & 0/6 & 0/6 & 0/6 \\
 & merged, still lost = wrong & 0/6 & 0/6 & 0/6 & 0/6 & 0/6 & 0/6 & 0/6 & 0/6 & 0/6 & 0/6 \\
 & merged with F6 and C503 & 0/6 & 0/6 & 0/6 & 0/6 & 0/6 & 0/6 & 0/6 & 0/6 & 0/6 & 0/6 \\
 & merged with F6 and C503, still lost = wrong & 0/6 & 0/6 & 0/6 & 0/6 & 0/6 & 0/6 & 0/6 & 0/6 & 0/6 & 0/6 \\
\midrule
Text CoT & returned only (paper) & 0/2 & 0/5 & 0/3 & 0/3 & 0/3 & 0/2 & 0/3 & 0/1 & 0/1 & 0/2 \\
 & merged with F6 & 0/6 & 0/6 & 0/6 & 0/6 & 0/6 & 0/6 & 0/6 & 0/6 & 0/6 & 0/6 \\
 & merged, still lost = wrong & 0/6 & 0/6 & 0/6 & 0/6 & 0/6 & 0/6 & 0/6 & 0/6 & 0/6 & 0/6 \\
 & merged with F6 and C503 & 0/6 & 0/6 & 0/6 & 0/6 & 0/6 & 0/6 & 0/6 & 0/6 & 0/6 & 0/6 \\
 & merged with F6 and C503, still lost = wrong & 0/6 & 0/6 & 0/6 & 0/6 & 0/6 & 0/6 & 0/6 & 0/6 & 0/6 & 0/6 \\
\bottomrule
\end{tabular}

%% file: gen_v6/tab_v6_f6_slope.tex
\begin{tabular}{lrrrrrrrr}
\toprule
version & slope & 95\% CI & $p$ & $R^2$ & leave-one-out & \texttt{3.8-flash} & \texttt{3.6-flash} & \texttt{3.1-pro} \\
\midrule
returned only (paper) & 3.5 & $[2.1, 5.0]$ & $1.5\times 10^{-4}$ & 0.71 & 2.8--3.8 & 5.6 & 2.4 & 2.1 \\
merged with F6 & 3.5 & $[2.1, 4.9]$ & $1.7\times 10^{-4}$ & 0.71 & 2.8--3.8 & 5.6 & 2.4 & 1.9 \\
merged, still lost = wrong & 3.5 & $[2.1, 4.9]$ & $1.7\times 10^{-4}$ & 0.71 & 2.8--3.8 & 5.6 & 2.4 & 1.9 \\
merged with F6 and C503 & 3.2 & $[2.1, 4.4]$ & $5.1\times 10^{-5}$ & 0.76 & 2.8--3.5 & 3.9 & 2.4 & 1.9 \\
merged with F6 and C503, still lost = wrong & 3.2 & $[2.1, 4.4]$ & $5.1\times 10^{-5}$ & 0.76 & 2.8--3.5 & 3.9 & 2.4 & 1.9 \\
\bottomrule
\end{tabular}

%% file: gen_v6/tab_v6_f6_other.tex
\begin{tabular}{lllrrrr}
\toprule
configuration & lost keys returned & version & core exact & XL exact & held-out exact & $\log_{10} m_{50}$ \\
\midrule
g36\_high text & 1/1 & returned only (paper) & 137/201 & -- & -- & 3.7 \\
 &  & merged with F6 & 138/202 & -- & -- & 3.7 \\
 &  & merged, still lost = wrong & 138/202 & -- & -- & 3.7 \\
 &  & merged with F6 and C503 & 138/202 & -- & -- & 3.7 \\
 &  & merged with F6 and C503, still lost = wrong & 138/202 & -- & -- & 3.7 \\
g36\_low visual & 1/1 & returned only (paper) & 205/391 & -- & -- & 2.2 \\
 &  & merged with F6 & 205/392 & -- & -- & 2.2 \\
 &  & merged, still lost = wrong & 205/392 & -- & -- & 2.2 \\
 &  & merged with F6 and C503 & 205/392 & -- & -- & 2.2 \\
 &  & merged with F6 and C503, still lost = wrong & 205/392 & -- & -- & 2.2 \\
g37\_low text & 3/3 & returned only (paper) & 119/199 & -- & -- & 2.9 \\
 &  & merged with F6 & 121/202 & -- & -- & 2.9 \\
 &  & merged, still lost = wrong & 121/202 & -- & -- & 2.9 \\
 &  & merged with F6 and C503 & 121/202 & -- & -- & 2.9 \\
 &  & merged with F6 and C503, still lost = wrong & 121/202 & -- & -- & 2.9 \\
g37\_low visual & 8/8 & returned only (paper) & 123/194 & -- & -- & 3.2 \\
 &  & merged with F6 & 128/202 & -- & -- & 3.2 \\
 &  & merged, still lost = wrong & 128/202 & -- & -- & 3.2 \\
 &  & merged with F6 and C503 & 128/202 & -- & -- & 3.2 \\
 &  & merged with F6 and C503, still lost = wrong & 128/202 & -- & -- & 3.2 \\
g38\_low hint & 1/1 & returned only (paper) & 257/391 & -- & -- & 3.5 \\
 &  & merged with F6 & 258/392 & -- & -- & 3.6 \\
 &  & merged, still lost = wrong & 258/392 & -- & -- & 3.6 \\
 &  & merged with F6 and C503 & 258/392 & -- & -- & 3.6 \\
 &  & merged with F6 and C503, still lost = wrong & 258/392 & -- & -- & 3.6 \\
g38\_low text & 1/1 & returned only (paper) & 238/391 & 0/60 & -- & 3.0 \\
 &  & merged with F6 & 238/392 & 0/60 & -- & 3.0 \\
 &  & merged, still lost = wrong & 238/392 & 0/60 & -- & 3.0 \\
 &  & merged with F6 and C503 & 238/392 & 0/60 & -- & 3.0 \\
 &  & merged with F6 and C503, still lost = wrong & 238/392 & 0/60 & -- & 3.0 \\
g38\_med text & 7/7 & returned only (paper) & 283/385 & -- & -- & 4.6 \\
 &  & merged with F6 & 285/392 & -- & -- & 4.5 \\
 &  & merged, still lost = wrong & 285/392 & -- & -- & 4.5 \\
 &  & merged with F6 and C503 & 285/392 & -- & -- & 4.5 \\
 &  & merged with F6 and C503, still lost = wrong & 285/392 & -- & -- & 4.5 \\
gm31\_high car & 0/4 & returned only (paper) & 23/25 & -- & -- & -- \\
 &  & merged with F6 & 23/25 & -- & -- & -- \\
 &  & merged, still lost = wrong & 23/29 & -- & -- & -- \\
 &  & merged with F6 and C503 & 23/25 & -- & -- & -- \\
 &  & merged with F6 and C503, still lost = wrong & 23/29 & -- & -- & -- \\
gm31\_high code & 0/7 & returned only (paper) & 18/19 & -- & -- & -- \\
 &  & merged with F6 & 18/19 & -- & -- & -- \\
 &  & merged, still lost = wrong & 18/26 & -- & -- & -- \\
 &  & merged with F6 and C503 & 18/19 & -- & -- & -- \\
 &  & merged with F6 and C503, still lost = wrong & 18/26 & -- & -- & -- \\
gm31\_high hint & 0/1 & returned only (paper) & 12/24 & -- & -- & -- \\
 &  & merged with F6 & 12/24 & -- & -- & -- \\
 &  & merged, still lost = wrong & 12/25 & -- & -- & -- \\
 &  & merged with F6 and C503 & 12/24 & -- & -- & -- \\
 &  & merged with F6 and C503, still lost = wrong & 12/25 & -- & -- & -- \\
gm31\_high text & 1/1 & returned only (paper) & 15/23 & -- & -- & -- \\
 &  & merged with F6 & 15/24 & -- & -- & -- \\
 &  & merged, still lost = wrong & 15/24 & -- & -- & -- \\
 &  & merged with F6 and C503 & 15/24 & -- & -- & -- \\
 &  & merged with F6 and C503, still lost = wrong & 15/24 & -- & -- & -- \\
gm31\_high visual & 0/1 & returned only (paper) & 12/23 & -- & -- & -- \\
 &  & merged with F6 & 12/23 & -- & -- & -- \\
 &  & merged, still lost = wrong & 12/24 & -- & -- & -- \\
 &  & merged with F6 and C503 & 12/23 & -- & -- & -- \\
 &  & merged with F6 and C503, still lost = wrong & 12/24 & -- & -- & -- \\
gm31\_min text & 1/1 & returned only (paper) & 131/391 & -- & -- & 0.7 \\
 &  & merged with F6 & 131/392 & -- & -- & 0.7 \\
 &  & merged, still lost = wrong & 131/392 & -- & -- & 0.7 \\
 &  & merged with F6 and C503 & 131/392 & -- & -- & 0.7 \\
 &  & merged with F6 and C503, still lost = wrong & 131/392 & -- & -- & 0.7 \\
pro\_low visual & 1/1 & returned only (paper) & 118/202 & 0/59 & -- & 2.6 \\
 &  & merged with F6 & 118/202 & 0/60 & -- & 2.6 \\
 &  & merged, still lost = wrong & 118/202 & 0/60 & -- & 2.6 \\
 &  & merged with F6 and C503 & 118/202 & 0/60 & -- & 2.6 \\
 &  & merged with F6 and C503, still lost = wrong & 118/202 & 0/60 & -- & 2.6 \\
\bottomrule
\end{tabular}

%% file: gen_v8p/gen_v8/tab_g9_main.tex
\begin{tabular}{llrrrrrrrrrrrl}
\toprule
Model & Cond. & $\beta_{\lg}\pm\text{SE}$ & odds/dec. & H1 $p_{\mathrm{Holm}}$ & $\beta_{\mathrm{bud}}\pm\text{SE}$ & H2 $p_{\mathrm{Holm}}$ & H3 $b/c$ & H3 $p_{\mathrm{Holm}}$ & $m_{50}(512)$ & $m_{50}(2048)$ & $m_{50}(8192)$ & $m_{50}(\text{free})$ & slope/dec.\ [95\% CI] \\
\midrule
\texttt{gemma-4-31B-it} & Visual & $-$1.22$\pm$0.11 & 0.30 & $<10^{-4}$ & 1.19$\pm$0.16 & $<10^{-4}$ & 1/5 & 1.00 & 0.04 & 0.43 & 0.99 & 1.15 & 0.79 [0.55, 1.06] \\
\texttt{gemma-4-31B-it} & Text & $-$1.46$\pm$0.12 & 0.23 & $<10^{-4}$ & 1.25$\pm$0.15 & $<10^{-4}$ & 0/8 & 1.00 & 0.43 & 0.68 & 1.50 & 1.84 & 0.88 [0.70, 1.12] \\
\midrule
\texttt{gemma-4-26B-A4B-it} & Visual & $-$1.70$\pm$0.21 & 0.18 & $<10^{-4}$ & 1.66$\pm$0.30 & $<10^{-4}$ & 1/1 & 1.00 & $-$1.40 & $-$0.42 & $-$0.16 & 0.40 & 1.03 [0.18, 2.73] \\
\texttt{gemma-4-26B-A4B-it} & Text & $-$1.65$\pm$0.14 & 0.19 & $<10^{-4}$ & 1.52$\pm$0.20 & $<10^{-4}$ & 2/3 & 1.00 & $-$0.03 & 0.14 & 0.72 & 1.54 & 0.62 [0.38, 0.90] \\
\midrule
\texttt{gemma-4-12B-it} & Visual & $-$1.49$\pm$0.18 & 0.23 & $<10^{-4}$ & 1.53$\pm$0.23 & $<10^{-4}$ & 1/2 & 1.00 & $-$0.96 & $-$0.15 & 0.34 & 0.41 & 1.07 [0.51, 2.07] \\
\texttt{gemma-4-12B-it} & Text & $-$1.51$\pm$0.15 & 0.22 & $<10^{-4}$ & 1.98$\pm$0.21 & $<10^{-4}$ & 1/1 & 1.00 & $-$0.67 & 0.29 & 0.88 & 1.03 & 1.28 [0.88, 1.79] \\
\midrule
\texttt{gemma-4-E4B-it} & Visual & $-$0.14$\pm$0.44 & 0.87 & 0.372 & 0.00$\pm$0.77 & 0.500 & 0/0 & -- & $-$1.92 & $-$3.55 & $-$4.58 & $-$42.53 & $-$2.21 [$-$2.87, $-$1.15] \\
\texttt{gemma-4-E4B-it} & Text & $-$1.72$\pm$0.21 & 0.18 & $<10^{-4}$ & 1.18$\pm$0.22 & $<10^{-4}$ & 0/0 & -- & $-$0.13 & 0.28 & 0.47 & 0.36 & 0.49 [0.22, 0.82] \\
\bottomrule
\end{tabular}

%% file: gen_v8/tab_g9_acc.tex
\begin{tabular}{lllrrrrrrrrrr}
\toprule
Model & Cond. & Budget & Core ($921$) & Bin ($200$) & Grid $\GapL$ ($291$) & Grid $\shP$ ($350$) & Euler ($80$) & Twins ($160$) & Twin $\sST$ & Twin $\sAM$ & Percept.\ exact (\%) & Percept.\ F1 \\
\midrule
\texttt{gemma-4-31B-it} & Visual & 512 & 24.5 & 82.5 & 7.5 & 6.7 & 14.8 & 0.6 & 1.2 & 0.0 & 34.5 & 0.971 \\
\texttt{gemma-4-31B-it} & Visual & 2048 & 27.3 & 82.5 & 9.6 & 9.3 & 23.4 & 0.0 & 0.0 & 0.0 & 35.5 & 0.973 \\
\texttt{gemma-4-31B-it} & Visual & 8192 & 36.2 & 94.0 & 13.5 & 17.9 & 39.8 & 5.0 & 3.8 & 6.2 & 39.8 & 0.973 \\
\texttt{gemma-4-31B-it} & Visual & free & 38.1 & 93.5 & 17.1 & 17.9 & 46.9 & 6.2 & 3.8 & 8.8 & 40.3 & 0.974 \\
\texttt{gemma-4-31B-it} & Text & 512 & 27.7 & 82.0 & 12.5 & 10.3 & 18.8 & 0.6 & 1.2 & 0.0 & -- & -- \\
\texttt{gemma-4-31B-it} & Text & 2048 & 33.1 & 94.0 & 16.7 & 13.1 & 22.7 & 0.0 & 0.0 & 0.0 & -- & -- \\
\texttt{gemma-4-31B-it} & Text & 8192 & 42.9 & 100.0 & 24.9 & 19.6 & 50.0 & 2.5 & 1.2 & 3.8 & -- & -- \\
\texttt{gemma-4-31B-it} & Text & free & 47.1 & 100.0 & 30.2 & 25.0 & 55.5 & 10.0 & 5.0 & 15.0 & -- & -- \\
\midrule
\texttt{gemma-4-26B-A4B-it} & Visual & 512 & 14.7 & 61.5 & 1.4 & 0.0 & 6.2 & 0.0 & 0.0 & 0.0 & 35.5 & 0.963 \\
\texttt{gemma-4-26B-A4B-it} & Visual & 2048 & 18.9 & 81.0 & 0.7 & 0.6 & 6.2 & 0.0 & 0.0 & 0.0 & 27.6 & 0.961 \\
\texttt{gemma-4-26B-A4B-it} & Visual & 8192 & 24.8 & 89.0 & 3.9 & 3.5 & 21.9 & 0.0 & 0.0 & 0.0 & 33.1 & 0.963 \\
\texttt{gemma-4-26B-A4B-it} & Visual & free & 28.1 & 90.5 & 5.7 & 5.1 & 35.9 & 1.2 & 1.2 & 1.2 & 32.2 & 0.962 \\
\texttt{gemma-4-26B-A4B-it} & Text & 512 & 13.7 & 47.0 & 2.1 & 3.2 & 12.5 & 0.0 & 0.0 & 0.0 & -- & -- \\
\texttt{gemma-4-26B-A4B-it} & Text & 2048 & 22.1 & 72.5 & 7.5 & 7.1 & 12.5 & 0.0 & 0.0 & 0.0 & -- & -- \\
\texttt{gemma-4-26B-A4B-it} & Text & 8192 & 30.6 & 87.5 & 12.5 & 11.9 & 27.3 & 1.2 & 1.2 & 1.2 & -- & -- \\
\texttt{gemma-4-26B-A4B-it} & Text & free & 43.2 & 100.0 & 27.8 & 17.9 & 50.0 & 3.1 & 2.5 & 3.8 & -- & -- \\
\midrule
\texttt{gemma-4-12B-it} & Visual & 512 & 14.9 & 59.5 & 1.4 & 2.6 & 4.7 & 0.0 & 0.0 & 0.0 & 25.7 & 0.953 \\
\texttt{gemma-4-12B-it} & Visual & 2048 & 18.9 & 68.5 & 3.9 & 4.2 & 10.2 & 0.6 & 1.2 & 0.0 & 25.7 & 0.955 \\
\texttt{gemma-4-12B-it} & Visual & 8192 & 25.1 & 78.5 & 5.7 & 9.0 & 23.4 & 0.0 & 0.0 & 0.0 & 28.8 & 0.959 \\
\texttt{gemma-4-12B-it} & Visual & free & 28.6 & 88.5 & 7.1 & 9.3 & 28.9 & 1.9 & 1.2 & 2.5 & 33.3 & 0.961 \\
\texttt{gemma-4-12B-it} & Text & 512 & 14.8 & 57.0 & 1.1 & 3.8 & 5.5 & 0.0 & 0.0 & 0.0 & -- & -- \\
\texttt{gemma-4-12B-it} & Text & 2048 & 23.1 & 72.5 & 6.4 & 8.0 & 19.5 & 0.0 & 0.0 & 0.0 & -- & -- \\
\texttt{gemma-4-12B-it} & Text & 8192 & 30.8 & 81.5 & 16.7 & 13.1 & 25.8 & 0.0 & 0.0 & 0.0 & -- & -- \\
\texttt{gemma-4-12B-it} & Text & free & 36.7 & 97.5 & 19.2 & 15.7 & 31.2 & 1.2 & 1.2 & 1.2 & -- & -- \\
\midrule
\texttt{gemma-4-E4B-it} & Visual & 512 & 6.5 & 29.5 & 0.0 & 0.3 & 0.0 & 0.0 & 0.0 & 0.0 & 2.6 & 0.786 \\
\texttt{gemma-4-E4B-it} & Visual & 2048 & 8.7 & 37.5 & 0.7 & 1.0 & 0.0 & 0.0 & 0.0 & 0.0 & 2.1 & 0.806 \\
\texttt{gemma-4-E4B-it} & Visual & 8192 & 8.7 & 39.5 & 0.4 & 0.0 & 0.0 & 0.0 & 0.0 & 0.0 & 2.0 & 0.815 \\
\texttt{gemma-4-E4B-it} & Visual & free & 8.0 & 36.5 & 0.0 & 0.3 & 0.0 & 0.0 & 0.0 & 0.0 & 2.1 & 0.817 \\
\texttt{gemma-4-E4B-it} & Text & 512 & 14.0 & 48.5 & 3.2 & 3.8 & 8.6 & 0.0 & 0.0 & 0.0 & -- & -- \\
\texttt{gemma-4-E4B-it} & Text & 2048 & 18.1 & 61.5 & 4.6 & 3.8 & 14.8 & 0.6 & 1.2 & 0.0 & -- & -- \\
\texttt{gemma-4-E4B-it} & Text & 8192 & 25.3 & 77.0 & 12.5 & 8.7 & 13.3 & 0.0 & 0.0 & 0.0 & -- & -- \\
\texttt{gemma-4-E4B-it} & Text & free & 25.6 & 81.0 & 11.7 & 9.0 & 10.2 & 0.0 & 0.0 & 0.0 & -- & -- \\
\bottomrule
\end{tabular}

%% file: gen_v8p/gen_v8/tab_g9_compliance.tex
\begin{tabular}{lllrrrrrrl}
\toprule
Model & Cond. & Budget & $n$ & Med.\ reason.\ tok. & Med.\ compl.\ tok. & Compliance ($\le b{+}16$, \%) & Length-trunc.\ (\%) & Class $\beta_{\mathrm{hard}}\pm\text{SE}$ (free) & Class $p$ (two-sided) \\
\midrule
\texttt{gemma-4-31B-it} & Visual & 512 & 1081 & 510 & 2030 & 100.0 & 33.9 & -- & -- \\
\texttt{gemma-4-31B-it} & Visual & 2048 & 1081 & 2046 & 3185 & 100.0 & 10.9 & -- & -- \\
\texttt{gemma-4-31B-it} & Visual & 8192 & 1081 & 8190 & 9188 & 100.0 & 1.0 & -- & -- \\
\texttt{gemma-4-31B-it} & Visual & free & 1081 & 10448 & 11573 & -- & 0.1 & 0.91$\pm$0.24 & 0.0002 \\
\texttt{gemma-4-31B-it} & Text & 512 & 1081 & 510 & 1875 & 100.0 & 28.3 & -- & -- \\
\texttt{gemma-4-31B-it} & Text & 2048 & 1081 & 2046 & 3020 & 100.0 & 6.3 & -- & -- \\
\texttt{gemma-4-31B-it} & Text & 8192 & 1081 & 8190 & 8930 & 100.0 & 0.6 & -- & -- \\
\texttt{gemma-4-31B-it} & Text & free & 1081 & 12184 & 12997 & -- & 0.0 & 0.41$\pm$0.24 & 0.085 \\
\midrule
\texttt{gemma-4-26B-A4B-it} & Visual & 512 & 1081 & 510 & 2560 & 100.0 & 74.1 & -- & -- \\
\texttt{gemma-4-26B-A4B-it} & Visual & 2048 & 1081 & 2046 & 3882 & 100.0 & 47.2 & -- & -- \\
\texttt{gemma-4-26B-A4B-it} & Visual & 8192 & 1081 & 8190 & 9429 & 100.0 & 28.0 & -- & -- \\
\texttt{gemma-4-26B-A4B-it} & Visual & free & 1081 & 21415 & 22460 & -- & 32.3 & 0.74$\pm$0.31 & 0.019 \\
\texttt{gemma-4-26B-A4B-it} & Text & 512 & 1081 & 510 & 2560 & 100.0 & 74.1 & -- & -- \\
\texttt{gemma-4-26B-A4B-it} & Text & 2048 & 1081 & 2046 & 3409 & 99.9 & 37.2 & -- & -- \\
\texttt{gemma-4-26B-A4B-it} & Text & 8192 & 1081 & 8190 & 9245 & 99.9 & 29.9 & -- & -- \\
\texttt{gemma-4-26B-A4B-it} & Text & free & 1081 & 28759 & 29420 & -- & 48.2 & $-$0.06$\pm$0.24 & 0.817 \\
\midrule
\texttt{gemma-4-12B-it} & Visual & 512 & 1081 & 510 & 2560 & 99.9 & 67.1 & -- & -- \\
\texttt{gemma-4-12B-it} & Visual & 2048 & 1081 & 2046 & 3840 & 100.0 & 44.4 & -- & -- \\
\texttt{gemma-4-12B-it} & Visual & 8192 & 1081 & 8190 & 9460 & 100.0 & 27.5 & -- & -- \\
\texttt{gemma-4-12B-it} & Visual & free & 1081 & 28565 & 29637 & -- & 49.3 & 0.71$\pm$0.29 & 0.016 \\
\texttt{gemma-4-12B-it} & Text & 512 & 1081 & 510 & 2560 & 99.2 & 65.9 & -- & -- \\
\texttt{gemma-4-12B-it} & Text & 2048 & 1081 & 2046 & 3339 & 100.0 & 31.8 & -- & -- \\
\texttt{gemma-4-12B-it} & Text & 8192 & 1081 & 8190 & 9264 & 100.0 & 25.1 & -- & -- \\
\texttt{gemma-4-12B-it} & Text & free & 1081 & 29996 & 30000 & -- & 58.6 & 0.60$\pm$0.26 & 0.018 \\
\midrule
\texttt{gemma-4-E4B-it} & Visual & 512 & 1081 & 510 & 2038 & 100.0 & 34.6 & -- & -- \\
\texttt{gemma-4-E4B-it} & Visual & 2048 & 1081 & 2046 & 2948 & 100.0 & 8.9 & -- & -- \\
\texttt{gemma-4-E4B-it} & Visual & 8192 & 1081 & 3364 & 4194 & 100.0 & 1.3 & -- & -- \\
\texttt{gemma-4-E4B-it} & Visual & free & 1081 & 3316 & 4152 & -- & 0.0 & sep. & 0.999 \\
\texttt{gemma-4-E4B-it} & Text & 512 & 1081 & 510 & 1127 & 100.0 & 13.0 & -- & -- \\
\texttt{gemma-4-E4B-it} & Text & 2048 & 1081 & 2046 & 2483 & 100.0 & 5.1 & -- & -- \\
\texttt{gemma-4-E4B-it} & Text & 8192 & 1081 & 3672 & 4072 & 100.0 & 0.3 & -- & -- \\
\texttt{gemma-4-E4B-it} & Text & free & 1081 & 3677 & 4116 & -- & 0.0 & 0.63$\pm$0.32 & 0.046 \\
\bottomrule
\end{tabular}

%% file: gen_v8p/gen_v8/tab_qw_main.tex
\begin{tabular}{llrrrrrrrrrrrl}
\toprule
Model & Cond. & $\beta_{\lg}\pm\text{SE}$ & odds/dec. & H1 $p_{\mathrm{Holm}}$ & $\beta_{\mathrm{bud}}\pm\text{SE}$ & H2 $p_{\mathrm{Holm}}$ & H3 $b/c$ & H3 $p_{\mathrm{Holm}}$ & $m_{50}(512)$ & $m_{50}(2048)$ & $m_{50}(8192)$ & $m_{50}(\text{free})$ & slope/dec.\ [95\% CI] \\
\midrule
\texttt{Qwen3-VL-32B-Thinking} & Visual & $-$1.34$\pm$0.33 & 0.26 & $<10^{-4}$ & 0.58$\pm$0.38 & 0.123 & 1/0 & 0.500 & $-$0.63 & $-$0.87 & $-$1.16 & $-$1.12 & $-$0.44 [$-$1.12, 0.35] \\
\texttt{Qwen3-VL-32B-Thinking} & Text & $-$1.61$\pm$0.17 & 0.20 & $<10^{-4}$ & 1.45$\pm$0.25 & $<10^{-4}$ & 0/0 & -- & $-$0.07 & $-$0.07 & 0.21 & 0.81 & 0.23 [$-$0.07, 0.59] \\
\midrule
\texttt{Qwen3-VL-8B-Thinking} & Visual & $-$2.77$\pm$1.17 & 0.06 & 0.018 & $-$1.43$\pm$0.57 & 0.994 & 0/0 & -- & $-$1.13 & $-$2.33 & $-$4.30 & $-$0.84 & $-$2.63 [$-$4.15, $-$1.32] \\
\texttt{Qwen3-VL-8B-Thinking} & Text & $-$1.50$\pm$0.19 & 0.22 & $<10^{-4}$ & 1.78$\pm$0.30 & $<10^{-4}$ & 0/0 & -- & $-$0.29 & $-$0.94 & 0.12 & 0.32 & 0.34 [$-$0.04, 0.84] \\
\midrule
\texttt{Qwen3-VL-8B-Instruct} & Visual & $-$1.08$\pm$0.60 & 0.34 & 0.036 & -- & -- & 0/0 & -- & -- & -- & -- & $-$3.08 & -- \\
\texttt{Qwen3-VL-8B-Instruct} & Text & $-$2.90$\pm$0.56 & 0.05 & $<10^{-4}$ & -- & -- & 0/0 & -- & -- & -- & -- & $-$0.15 & -- \\
\bottomrule
\end{tabular}

%% file: gen_v8/tab_qw_acc.tex
\begin{tabular}{lllrrrrrrrrrr}
\toprule
Model & Cond. & Budget & Core ($921$) & Bin ($200$) & Grid $\GapL$ ($291$) & Grid $\shP$ ($350$) & Euler ($80$) & Twins ($160$) & Twin $\sST$ & Twin $\sAM$ & Percept.\ exact (\%) & Percept.\ F1 \\
\midrule
\texttt{Qwen3-VL-32B-Thinking} & Visual & 512 & 6.3 & 24.5 & 0.4 & 0.6 & 4.7 & 0.0 & 0.0 & 0.0 & 11.7 & 0.920 \\
\texttt{Qwen3-VL-32B-Thinking} & Visual & 2048 & 8.8 & 36.5 & 0.0 & 0.3 & 5.5 & 0.0 & 0.0 & 0.0 & 15.7 & 0.941 \\
\texttt{Qwen3-VL-32B-Thinking} & Visual & 8192 & 18.7 & 78.0 & 2.1 & 2.2 & 2.3 & 0.0 & 0.0 & 0.0 & 18.8 & 0.944 \\
\texttt{Qwen3-VL-32B-Thinking} & Visual & free & 20.0 & 83.0 & 2.1 & 1.6 & 5.5 & 0.6 & 1.2 & 0.0 & 19.6 & 0.944 \\
\texttt{Qwen3-VL-32B-Thinking} & Text & 512 & 4.8 & 11.5 & 0.4 & 1.6 & 11.7 & 0.0 & 0.0 & 0.0 & -- & -- \\
\texttt{Qwen3-VL-32B-Thinking} & Text & 2048 & 8.1 & 25.5 & 1.1 & 2.9 & 9.4 & 0.0 & 0.0 & 0.0 & -- & -- \\
\texttt{Qwen3-VL-32B-Thinking} & Text & 8192 & 22.8 & 72.0 & 8.2 & 9.6 & 10.2 & 0.0 & 0.0 & 0.0 & -- & -- \\
\texttt{Qwen3-VL-32B-Thinking} & Text & free & 34.2 & 98.5 & 16.0 & 15.7 & 18.8 & 0.0 & 0.0 & 0.0 & -- & -- \\
\midrule
\texttt{Qwen3-VL-8B-Thinking} & Visual & 512 & 3.9 & 12.5 & 1.1 & 0.6 & 4.7 & 0.0 & 0.0 & 0.0 & 5.2 & 0.886 \\
\texttt{Qwen3-VL-8B-Thinking} & Visual & 2048 & 8.5 & 36.5 & 0.4 & 0.0 & 3.1 & 0.0 & 0.0 & 0.0 & 5.8 & 0.881 \\
\texttt{Qwen3-VL-8B-Thinking} & Visual & 8192 & 10.0 & 45.0 & 0.0 & 0.6 & 0.0 & 0.0 & 0.0 & 0.0 & 5.6 & 0.868 \\
\texttt{Qwen3-VL-8B-Thinking} & Visual & free & 12.6 & 56.0 & 0.0 & 0.3 & 2.3 & 0.0 & 0.0 & 0.0 & 5.4 & 0.852 \\
\texttt{Qwen3-VL-8B-Thinking} & Text & 512 & 4.8 & 15.5 & 1.8 & 0.0 & 6.2 & 0.0 & 0.0 & 0.0 & -- & -- \\
\texttt{Qwen3-VL-8B-Thinking} & Text & 2048 & 7.9 & 30.0 & 1.8 & 1.3 & 3.1 & 0.0 & 0.0 & 0.0 & -- & -- \\
\texttt{Qwen3-VL-8B-Thinking} & Text & 8192 & 19.9 & 64.0 & 6.0 & 9.3 & 7.0 & 0.0 & 0.0 & 0.0 & -- & -- \\
\texttt{Qwen3-VL-8B-Thinking} & Text & free & 29.3 & 96.0 & 11.7 & 11.5 & 7.0 & 0.0 & 0.0 & 0.0 & -- & -- \\
\midrule
\texttt{Qwen3-VL-8B-Instruct} & Visual & free & 7.7 & 33.5 & 0.0 & 1.0 & 0.8 & 0.0 & 0.0 & 0.0 & 3.6 & 0.484 \\
\texttt{Qwen3-VL-8B-Instruct} & Text & free & 12.1 & 45.5 & 1.1 & 1.3 & 10.2 & 0.0 & 0.0 & 0.0 & -- & -- \\
\bottomrule
\end{tabular}

%% file: gen_v8p/gen_v8/tab_qw_compliance.tex
\begin{tabular}{lllrrrrrrl}
\toprule
Model & Cond. & Budget & $n$ & Med.\ reason.\ tok. & Med.\ compl.\ tok. & Compliance ($\le b{+}16$, \%) & Length-trunc.\ (\%) & Class $\beta_{\mathrm{hard}}\pm\text{SE}$ (free) & Class $p$ (two-sided) \\
\midrule
\texttt{Qwen3-VL-32B-Thinking} & Visual & 512 & 1081 & 511 & 2560 & 100.0 & 53.2 & -- & -- \\
\texttt{Qwen3-VL-32B-Thinking} & Visual & 2048 & 1081 & 2047 & 2348 & 100.0 & 27.8 & -- & -- \\
\texttt{Qwen3-VL-32B-Thinking} & Visual & 8192 & 1081 & 8191 & 8400 & 100.0 & 10.7 & -- & -- \\
\texttt{Qwen3-VL-32B-Thinking} & Visual & free & 1081 & 17935 & 18205 & -- & 15.9 & $-$0.25$\pm$0.58 & 0.662 \\
\texttt{Qwen3-VL-32B-Thinking} & Text & 512 & 1081 & 511 & 2560 & 100.0 & 78.8 & -- & -- \\
\texttt{Qwen3-VL-32B-Thinking} & Text & 2048 & 1081 & 2047 & 4096 & 100.0 & 51.8 & -- & -- \\
\texttt{Qwen3-VL-32B-Thinking} & Text & 8192 & 1081 & 8191 & 8648 & 100.0 & 22.5 & -- & -- \\
\texttt{Qwen3-VL-32B-Thinking} & Text & free & 1081 & 24376 & 24876 & -- & 39.3 & 0.39$\pm$0.27 & 0.148 \\
\midrule
\texttt{Qwen3-VL-8B-Thinking} & Visual & 512 & 1081 & 511 & 800 & 100.0 & 24.6 & -- & -- \\
\texttt{Qwen3-VL-8B-Thinking} & Visual & 2048 & 1081 & 2047 & 2277 & 100.0 & 10.6 & -- & -- \\
\texttt{Qwen3-VL-8B-Thinking} & Visual & 8192 & 1081 & 8191 & 8371 & 100.0 & 5.3 & -- & -- \\
\texttt{Qwen3-VL-8B-Thinking} & Visual & free & 1081 & 16104 & 16380 & -- & 13.4 & 0.84$\pm$1.52 & 0.581 \\
\texttt{Qwen3-VL-8B-Thinking} & Text & 512 & 1081 & 511 & 535 & 100.0 & 46.6 & -- & -- \\
\texttt{Qwen3-VL-8B-Thinking} & Text & 2048 & 1081 & 2047 & 2059 & 100.0 & 26.5 & -- & -- \\
\texttt{Qwen3-VL-8B-Thinking} & Text & 8192 & 1081 & 8191 & 8203 & 100.0 & 13.7 & -- & -- \\
\texttt{Qwen3-VL-8B-Thinking} & Text & free & 1081 & 28778 & 28855 & -- & 46.1 & 0.79$\pm$0.31 & 0.010 \\
\midrule
\texttt{Qwen3-VL-8B-Instruct} & Visual & free & 1081 & 0 & 1647 & -- & 17.8 & sep. & 0.999 \\
\texttt{Qwen3-VL-8B-Instruct} & Text & free & 1081 & 0 & 30000 & -- & 52.1 & 0.61$\pm$0.67 & 0.361 \\
\bottomrule
\end{tabular}

%% file: gen_v8p/gen_v8/tab_trunc_finished.tex
\begin{tabular}{lllrrrrrl}
\toprule
Exp. & Model & Cond. & frozen & finished & $p$ & $p_{\mathrm{Holm}}$ & $n_{\mathrm{fin}}$ & survives \\
\midrule
\multicolumn{9}{l}{\textit{H1 (magnitude law): $\beta_{\lg}$ at \texttt{free} budget}} \\
\midrule
G9 & \texttt{gemma-4-31B-it} & Visual & $-$1.22 & $-$1.22 & $<10^{-4}$ & $<10^{-4}$ & 720 & yes \\
G9 & \texttt{gemma-4-31B-it} & Text & $-$1.46 & $-$1.46 & $<10^{-4}$ & $<10^{-4}$ & 721 & yes \\
\midrule
G9 & \texttt{gemma-4-26B-A4B-it} & Visual & $-$1.70 & $-$1.75 & $<10^{-4}$ & $<10^{-4}$ & 465 & yes \\
G9 & \texttt{gemma-4-26B-A4B-it} & Text & $-$1.65 & $-$1.63 & $<10^{-4}$ & $<10^{-4}$ & 309 & yes \\
\midrule
G9 & \texttt{gemma-4-12B-it} & Visual & $-$1.49 & $-$1.73 & $<10^{-4}$ & $<10^{-4}$ & 327 & yes \\
G9 & \texttt{gemma-4-12B-it} & Text & $-$1.51 & $-$2.09 & $<10^{-4}$ & $<10^{-4}$ & 234 & yes \\
\midrule
G9 & \texttt{gemma-4-E4B-it} & Visual & $-$0.14 & $-$0.14 & 0.372 & 0.372 & 721 & no \\
G9 & \texttt{gemma-4-E4B-it} & Text & $-$1.72 & $-$1.72 & $<10^{-4}$ & $<10^{-4}$ & 721 & yes \\
\midrule
QW & \texttt{Qwen3-VL-32B-Thinking} & Visual & $-$1.34 & $-$1.23 & $<10^{-4}$ & 0.0002 & 585 & yes \\
QW & \texttt{Qwen3-VL-32B-Thinking} & Text & $-$1.61 & $-$2.00 & $<10^{-4}$ & $<10^{-4}$ & 384 & yes \\
\midrule
QW & \texttt{Qwen3-VL-8B-Thinking} & Visual & $-$2.77 & $-$2.64 & 0.011 & 0.022 & 614 & yes \\
QW & \texttt{Qwen3-VL-8B-Thinking} & Text & $-$1.50 & $-$2.92 & $<10^{-4}$ & $<10^{-4}$ & 326 & yes \\
\midrule
QW & \texttt{Qwen3-VL-8B-Instruct} & Visual & $-$1.08 & $-$0.76 & 0.125 & 0.125 & 612 & no$^\dagger$ \\
QW & \texttt{Qwen3-VL-8B-Instruct} & Text & $-$2.90 & $-$2.46 & 0.0004 & 0.001 & 277 & yes \\
\midrule
\multicolumn{9}{l}{\textit{H2 (budget effect): $\beta_{\mathrm{bud}}$ across $b\in\{512,2048,8192\}$}} \\
\midrule
G9 & \texttt{gemma-4-31B-it} & Visual & 1.19 & 0.65 & 0.0001 & 0.0007 & 1{,}799 & yes \\
G9 & \texttt{gemma-4-31B-it} & Text & 1.25 & 0.94 & $<10^{-4}$ & $<10^{-4}$ & 1{,}886 & yes \\
\midrule
G9 & \texttt{gemma-4-26B-A4B-it} & Visual & 1.66 & 1.98 & 0.0003 & 0.002 & 942 & yes \\
G9 & \texttt{gemma-4-26B-A4B-it} & Text & 1.52 & 0.58 & 0.045 & 0.091 & 1{,}059 & no$^\dagger$ \\
\midrule
G9 & \texttt{gemma-4-12B-it} & Visual & 1.53 & 0.70 & 0.005 & 0.019 & 1{,}063 & yes \\
G9 & \texttt{gemma-4-12B-it} & Text & 1.98 & 0.72 & 0.007 & 0.021 & 1{,}220 & yes \\
\midrule
G9 & \texttt{gemma-4-E4B-it} & Visual & 0.00 & 0.32 & 0.382 & 0.382 & 1{,}817 & no \\
G9 & \texttt{gemma-4-E4B-it} & Text & 1.18 & 0.78 & 0.0005 & 0.003 & 1{,}995 & yes \\
\midrule
QW & \texttt{Qwen3-VL-32B-Thinking} & Visual & 0.58 & 1.02 & 0.019 & 0.038 & 1{,}584 & yes$^\dagger$ \\
QW & \texttt{Qwen3-VL-32B-Thinking} & Text & 1.45 & 1.56 & $<10^{-4}$ & $<10^{-4}$ & 1{,}140 & yes \\
\midrule
QW & \texttt{Qwen3-VL-8B-Thinking} & Visual & $-$1.43 & $-$1.52 & 0.993 & 0.993 & 1{,}973 & no \\
QW & \texttt{Qwen3-VL-8B-Thinking} & Text & 1.78 & 3.06 & $<10^{-4}$ & $<10^{-4}$ & 1{,}624 & yes \\
\bottomrule
\end{tabular}

%% file: gen_v8p/gen_v8/tab_r9_verdicts.tex
\begin{tabular}{@{}lp{0.36\linewidth}ll@{}}
\toprule
Rule & Registered criterion & \texttt{3.8-flash} & \texttt{3.1-pro} \\
\midrule
H1a & Code $\ge$75\%, $\ge$+40\,pp over Direct ($p<10^{-4}$), $\ge$+30\,pp in every style & fail: 83.3\%, +23.3\,pp, min style +16.7 & fail: 70.0\%, +43.3\,pp, min style +0.0 \\
H1b & Direct falls over 4 magnitude bins; slope $<0$, $p<0.01$ & pass: slope $-$0.36, $p=3.0\times 10^{-4}$ & pass: slope $-$0.60, $p=5.1\times 10^{-6}$ \\
H2a & matched twins: $|\sST-\sAM|\le$15\,pp ($p>0.05$), log-error gap $\le$0.35, Code $\ge$75\% on both & fail: 20.0\,pp, Code 87.5/82.5\% & fail: 15.0\,pp, Code 75.0/67.5\% \\
H2b & $\sPM$ exceeds $\sAM$ by $\ge$35\,pp ($p<10^{-4}$) & fail: +32.5\,pp, $p=0.0012$ & fail: +42.5\,pp, $p=1.1\times 10^{-4}$ \\
H3a & $\mathrm{FNR}_H>\mathrm{FNR}_V$, Wilcoxon $p<10^{-4}$, in all 4 styles & pass: $p=7.6\times 10^{-7}$, 4/4 styles & fail: $p=0.0083$, 3/4 styles \\
H3b-i & two-view vs.\ one view: $\ge$+20\,pp graphs and tasks, $p<0.005$ & pass: +45.8\,pp, $p=5.7\times 10^{-14}$ & pass: +33.3\,pp, $p=2.3\times 10^{-10}$ \\
H3b-ii & two-view vs.\ quadrant zoom: $\ge$+15\,pp, $p<0.01$ & pass: +35.4\,pp, $p=4.1\times 10^{-6}$ & pass: +54.2\,pp, $p=4.5\times 10^{-13}$ \\
H3b-iii & $90^\circ$ vs.\ $180^\circ$ second view: $\ge$+15\,pp, $p<0.01$ & fail: $-$4.2\,pp, $p=0.83$ & fail: $-$8.3\,pp, $p=0.97$ \\
\bottomrule
\end{tabular}

%% file: gen_v8/tab_r9.tex
\begin{tabular}{llrrrrrr}
\toprule
 & & \multicolumn{3}{c}{\texttt{gemini-3.8-flash} (low)} & \multicolumn{3}{c}{\texttt{gemini-3.1-pro-preview} (low)} \\
\cmidrule(lr){3-5}\cmidrule(lr){6-8}
Slice / Condition & Metric / Subset & Graph / Dir. & Task / Code & Edge $F_1$ / $\Delta$ & Graph / Dir. & Task / Code & Edge $F_1$ / $\Delta$ \\
\midrule
\multicolumn{8}{l}{\emph{R9-A: Large realistic diagrams ($7{\times}7$ and $8{\times}8$, $N=48$ images, $96$ task pairs: $\sST$ and $\sAM$)}} \\
Direct (CoT) & prose JSON + count & 12/48 (25.0\%) & 0/96 (0.0\%) & 0.987 & 6/48 (12.5\%) & 0/96 (0.0\%) & 0.985 \\
One-shot Code & single-turn Python & 2/48 (4.2\%) & 14/96 (14.6\%) & 0.946 & 9/48 (18.8\%) & 7/96 (7.3\%) & 0.962 \\
TwoStage-1v & 1 upright view + solver & 13/48 (27.1\%) & 26/96 (27.1\%) & 0.977 & 18/48 (37.5\%) & 36/96 (37.5\%) & 0.988 \\
TwoStage-Zoom & $2{\times}2$ quadrant crops + solver & 18/48 (37.5\%) & 36/96 (37.5\%) & 0.989 & 8/48 (16.7\%) & 16/96 (16.7\%) & 0.990 \\
TwoStage-rot180 & $0^\circ{+}180^\circ$ + pair crops + solver & 37/48 (77.1\%) & 74/96 (77.1\%) & 0.998 & 38/48 (79.2\%) & 76/96 (79.2\%) & 0.998 \\
TwoStage-2v & $0^\circ{+}90^\circ$ + pair crops + solver & 35/48 (72.9\%) & 70/96 (72.9\%) & 0.998 & 34/48 (70.8\%) & 68/96 (70.8\%) & 0.997 \\
\midrule
\multicolumn{8}{l}{\emph{R9-A: Single-view horizontal ($\mathrm{FNR}_H$) vs.\ vertical ($\mathrm{FNR}_V$) edge false-negative rate ($N=48$ images)}} \\
All 4 domains & $N=48$ grids & $\mathrm{FNR}_H=7.3\%$ & $\mathrm{FNR}_V=1.6\%$ & $4.7\times$ ($p=7.6\times 10^{-7}$) & $\mathrm{FNR}_H=1.8\%$ & $\mathrm{FNR}_V=0.6\%$ & $3.0\times$ ($p=0.0083$) \\
\quad PCB copper trace & $n=12$ grids & 10.9\% & 2.4\% & $4.5\times$ & 4.9\% & 1.3\% & $3.7\times$ \\
\quad Blueprint floorplan & $n=12$ grids & 9.5\% & 1.9\% & $5.0\times$ & 1.1\% & 0.2\% & $6.1\times$ \\
\quad Transit schematic & $n=12$ grids & 6.5\% & 1.9\% & $3.5\times$ & 0.5\% & 1.0\% & $0.5\times$ \\
\quad Camera desk shadow & $n=12$ grids & 2.4\% & 0.0\% & all misses H & 0.9\% & 0.0\% & all misses H \\
\midrule
\multicolumn{8}{l}{\emph{R9-B: Magnitude-matched twins on realistic diagrams ($4{\times}4$ to $5{\times}6$, $N=40$ images, $120$ evaluations)}} \\
All R9-B tasks & $N=120$ pairs & 72/120 (60.0\%) & 100/120 (83.3\%) & +23.3\text{ pp} & 32/120 (26.7\%) & 84/120 (70.0\%) & +43.3\text{ pp} \\
\quad PCB copper trace & $n=30$ pairs & 16/30 (53.3\%) & 21/30 (70.0\%) & +16.7\text{ pp} & 7/30 (23.3\%) & 22/30 (73.3\%) & +50.0\text{ pp} \\
\quad Blueprint floorplan & $n=30$ pairs & 15/30 (50.0\%) & 23/30 (76.7\%) & +26.7\text{ pp} & 10/30 (33.3\%) & 28/30 (93.3\%) & +60.0\text{ pp} \\
\quad Transit schematic & $n=30$ pairs & 18/30 (60.0\%) & 28/30 (93.3\%) & +33.3\text{ pp} & 7/30 (23.3\%) & 7/30 (23.3\%) & +0.0\text{ pp} \\
\quad Camera desk shadow & $n=30$ pairs & 23/30 (76.7\%) & 28/30 (93.3\%) & +16.7\text{ pp} & 8/30 (26.7\%) & 27/30 (90.0\%) & +63.3\text{ pp} \\
Twin $\sST$ ($\GapL$) & $n=40$ grids & 25/40 (62.5\%) & 35/40 (87.5\%) & +25.0\text{ pp} & 9/40 (22.5\%) & 30/40 (75.0\%) & +52.5\text{ pp} \\
Twin $\sAM$ ($\shP$-c) & $n=40$ grids & 17/40 (42.5\%) & 33/40 (82.5\%) & +40.0\text{ pp} & 3/40 (7.5\%) & 27/40 (67.5\%) & +60.0\text{ pp} \\
Control $\sPM$ ($\GapL$, small) & $n=40$ grids & 30/40 (75.0\%) & 32/40 (80.0\%) & +5.0\text{ pp} & 20/40 (50.0\%) & 27/40 (67.5\%) & +17.5\text{ pp} \\
\bottomrule
\end{tabular}

%% file: gen_v8/tab_code_all_v84.tex
\begin{tabular}{llrrrrrrrr}
\toprule
Model & Condition & $n$ & acc (\%) & exec ok (\%) & timeout (\%) & libs (\%) & algebraic (\%) & DP (\%) & tokens \\
\midrule
Gemini 3.6 Flash (low) & Visual$\to$Code & 202 & 94 & 97 & 3 & 13 & 54 & 22 & 1.6k \\
 & \car{} & 202 & 95 & 99 & 1 & 7 & 74 & 34 & 1.9k \\
Gemini 3.8 Flash (low) & Visual$\to$Code & 392 & 93 & 100 & 0 & 6 & 62 & 27 & 1.7k \\
 & Text$\to$Code & 392 & 96 & 98 & 0 & 3 & 64 & 32 & 0.9k \\
 & Hint$\to$Code & 202 & 89 & 99 & 0 & 0 & 89 & 36 & 1.8k \\
 & Perceive+Code & 392 & 93 & 98 & 0 & 1 & 66 & 23 & 1.5k \\
 & \car{} & 392 & 91 & 100 & 0 & 0 & 72 & 41 & 1.5k \\
 & Text CAR & 392 & 98 & 100 & 0 & 0 & 67 & 50 & 1.2k \\
Gemma 4 31B (no thinking) & Visual$\to$Code & 392 & 67 & 100 & 0 & 10 & 67 & 33 & 0.9k \\
 & Hint$\to$Code & 202 & 63 & 100 & 0 & 0 & 99 & 41 & 1.0k \\
 & Perceive+Code & 202 & 64 & 100 & 0 & 13 & 67 & 33 & 1.1k \\
 & \car{} & 392 & 62 & 100 & 0 & 0 & 74 & 42 & 1.2k \\
Gemma 4 31B (thinking) & Visual$\to$Code & 19 & 95 & 95 & 0 & 0 & 75 & 18 & 8.0k \\
 & \car{} & 25 & 92 & 100 & 0 & 0 & 83 & 54 & 7.3k \\
Gemini 3.1 Pro (low) & Visual$\to$Code & 202 & 86 & 100 & 0 & 0 & 62 & 27 & 2.5k \\
 & Perceive+Code & 202 & 91 & 100 & 0 & 0 & 58 & 26 & 2.2k \\
 & \car{} & 202 & 88 & 100 & 0 & 0 & 74 & 51 & 2.4k \\
\bottomrule
\end{tabular}